\documentclass[10pt,journal,compsoc]{IEEEtran}

\usepackage [sort,compress]{cite}
\usepackage{graphicx}
\usepackage{grffile}
\usepackage{color}
\usepackage[table,usenames,dvipsnames]{xcolor}
\usepackage{amsmath}
\usepackage{array}
\usepackage[caption=false,font=footnotesize]{subfig}
\usepackage{fixltx2e}
\usepackage{url}
\usepackage{amssymb}
\usepackage{lineno}
\usepackage{adjustbox}
\usepackage{multirow}
\usepackage{booktabs}
\usepackage{dblfloatfix}
\usepackage{balance}
\usepackage{rotating}
\usepackage{tabularx}
\usepackage{xparse}

\NewDocumentCommand{\rot}{O{45} O{1em} m}{\makebox[#2][l]{\rotatebox{#1}{#3}}}

\definecolor{lightgray}{gray}{0.875}

\definecolor{markcolor}{rgb}{0.0, 0.0, 0.8}

\begin{document}

\title{SIFT Matching by Context Exposed}

\author{Fabio~Bellavia
\thanks{F. Bellavia is with the Department of Mathematics and Computer Science, Universit\`{a} degli Studi di Palermo, Via Archirafi, 34, 90123 Palermo, Italy, e-mail: fabio.bellavia@unipa.it .}}

\markboth{Draft}%
{F. Bellavia: SIFT Matching by Context Exposed}

\IEEEtitleabstractindextext{
\begin{abstract}
This paper investigates how to step up local image descriptor matching by exploiting matching context information. Two main contexts are identified, originated respectively from the descriptor space and from the keypoint space. The former is generally used to design the actual matching strategy while the latter to filter matches according to the local spatial consistency. On this basis, a new matching strategy and a novel local spatial filter, named respectively blob matching and Delaunay Triangulation Matching (DTM) are devised. Blob matching provides a general matching framework by merging together several strategies, including rank-based pre-filtering as well as many-to-many and symmetric matching, enabling to achieve a global improvement upon each individual strategy. DTM alternates between Delaunay triangulation contractions and expansions to figure out and adjust keypoint neighborhood consistency. Experimental evaluation shows that DTM is comparable or better than the state-of-the-art in terms of matching accuracy and robustness. Evaluation is carried out according to a new benchmark devised for analyzing the matching pipeline in terms of correct correspondences on both planar and non-planar scenes, including several state-of-the-art methods as well as the common SIFT matching approach for reference. This evaluation can be of assistance for future research in this field.
\end{abstract}

}

\maketitle

\IEEEdisplaynontitleabstractindextext

\IEEEpeerreviewmaketitle

\IEEEraisesectionheading{\section{Introduction}}
\IEEEPARstart{K}{eypoint} correspondences play a crucial role in many computer vision algorithms dealing with spatial localization. These include~\cite{szeliski_book}: Structure from Motion (SfM), image stitching, large-scale image retrieval and Simultaneous Localization And Mapping (SLAM), whose practical applications are more and more affecting everyday life in providing assistance or for mere entertainment. The emerging autonomous driving systems, together with the various applications of the augmented reality from medicine to gaming, represent some relevant examples in that sense.

This state of things has granted an active interest on this research topic over the decades, continuously evolving side by side with the novel advancements and challenges arising in the field. In this scenario, despite its age, the Scale Invariant Feature Transform (SIFT)~\cite{sift}, both as keypoint detector and as local image descriptor, is in good health. As a matter of fact, SIFT is still popular~\cite{sift_matching,imw2020} and revisited~\cite{root_sift,sgloh2_pami,sift_matching}. Moreover, SIFT has obtained satisfactory results in recent benchmarks~\cite{wisw,sift_matching,imw2020} and many applications still rely on it~\cite{colmap}. At the current state of the research, the keypoint extraction process and the computation of the associated local image descriptors, that must be synergically used to establishing correspondences, seem to have reached somewhat their limits when the matching process is considered untied from the context provided by the source images~\cite{contextdesc}. This observation is reflected in the progress done by deep learning descriptors~\cite{tfeat,hardnet,geodesc,contextdesc,sosnet} in conjunction with the availability of ever more big datasets~\cite{sun3d,hpatches,megadepth,imw2020}. In this paper two different matching contexts are discussed.

The first one is provided by the \emph{descriptor space}. Mutual Nearest Neighbor (NN) and the Nearest Neighbor Ratio (NNR)~\cite{sift}, two of the most commonly employed matching strategies are examples of this context exploitation. NN requires a match to be the best on both the input pair images, i.e. NN considers a \emph{inter-relation} in the descriptor space. NNR selects matches according to the ratio between the first and second best distances inside the reference image, i.e. NNR considers a \emph{intra-relation} in the descriptor space.

The second matching context is the one provided by the \emph{keypoint space} inside the images. This kind of scene knowledge includes patch relative orientations~\cite{sgloh2_pami} and keypoint spatial relations~\cite{spatial_filter_eval,matchers_survey}, which have been exploited to successfully disambiguate matches, boosting the final matching accuracy. Model constraints such those imposed by planar scenes and epipolar geometry~\cite{multiview}, mainly built upon the Random SAmple COnsensus (RANSAC)~\cite{ransac,prosac,scramsac,groupsac,gc_ransac,adalam}, also operate on the keypoint space and generally represent the final post-filtering step of the entire process. Although very effective, this last kind of model-based matching is somewhat less general since it is not valid in all situations, such as in the case of moving or deformable objects that violate the constraint of a single rigid structure~\cite{lpm}.

The recent evolution of deep networks jointly training a keypoint detector and descriptor~\cite{superpoint,disk} has brought fresh attention to the matching step, inherently correlated to scene context, as the next step to be included in an all-in-one deep matching network~\cite{superglue,contextdesc,d2d} or as a replacement for RANSAC~\cite{point_net,nm_net,oanet,acne}. In this paper existent and novel general matching strategies exploiting image context in terms of both descriptor and keypoint spaces are presented and discussed. As baseline, their applications to SIFT are considered, but results on other keypoint detectors and descriptors are also shown. The contributions of this study can be mainly divided into three parts:
\begin{itemize}
	\item Descriptor space context: Starting from a known greedy matching strategy using one-to-one NN together with NNR, several potential enhancements that include rank-based filtering as well as many-to-many and symmetric matching are investigated, extensively combined, and evaluated. The overall matching strategy obtained by merging altogether the best approaches, named blob matching, is proved to give more correct correspondences with respect to each individual matching strategy it is based upon.
	\item Keypoint space context: A new robust and effective local spatial filter named Delaunay Triangulation Matching (DTM) is designed. DTM associates matches greedly, according to the consistency of the keypoint spatial local neighborhoods. Neighborhood relations are obtained on the basis of Delaunay triangulations on the source images. DTM can guarantee comparable or better matching results with respect to the state-of-the-art.
	\item Benchmarking image matching by context: A new evaluation aimed at comparing state-of-the-art spatial matching strategies is devised, evaluating their behaviors on different keypoint detectors and recent descriptors on both planar and non-planar scenarios. In the non-planar case, when no 3D data are available, ground-truth correct matches are checked according to~\cite{sift_matching}. With respect to other evaluations merely relying on the epipolar distance~\cite{fund_mat_eval}, this setup avoid incorrect associations due to epipolar ambiguity and, differently from benchmarks based on camera pose estimation error~\cite{imw2020}, provides direct evaluation scores of the matches.
\end{itemize}

The rest of the paper is organized as follows: Related work is presented in Sec.~\ref{related_work}, blob matching is defined in Sec.~\ref{ds}, DTM in Sec.~\ref{ks}, and proposed benchmark setup and results are discussed in Sec.~\ref{eval}. Finally, conclusions and future work are outlined in Sec.~\ref{conclusion}.

\section{Related work}\label{related_work}
Assigning correspondences requires the close cooperation between three main subjects: The keypoint detector, the local image descriptor and the matching strategy.

A good keypoint detector for this task must be able to extract distinctive yet repeatable characteristic points on the input images. Moreover, the number of detected keypoints should be sufficient to provide a good coverage of relevant structures of the scene but, at the same times, the number of keypoints should be limited so as to make the computational process feasible and to reduce the chance of false matches. Keypoint distinctiveness decreases as the number of detected keypoints increases, especially in the presence of repeated structures on the scene. 

Classical keypoint detectors usually extract corners or blob-like structures, the reader can refer to~\cite{szeliski_book} for a general overview. Recently, keypoint detectors based on deep learning have started to emerge, especially in frameworks where detectors and descriptors sharing a common network are jointly estimated~\cite{superpoint,disk,d2d}, or by combining handcrafted and deep learned filters~\cite{keynet}. Experimental evaluation in Sec.~\ref{eval} will be carried out considering the state-of-the art SIFT and the HarrisZ detector~\cite{harrisz}. SIFT is a well known blob-like DoG multi-scale detector, while HarrisZ is an affine multi-scale corner detector. Relying on Harris corners computed on scale enhanced gradient derivatives, HarrisZ uses an adaptive filter response and a rough edge mask to select keypoints. In recent image matching challenges~\cite{imw2021}, a matching pipeline built on HarrisZ corners provided better results than other keypoint detectors based on DoG (i.e. SIFT) keypoints.

Local image descriptors are used to obtain a meaningful numerical vector encoding the distinctive attributes of a keypoint patch, i.e. the local keypoint neighborhood. Ideally, descriptors for the same keypoint undergoing both geometrical or color distortions must be close in the descriptor vector space, and the opposite must hold for distinct keypoints. A trade-off between the descriptor tolerance to image deformation and its discriminability is often required, since high descriptor invariance decreases descriptor discriminability.

Local image descriptors can be divided into handcrafted and data-driven, the reader can refer to~\cite{szeliski_book} for a general overview. The popular handcrafted SIFT descriptor is based on the gradient orientation histogram. Among data-driven descriptors, deep descriptors nowadays outperform any others~\cite{wisw,sift_matching,imw2020} thanks to the modern GPU hardware capabilities and the availability of large training datasets~\cite{sun3d,hpatches,megadepth,imw2020}. Triplet loss~\cite{tfeat}, hard negative mining~\cite{hardnet}, second-order similarity~\cite{sosnet}, geometric constraint integration~\cite{geodesc} and jointly detector-descriptor optimization~\cite{superpoint,contextdesc,disk,d2d} are some of the techniques employed to get state-of-the-art results. 

The matching strategy evaluation carried out in Sec.~\ref{eval} will employ the handcrafted descriptors RootSIFT~\cite{root_sift} and the double square-rooting shifting Gradient Local Orientation Histogram (RootsGLOH2)~\cite{rootsgloh2}. Furthermore, the deep Second Order Similarity Network (SOSNet)~\cite{sosnet} and HardNet2~\cite{hardnetamos} will be used. RootSIFT improves upon SIFT by replacing the Euclidean distance with the Hellinger's distance and, in addition to RootSIFT, RootsGLOH2 is able to better handle patch orientation estimation. Deep-based SOSNet and HardNet2 are the current state-of-the-art, while RootsGLOH2 has been shown to be among the current best handcrafted descriptor~\cite{rootsgloh2}.

Patch normalization is the interface between keypoint extraction and local descriptor computation, and it is often addressed together with this last step. The most common patch normalization approach is the one employed by SIFT, yet other approaches exists~\cite{orb,learning_ori,affnet}. In particular, it has been reported in~\cite{mrogh,sift_matching} that the orientation estimation is one the most critical aspect that needs to be handled. The deep patch orientation assignment designed in~\cite{learning_ori} was proved to improve the matching accuracy noticeably~\cite{sift_matching} and will be employed for the evaluation presented in Sec.~\ref{eval}.

Matches are assigned by the pairwise inspection of the similarity in the descriptor space and optionally by considering the keypoint displacement in the images. The most common pipeline uses mutual NN or NNR matching followed by RANSAC. Using a symmetric variant of NNR~\cite{sift_matching} or considering the first geometrical inconsistent match in NNR~\cite{fginn} have been shown to generally improve the matching process. Further improvements have been observed when considering many-to-many putative matches instead of constraining matches to be one-to-one~\cite{generalized_ransac}. Many-to-many matches can be also related to the employment of multiple synthesized views to enrich the candidate matches~\cite{mods}.

Correspondence filtering on the basis of spatial constraints is a wide research topic, the reader can refer to~\cite{spatial_filter_eval,matchers_survey} for a more comprehensive presentation. The scene model can provide effective constraints, as in the case of planar and epipolar geometries. Although quite powerful, robust model regression strategies relying on RANSAC~\cite{ransac,prosac,scramsac,groupsac,gc_ransac}, can be limiting when violating the assumption of rigid scenes such in the case of moving and deformable objects~\cite{lpm}, unless handling multiple models~\cite{progx}. More non-global and relaxed spatial constraints have been designed to overpass this limitation, also able to boost RANSAC in terms of both efficiency and accuracy when employed to pre-filter candidate matches. When feasible, executing RANSAC after any other kind of match selection is always the best practice.

An early example of spatial filtering is the topological filter~\cite{topological_filter}, which checks the relative positions of the matched keypoints across the images. Better filtering approaches consider the local consistency around the keypoints of the pair defining the match. Local neighborhood can be defined by using a fixed circular radius and measuring consistency in terms of the number of other matches having both keypoints on the respective images falling into the neighborhoods associated to the considered match. In addition, the relative local image transformation between corresponding keypoints inducted by patch normalization can provide a further consistency check to be exploited to refine local neighborhoods. This can be done by comparing the transformation parameters or by considering the reprojection errors according to the transformations. Circular neighborhoods refined according to patch-based consistency on local similarity transformations are checked in~\cite{scramsac} to pre-filter matches before RANSAC. Grids can also be employed to define neighborhoods efficiently, as for the Grid-based Motion Statistics (GMS)~\cite{gms}. By defining neighborhoods as $3\times3$ grid blocks while measuring the consistency block-wise, GMS is able to take into account the relative positions of the matches inside the neighborhoods. Moreover, since the best neighborhood size is generally not known a priori, GMS makes use of different grid sizes to get a multi-scale approach. Another possible choice, employed in the Locality Preserving Matching (LPM)~\cite{lpm} is to define circular neighborhoods by considering only the closest $\mathrm{k}$ matches in terms of keypoint proximity. Unlikely GMS, LPM limits the neighbors to those matches having motion flow vectors similar to that of considered match, and considers multiple values of $\mathrm{k}$ to be more robust. In the case of the Guided LPM (GLPM)~\cite{glpm}, NNR match pre-filtering precedes LPM. LPM neighborhood definition is also employed by the Learning for Mismatch Removal (LMR)~\cite{lmr} to extrapolate local correspondence relations to learn how to classify correspondences. Neighborhoods can be also defined in terms of the edges of the Delaunay triangulation, as for the Progressive Feature Matching (PFM)~\cite{pfm}, employing affine patch-based neighborhood consistency to cast the problem as a Markov random field function optimization. The Progressive Graph Matching (PGM)~\cite{pgm} uses instead another graph matching formulation with neighborhoods defined by the $\mathrm{k}$ closest matches and affine patch-based consistency, while in~\cite{gc_ransac} graph-based optimization of fixed circular neighborhoods is used to improve the best selected RANSAC model. 

Spatial filtering can be also formulated as the estimation and outlier rejection of the motion vector field inducted by the scene on the images. In that sense, both PFM and PGM estimate local affine motion fields from sparse correspondences. According to this idea, grid neighborhood is also employed in~\cite{viso2} to collect motion hypotheses and discard the outlier matches violating them. More complex approaches such as the Locally Linear transforming (LTT)~\cite{llt}, the Vector Field Consensus (VFC)~\cite{vfc} and the Bilateral Model (BM)~\cite{bm} simultaneously estimate a smooth motion field from the putative matches while removing outliers. To a certain extend, also the Sequential Correspondence Verification (SCV)~\cite{scv}, which considers the region growing progression around the local affine patch neighborhoods to make decisions about the correctness of a match, can be framed in this kind of design.

Other solutions relying on the spatial relations between correspondences are designed as the clustering of the motion vector field. In~\cite{groupsac} images are segmented according to their spatial and motion information prioritizing RANSAC model sampling from large, more consistent groups. In a similar spirit, the recent Adaptive Locally-Affine Matching (AdaLAM)~\cite{adalam} executes multiple local affine RANSACs inside the circular neighborhoods of seed matches, i.e. those matches with high confidence according to their descriptor similarity. The Robust Feature Matching Spatial Clustering of Applications with Noise (RFM-SCAN)~\cite{rfm_scan} uses instead clustering to directly identify outlier correspondences not belonging to any cluster. A game theoretic formulation is instead proposed in~\cite{game_theoretic_journal}, which iteratively outputs the best clusters only on the basis of the patch-based local transformation consistency in terms of reprojection error, disregarding keypoint proximity\footnote{However, given two matches, it usually holds that the closer are the respective keypoints on the same image, the more consistent is their patch-based consistency according to the reprojection error.}.

More recently, deep networks have reached the state-of-the art in the case of planar or epipolar geometry scene constraint matching~\cite{point_net,nm_net,oanet,acne}. In~\cite{point_net} the context normalization is introduced to exploit contextual information while preserving permutation equivariance. The network architecture of~\cite{nm_net} defines instead grouping operations to supply each correspondence with data from its nearest correspondences only in terms of patch-based local transformation consistency as in~\cite{game_theoretic_journal}. In addition to context normalization, the Order-Aware Network (OANet)~\cite{oanet} adds network layers to learn how to cluster unordered sets of correspondences so as to incorporate the data context and the spatial correlation, while the Attentive Context Network (ACNe)~\cite{acne} employs local and global attention to exclude outliers from context normalization. Attentional graph neural networks is instead proposed in~\cite{superglue} to infer global and local spatial correlations. More recently, the network designed in~\cite{loftr} was able to get keypoint-free correspondences for a coarse-to-fine matching strategy, with initial rough dense correspondences provided by exploiting self and cross attention inferred by transformers.  

\section{Blob matching}\label{ds}
Blob matching mainly works by filtering matches through descriptor space context heuristics. Before defining the blob matching, the base matching strategies it relies upon are reviewed and discussed. The notation adopted hereafter assumes that $\mathrm{D}\in\mathbb{R}^{n\times m}$ is a matrix such that the entry $\mathrm{D}_{ij}$ is the distance between the descriptors associated to the $i$-th keypoint on the first image $I_1$ and the $j$-th keypoint on the second image $I_2$. $\mathrm{D}_{ij^w}$ denotes the $w$-th lowest value found on row $i$, i.e. the $w$-th best descriptor association. Likewise, $\mathrm{D}_{i^wj}$ is the $w$-th lowest value on column $j$, and the subscript $_{\downdownarrows}$ denotes the extraction of the index pair $(i,j)$ from $\mathrm{D}_{ij}$, done element-wise in the case of a set of matches. NN is the basic way to associate correspondences
\begin{equation}
	S_{NN}=\{\mathrm{D}_{ij^1}\}_\downdownarrows
\end{equation}
Here, the index spans $1\leq i \leq n$ and $1\leq j \leq m$ are omitted for convenience. Mutual NN constrains even more $S_{NN}$ by requiring that selected matches must be the best in both images, i.e. simultaneously on both the rows and columns of $\mathrm{D}$
\begin{equation}\label{nnr}
	S_{mNN}=\{\mathrm{D}_{ij^1}=\mathrm{D}_{i^1j}\}_\downdownarrows
\end{equation}
It is easy to see that $S_{mNN}\subseteq S_{NN}$. NNR~\cite{sift} considers instead a matrix $\mathrm{D}'$ such that
\begin{equation}
	\mathrm{D}'_{ij}=\frac{\mathrm{D}_{ij}}{\mathrm{D}_{ij^2}}
\end{equation}
and the threshold $t_r$, usually set to 0.8, so that
\begin{equation}
	S_{NNR}=\{\mathrm{D}'_{ij^1}\leq t_r\}_\downdownarrows
\end{equation}
NNR can be related to the triplet matching learning adopted by deep descriptors starting from~\cite{tfeat}. Notice that $\mathrm{D}'_{ij}\in[0,1]$ by definition, and $\mathrm{D}'_{ij^1\downdownarrows}=\mathrm{D}_{ij^1\downdownarrows}$ for any row since scalar multiplication does not affect the ordering. NNR is usually more accurate than mutual NN since NNR relative values can better express the image context than NN absolute values. Nevertheless, $S_{mNN}$ is symmetric since it takes both the images as reference, and provides a one-to-one matching relation. By contrast, $S_{NNR}$ considers only the first image $I_1$ as reference, since the denominator in Eqs.~\ref{nnr} is computed on rows, providing a one-to-many matching relation. In order to relax the strict requirements of mutual NN, the following greedy strategy, to the best of the author's knowledge first mentioned in~\cite{descriptor_eval}, can be employed. As notation, $\mathrm{D}_k$ is the $k$-th best entry of the matrix $\mathrm{D}$ considered as linear vector and $G$ is the set of matches, initially empty. At each iteration $1\leq k\leq n\times m$, $\mathrm{D}_k$ is included into $G$ if both $\mathrm{D_{k\downarrow\vert}}\notin G_{\downarrow\vert}$ and $\mathrm{D_{k\vert\downarrow}}\notin G_{\vert\downarrow}$, where the subscripts $_{\downarrow\vert}$ and $_{\vert\downarrow}$ denote the operators that extract the row and column index of the entry, respectively. The final matching set is
\begin{equation}\label{gnn}
	S_{gNN}=G_\downdownarrows
\end{equation}
\begin{figure*}[h!]
	\center
	\hfil
	\subfloat[]{\label{blob_pairs}
		\resizebox{0.21\textwidth}{!}{
			\begin{tabular}[b]{|c|c|c|c|c|}
				\multicolumn{5}{l}{$\mathrm{D}$} \\
				\hline
				1.6 & 2.5 & 1 & 4 & 2.3 \\
				\hline
				4.2 & 0.5 & 1.7 & 3 & 1.1 \\
				\hline
				5.1 & 3.5 & 3.1 & 1.2 & 2 \\
				\hline
				2.8 & 0.6 & 2.1 & 4.1 & 5 \\
				\hline
				4.4 & 3.4 & 2.4 & 4.3 & 4.5 \\
				\hline
				3.2 & 5.5 & 5.8 & 6.1 & 3.6 \\
				\hline
				1.3 & 6 & 3.7 & 2.7 & 1.4 \\
				\hline
				\multicolumn{5}{l}{\hphantom{0}} \\				
			\end{tabular}
		}
		\hspace{2em}
		\resizebox{0.46\textwidth}{!}{
			\begin{tabular}[b]{r|l}
				\multicolumn{1}{r}{setup} & \multicolumn{1}{l}{$G_\downdownarrows$} \\
				\toprule
				NN by row & $S_{NN}=\{(1,3),(2,2),(3,4),(4,2),(5,3),(6,1),(7,1)\}$ \\
				NN by column & $S_{NN}=\{(1,3),(2,2),(2,5),(3,4),(7,1)\}$ \\
				$f=1_\cap$, $f'>0$ & $S_{mNN}=\{(2,2),(1,3),(3,4),(7,1)\}$ \\
				$f=1_\cup$, $f'=1$ & $S_{mNN}=\{(2,2),(1,3),(3,4),(7,1)\}$ \\
				$f\ge\Omega$, $f'=1$ & $S_{gNN}=\{(2,2),(1,3),(3,4),(7,1),(6,5)\}$ \\
				$f>0_\cup$, $f'=1$ & $S_{gNN}=\{(2,2),(1,3),(3,4),(7,1),(6,5)\}$ \\				
			    $f=3_\cap$, $f'=1$ & $\{(2,2),(1,3),(3,4),(7,1)\}$ \\
				$f=1_\cup$, $f'=2$ & $\{(2,2),(4,2),(1,3),(2,5),(3,4),(7,1),(5,3),(6,1)\}$ \\
				$f=3_\cap$, $f'=2$ & $\{(2,2),(4,2),(1,3),(2,5),(3,4),(7,1),(7,5),(1,1),(4,3)\}$ \\
				$f\ge\Omega$, $f'=2$ & $\{(2,2),(4,2),(1,3),(2,5),(3,4),(7,1),(7,5),(1,1),(4,3),(5,4)\}$ \\
				$f=3_\cup$, $f'=2$ & $\{(2,2),(4,2),(1,3),(2,5),(3,4),(7,1),(7,5),(1,1),(4,3),(5,4)\}$ \\
				\midrule[\heavyrulewidth]
				\multicolumn{2}{l}{$\Omega=\max(n,m)$ with $\mathrm{D}\in\mathbb{R}^{n\times m}$, in this case $F_\cap=F_\cup$}
			\end{tabular}
		}
	}
	\hspace{3em}		
	\subfloat[]{\label{blob_fginn}
		\begin{tabular}[b]{@{}c@{}}
			\includegraphics[height=0.18\textwidth]{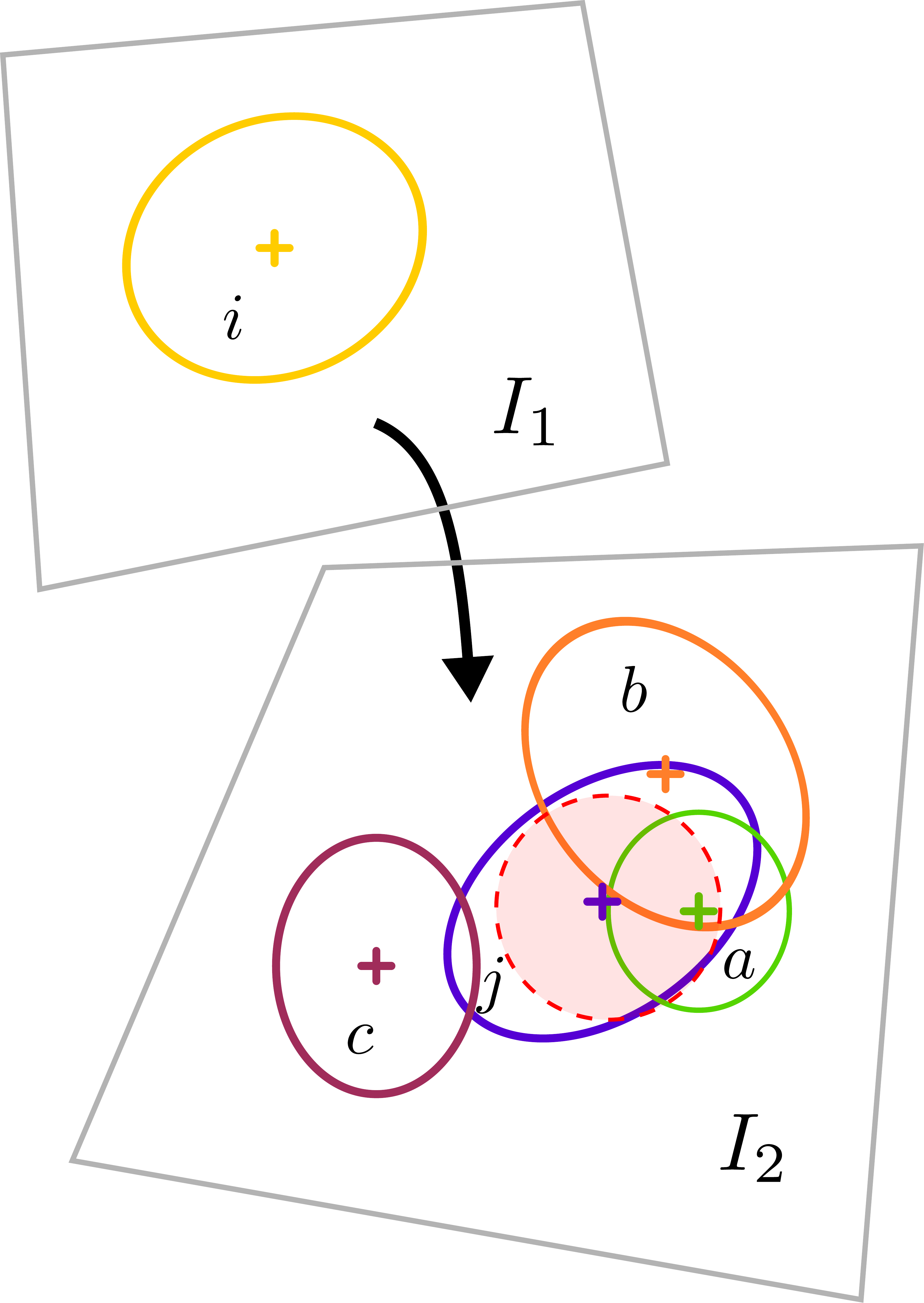} \\
		\end{tabular}
	}
	\hfil
	\caption{\label{blob_fig}
		\protect\subref{blob_pairs}~The candidate match sets $G_\downdownarrows$ for different blob matching configurations on a toy example input $\mathrm{D}$. Some setups may always lead to the same $G_\downdownarrows$ set, regardless of $\mathrm{D}$, as for the third and fourth setup rows. This is not the case of the seventh setup row, where the same $G_\downdownarrows$ output is due to the choice of $\mathrm{D}$.~\protect\subref{blob_fginn}~Visual representations of NNR and FGINN. If it holds for $\mathrm{D}$ that $\mathrm{D}_{ij}\leq \mathrm{D}_{ia}\leq \mathrm{D}_{ib}\leq \mathrm{D}_{ic}$ and the spatial configuration of the associated patch ellipses is the one reported in the figure, for the match $(i,j)$ then $\mathrm{D}_{ij}/\mathrm{D}_{ia}$, $\mathrm{D}_{ij}/\mathrm{D}_{ib}$ and $\mathrm{D}_{ij}/\mathrm{D}_{ic}$ are the values of NNR, FGINN using keypoint distance and FGINN using the overlap error, respectively (see the text for details, best viewed in color).}
\end{figure*}
Under the assumption that $\mathrm{D}$ entries have unique values, it holds that $S_{mNN}\subseteq S_{gNN}$ and $\vert S_{gNN}\vert=\min(n,m)$. If a mutual match $(i,j)$ of $S_{mNN}$ has not been included in $S_{gNN}$, there would have been a match $(\widetilde{i},\widetilde{j})$ with $i=\widetilde{i}$ or $j=\widetilde{j}$ in a previous iteration such that $\mathrm{D}_{\,\widetilde{i}\,\widetilde{j}}<\mathrm{D}_{ij}$. This is a contradiction since $\mathrm{D}_{ij}=\mathrm{D}_{i^1j}=\mathrm{D}_{ij^1}$ by definition. Figure~\ref{blob_pairs} shows the differences between $S_{NN}$, $S_{mNN}$ and $S_{gNN}$.

Mutual or greedy NN can be combined with NNR, in the sense that matches are extracted by NN but sorted and eventually filtered according to the NNR ranking. In this case, when the greedy NN is employed with NNR, one have to replace Eq.~\ref{nnr} with an alternative definition
\begin{equation}\label{nnr2}
	\mathrm{D}'_{ij}=\frac{\mathrm{D}_{ij}}{\mathrm{D}_{ij^2_{\geq}}}
\end{equation}
$\mathrm{D}_{ij^w_{\geq}}$ and $\mathrm{D}_{i^w_{\geq}j}$ are the lowest $w$-th values greater or equal to $\mathrm{D}_{ij}$ on the $j$-th column and on the $i$-th row, respectively. This further constraint is necessary since $\mathrm{D}_{ij}$ may not be equal to $\mathrm{D}_{ij^1}$ so that $\mathrm{D}'_{ij}>1$. In order to improve the matching process, a symmetric NNR is proposed in~\cite{sift_matching} as the harmonic mean between the two entries obtained by swapping the reference image, corresponding to operate on the matrix transpose $\mathrm{D}^\top$ of $\mathrm{D}$
\begin{equation}\label{snnr}
	\mathrm{D}''_{ij}=\frac{2\mathrm{D}'_{ij}(\mathrm{D}^\top)'_{ji}}{\mathrm{D}'_{ij}+(\mathrm{D}^\top)'_{ji}}
\end{equation}
In the particular case of Eq.~\ref{nnr2} the harmonic mean becomes
\begin{equation}
	\mathrm{D}''_{ij}=\frac{2\mathrm{D}_{ij}}{\mathrm{D}_{ij^2_{\geq}}+\mathrm{D}_{i^2_{\geq}j}}
\end{equation}
The First Geometrically Inconsistent NN (FGINN)~\cite{fginn} is another possible improvement to NNR
\begin{equation}\label{nnr4}
	\mathrm{D}'_{ij}=\frac{\mathrm{D}_{ij}}{\mathrm{D}_{ij^2_{\circledcirc}}}
\end{equation}
Here, the second lowest value in the denominator of Eq.~\ref{nnr} is intended among those keypoints far at least $t_o=10$ pixels from the keypoint $j$ in the corresponding image, denoted as $\mathrm{D}_{ij^2_{\circledcirc}}$. The choice of the second lowest value according to FGINN is shown in Fig.~\ref{blob_fginn}. Although the keypoint position implies to work on the keypoint space, this approach mainly deals with the descriptor space and it is discussed here. Finally, in~\cite{generalized_ransac}, many-to-may match relations have been shown to improve the recall of the matching process, leading to better samples for the RANSAC hypothesis generation.

Aimed at incorporating all the matching strategies discussed so far, the blob matching is now formulated according to the following steps.
\begin{enumerate}
\item The similarity matrix $\mathrm{D}$ is pre-filtered so that only matches appearing among the $f$ best matches for both or any of the input pair images will be considered~\cite{imw2020}, giving rise respectively to one of the sets
\begin{equation}\label{cap}
	F_\cap=\{\mathrm{D}_{ij}\leq\mathrm{D}_{ij^f}\}\cap\{\mathrm{D}_{ij}\leq\mathrm{D}_{i^fj}\}
\end{equation}
\begin{equation}\label{cup}
	F_\cup=\{\mathrm{D}_{ij}\leq\mathrm{D}_{ij^f}\}\cup\{\mathrm{D}_{ij}\leq\mathrm{D}_{i^fj}\}
\end{equation}
As additional notation, the subscript $\cap$ or $\cup$ will be appended to the value of $f$ to shortly refers to $F_\cap$ or $F_\cup$, respectively, while $F$ will indicate one of the two sets indistinctly.
\item Surviving matches are then filtered according to the greedy approach, modified to take into account the first $f'$ best matches instead of only the first one. In detail, after being sorted by increasing values, $\mathrm{D}_{ij}\in F$ is added to the multiset $G$ if both $(G\cup\mathrm{D_{ij}})_{\vert\downarrow}$ and $(G\cup\mathrm{D_{ij}})_{\downarrow\vert}$ do not contain elements counted more than $f'$ times (see again Fig.~\ref{blob_pairs}).
\item NNR-like similarity values for the matches in $G$ are obtained by taking into account a many-to-many scheme but also FGINN. Equation~\ref{nnr2} is modified into
\begin{equation}\label{nnr25}
	\mathrm{D}'_{ij}=\mathcal{D}
\end{equation}
where $\mathcal{D}$ can be further specified as
\begin{equation}\label{nnr3}
	\mathcal{D}_\geq=\frac{\mathrm{D}_{ij}}{\mathrm{D}_{ij^2_{\geq\circledcirc}}}
\end{equation}
The symbols $\geq$ and $\circledcirc$ are the same as for Eqs.~\ref{nnr2} and~\ref{nnr3}, respectively. The threshold $t_o$ can also express a relative threshold and not only an absolute pixel distance. In the first case, a filtering based on the overlap error between the elliptical patches is considered instead of the original FGINN criterion based on the keypoint center distance (see again Fig.~\ref{blob_fginn}). This allows to rely on non-absolute values. A further specialization for $\mathcal{D}$ is also considered
\begin{equation}\label{nnr5}
	\mathcal{D}^+_\ast=\frac{\mathrm{D}_{ij}}{\mathrm{D}_{ij}+\mathrm{D}_{ij^2_{\ast\circledcirc}}}
\end{equation}
The $\ast$ symbol is a placeholder for $\geq$, and can be possible empty, since $\geq$ is not strictly required to accommodates values into the range $[0,1]$. This happens because the whole column or row spans can be considered instead of limiting the selection only to the values greater than the current one as it was required for $\mathrm{D}_{ij^2_{\geq}}$ in Eq.~\ref{nnr2}.
\item Lastly, in order to provide the final similarity score $\overline{\mathrm{D}}_{ij}$ for matches in $G$ to be used for sorting or thresholding the keypoint pairs, a function $\mathcal{W}$ is applied so as to combine the two possible matching similarity values obtained when considering each image as reference. In detail, using the first image as reference corresponds to employ $\mathrm{D}'_{ij}$, while using the other image corresponds to $(\mathrm{D}^\top)'_{ji}$ so that
\begin{equation}
	\overline{\mathrm{D}}_{ij}=\mathcal{W}(\mathrm{D}'_{ij},(\mathrm{D}^\top)'_{ji})
\end{equation}
$\mathcal{W}(a,b)$ can simply be the projection on one of the two arguments ($\mathcal{W}(a,b)=a$ or $\mathcal{W}(a,b)=b$), the harmonic mean of Eq.~\ref{snnr} ($\mathcal{W}(a,b)=(2ab)/(a+b)$), and the minimum ($\mathcal{W}(a,b)=\min(a,b)$) or maximum ($\mathcal{W}(a,b)=\max(a,b)$) values of the two arguments, likewise respectively the intersection and union of the matching sets employed in~\cite{imw2020}.
\end{enumerate}
The first two steps of blob matching extract the set $G_\downdownarrows$ of the candidate matches, ranked by the remaining two steps. Clearly, it may happen that different initial configurations lead to the same $G_\downdownarrows$ set (see again Fig.~\ref{blob_pairs}). It comes out in Sec.~\ref{eval_blob} that the best blob matching configuration is $f=10_\cup$, $f'=5$, $\mathcal{D}^+$ with $t_o=10$ px or $t_o=75$\%, and $\mathcal{W}(a,b)=(2ab)/(a+b)$. Notice also that blob matching is quite general: By setting $f=1$ only mutual NN matches are considered, moving up $f'$ from 1 when $f>1$ the one-to-one match relation becomes a many-to-many relation, FGINN is turned off when $t_o=\infty$, and only the first image is used as reference when $\mathcal{W}(a,b)=a$.

\section{Delaunay triangulation matching}\label{ks}
As discussed in Sec.~\ref{related_work}, spatial filtering relies on the concept of the image local neighborhood, often assumed to be isotropic, circular or squared, for an efficient and fast computation. In the general case, the optimal circular radius needed to define the neighborhood is not known a priori and can vary among different image regions due to the non-homogeneous distribution of keypoints on the image. In order to alleviate this issue, the neighborhood estimation can proceed in a corse-to-fine manner using different radius, or it can consider the $\mathrm{k}$ closest keypoints constrained by the similarity of the motion flow or by the inducted local patch-based transformations. DTM employs an alternative neighborhood definition based on the Delaunay triangulation, which naturally fits into the keypoint distribution of the image and its structure, implicitly providing a sort of dynamic neighborhood without requiring to supply the neighborhood size. Unlikely~\cite{pfm}, appeared after the original submission of this manuscript, DTM considers Delaunay triangulations from both the images of the input pair to get more consistent and refined neighborhoods as intersection. This approach was already proposed in~\cite{sift_matching} with the aim of benchmarking descriptors for growing up good matches from an initial set of ground-truth correspondences.

\begin{figure}
	\center
	\subfloat[]{\label{boundary_img}
		\includegraphics[width=0.22\textwidth]{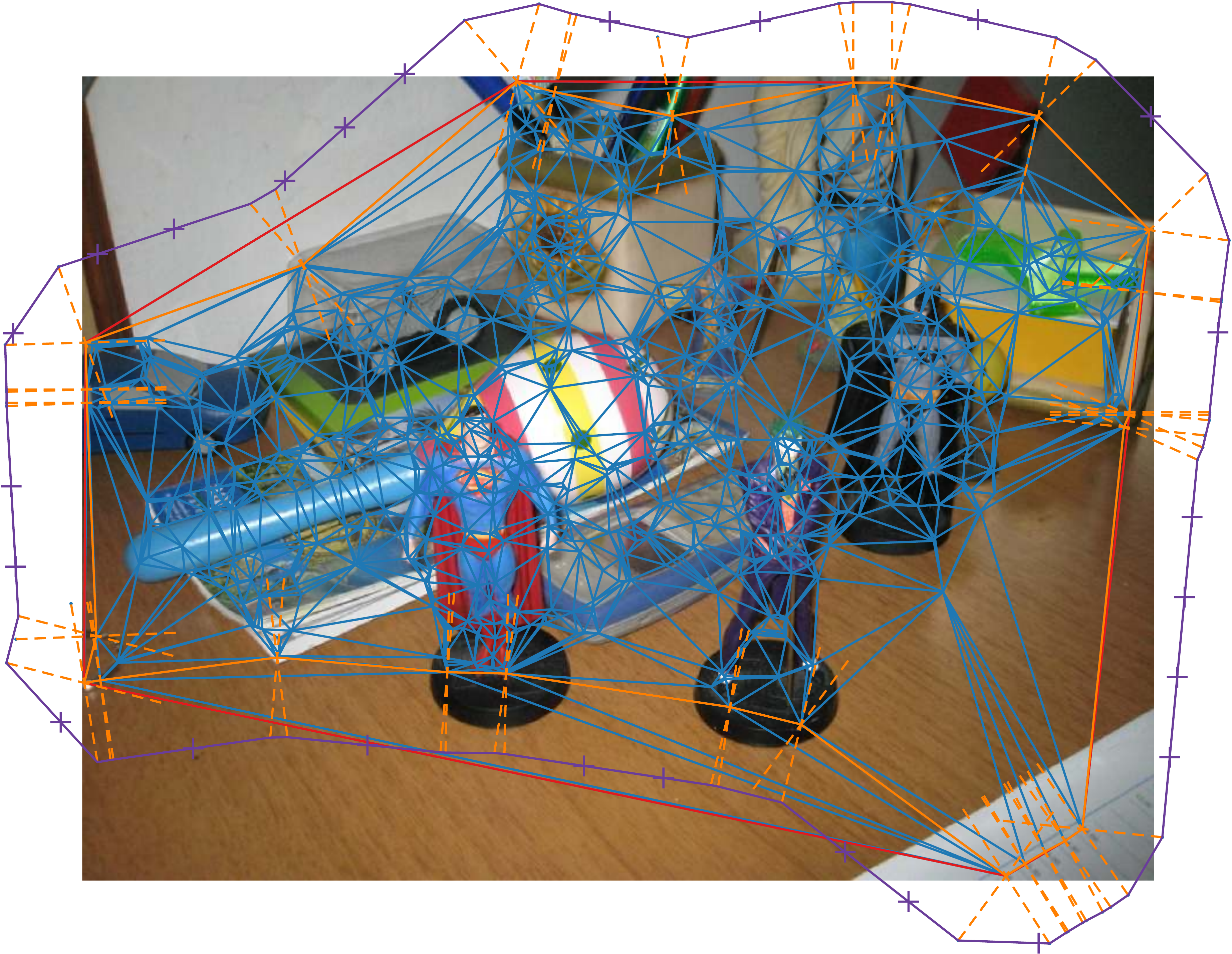}
	}
	\hfil
	\subfloat[]{\label{boundary_mesh_img}
		\includegraphics[width=0.22\textwidth]{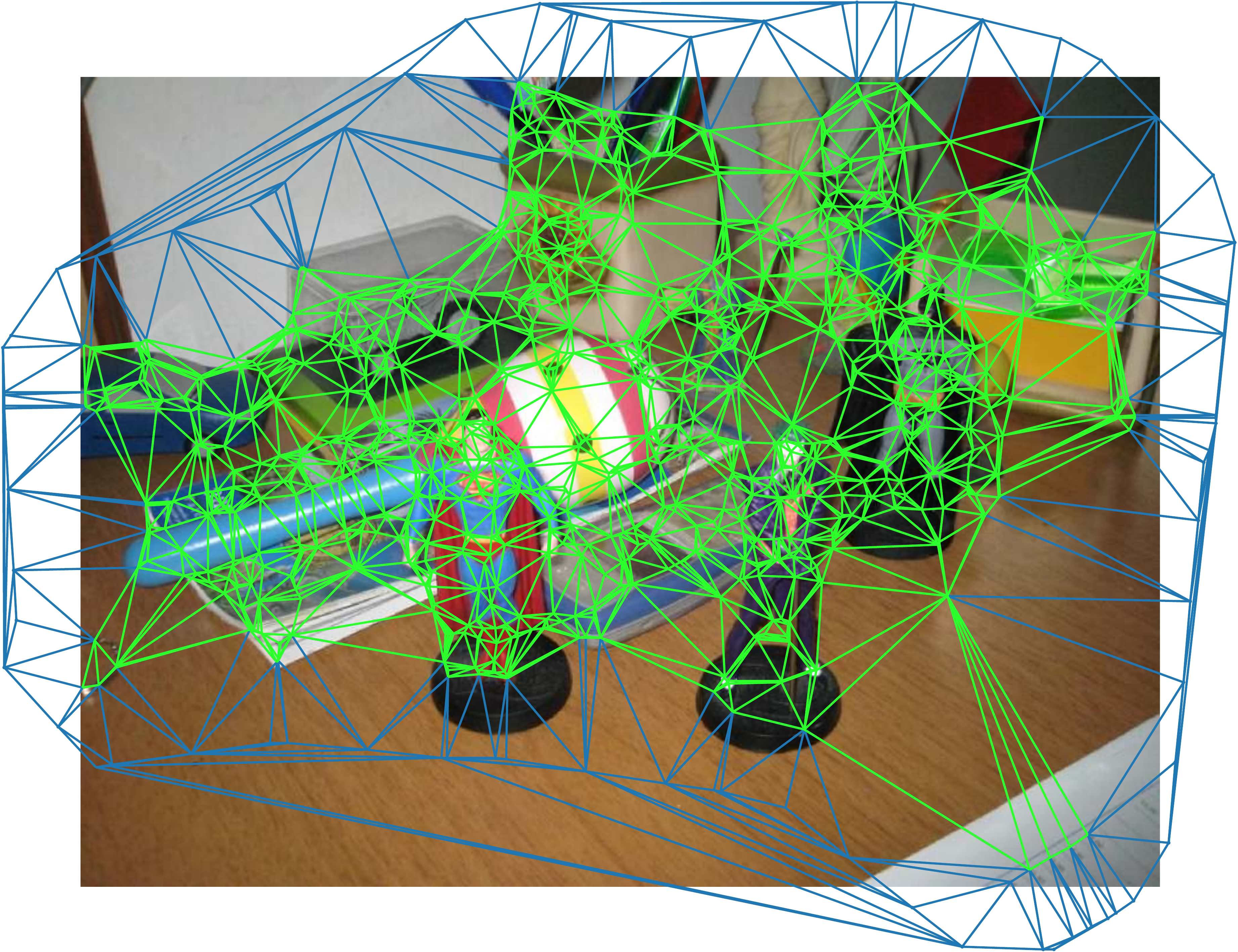}
	}
	\\[-0.5em]
	\caption{\label{border_steps}
		(left) original Delaunay triangulation (blue), convex hull (red), alpha shape boundary edges (solid orange), border fattening (dashed orange), and final contour edges (purple). (right) final Delaunay triangulation with keypoint-only marked edges (green) (see text for details, best viewed in color and zoomed in).}
	\vspace{-1em}
\end{figure}

When computing the Delaunay triangulation, only boundaries need some attention. Boundary edges of any Delaunay triangulation correspond to the convex hull of the considered keypoints, and triangles for keypoints on the edges of convex hull are generally not well-shaped and do not lead to appropriate neighborhoods. Adding the image corners does not solve the problem as well as breaking the image canvas borders into multiple lines. A more feasible solution is to expand the boundary edges and split them, where the boundary can be determined by the convex hull or, better, by alpha-shapes\footnote{Using the Matlab \texttt{boundary} function with default parameters.}, which also relies on Delaunay triangulation. A visual explanation of the different boundary choices is reported in Appendix~\ref{appendix_border}, while the alpha shape border computation adopted for DTM is illustrated in Fig.~\ref{border_steps}. Being $V$ a keypoint set, alpha-shape boundary edges (solid orange) are extracted and expanded. Specifically, from the two vertexes of each alpha-shape edge, the four points at a fixed distance $s$ from the edge\footnote{The value of $s$ is empirically set the to minimum between the width and the height of the image divided by 10.} lying on the line perpendicular to it are included in the set $B'$ (dashed orange). Alpha-shape boundary edges for $V\cup B'$ are then extracted and break down into segments of length $s$, whose vertexes provide the final boundary point set $B$ for the triangulation (purple).

The use of Delaunay triangulation in DTM is twofold: In the first DTM$_1$ stage, Delaunay triangulation is combined with the greedy matching strategy employed to define $S_{gNN}$ in Eq.~\ref{gnn} to iteratively prune matches. In the second DTM$_2$ stage, it is employed to grow up consistent matches from the previously surviving matches. Note that DTM input is only the set of keypoint correspondences and their similarity in the descriptor space context, without considering additional patch-based local transformation information for consistency. This make DTM more general and potentially robust when the relative patch-based local transformations are unavailable, unreliable or unable to provide a good approximation of the specific input motion field, e.g. when relying on patch-based similarity or affine local transformations in case of severe perspective distortions.

DTM$_1$ repeatedly and greedily removes inconsistent neighbor matches and restores the consistent ones, progressively adjusting neighborhoods. It is known that both the neighborhood graph and the minimum spanning tree are subsets of the Delaunay triangulation, which also maximizes the minimum angle of each triangle mesh. This late property gives rise to neighborhoods well spread among all directions. Moreover, as shown in Appendix~\ref{appendix_neighborhood}, in the ideal case Delaunay-based neighborhoods have intuitively better chances to contain correspondences consistent with the considered match and hence to restore accidentally removed good matches. 

\begin{figure}
	\center
	\subfloat[]{\label{dtm_step1}
		\includegraphics[width=0.22\textwidth]{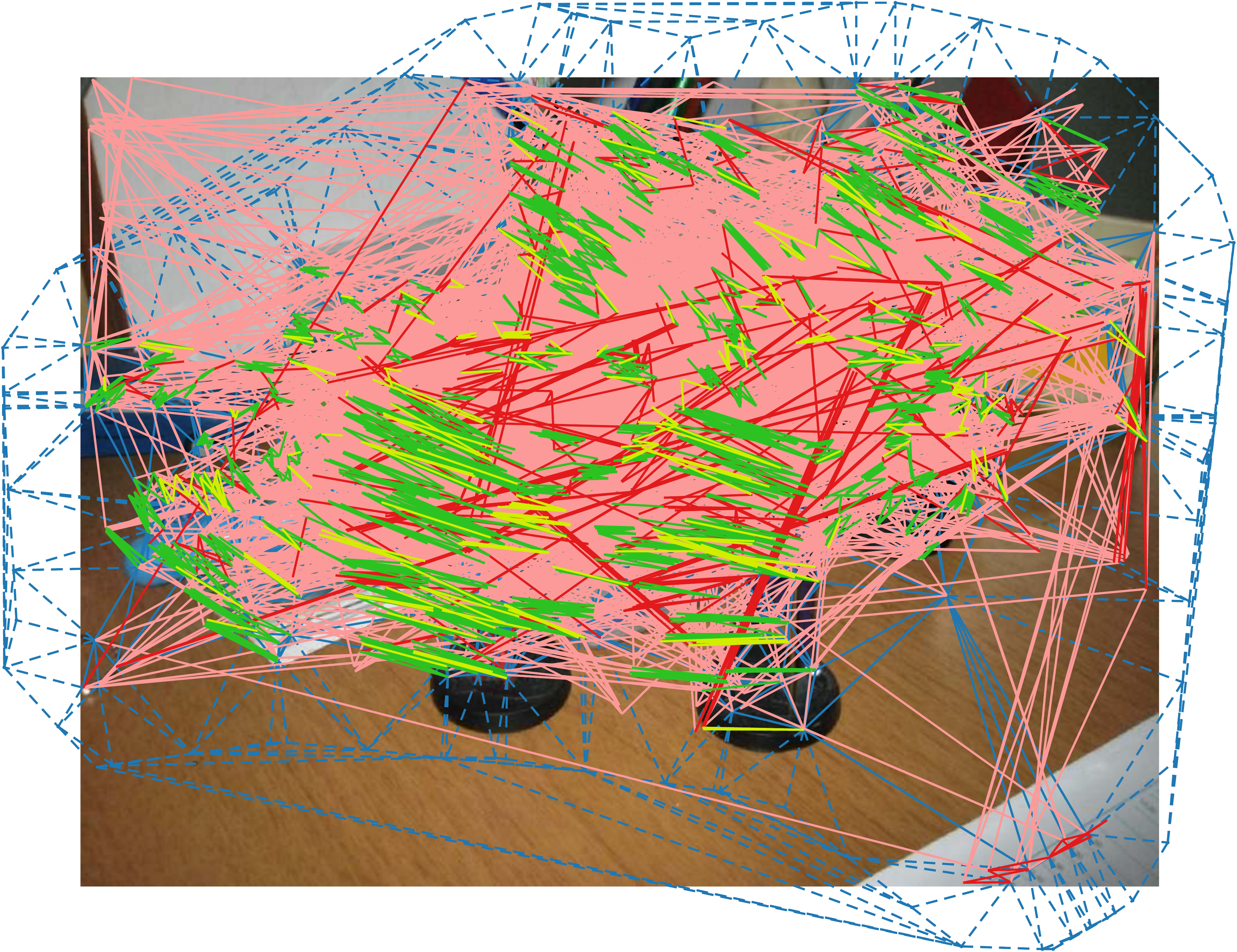}
	}
	\hspace{0.2em}
	\hfil
	\hspace{0.2em}
	\subfloat[]{\label{dtm_step1_ce}
		\includegraphics[width=0.22\textwidth]{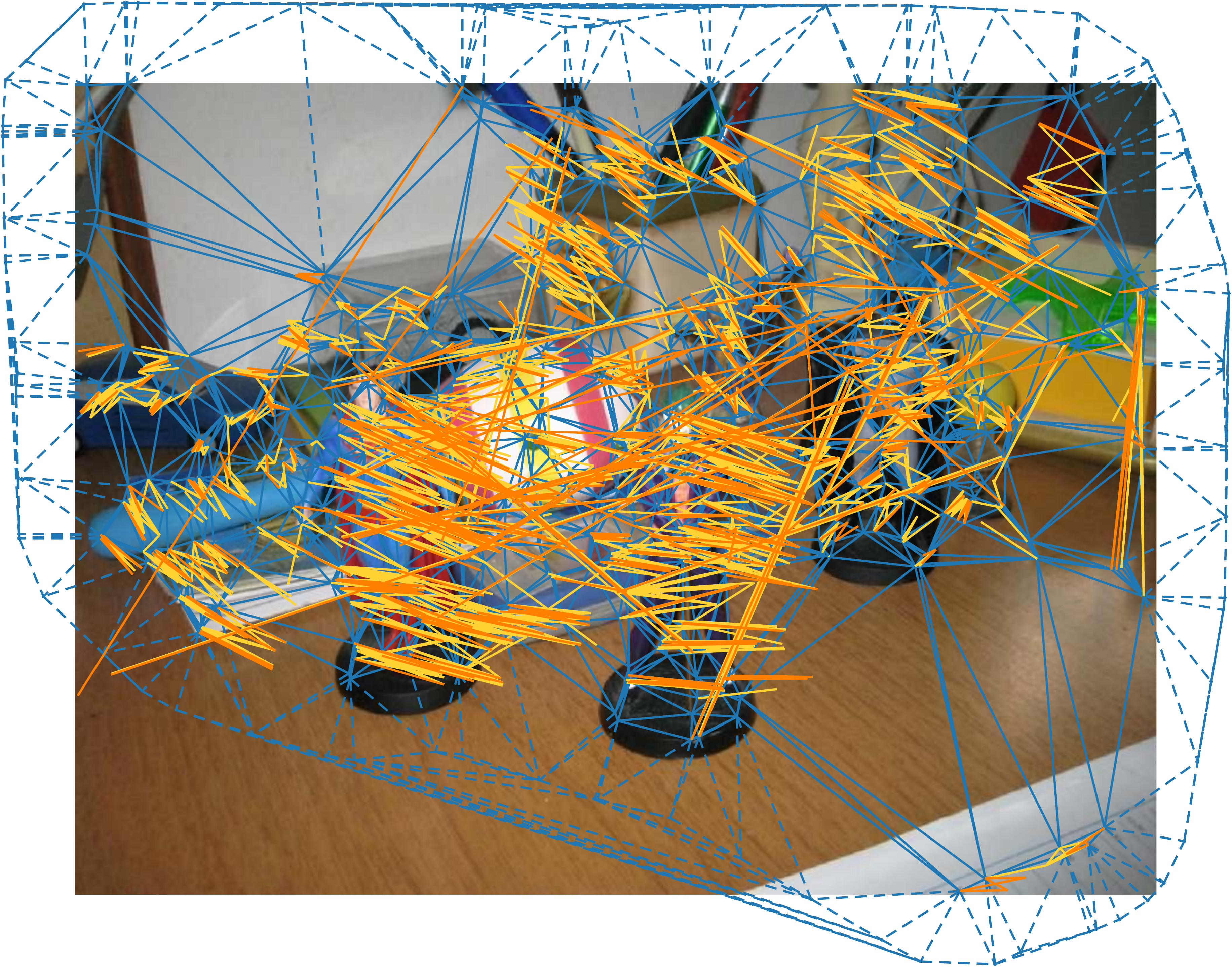}
	}
	\\[-0.75em]
	\subfloat[]{\label{dtm_step2}
		\includegraphics[width=0.22\textwidth]{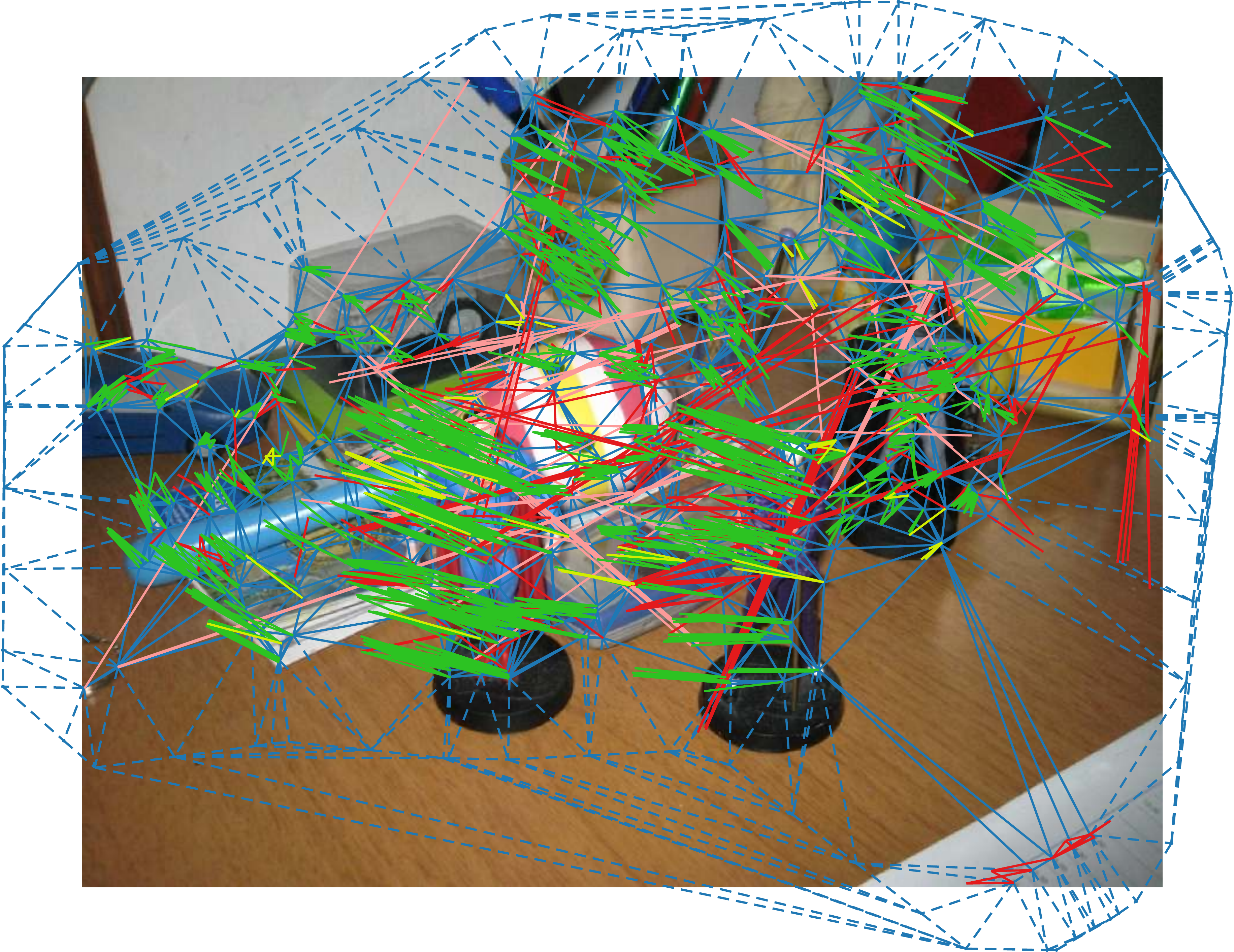}
	}
	\hfil
	\subfloat[]{\label{dtm_step2_ce}
		\includegraphics[width=0.22\textwidth]{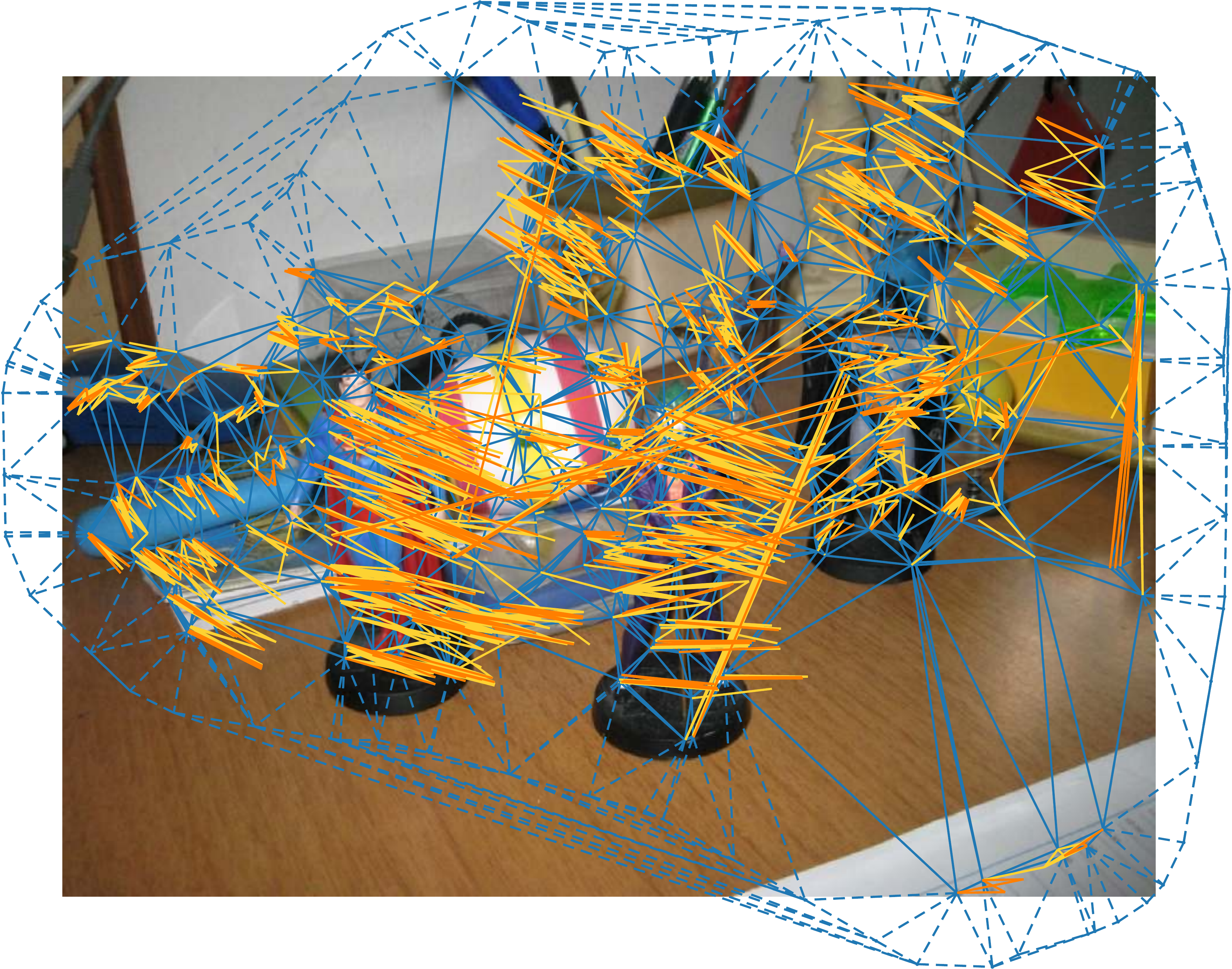}
	}
	\\[-0.75em]
	\subfloat[]{\label{dtm_step_end}
		\includegraphics[width=0.22\textwidth]{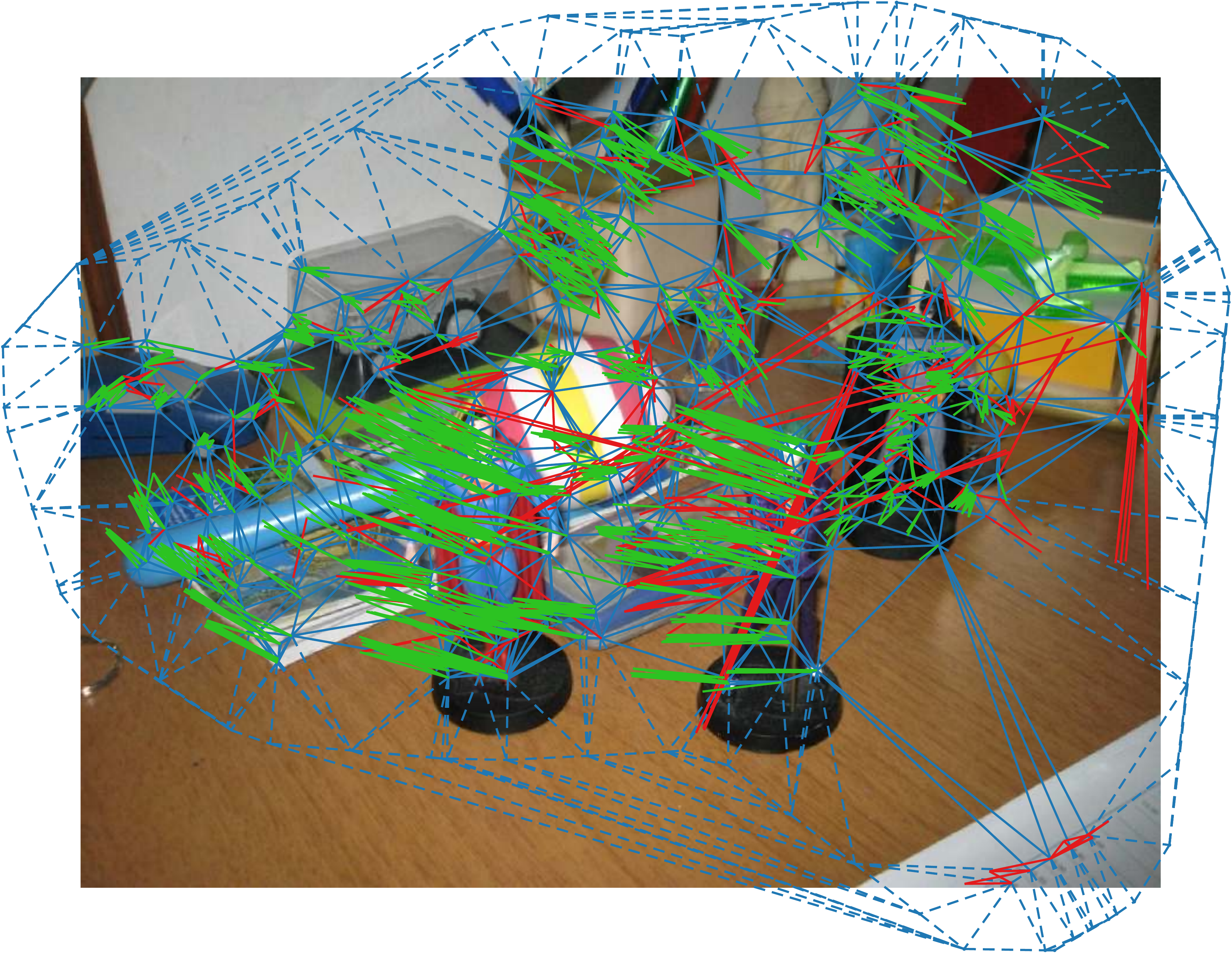}
	}
	\hfil
	\subfloat[]{\label{dtm_added}
		\includegraphics[width=0.22\textwidth]{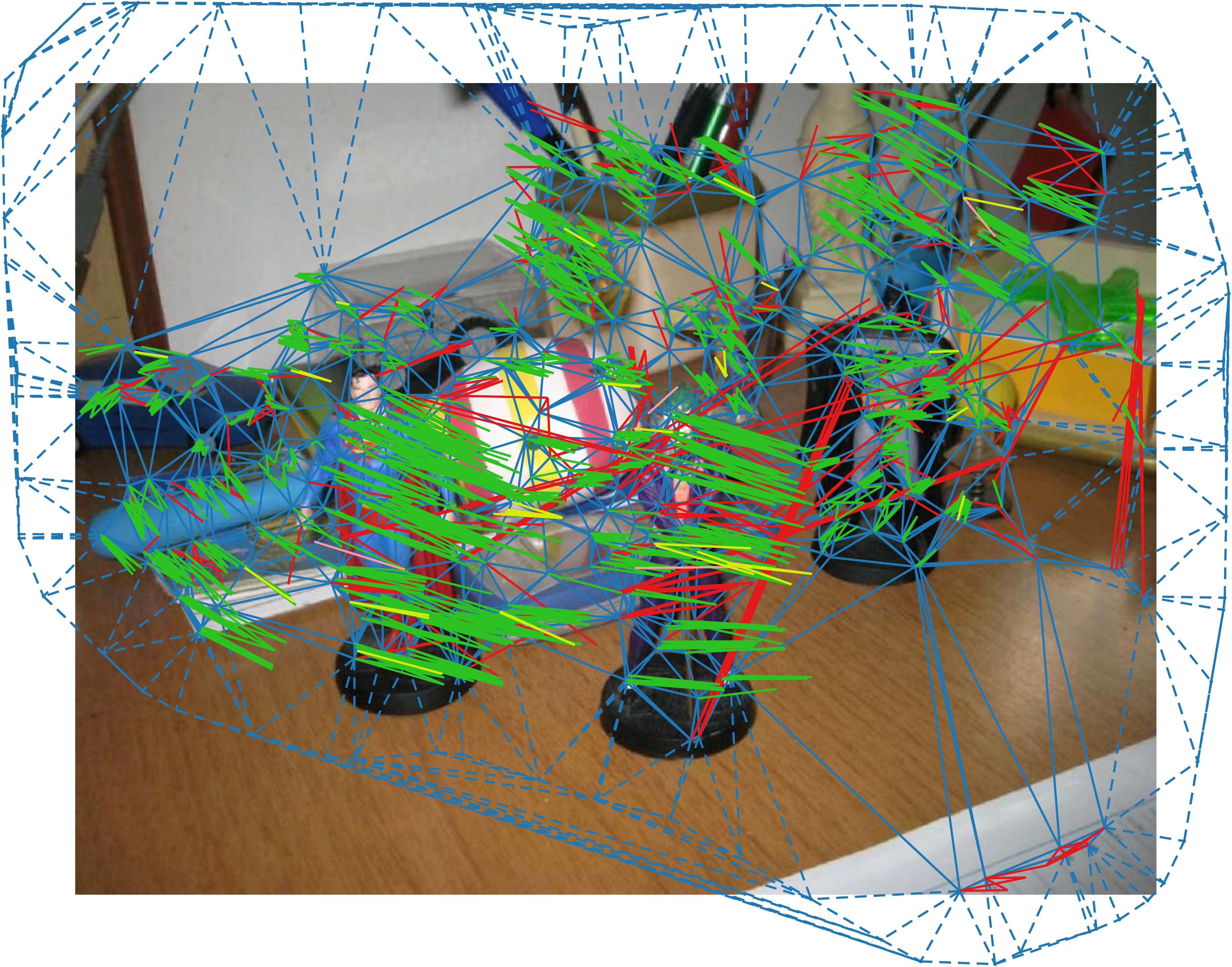}
	}
	\\[-0.5em]
	\caption{\label{dtm_steps}
		DTM computation steps. The first and second images of the input pair are superimposed on the left and on the right, respectively, together with their associated Delaunay triangulations (blue). Initial matches are obtained from the best configuration given by HarrisZ+SOSNet with blob matching (see Sec.~\ref{eval}). The first and second iterations $i$ of DTM$_1$ are reported as the first and second rows, respectively. For these rows, on the left, the clusters of vector flows for the retained (green and red) and pruned (yellow and light red) matches are shown. The clusters of correct (green and yellow) and wrong (red and light red) matches can be established as well, according to the evaluation protocol described in Sec.~\ref{eval}. Corresponding contraction (orange) and expansion (yellow) clusters are indicated on the right. Image \protect\subref{dtm_step_end} reports the final filtered matches at the fourth last iteration $\overline{i}$ of DTM$_1$, while image \protect\subref{dtm_added} shows the final matches after the last iteration $i=0$ of DTM$_2$. In this last image the colored clusters indicate the matches before DTM$_2$ (green and red) and those added (yellow and light red), correct (green and yellow) or wrong (red and light red) (see text for details, best viewed in color and zoomed in).}
		\vspace{-2em}
\end{figure}

Given an initial set of matches $M^0$, each iteration $i$ of DTM$_1$ iteratively prunes $M^i$ until $M^i=M^{i-1}$ (see Fig.~\ref{dtm_steps}) as described by the following steps:
\begin{enumerate}
	\item \emph{Extract keypoint locations for current (surviving) matches}. Set $K^i_1=M^{i-1}_{\downarrow\vert}$ and $K^i_2=M^{i-1}_{\vert\downarrow}$.
	\item \emph{Construct the current Delaunay triangulation of each image}. Round-off keypoint coordinates of $K_1$ to define the vertex sets $V_1=\{ (\lceil k_x \rfloor ,\lceil k_y \rfloor) \in K_1\}$. From $V_1$, compute the boundary set $B^i_1$ using alpha-shapes as described before, and build the Delaunay triangulation $\mathcal{T}_1$ for $I_1$ from $V_1\cup B^i_1$ (see Fig.~\ref{dtm_steps}, left column). Analogously, define $V_2$, $B^i_2$ and $\mathcal{T}_2$ (see Fig.~\ref{dtm_steps}, right column). Vertex collapsing by rounding-off avoids many too small triangles that can slow-down the computation.
	\item \emph{Define the local non-isotropic neighborhoods}. For each vertex $v\in V_1$, define $A^1_v$ as the set of vertexes adjacent to $v$ in the triangulation $\mathcal{T}_1$, including $v$ itself. Define also $M^i_{A^1_v}$ as the set of matches in $M^{i-1}$ each having the keypoint lying on $I_1$ collapsed into a vertex in $A^1_v$ by rounding-off, as described before. $A^2_v$ and $M^i_{A^2_v}$ are defined analogously for $I_2$.
	\item\label{sim_rank} \emph{Rank matches according to their coherence}. Assign a rank $r(m)$ to matches $m\in M^{i-1}$ collapsed into vertex pair $(v_l,v_r)$, by sorting them first according to their increasing descriptor similarity and then by the decreasing cardinalities $\vert M^i_{A^1_{v_l}}\cap M^i_{A^2_{v_r}}\vert$.
	\item \emph{Contract the Delaunay triangulations}. Set $T=\emptyset$, $M'=M^{i-1}$, and add the match $m\in M'$ ranked first according to $r(m)$ to $T$. Then remove this match $m$ from $M'$ as well as matches in $(M^i_{A^1_{v_l}}\cup M^i_{A^2_{v_r}})\setminus(M^i_{A^1_{v_l}}\cap M^i_{A^2_{v_r}})$, where $m$ collapsed into $(v_l,v_r)$. Repeat until $M'=\emptyset$ (see Fig.~\ref{dtm_step1_ce},\subref*{dtm_step2_ce}).
	\item \emph{Expand the Delaunay triangulations}. Define $M^i$ as the union of the sets $(M^i_{A^1_{v_l}}\cap M^i_{A^2_{v_r}})$, each one  obtained from the collapsing pair $(v_l, v_r)$ a match $m\in T$ corresponds to (see again Fig.~\ref{dtm_step1_ce},\subref*{dtm_step2_ce}).
\end{enumerate}
The set $M^{\overline{i}}$ at the last iteration $\overline{i}$ contains the matches survived to the Delaunay triangulation ``pulses'' (see Fig.~\ref{dtm_step_end}). The convergence is always guarantee since by construction the cardinality of $M^i$ cannot increase with the iterations $i$. In the worst case, no sufficient matches for the Delaunay triangulations are found in the last iteration, and DTM outputs no correspondences. Notice that DTM$_1$ greedy formulation does not involve parameters, unlikely other approaches requiring parameters to define energy minimization cost functions and criteria to stop the execution either to select the initial or final matches.  

DTM$_2$ employs the Delaunay triangulations of the correspondences survived to DTM$_1$ to approximate the motion field and hopefully restore consistent matches previously discarded. Delaunay triangulation allows a weak model assumption for motion field, i.e. correct matches should be inside corresponding triangular meshes, without any explicit motion field characterization as imposed by other formulations. 

In order to pick up good the matches accidentally discarded by DTM$_1$ since surrounded only by wrong matches, DTM$_2$ proceeds in reverse order, starting from iteration $i=\overline{i}-1$ downto 0, with $E^{\overline{i}-1}=M^{\overline{i}}=M^{\overline{i}-1}$ by the definition of $\overline{i}$ in DTM$_1$:
\begin{enumerate}
	\item \emph{Construct the Delaunay triangulations for the (estimated) good matches}. At step $i$, compute the Delaunay triangulations $\mathcal{T}'_1$ and $\mathcal{T}'_2$ as before, but from the collapsed vertex set of $E^i$ plus the boundary sets $B_1^i$ and $B_2^i$ computed in the DTM$_1$ stage.
	\item \emph{Add coherent matches according to the Delaunay triangulations}. Initially set $E^{i-1}=E^i$. For each match $m$ in $M_i\setminus M^{i-1}$, find the triangle $W_1$ of $\mathcal{T}'_1$ where the corresponding keypoint of $m$ on $I_1$ falls into. If the other keypoint of $m$ on $I_2$ falls into any triangle on $I_2$ formed by the corresponding vertexes of $W_1$, and this also holds when swapping the role of the images, add the match to $E^{i-1}$.
\end{enumerate}
The final set $E=E^0$ contains the enhanced matches (see Fig.~\ref{dtm_added}). Notice that DTM$_2$ uses the boundary sets $B_1$ and $B_2$ computed in DTM$_1$ in order to increase the chance to include previously discarded keypoints close to the boundary. Like the previous stage, no parameters get involved in the computation.

\section{Evaluation}\label{eval}
\subsection{Setup}\label{eval_setup}
The evaluation pipeline is composed by the following steps: keypoint extraction, local descriptor computation, descriptor matching, local spatial filtering and model fitting. For the keypoint extraction the SIFT and HarrisZ detectors are considered, while RootSIFT, RootsGLOH2, HardNet2 and SOSNet are employed as local descriptors. SIFT and RootSIFT are included as baselines. Patch orientation is estimated according to~\cite{learning_ori} for all descriptors, except for RootsGLOH2 that needs no orientation-adjusted patches. Descriptor matching, the next step of the pipeline, is achieved only by blob matching, since it can behave as the common descriptor matching strategies with a proper tuning of the parameters. The goal of this evaluation step is to check the advantages offered by match pre-filtering, many-to-many matches, and the alternative distance definitions and combinations (referring in order to the four different steps of blob matching in Sec.~\ref{ds}). The next step of the pipeline evaluates spatial matching filters, including the proposed DTM and fourteen state-of-the-art filters, learned or not: LMR, LPM, GLPM, GMS, VFC, LLT, RFM-SCAN, AdaLAM, OANet, ACNe, PFM, PGM, SCV and BM (see Sec.~\ref{related_work}). For reference, the standard 0.8 NNR threshold is also included, although not properly a local spatial filter, and indicated as `th'. For the last step of the pipeline, being $q$ the minimal number of matches required to estimate the model\footnote{The value of $q$ is respectively set to 4, 8 for homography, fundamental matrix.}, only a simple global model estimation using uniquely one sample made up of the $3\times q$ top-ranked surviving matches at the previous step is employed. Although this approach, named 1SAC (one SAmple Consensus) is quite naive, it can give insights on more complex RANSAC approaches. Notice that except for th, GLPM and DTM (step~\ref{sim_rank} of stage DTM$_1$ in Sec.~\ref{ks}), other pruning methods do not take into account the descriptor context, i.e. the descriptor similarity.

For the evaluation both planar and non-planar scenes are considered, the latter being more complex due to the inclusion of spatial discontinuities caused by occlusion and parallax. In the case of planar scenes, the 15 sequences employed in~\cite{sift_matching} were used. Each sequence is made up of 6 images where the first one is fixed as reference, for a total of $19\times(6-1)=75$ image pairs. In the case of non-planar scenes, two different datasets with distinct evaluation protocols are considered. The first one, explicitly referred to as the non-planar dataset, includes the 72 image pairs from~\cite{sift_matching}, and additionally 27 image pairs already known in the computer vision community, for a total of $72+27=99$ image pairs belonging to 61 different scenes. For each image pairs of this dataset, a sparse set of hand-taken ground-truth correspondences and occluded points is available. Appendix~\ref{appendix_dataset} shows the scenes contained in the planar and non-planar datasets. The second dataset, SUN3D~\cite{sun3d}, contains 415 indoor sequences, whose only 401 of them were supplied with the additional data required to extrapolate almost dense ground-truth correspondences. The training sequences of ACNe, corresponding to half of the SUN3D dataset are also included in the evaluation, since as reported in the additional material no relevant differences were observed when taken into account. For each sequence in SUN3D, a maximum of 30 image pairs, uniformly distributed among the sequence time interval, were chosen, for a total of 11231 pairs. Images making each SUN3D pair correspond to a time step of 80 frames, unless the maximal visual overlap between the images is less than 25\%, in this case the frame step is lower.

The proposed evaluation relies on the computation of ground-truth matches. On the planar dataset, ground-truth correspondences can be easily obtained by estimating the homography that maps one-to-one points across the two images~\cite{descriptor_eval,hpatches}. On the non-planar dataset, unless 3D data are supplied, only a point-to-line mapping through the fundamental matrix is available according to the epipolar geometry. A common approach to evaluate RANSAC-like methods defines inlier matches (hence ground-truth matches) according to the distance between a point and the epipolar line imaging the corresponding point to be matched in the other image (here denoted as method A). However, this approach can lead to many false positive ground-truth matches due to the ambiguity of the map. Other approaches measure the error on the fundamental matrix obtained at the end of matching process with respect to a ground-truth one, which can be estimated by hand-taken correspondences~\cite{noransac} or on the basis of robust SfM approaches~\cite{imw2020,imw2021}. The error distance between corresponding ground-truth and estimated epipolar lines can be considered to judge the goodness of the matching, statistically~\cite{zhang_fun_mat}, or on the basis of an uniform image sampling~\cite{noransac}. The former approach has been employed in a recent evaluation~\cite{fund_mat_eval}. Alternatively, the pose error of the camera~\cite{imw2020} is obtained from the estimated fundamental matrix (to be understood in a broad sense). Nevertheless, all these evaluations give an indirect measure on the goodness of matches that does not guarantee a true evaluation of the matching process. This happens since there is no direct and clear formula relating fundamental matrix correctness and correct matches, so that from the perspective of evaluating the accuracy of the correspondences, this kind of solutions can be associated to method A. Notice that although SfM can estimate depth data to classify correct matches, these are in most cases incomplete or contain errors, so that a further possible solution can be addressed by limiting the evaluation to synthetic datasets~\cite{imw2021}. Finally, another approach is to compute ground-truth matches by relying on additional constraints inducted for instance by employing more images~\cite{perona} or by manually limiting the spatial localization of the matches~\cite{sift_matching}. This last solution is adopted in following evaluation according to the protocol described in~\cite{sift_matching} based on the approximated overlap error (denoted as method B), extended to cope with the issues discussed hereafter.
\begin{figure}
	\center
	\includegraphics[width=0.17\textwidth]{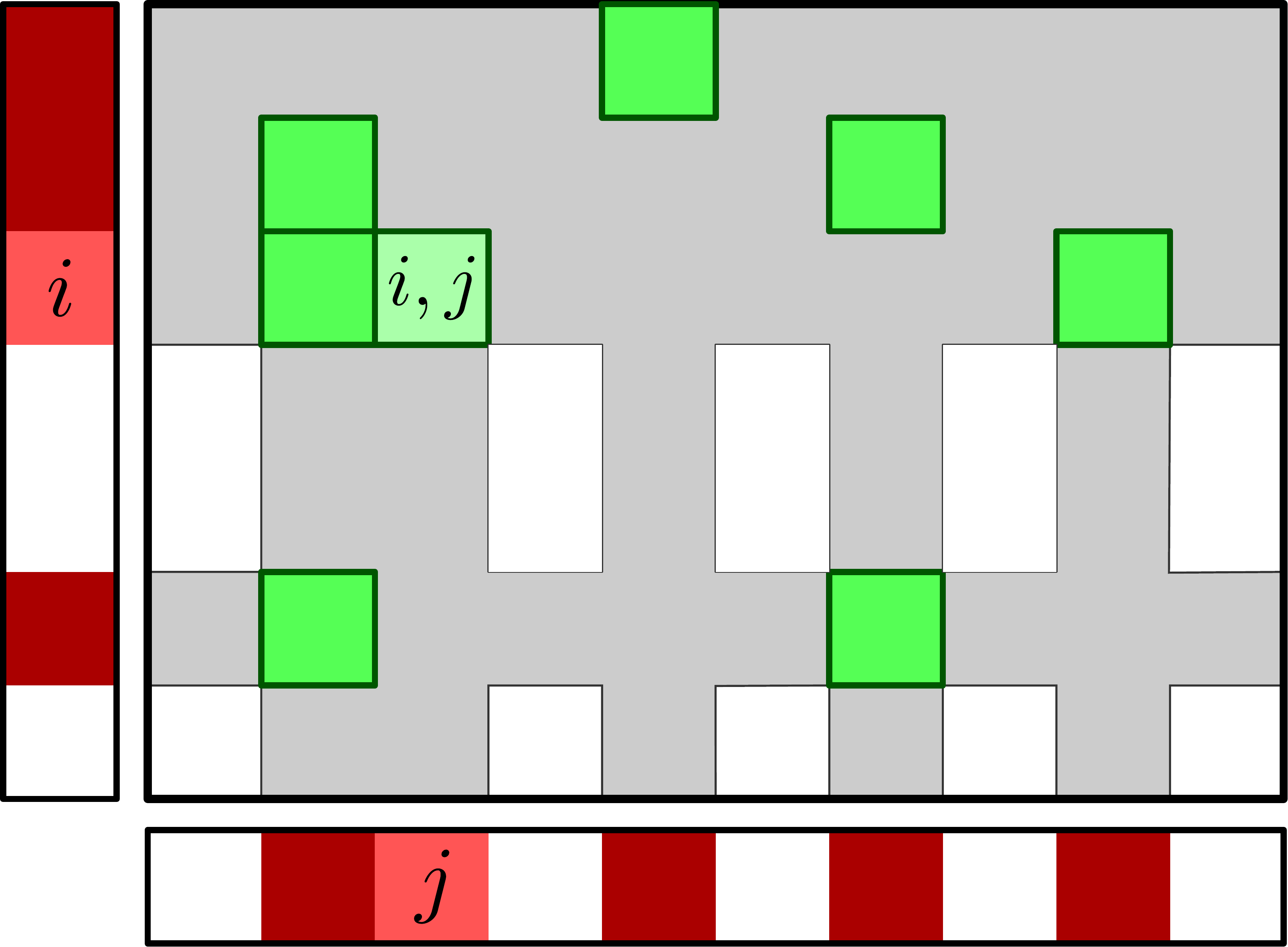}
	\caption{\label{norm_repeatability}
		Visual representation of the normalized recall on a given set of many-to-many matches $(i,j)$ (green). $\vert Z_{\downarrow\vert}\vert=4$ and $\vert Z_{\vert\downarrow}\vert=5$ by counting the numbers of elements (red) in the row and column outer sets, respectively. In the example $\vert Z\vert=\min(4,5)=4$ (see text for details, best viewed in color and zoomed in).}
	\vspace{-1em}
\end{figure}
True or approximated patch overlap can lead to the presence of large patches with low overlap error but where the keypoint center distance measured in pixels not acceptable by visual inspection. To cope with this, homography reprojection and the epipolar line distance errors are employed (limited to 30 pixels) as additional constraints to the true or approximated overlap errors (limited to 50\%), providing visually adequate results (method C). Moreover, taking into account that not all the local spatial filters handle the shape and size of the patch, an evaluation based only on the keypoint center distance disregarding the patch overlap can be conceived to define ground-truth matches (method D). Specifically, in the planar case the distance between a keypoint and the reprojection of the corresponding keypoint in the other image must not exceed 15 pixels. For the non-planar case the epipolar distance must not exceed 15 pixels but the matches must also pass the check defined on the basis of their spatial localization~\cite{sift_matching}. With respect to method D, method C obtains a lower number of correct matches due to the scale constraints according to the patch shapes. A comparison of the ground-truth estimation methods A, B, C and D is reported in Appendix~\ref{appendix_setup}, upon which after visual inspection method D has been selected for the planar and non-planar datasets. For the SUN3D dataset, an analogous of method D is employed. More specifically, each of the corresponding keypoints of a match is reprojected from one image to the other by exploiting depth and the extrinsic camera matrix data. The match is considered correct if at least in one case the reprojection error is less than 15 pixels or, in the remote eventuality that no depth estimation is available, if the maximum epipolar error is less than 15 pixels.    

Lastly, the definition of the recall is adjusted to handle many-to-many matches and avoiding apparent boosted results. In particular, using the same notation of Sec.~\ref{ds}, given the set $Z$ of good matches according to the ground-truth, the associated number of correct matches necessary to compute the recall is defined as $\min(\vert Z_{\downarrow\vert}\vert,\vert  Z_{\vert\downarrow}\vert)$ instead of $\vert Z\vert$, so that multiple keypoint associations are only taken into account once (see Fig.~\ref{norm_repeatability}). The precision is instead computed as usual. Notice that only in the case of mutual one-to-one matches $\vert Z\vert=\min(\vert Z_{\downarrow\vert}\vert,\vert  Z_{\vert\downarrow}\vert)$, so that the new recall definition extends the standard one. This sort of normalization is also employed to compose the top-ranked sample in 1SAC to handle many-to-many matches.

The evaluation code and data are freely available to support the reproducibility of the results~\footnote{\url{https://sites.google.com/view/fbellavia}}.

\begin{figure*}[h!]
	\center
	\subfloat[]{\label{heat_map}
		\includegraphics[width=0.30\textwidth]{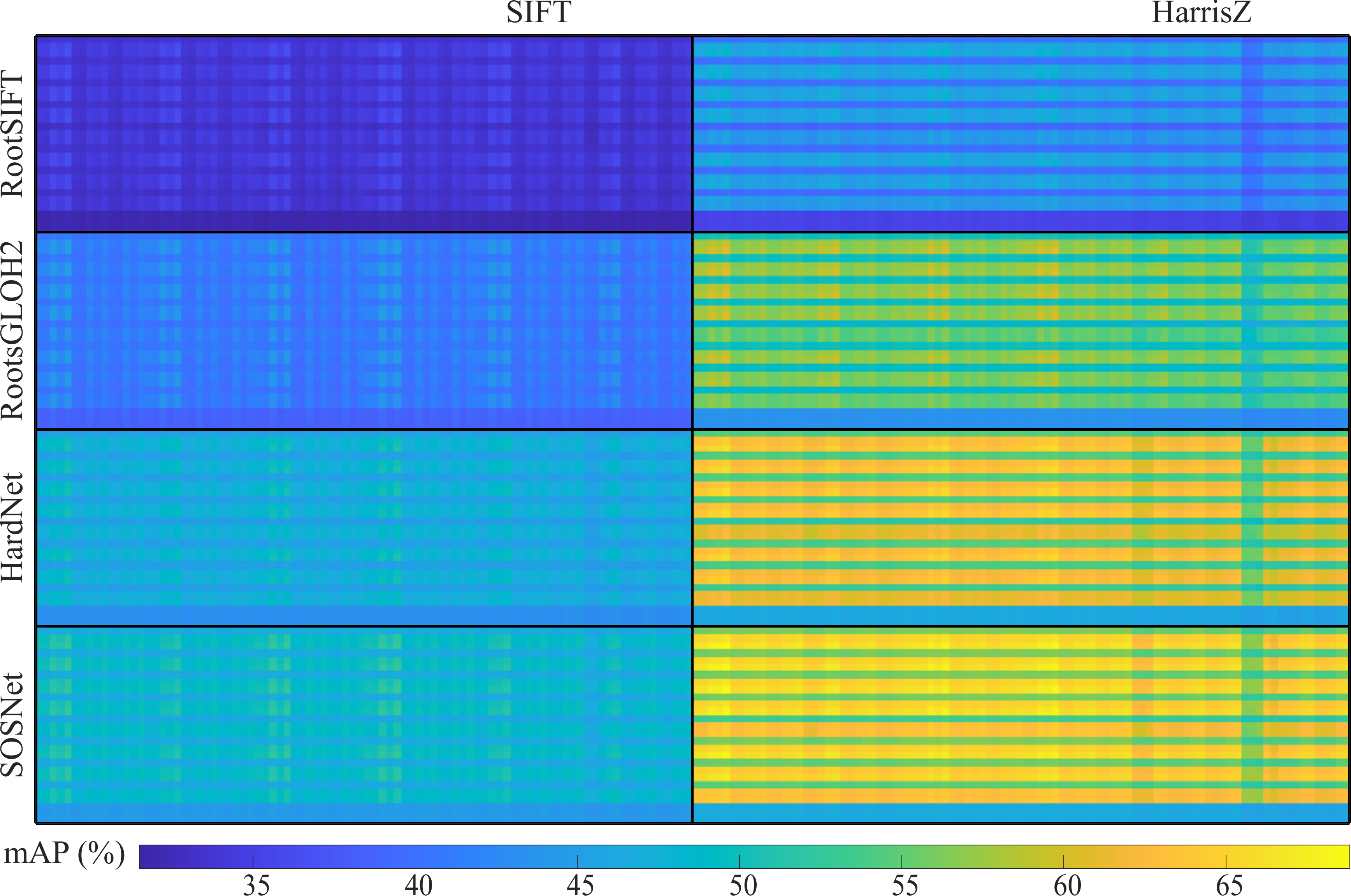}
	}
	\hfil
	\subfloat[]{\label{heat_map_details}
		\includegraphics[width=0.62\textwidth]{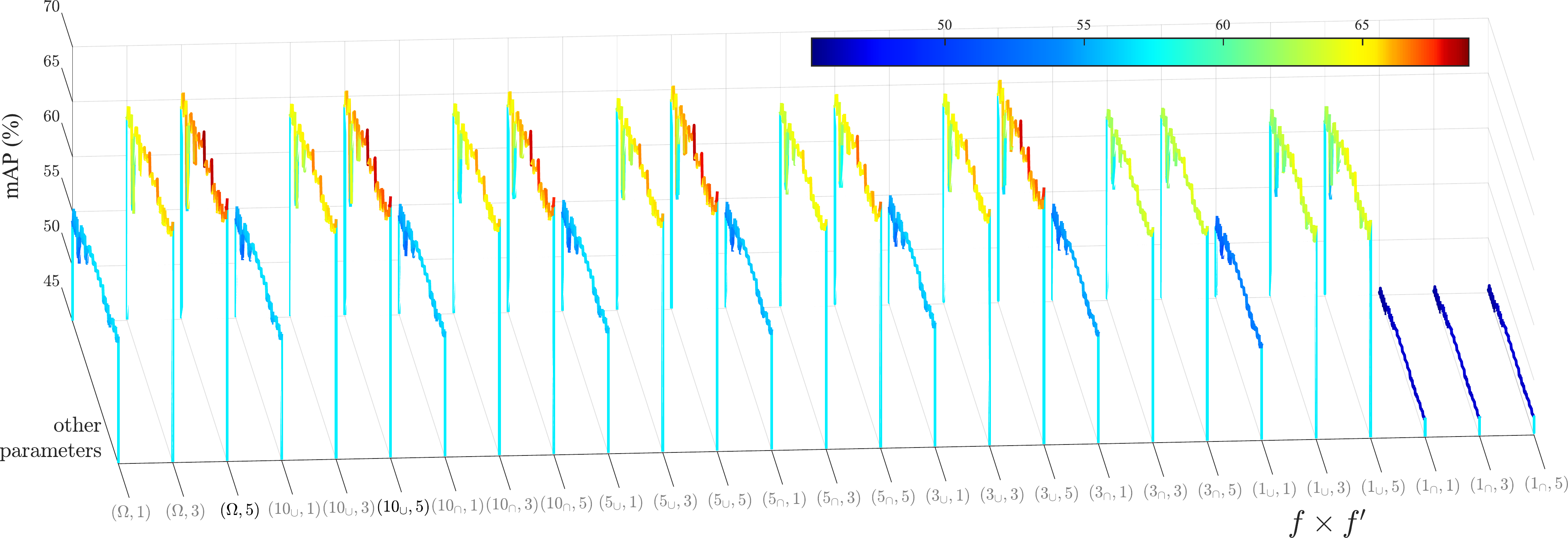}
	}
	\caption{\label{blob_plots}
		\protect\subref{heat_map} mAP for blob matching averaged on the image pairs of the planar and non-planar datasets for different setups and~\protect\subref{heat_map_details} for Harrisz+SOSNet only, with ground-truth matches estimated by method D (see text for details, best viewed in color and zoomed in).} 
	\vspace{-1em}	
\end{figure*}

\subsection{Blob matching results}\label{eval_blob}
Figure~\ref{heat_map} shows the heat map depicting the mean Average Precision (mAP) of blob matching for different setups. The mAP values are averaged on all the image pairs considering the planar and non-planar datasets, with ground-truth estimated according to method D. For method C, based on the original patch overlap, a similar ranking has been obtained (not shown). Each rectangular area in Fig.~\ref{heat_map} corresponds to a different detector+descriptor pair. Inside each rectangle, a row corresponds to a different $f\times f'$ pair with $f\in\{1_\cup,1_\cap,3_\cup,3_\cap,5_\cup,5_\cap,10_\cup,10_\cap,\Omega\}$ and $f'\in\{1,3,5\}$ (see Fig.~\ref{blob_fig} for the definition of $\Omega$). Likewise, each column represents the remaining parameter triplets $t_o\times\mathcal{D}\times\mathcal{W}$ with $t_o\in\{\infty,50\%,75\%,99\%,5\,\text{px},10\,\text{px}\}$, $\mathcal{D}=\{\mathcal{D}_\geq,\mathcal{D}^+_\geq,\mathcal{D}^+\}$ and $\mathcal{W}(a,b)\in\{a,b,\min(a,b),\max(a,b),(2ab)/(a+b)\}$. According to Fig.~\ref{heat_map}, HarrisZ provides better mAP results than SIFT, probably due to the more strict keypoint selection criteria of HarrisZ with respect to SIFT. The average number of ground-truth matches is 847/701, 750/532 in the case of the planar/non-planar datasets for HarrisZ and SIFT, respectively. Moreover, confirming previous benchmarks SOSNet and HardNet provide the best accuracy results followed by RootsGLOH2 and RootSIFT.

Under the same blob matching setup, mAP correlation between different detector+descriptor pairs is high. Specifically, the correlation is more than 90\% with the exception of SIFT+RootsGLOH2 with respect to any other detector+descriptor pairs, for which is yet higher than 60\%. According to these observations, only the best pair HarrisZ+SOSNet is chosen for a more detailed analysis. Figure~\ref{heat_map_details} plots the mAP values for the HarrisZ+SOSNet pair. The $f\times f'$ pairs $(\Omega,5)$ and $(10_\cup,5)$ are those providing the best mAP values (respectively 68.8\% and 68.7\%, highlighted in Fig.~\ref{heat_map_details}), while the corresponding one-to-one matching setup $(\Omega,1)$ and $(10_\cup,1)$ obtain mAP values around 57\%. This suggests that one-to-one matching can discard a lot of correct candidate matches. Moreover, the close results obtained by $f=\Omega$ and $f=10_\cup$ indicate that it is sufficient to inspect only the first 10 top-ranked matches when designing a matching strategy. This observation can be exploited to improve the computational efficiency since many wrong matches can be discarded a priori.

\begin{figure}[!h]
	\center
	\subfloat[]{\label{detail_sosnet}
		\includegraphics[width=0.46\textwidth]{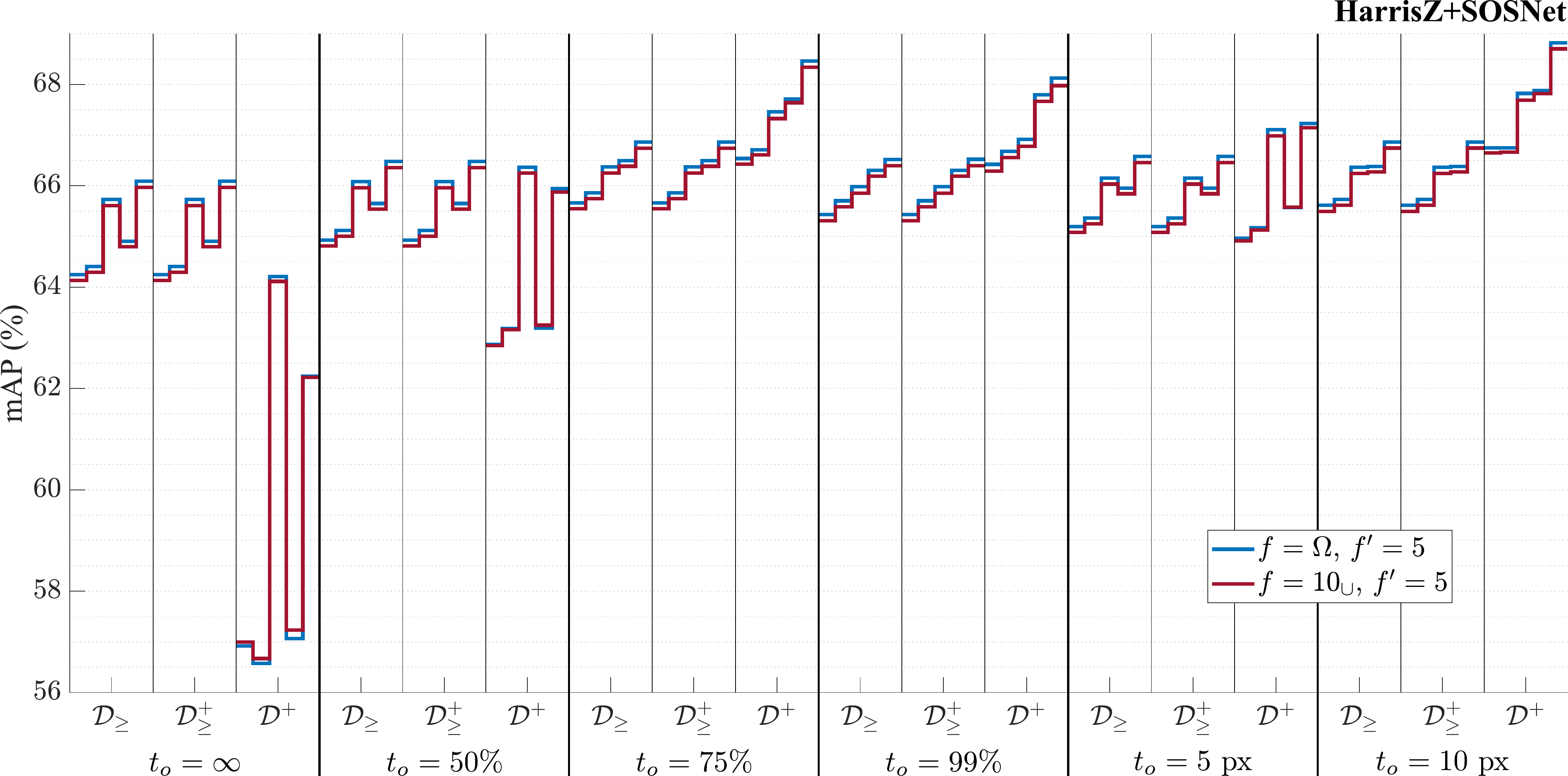}
	}
	\\
	\vspace{-0.33em}
	\subfloat[]{\label{detail_rootsift}
		\includegraphics[width=0.46\textwidth]{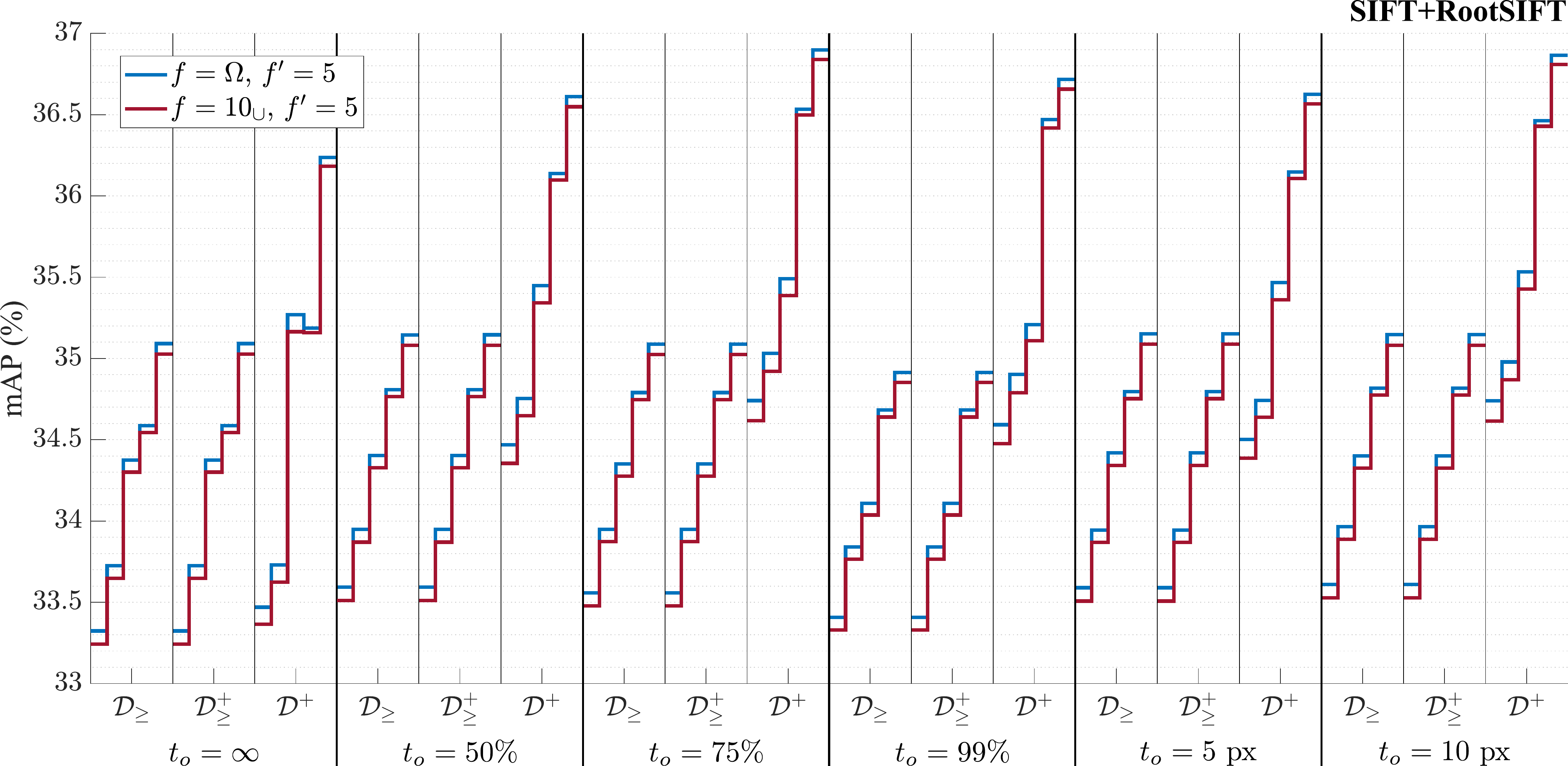}
	}
	\\
	\caption{Detailed mAP of blob matching with~\protect\subref{detail_sosnet} HarrisZ+SOSNet and~\protect\subref{detail_rootsift} SIFT+RootSIFT for different setups averaged on the whole dataset. $\mathcal{W}(a,b)$ values corresponding to $a$, $b$, $\min(a,b)$, $\max(a,b)$ and $(2ab)/(a+b)$ are reported in order inside each vertical sub-band (see text for details, best viewed in color and zoomed in).}
	\vspace{-1em}
\end{figure}

Figure~\ref{detail_sosnet} plots the mAP values for the remaining blob matching parameters for the best configuration found so far. For reference, the corresponding plots in the case of SIFT+RootSIFT is reported in Fig.~\ref{detail_rootsift}. Among NNR-like similarities, $\mathcal{D}^+$ is generally the best choice. This is reasonably expected, since $\mathcal{D}^+$ is designed for many-to-many matches ($f'=5$ in the evaluated configurations), while $\mathcal{D}^+_\geq$ does not provide any improvement with respect to $\mathcal{D}_\geq$. FGINN with \mbox{$t_o=75$} \% or $t_o=10$ px provides substantial improvements with respect to the base setup with no FGINN ($t_o=\infty$), but only when the setup includes $\mathcal{D}^+$. Inside each sub-band of the plots, mAP values are reported in order considering the different $\mathcal{W}$ choices. While there is no evident difference using one image or another as reference, their combined distances improve the results. In particular, the minimum seems to achieve better results when FGINN has no or small ranges on the opposite of the maximum, and in any case the harmonic mean equals or surpasses the best among the previous $\mathcal{W}$ choices. As observed in~\cite{imw2020}, the more the matches are discriminative, that happens when FGINN is employed with a sufficient range, the more their combination by union, which is equivalent to use the maximum, is better. The opposite holds for their intersection, which is equivalent to use the minimum. According to these evaluation the best blob matching setup is $f=10_\cup$, $f'=5$, $\mathcal{D}^+$ with $t_o=10$ px or $t_o=75$ \%, and $\mathcal{W}(a,b)=(2ab)/(a+b)$.

\subsection{Delaunay Triangulation Matching results}\label{eval_dt}
\begin{figure*}
	\center
	\includegraphics[height=0.065\textwidth]{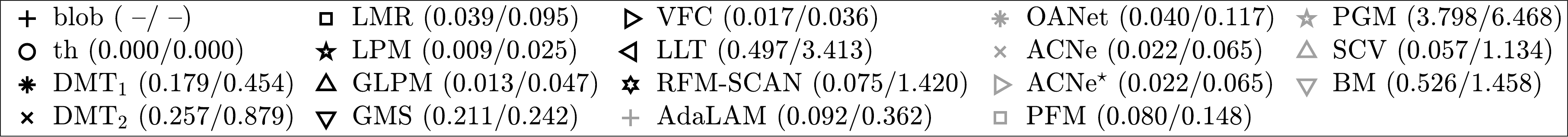}\\
	\rotatebox[origin=c]{90}{\hfil non-planar\hspace{7em}planar\hfil}
	\subfloat[global behaviour]{\label{plot_all}
		\begin{minipage}[c]{0.21\textwidth}
			\includegraphics[height=\textwidth]{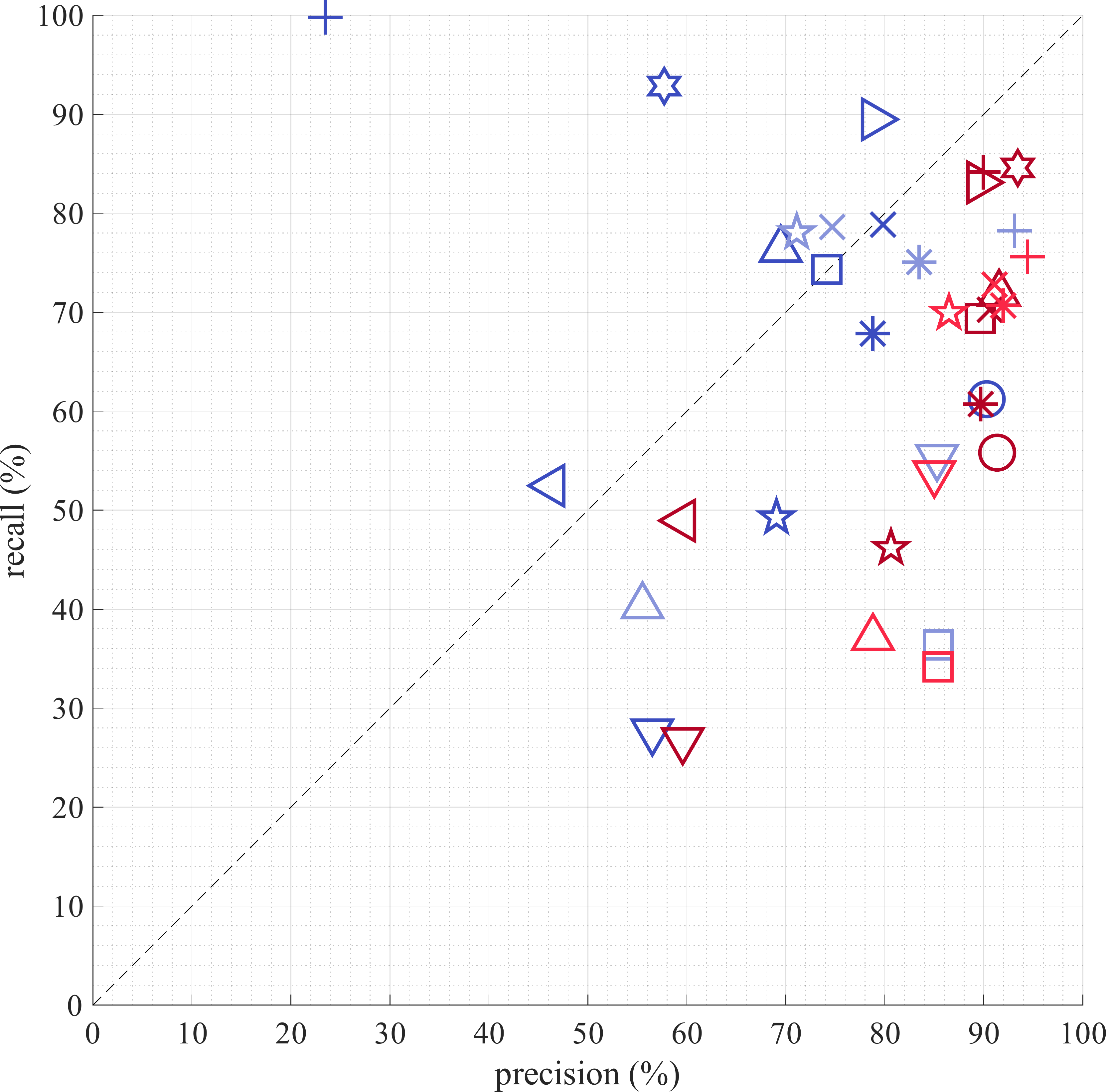}
			\\
			\includegraphics[height=\textwidth]{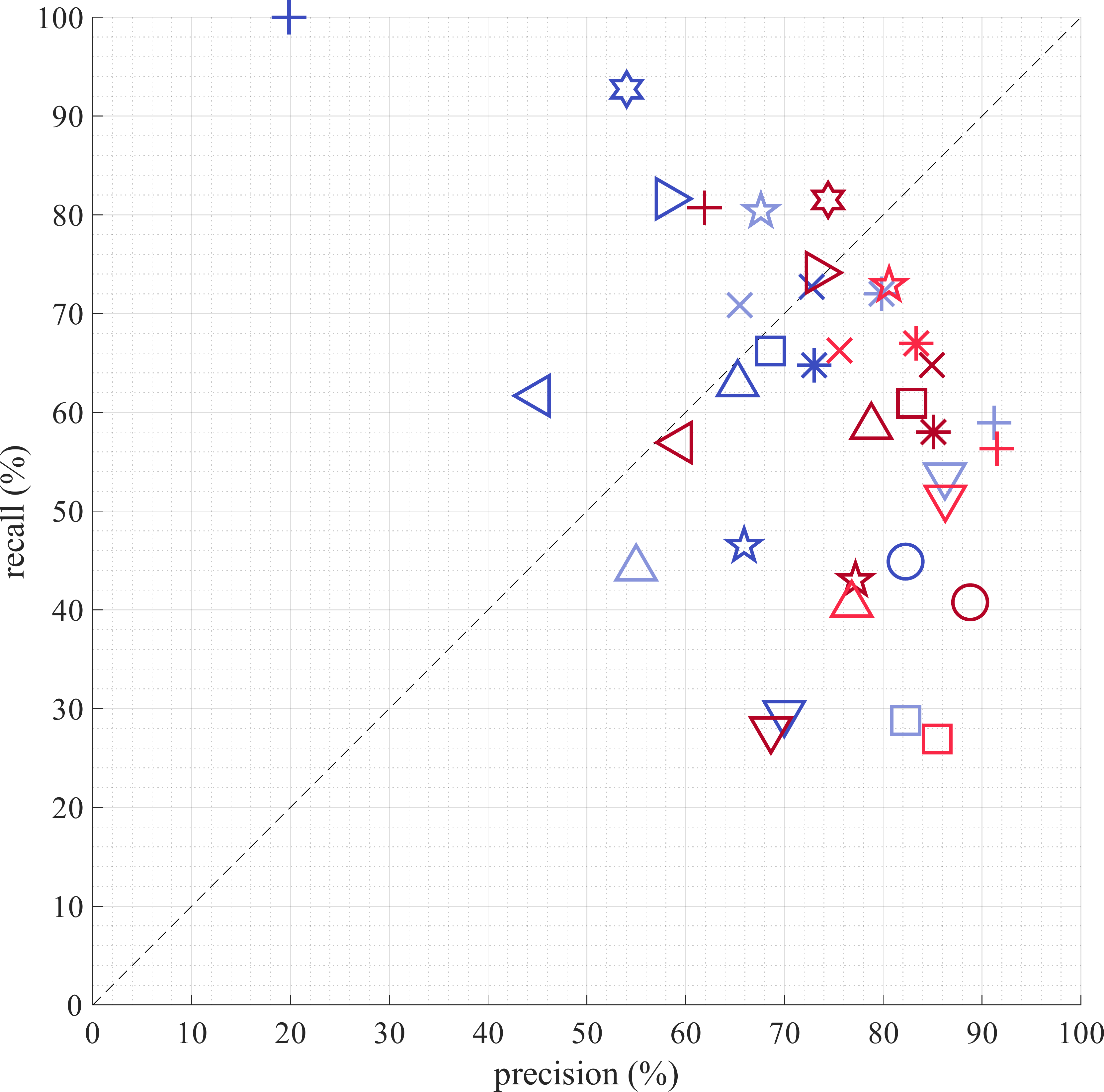}
		\end{minipage}
	}
	\subfloat[baseline configuration]{\label{plot_sift_rsift}
		\begin{minipage}[c]{0.21\textwidth}
			\includegraphics[height=\textwidth]{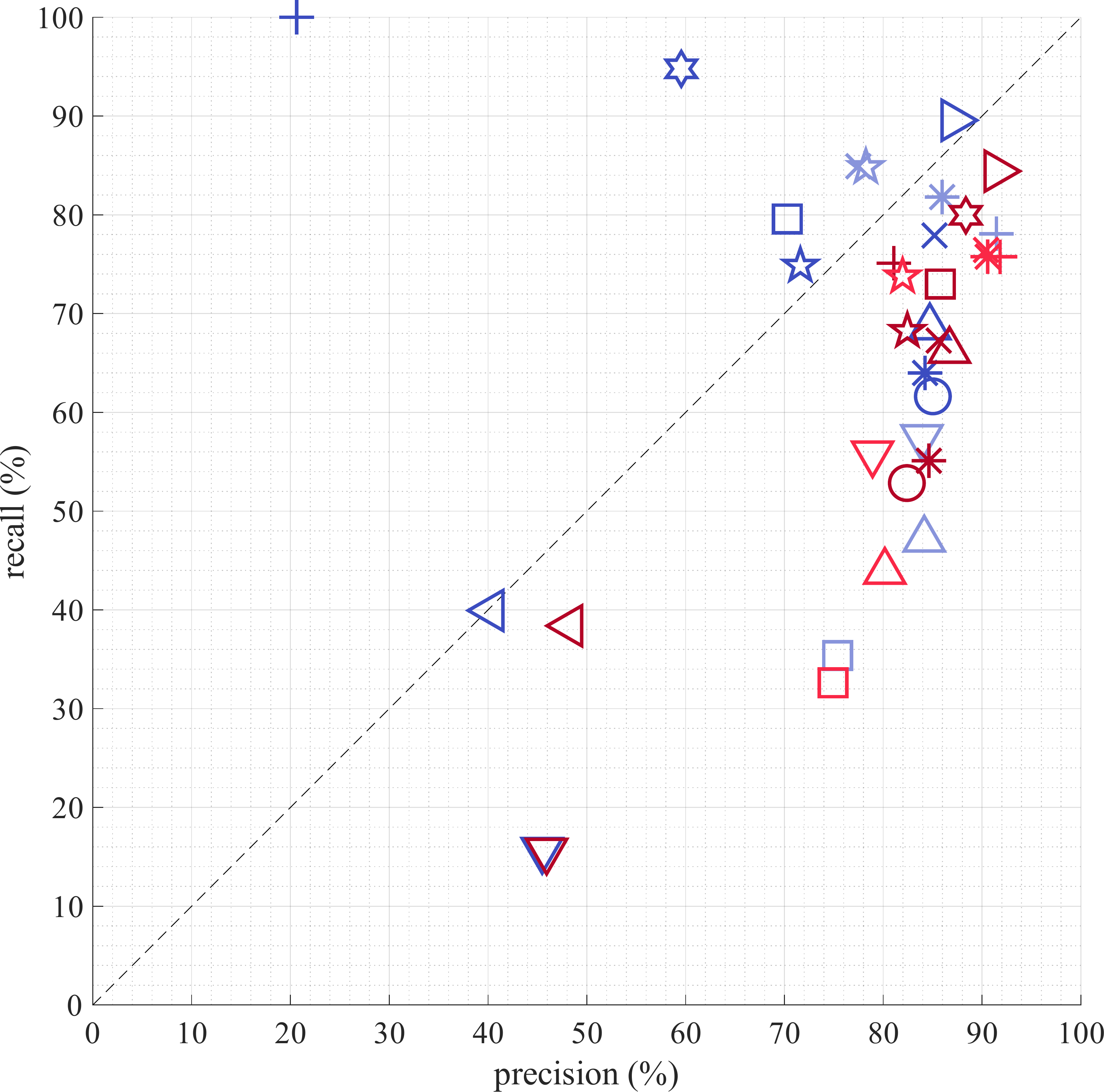}
			\\
			\includegraphics[height=\textwidth]{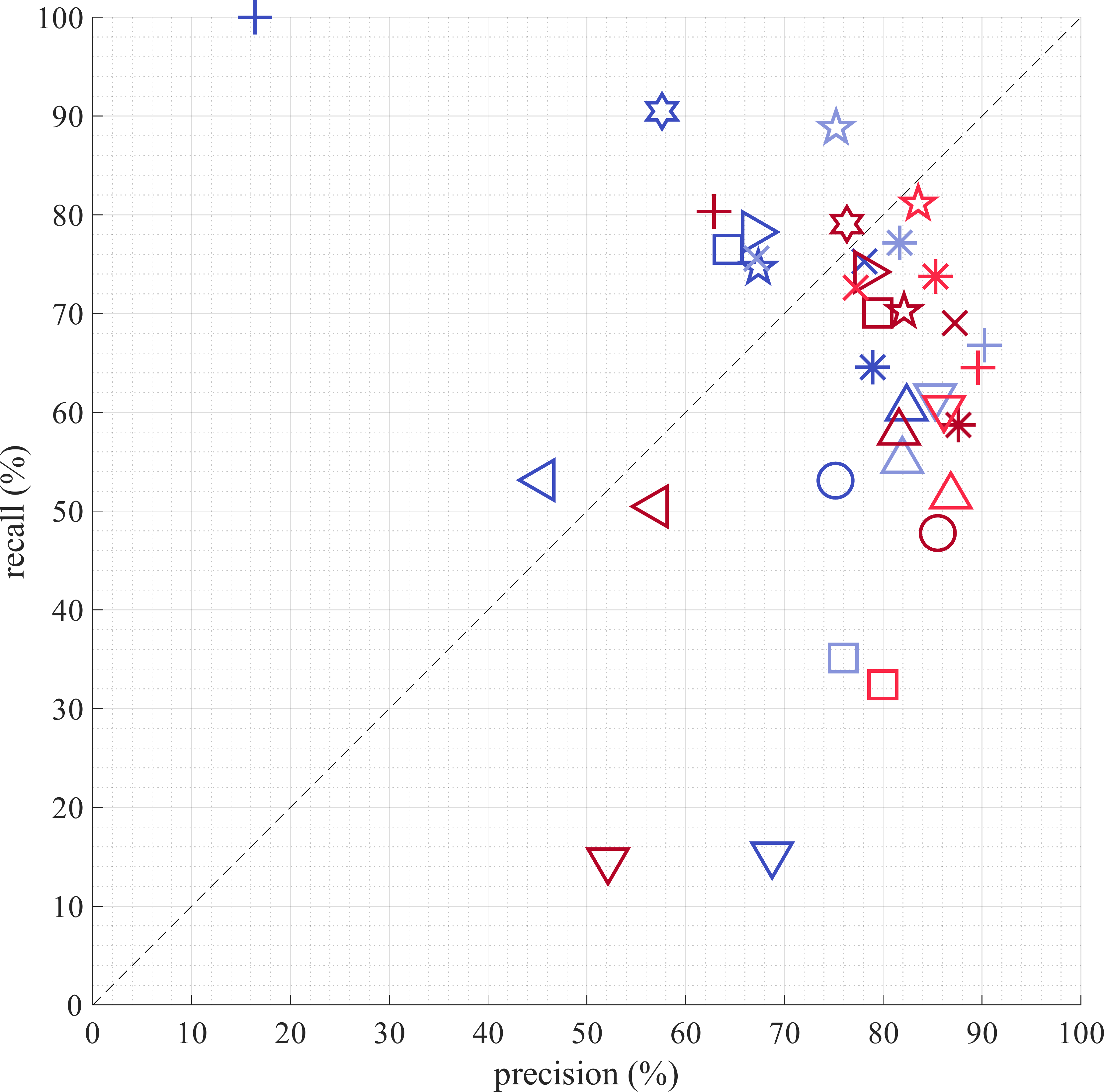}
		\end{minipage}
	}
	\subfloat[best configuration]{\label{plot_hz_sos}
		\begin{minipage}[c]{0.21\textwidth}
			\includegraphics[height=\textwidth]{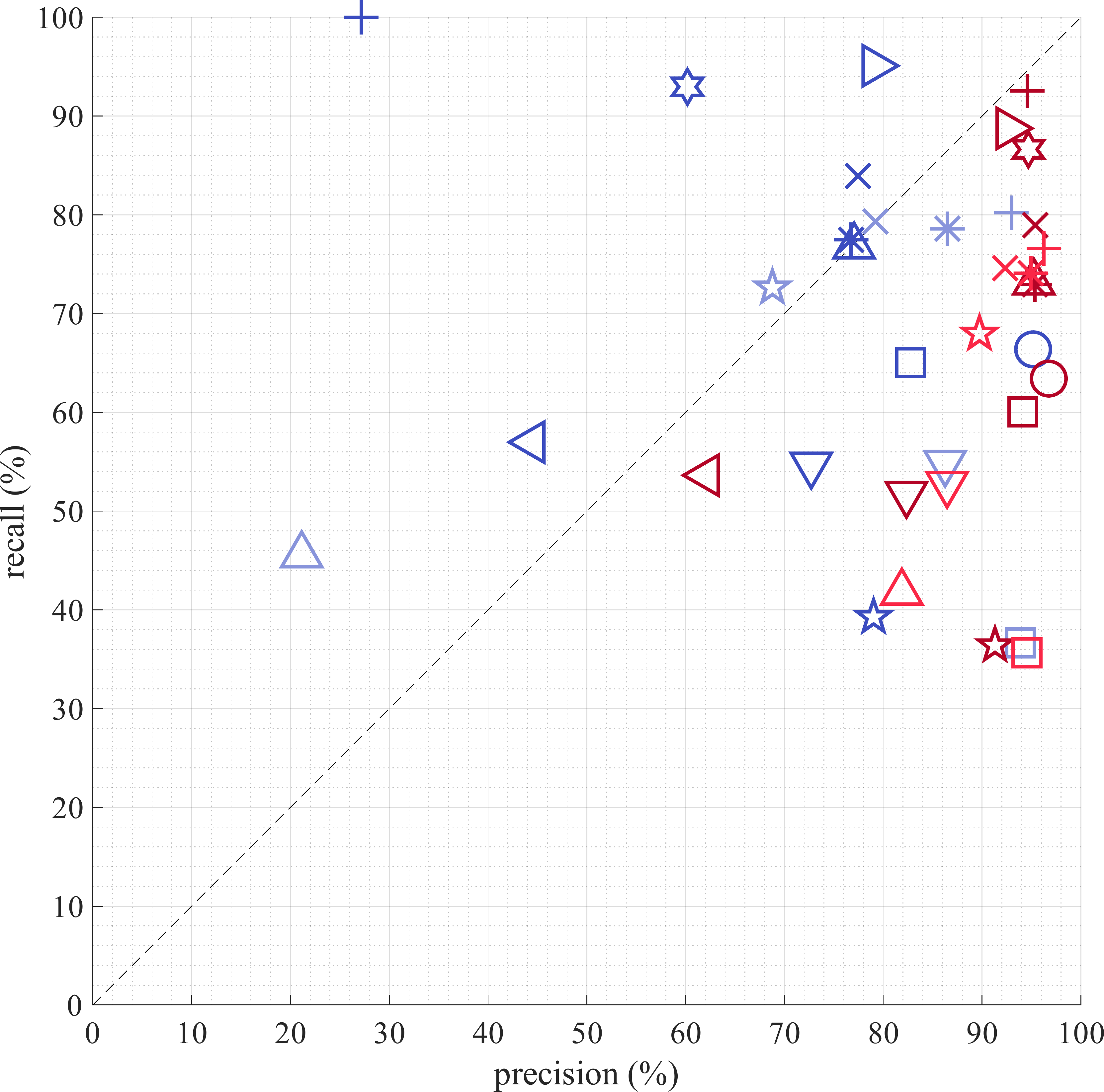}
			\\
			\includegraphics[height=\textwidth]{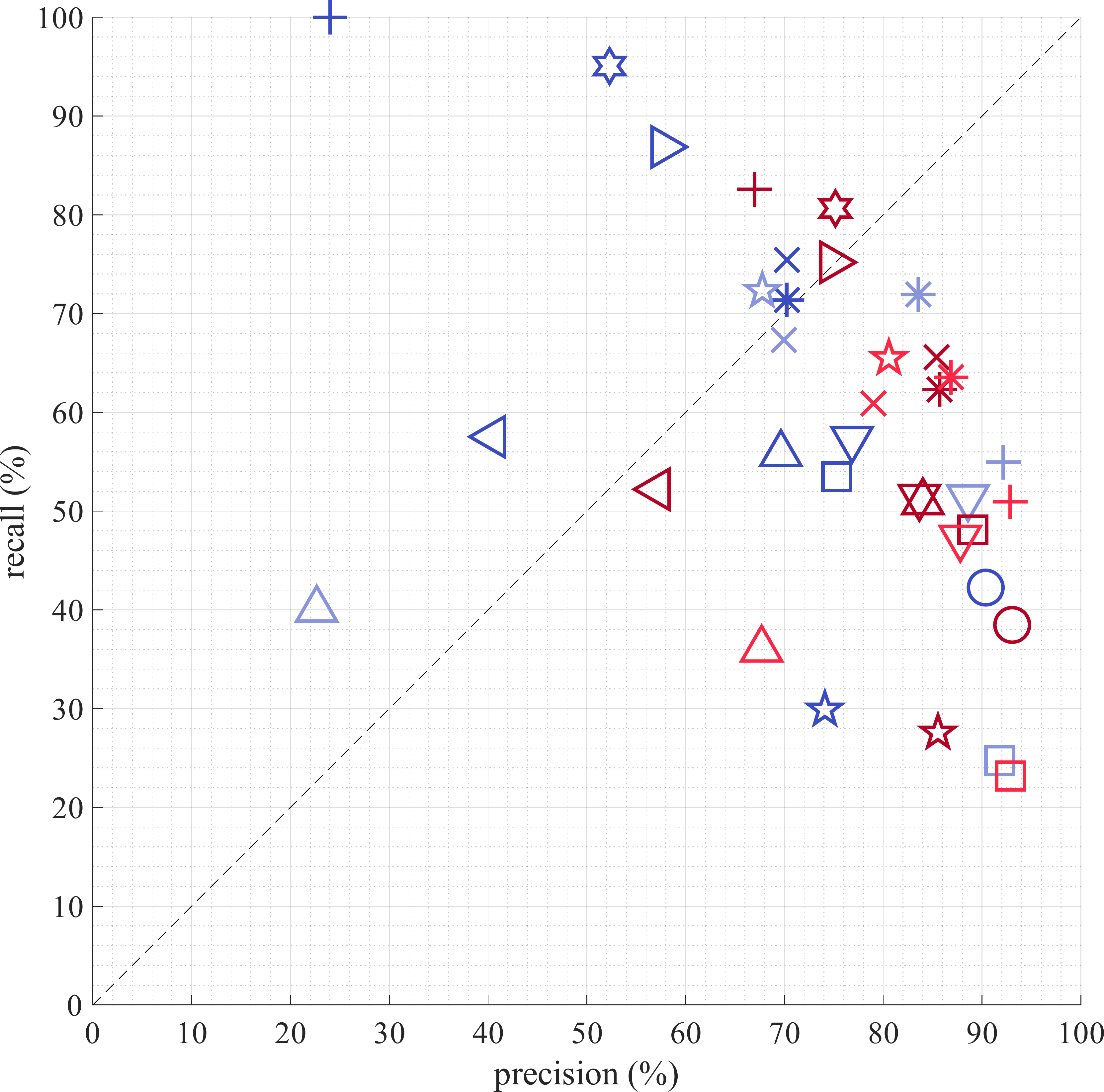}
		\end{minipage}
	}
	\hspace{0.5em}	
	\rotatebox[origin=c]{90}{\hfil\hspace{1.5em} SIFT+SOSNet\hspace{4em}HarrisZ+SOSNet\hfil}
	\subfloat[SUN3D]{\label{plot_sun3d_}
		\begin{minipage}[c]{0.21\textwidth}
			\includegraphics[height=\textwidth]{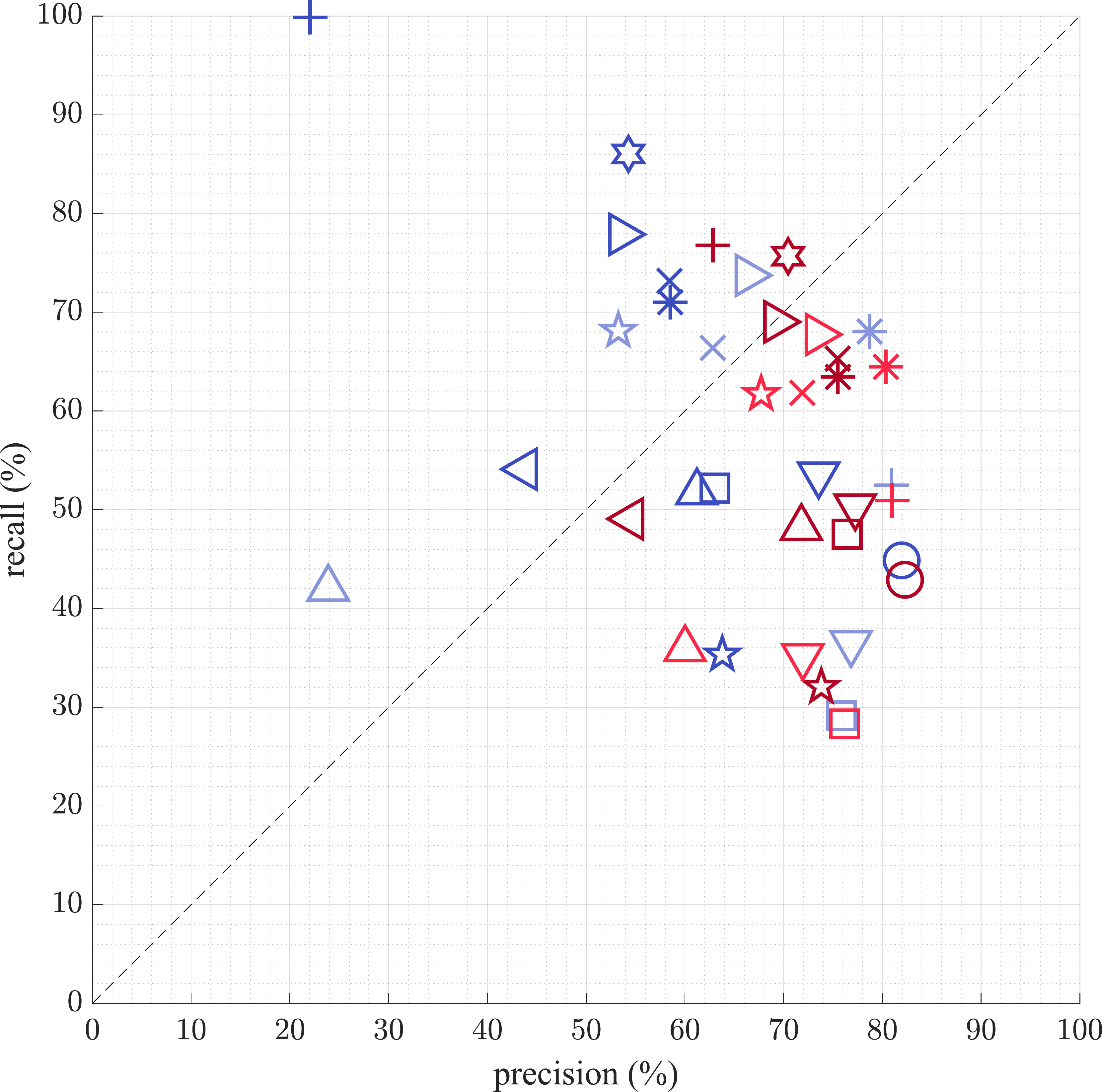}
			\\
			\includegraphics[height=\textwidth]{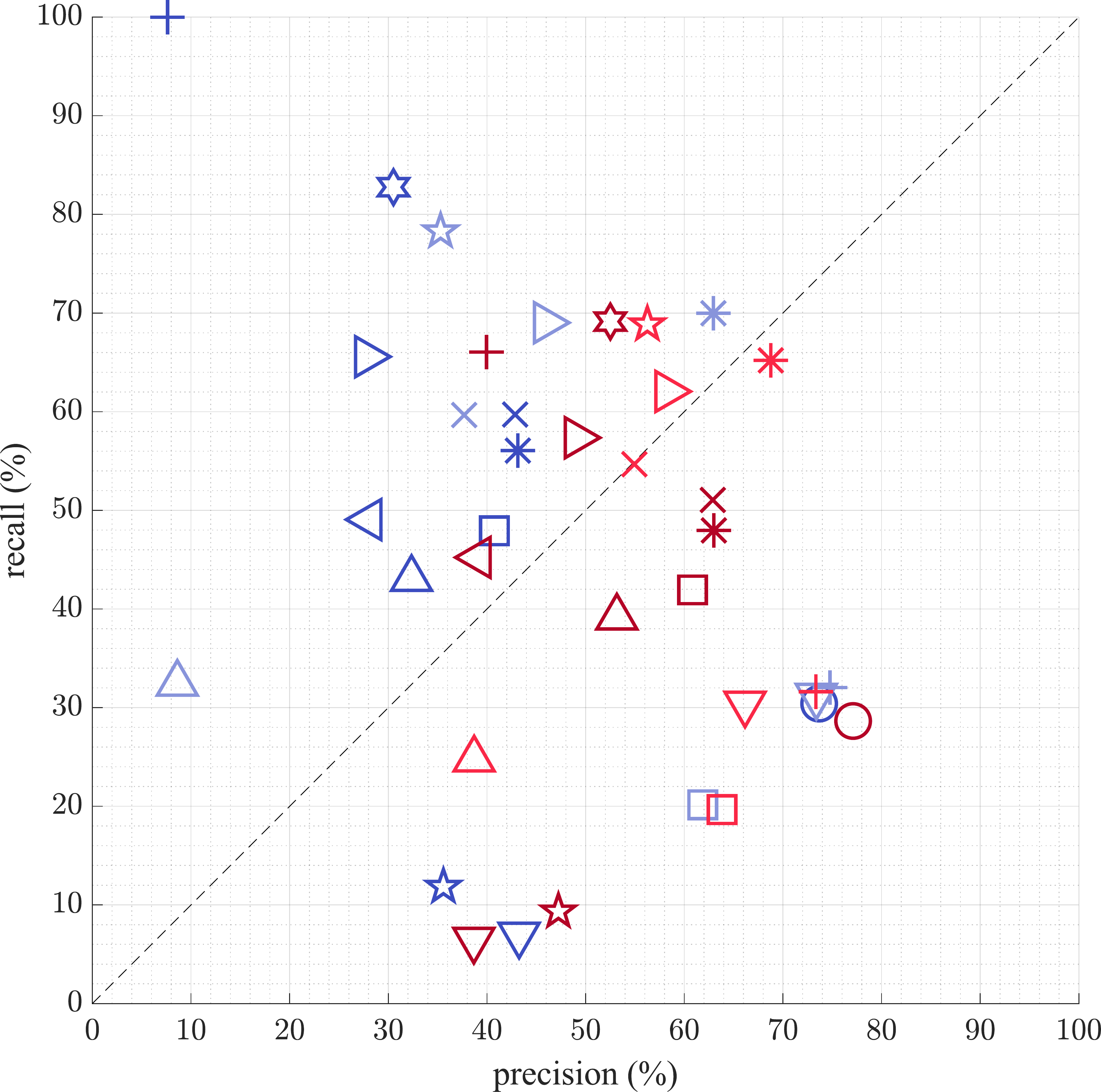}
		\end{minipage}
	}	
	\caption{\label{pr_plot}
		Average precision and recall values of the local spatial filters on the planar, non-planar and SUN3D datasets. The recall is computed with respect to ground truth matches obtained by blob matching, results without/with 1SAC are in blue/red. Average running times (s) for the baseline/best configuration are reported alongside in the legend (see text for details, best viewed in color and zoomed in).}
	\vspace{-1em}	
\end{figure*}

The scatter plots of Fig.~\ref{pr_plot} refer to the average precision and recall values of the evaluated local spatial filters, without or with 1SAC as post-processing, on the planar, non-planar and SUN3D datasets. For the planar and non-planar datasets, results are reported by considering: The global behavior (Fig.~\ref{plot_all}), i.e. averaging the results over all the considered detectors, descriptors and blob matching setups; the baseline configuration as reference (Fig.~\ref{plot_sift_rsift}), i.e. SIFT+RootSIFT with the one-to-one NNR greedy matching obtained by setting $f=\Omega$, $f'=1$, $t_o=\infty$, $\mathcal{D}_\geq$, $\mathcal{W}=a$; the best configuration in terms of mAP (Fig.~\ref{plot_hz_sos}), i.e. HarrisZ+SOSNet with blob matching setup $f=10_\cup$, $f'=5$, $t_o=0.75$, $\mathcal{D}^+$, $\mathcal{W}=2ab/(a+b)$. For the SUN3D dataset (Fig.~\ref{plot_sun3d_}), results are reported for the best configuration, also replacing HarrisZ with SIFT, and ACNe$^\star$ indicates that ACNE is trained with the SUN3D indoor dataset instead of the YFCC100M outdoor dataset~\cite{oanet}. For the planar and non-planar datasets, detailed average statistics including the precision and recall, the number of correct and output matches, the number of times a method failed, and the running time are reported in Appendix~\ref{appendix_dtm}. The recall is computed by considering only ground-truth matches from the specific blob matching setup used in the pipeline. No precision/recall aggregated measures, such as mAP or F$_\beta$ score are considered in the evaluation. On one hand, mAP requires that the number of output matches should be approximately the same for all methods since it is very sensitive to the recall, otherwise the highest scores would be assigned to the methods providing more output matches, including the initial blob matching. On the other hand, the choice of the $\beta$ parameter in F$_\beta$ can be questionable, as well as the choice of the recall normalization factor (see again Appendix~\ref{appendix_dtm} for further details), promoting one method or another without reflecting their effective performances. Nevertheless, the mAP and the F$_1$ and F$_{0.5}$ scores are reported for completeness in the additional material.

The average number of ground-truth matches per image pair, limited to blob matching only, for the planar/non-planar dataset are 551/429, 367/257, 681/575 for the global, reference and best configurations, and 963/691, 997/696, 928/686 when considering all the possible matches. According to the number of correct retrieved matches, the planar case is easier than the non-planar case. For SUN3D, the average number of ground-truth matches per image pair, limited to blob matching, is 481 for the best configuration and 229 when replacing HarrisZ with SIFT. 
Given the best configuration HarrisZ+SOSNet, the relative distribution of the spatial filters over the plots for the non-planar and SUN3D datasets is quite similar. By inspecting Fig.~\ref{pr_plot}, all methods improve the precision with respect to the blob matching, that obviously achieve the highest recall.

Focusing on the results without 1SAC, the full-stage DTM (DTM$_2$) provides high levels of precision and recall, comparable with OANet, ACNe and PGM. OANet obtains somewhat better precision than the other match filters. Notice also that ACNe trained on the same kind of images to be processed (ACNe$^\star$ on SUN3D) achieves boosted performances, being or not the training set included (see Sec.~\ref{eval_setup} and the additional material). The precision of the first stage of DTM alone (DTM$_1$) is similar to that of the complete DTM, but the recall is lower, underlining the goodness of the DTM$_2$ stage. LPM and LMR obtain recall values equivalent to those of the previously mentioned spatial filters, but with lower precision in the baseline configuration, while they regain in terms of precision at expense of the recall in the best configuration. GPLM behaves likewise LPM but with better recall in the best configuration. AdaLAM achieves very high precision but lower recall with respect to the previous spatial filters, unless running on planar scenes. Simple thresholding (th) gets precision values similar to AdaLAM but with lower recall. BM recall and precision are in-between those of AdaLAM and simple thresholding, while PFM obtains a lower recall than the simple thesholding for a similar precision. GMS improves upon the simple thresholding in the case of the best configuration, which is probably closer to the original setup GMS was designed for in terms of the number of input matches and keypoint type. VFC gets very high recall and a reasonable precision on the planar dataset, while for the other datasets it achieves still very high recall but low precision. RFM-SCAN is generally equal or worse than VFC. Finally, LLT and SVC obtained in this evaluation the worst results.

1SAC, without being a full RANSAC, is able to improve the precision with an acceptable loss in terms of recall. This effect decreases when simultaneously the precision is high and the recall is low, such as for AdaLAM. It is reasonable that 1SAC post-process does not affect AdaLAM, which consists of multiple local RANSACs. Moreover, 1SAC generally reduces the gain in terms of precision of OANet with respect to DTM, PGM and ACNe. Notice that OANet and ACNe network design and training takes more or less explicitly into account the same global model constraints based on the epipolar geometry of 1SAC, not considered in the design of the other spatial filters. Blob matching after 1SAC post-filtering achieves almost the best results in terms of both precision and recall in the planar case. This holds because planar images are relatively easy, homographies provide one-to-one point maps between images, and 1SAC re-filters the output of the local spatial filters. These observations indicate that homographies are correctly estimated in any case, so that the highest recall is obtained as more input matches are provided. With 1SAC, going from the baseline to the best configuration, a general expansion towards the top-right area of scatter plots can be observed. This agrees with the fact that the absolute number of correct matches gets roughly doubled while the total number of output matches becomes five times ($f'$ changes from 1 to 5) as moving from the baseline to the best configuration, so that top-ranked matches employed by 1SAC becomes less contaminated by outliers\footnote{Note that in general the average precision is not equal to the ratio between the average number of correct matches and the average number of output matches.}.   

Blob matching precision of each plot gives also an indication of the average inlier ratio, so that SIFT+SOSNet on SUN3D (about 8\% of average inlier ratio according to blob matching) is more contaminated by outliers than the other presented plots. On this setup, OANet followed by ACNe$^\star$ suffer less the outlier contamination. Nevertheless, their differences with the other spatial filters after 1SAC are quite reduced since they implicitly contains epipolar constraints, unlike other methods. Under this consideration DTM neighborhoods based only on triangulation without considering any form of consistency based on patch relative local transformations are quite robust. More detailed histograms supporting these observations according to the inlier ratio can be found in Appendix~\ref{appendix_dtm}.   

The number of times the local spatial filters failed to get at least one correct match as output, reported on Appendix~\ref{appendix_dtm}, can provide further clues about the robustness of each match filter. For both planar or non-planar scenes, excluding obviously the blob matching, DTM, OANet, ACNe and PGM are among those methods which fail less. Notice also that when 1SAC is applied more failures arise, due to the reduced number of the candidate matches.

A visual qualitative analysis on the best configuration according to the examples reported in Appendix~\ref{appendix_dtm} agrees with the quantitative results discussed above. It can be noted that for DTM, with respect to other spatial filters such as ACNe and PGM, the wrong matches often concern the image regions near the triangulation boundary, which lacks a neighborhood covering in all directions. These wrong matches are generally removed by 1SAC which, unlike the case of the other spatial filters, seems to remove less inliers in the case of DTM, maybe due to the kind of outliers.

Average running times are reported in Fig.~\ref{pr_plot} alongside the legend for the baseline/best configurations, implying respectively one-to-one or many-to-many matching relations. Results have been obtained with Ubuntu 20.04 running on an Intel Core I9 10900K with 64 GB of RAM equipped with a NVIDIA GeForce RTX 2080 Ti GPU. The original code was used for each implementation, only modified to works with blob matches as input. The code is implemented in Python or Matlab with different level of optimization, ranging to Matlab mex C functions to GPU parallel optimization. From the baseline to the best configuration the running time increases proportionally to the number of processed matches by at least a linear factor, theoretically expected to be around $\sum_{i=1}^{f'}\frac{i}{f'}=3$ since $f'=5'$ for the many-to-many match in the best configuration. The code of PGM, BM and LLT is the slowest, since implemented in Matlab with almost no optimization. DTM and GMS, respectively written in Matlab and Python without any type of optimization, follow in the list, and the faster code of the remaining spatial filters contains several optimizations. Better running times are expected for DTM after code optimization since the various steps of DTM can be highly parallelized and, in the same manner of LPM, portions of DTM code would benefit of mex C code rewriting\footnote{According to the author's experience Matlab is not the best environment for graph manipulations.}.

\vspace{-1em}
\section{Conclusions and Future Work}\label{conclusion}
This paper analyzes the problem of the local image descriptor matching according to two possible contexts that can be used to characterize the images, and to improve the number of correct correspondences.

The first context is provided by the descriptor space. The novel general blob matching strategy is designed to incorporate different approaches for a clear and detailed analysis of the aspects that characterize the basic matching strategies. According to the evaluation, pre-filtering, many-to-many matches, two-way comparisons using symmetric distances and a good choice of the second best match in NNR can improve the matching process.

The second context is provided by the keypoint space, i.e. the actual image space. A new local spatial filter named DTM is proposed. DTM extracts spatial neighborhood relations between keypoints by Delaunay triangulation, alternating triangulation contractions and expansions to remove inconsistent matches and to include consistent matches. DTM is robust and obtains comparable or better results than the state-of-the-art. Furthermore, DTM neighborhoods do not rely on parameters to be defined, but are implicitly derived by the keypoint distribution onto the images. Moreover, DTM does not require patch relative local transformations for validate the neighborhood consistency. 

Although blob matching and DTM mainly operate respectively on the descriptor and space contexts, they both betray contaminations from the opposite contexts, underlining the need of fully integrating different contexts to go beyond the current state-of-the-art. To be noted that blob matching and DTM have been employed recently as part of a very competitive matching pipeline which achieved among the best results in the recent Image Matching Challenge 2021 (IMC2021) and SimLocMatch contests~\cite{imw2021}.   

Finally, a comprehensive evaluation of the main phases of the matching pipeline is carried out, based on a new benchmark, focusing on the estimation of correct matches and not on their effects on the scene. The analysis considers both blob and corner like keypoints, among the current best local image descriptors, several image matching strategies, state-of-the-art local spatial filters, and also the simple model-based filter. It clearly emerges that the combination of the different methods can offer a clear advantage with respect to the baseline SIFT matching strategy.

As future work, further possibilities for merging matching strategies will be investigated, as well as mesh-based applications of the triangulation in order to grow up matches and obtain semi-dense correspondences. Additionally, it would be interesting to analyze how triangulation-based neighborhoods can be used for clustering and spatially characterizing the objects in the scene. Further research directions will be also aimed at improving the benchmark, by extending the datasets with more scenes and by designing better error metrics to compare the different approaches.

\section*{Acknowledgment}
The Titan Xp used for this research was generously donated by the NVIDIA Corporation.

This research is funded by the Italian Ministry of Education and Research (MIUR) under the program PON Ricerca e Innovazione 2014-2020, cofunded by the European Social Fund (ESF), CUP B74I18000220006, id. proposta AIM 1875400, linea di attivit\`{a} 2, Area Cultural Heritage.

\bibliographystyle{IEEEtran}
\bibliography{paper.bib}

\begin{thebibliography}{10}
\providecommand{\url}[1]{#1}
\csname url@samestyle\endcsname
\providecommand{\newblock}{\relax}
\providecommand{\bibinfo}[2]{#2}
\providecommand{\BIBentrySTDinterwordspacing}{\spaceskip=0pt\relax}
\providecommand{\BIBentryALTinterwordstretchfactor}{4}
\providecommand{\BIBentryALTinterwordspacing}{\spaceskip=\fontdimen2\font plus
\BIBentryALTinterwordstretchfactor\fontdimen3\font minus
  \fontdimen4\font\relax}
\providecommand{\BIBforeignlanguage}[2]{{%
\expandafter\ifx\csname l@#1\endcsname\relax
\typeout{** WARNING: IEEEtran.bst: No hyphenation pattern has been}%
\typeout{** loaded for the language `#1'. Using the pattern for}%
\typeout{** the default language instead.}%
\else
\language=\csname l@#1\endcsname
\fi
#2}}
\providecommand{\BIBdecl}{\relax}
\BIBdecl

\bibitem{szeliski_book}
R.~Szeliski, \emph{Computer Vision: Algorithms and Applications, 2nd
  edition}.\hskip 1em plus 0.5em minus 0.4em\relax Springer, 2021.

\bibitem{sift}
D.~Lowe, ``Distinctive image features from scale-invariant keypoints,''
  \emph{International Journal of Computer Vision}, vol.~60, no.~2, pp. 91--110,
  2004.

\bibitem{sift_matching}
F.~Bellavia and C.~Colombo, ``Is there anything new to say about {SIFT}
  matching?'' \emph{International Journal of Computer Vision}, vol. 128, pp.
  1847--1866, 2020.

\bibitem{imw2020}
Y.~Jin, D.~Mishkin, A.~Mishchuk, J.~Matas, P.~Fua, K.~M. Yi, and E.~Trulls,
  ``Image matching across wide baselines: From paper to practice,''
  \emph{International Journal of Computer Vision}, vol. 129, pp. 517--547,
  2021.

\bibitem{root_sift}
R.~Arandjelovi\'c and A.~Zisserman, ``Three things everyone should know to
  improve object retrieval,'' in \emph{Proceedings of the {IEEE} Conference on
  Computer Vision and Pattern Recognition ({CVPR})}, 2012, pp. 2911--2918.

\bibitem{sgloh2_pami}
F.~Bellavia and C.~Colombo, ``Rethinking the {sGLOH} descriptor,'' \emph{{IEEE}
  Transactions on Pattern Analysis and Machine Intelligence}, vol.~40, no.~4,
  pp. 931--944, 2018.

\bibitem{wisw}
\BIBentryALTinterwordspacing
------, ``{`Which is Which?'} {E}valuation of local descriptors for image
  matching in real-world scenarios,'' in \emph{International Conference on
  Computer Analysis of Images and Patterns ({CAIP})}, 2019. [Online].
  Available: \url{http://cvg.dsi.unifi.it/wisw.caip2019}
\BIBentrySTDinterwordspacing

\bibitem{colmap}
J.~L. Sch\"{o}nberger and J.~M. Frahm, ``Structure-from-{M}otion revisited,''
  in \emph{Conference on Computer Vision and Pattern Recognition (CVPR)}, 2016.

\bibitem{contextdesc}
Z.~Luo, T.~Shen, L.~Zhou, J.~Zhang, Y.~Yao, S.~Li, T.~Fang, and L.~Quan,
  ``{ContextDesc}: Local descriptor augmentation with cross-modality context,''
  \emph{Computer Vision and Pattern Recognition (CVPR)}, 2019.

\bibitem{tfeat}
V.~Balntas, E.~Riba, D.~Ponsa, and K.~Mikolajczyk, ``Learning local feature
  descriptors with triplets and shallow convolutional neural networks,'' in
  \emph{Proceedings of the British Machine Vision Conference ({BMVC})}, 2016,
  pp. 119.1--119.11.

\bibitem{hardnet}
A.~Mishchuk, D.~Mishkin, F.~Radenovic, and J.~Matas, ``Working hard to know
  your neighbor's margins: Local descriptor learning loss,'' in \emph{Advances
  in Neural Information Processing Systems 30: Annual Conference on Neural
  Information Processing Systems ({NIPS})}, 2017, pp. 4829--4840.

\bibitem{geodesc}
Z.~Luo, T.~Shen, L.~Zhou, S.~Zhu, R.~Zhang, Y.~Yao, T.~Fang, and L.~Quan,
  ``Geodesc: Learning local descriptors by integrating geometry constraints,''
  in \emph{Proceedings of the European Conference on Computer Vision ({ECCV})},
  2018.

\bibitem{sosnet}
Y.~Tian, X.~Yu, B.~Fan, F.~Wu, H.~Heijnen, and V.~Balntas, ``{SOSNet}: Second
  order similarity regularization for local descriptor learning,'' in
  \emph{CVPR}, 2019.

\bibitem{sun3d}
J.~Xiao, A.~Owens, and A.~Torralba, ``{SUN3D}: A database of big spaces
  reconstructed using sfm and object labels,'' in \emph{Proceedings of the
  International Conference on Computer Vision ({ICCV})}, 2013, pp. 1625--1632.

\bibitem{hpatches}
V.~Balntas, K.~Lenc, A.~Vedaldi, and K.~Mikolajczyk, ``{HPatches:} {A}
  benchmark and evaluation of handcrafted and learned local descriptors,'' in
  \emph{{IEEE} Conference on Computer Vision and Pattern Recognition ({CVPR})},
  2017, pp. 3852--3861.

\bibitem{megadepth}
Z.~Li and N.~Snavely, ``{MegaDepth}: Learning single-view depth prediction from
  internet photos,'' in \emph{Computer Vision and Pattern Recognition (CVPR)},
  2018.

\bibitem{spatial_filter_eval}
C.~{Zhao}, Z.~{Cao}, J.~{Yang}, K.~{Xian}, and X.~{Li}, ``Image feature
  correspondence selection: A comparative study and a new contribution,''
  \emph{IEEE Transactions on Image Processing}, vol.~29, pp. 3506--3519, 2020.

\bibitem{matchers_survey}
J.~Ma, J.~Jiang, A.~Fan, J.~Jiang, and J.~Yan, ``Image matching from
  handcrafted to deep features: A survey,'' \emph{International Journal of
  Computer Vision}, vol. 129, pp. 23--79, 2021.

\bibitem{multiview}
R.~I. Hartley and A.~Zisserman, \emph{Multiple View Geometry in Computer
  Vision}.\hskip 1em plus 0.5em minus 0.4em\relax Cambridge University Press,
  2000.

\bibitem{ransac}
M.~Fischler and R.~Bolles, ``Random sample consensus: A paradigm for model
  fitting with applications to image analysis and automated cartography,''
  \emph{Communications of the ACM}, vol.~24, no.~6, pp. 381--395, 1981.

\bibitem{prosac}
O.~{Chum} and J.~{Matas}, ``Matching with {PROSAC} - progressive sample
  consensus,'' in \emph{Proceedings of the IEEE Conference on Computer Vision
  and Pattern Recognition ({CVPR})}, 2005, pp. 220--226.

\bibitem{scramsac}
T.~Sattler, B.~Leibe, and L.~Kobbelt, ``{SCRAMSAC}: Improving {RANSAC}'s
  efficiency with a spatial consistency filter,'' in \emph{Proceedings of the
  IEEE International Conference on Computer Vision ({ICPR})}, 2009, pp.
  2090--2097.

\bibitem{groupsac}
N.~Ni, J.~Hailin, and F.~Dellaert, ``{GroupSAC}: Efficient consensus in the
  presence of groupings,'' in \emph{Proceedings of the {IEEE} International
  Conference on Computer Vision ({ICCV})}, 2009, pp. 2193--2200.

\bibitem{gc_ransac}
D.~{Barath} and J.~{Matas}, ``{Graph-Cut RANSAC},'' in \emph{Proceedings of the
  IEEE Conference on Computer Vision and Pattern Recognition ({CVPR})}, 2018,
  pp. 6733--6741.

\bibitem{adalam}
L.~Cavalli, V.~Larsson, M.~R. Oswald, T.~Sattler, and M.~Pollefeys,
  ``Handcrafted outlier detection revisited,'' in \emph{Proceedings of the
  European Conference on Computer Vision ({ECCV})}, 2020, pp. 770--787.

\bibitem{lpm}
J.~Ma, J.~Zhao, J.~Jiang, H.~Zhou, and X.~Guo, ``Locality preserving
  matching,'' \emph{International Journal of Computer Vision}, vol. 127, no.~5,
  pp. 512--531, May 2019.

\bibitem{superpoint}
D.~DeTone, T.~Malisiewicz, and A.~Rabinovich, ``Superpoint: Self-supervised
  interest point detection and description,'' in \emph{The IEEE Conference on
  Computer Vision and Pattern Recognition (CVPR) Workshops}, June 2018.

\bibitem{disk}
M.~J. Tyszkiewicz, P.~Fua, and E.~Trulls, ``{DISK}: Learning local features
  with policy gradient,'' in \emph{Proceedings of the 32nd Conference on Neural
  Information Processing Systems ({NeurIPS})}, 2020.

\bibitem{superglue}
P.~E. Sarlin, D.~DeTone, T.~Malisiewicz, and A.~Rabinovich, ``{S}uper{G}lue:
  Learning feature matching with graph neural networks,'' in \emph{arXiv},
  2019.

\bibitem{d2d}
Y.~Tian, V.~Balntas, T.~Ng, A.~B. Laguna, Y.~Demiris, and K.~Mikolajczyk,
  ``{D2D:} keypoint extraction with describe to detect approach,'' in
  \emph{Proceedings of the 15th Asian Conference on Computer Vision ({ACCV})},
  2020.

\bibitem{point_net}
K.~M. {Yi}, E.~{Trulls}, Y.~{Ono}, V.~{Lepetit}, M.~{Salzmann}, and P.~{Fua},
  ``Learning to find good correspondences,'' in \emph{Proceedings of the IEEE
  Conference on Computer Vision and Pattern Recognition ({CVPR})}, 2018.

\bibitem{nm_net}
C.~Zhao, Z.~Cao, C.~Li, X.~Li, and J.~Yang, ``{NM-Net}: Mining reliable
  neighbors for robust feature correspondences,'' in \emph{Proceedings of the
  {IEEE} Conference on Computer Vision and Pattern Recognition ({CVPR})}, 2019,
  pp. 215--224.

\bibitem{oanet}
J.~{Zhang}, D.~{Sun}, Z.~{Luo}, A.~{Yao}, L.~{Zhou}, T.~{Shen}, Y.~{Chen},
  H.~{Liao}, and L.~{Quan}, ``Learning two-view correspondences and geometry
  using order-aware network,'' in \emph{Proceedings of the IEEE International
  Conference on Computer Vision ({ICCV})}, 2019, pp. 5844--5853.

\bibitem{acne}
W.~Sun, W.~Jiang, E.~Trulls, A.~Tagliasacchi, and K.~M. Yi, ``{ACNe}: Attentive
  context normalization for robust permutation-equivariant learning,'' in
  \emph{Proceedings of the {IEEE} Conference on Computer Vision and Pattern
  Recognition ({CVPR})}, 2020.

\bibitem{fund_mat_eval}
J.~W. Bian, Y.~H. Wu, J.~Zhao, Y.~Liu, L.~Zhang, M.~M. Cheng, and I.~Reid, ``An
  evaluation of feature matchers for fundamental matrix estimation,'' in
  \emph{Proceedings of the British Machine Vision Conference ({BMVC})}, 2019.

\bibitem{keynet}
A.~Barroso-Laguna, E.~Riba, D.~Ponsa, and K.~Mikolajczyk, ``{Key.Net}: Keypoint
  detection by handcrafted and learned {CNN} filters,'' in \emph{Proceedings of
  the International Conference on Computer Vision ({ICCV})}, 2019.

\bibitem{harrisz}
F.~Bellavia, D.~Tegolo, and C.~Valenti, ``Improving {H}arris corner selection
  strategy,'' \emph{{IET} Computer Vision}, vol.~5, no.~2, pp. 86--96, 2011.

\bibitem{imw2021}
\BIBentryALTinterwordspacing
``{I}mage {M}atching {W}orkshop ({IMW}) challenge at ({CVPR2021}),'' 2021.
  [Online]. Available: \url{https://image-matching-workshop.github.io}
\BIBentrySTDinterwordspacing

\bibitem{rootsgloh2}
F.~Bellavia and C.~Colombo, ``{RootsGLOH2}: embedding {RootSIFT} ``square
  rooting'' in {sGLOH2},'' \emph{IET Computer Vision}, 2020.

\bibitem{hardnetamos}
M.~Pultar, D.~Mishkin, and J.~Matas, ``Leveraging outdoor webcams for local
  descriptor learning,'' in \emph{Proceedings of Computer Vision Winter
  Workshop ({CVWW}) 2019}, 2019.

\bibitem{orb}
E.~Rublee, V.~Rabaud, K.~Konolige, and G.~Bradski, ``{ORB}: an efficient
  alternative to {SIFT} or {SURF},'' in \emph{Proceedings of the {IEEE}
  International Conference on Computer Vision ({ICCV})}, 2011, pp. 2564--2571.

\bibitem{learning_ori}
K.~Yi, Y.~Verdie, P.~Fua, and V.~Lepetit, ``Learning to assign orientations to
  feature points,'' in \emph{Proceedings of the {IEEE} Conference on Computer
  Vision and Pattern Recognition ({CVPR})}, 2016, pp. 1--8.

\bibitem{affnet}
D.~Mishkin, F.~Radenovic, and J.~Matas, ``Repeatability is not enough: Learning
  affine regions via discriminability,'' in \emph{Proceedings of the European
  Conference on Computer Vision ({ECCV})}, 2018.

\bibitem{mrogh}
B.~Fan, F.~Wu, and Z.~Hu, ``Rotationally invariant descriptors using intensity
  order pooling,'' \emph{{IEEE} Transactions on Pattern Analysis and Machine
  Intelligence}, vol.~34, no.~10, pp. 2031--2045, 2012.

\bibitem{fginn}
K.~Lenc, J.~Matas, and D.~Mishkin, ``A few things one should know about feature
  extraction, description and matching,'' in \emph{Proceedings of the Computer
  Vision Winter Workshop ({CVWW})}, 2014, pp. 67--74.

\bibitem{generalized_ransac}
W.~{Zhang} and J.~{Kosecka}, ``Generalized {RANSAC} framework for relaxed
  correspondence problems,'' in \emph{Proceeding of the International Symposium
  on {3D} Data Processing, Visualization, and Transmission ({3DPVT})}, 2006,
  pp. 854--860.

\bibitem{mods}
D.~Mishkin, J.~Matas, and M.~Perdoch, ``{MODS}: Fast and robust method for
  two-view matching,'' \emph{Computer Vision and Image Understanding}, 2015.

\bibitem{progx}
D.~Barath and J.~Matas, ``{Progressive-X}: Efficient, anytime, multi-model
  fitting algorithm,'' in \emph{Proceedings of the IEEE International
  Conference on Computer Vision ({ICCV})}, 2019.

\bibitem{topological_filter}
V.~{Ferrari}, T.~{Tuytelaars}, and {Luc Val Gool}, ``Wide-baseline
  multiple-view correspondences,'' in \emph{Proceedings of the {IEEE}
  Conference on Computer Vision and Pattern Recognition ({CVPR})}, 2003.

\bibitem{gms}
J.~W. Bian, W.~Y. Lin, Y.~Liu, L.~Zhang, S.~K. Yeung, M.~M. Cheng, and I.~Reid,
  ``{GMS}: Grid-based motion statistics for fast, ultra-robust feature
  correspondence,'' \emph{International Journal of Computer Vision}, vol. 128,
  pp. 1580--1593, 2020.

\bibitem{glpm}
J.~{Ma}, J.~{Jiang}, H.~{Zhou}, J.~{Zhao}, and X.~{Guo}, ``Guided locality
  preserving feature matching for remote sensing image registration,''
  \emph{IEEE Transactions on Geoscience and Remote Sensing}, vol.~56, no.~8,
  pp. 4435--4447, 2018.

\bibitem{lmr}
J.~{Ma}, X.~{Jiang}, J.~{Jiang}, J.~{Zhao}, and X.~{Guo}, ``{LMR}: Learning a
  two-class classifier for mismatch removal,'' \emph{IEEE Transactions on Image
  Processing}, vol.~28, no.~8, pp. 4045--4059, 2019.

\bibitem{pfm}
S.~Lee, J.~Lim, and I.~H. Suh, ``Progressive feature matching: Incremental
  graph construction and optimization,'' \emph{{IEEE} Transactions on Image
  Processing}, vol.~29, pp. 6992--7005, 2020.

\bibitem{pgm}
M.~Cho and K.~M. Lee, ``Progressive graph matching: Making a move of graphs via
  probabilistic voting,'' in \emph{Proceedings of the IEEE Conference on
  Computer Vision and Pattern Recognition ({CVPR})}, 2012, pp. 398--405.

\bibitem{viso2}
A.~{Geiger}, J.~{Ziegler}, and C.~{Stiller}, ``Stereoscan: Dense {3D}
  reconstruction in real-time,'' in \emph{Proceedings of the {IEEE} Intelligent
  Vehicles Symposium ({IV})}, 2011.

\bibitem{llt}
J.~{Ma}, H.~{Zhou}, J.~{Zhao}, Y.~{Gao}, J.~{Jiang}, and J.~{Tian}, ``Robust
  feature matching for remote sensing image registration via locally linear
  transforming,'' \emph{IEEE Transactions on Geoscience and Remote Sensing},
  vol.~53, no.~12, pp. 6469--6481, 2015.

\bibitem{vfc}
J.~Ma, J.~Zhao, J.~Tian, A.~L. Yuille, and Z.~Tu, ``Robust point matching via
  vector field consensus,'' \emph{IEEE Transactions on Image Processing},
  vol.~23, no.~4, pp. 1706--1721, 2014.

\bibitem{bm}
W.~Y.~D. Lin, M.~M. Cheng, J.~Lu, H.~Yang, M.~N. Do, and P.~Torr, ``Bilateral
  functions for global motion modeling,'' in \emph{Proceedings of the European
  Conference on Computer Vision ({ECCV})}, 2014, pp. 341--356.

\bibitem{scv}
J.~Cech, J.~Matas, and M.~Perdoch, ``Efficient sequential correspondence
  selection by cosegmentation,'' \emph{{IEEE} Transactions on Pattern Analysis
  and Machine Intelligence}, vol.~32, no.~9, pp. 1568--1581, 2010.

\bibitem{rfm_scan}
X.~{Jiang}, J.~{Ma}, J.~{Jiang}, and X.~{Guo}, ``Robust feature matching using
  spatial clustering with heavy outliers,'' \emph{IEEE Transactions on Image
  Processing}, vol.~29, pp. 736--746, 2020.

\bibitem{game_theoretic_journal}
A.~Albarelli, E.~Rodol\`{a}, and A.~Torsello, ``Imposing semi-local geometric
  constraints for accurate correspondences selection in structure from motion:
  A game-theoretic perspective,'' \emph{International Journal of Computer
  Vision}, vol.~92, pp. 36--53, 2012.

\bibitem{loftr}
J.~Sun, Z.~Shen, Y.~Wang, H.~Bao, and X.~Zhou, ``{LoFTR}: Detector-free local
  feature matching with transformers,'' 2021.

\bibitem{descriptor_eval}
K.~Mikolajczyk and C.~Schmid, ``A performance evaluation of local
  descriptors,'' \emph{{IEEE} Transactions on Pattern Analysis and Machine
  Intelligence}, vol.~27, no.~10, pp. 1615--1630, 2005.

\bibitem{noransac}
D.~Tegolo and F.~Bellavia, ``no{RANSAC} for fundamental matrix estimation,'' in
  \emph{Proceedings of the British Machine Vision Conference ({BMVC})}, 2011.

\bibitem{zhang_fun_mat}
Z.~Zhang, ``Determining the epipolar geometry and its uncertainty: A review,''
  \emph{International Journal of Computer Vision}, vol.~27, no.~2, pp.
  161--195, 1998.

\bibitem{perona}
P.~Moreels and P.~Perona, ``Evaluation of features detectors and descriptors
  based on 3d objects,'' \emph{International Journal of Computer Vision},
  vol.~73, no.~3, pp. 263--284, 2007.

\end{thebibliography}


\vfill
\clearpage
\newpage

\appendices
\section{DTM border computation}\label{appendix_border}

\begin{figure}[h!]
	\center
	\subfloat[]{\label{corner_mesh_img}
		\includegraphics[width=0.185\textwidth]{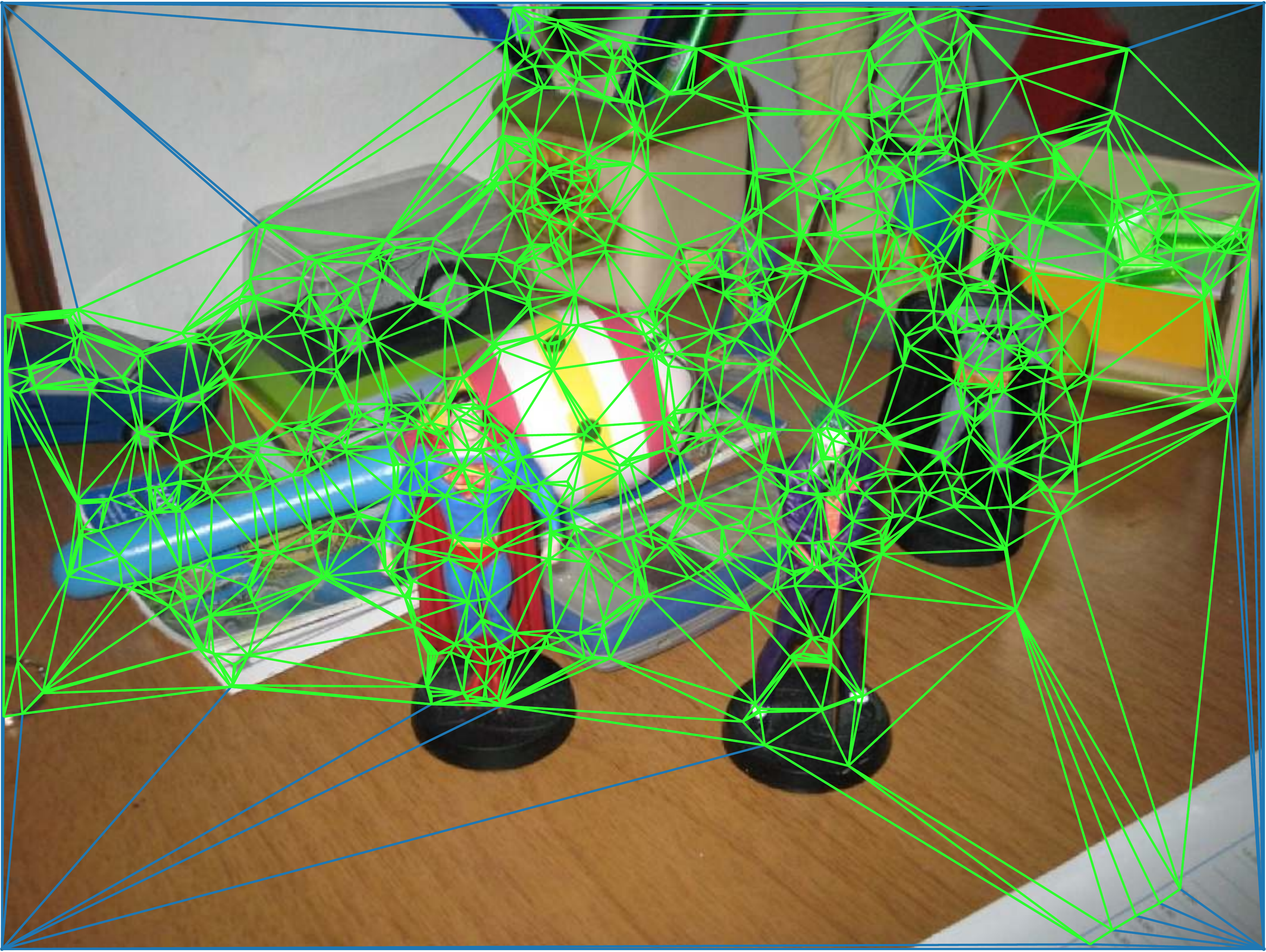}
	}
	\hspace{0.2em}
	\hfil
	\hspace{0.2em}
	\subfloat[]{\label{corner_mesh_split_img}
		\includegraphics[width=0.185\textwidth]{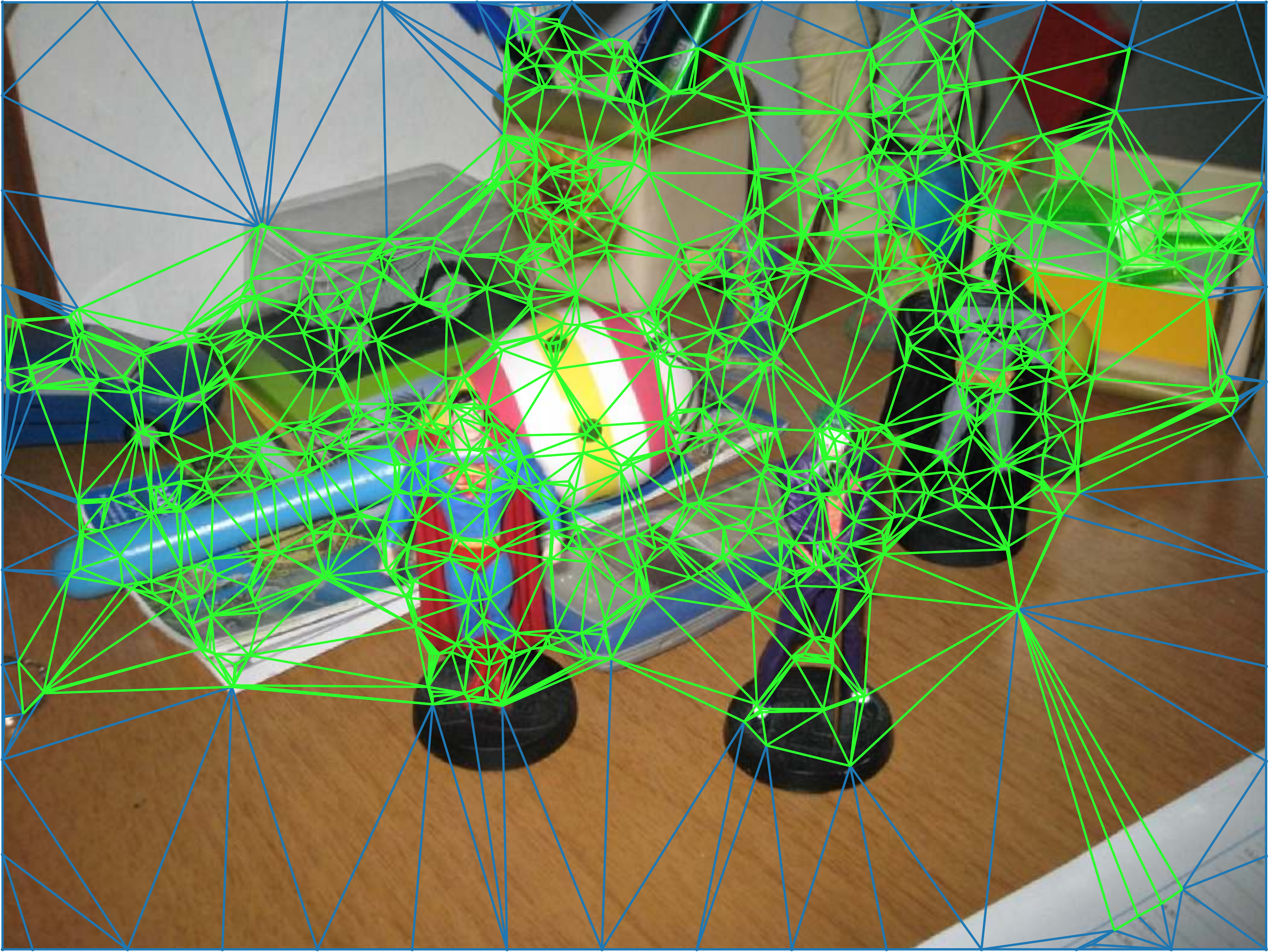}
	}
	\\
	\subfloat[]{\label{convex_hull_img}
		\includegraphics[width=0.22\textwidth]{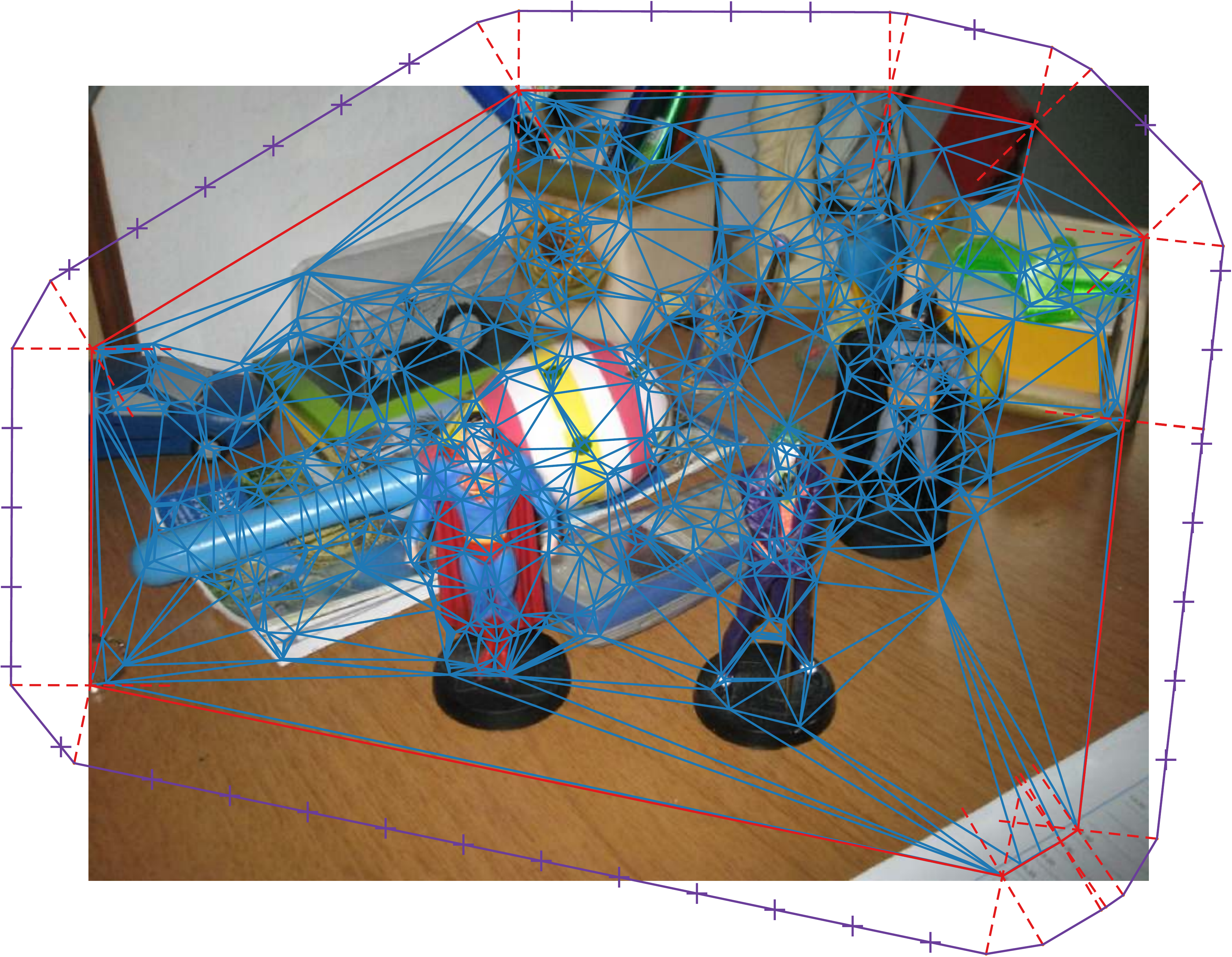}
	}
	\hfil
	\subfloat[]{\label{convex_hull_mesh_img}
		\includegraphics[width=0.22\textwidth]{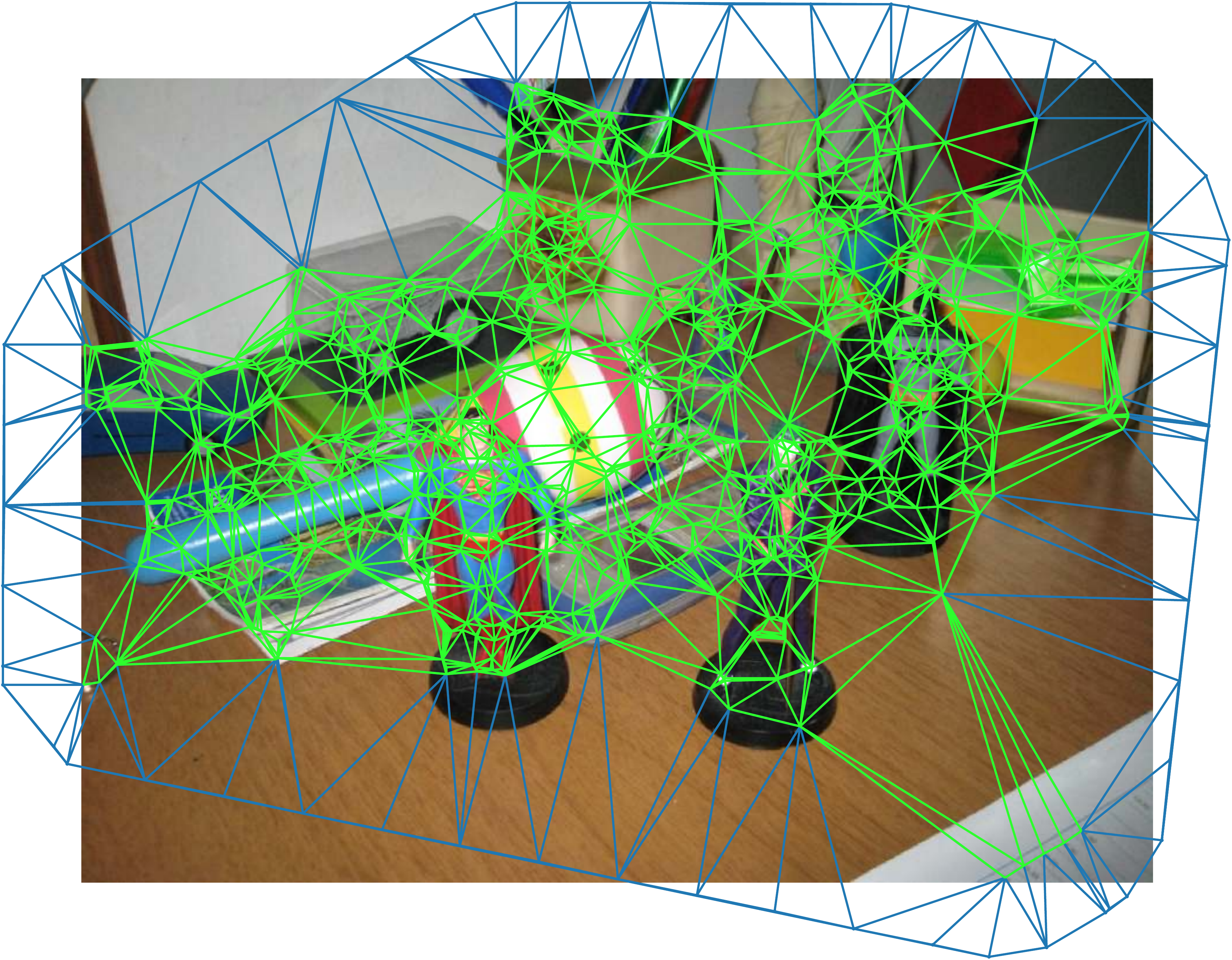}
	}
	\\
	\subfloat[]{\label{boundary_img_}
		\includegraphics[width=0.22\textwidth]{imgs_compressed/boundary.pdf}
	}
	\hfil
	\subfloat[]{\label{boundary_mesh_img_}
		\includegraphics[width=0.22\textwidth]{imgs_compressed/boundary_mesh.pdf}
	}
	\caption{\label{boundary_imgs}
		Delaunay triangulations by including image corners \protect\subref{corner_mesh_img} and by breaking the canvas borders into multiple lines \protect\subref{corner_mesh_split_img}. \protect\subref{convex_hull_img} Convex hull (solid red) fattening (dashed red) in order to obtain expanded contour edges (purple) with the resulting Delaunay triangulation~\protect\subref{convex_hull_mesh_img}. \protect\subref{boundary_img_} Analogous results using alpha-shape boundary edges (orange) and \protect\subref{boundary_mesh_img_} the final Delaunay triangulation employed in DTM. Green Delaunay edges are used to distinguish the edges connecting keypoints from the others (see Sec.~\ref{ks} for details, best view in color and zoomed in).}
\end{figure}

\section{Neighborhood formulation differences}\label{appendix_neighborhood}
\begin{figure}[h!]
	\center
	\subfloat[]{\label{plane_h}
		\includegraphics[width=0.11\textwidth]{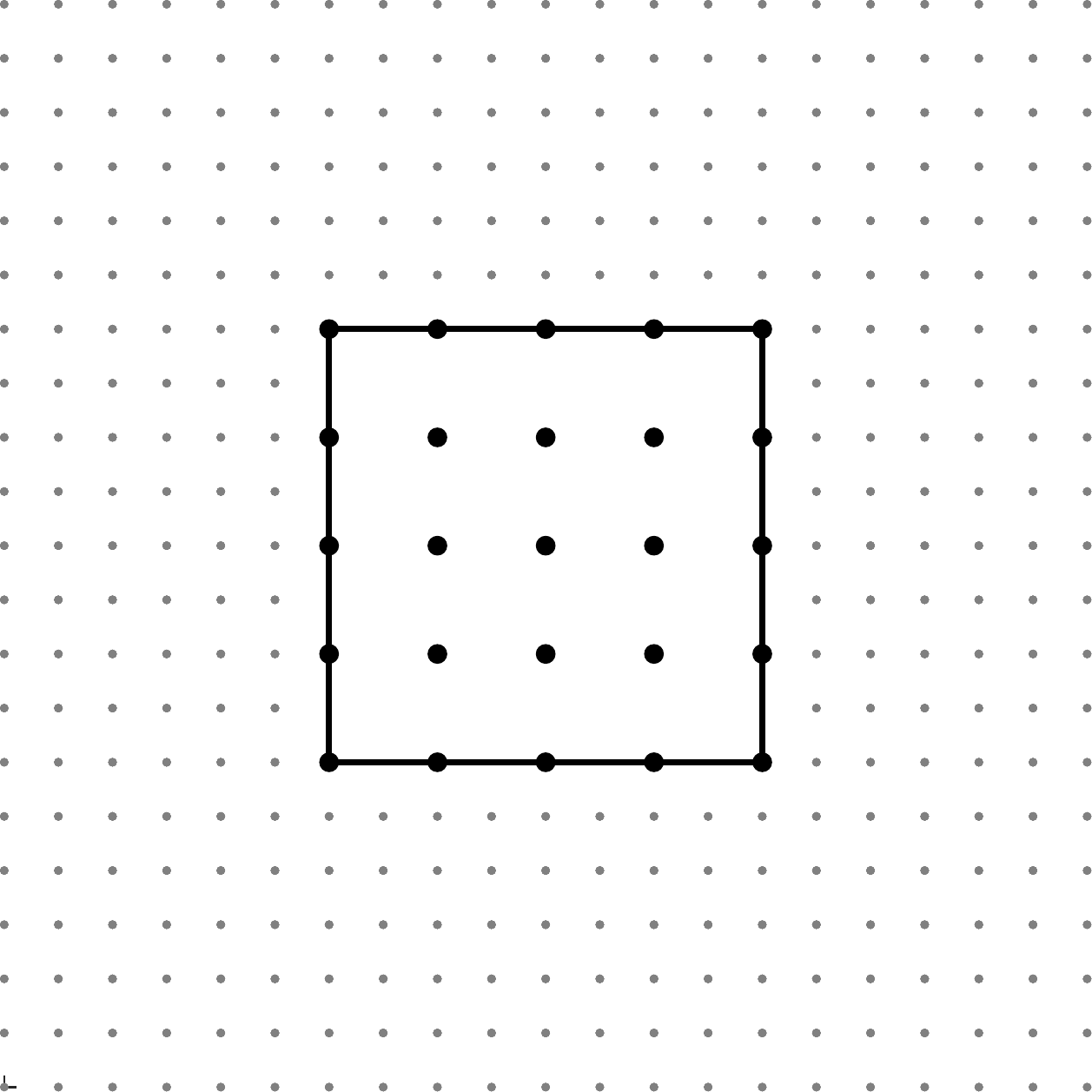}
	}
	\hfil
	\subfloat[]{\label{plane_e}
		\includegraphics[width=0.11\textwidth]{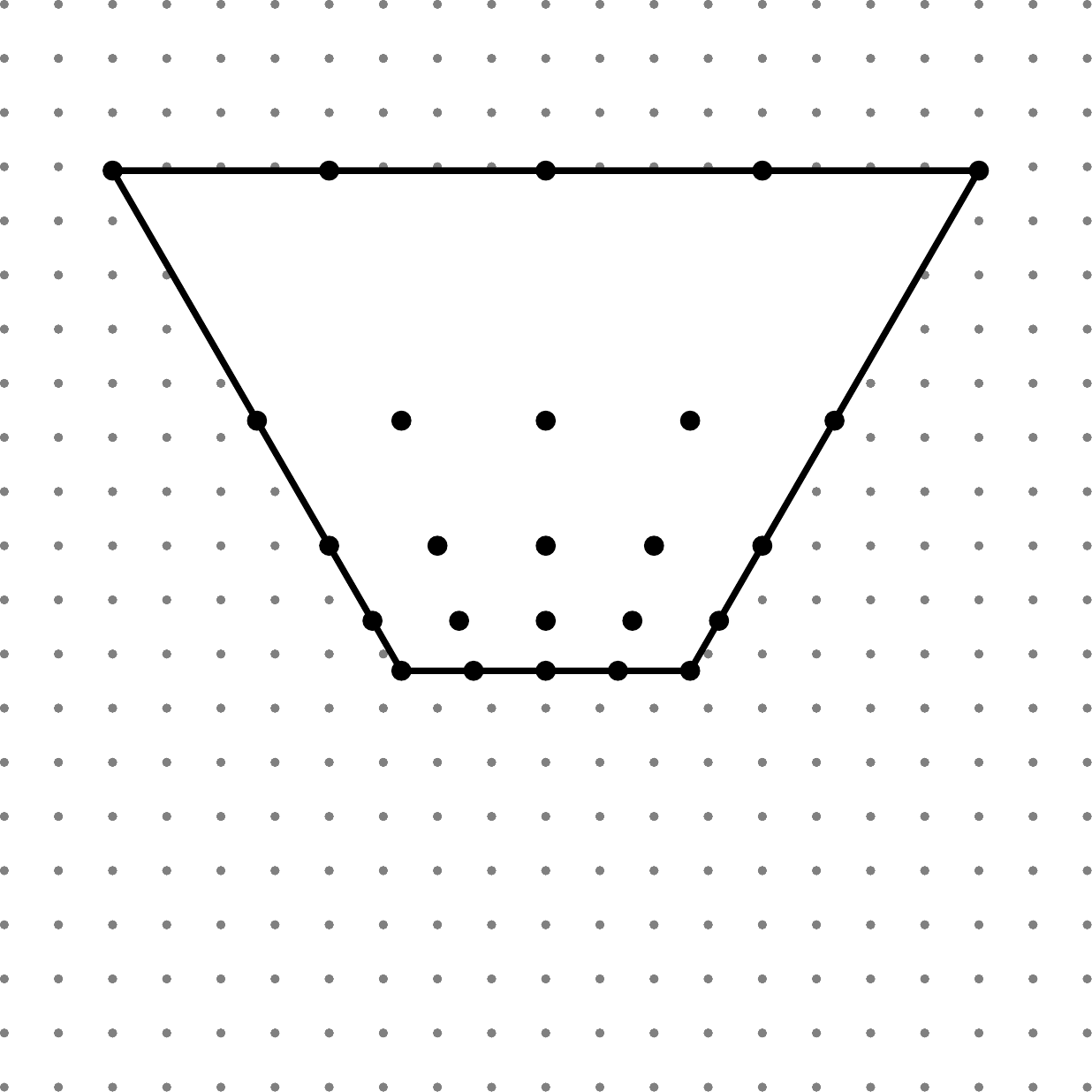}
	}
	\hfil
	\subfloat[]{\label{plane_b}
		\includegraphics[width=0.11\textwidth]{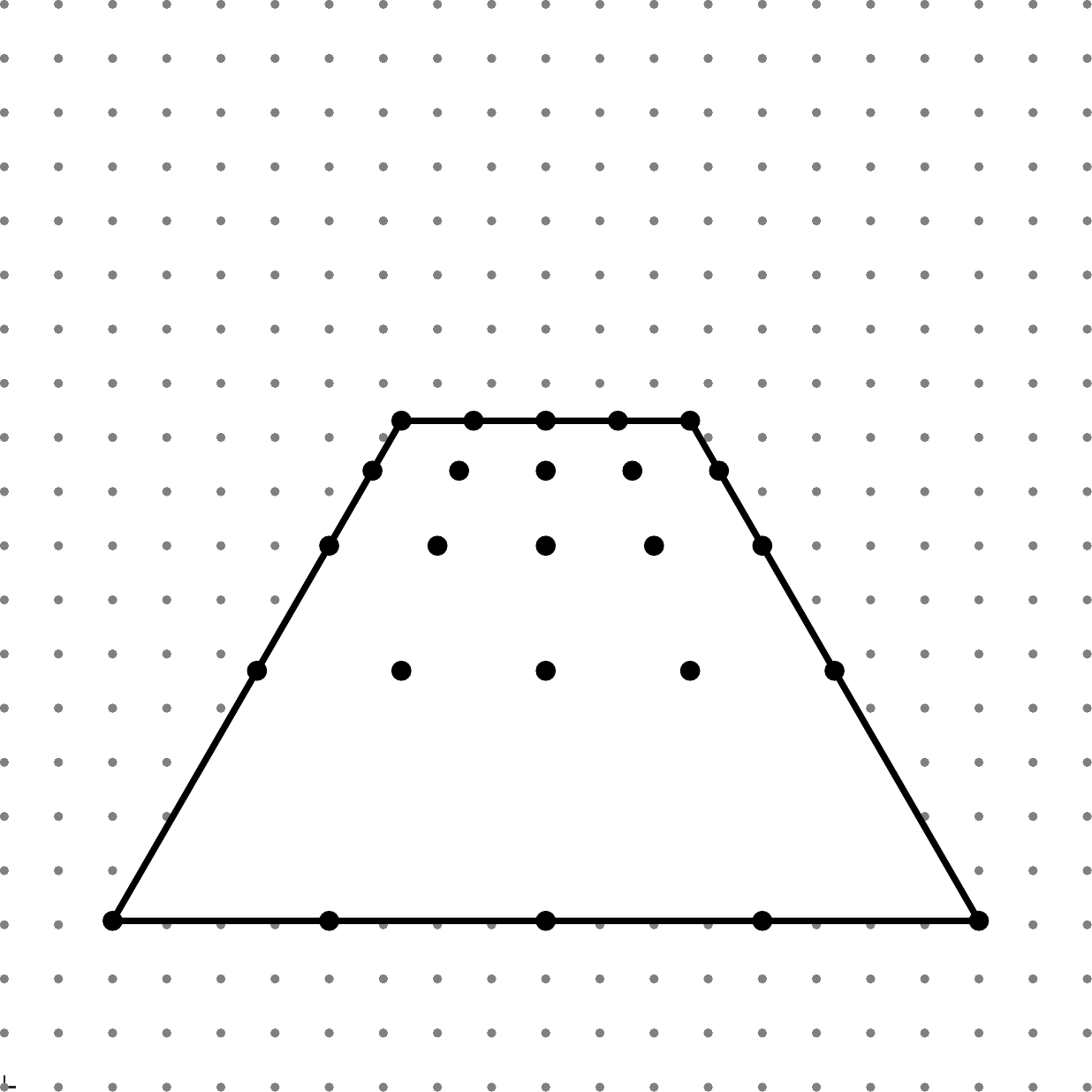}
	}
	\caption{\label{nplane1}
	\protect\subref{plane_h}~Frontal view of a toy example scene with a static background (gray dots) and a planar surface as foreground (black edges and dots). Other views of the scene with the foreground plane slanted~\protect\subref{plane_e}~upward and~\protect\subref{plane_e}~downward (best view in color and zoomed in).}
\end{figure}

\begin{figure}[h!]
	\center
	\subfloat[]{\label{plane_f}
		\includegraphics[width=0.11\textwidth]{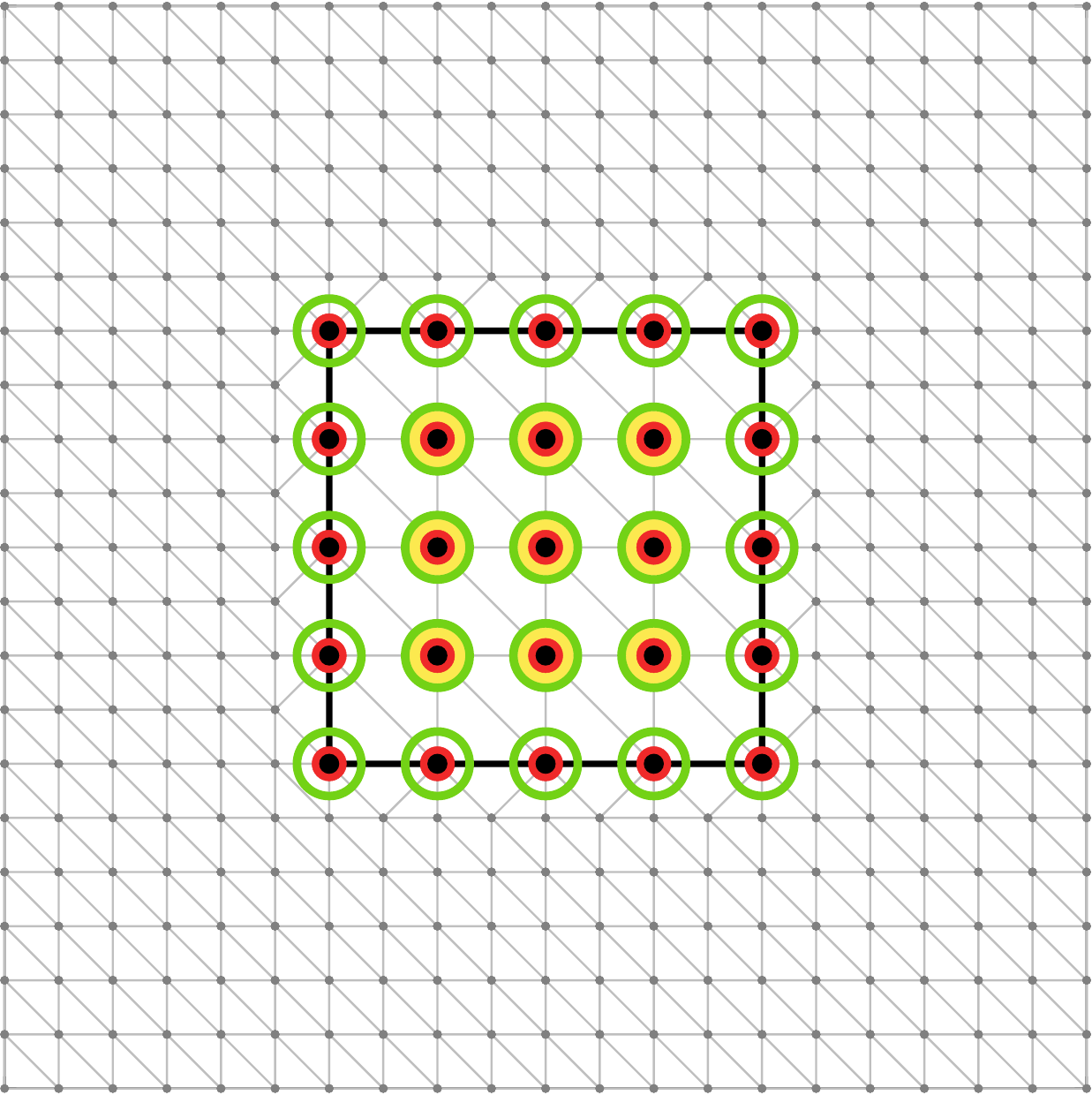}
	}
	\hfil
	\subfloat[]{\label{plane_c}
		\includegraphics[width=0.11\textwidth]{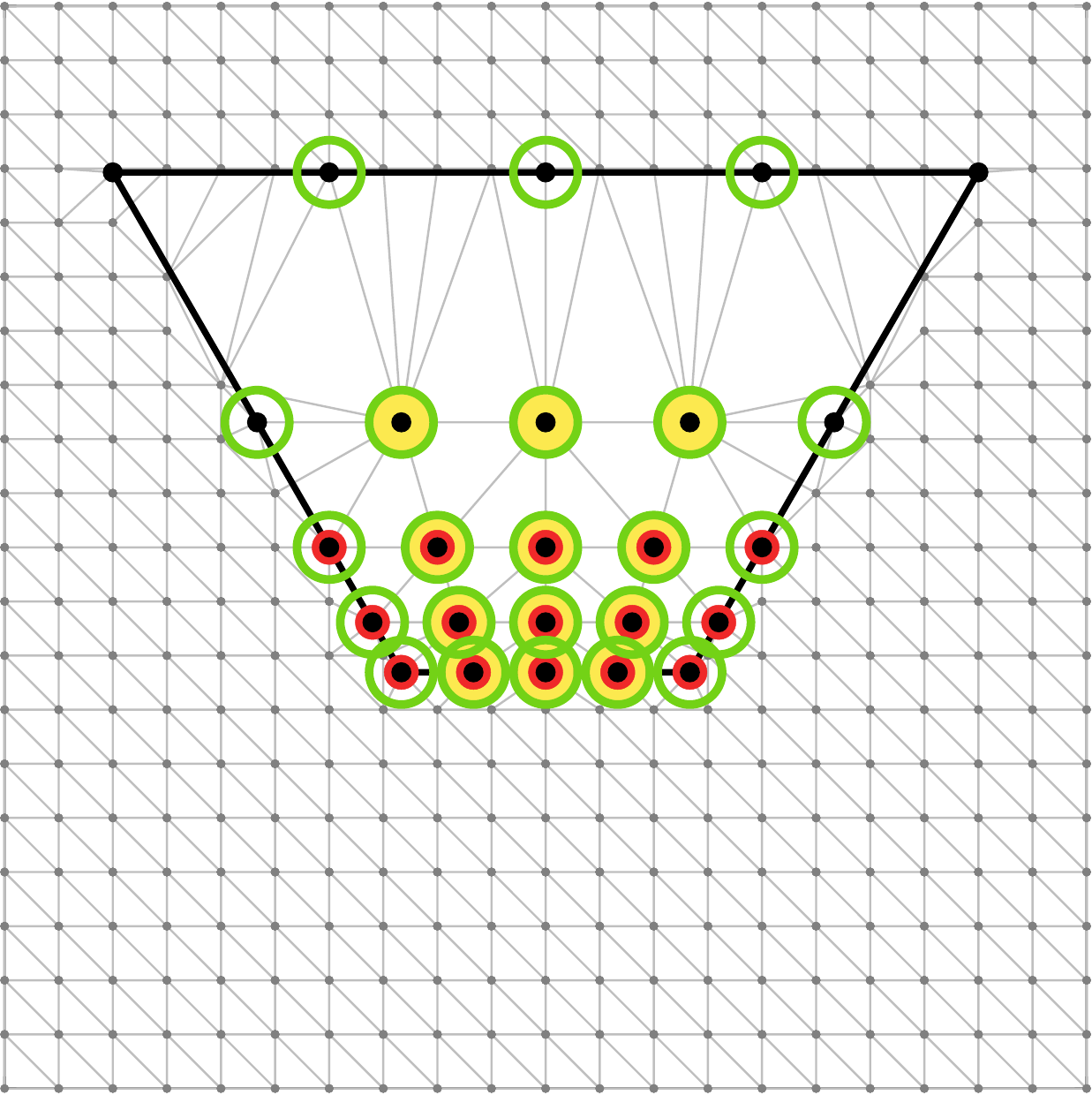}
	}
	\hfil
	\subfloat[]{\label{plane_a}
		\includegraphics[width=0.11\textwidth]{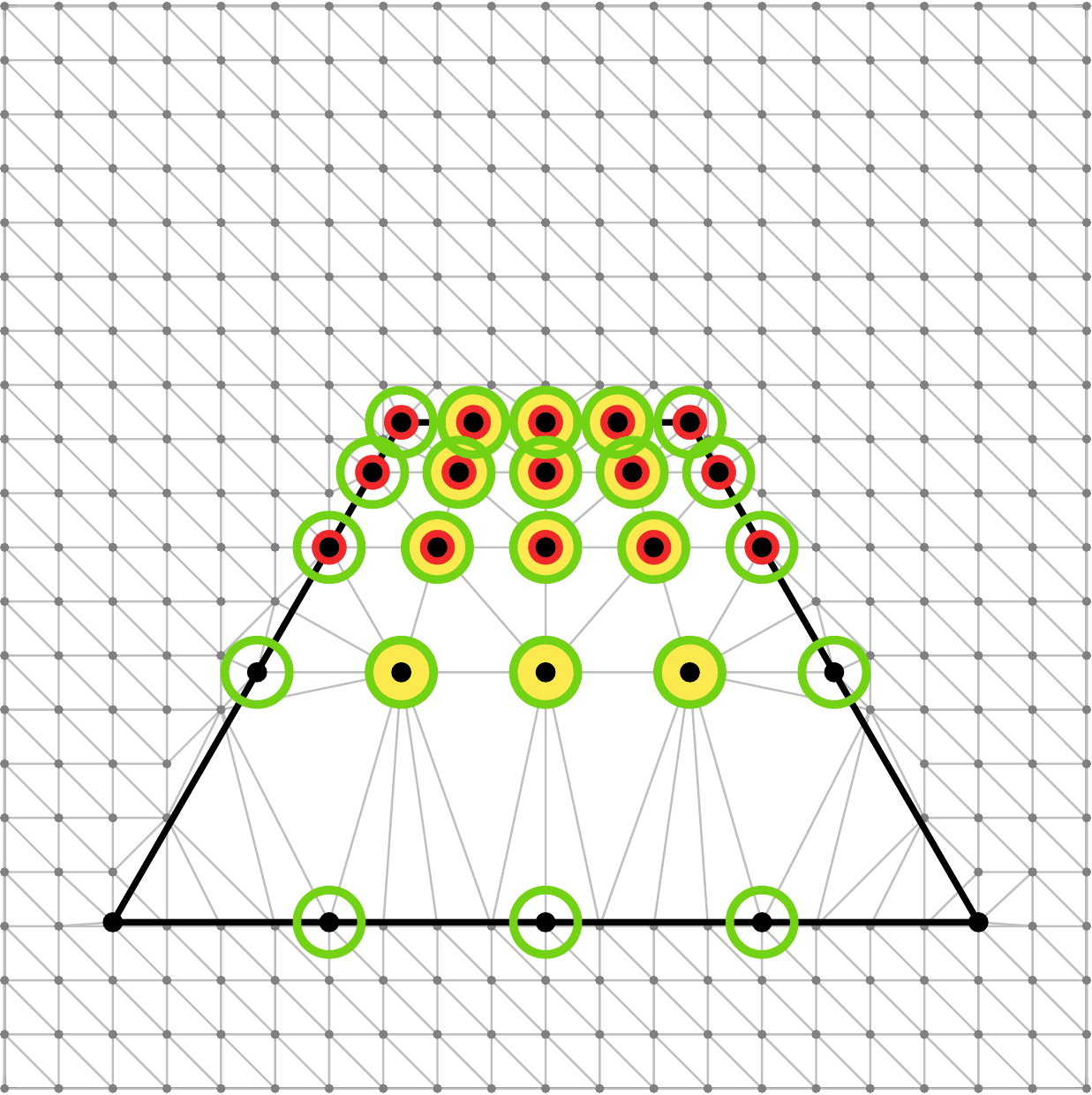}
	}
	\vspace{-0.35em}
	\caption{\label{nplane2}
	Neighborhoods for the foreground keypoints of each scene view of Fig.~\protect\ref{nplane1} containing at least another foreground keypoint, i.e. sharing the motion field cluster. These neighborhoods are indicated respectively in red, yellow and green for the circular radius neighborhood, the closest $\mathrm{k}$ nearest neighborhood and the Delaunay-based neighborhood. Delaunay triangulation edges are in gray (best view in color and zoomed in).}
\end{figure}

\begin{figure}[h!]
	\center
	\subfloat[]{\label{plane_g}
		\includegraphics[width=0.11\textwidth]{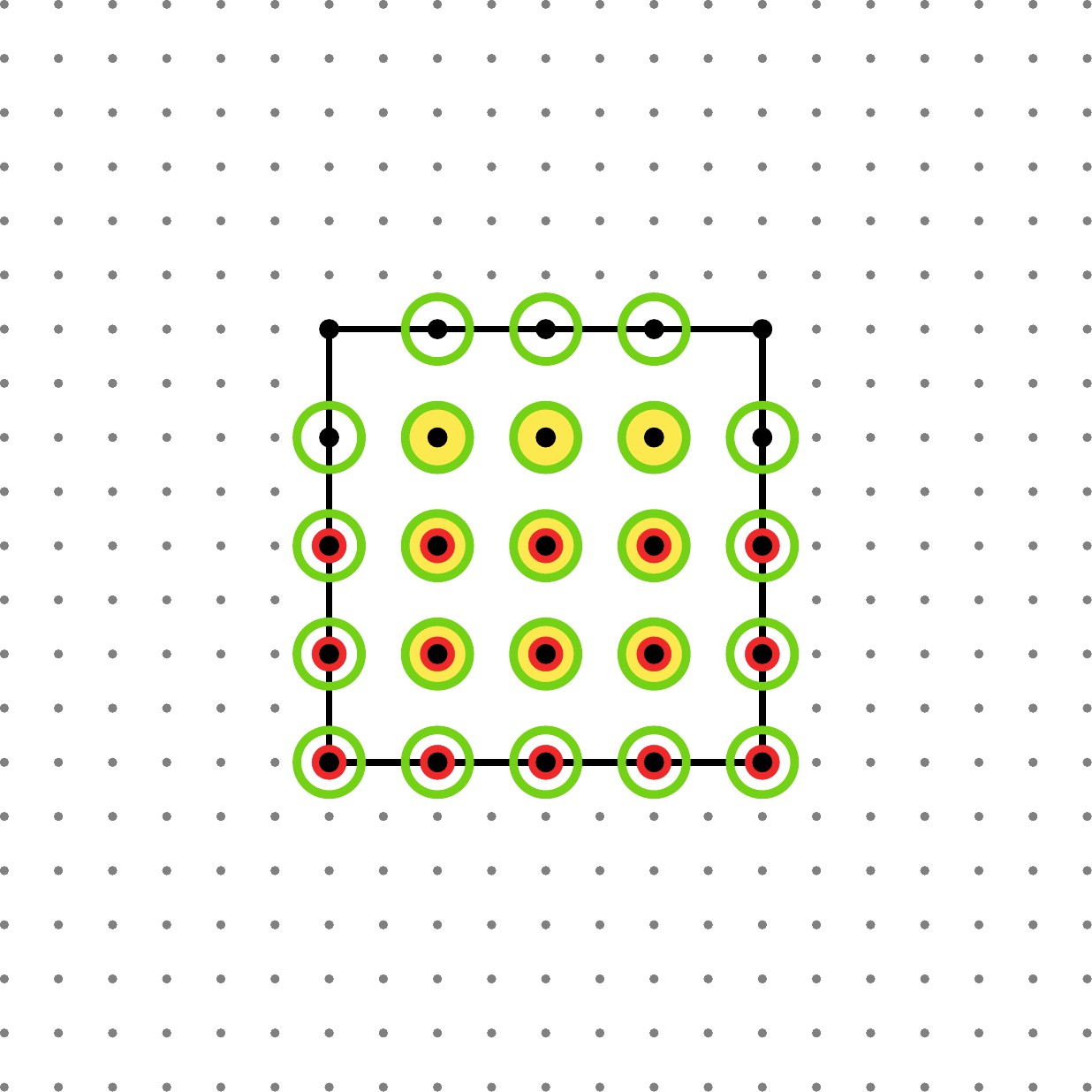}
	}
	\hfil
	\subfloat[]{\label{plane_d}
		\includegraphics[width=0.11\textwidth]{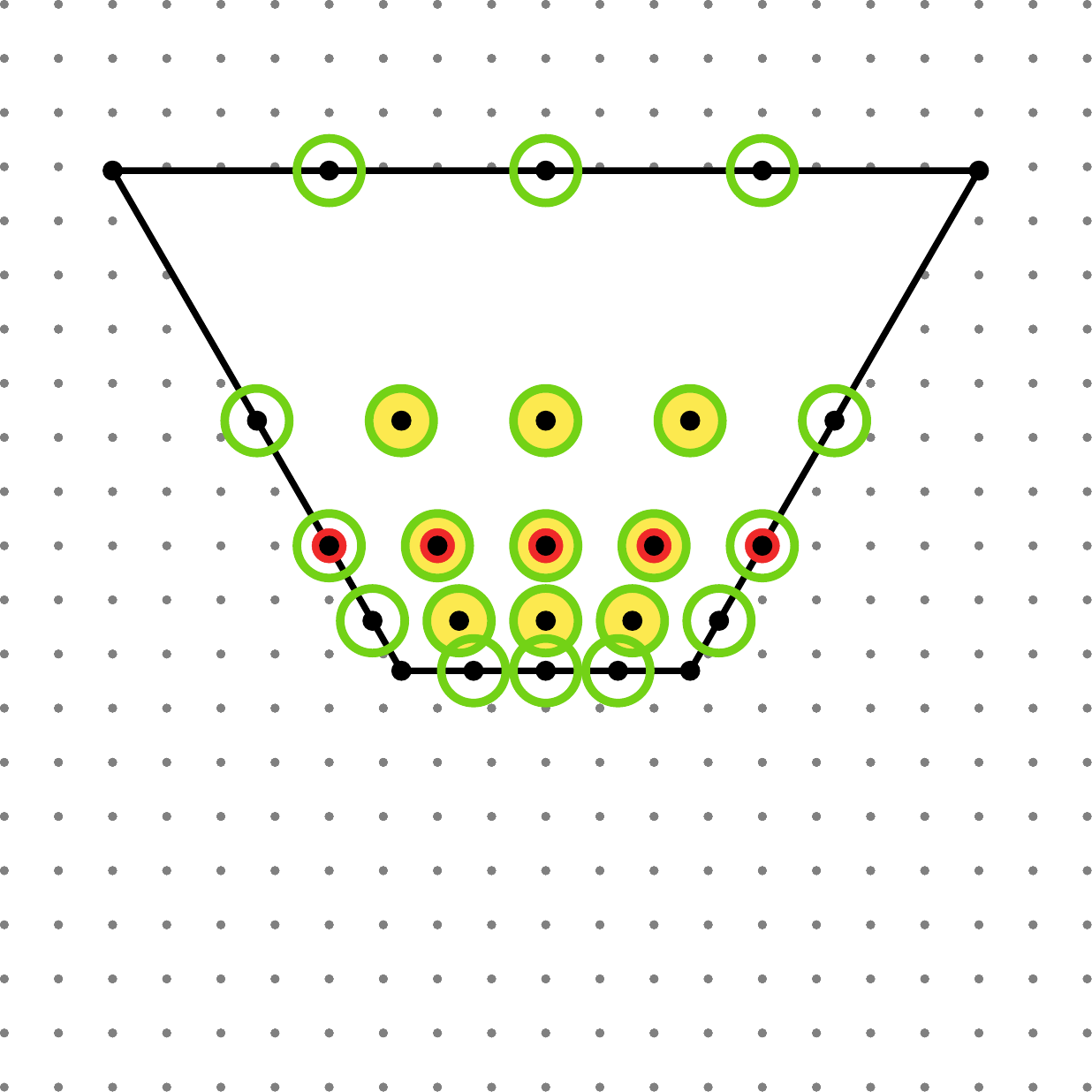}
	}
	\vspace{-0.33em}
	\caption{\label{nplane3}
	The intersections of the corresponding neighborhoods between~\protect\subref{plane_g} Fig.~\protect\ref{plane_f} and Fig.~\protect\ref{plane_c}, and~\protect\subref{plane_d} Fig.~\protect\ref{plane_c} and Fig.~\protect\ref{plane_a} on the first view according to Fig.~\protect\ref{nplane2} notation (best view in color and zoomed in).}
\end{figure}

Figure~\ref{nplane2} shows the intuition behind the different neighborhood formulations, discussed in the main text, on the toy example scene of Fig.~\ref{nplane1}. Assuming that dots are detected as keypoints, the aim is to check whether for the foreground object the keypoint neighborhood contains at least another foreground keypoints, i.e. beloging to the same motion field cluster. As neighborhood parameters, the circular radius is set equal to $1/4$ of the foreground square side in the frontal view, while $\mathrm{k}=3$. These parameters are set empirically. It can be noted that the Delaunay-based neighborhood allows to have more neighborhoods containing keypoints of the same motion field cluster than the other neighborhood formulations in all the views, avoiding to setup parameters in the neighborhood design. Clearly, background keypoints get included, but their number decreases when considering the intersection of the corresponding neighborhoods among the images, as shown in Fig.~\ref{nplane3}. Notice that apart from the Delaunay-based neighborhood, intersection causes a great drop of neighborhoods within the same motion field cluster. Moreover, patch-based local similarity or affine transformations for this scene is very unlikely to be useful in checking consistency.\vspace{-1.5em}

\vspace{-1.7em}
\vspace{-1.7em}
\vspace{-1.7em}
\section{Evaluation dataset}\label{appendix_dataset}
\begin{figure}[h!]
	\center
	\includegraphics[width=0.46\textwidth]{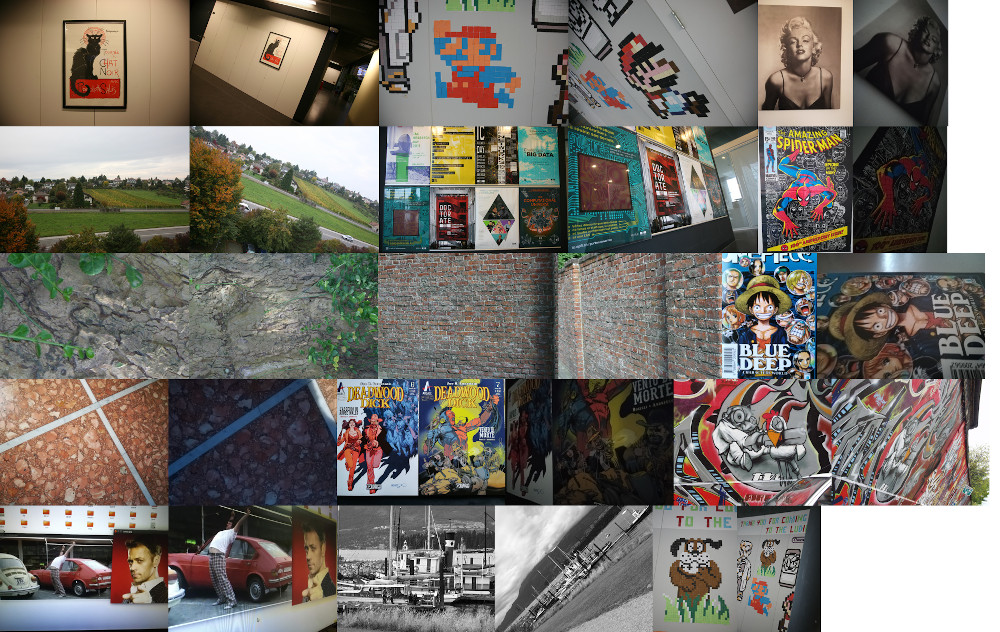}
	\caption{\label{planar_dataset}
		Exemplar image thumbnails for each scene of the planar dataset. Each scene is made up of six images, only two of these are shown (see Sec.~\ref{eval_setup} for details, best viewed in color and zoomed in).}
\end{figure}

\begin{figure}[h!]
	\center
		\includegraphics[width=0.48\textwidth]{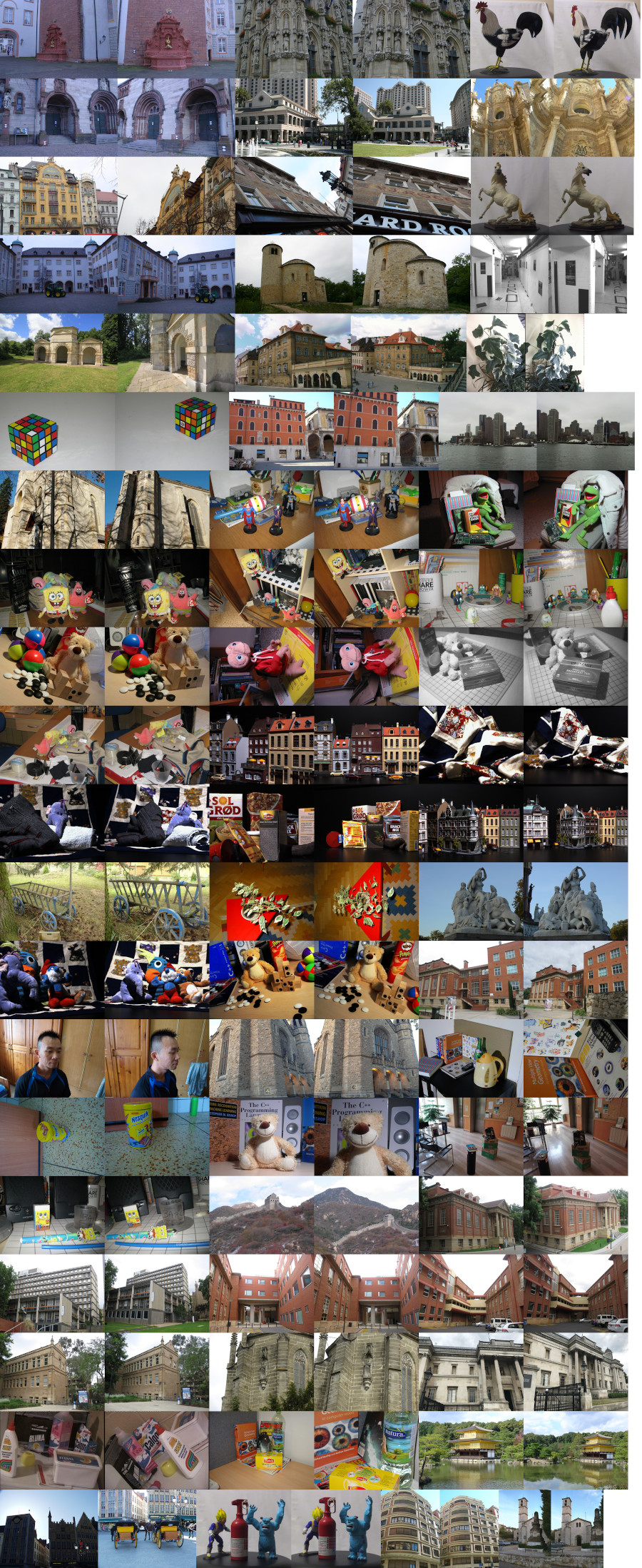}
	\caption{\label{non_planar_datasets}
		Exemplar image thumbnails for each scene of the non-planar dataset. Each scene is made up of two or three images, only two of these are shown (see Sec.~\ref{eval_setup} for details, best viewed in color and zoomed in).}
\end{figure}

\vfill
\vspace{6em}\hphantom{a}
\newpage

\section{Ground truth estimation methods}\label{appendix_setup}
\vspace{-1em}
\begin{figure}[h!]
	\center
	\subfloat[]{\label{setup_planar_sift}
		\includegraphics[width=0.22\textwidth]{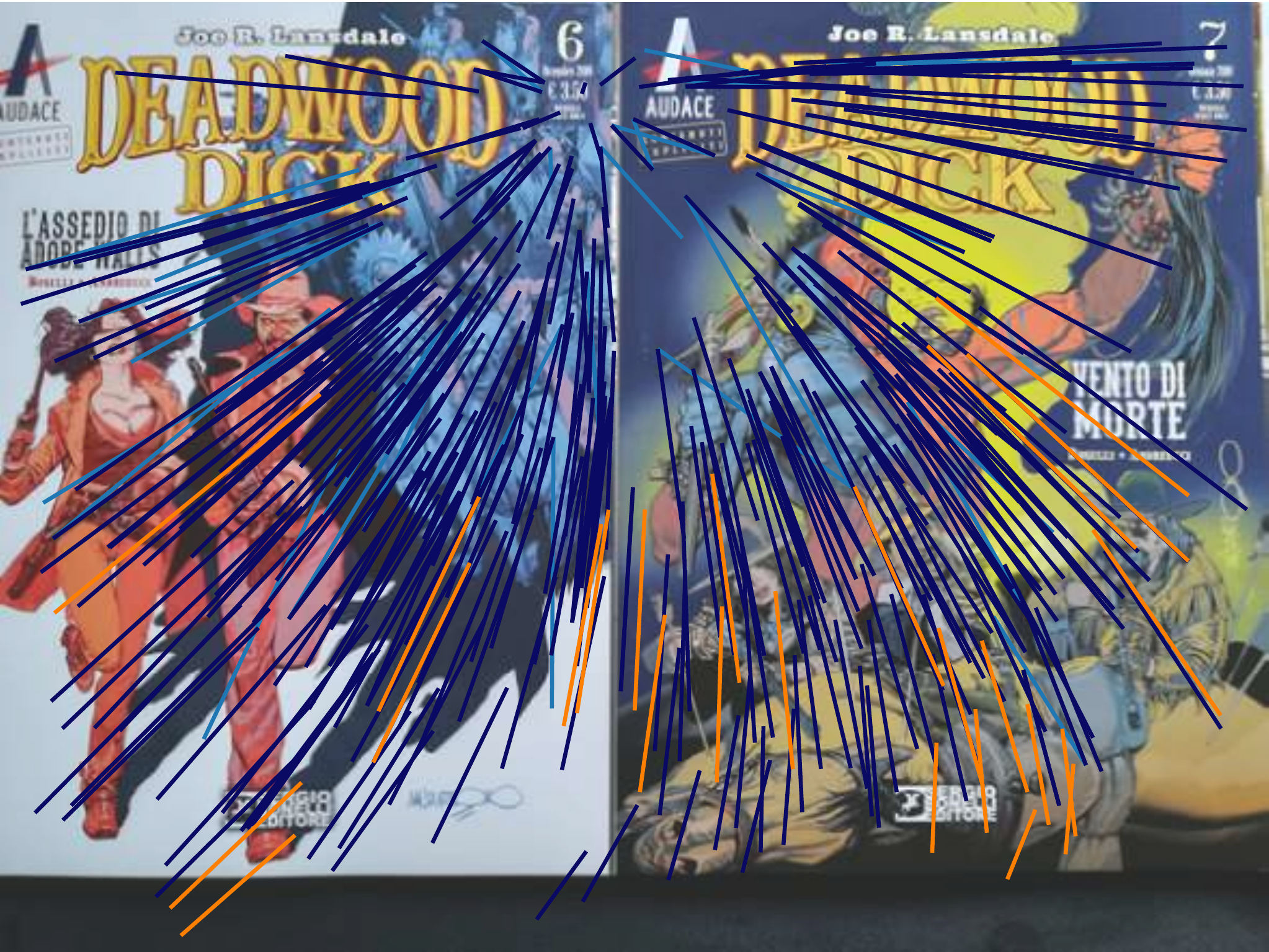}
	}
	\hfil
	\subfloat[]{\label{setup_planar_harrisz}
		\includegraphics[width=0.22\textwidth]{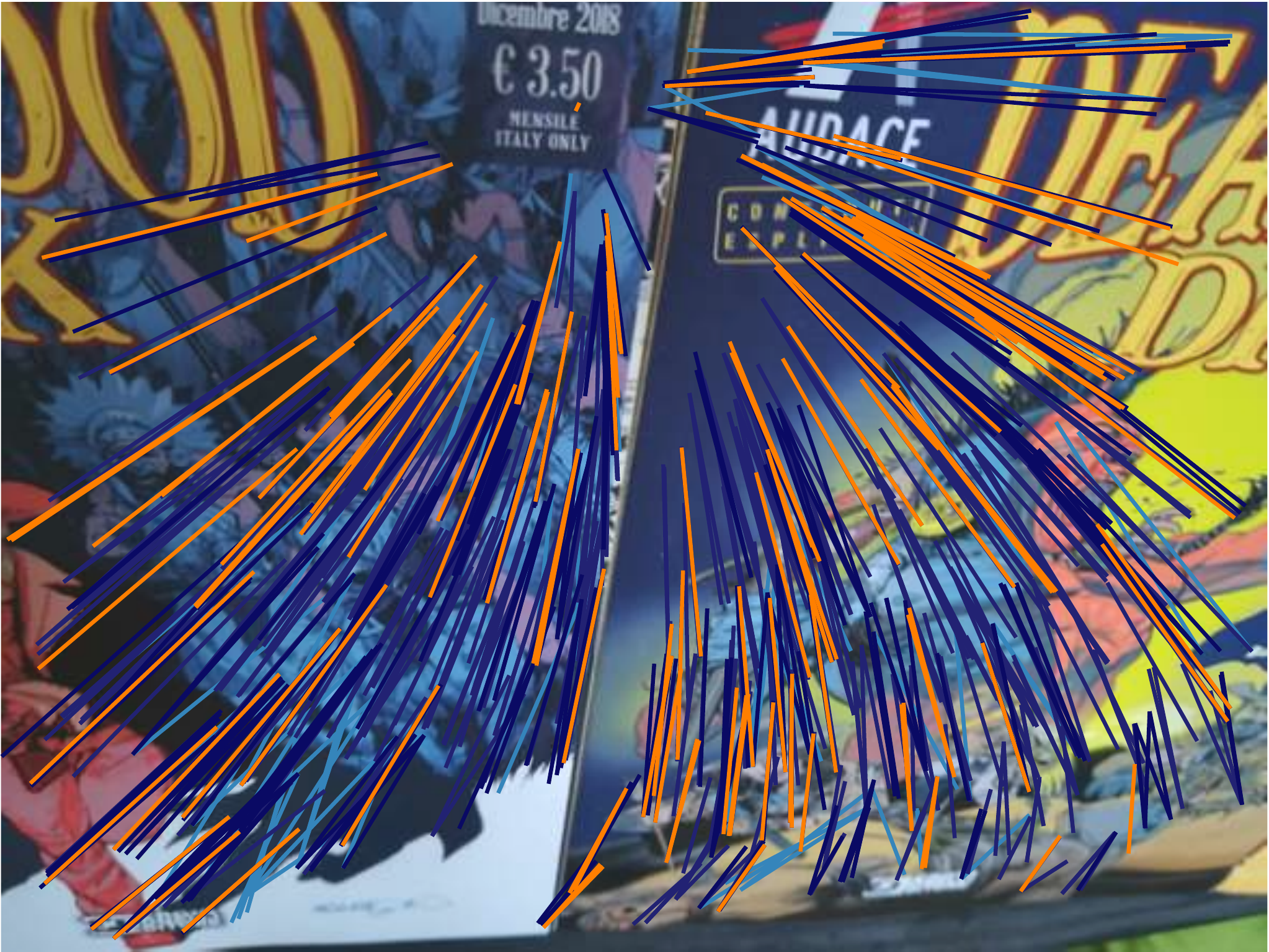}
	}
	\\[-0.5em]
	\subfloat[]{\label{setup_planar_sift_patches}
		\includegraphics[width=0.22\textwidth]{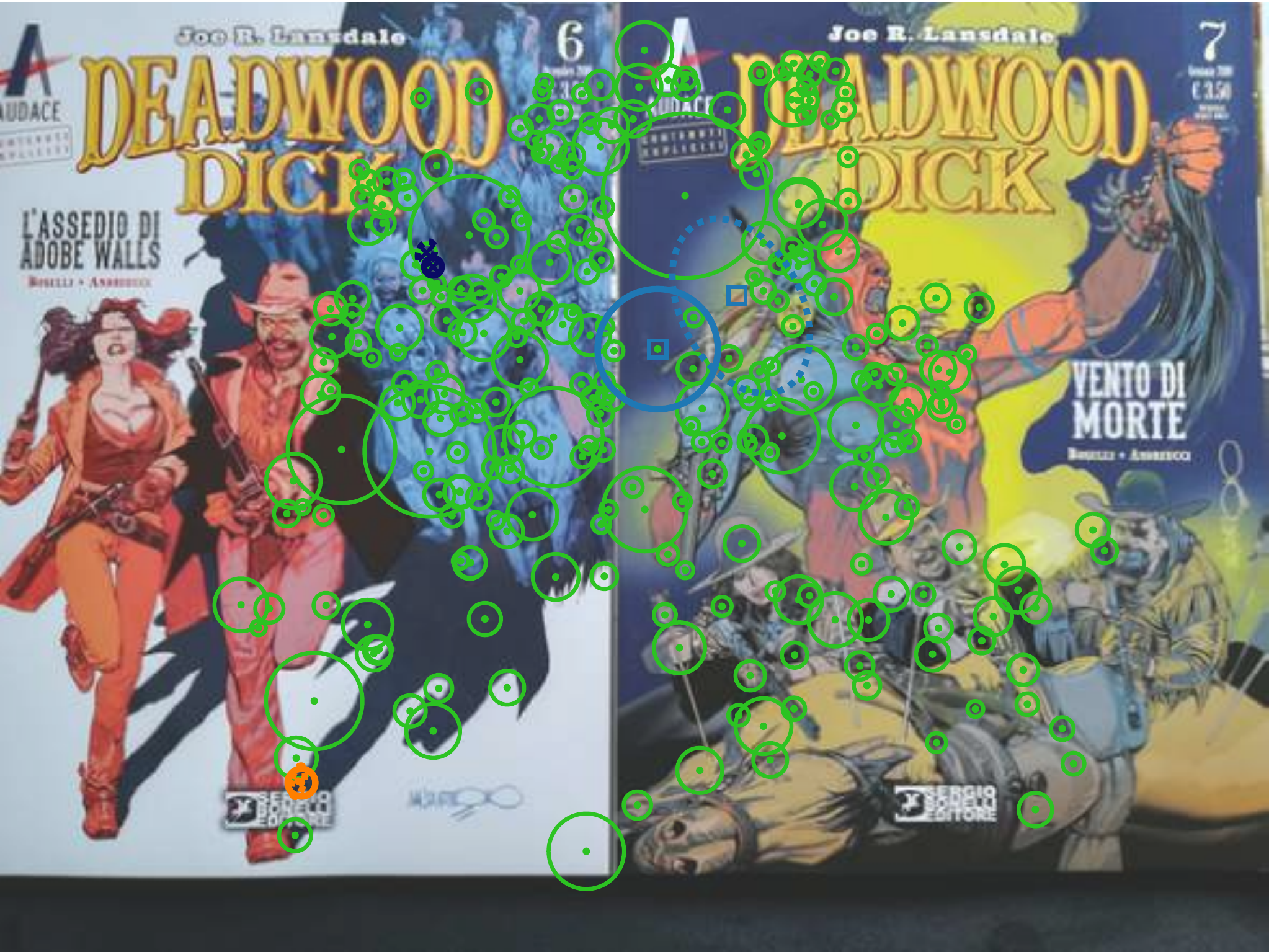}
	}
	\hfil
	\subfloat[]{\label{setup_planar_harrisz_patches}
		\includegraphics[width=0.22\textwidth]{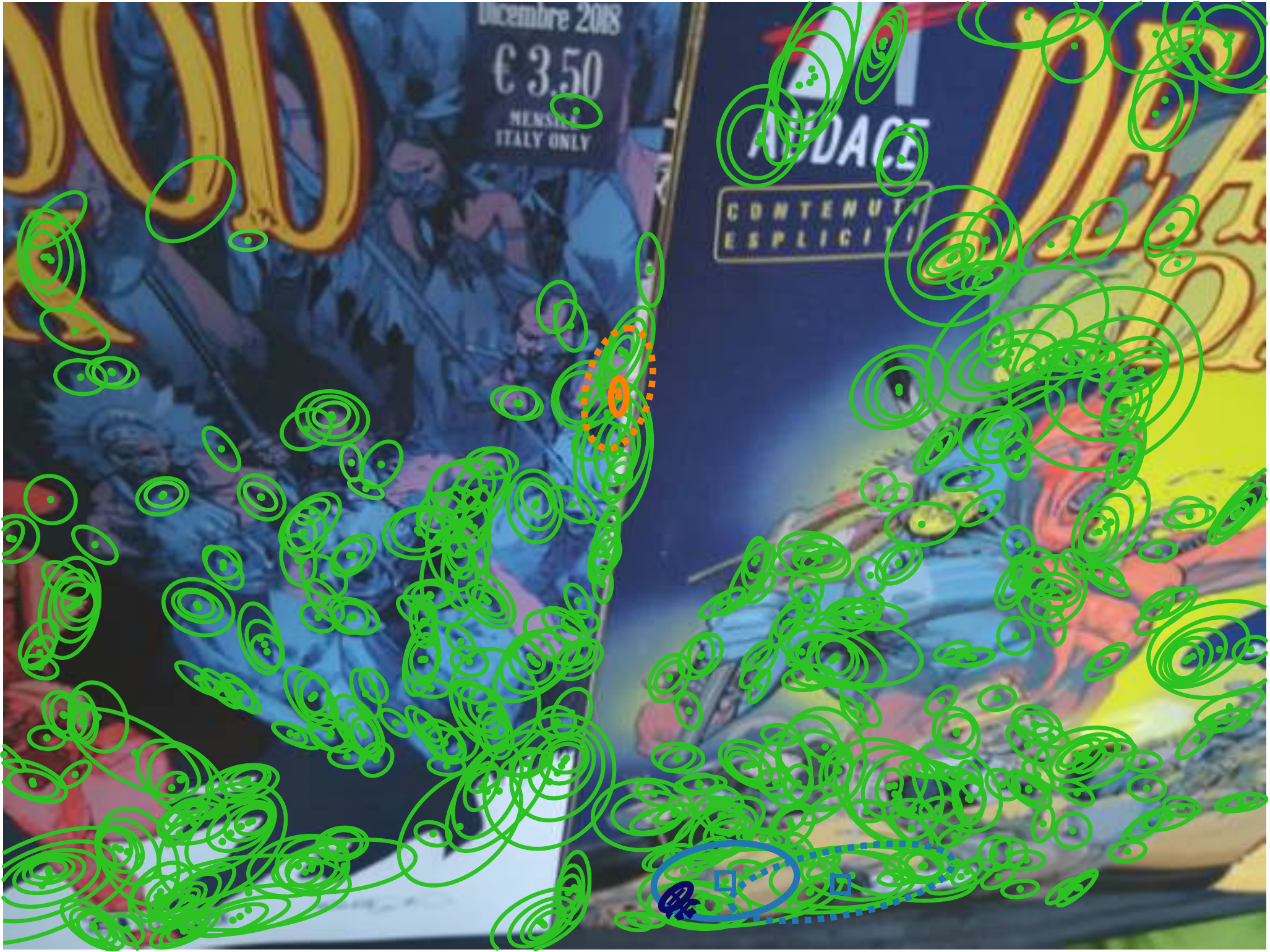}
	}
	\\[-0.5em]
	\subfloat[]{\label{setup_non_planar_sift}
		\includegraphics[width=0.22\textwidth]{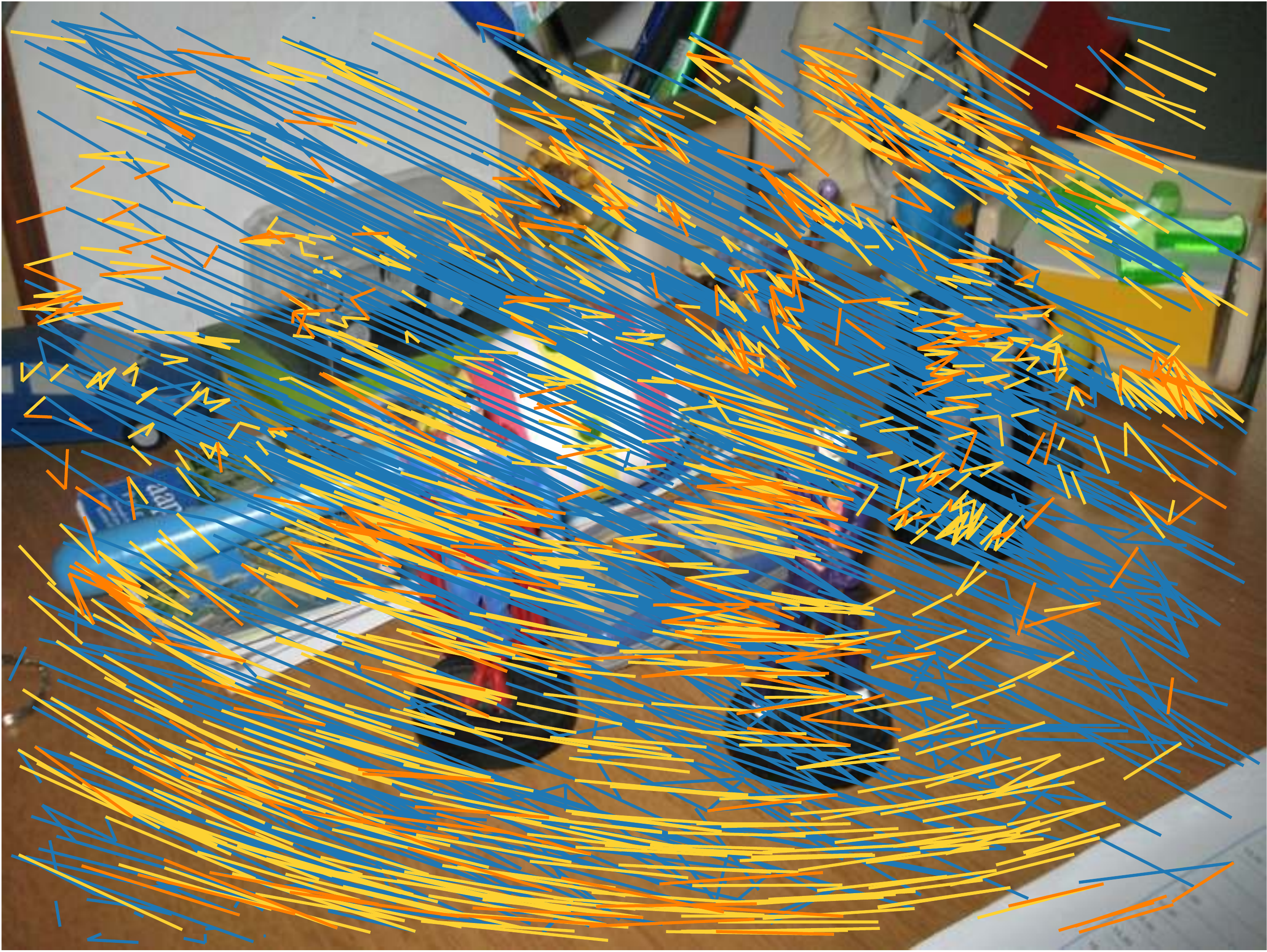}
	}
	\hfil
	\subfloat[]{\label{setup_non_planar_harrisz}
		\includegraphics[width=0.22\textwidth]{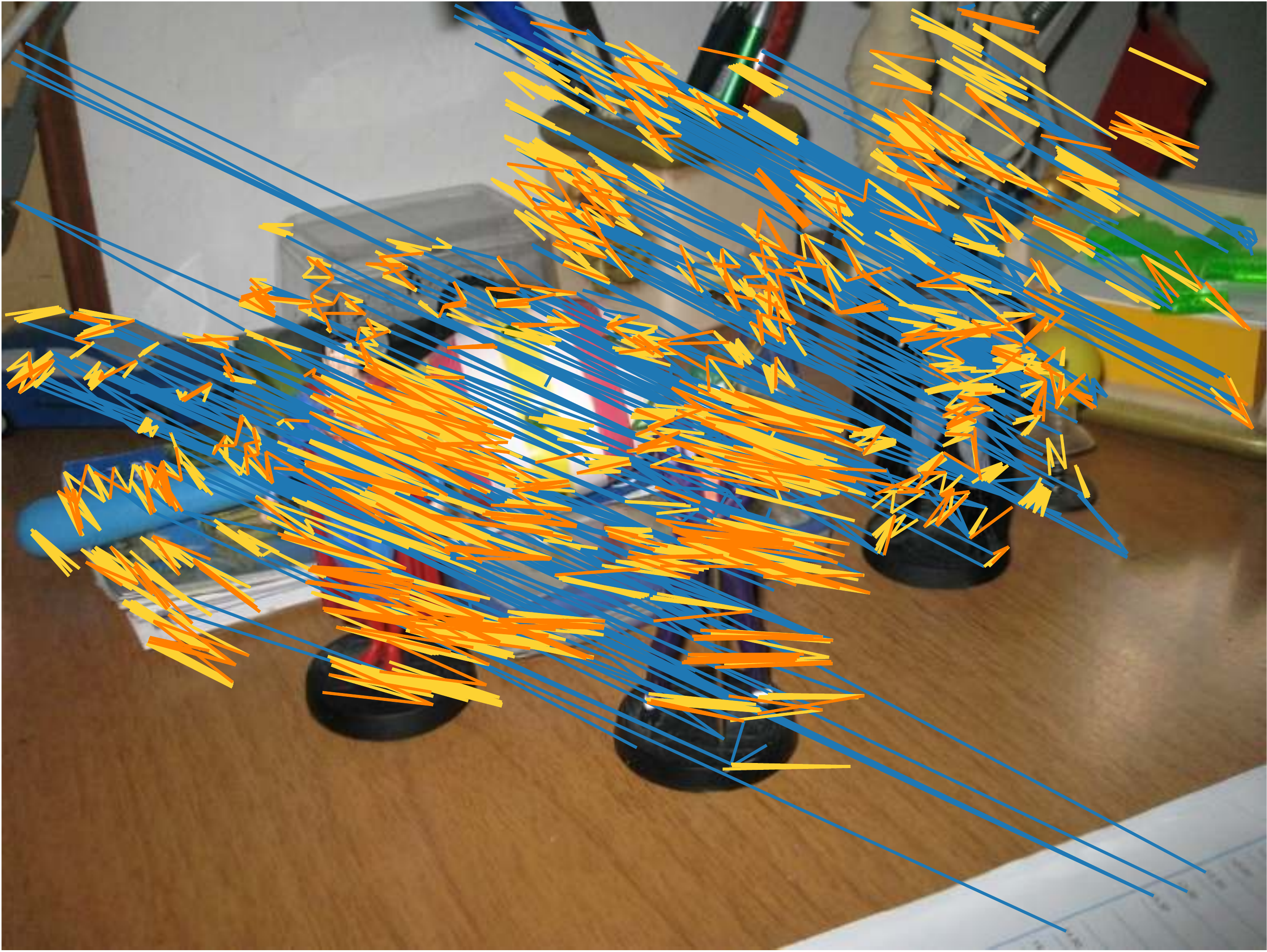}
	}
	\\[-0.5em]
	\subfloat[]{\label{setup_non_planar_sift_ov_epi}
		\includegraphics[width=0.22\textwidth]{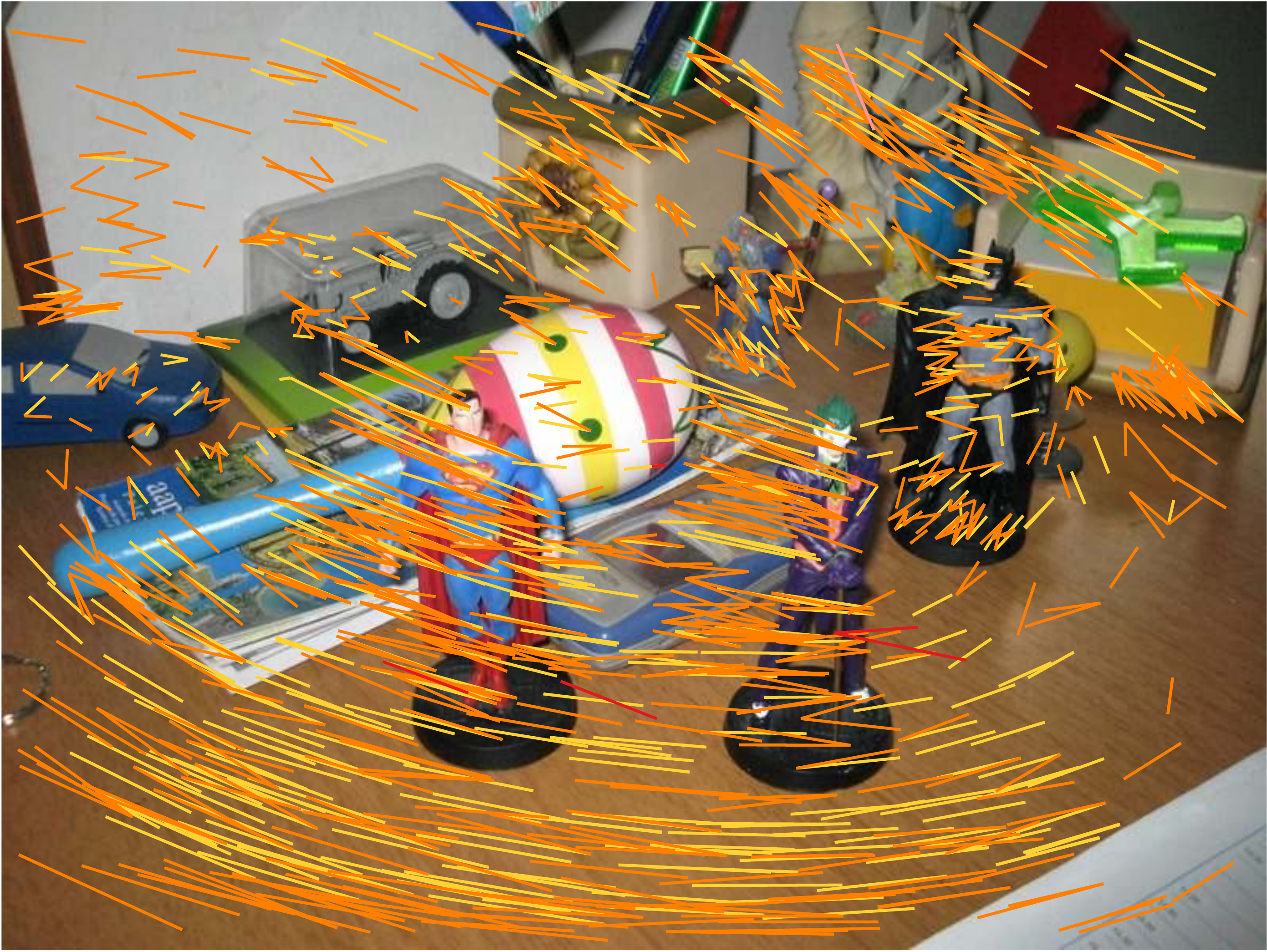}
	}
	\hfil
	\subfloat[]{\label{setup_non_planar_harrisz_ov_epi}
		\includegraphics[width=0.22\textwidth]{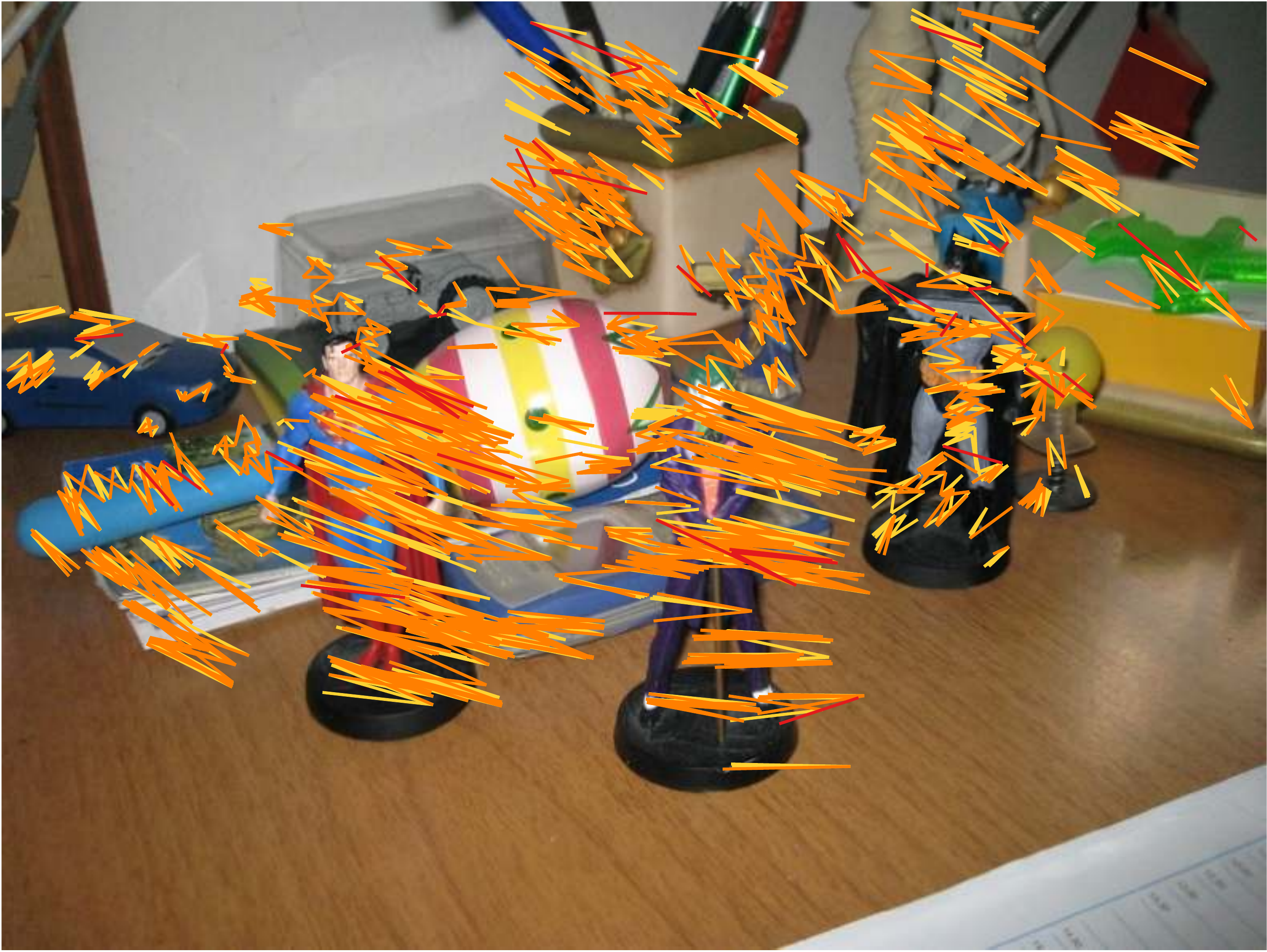}
	}
	\\[-0.5em]
	\subfloat[]{\label{setup_non_planar_sift_patches}
		\includegraphics[width=0.22\textwidth]{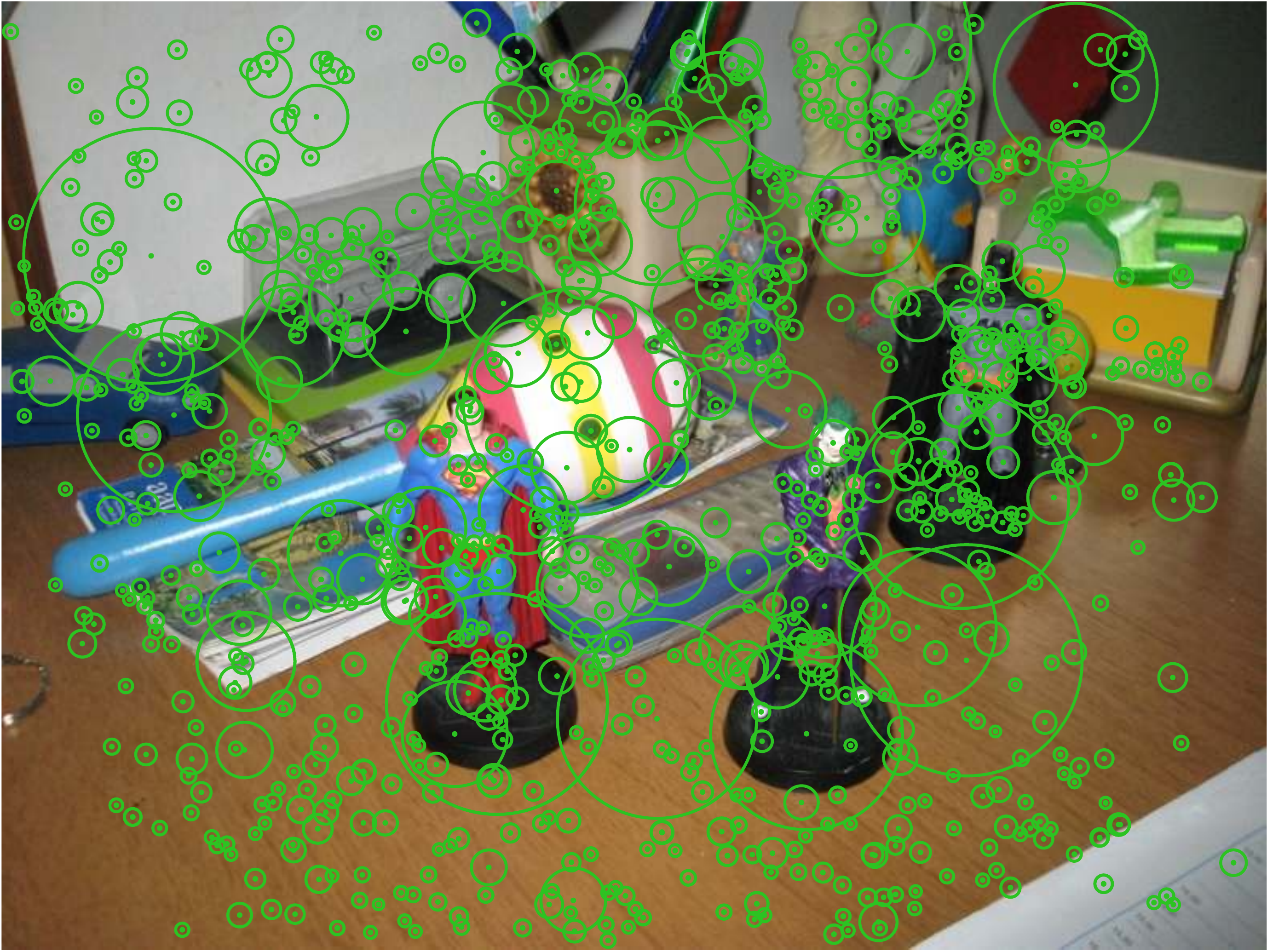}
	}
	\hfil
	\subfloat[]{\label{setup_non_planar_harrisz_patches}
		\includegraphics[width=0.22\textwidth]{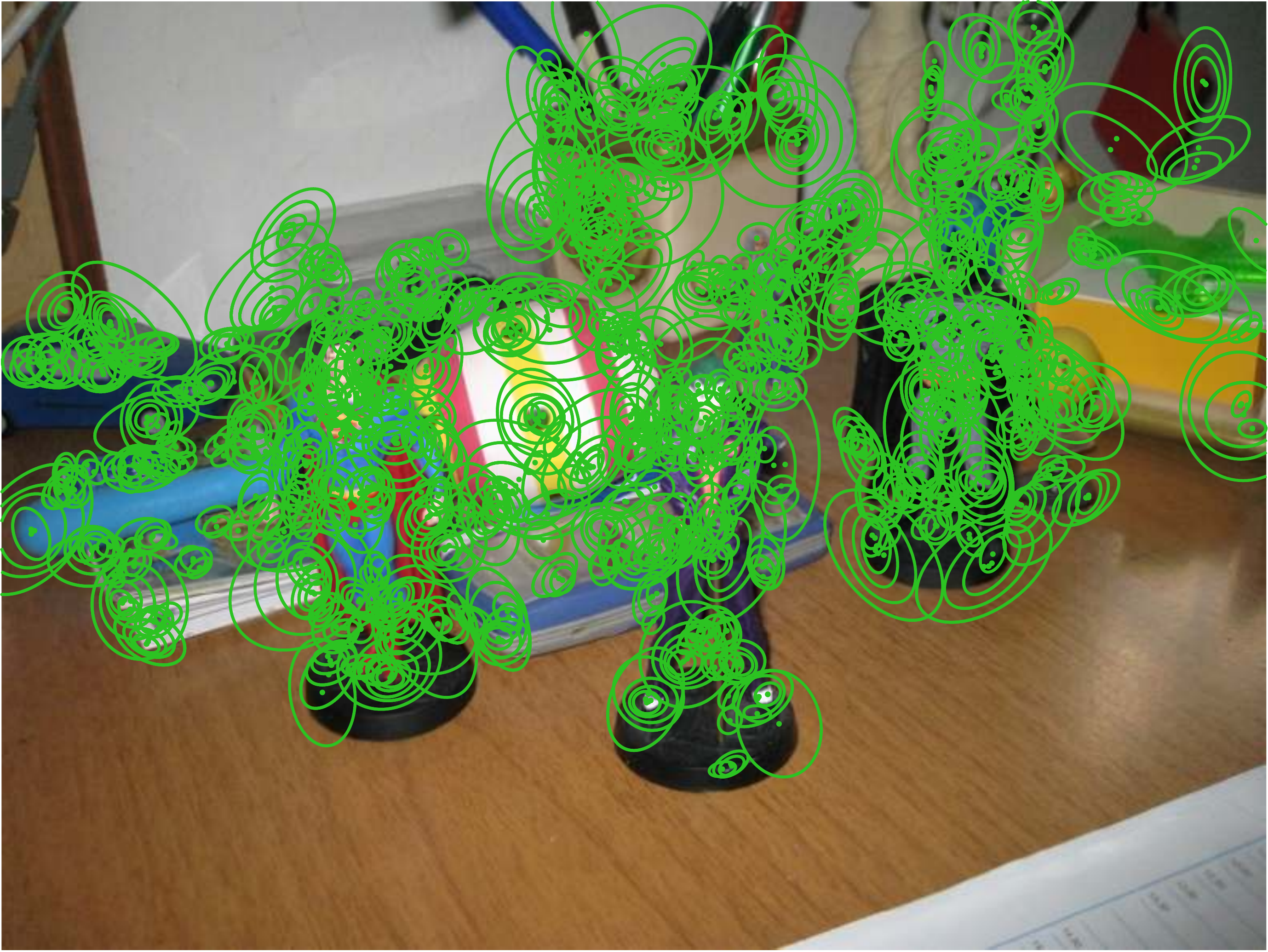}
	}
	\caption{\label{setup_imgs}
		Clusters of motion flows representing matches and keypoint support regions superimposed on planar and non-planar image pairs according to different ground-truth estimation methods when using SIFT (left column) and HarrisZ (right column). (first row) Method B (blue and light blue), method C (blue) and method D (blue and orange). (second row) Descriptor support regions for each keypoint match found in the previous row. The worst matches according to the reprojection error for each colored cluster are evidenced, the dashed ellipses are reprojected from the other image and the overlap error is computed by doubling the shape size. (third row) Method A (light blue and yellow) by setting the epipolar line error distance threshold to 7 pixels and method D (yellow and orange). (fourth row) Method B (yellow, red and light red), C (yellow and red) and D (yellow and orange). (fifth row) Descriptor support regions for each correspondence found by methods C and D (see Sec.~\ref{eval_setup}, best viewed in color and zoomed in).}
\end{figure}

\newpage
\section{DTM evaluation}\label{appendix_dtm}
Tables~\ref{table_all}-\ref{table_sosnet} report the results of the local spatial filters evaluation for the planar and non-planar datasets, without or with 1SAC applied as post-processing. The following statistics, averaged on the image pairs, are included: The precision, the recall, the number of correct and output matches, the number failures (i.e. the number of times the spatial filter did not provide at least one ground-truth match), and the running time. The recall is computed with respect to the ground-truth matches of the specific pipelined blob matching. The tables also include recall$^\star$, that considers overall the ground-truth matches among all the tested configurations. For the comparison, the former recall definition is more reasonable, since the local spatial filters cannot include matches outside those provided to them as input, i.e. the blob matching output. Nevertheless, recall$^\star$ can give clues of the whole pipeline performances. More detailed data can be found in the additional material.

\begin{figure*}
	\center
	\rotatebox[origin=l]{90}{\mbox{\hspace{3em}th}}
	\includegraphics[height=7.5em]{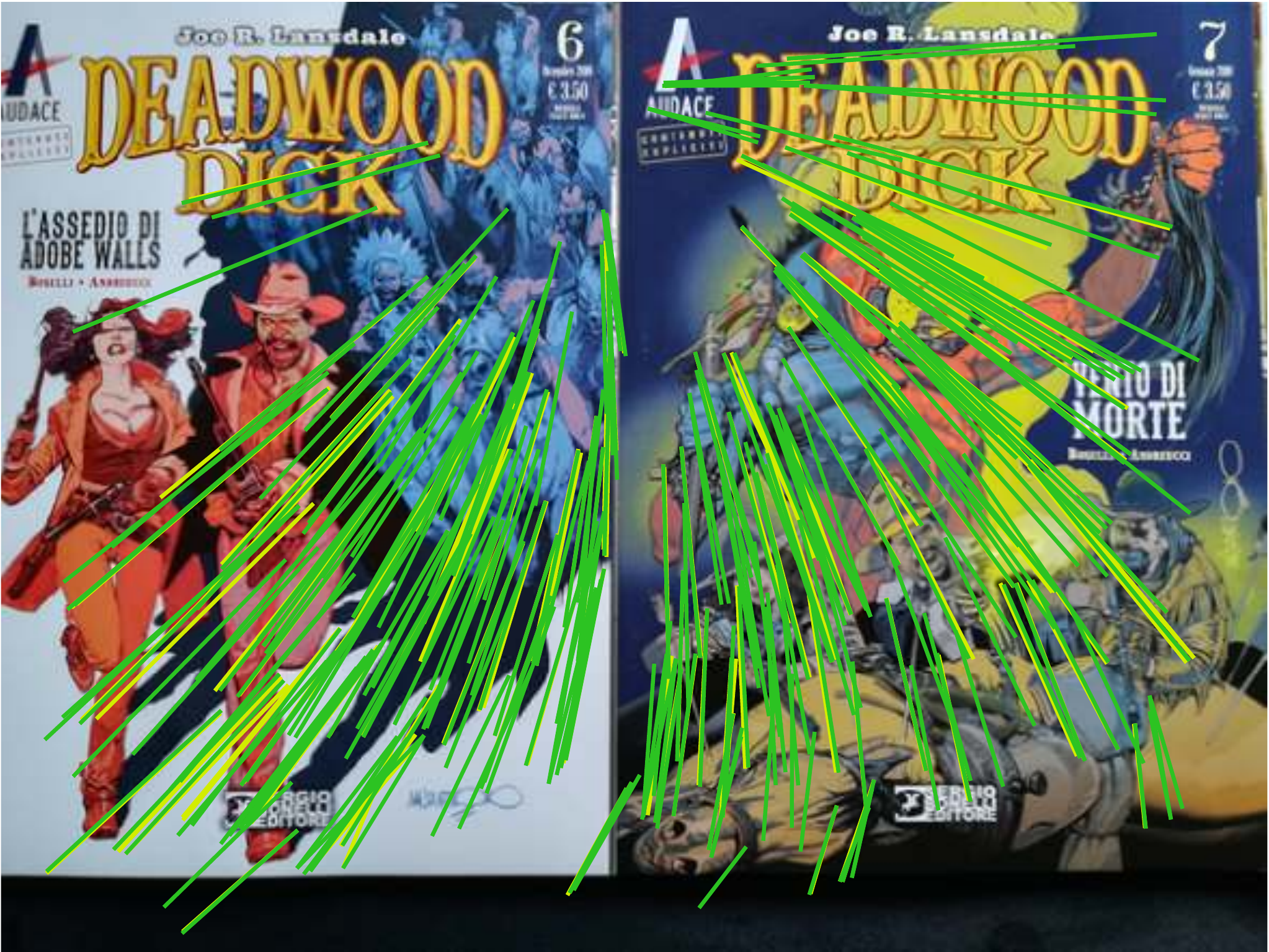}
	\includegraphics[height=7.5em]{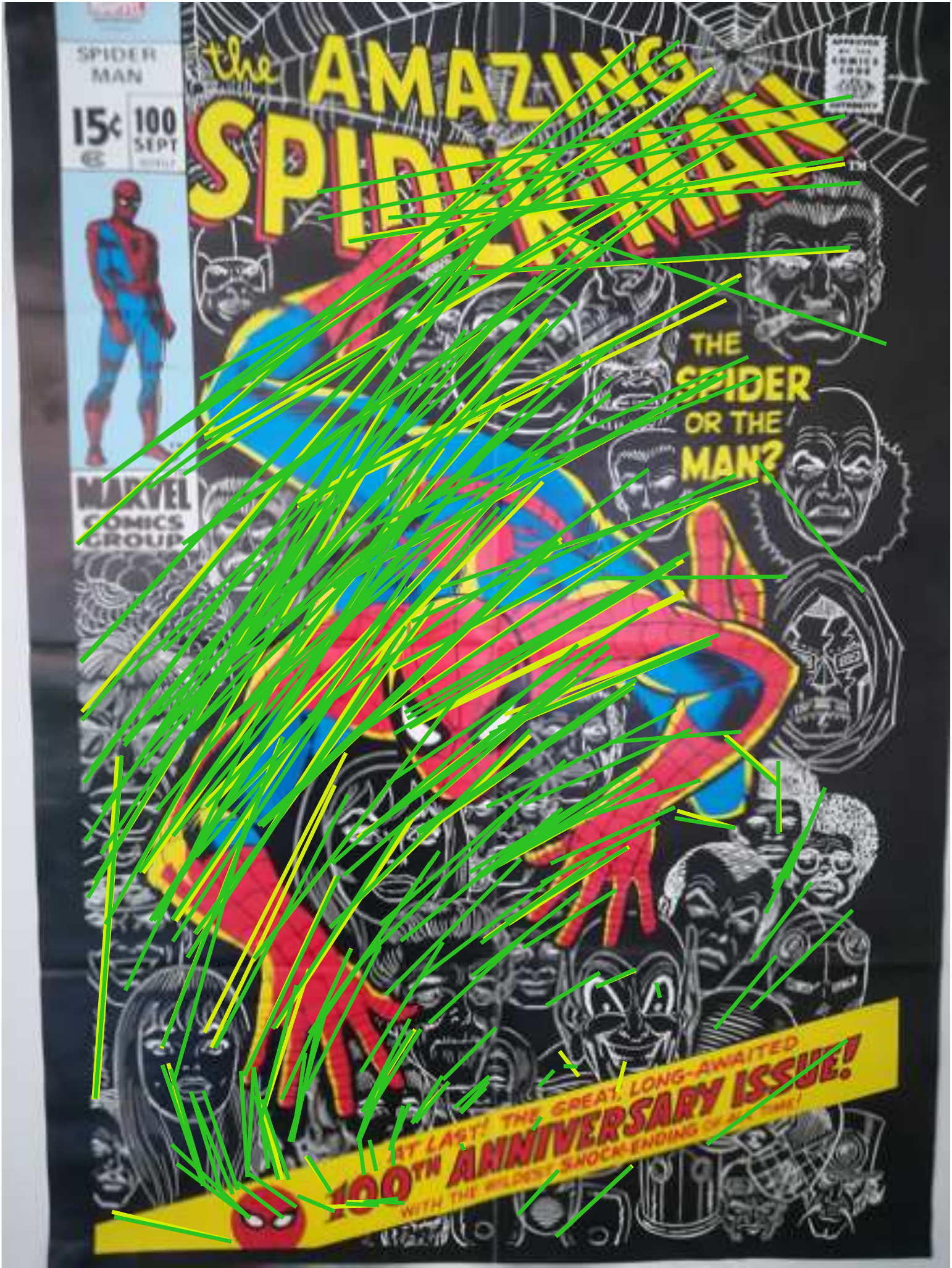}
	\includegraphics[height=7.5em]{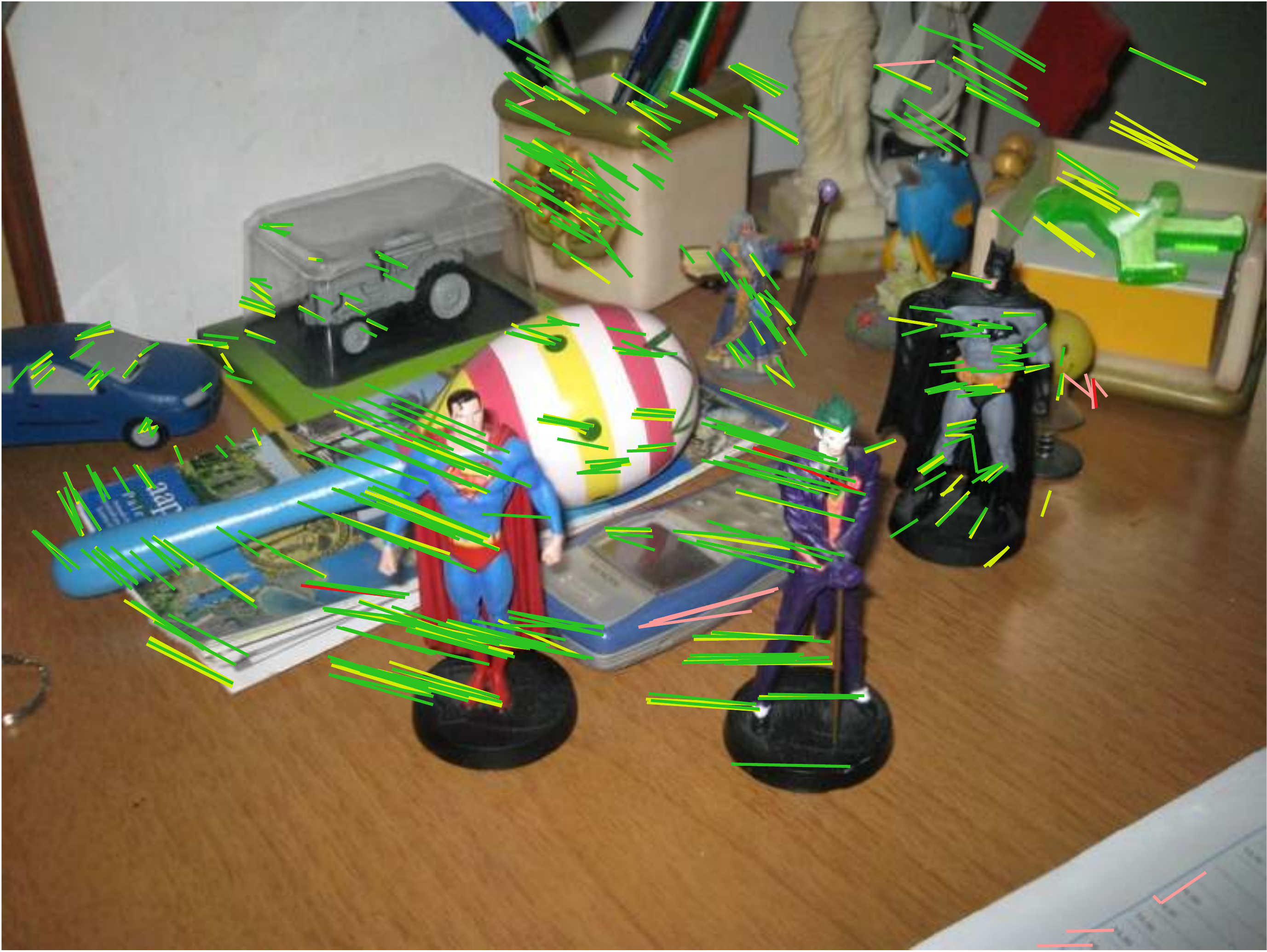}
	\includegraphics[height=7.5em]{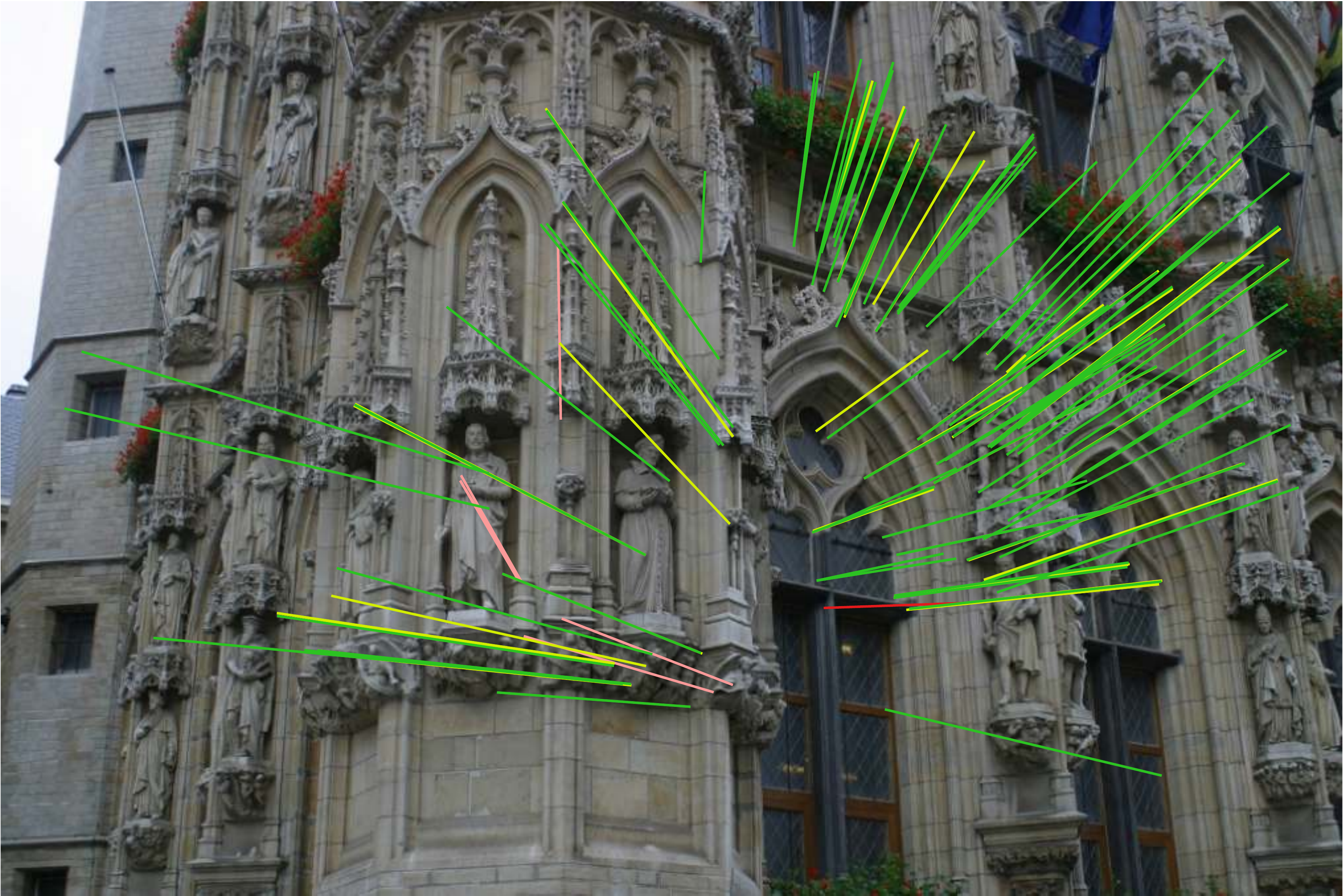}
	\includegraphics[height=7.5em]{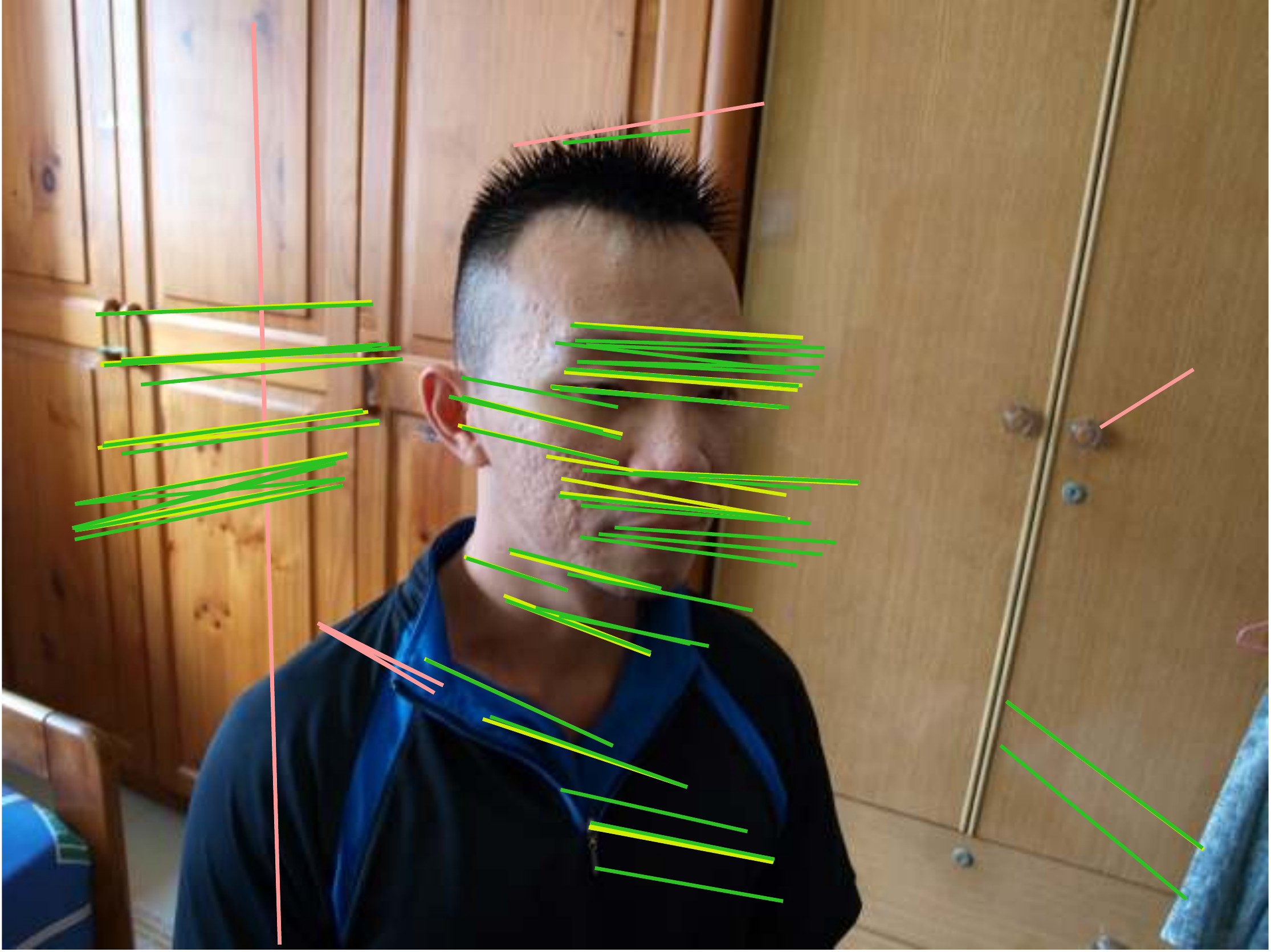}
	\\
	\vspace{0.5em}
	\rotatebox[origin=l]{90}{\mbox{\hspace{2em}DTM}}
	\includegraphics[height=7.5em]{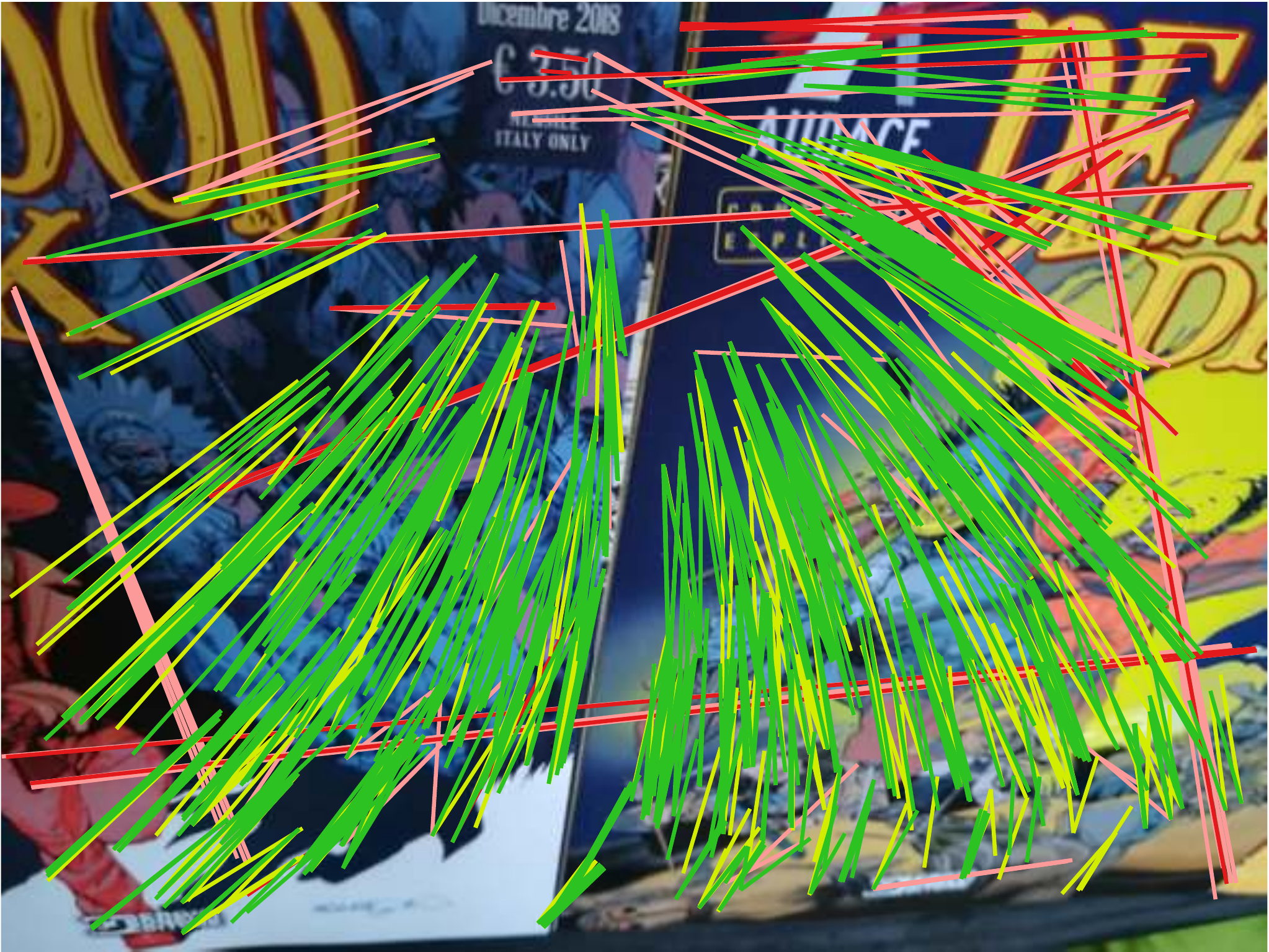}
	\includegraphics[height=7.5em]{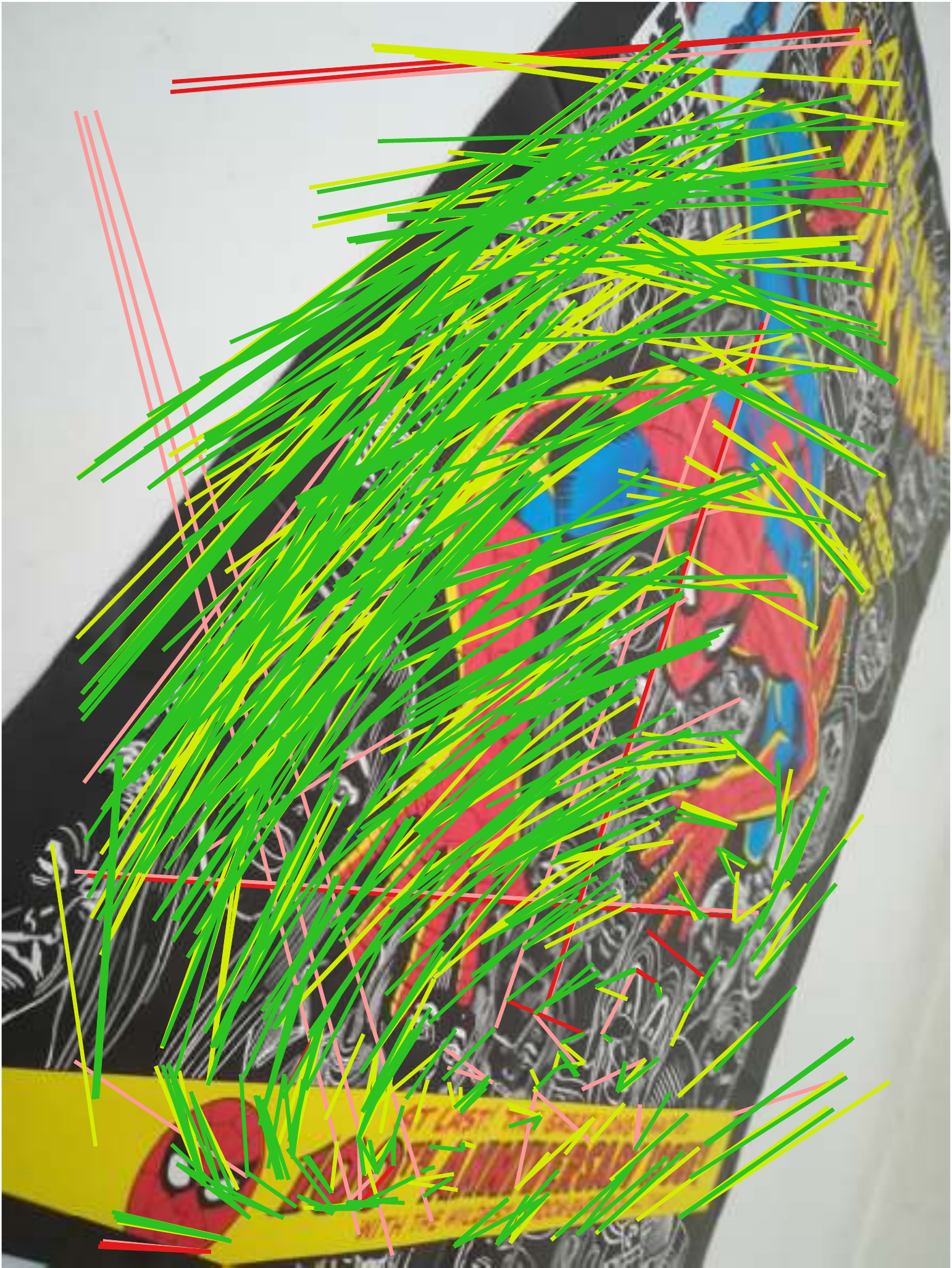}
	\includegraphics[height=7.5em]{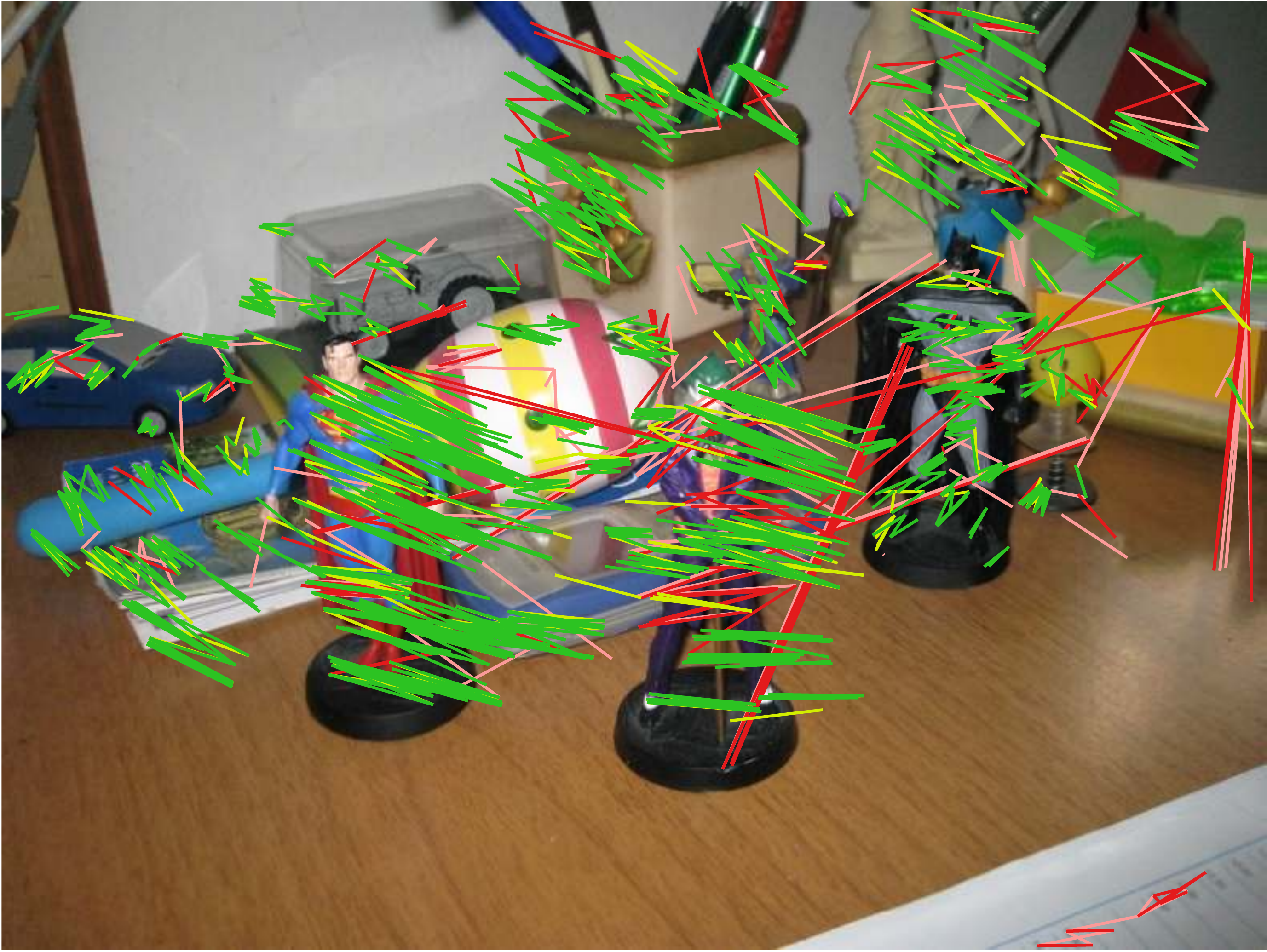}
	\includegraphics[height=7.5em]{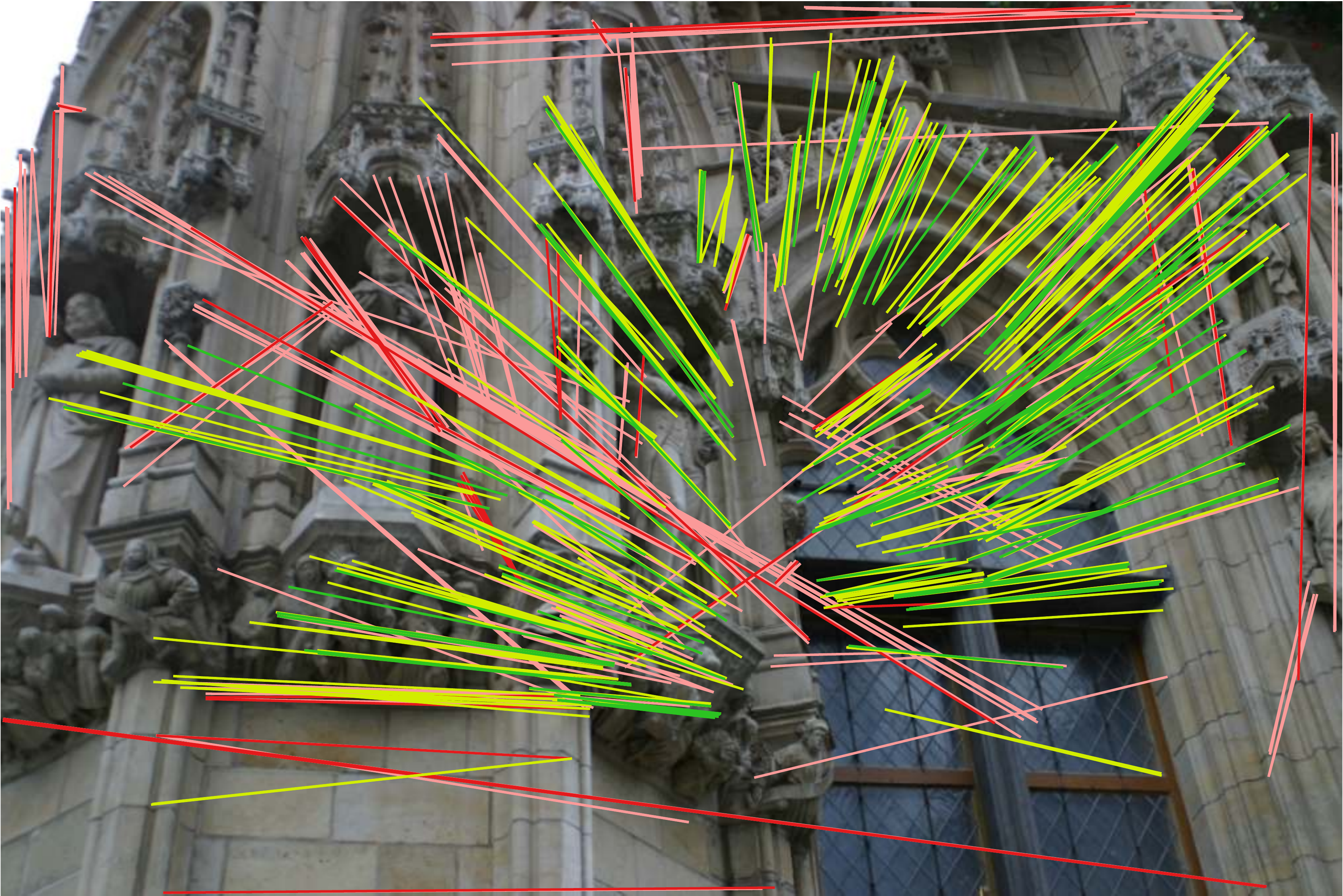}
	\includegraphics[height=7.5em]{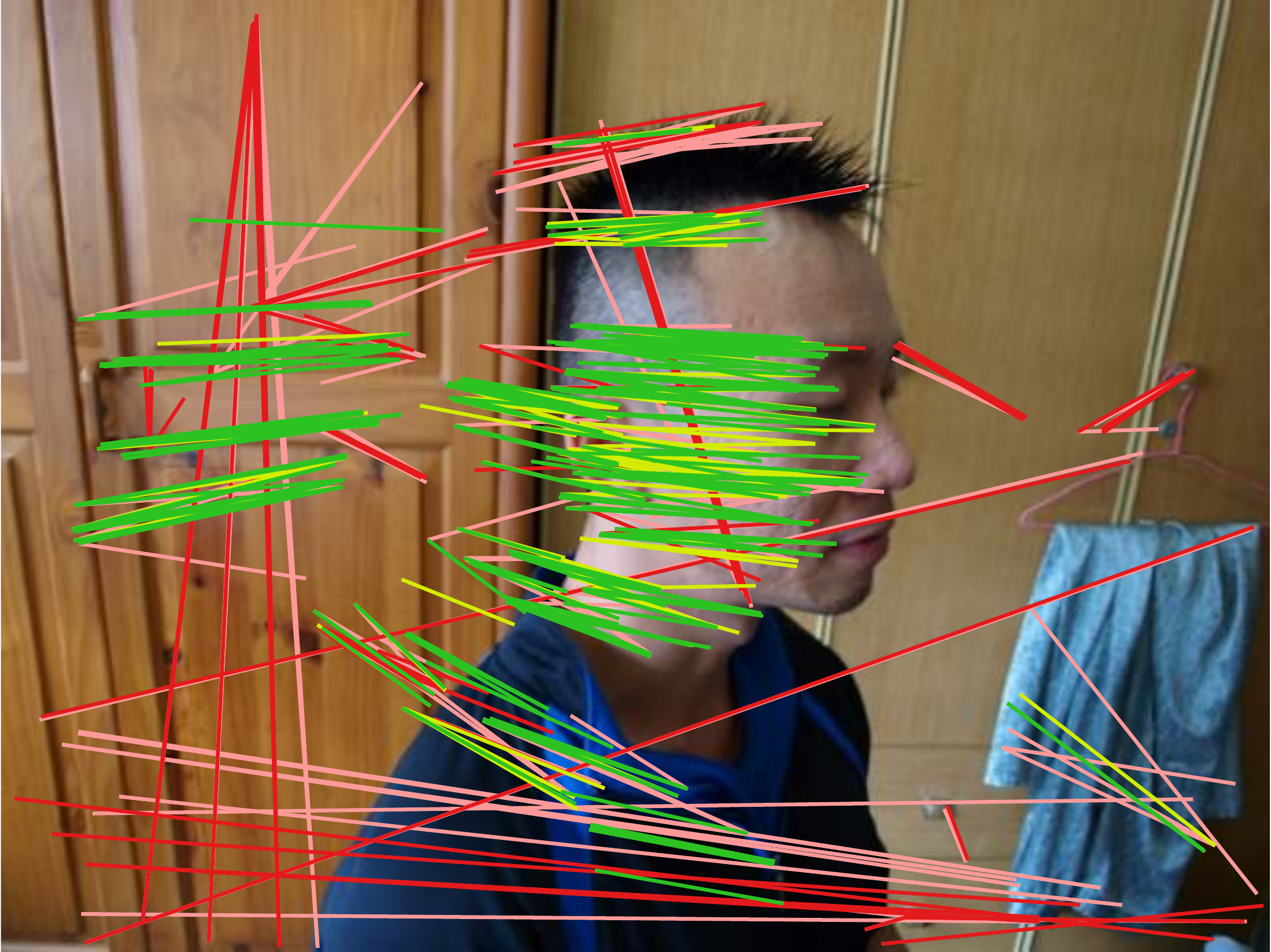}
	\\
	\vspace{0.5em}
	\rotatebox[origin=l]{90}{\mbox{\hspace{2em}LMR}}
	\includegraphics[height=7.5em]{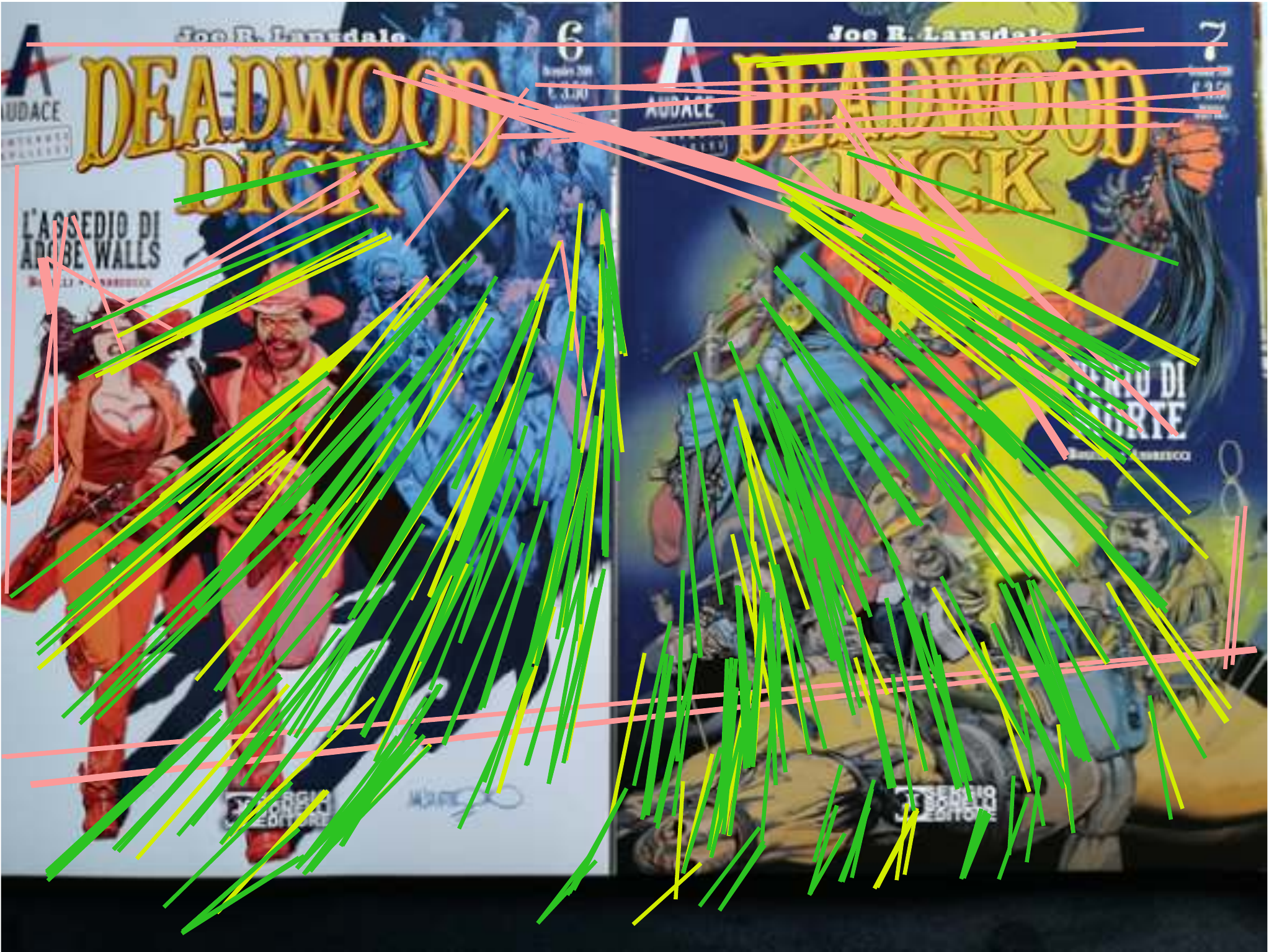}
	\includegraphics[height=7.5em]{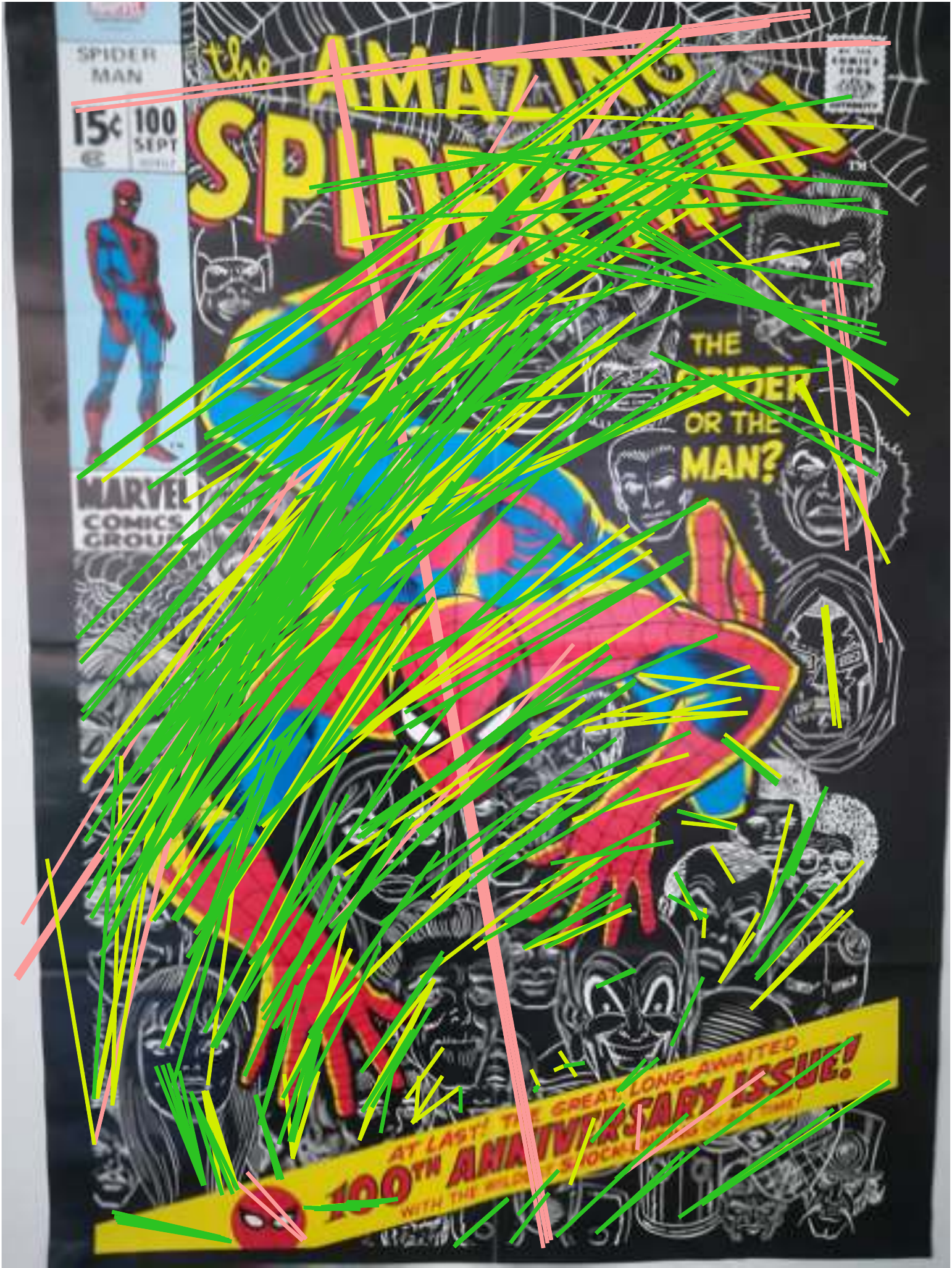}
	\includegraphics[height=7.5em]{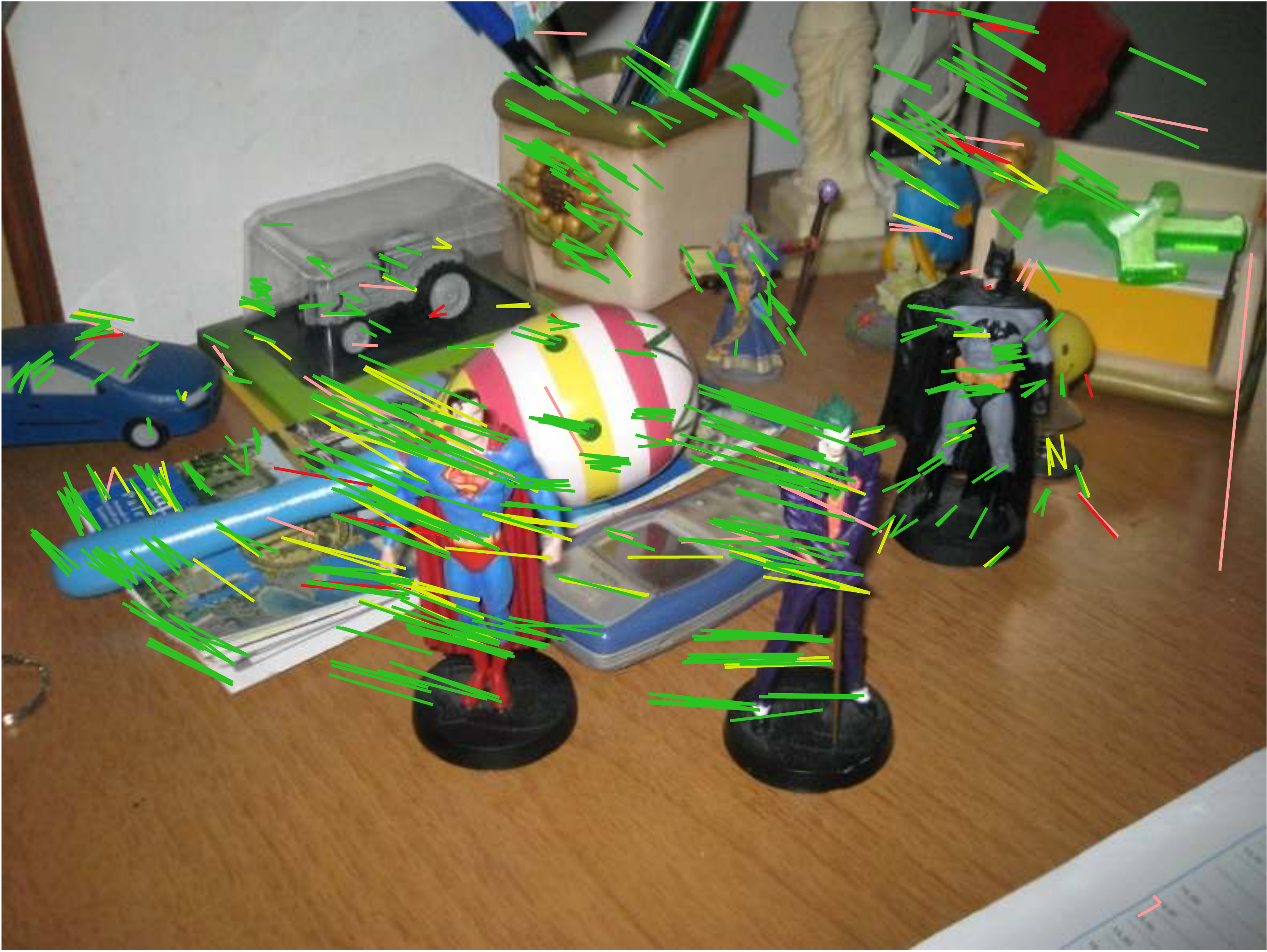}
	\includegraphics[height=7.5em]{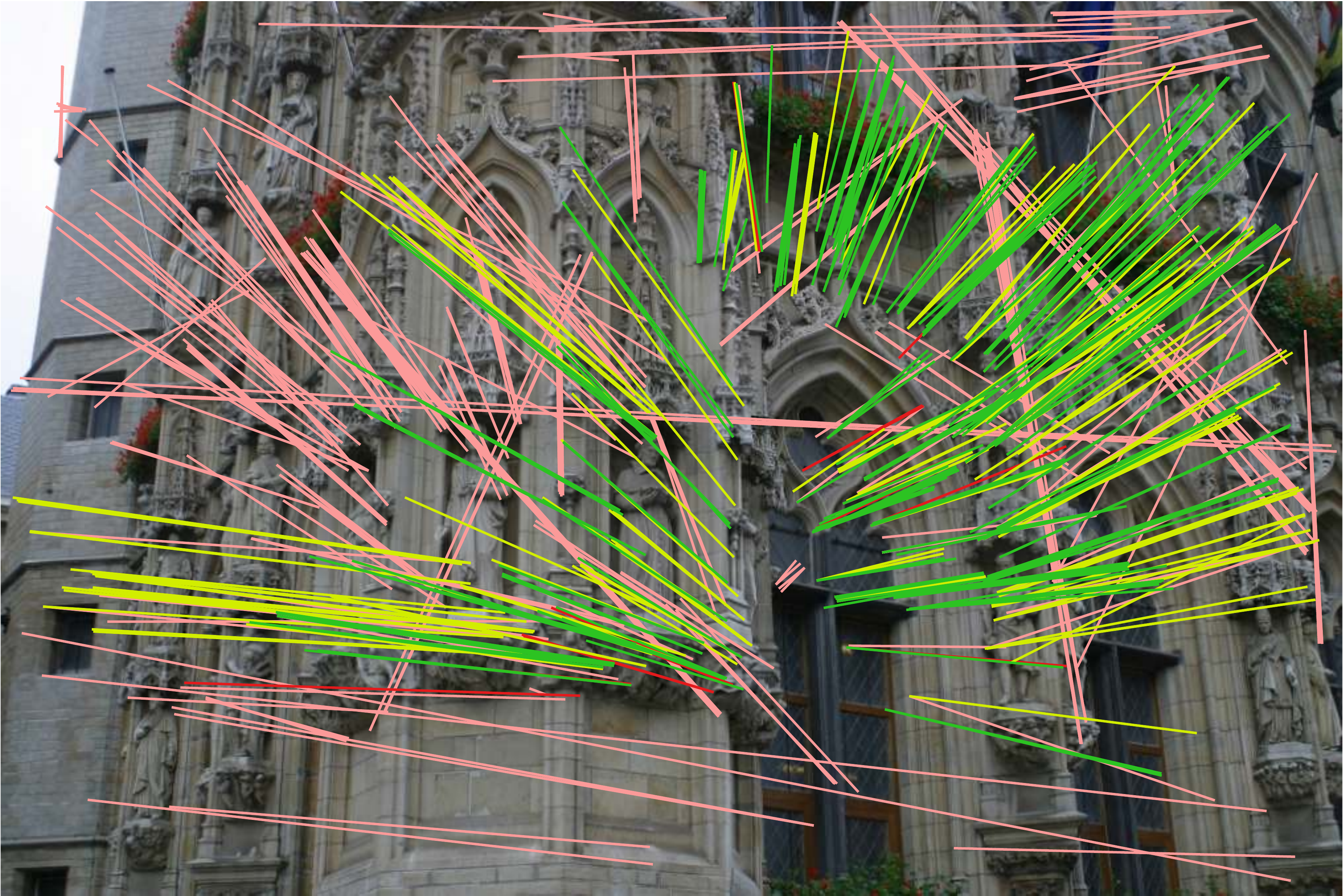}
	\includegraphics[height=7.5em]{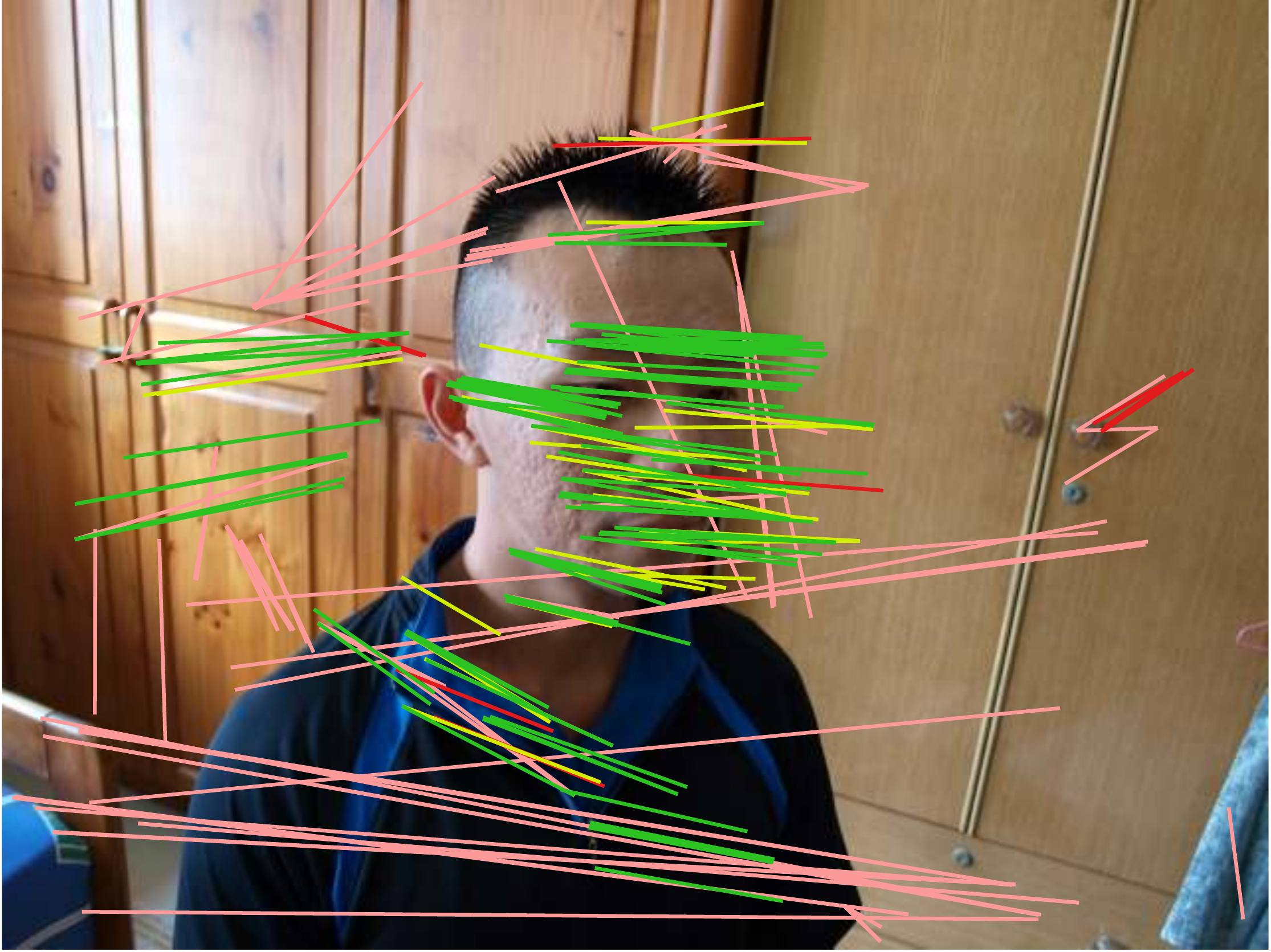}
	\\
	\vspace{0.5em}
	\rotatebox[origin=l]{90}{\mbox{\hspace{2em}LPM}}
	\includegraphics[height=7.5em]{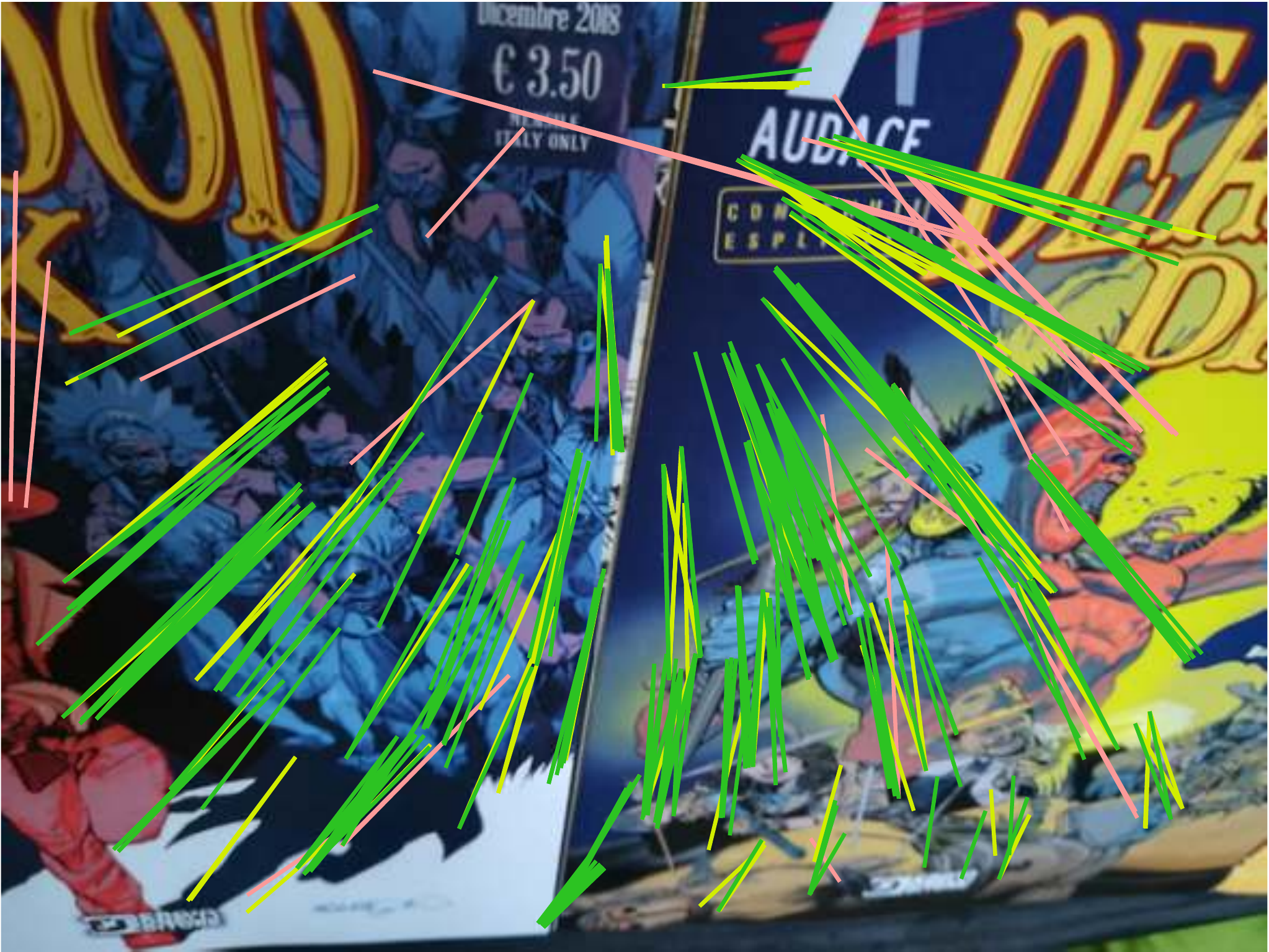}
	\includegraphics[height=7.5em]{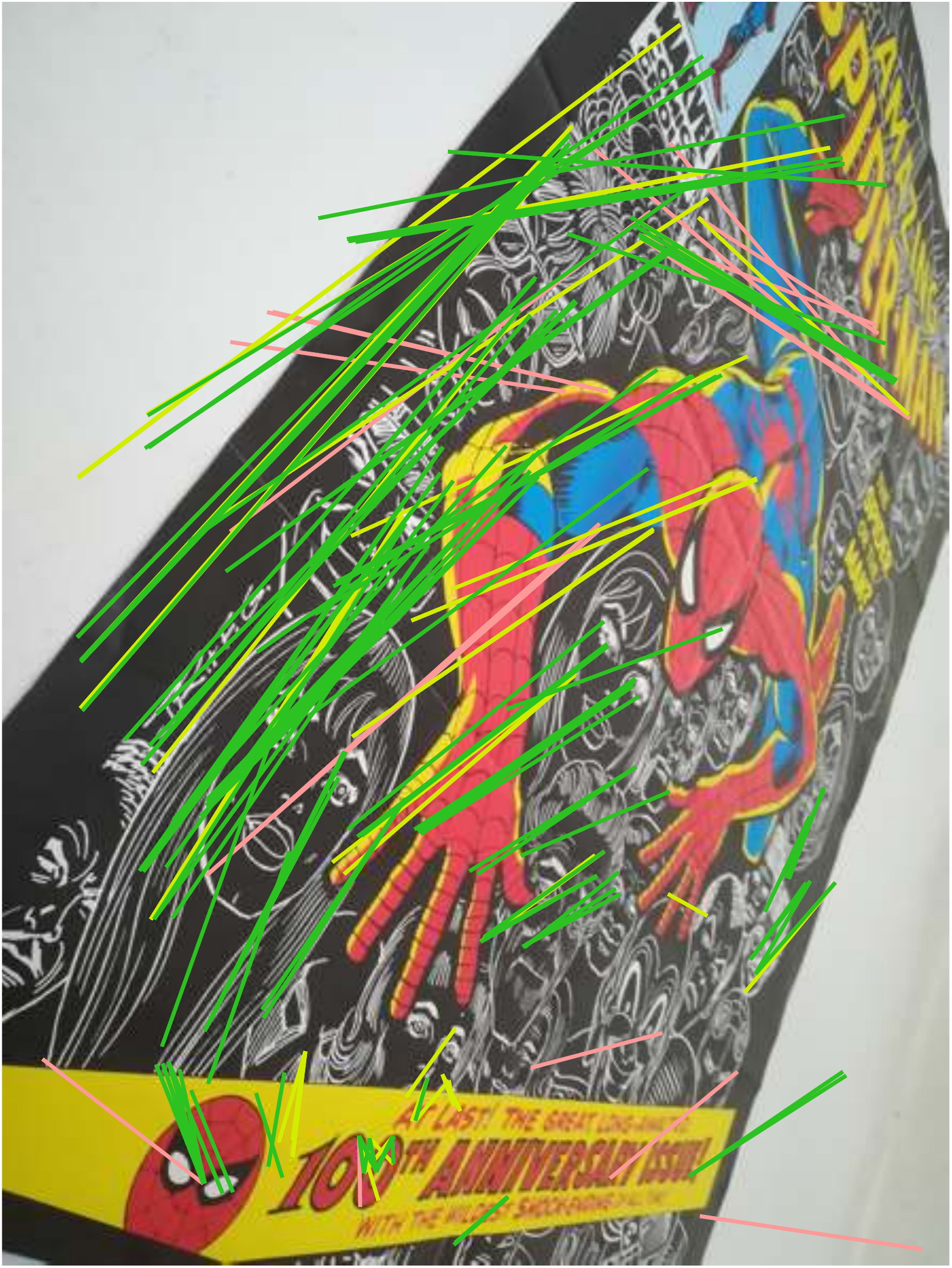}
	\includegraphics[height=7.5em]{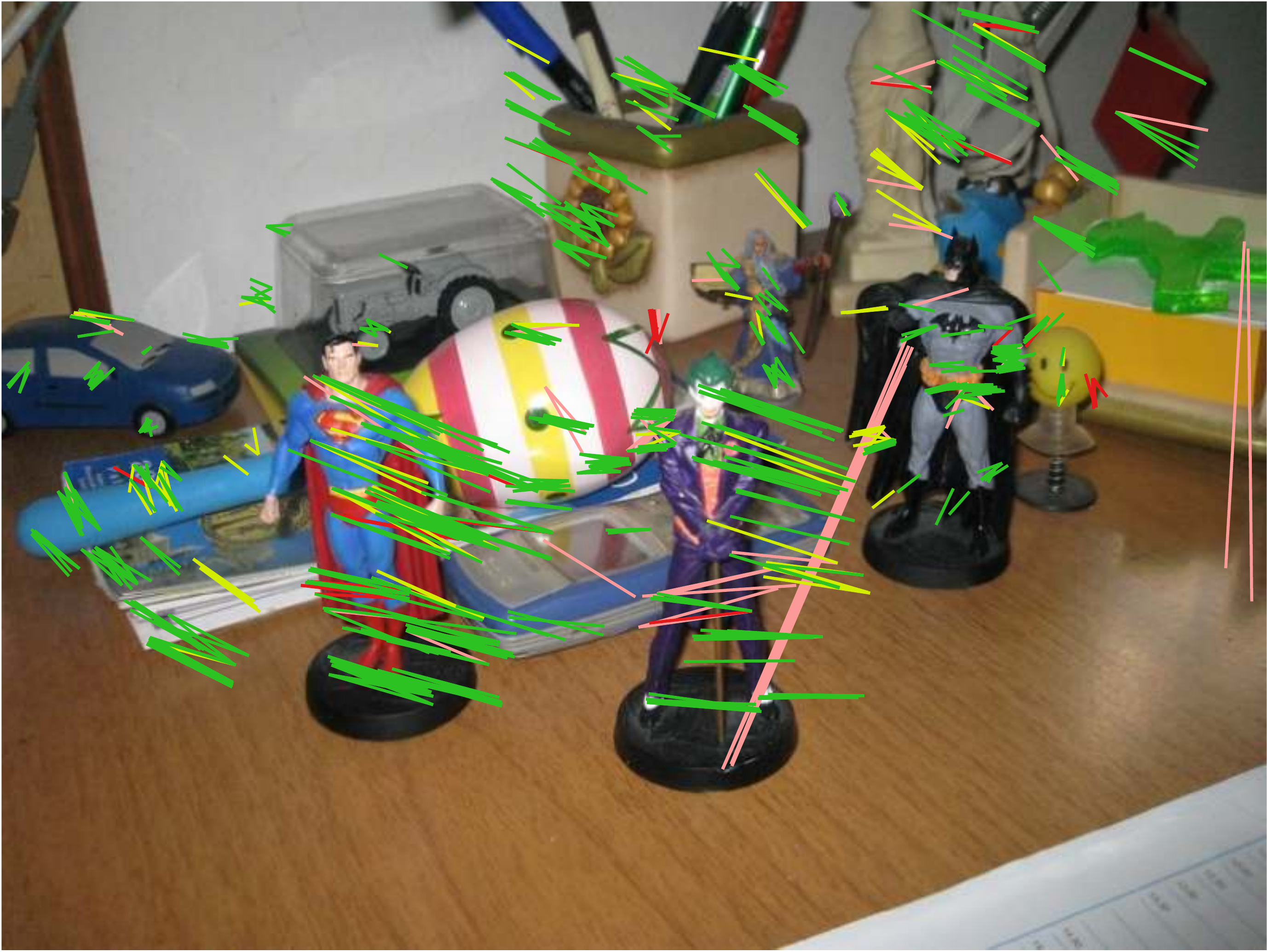}
	\includegraphics[height=7.5em]{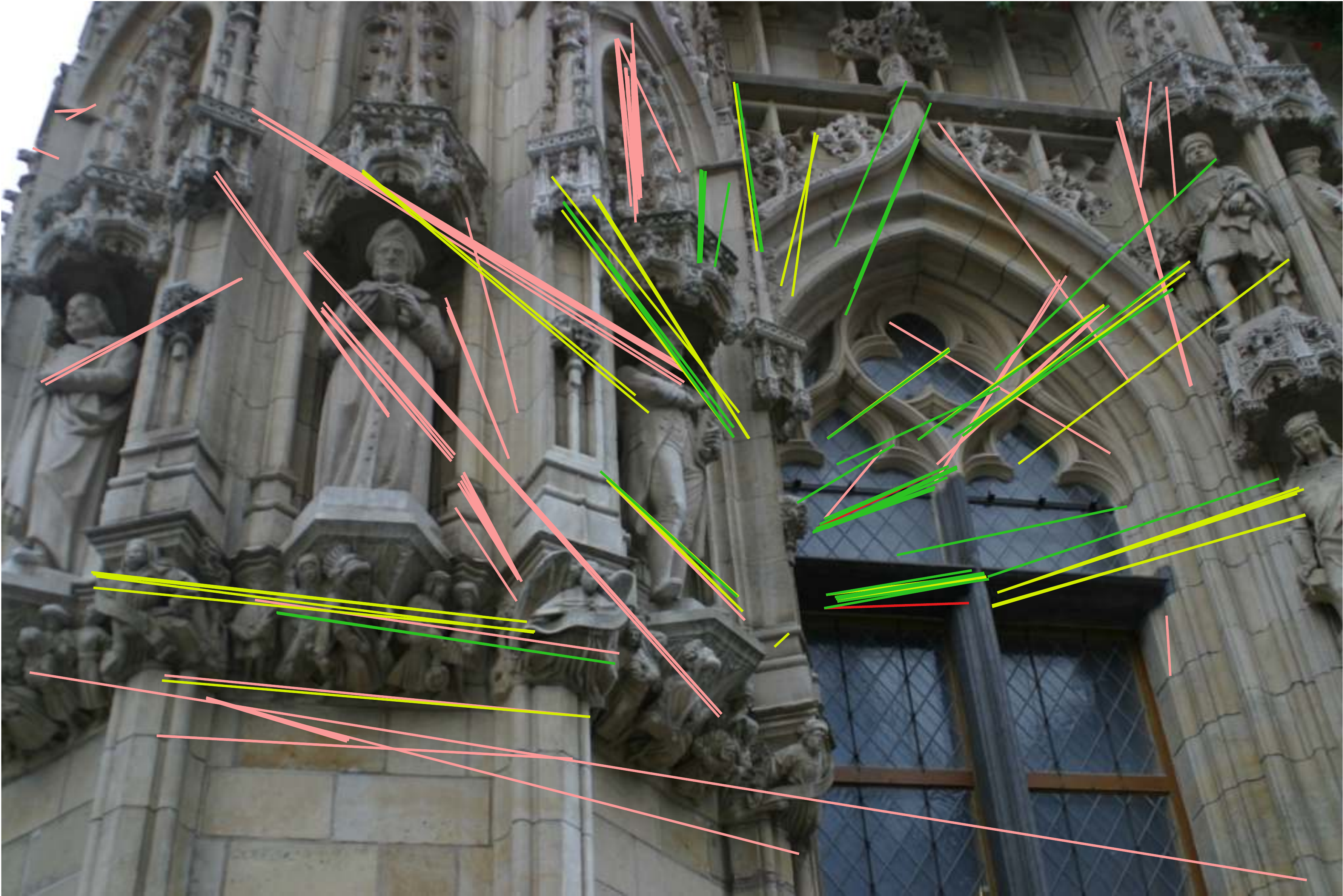}
	\includegraphics[height=7.5em]{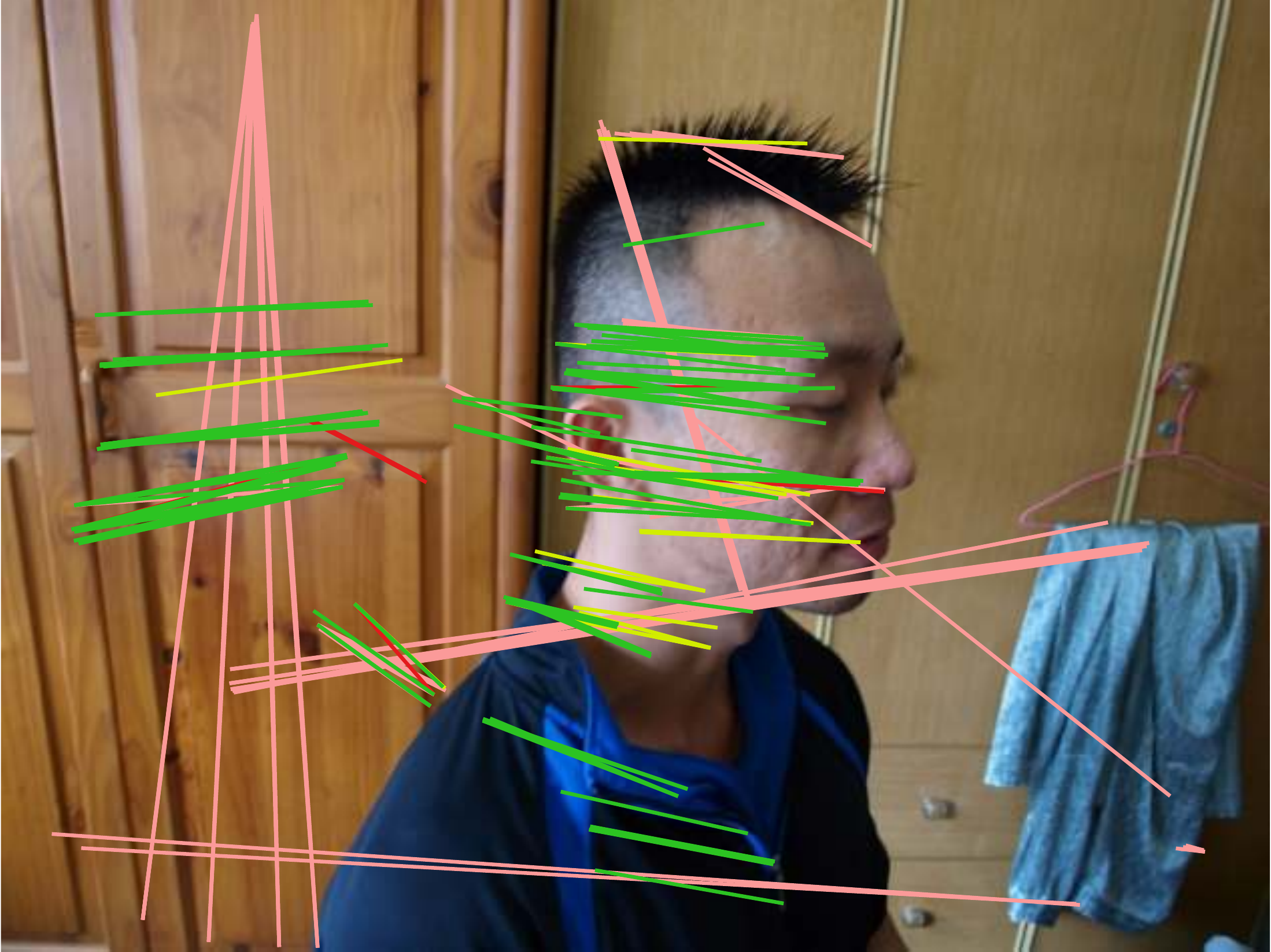}
	\\
	\vspace{0.5em}
	\rotatebox[origin=l]{90}{\mbox{\hspace{2em}GLPM}}
	\includegraphics[height=7.5em]{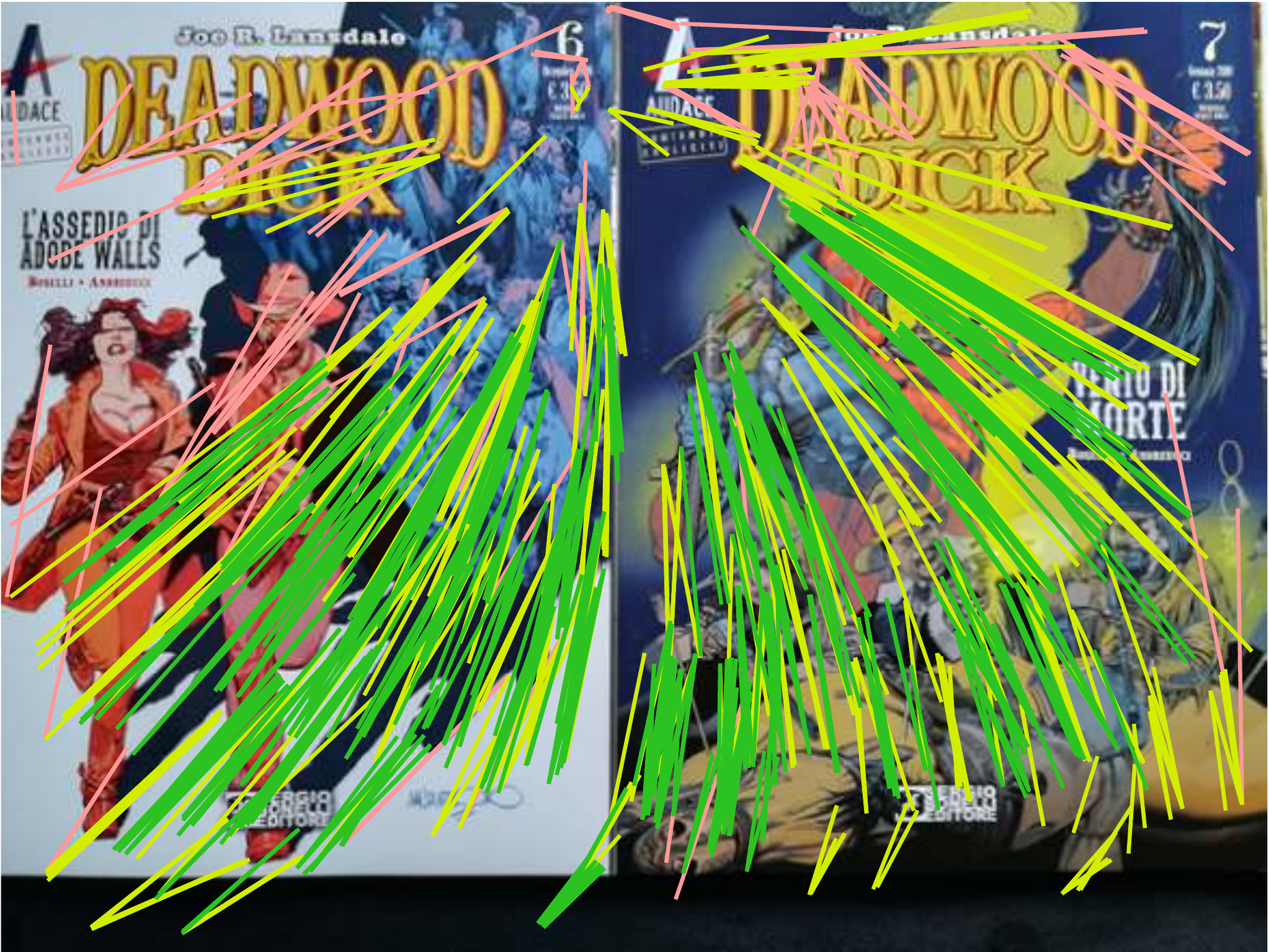}
	\includegraphics[height=7.5em]{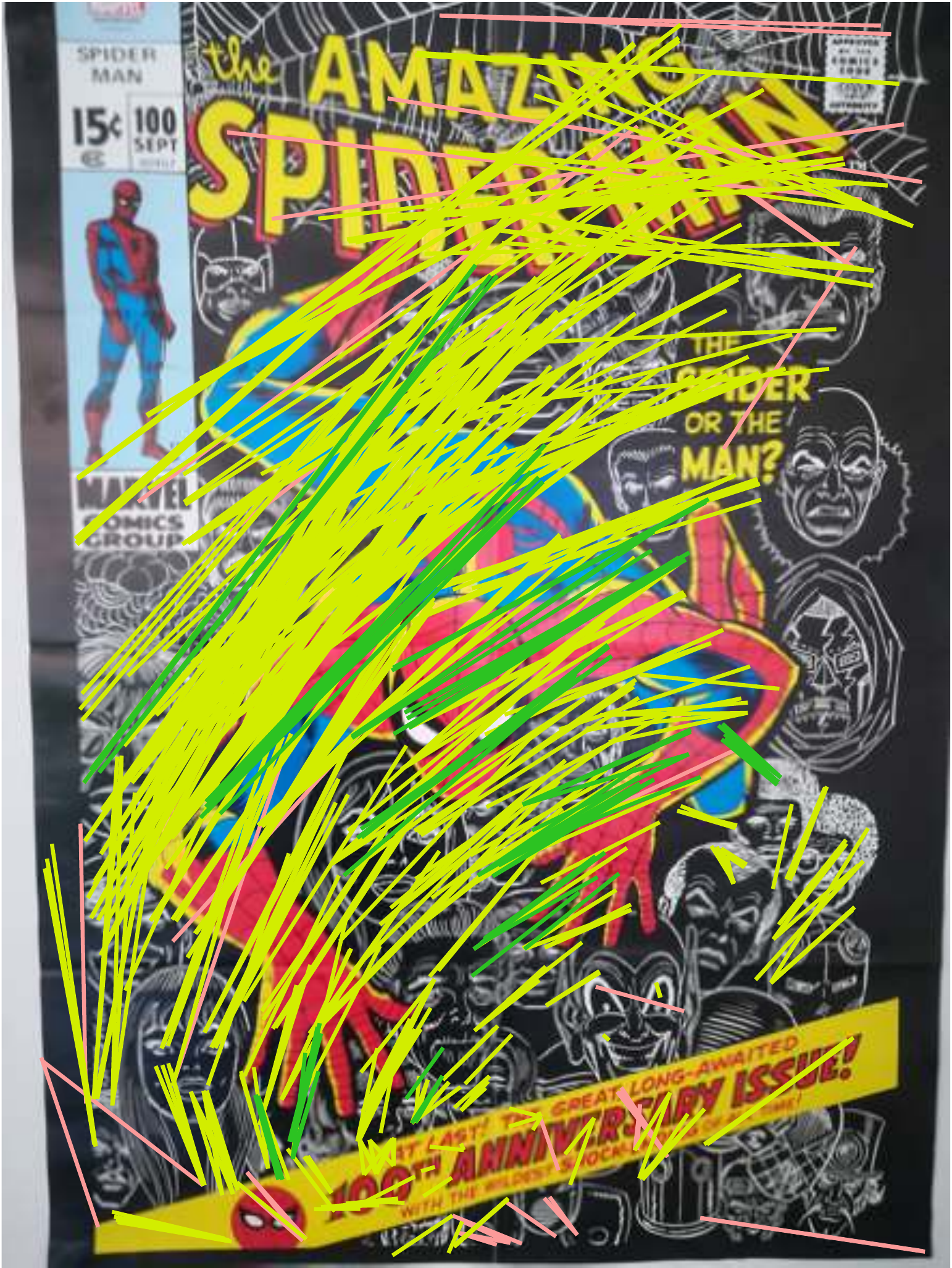}
	\includegraphics[height=7.5em]{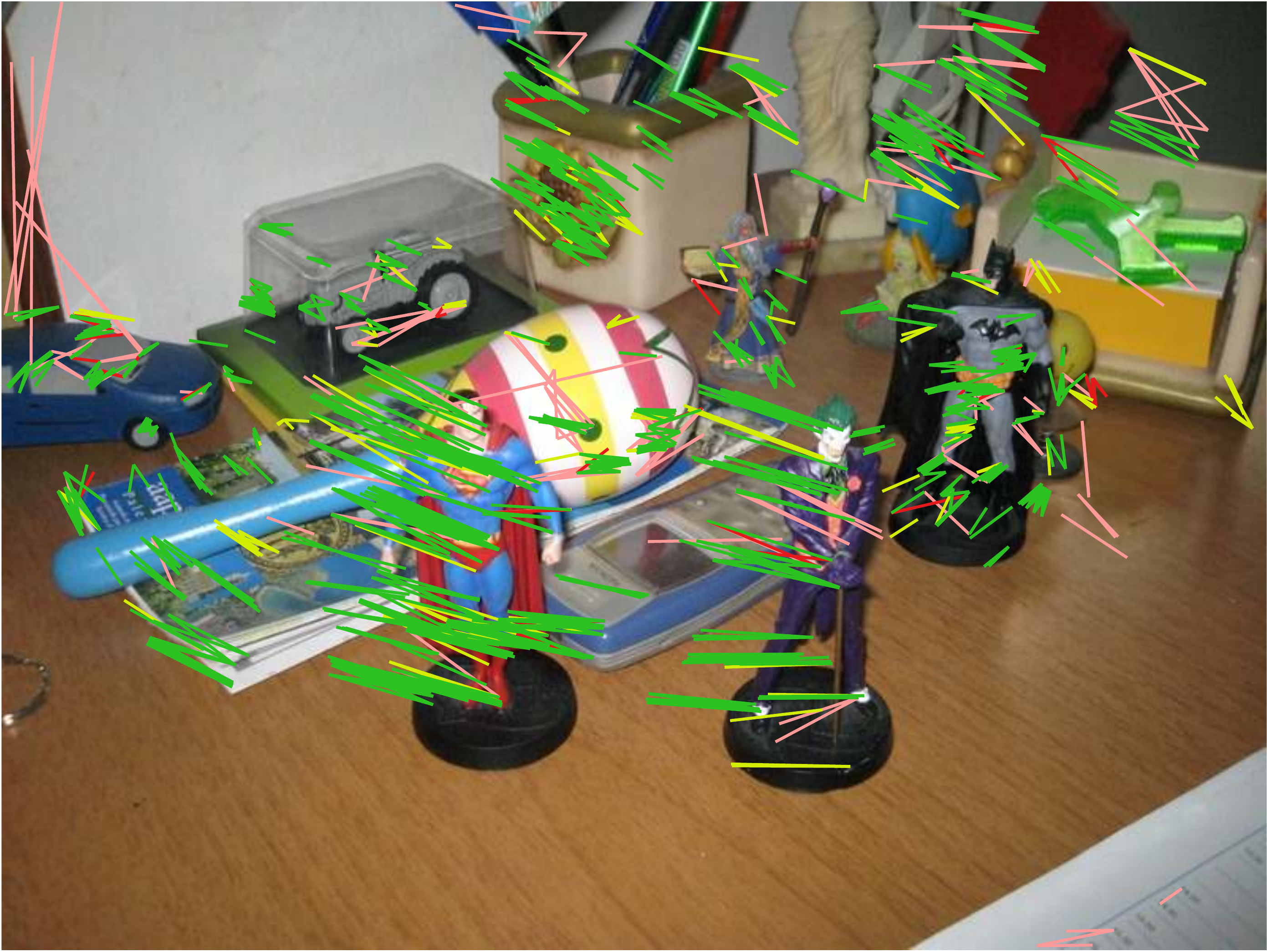}
	\includegraphics[height=7.5em]{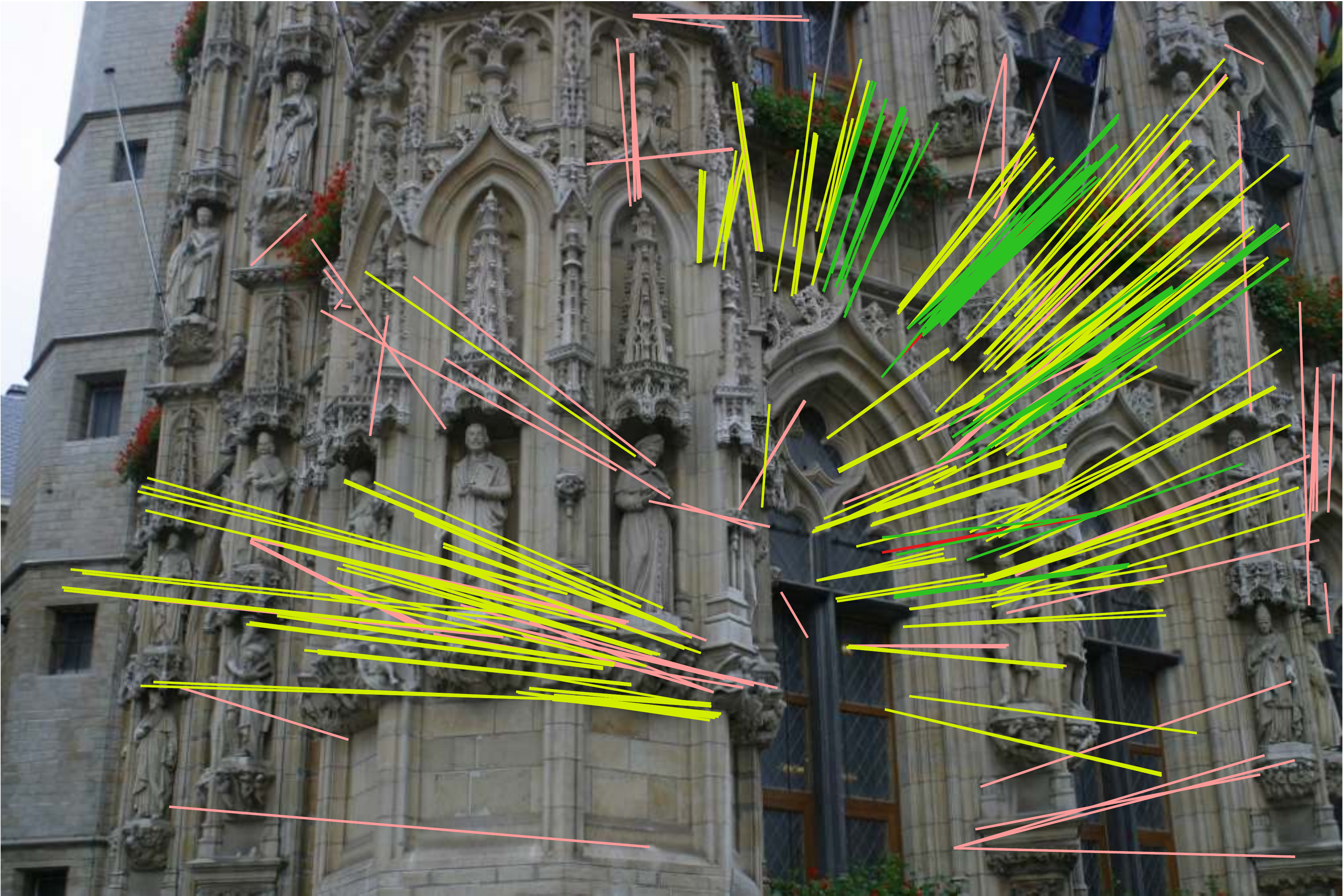}
	\includegraphics[height=7.5em]{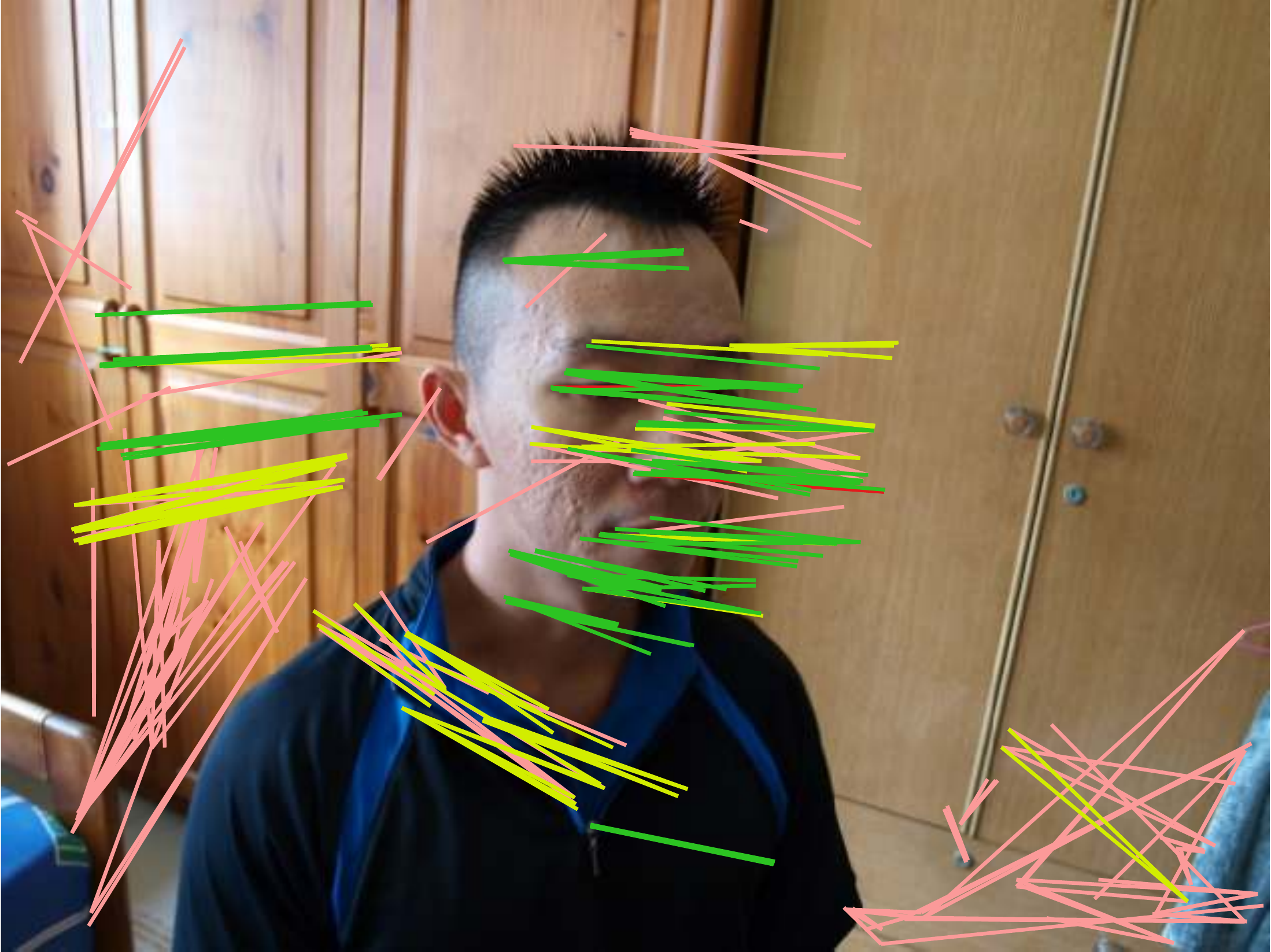}
	\\
	\vspace{0.5em}
	\rotatebox[origin=l]{90}{\mbox{\hspace{2em}GMS}}
	\includegraphics[height=7.5em]{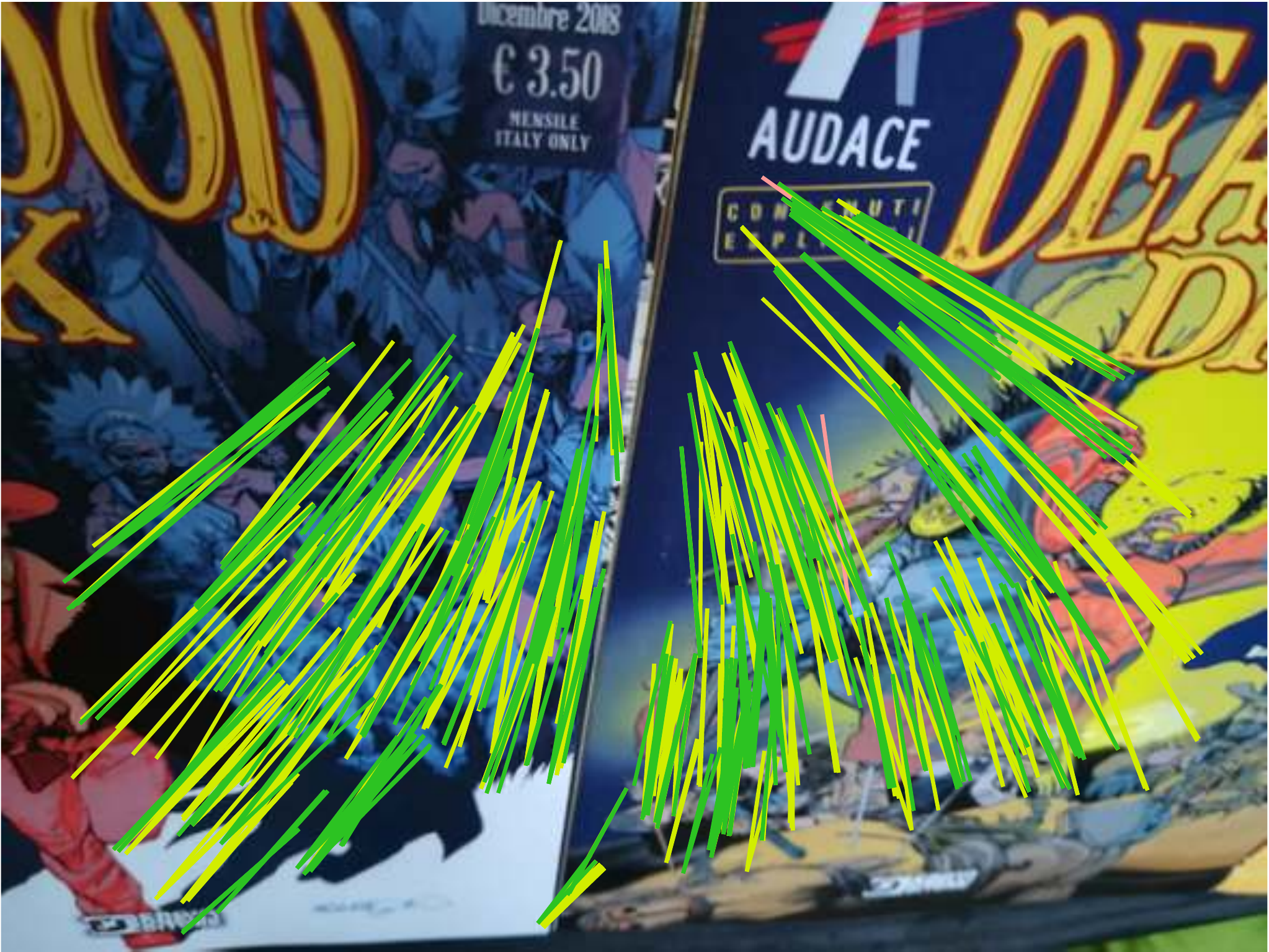}
	\includegraphics[height=7.5em]{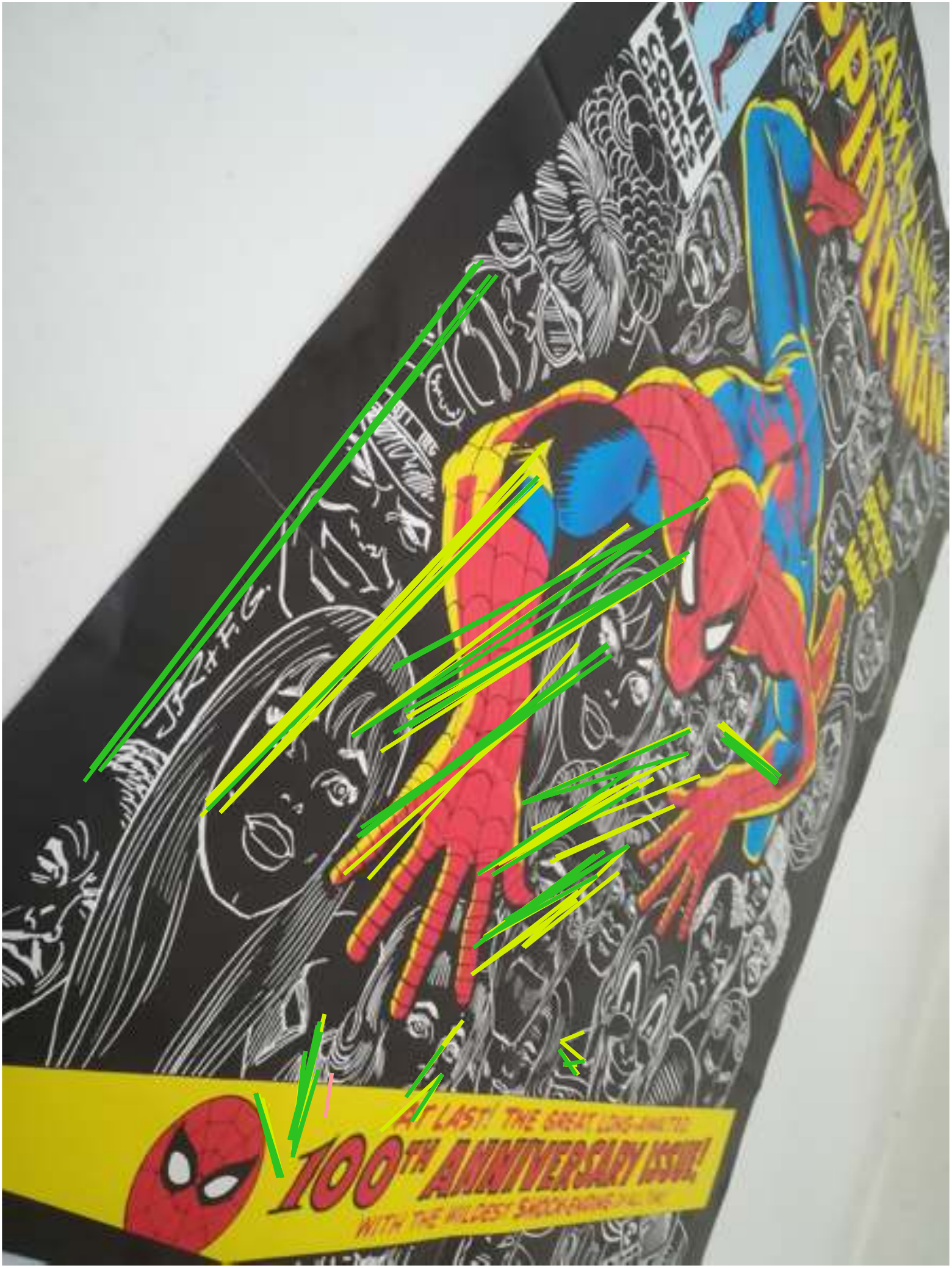}
	\includegraphics[height=7.5em]{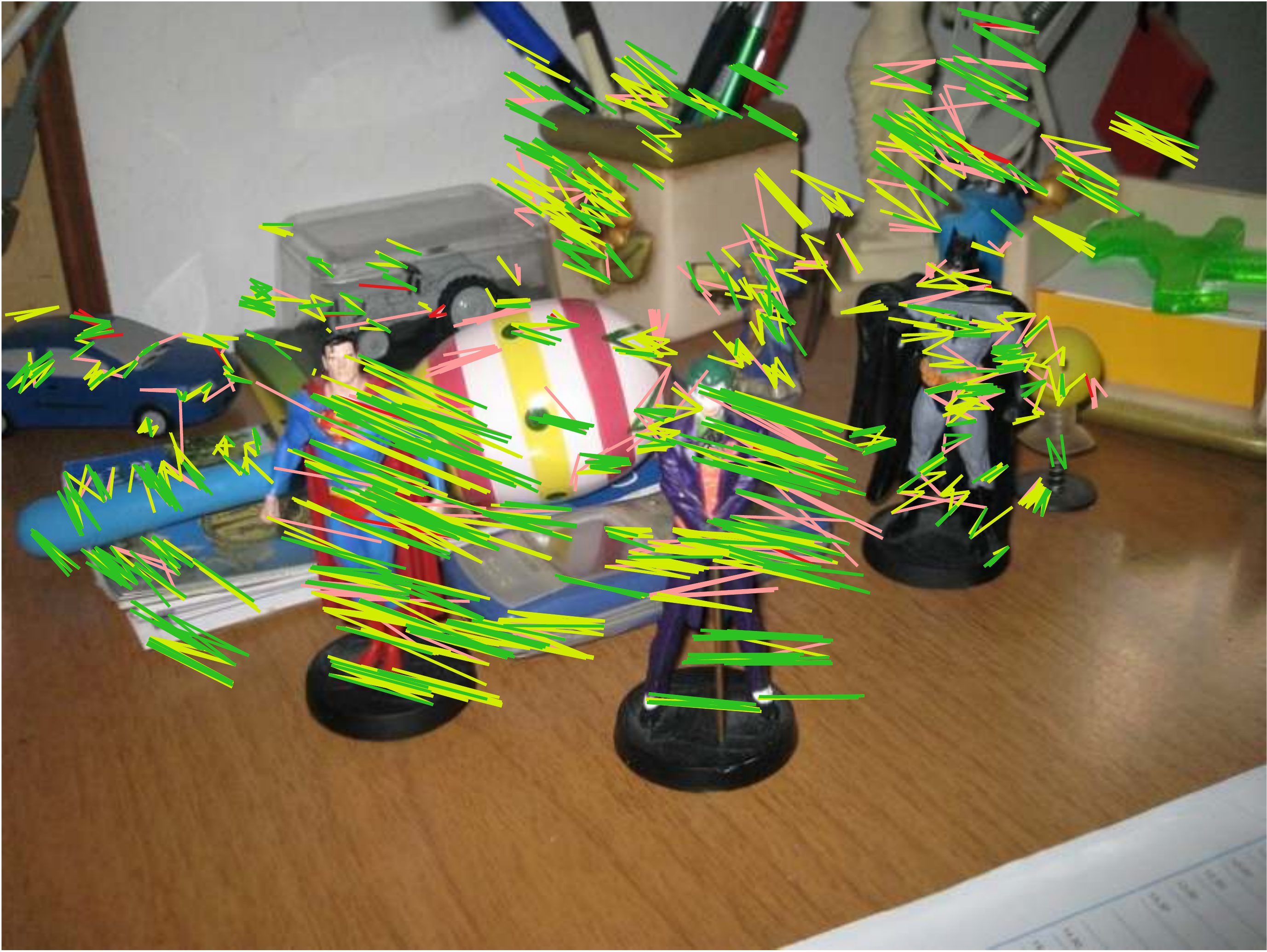}
	\includegraphics[height=7.5em]{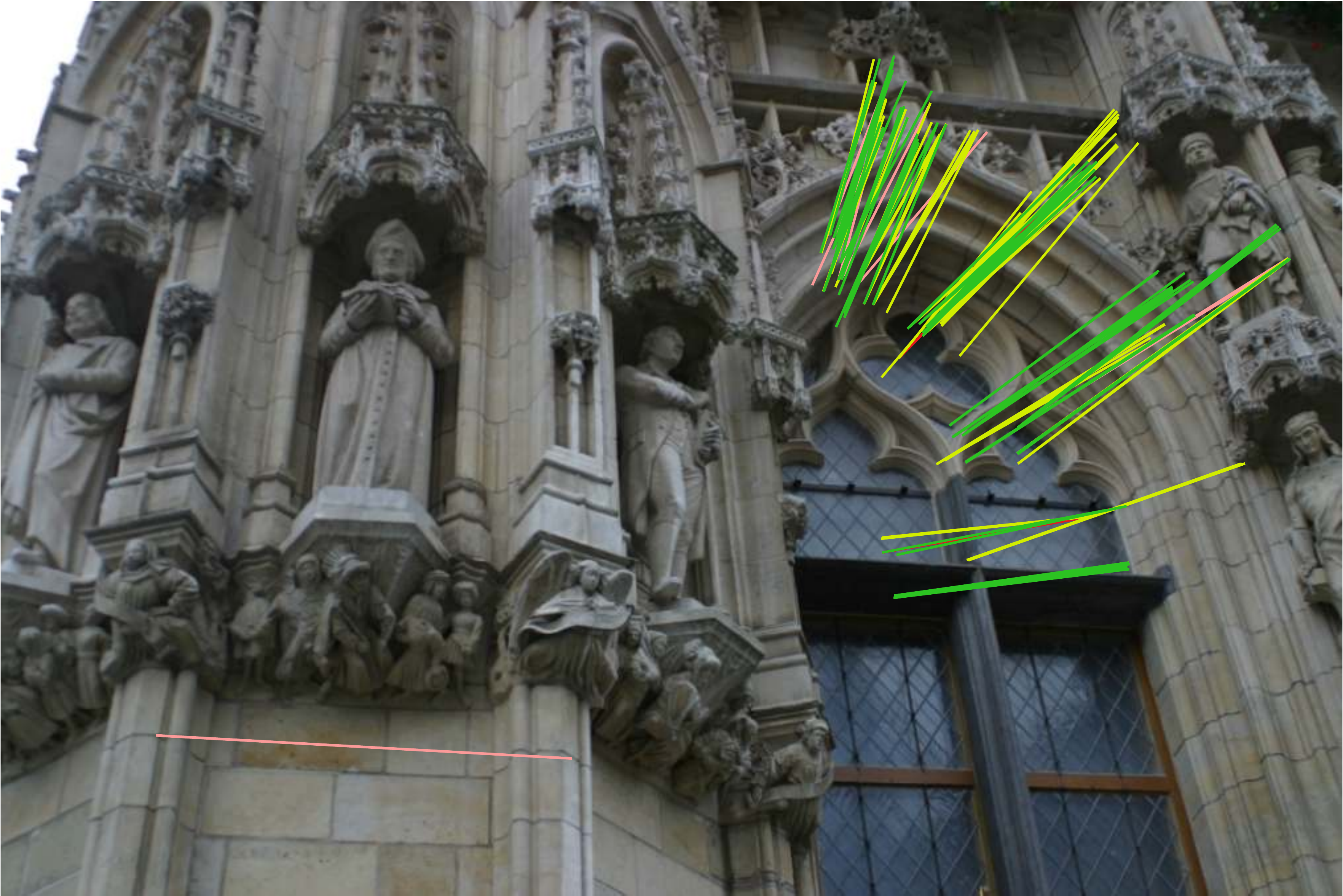}
	\includegraphics[height=7.5em]{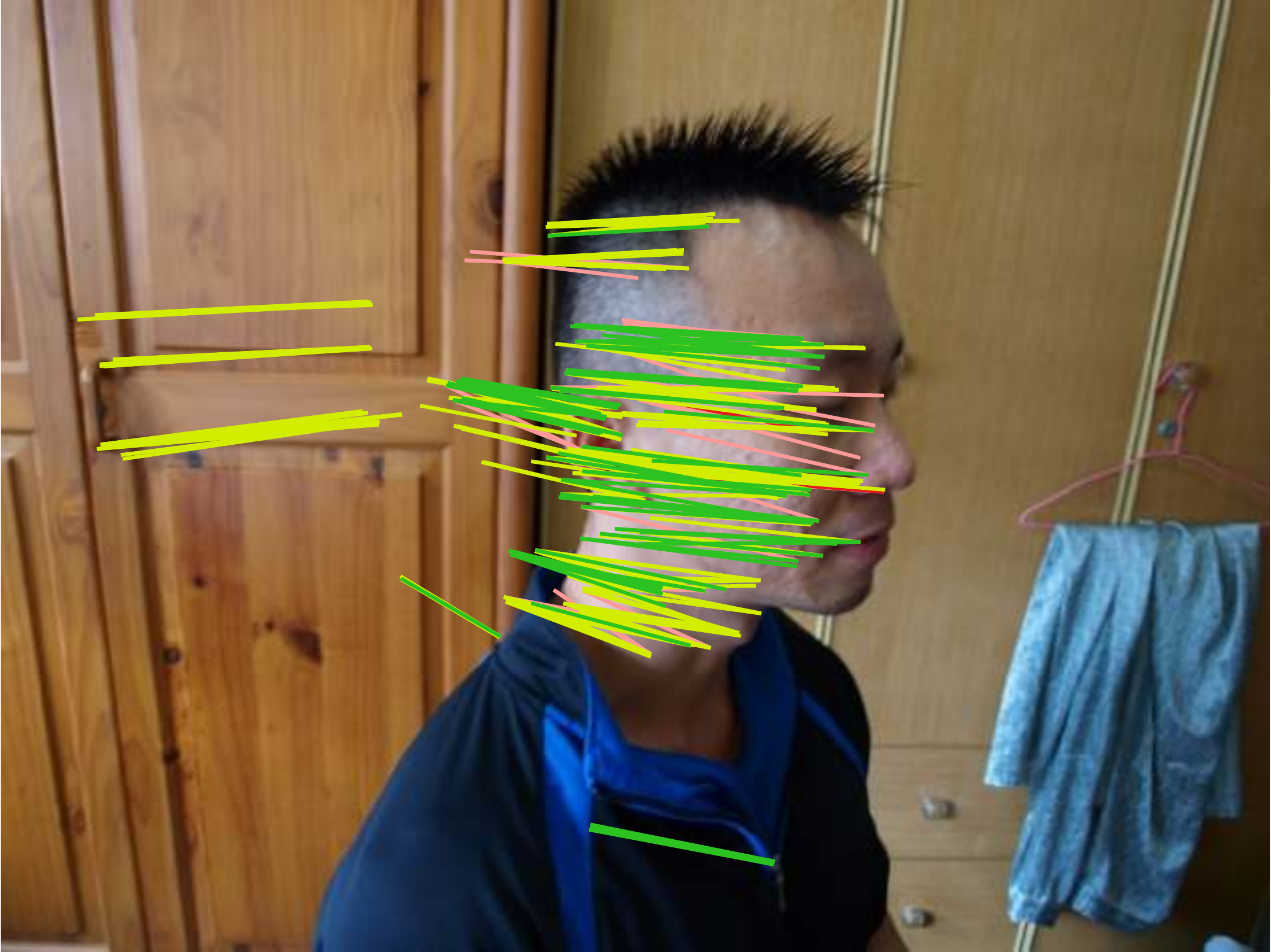}
	\\
	\vspace{0.5em}
	\rotatebox[origin=l]{90}{\mbox{\hspace{2em}VFC}}
	\includegraphics[height=7.5em]{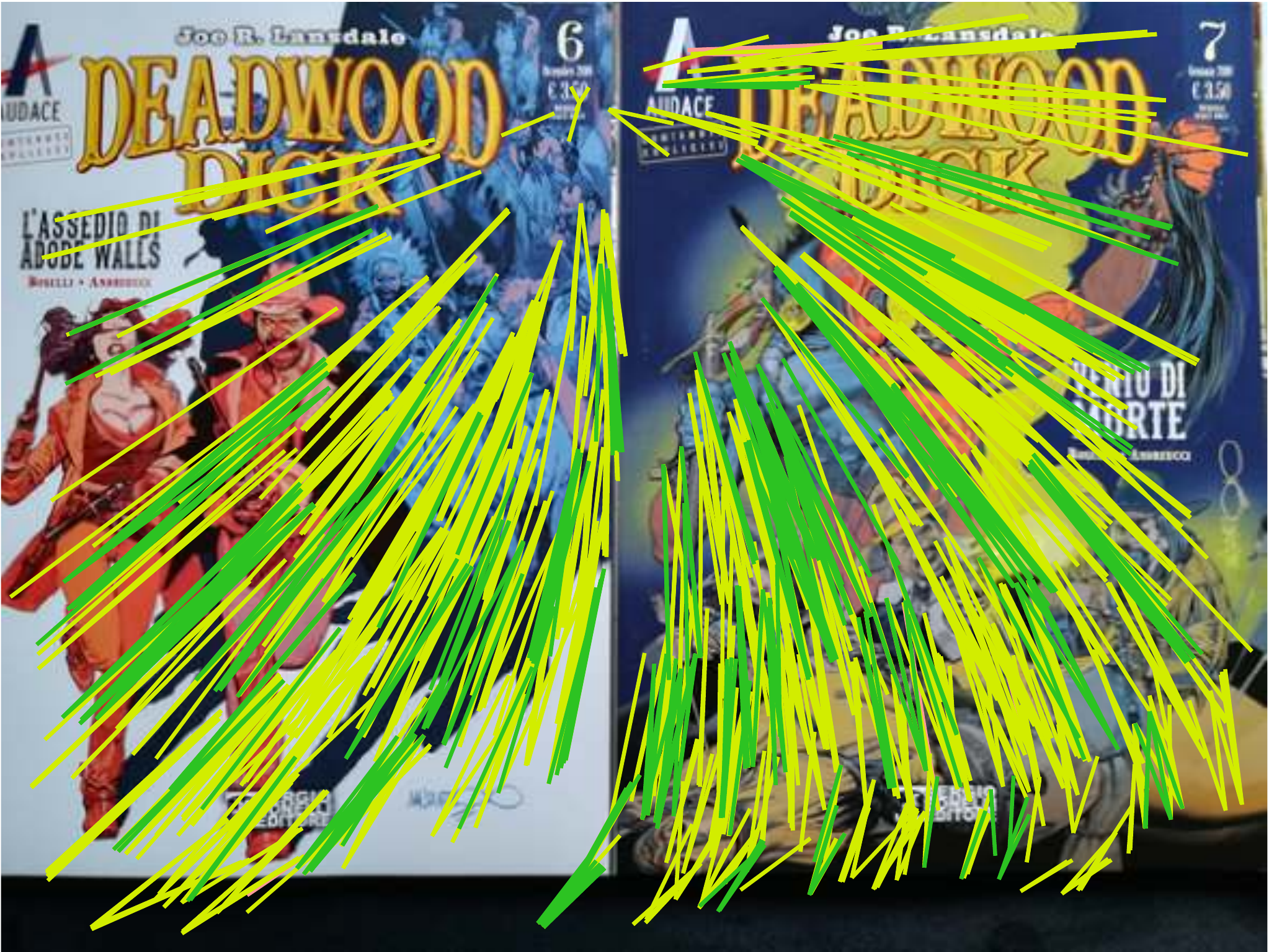}
	\includegraphics[height=7.5em]{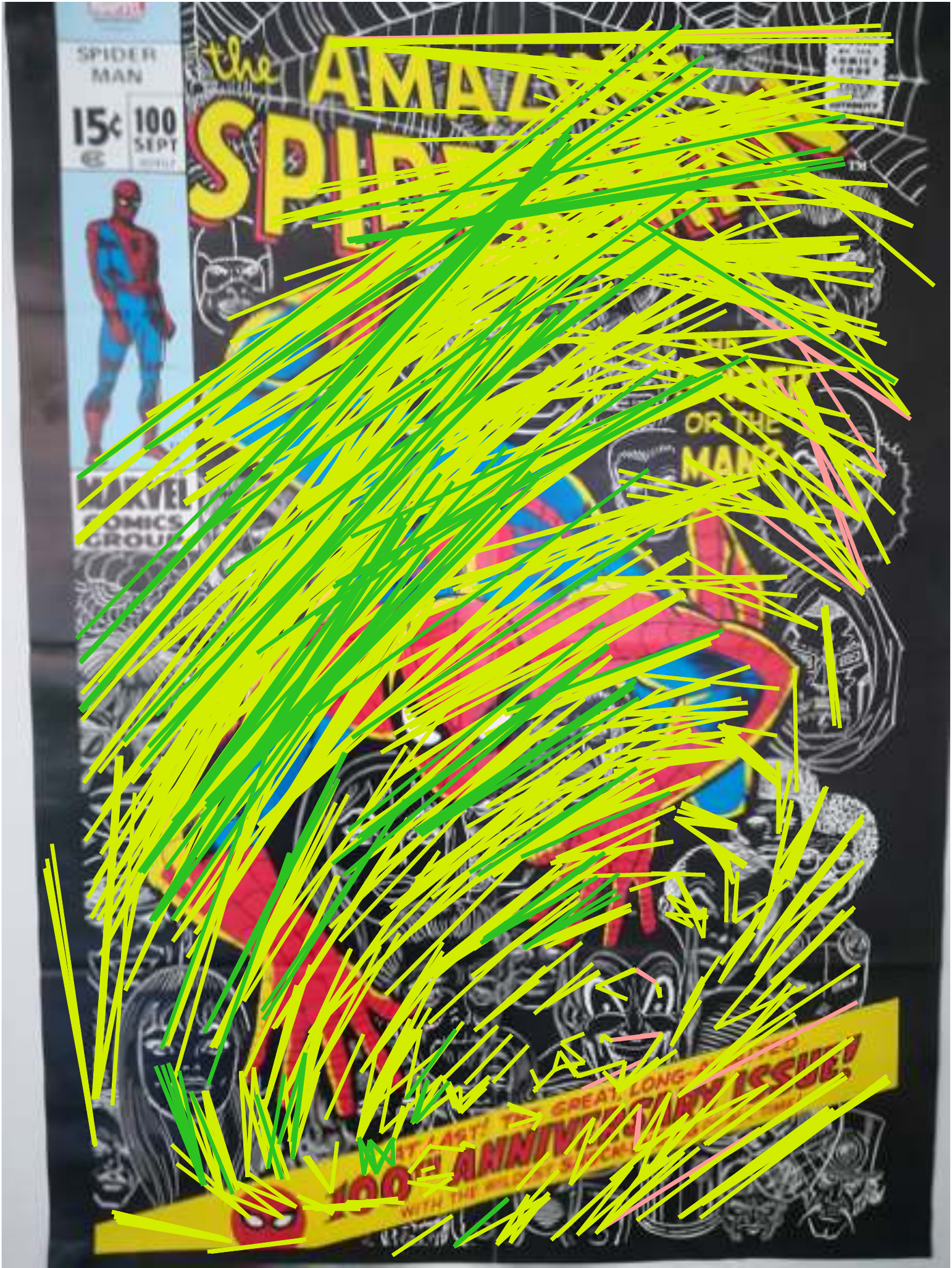}
	\includegraphics[height=7.5em]{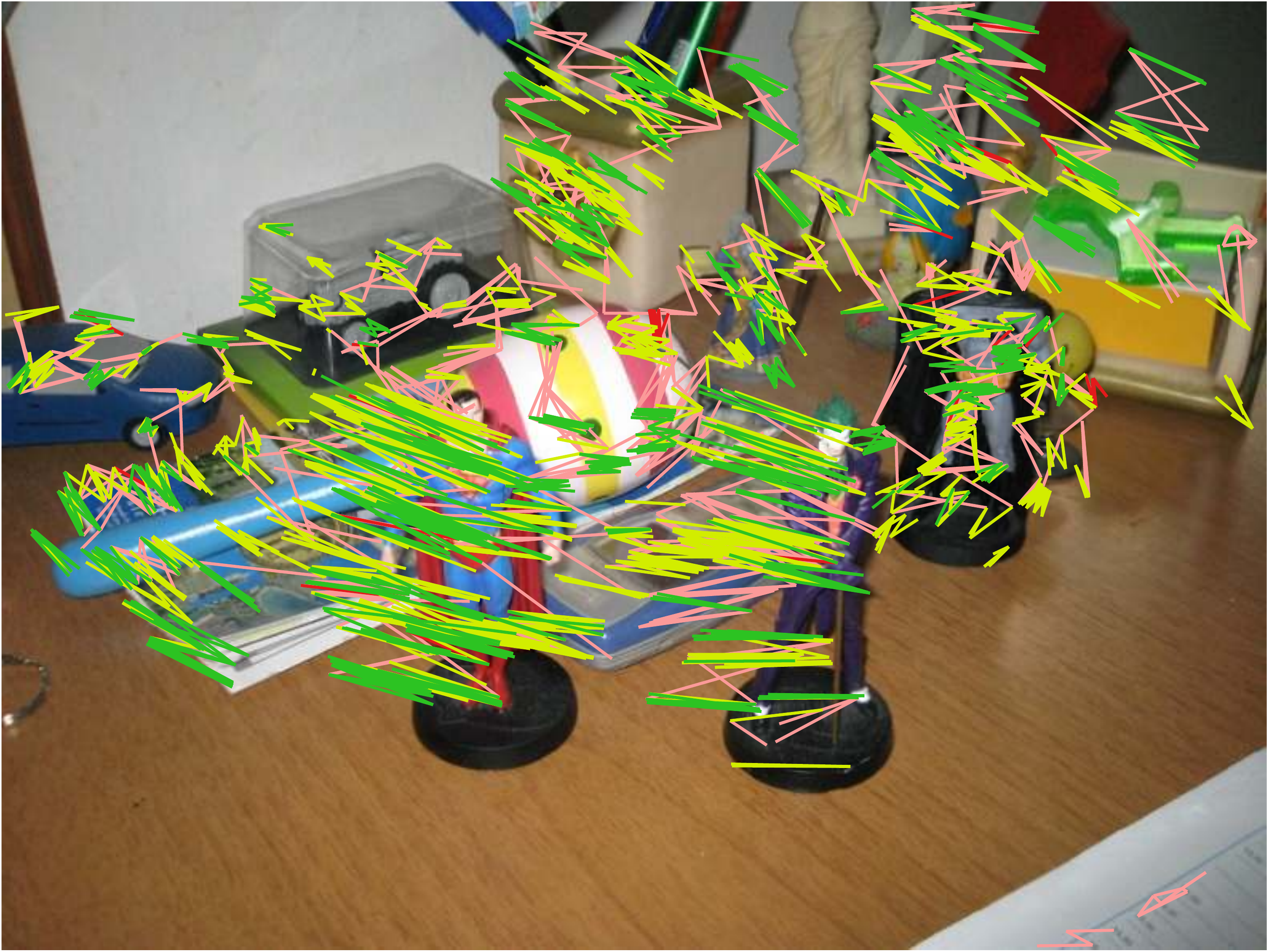}
	\includegraphics[height=7.5em]{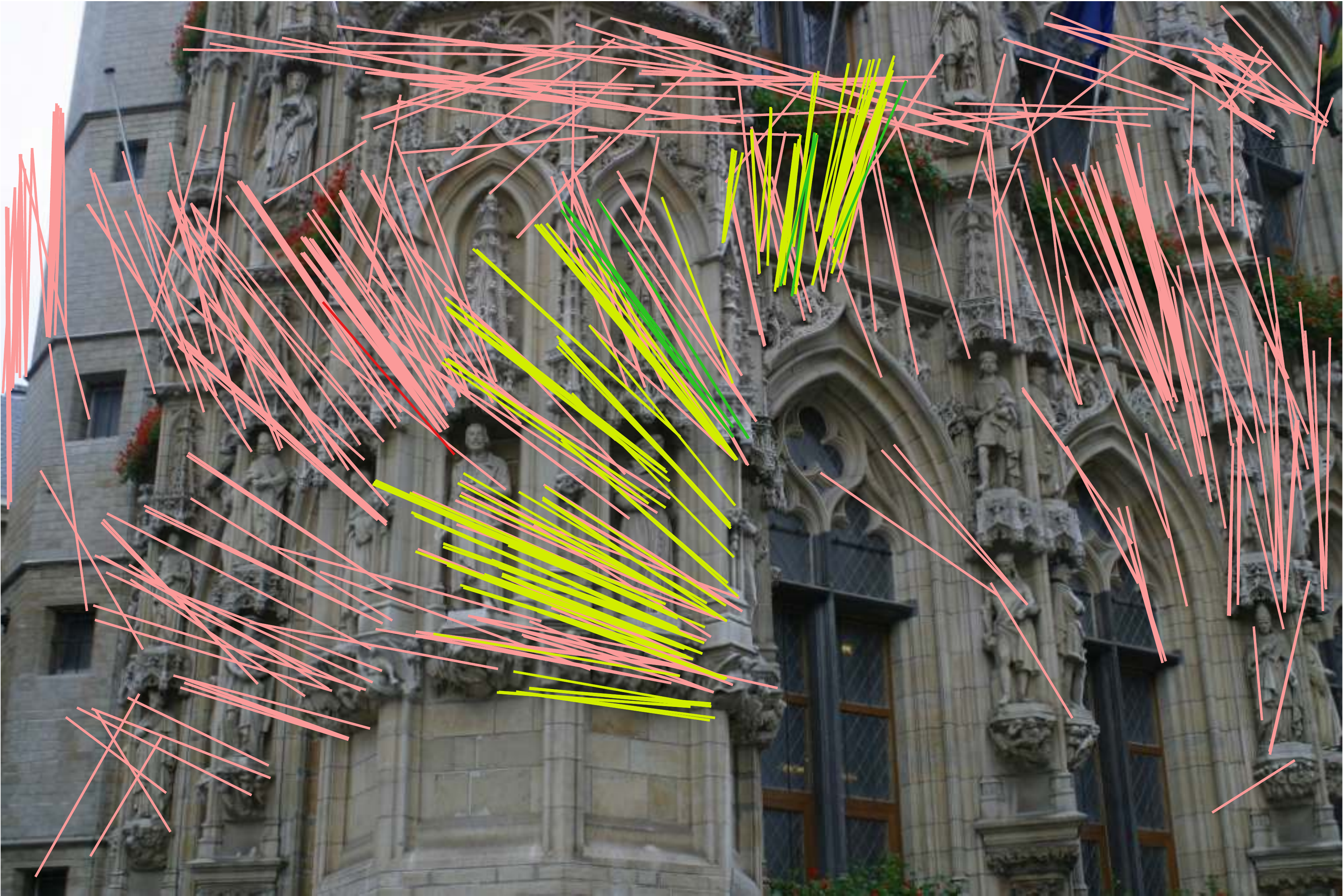}
	\includegraphics[height=7.5em]{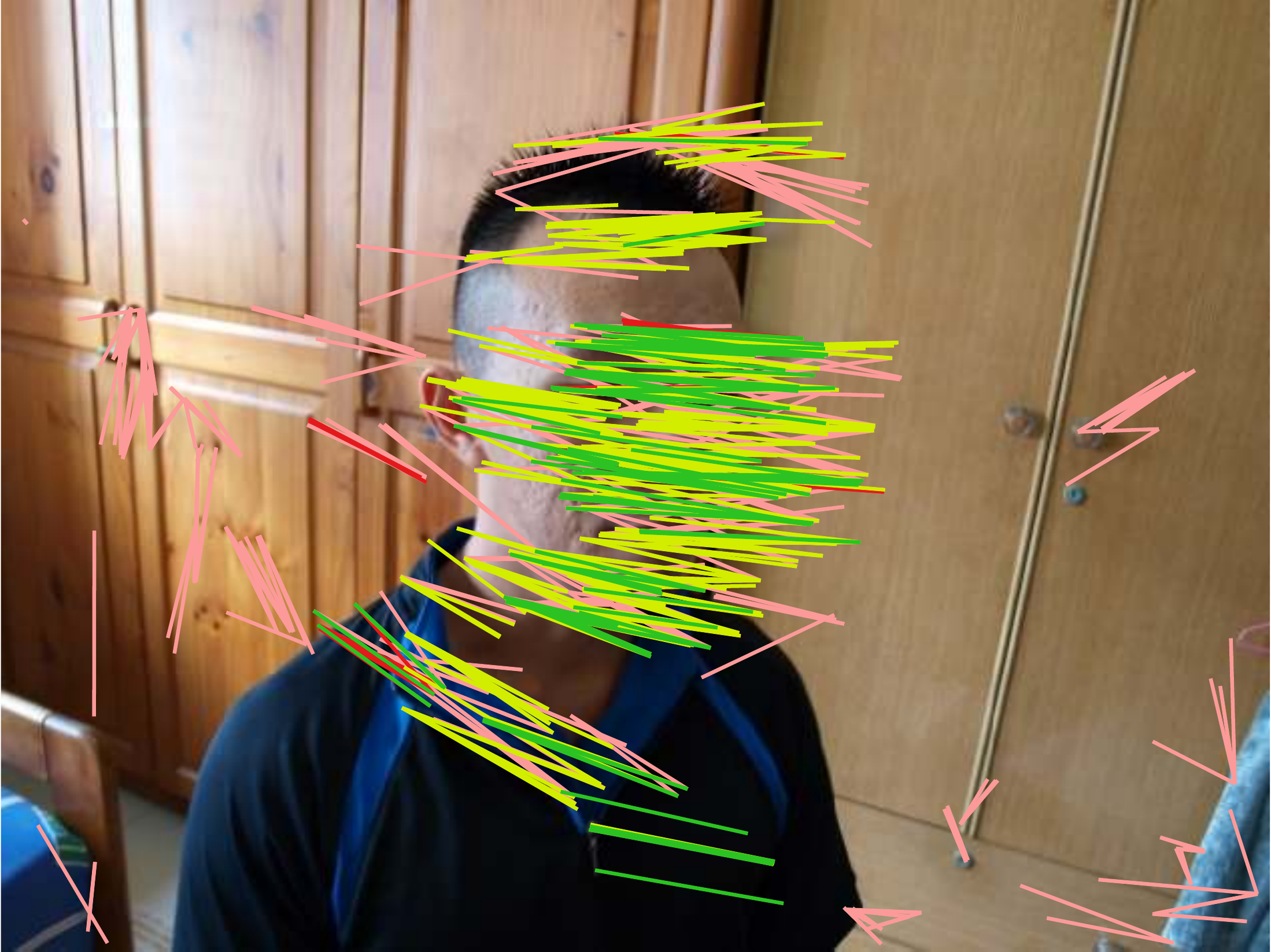}
	\\
	\vspace{0.5em}
	\rotatebox[origin=l]{90}{\mbox{\hspace{2em}LLT}}
	\includegraphics[height=7.5em]{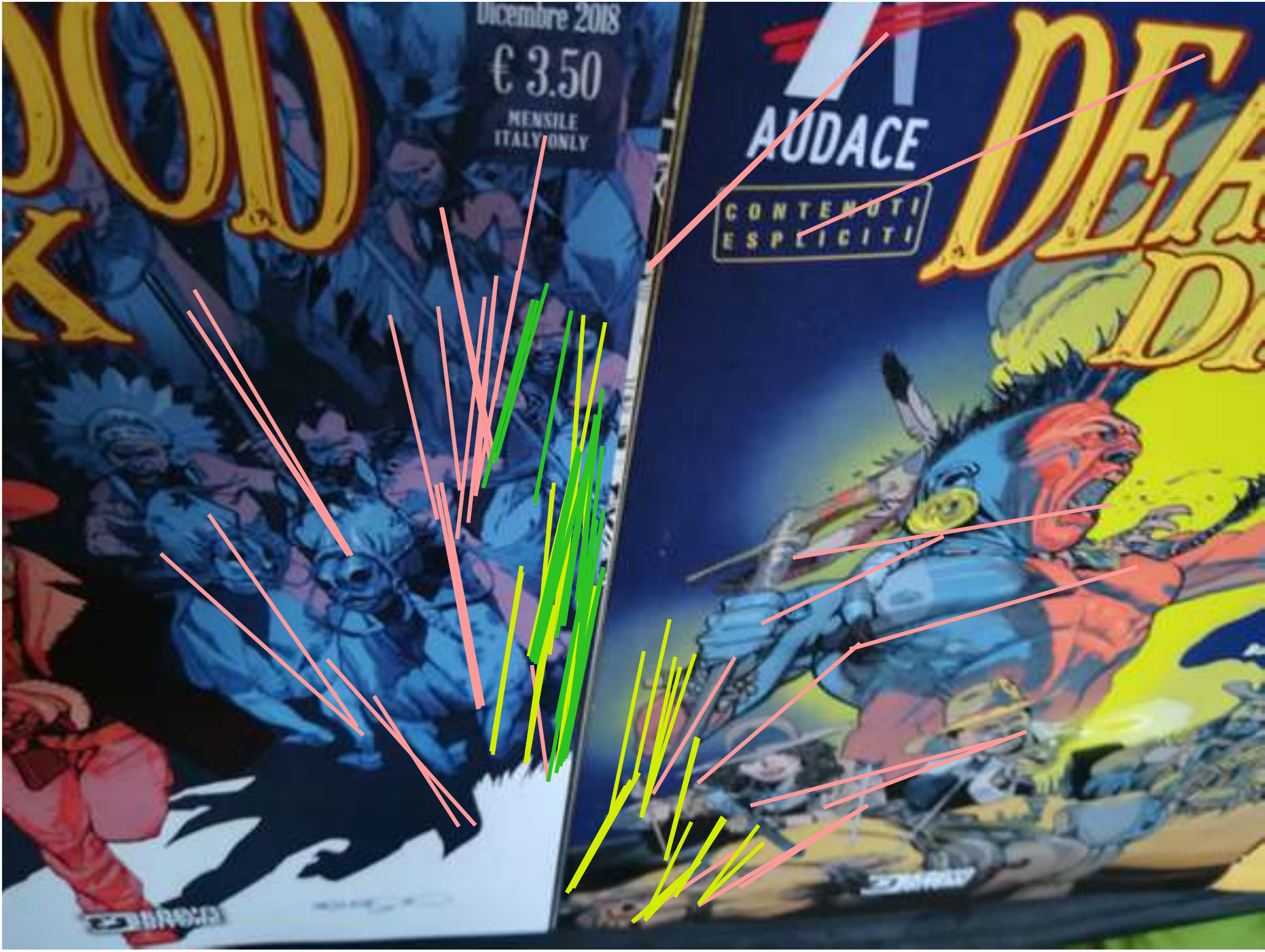}
	\includegraphics[height=7.5em]{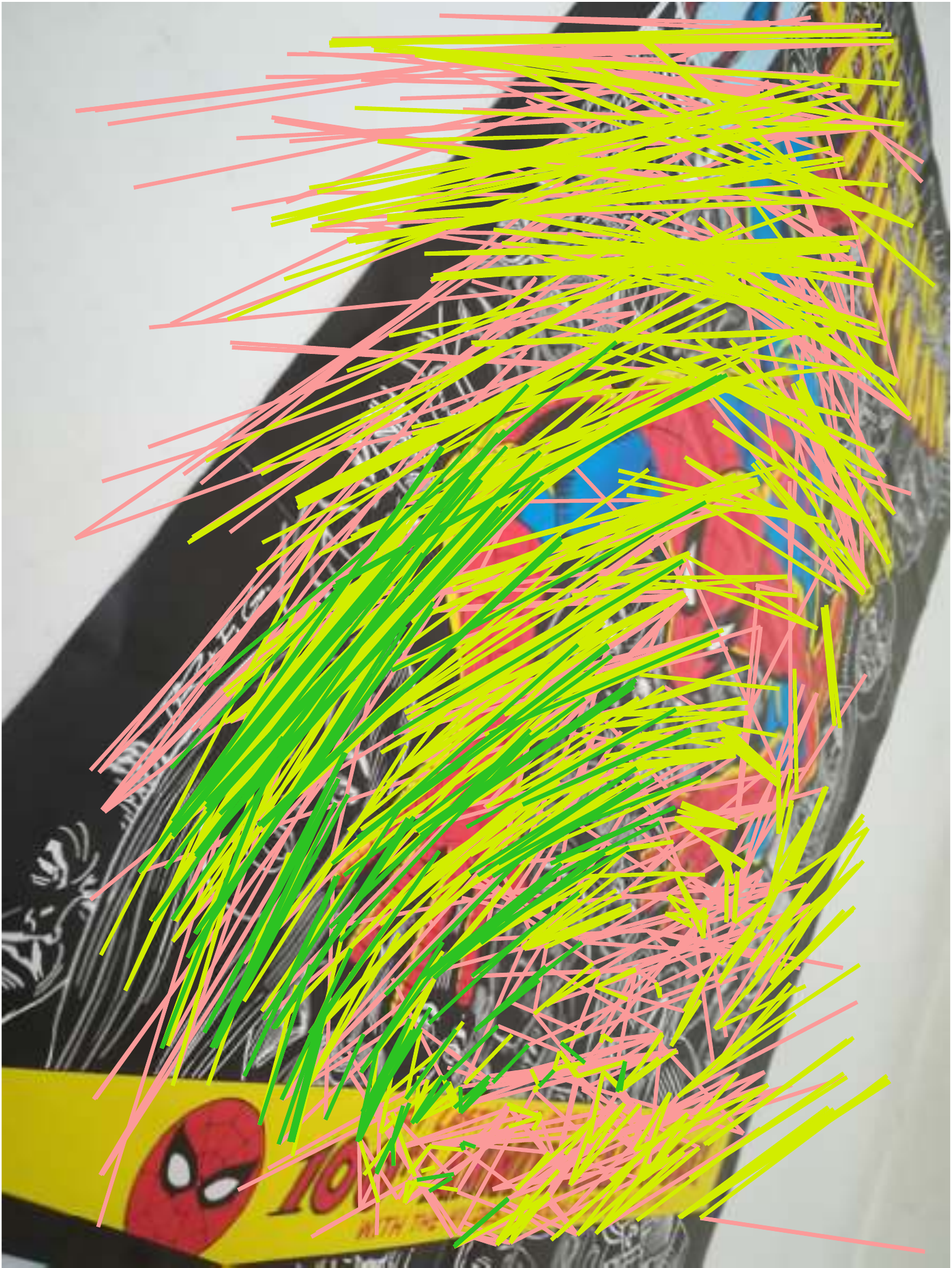}
	\includegraphics[height=7.5em]{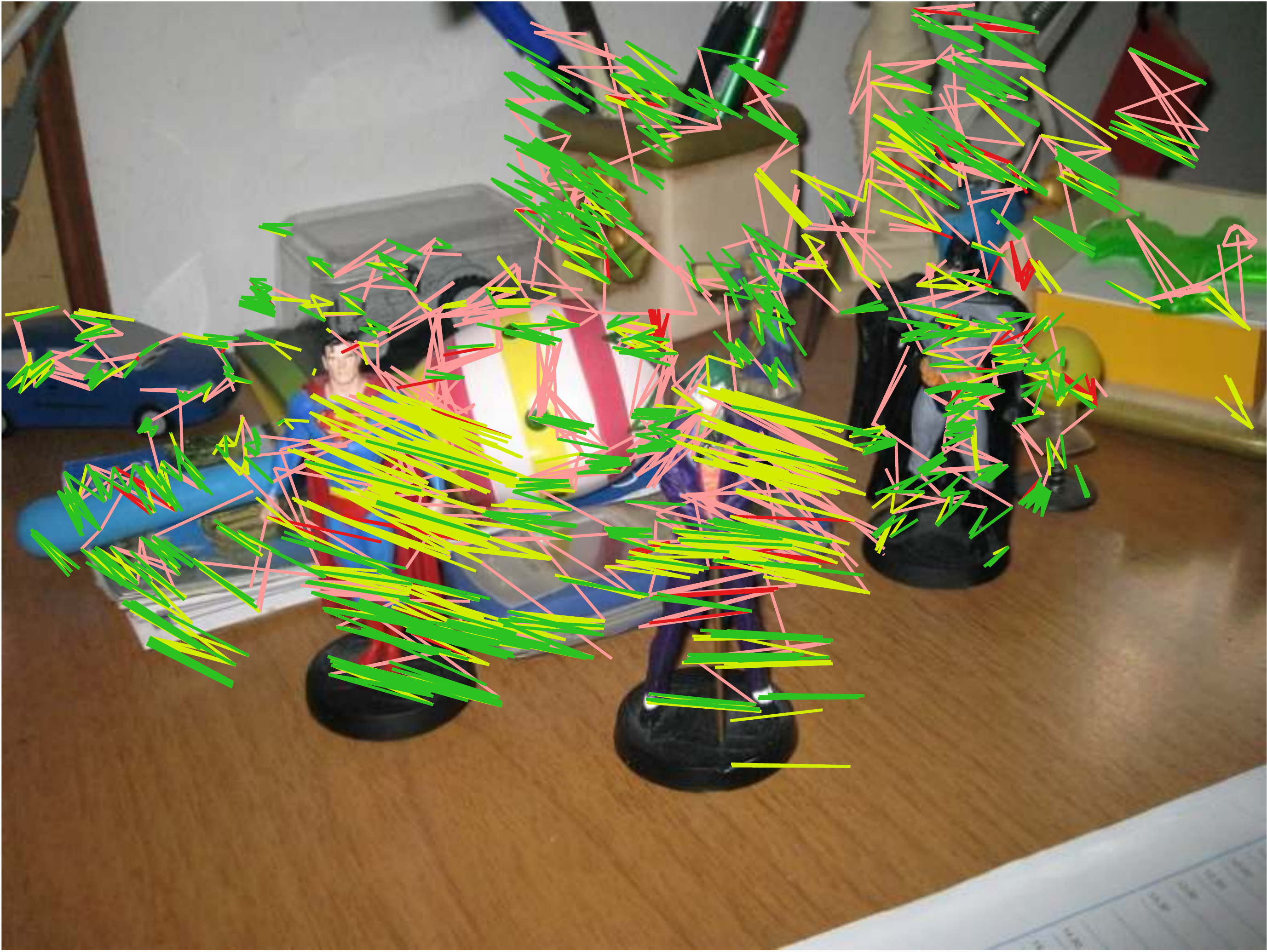}
	\includegraphics[height=7.5em]{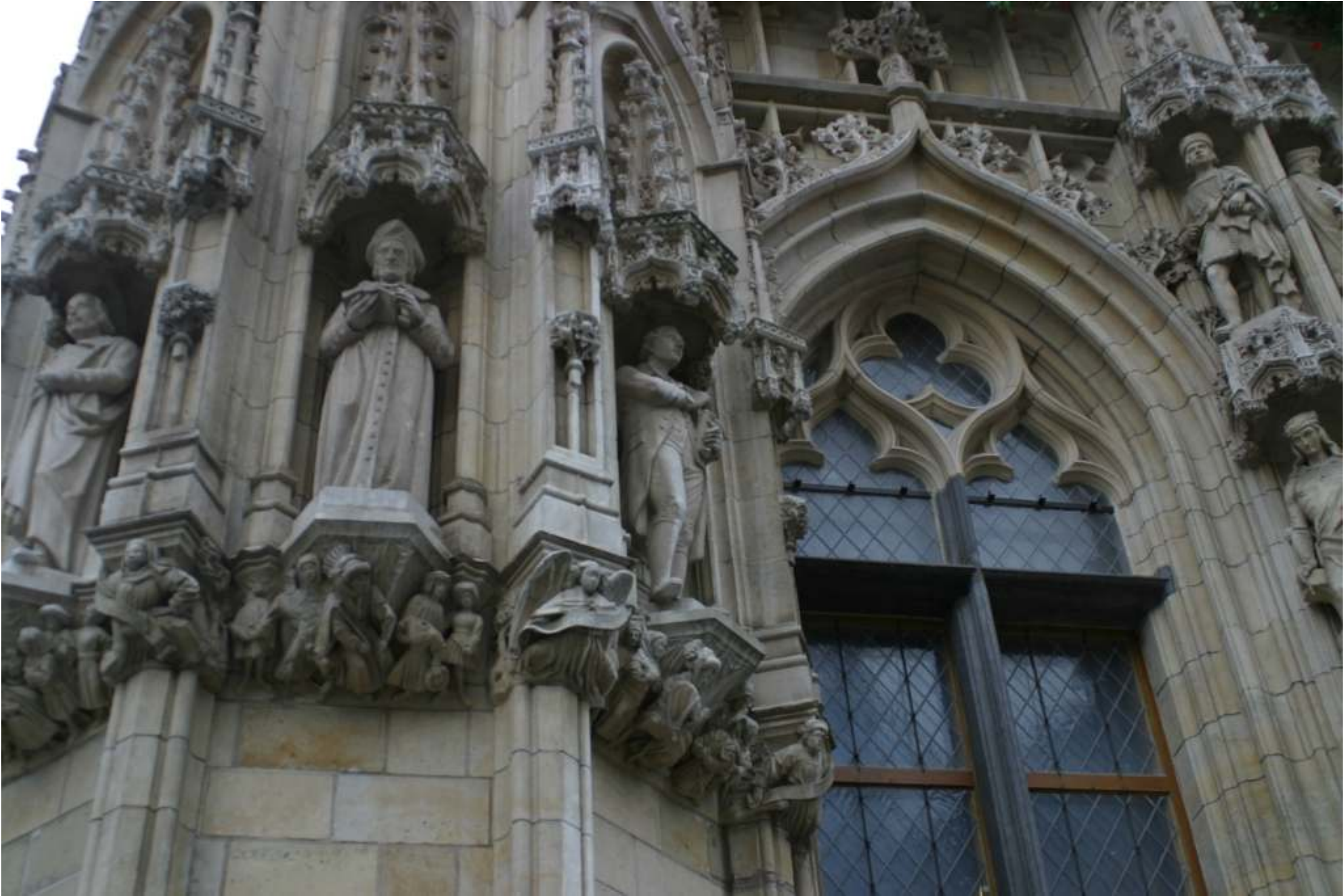}
	\includegraphics[height=7.5em]{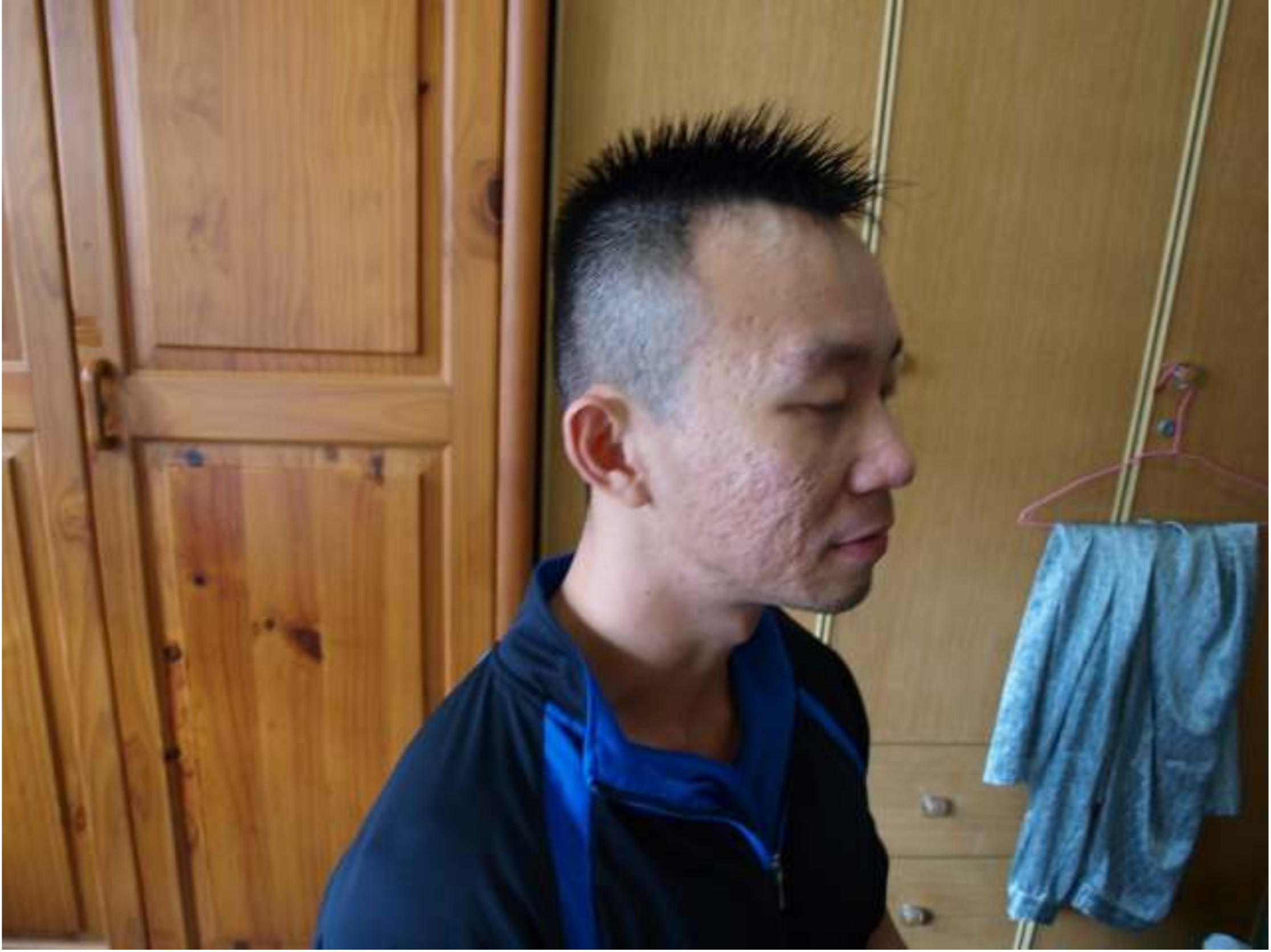}
	\\
	\begin{flushleft}
		\hspace*{7.5em}DD15\hspace{4.75em}Spidey13\hspace{4.75em}DC01\hspace{8em}LeuvenB\hspace{7.75em}BF
	\end{flushleft}
	\caption{\label{example_1a}
		Planar and non-planar local spatial filter matches according to the best configuration setup, the images of the input pair alternate among the rows. Image indexes are reported as suffix when the sequence contains more than two images. For each method inlier (yellow, green) and outlier (red and light red) clusters are shown, as well as the 1SAC filtered matches (green, red) (see Sec.~\ref{eval_dt}, best viewed in color and zoomed in).}
\end{figure*}

\begin{figure*}
	\center
	\rotatebox[origin=l]{90}{\mbox{\hspace{0.5em}RFM-SCAN}}
	\includegraphics[height=7.5em]{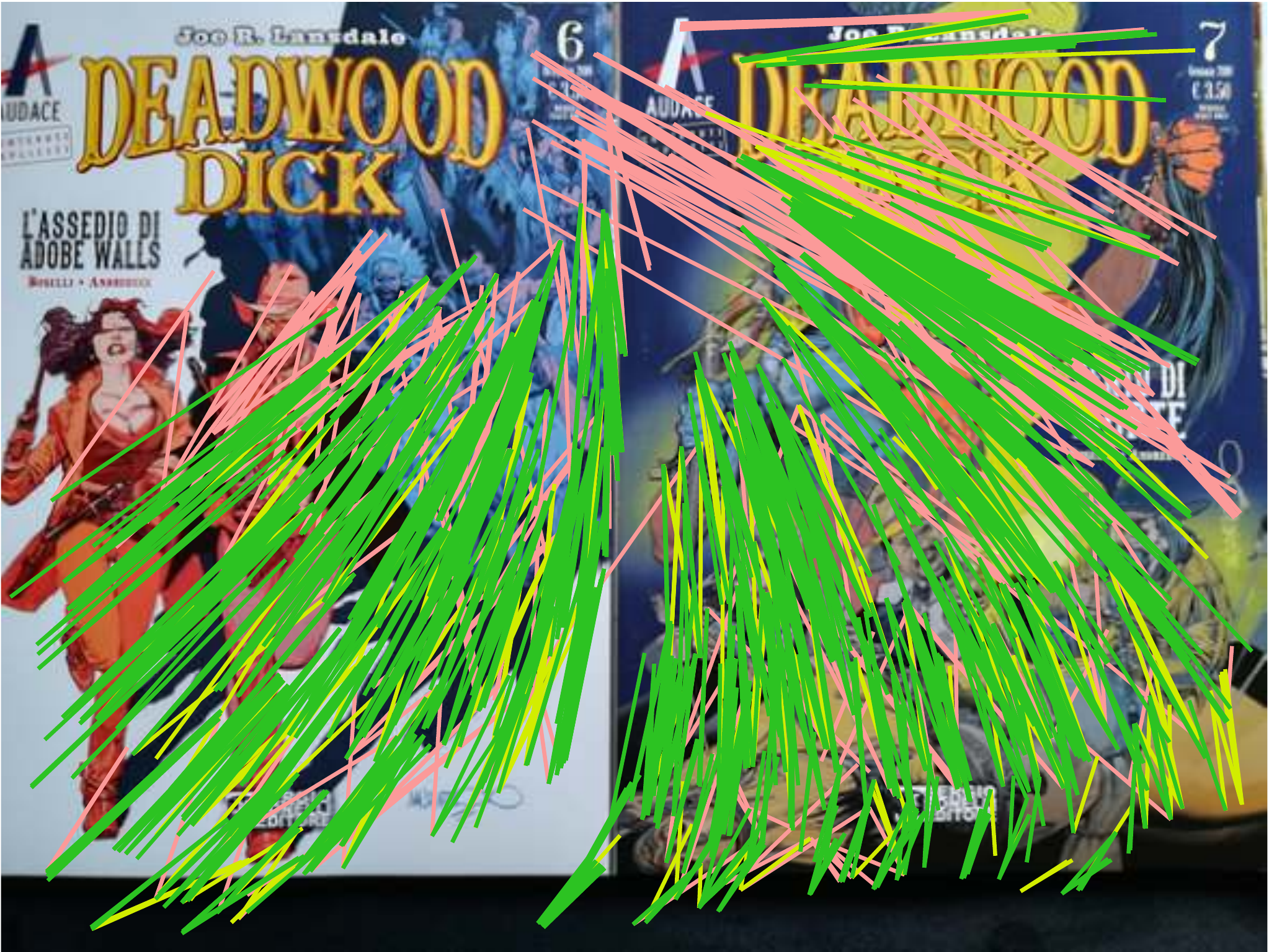}
	\includegraphics[height=7.5em]{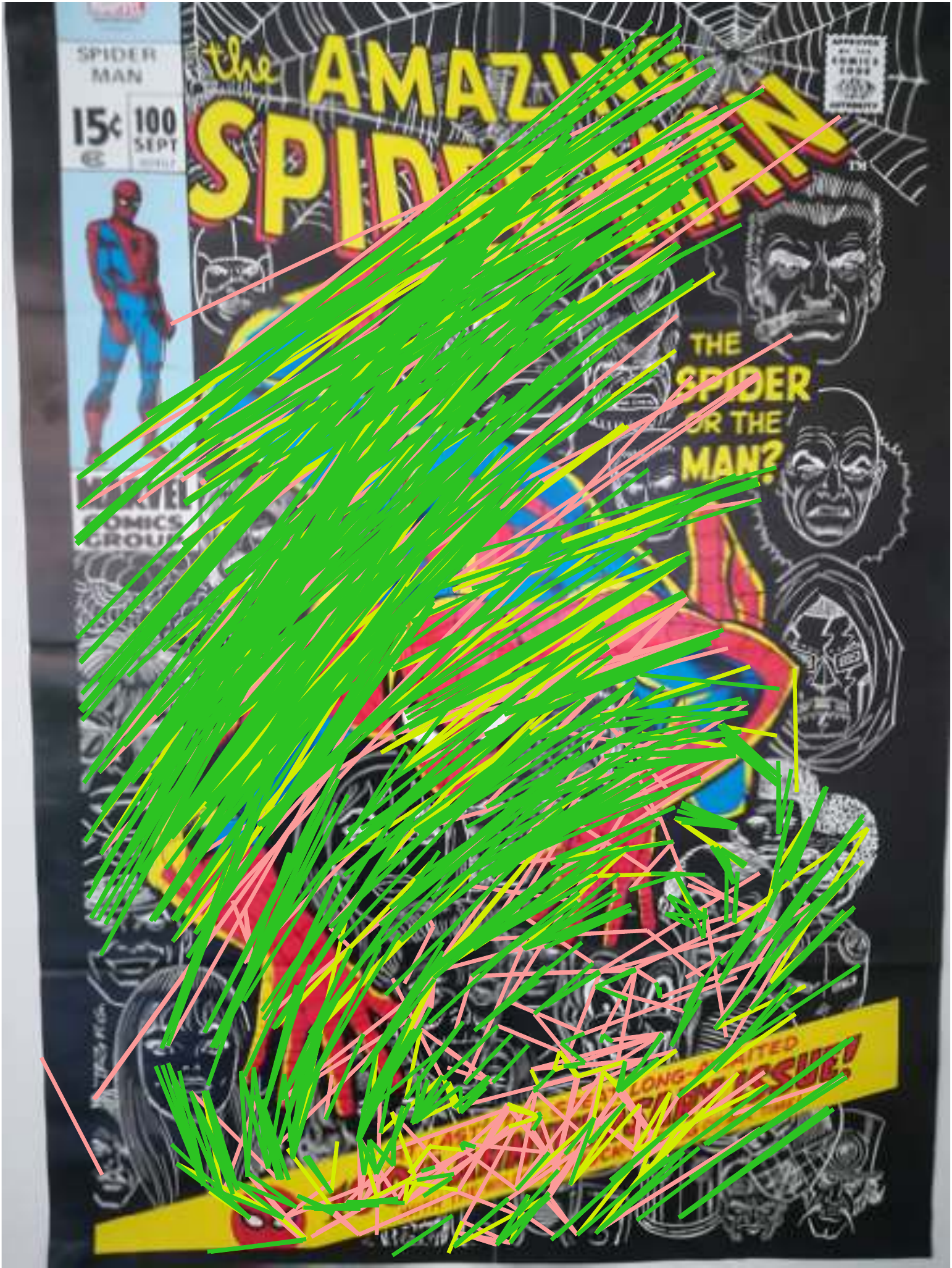}
	\includegraphics[height=7.5em]{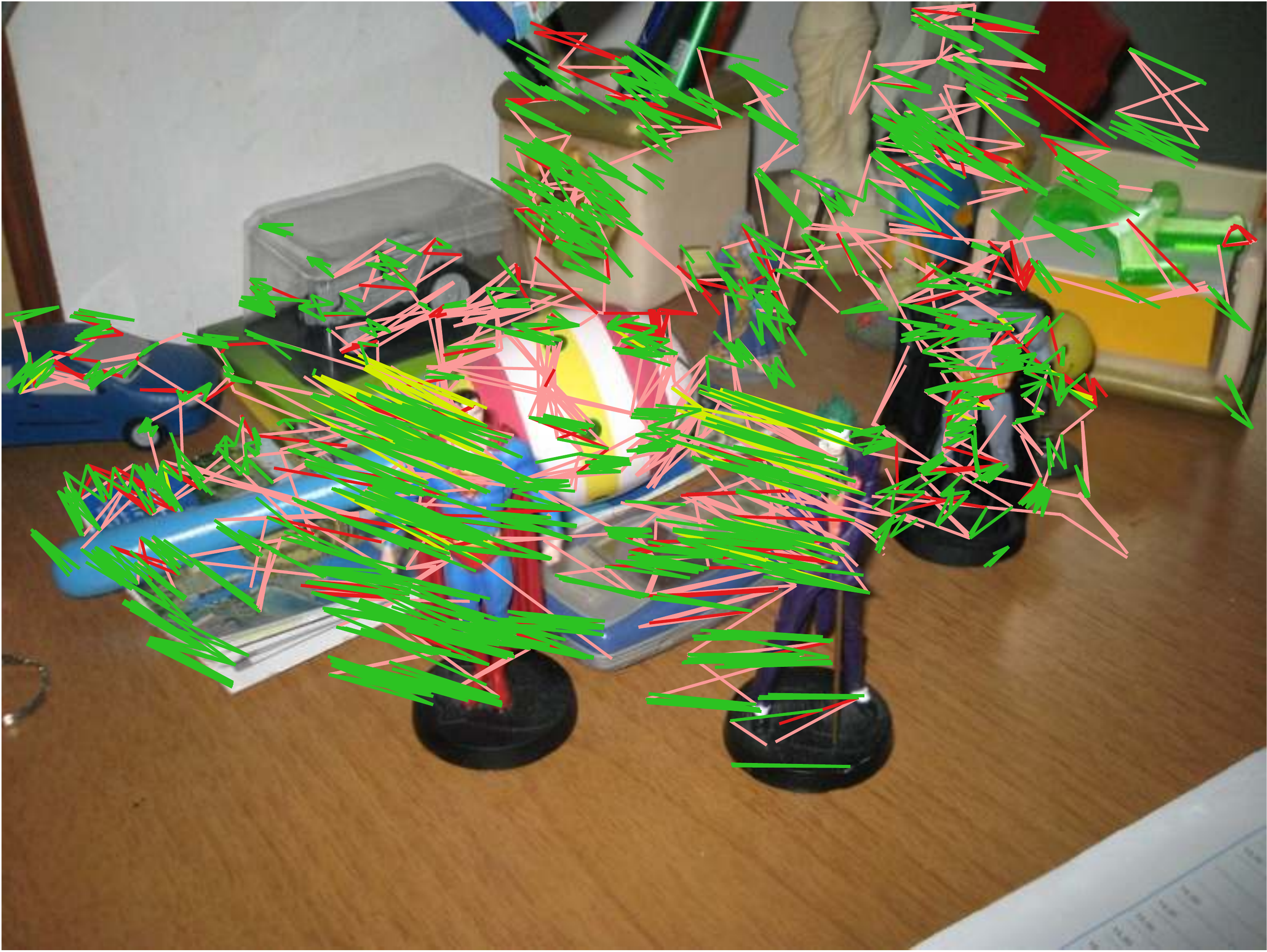}
	\includegraphics[height=7.5em]{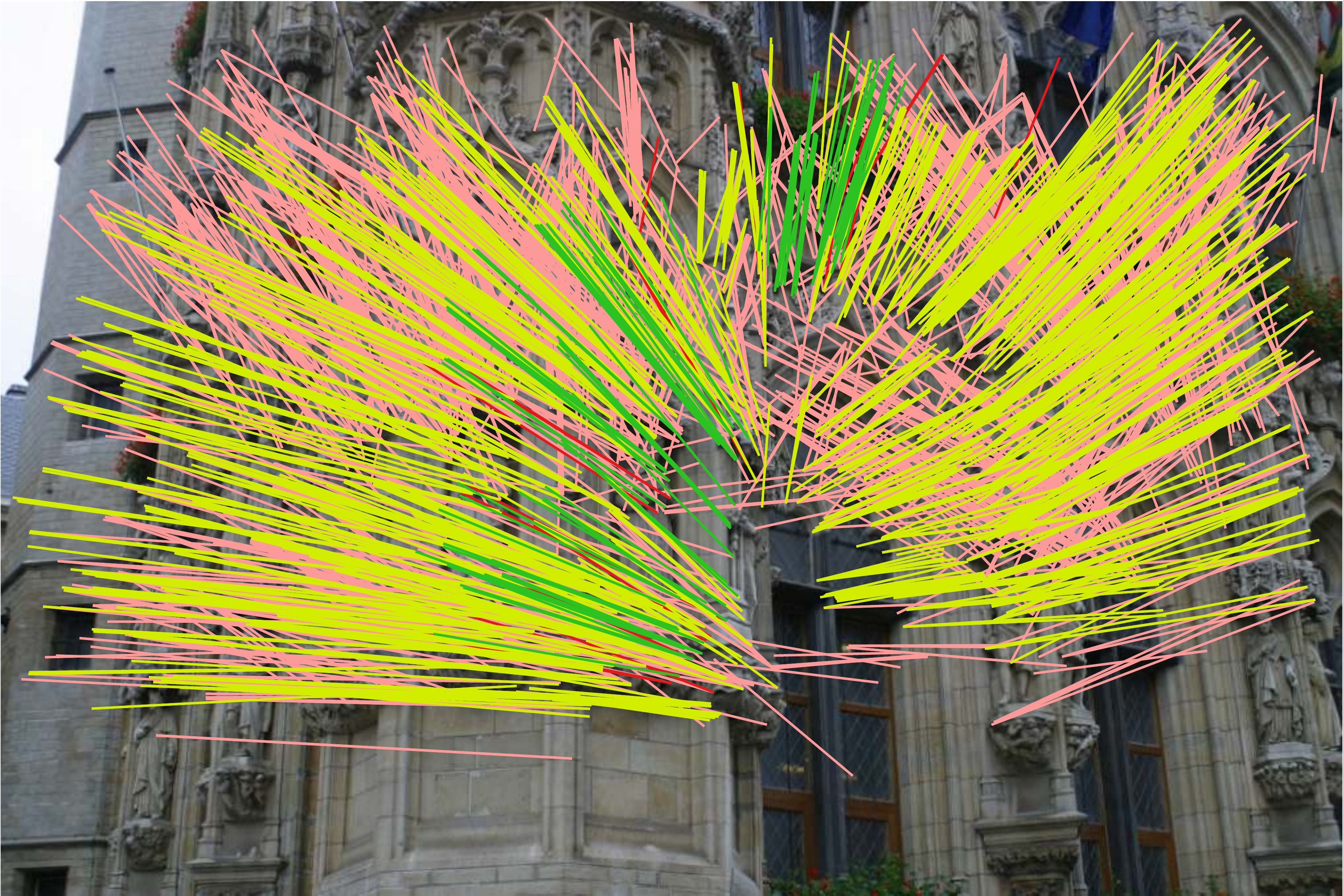}
	\includegraphics[height=7.5em]{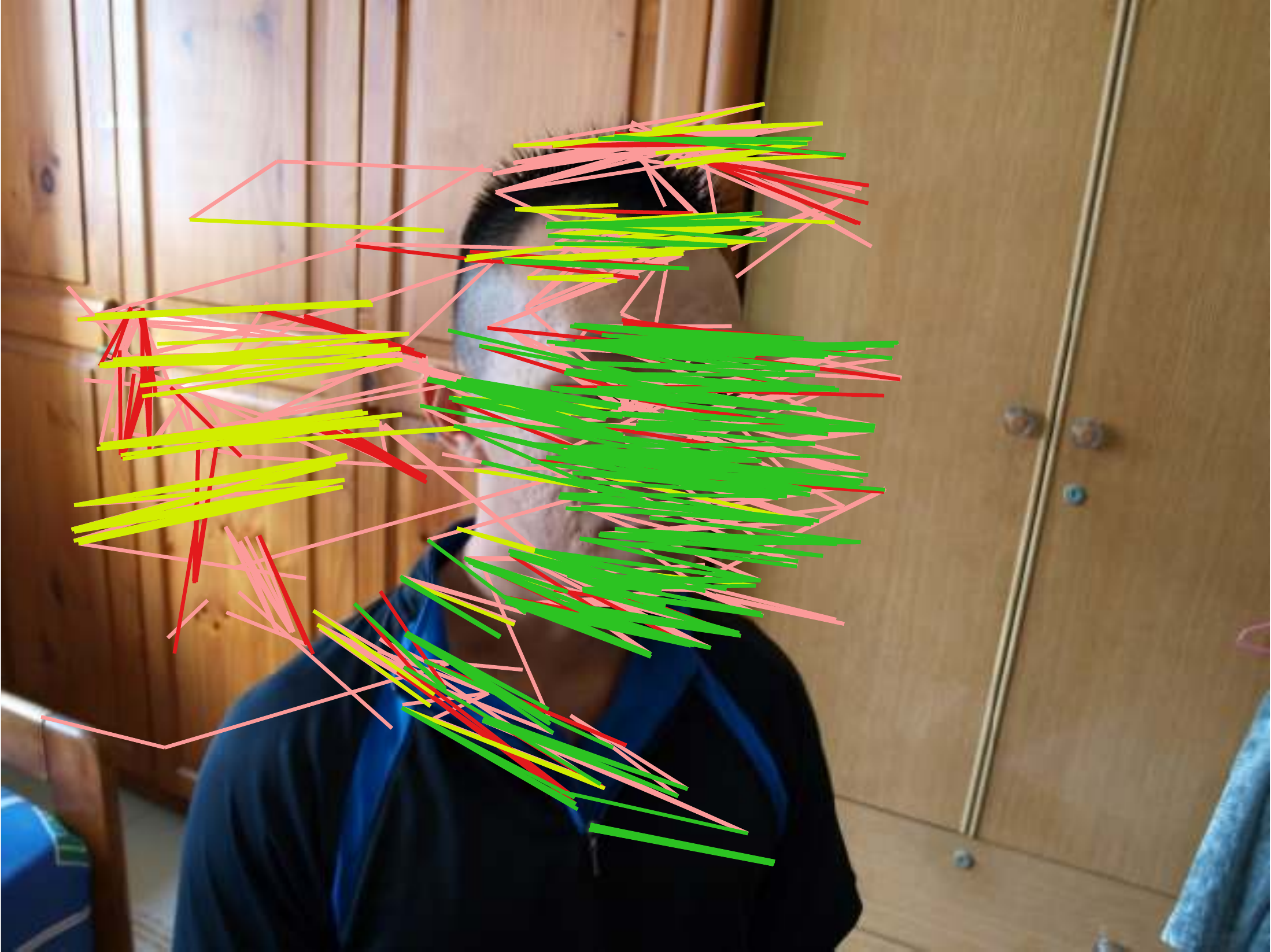}
	\\
	\vspace{0.5em}
	\rotatebox[origin=l]{90}{\mbox{\hspace{1em}AdaLAM}}
	\includegraphics[height=7.5em]{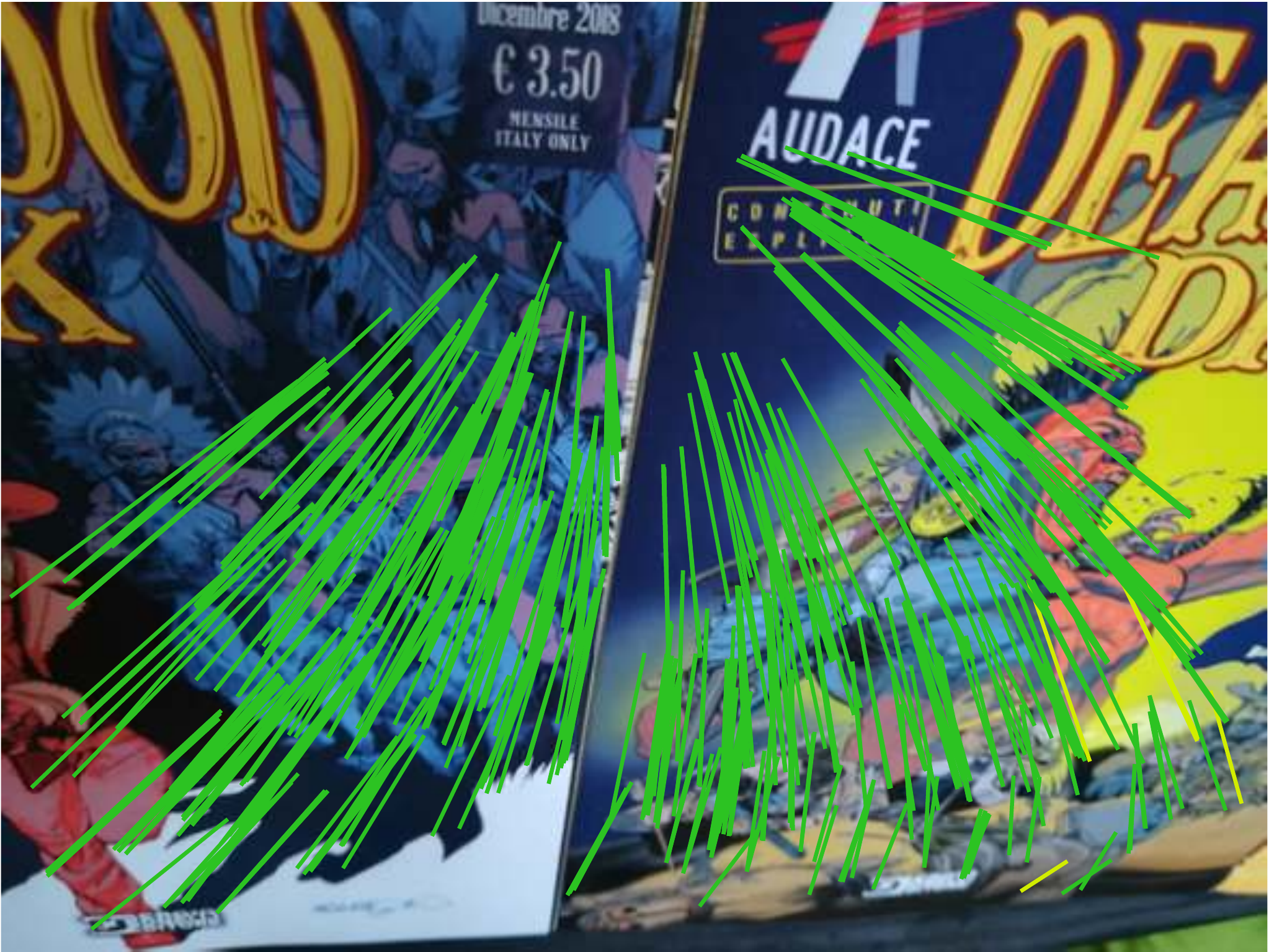}
	\includegraphics[height=7.5em]{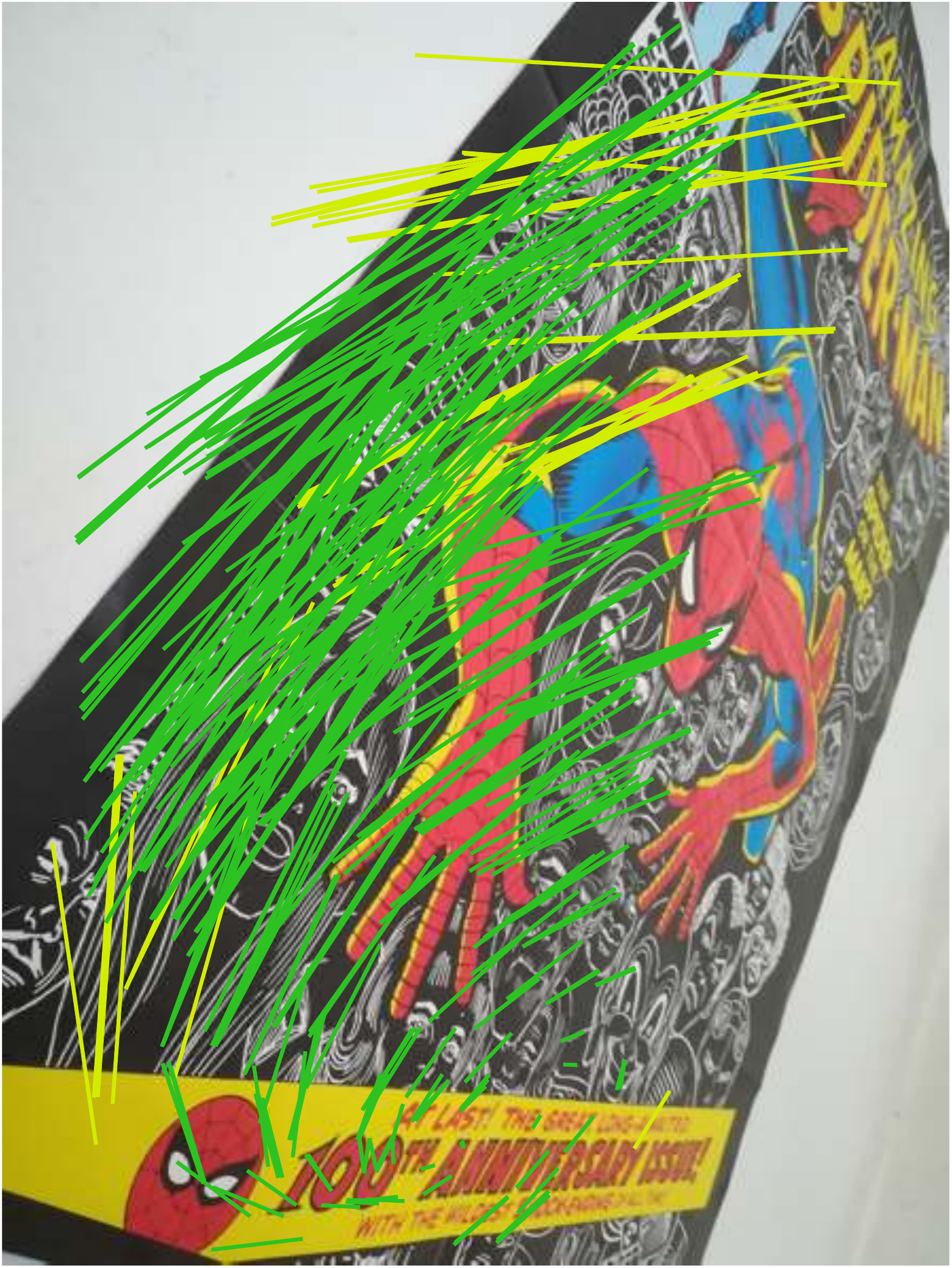}
	\includegraphics[height=7.5em]{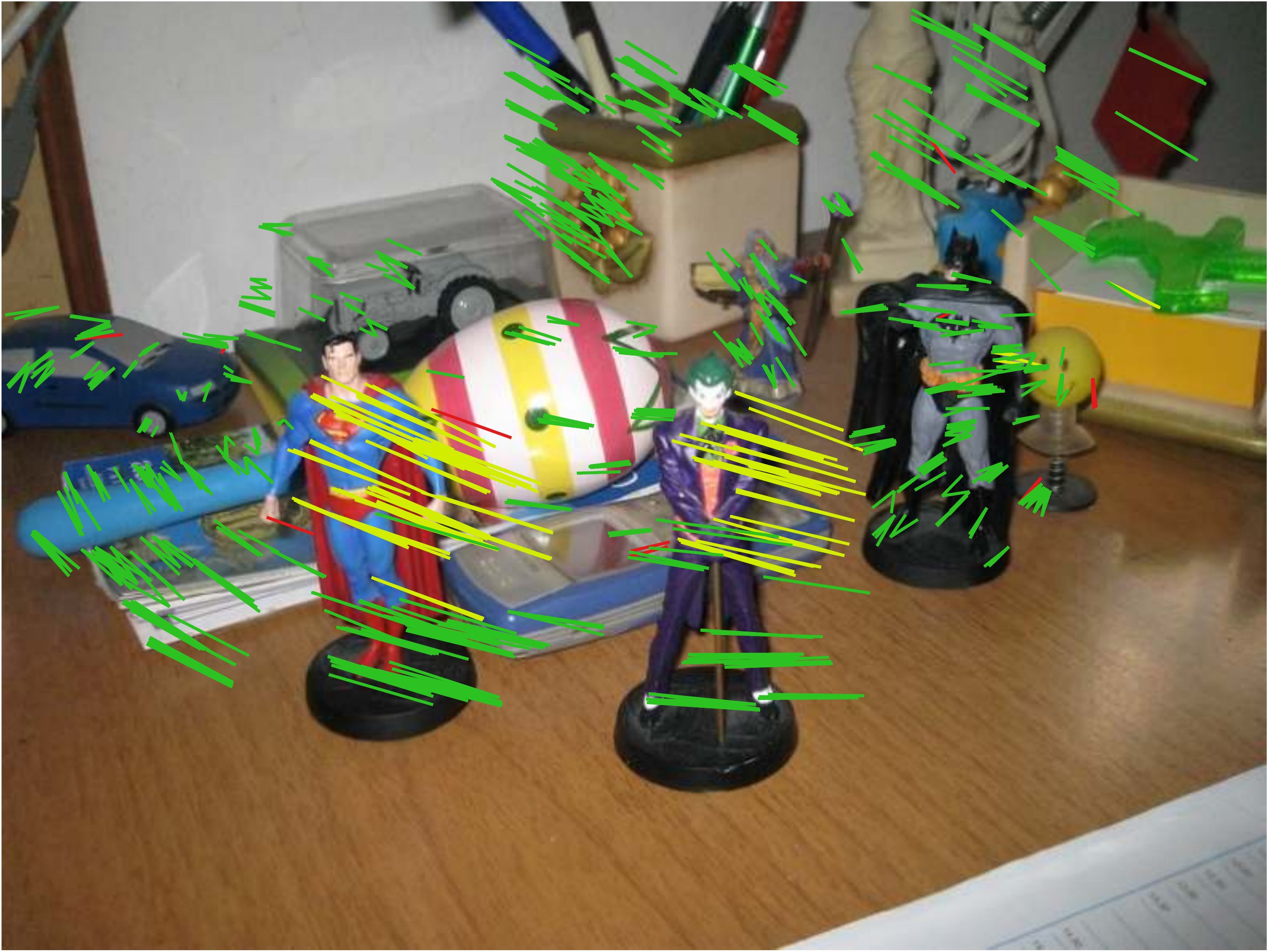}
	\includegraphics[height=7.5em]{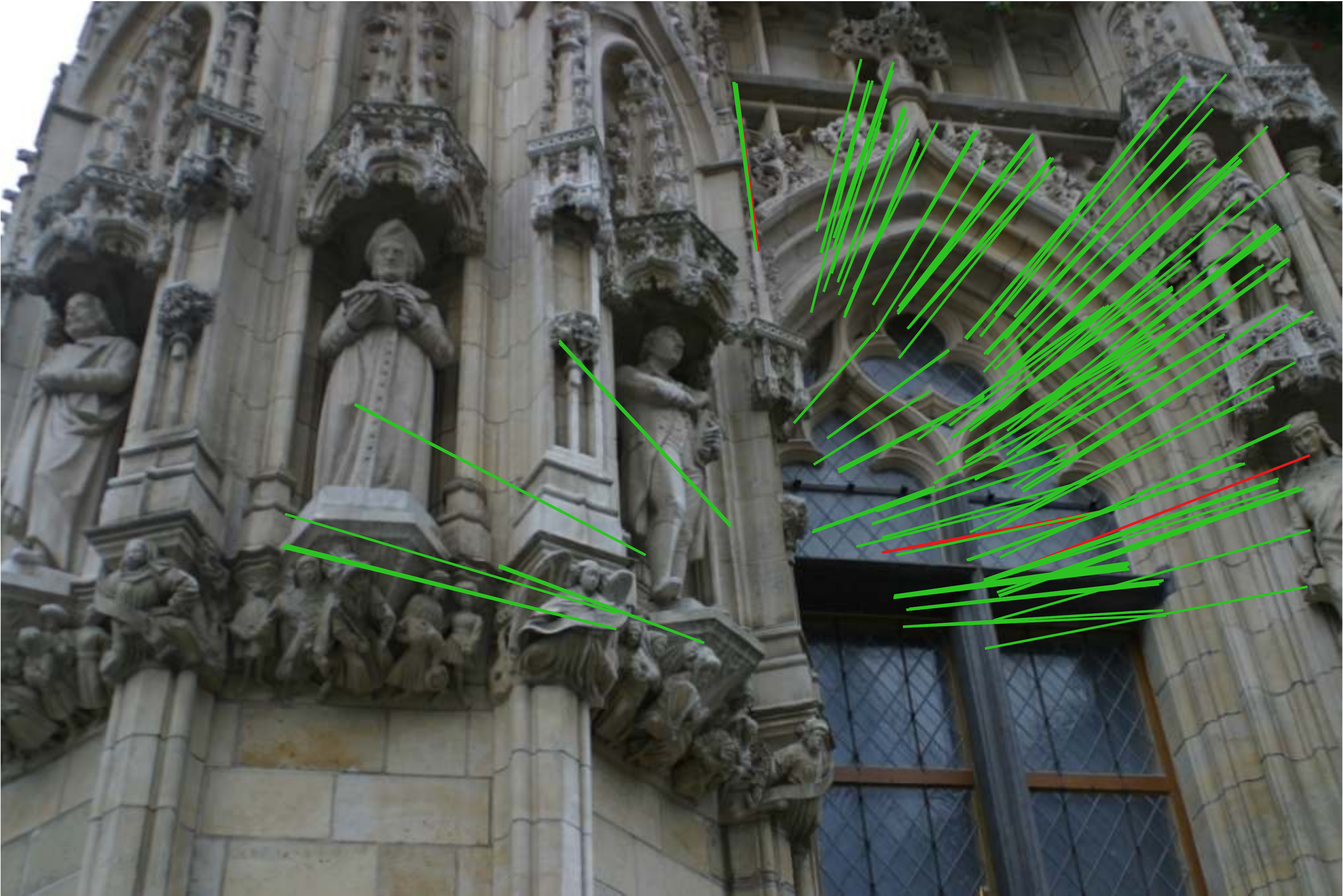}
	\includegraphics[height=7.5em]{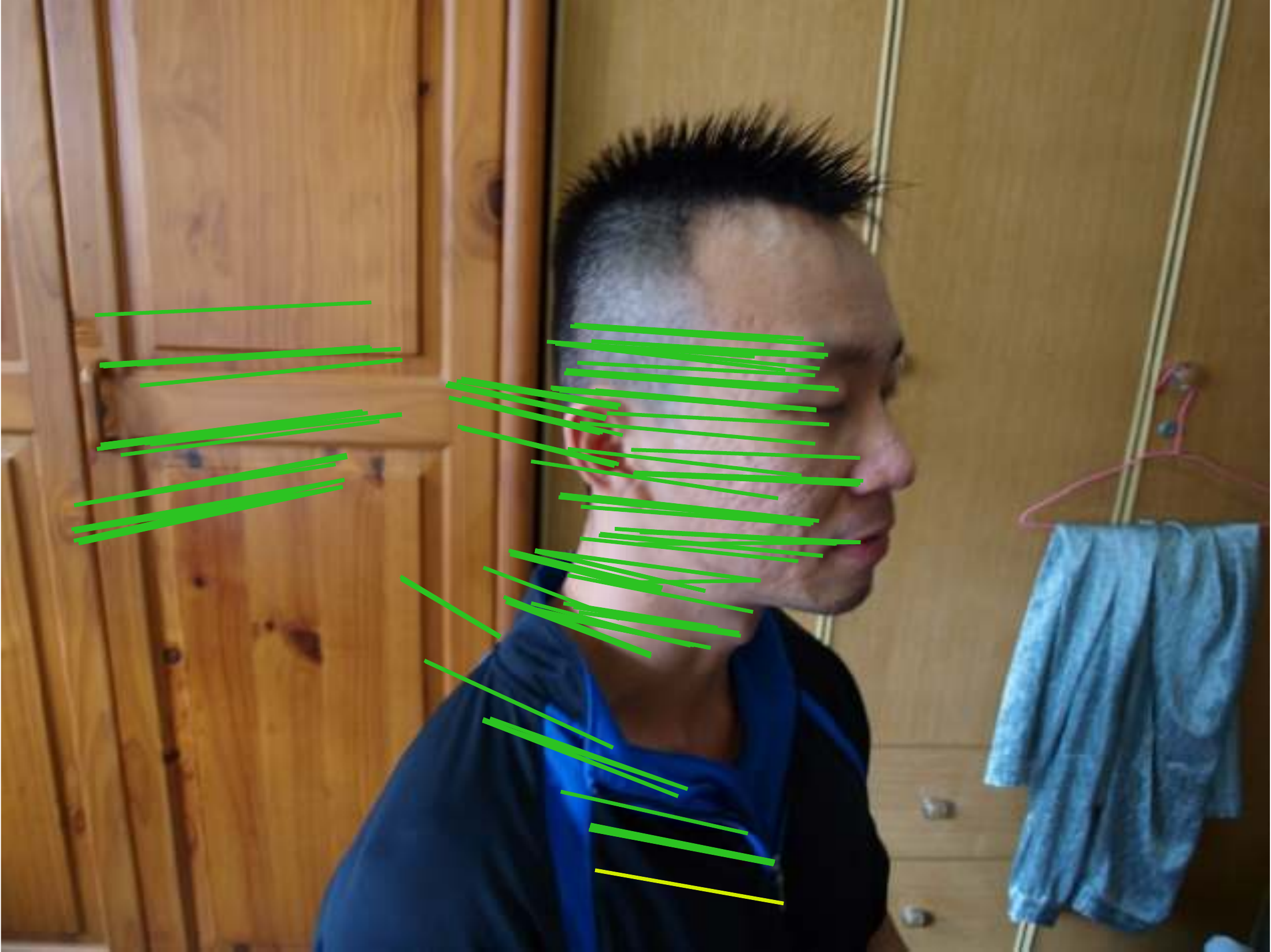}
	\\
	\vspace{0.5em}
	\rotatebox[origin=l]{90}{\mbox{\hspace{2em}OANet}}
	\includegraphics[height=7.5em]{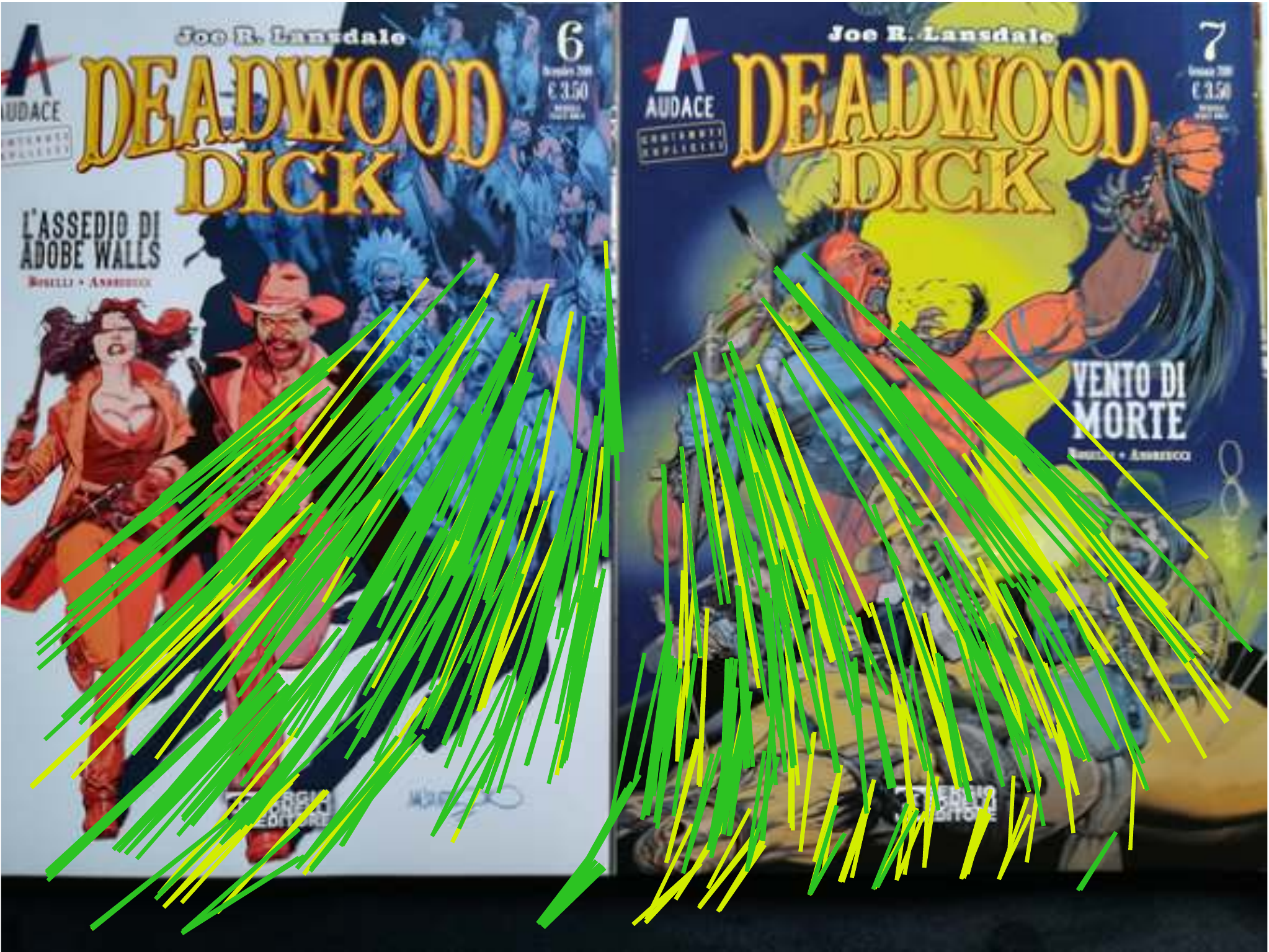}
	\includegraphics[height=7.5em]{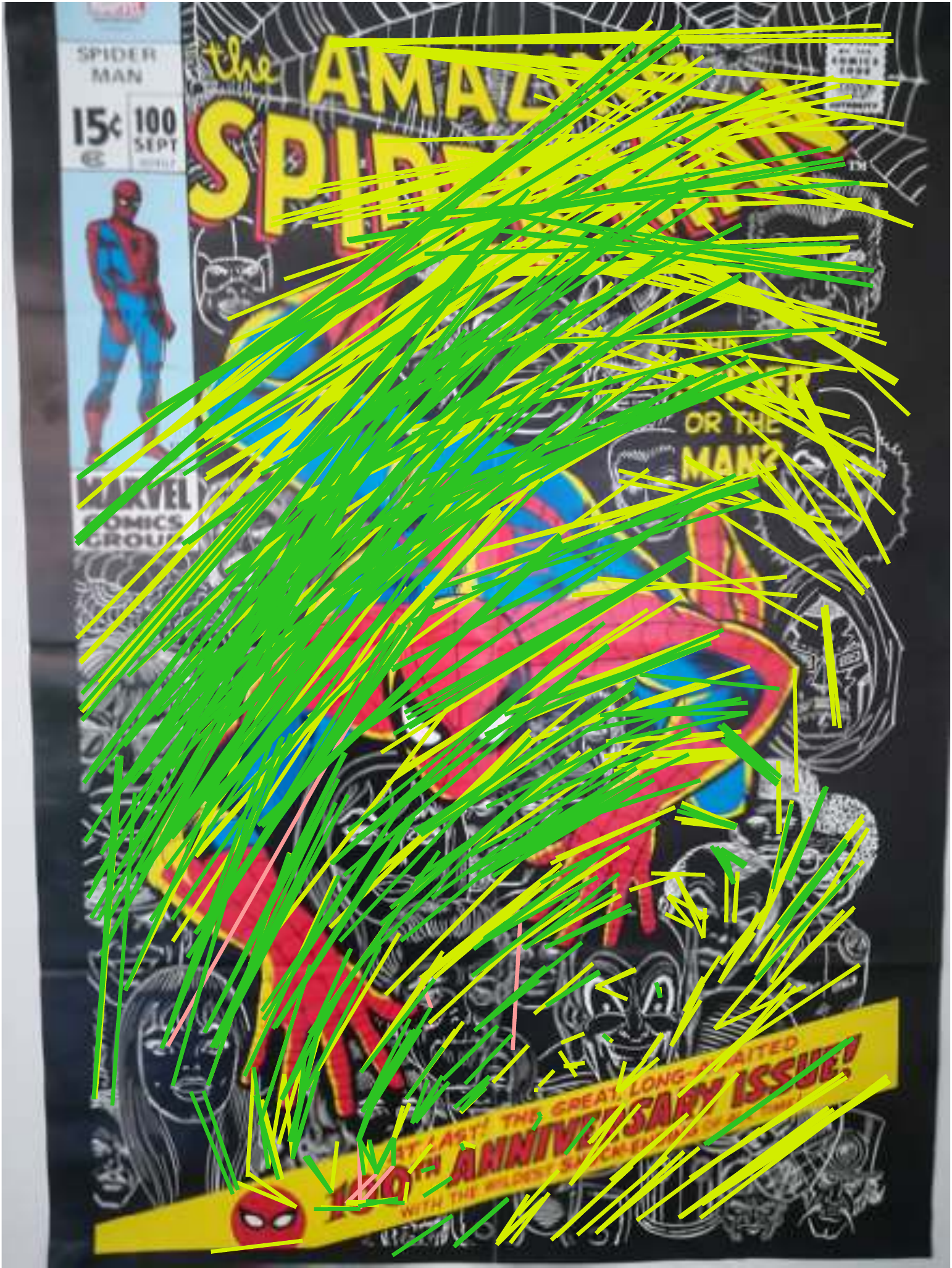}
	\includegraphics[height=7.5em]{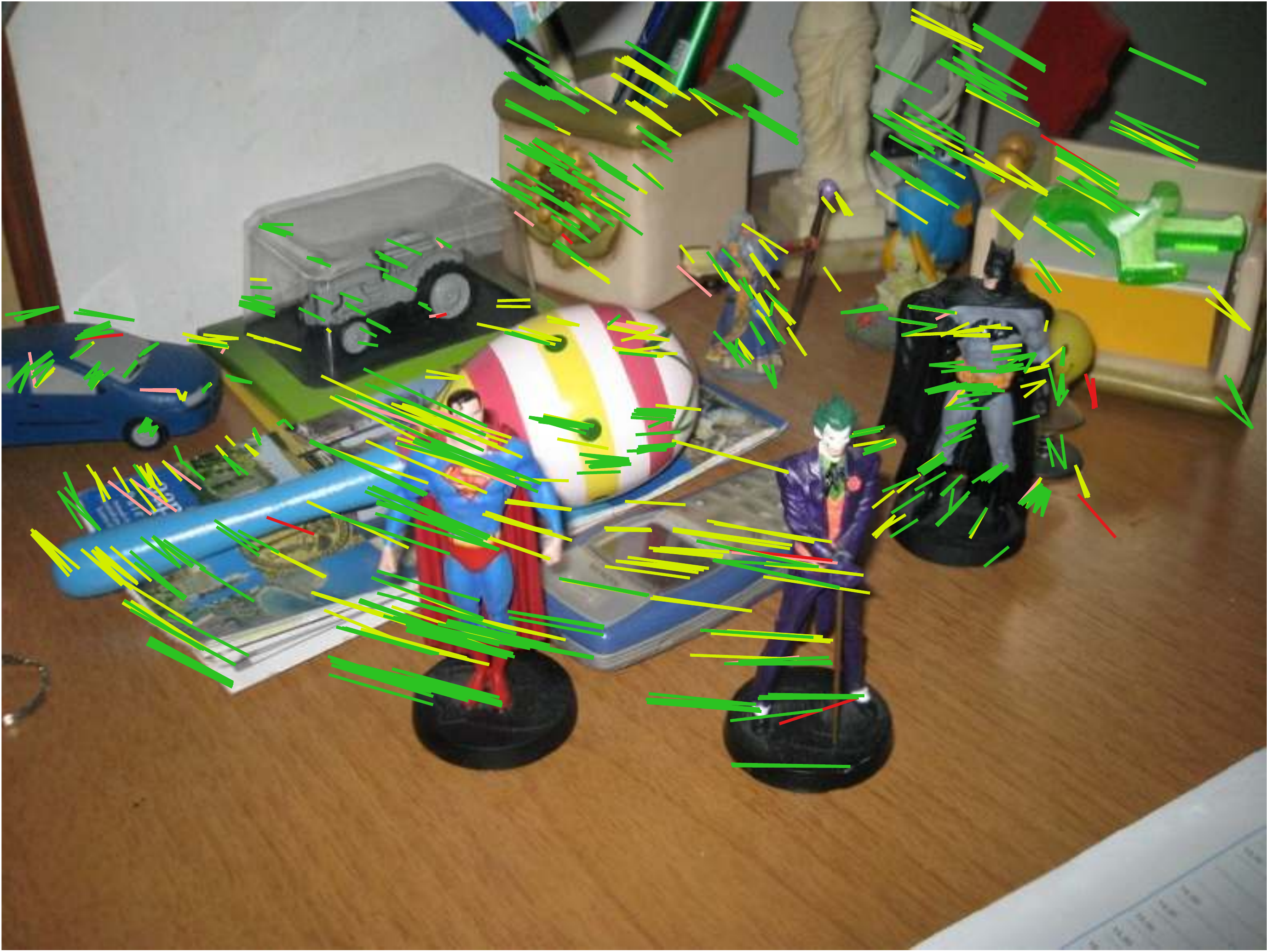}
	\includegraphics[height=7.5em]{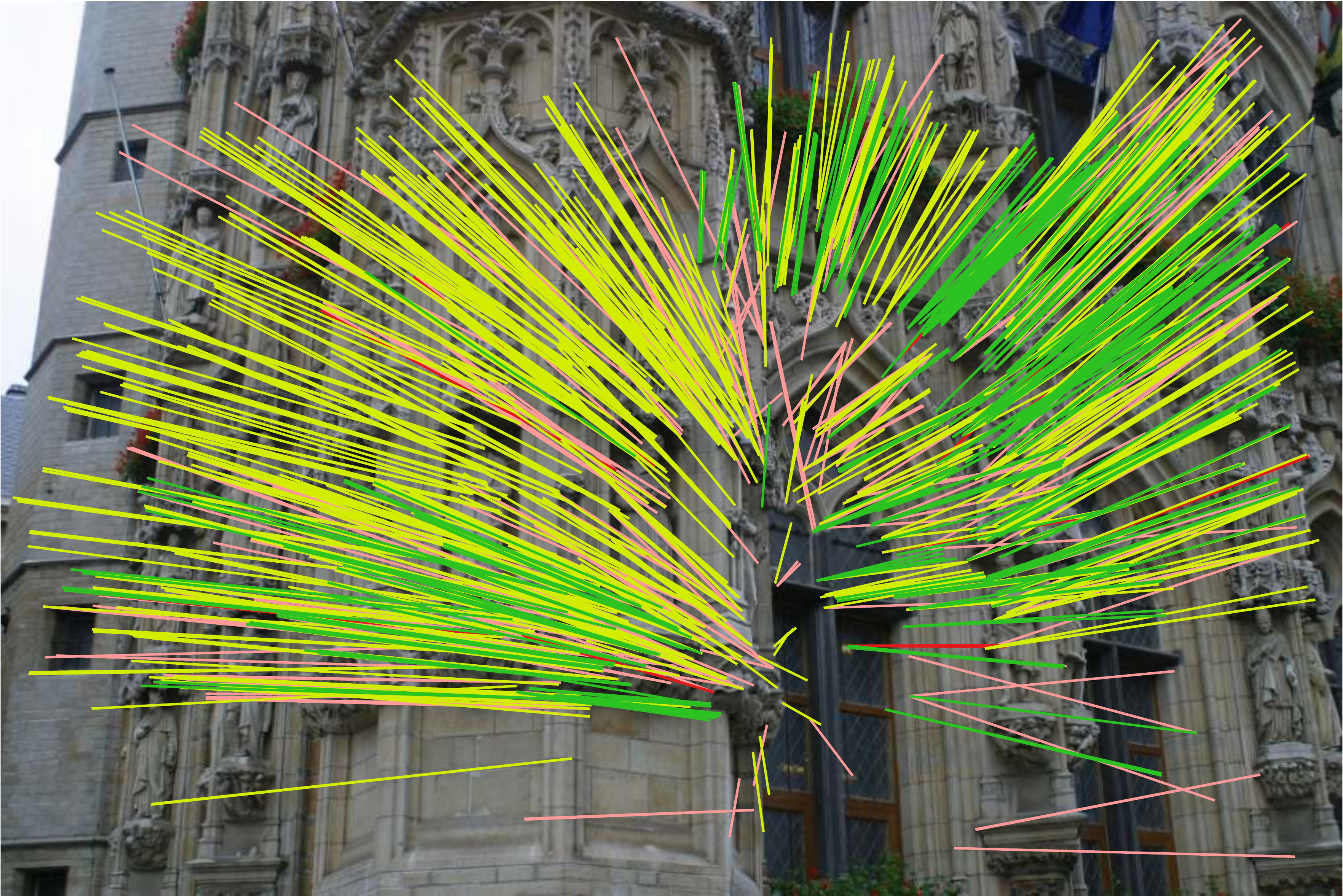}
	\includegraphics[height=7.5em]{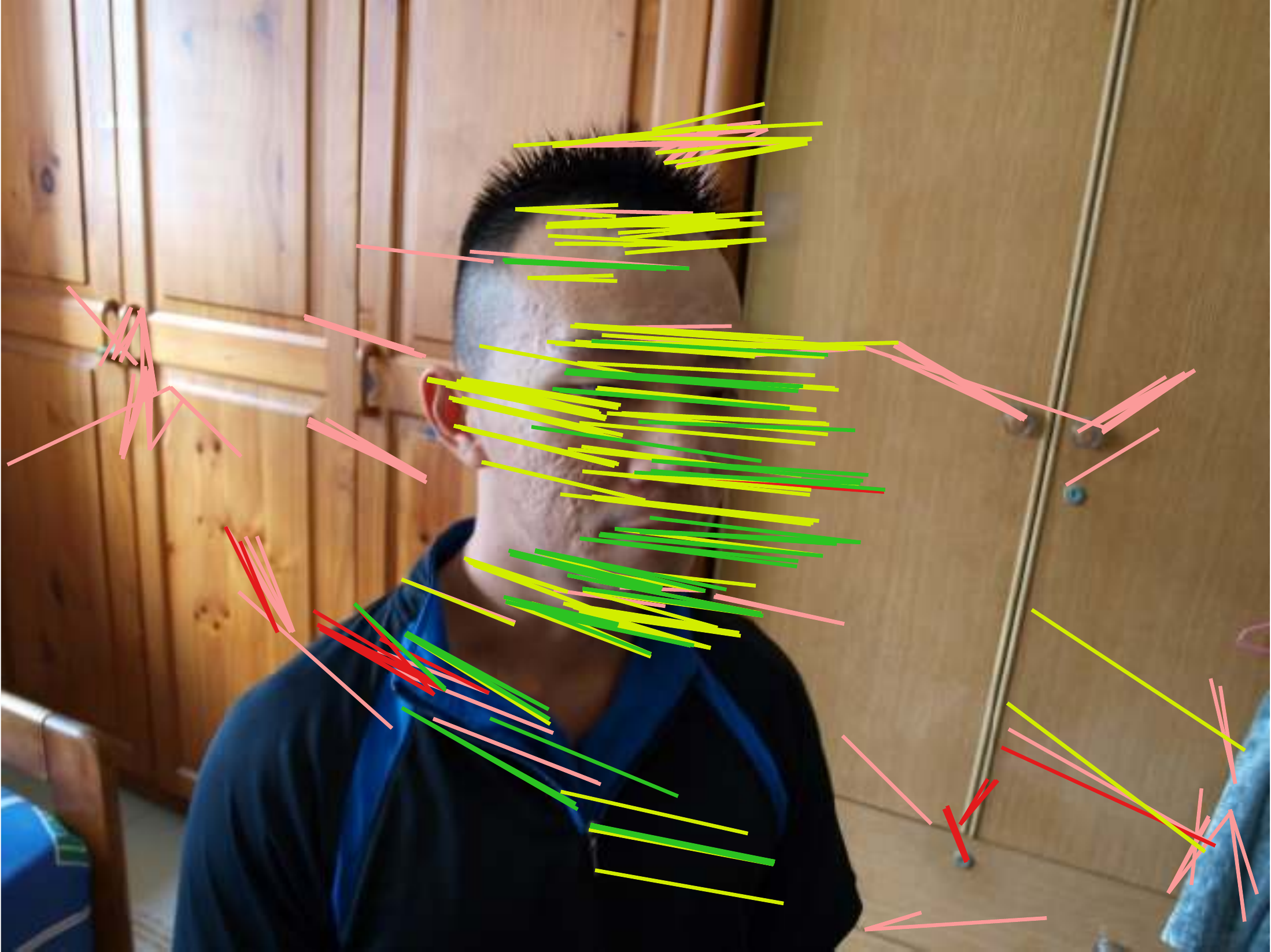}
	\\
	\vspace{0.5em}
	\rotatebox[origin=l]{90}{\mbox{\hspace{2em}ACNe}}
	\includegraphics[height=7.5em]{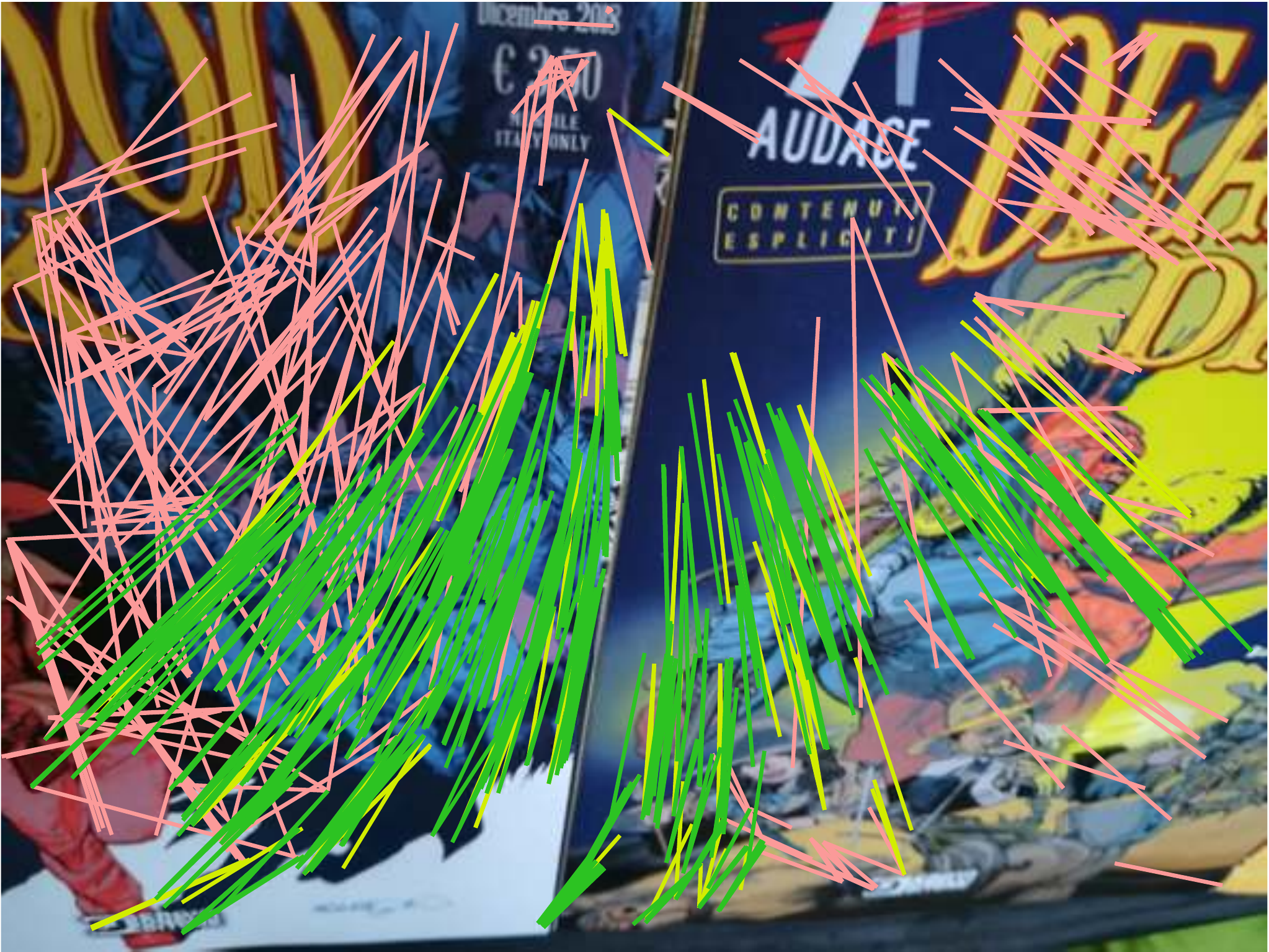}
	\includegraphics[height=7.5em]{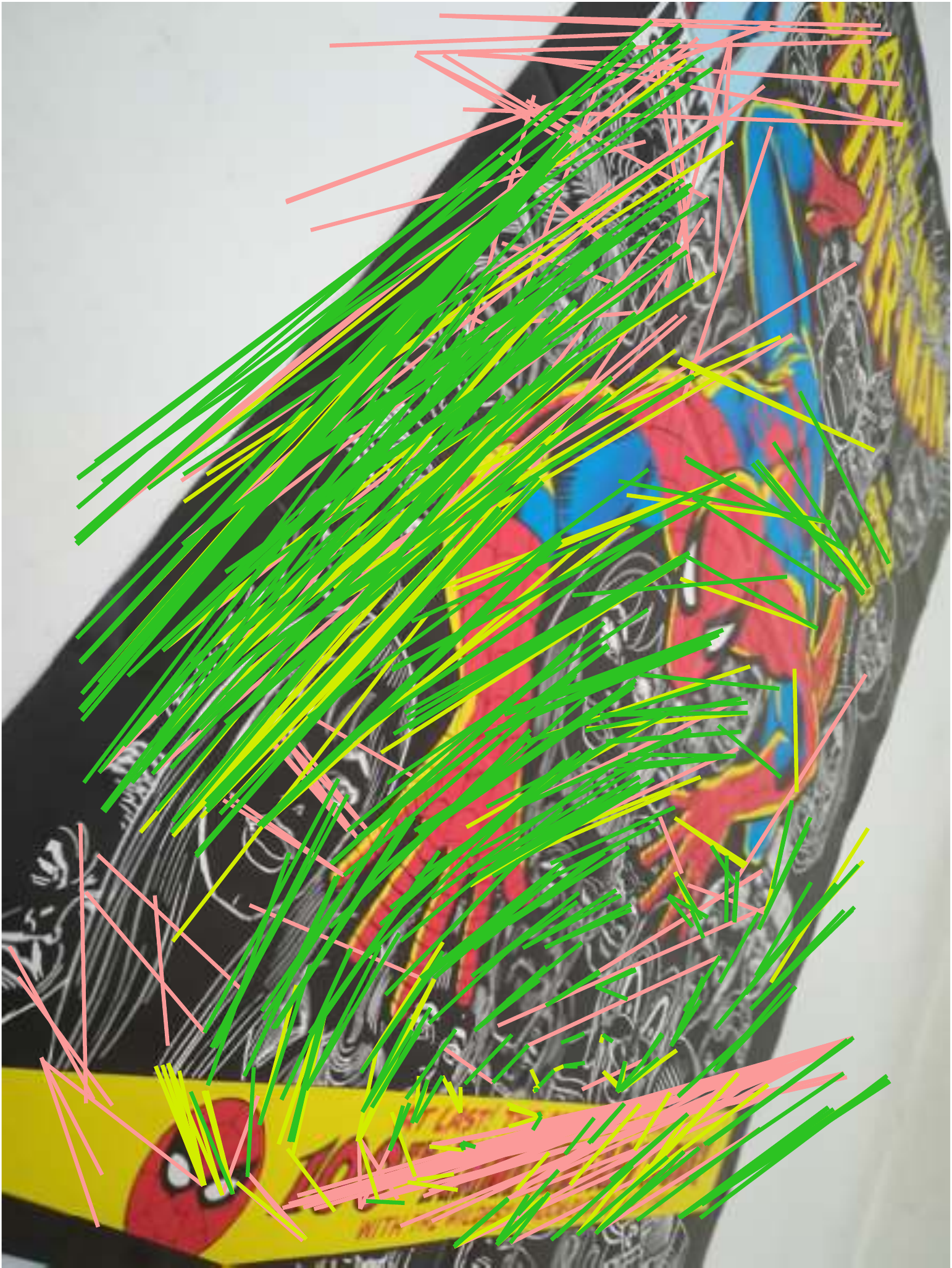}
	\includegraphics[height=7.5em]{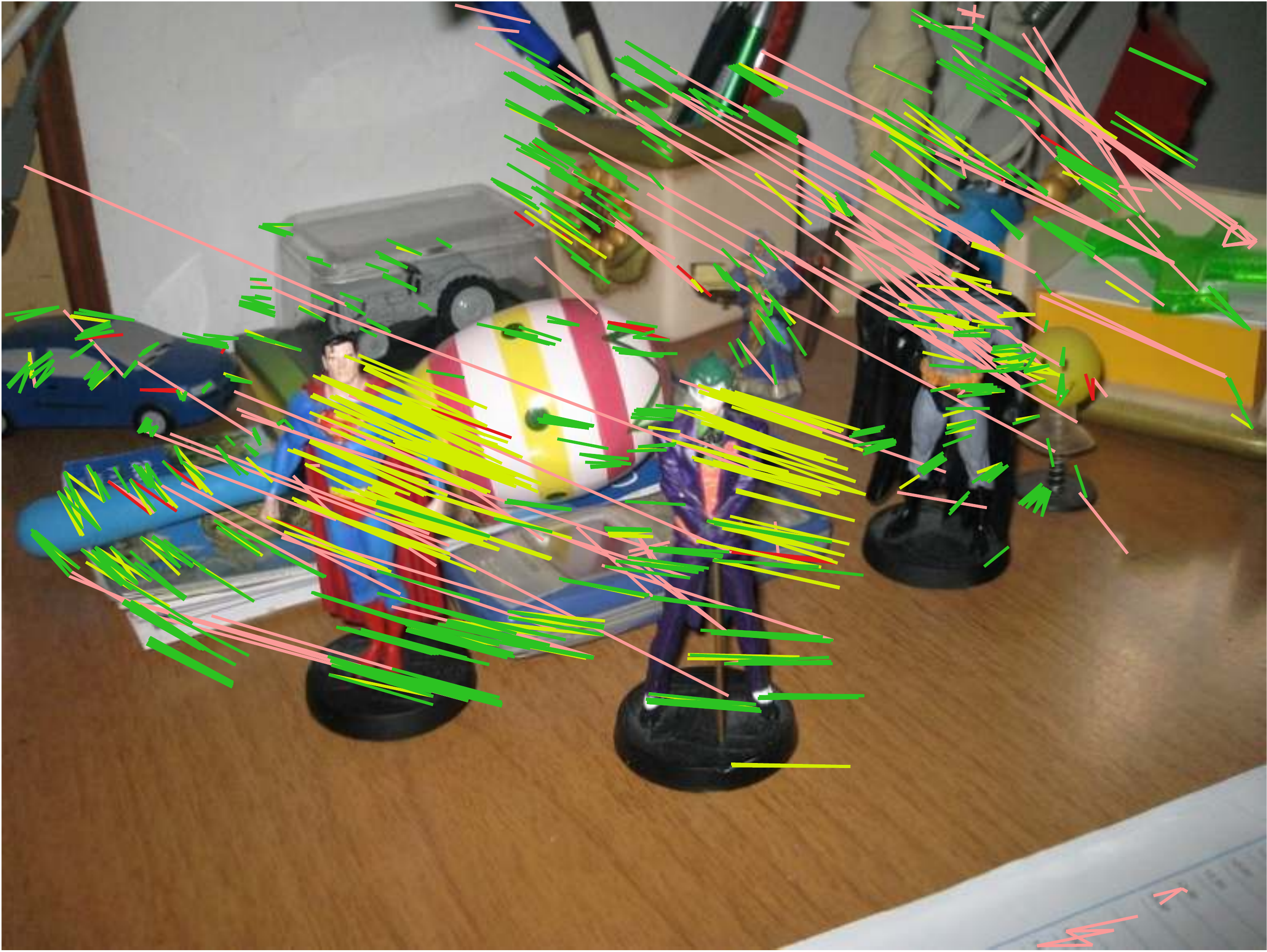}
	\includegraphics[height=7.5em]{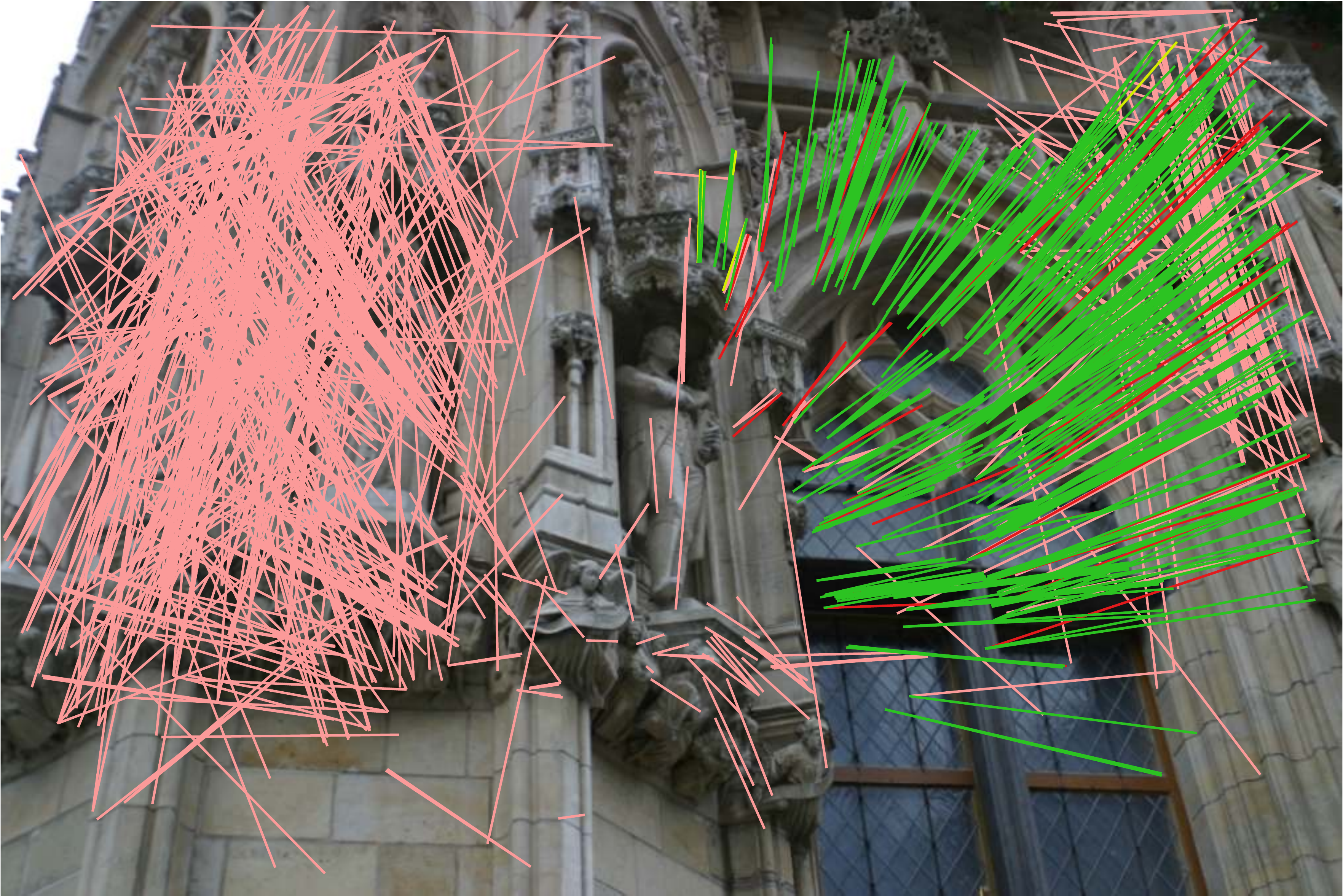}
	\includegraphics[height=7.5em]{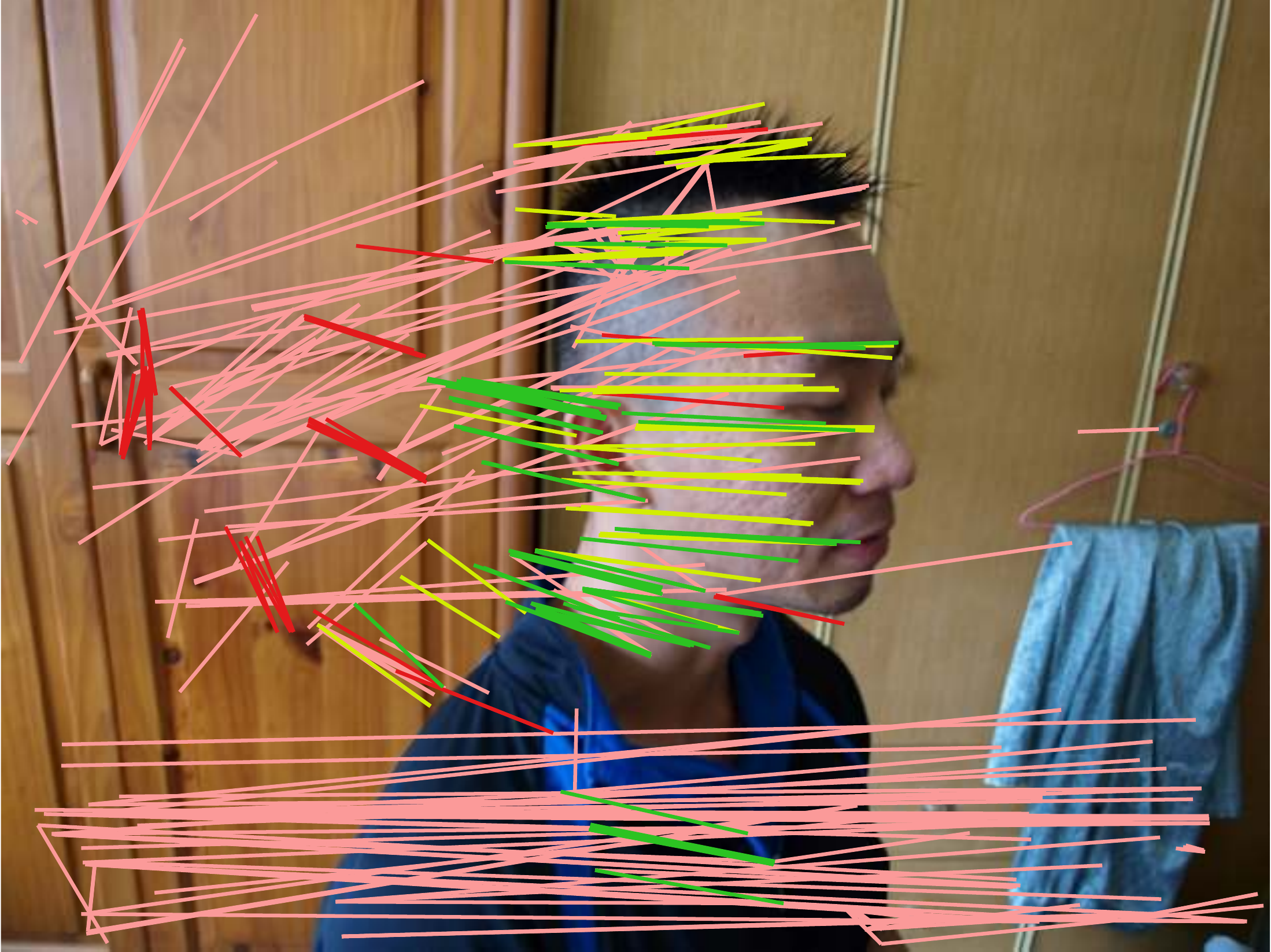}
	\\
	\vspace{0.5em}
	\rotatebox[origin=l]{90}{\mbox{\hspace{2em}PFM}}
	\includegraphics[height=7.5em]{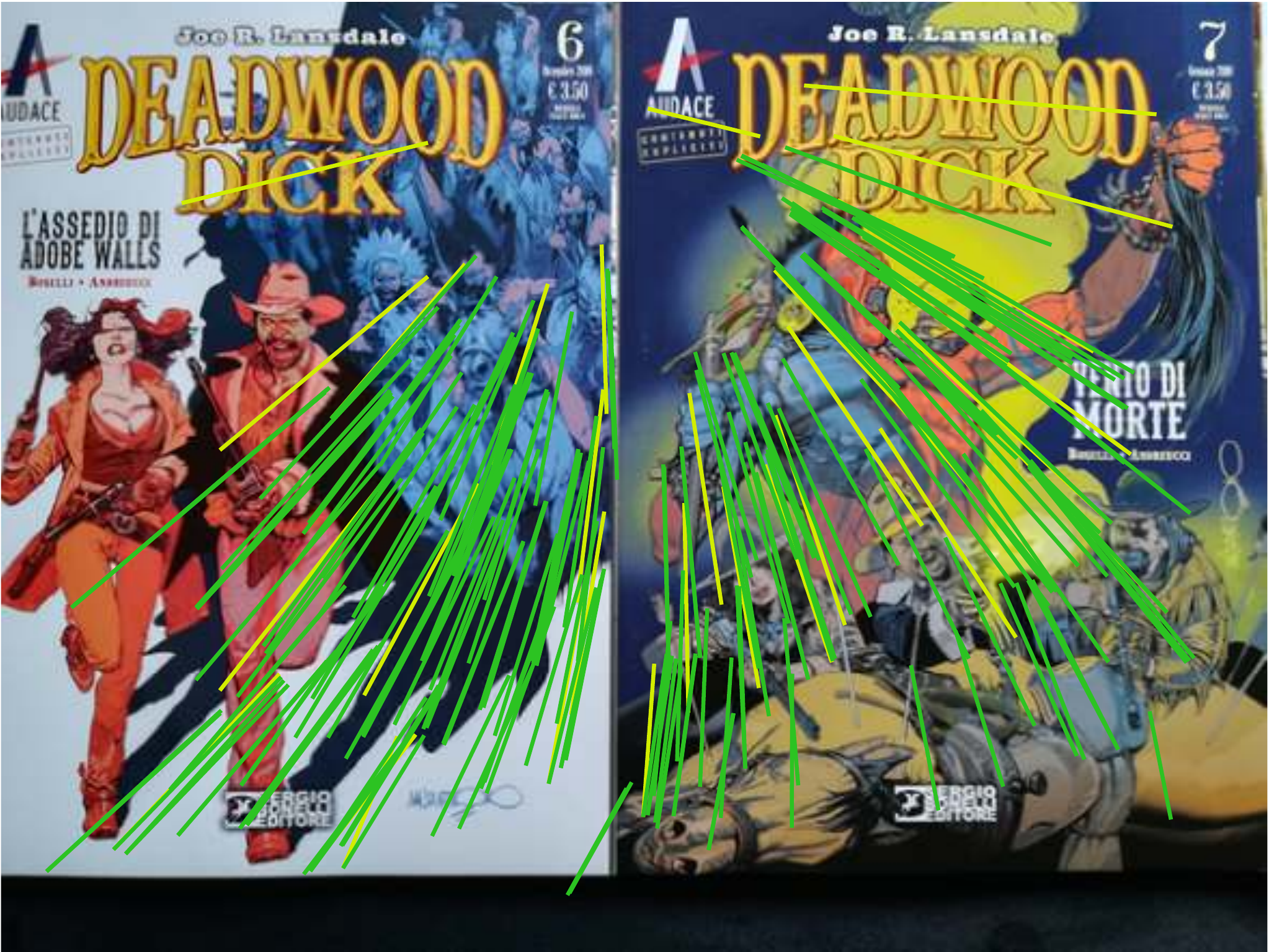}
	\includegraphics[height=7.5em]{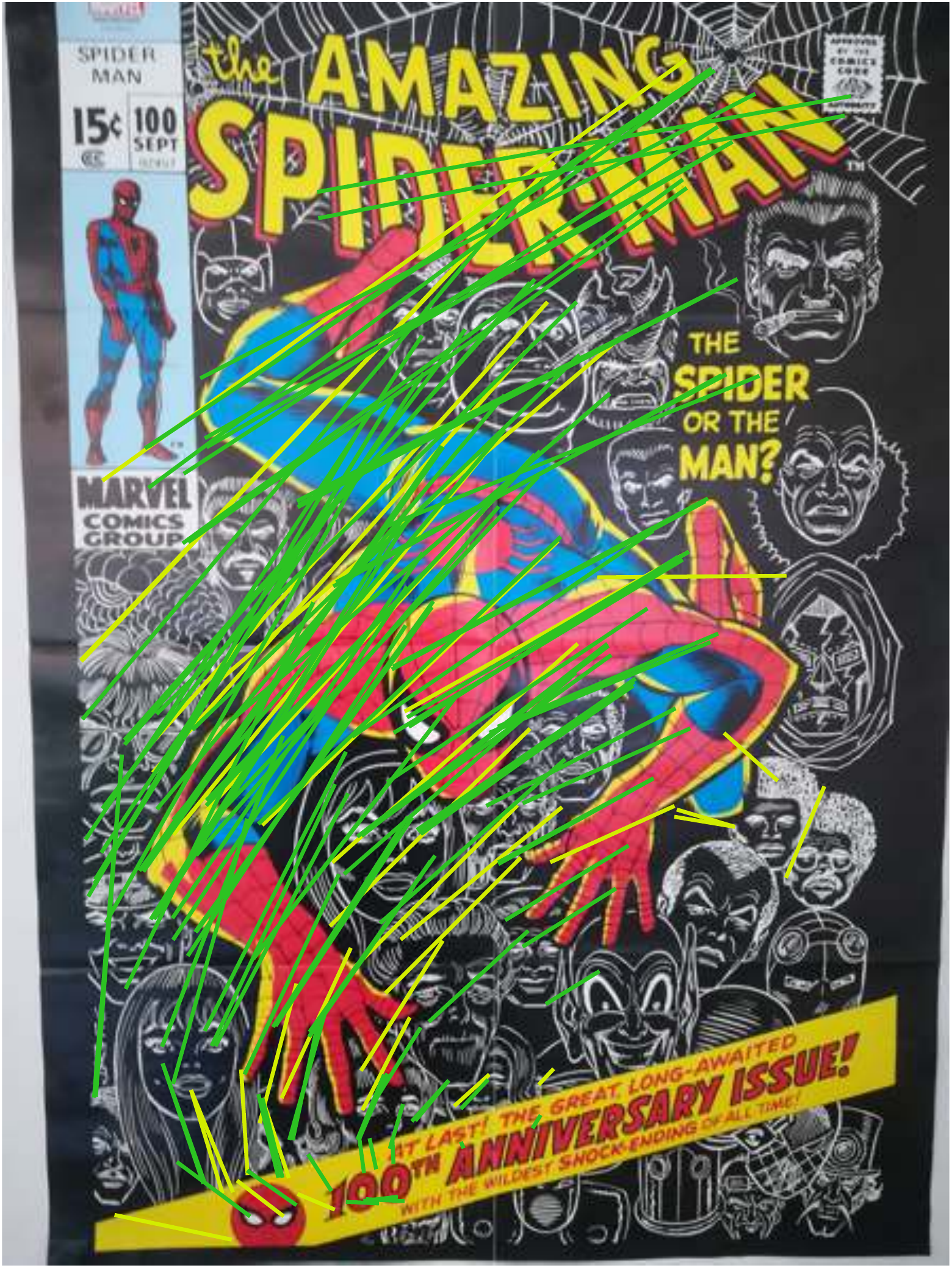}
	\includegraphics[height=7.5em]{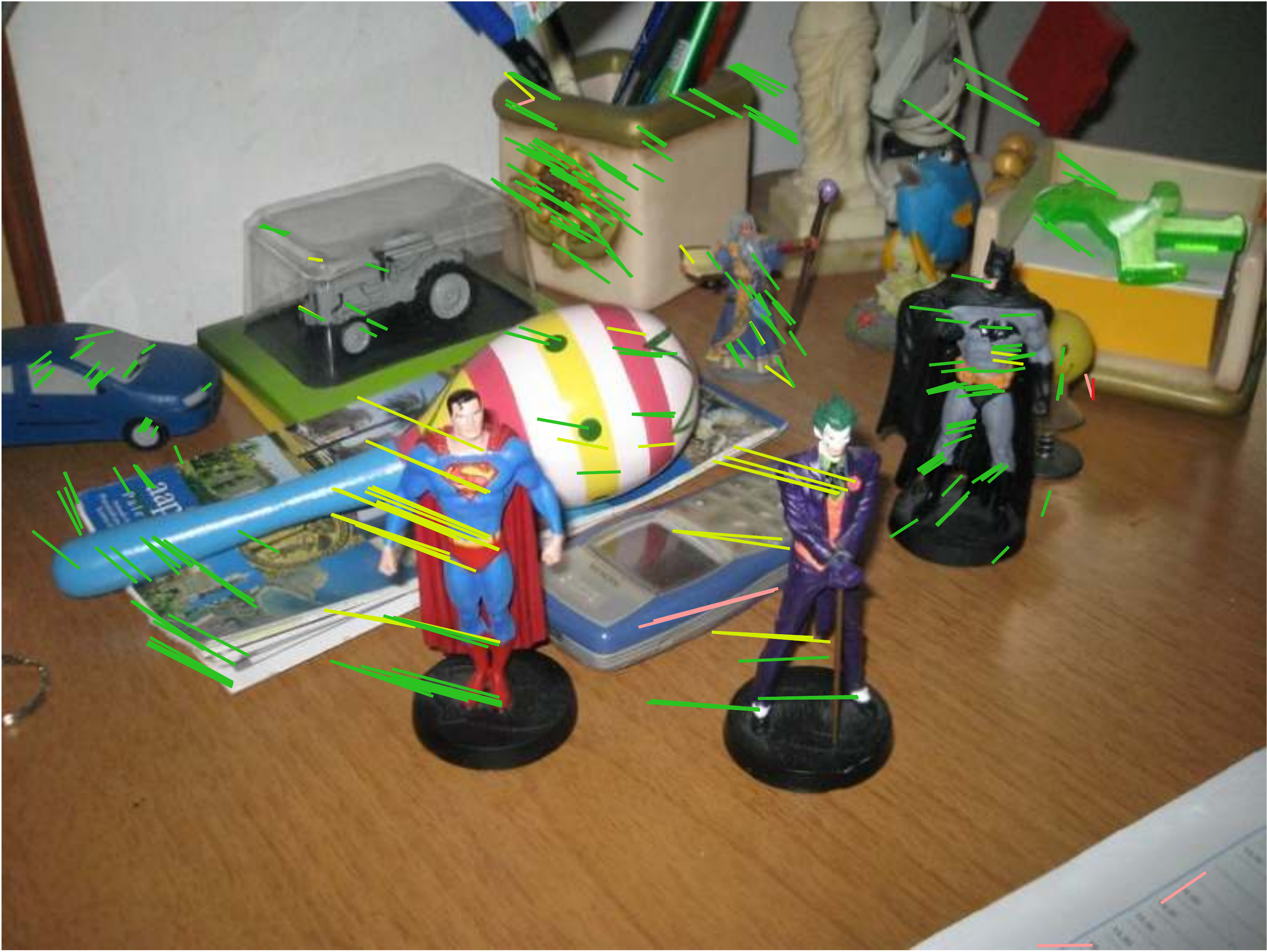}
	\includegraphics[height=7.5em]{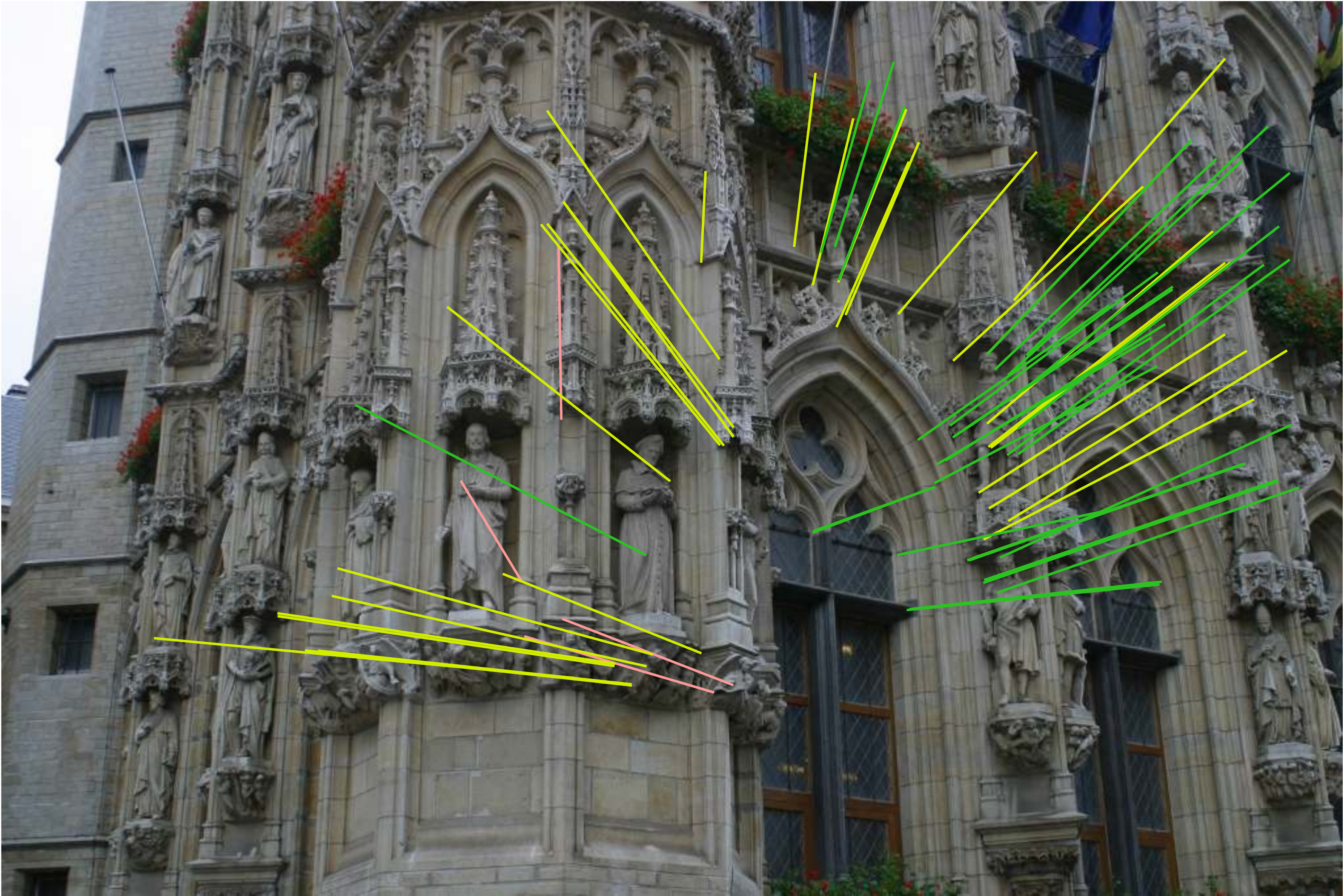}
	\includegraphics[height=7.5em]{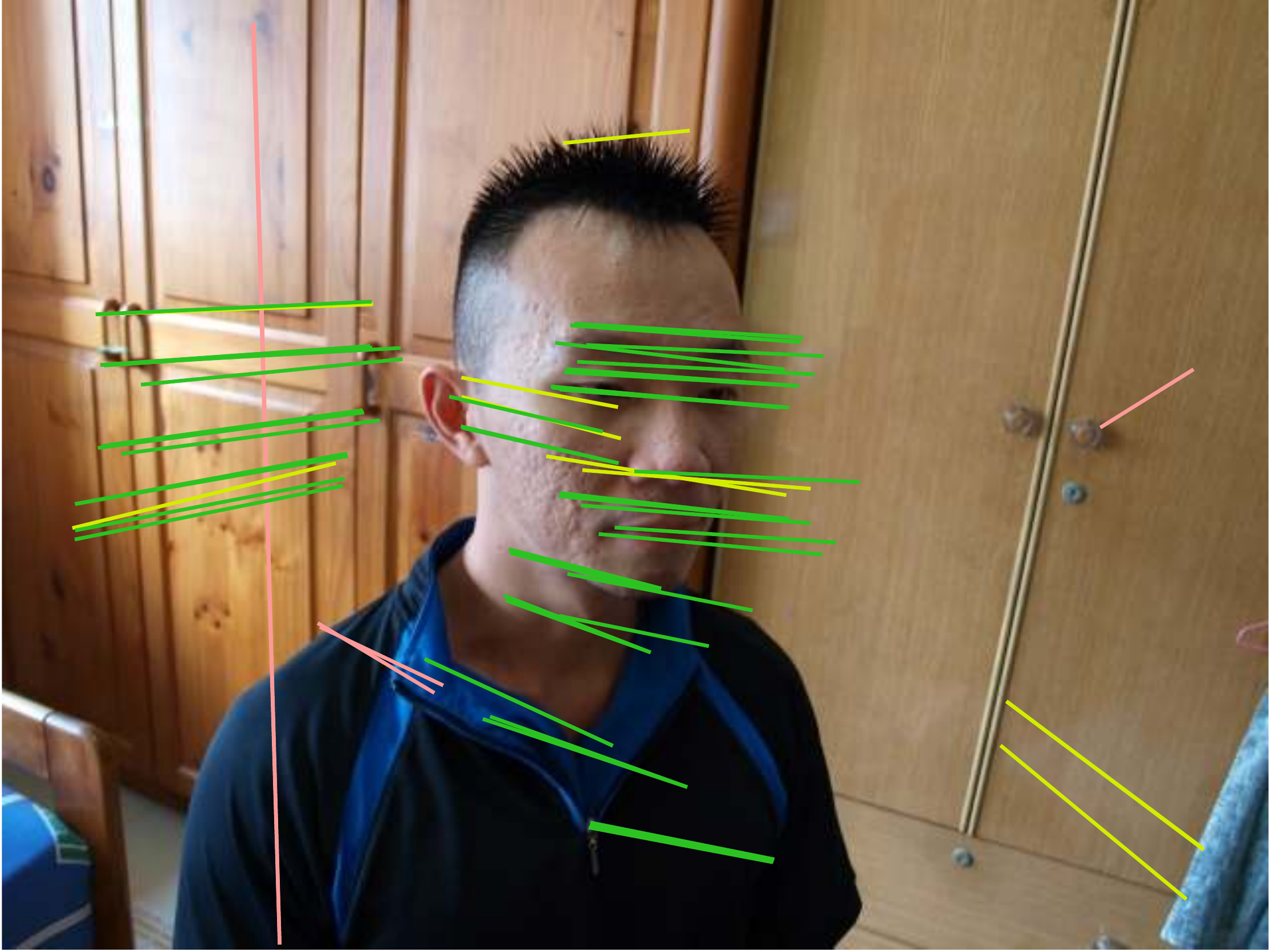}
	\\
	\vspace{0.5em}
	\rotatebox[origin=l]{90}{\mbox{\hspace{2em}PGM}}
	\includegraphics[height=7.5em]{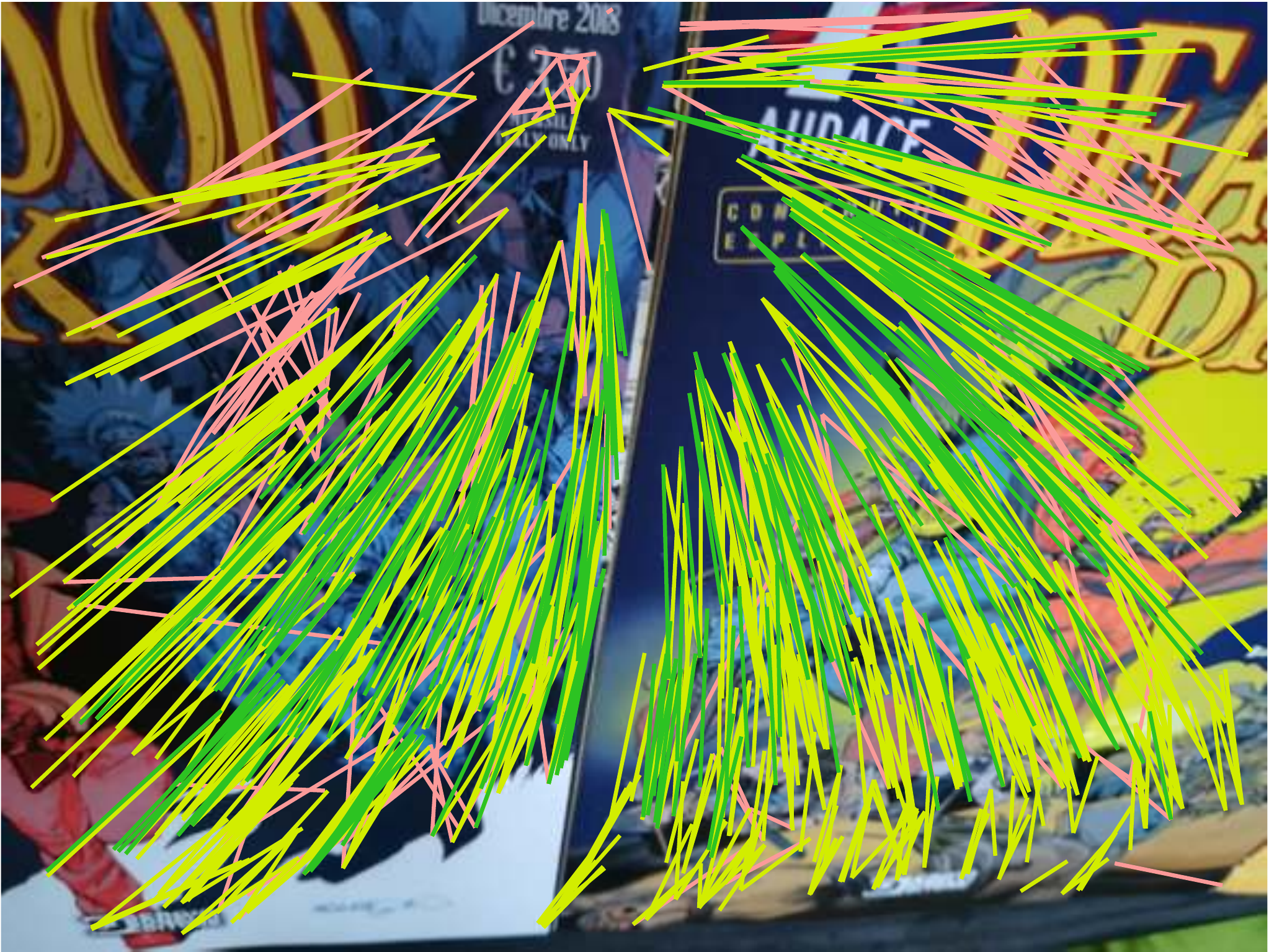}
	\includegraphics[height=7.5em]{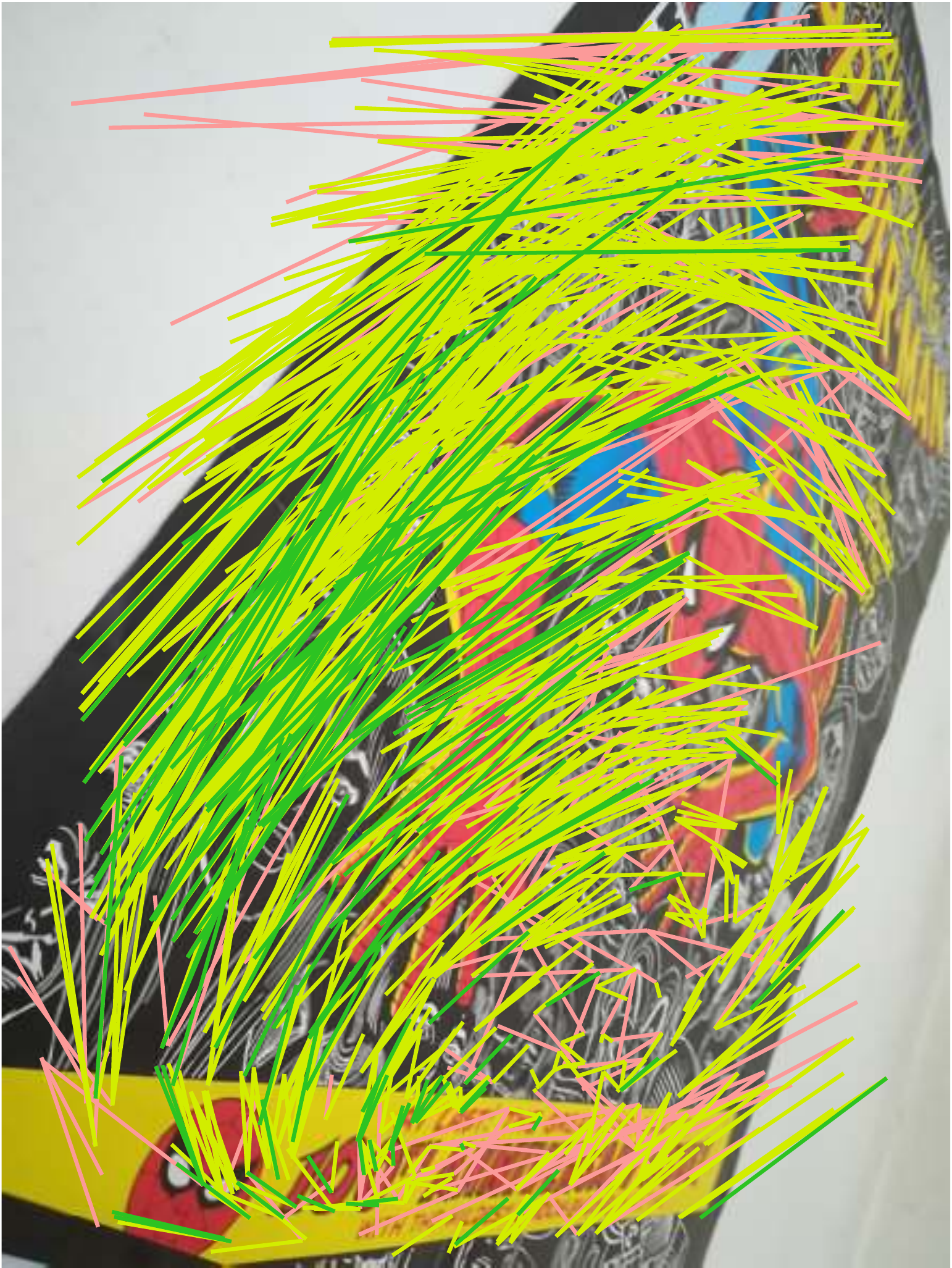}
	\includegraphics[height=7.5em]{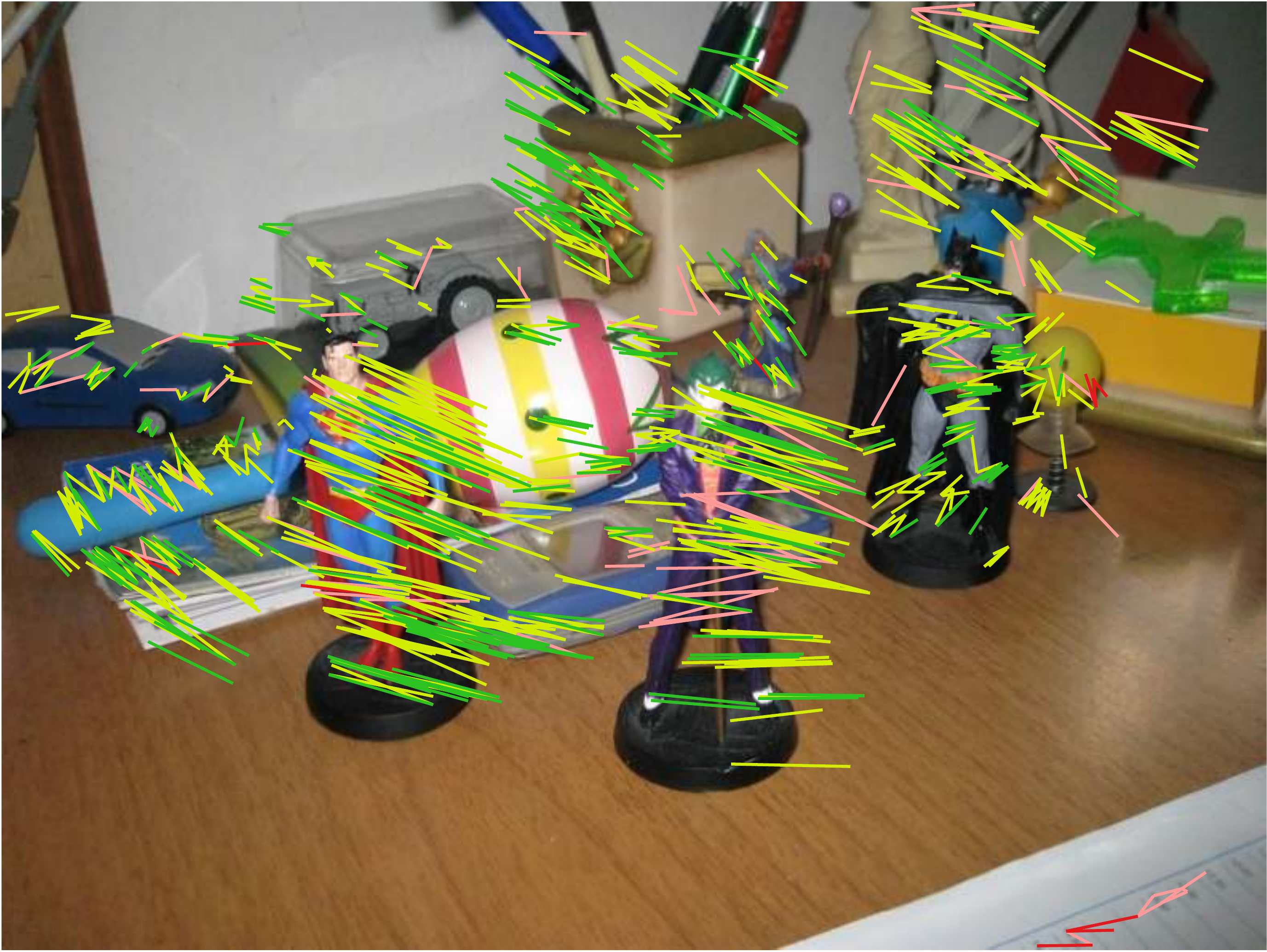}
	\includegraphics[height=7.5em]{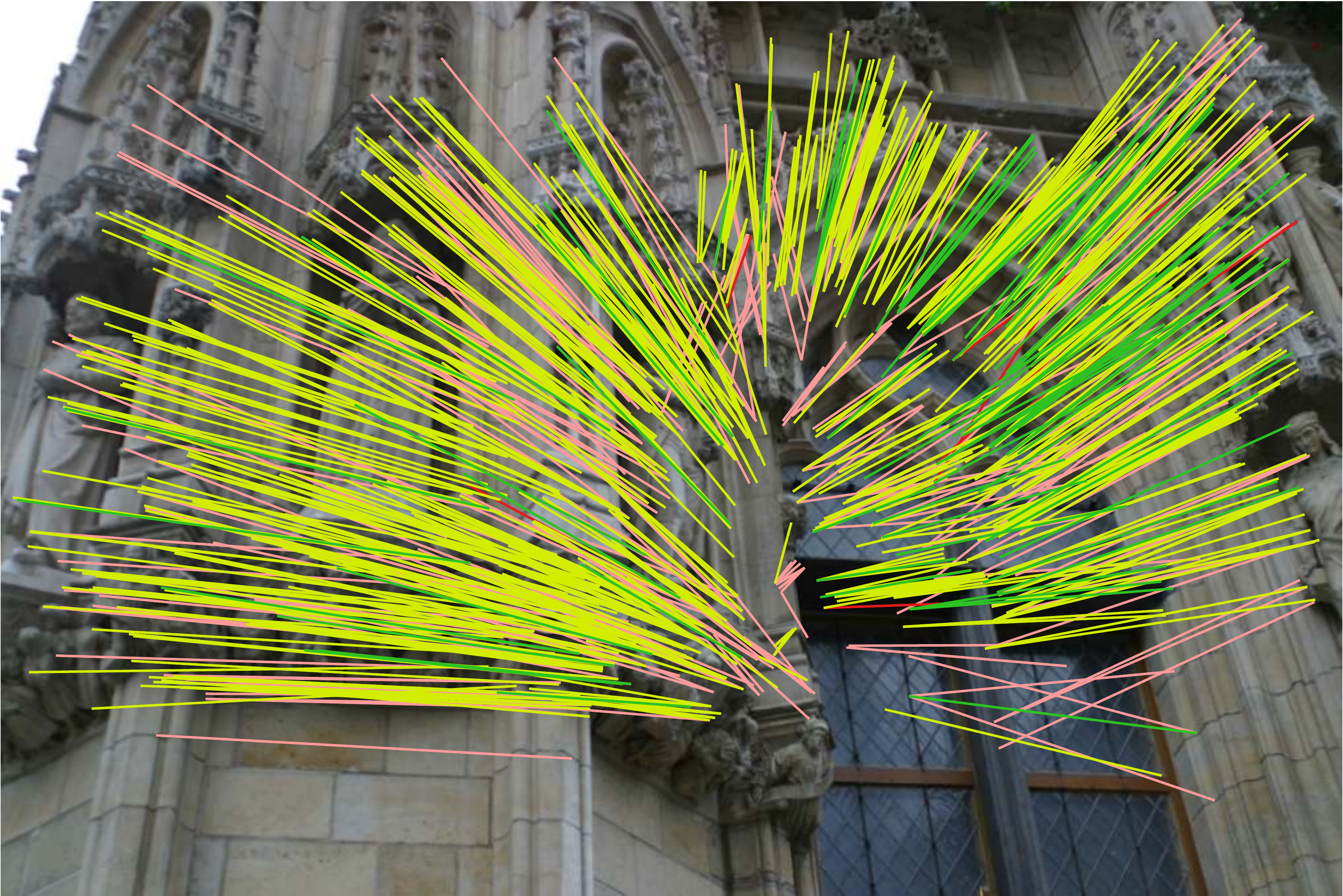}
	\includegraphics[height=7.5em]{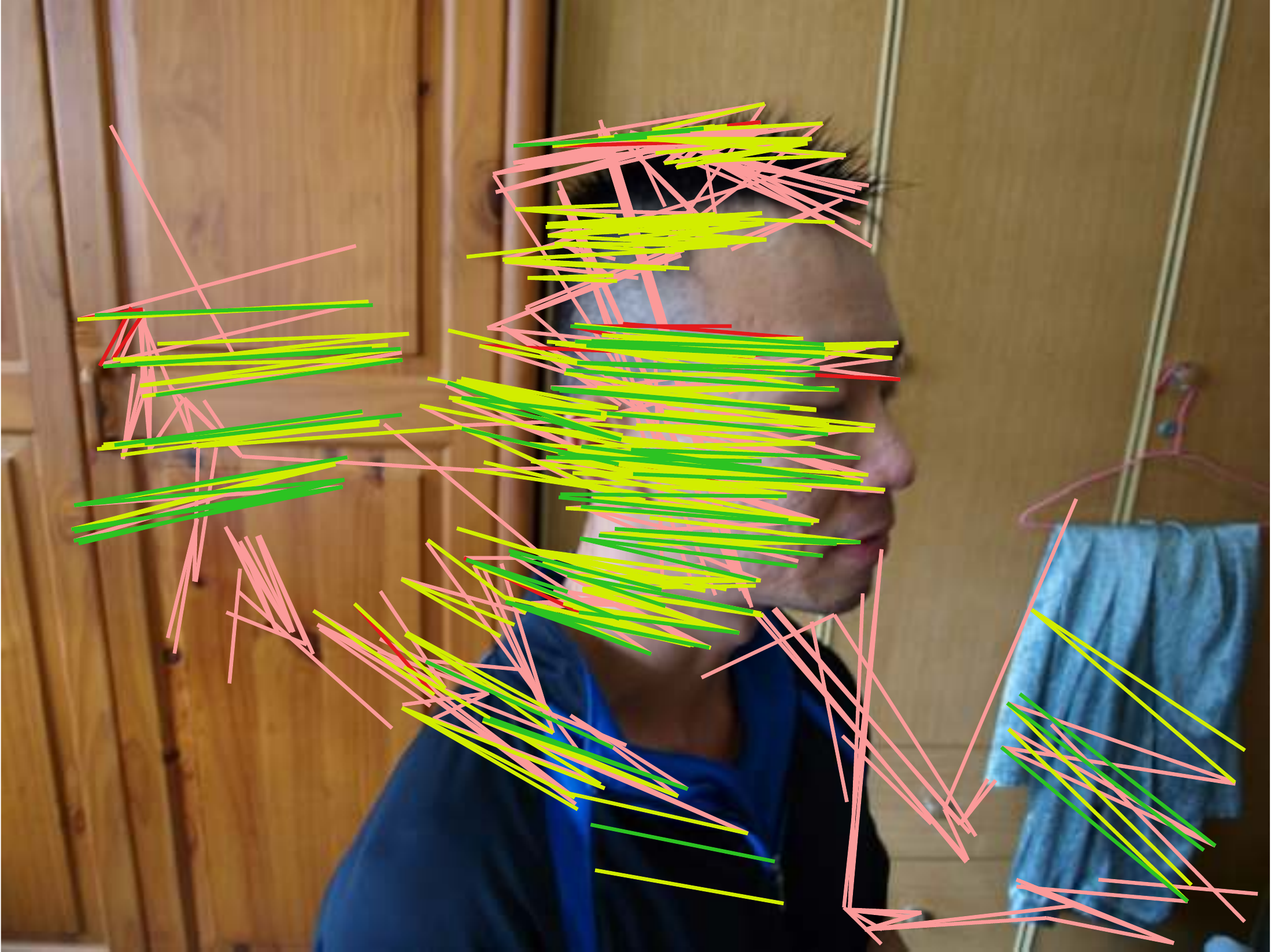}
	\\
	\vspace{0.5em}
	\rotatebox[origin=l]{90}{\mbox{\hspace{2em}SCV}}
	\includegraphics[height=7.5em]{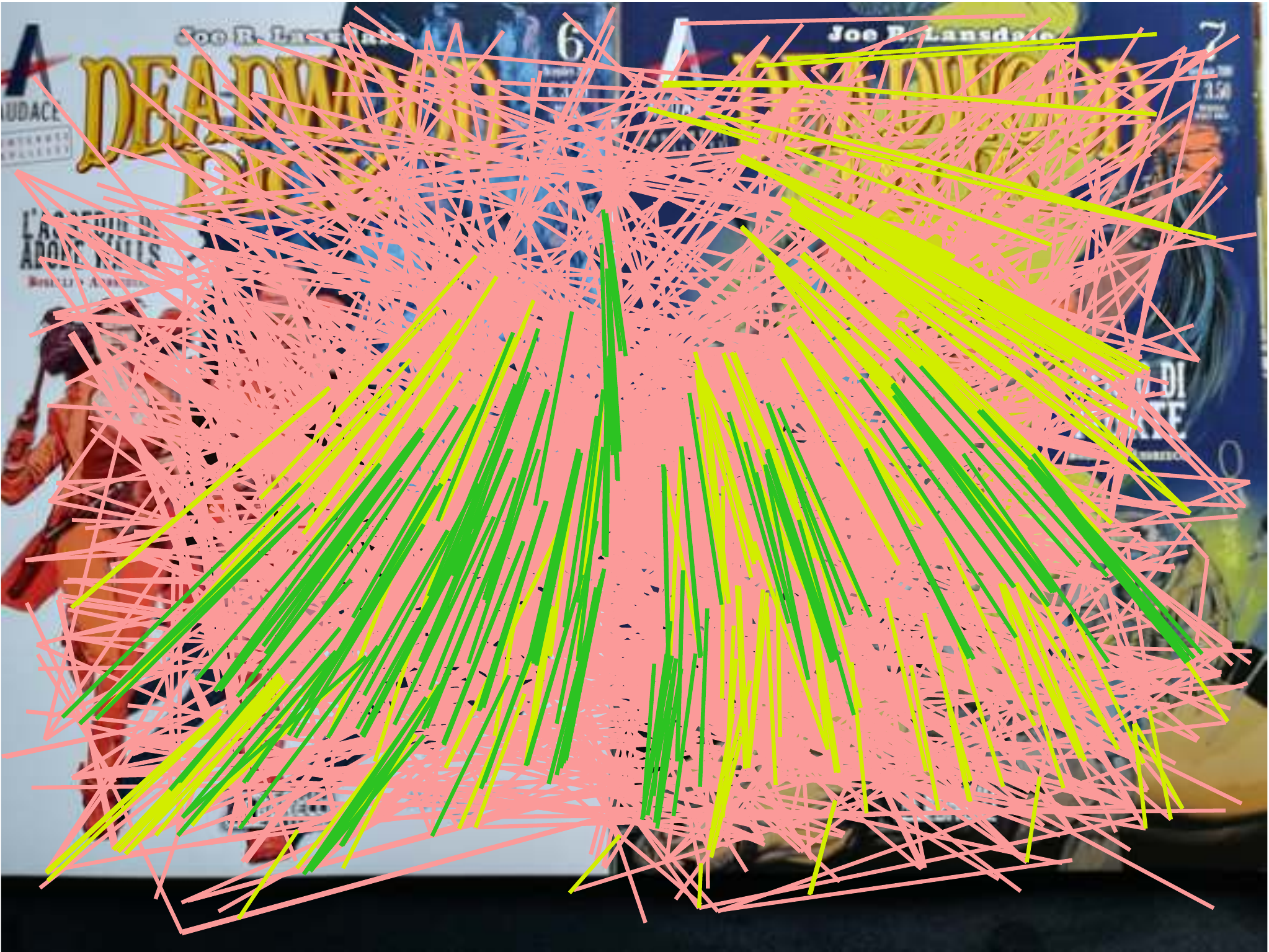}
	\includegraphics[height=7.5em]{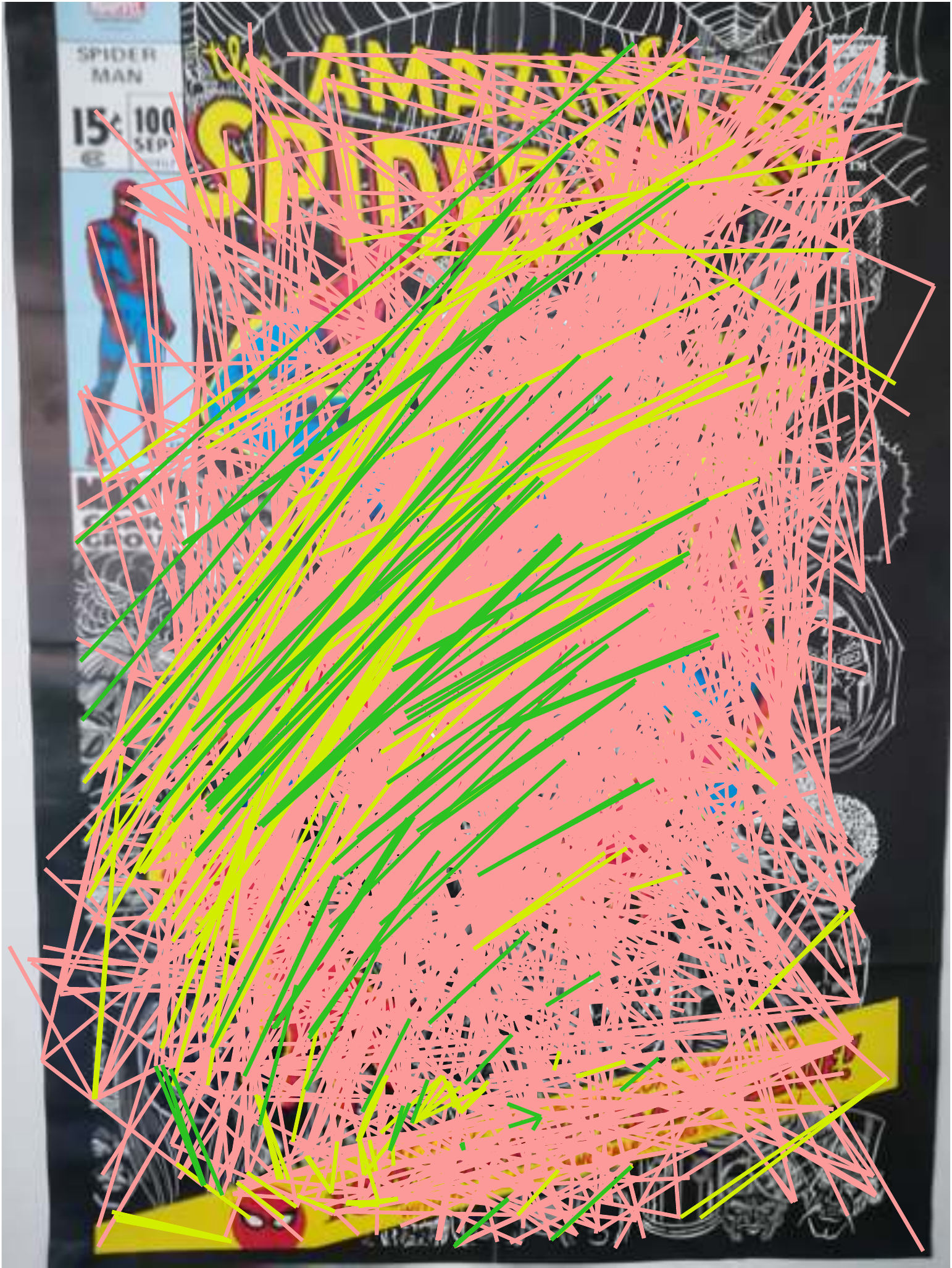}
	\includegraphics[height=7.5em]{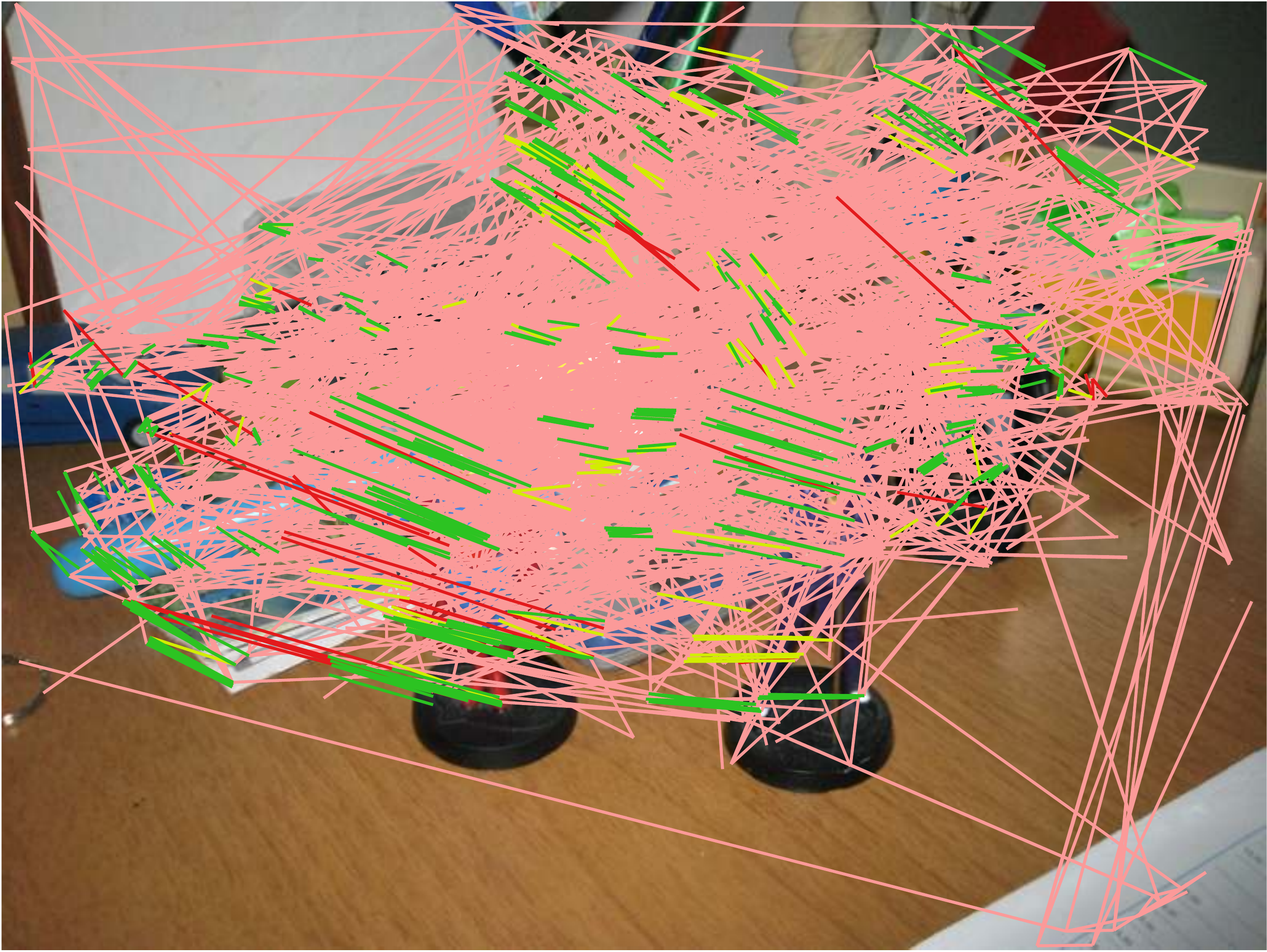}
	\includegraphics[height=7.5em]{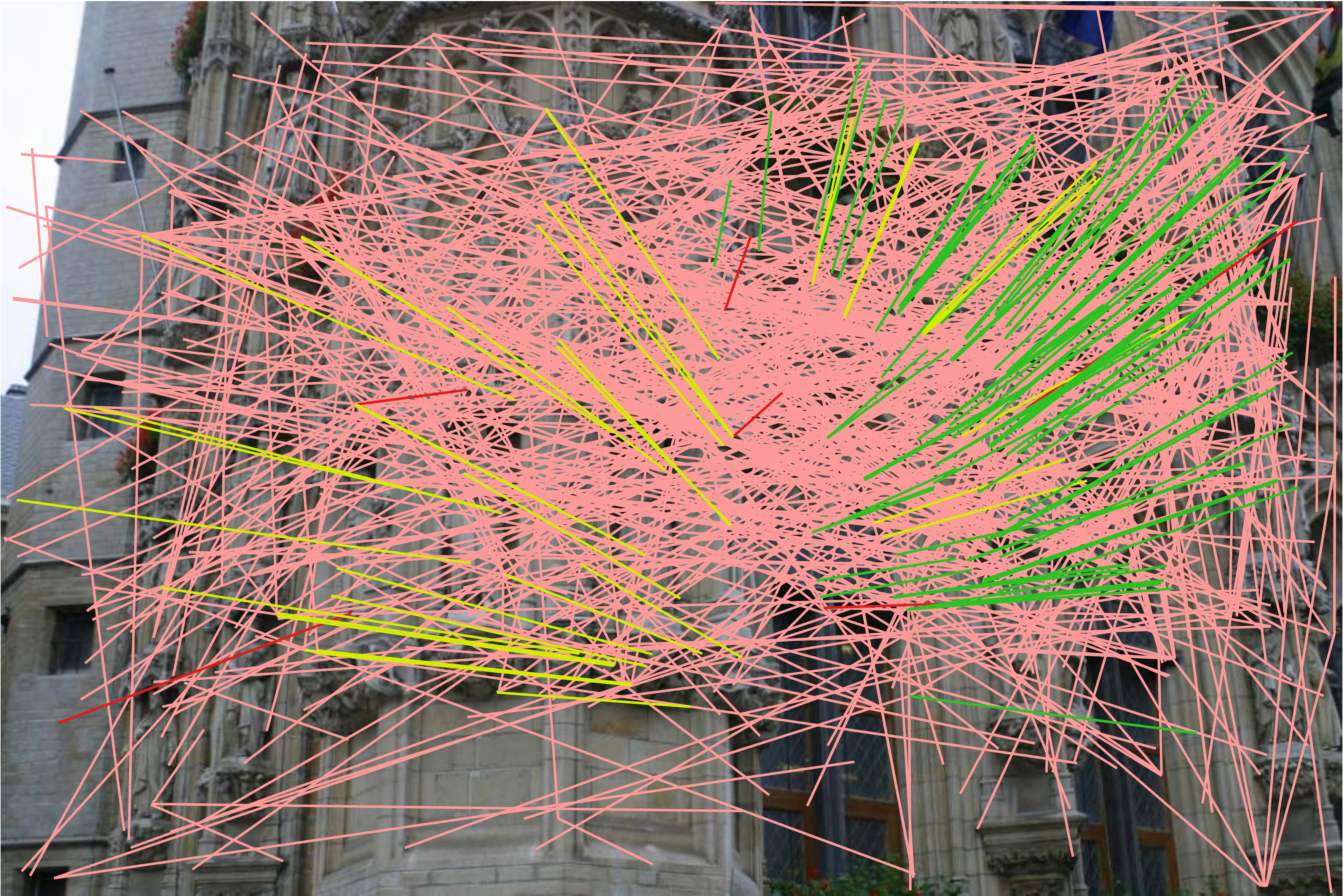}
	\includegraphics[height=7.5em]{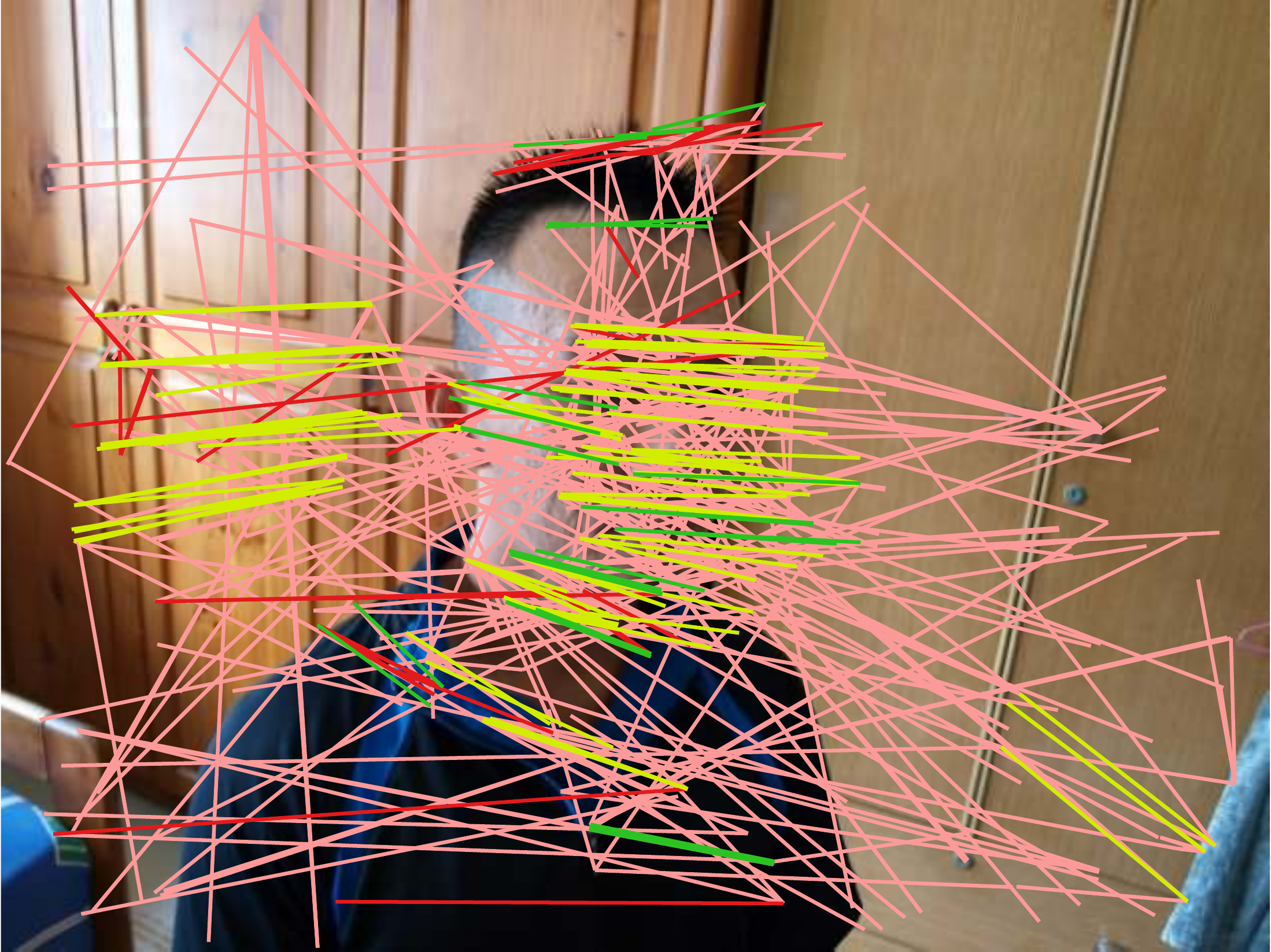}
	\\
	\vspace{0.5em}
	\rotatebox[origin=l]{90}{\mbox{\hspace{2em}BM}}
	\includegraphics[height=7.5em]{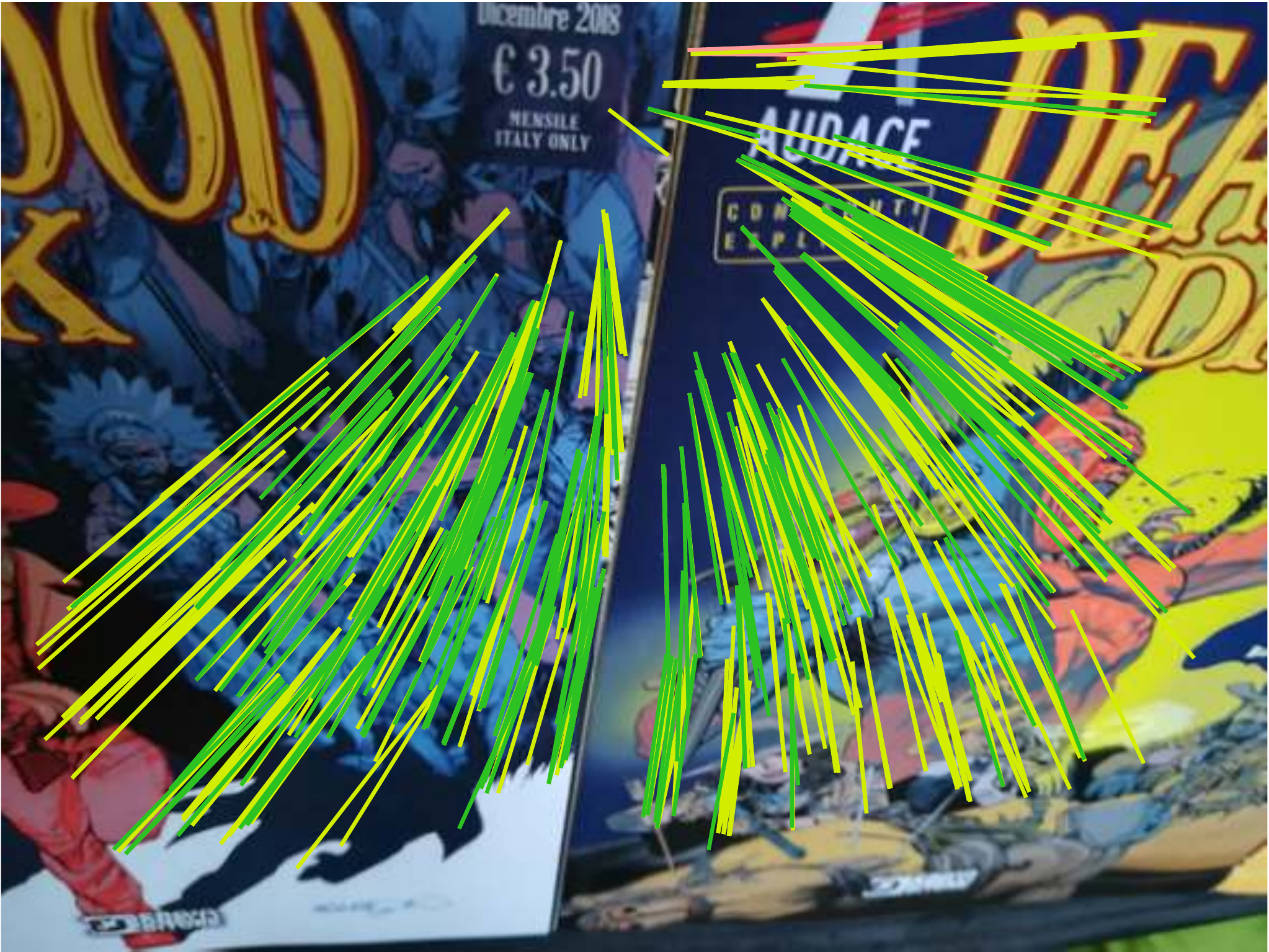}
	\includegraphics[height=7.5em]{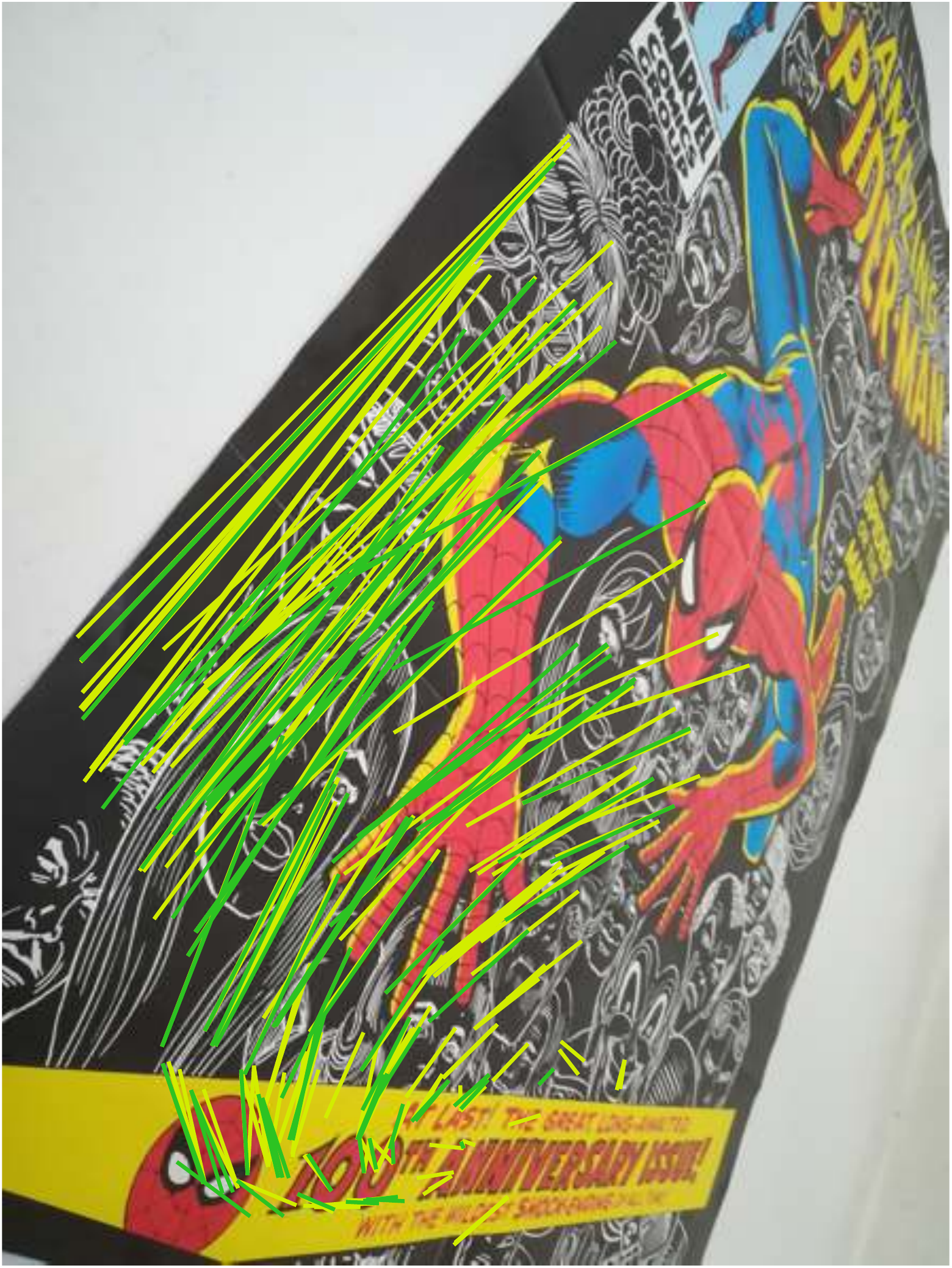}
	\includegraphics[height=7.5em]{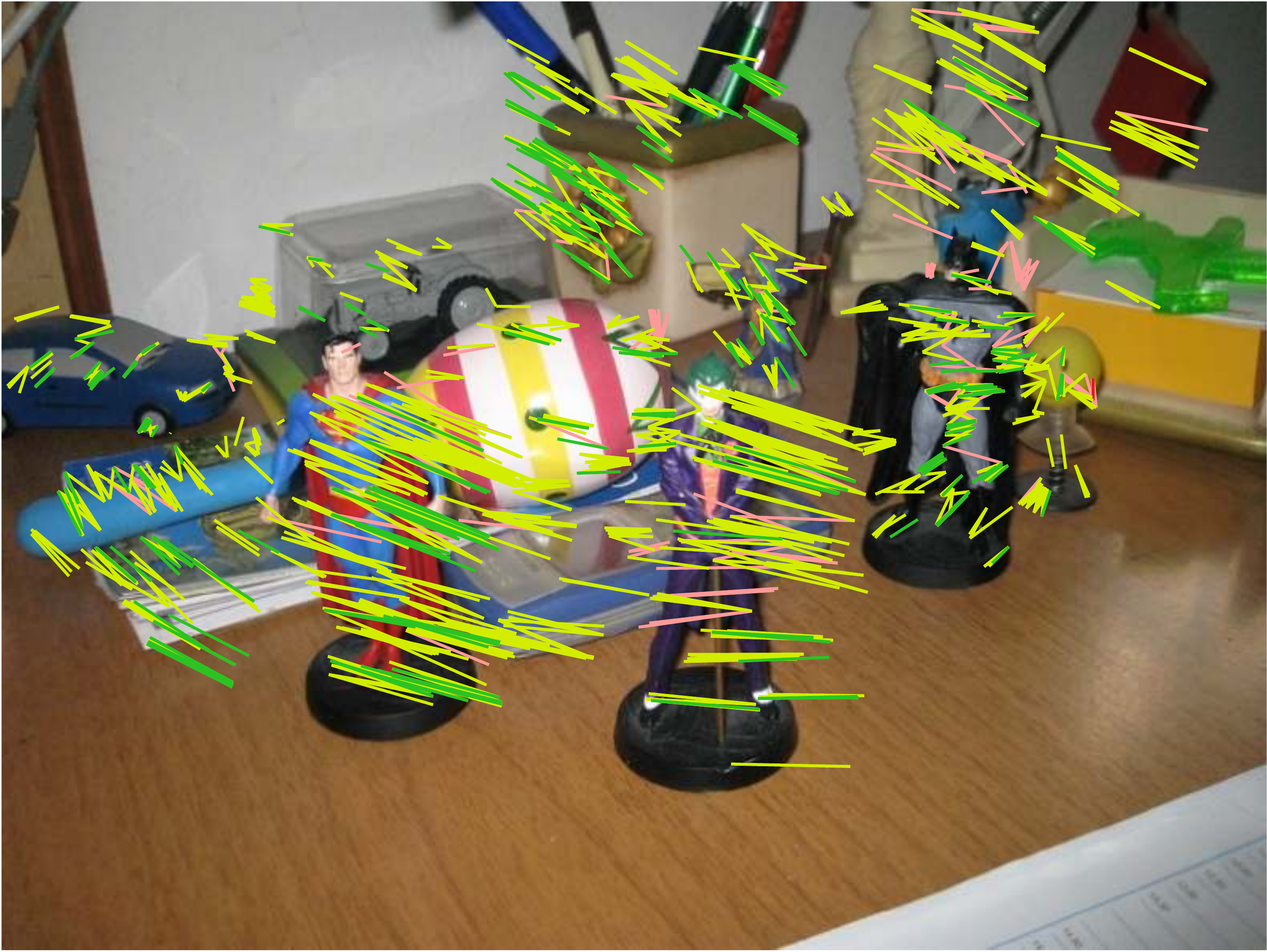}
	\includegraphics[height=7.5em]{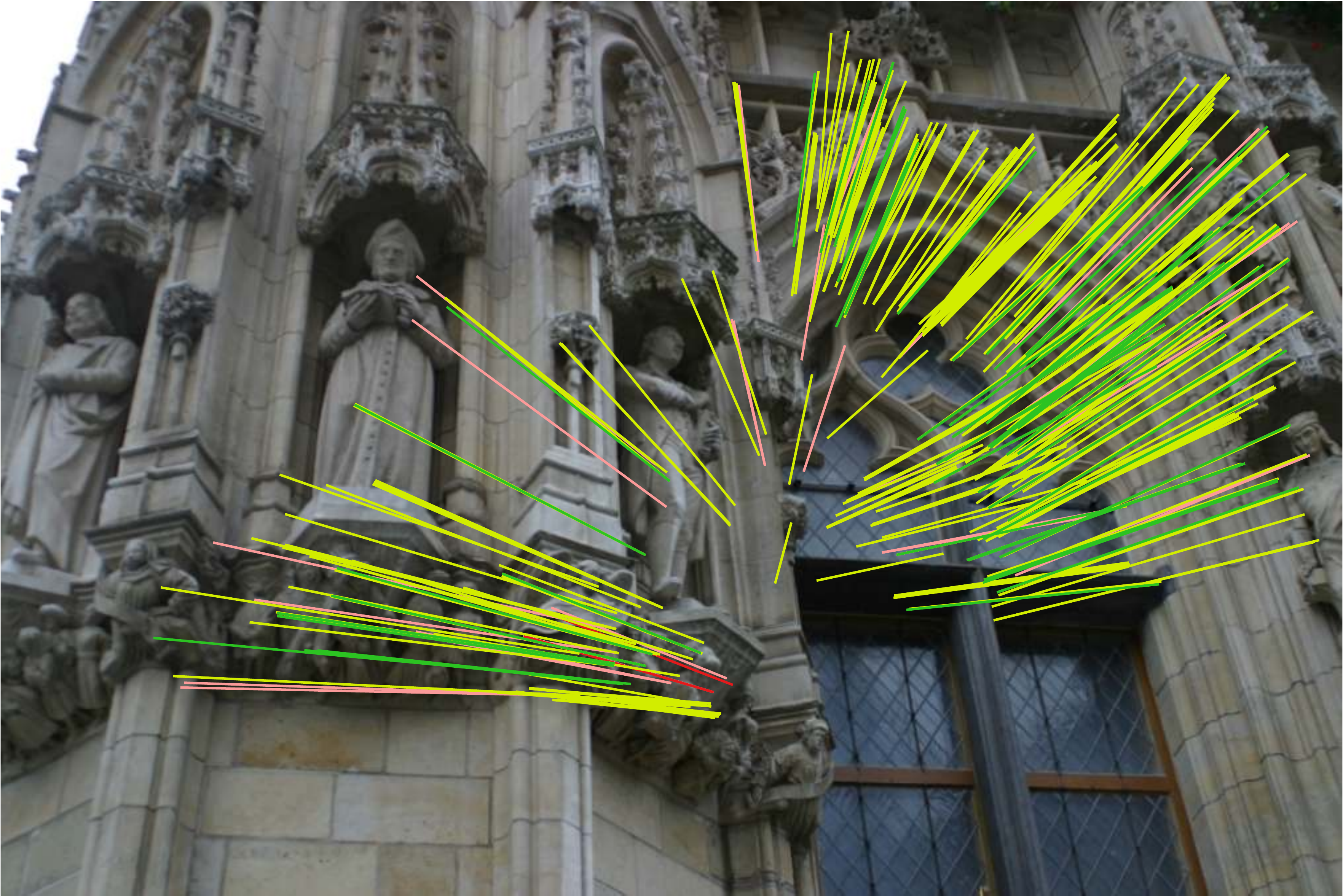}
	\includegraphics[height=7.5em]{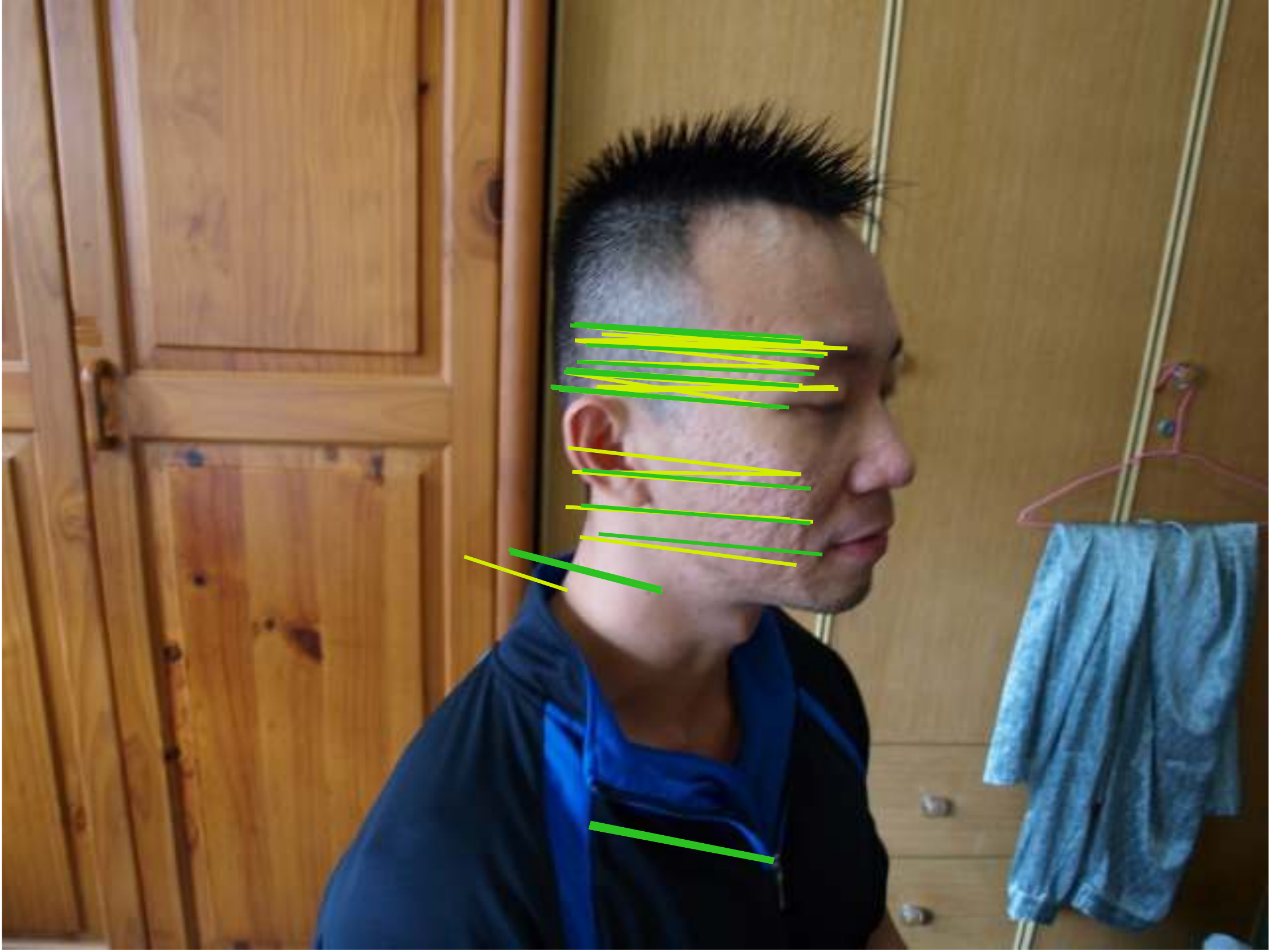}
	\\
	\begin{flushleft}
		\hspace*{7.5em}DD15\hspace{4.75em}Spidey13\hspace{4.75em}DC01\hspace{8em}LeuvenB\hspace{7.75em}BF
	\end{flushleft}
	\caption{\label{example_1b}
		Planar and non-planar local spatial filter matches according to the best configuration setup, the images of the input pair alternate among the rows. Image indexes are reported as suffix when the sequence contains more than two images. For each method inlier (yellow, green) and outlier (red and light red) clusters are shown, as well as the 1SAC filtered matches (green, red) (see Sec.~\ref{eval_dt}, best viewed in color and zoomed in).}
\end{figure*}

\begin{figure*}
	\center
	\rotatebox[origin=l]{90}{\mbox{\hspace{3em}th}}
	\includegraphics[height=7.5em]{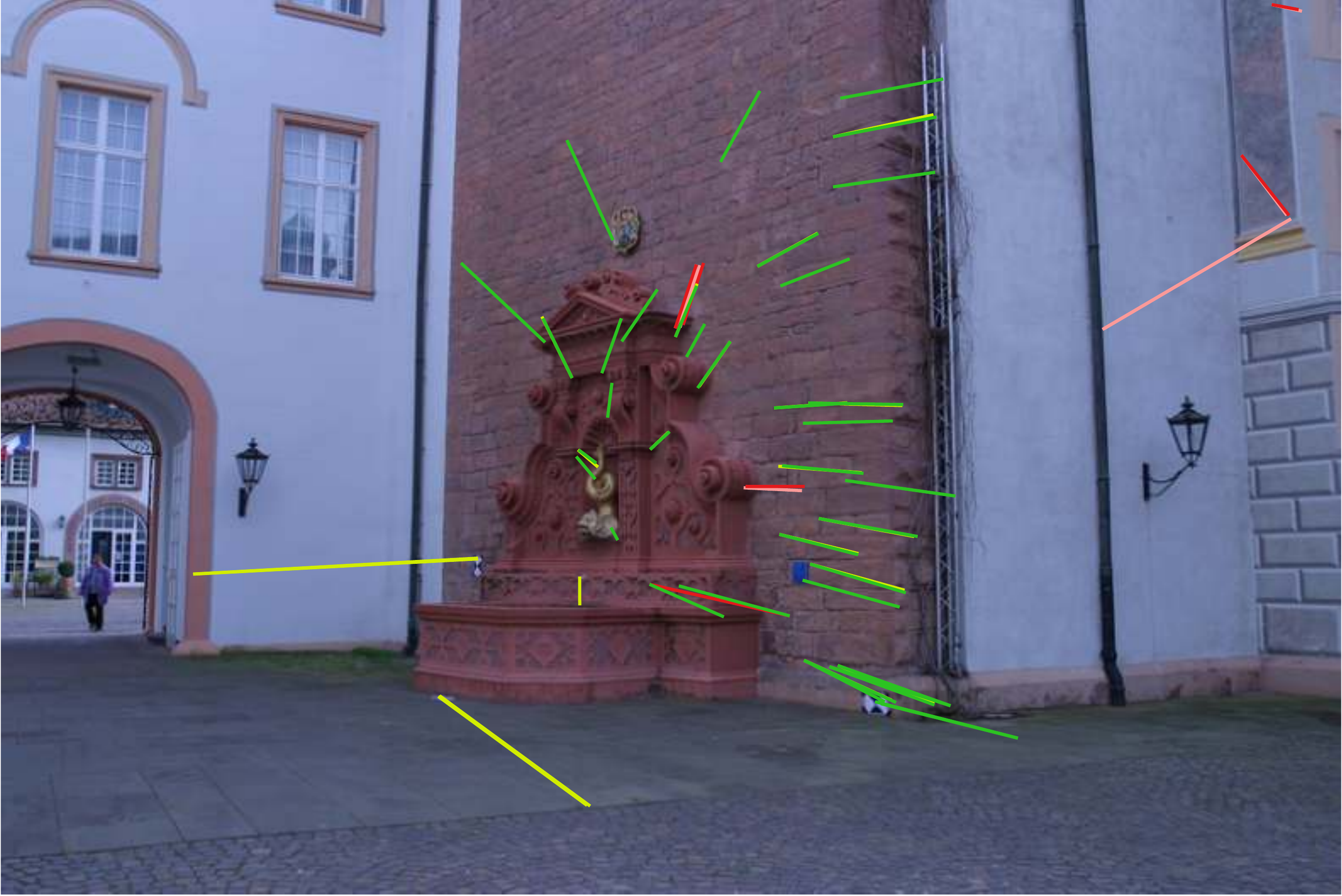}
	\includegraphics[height=7.5em]{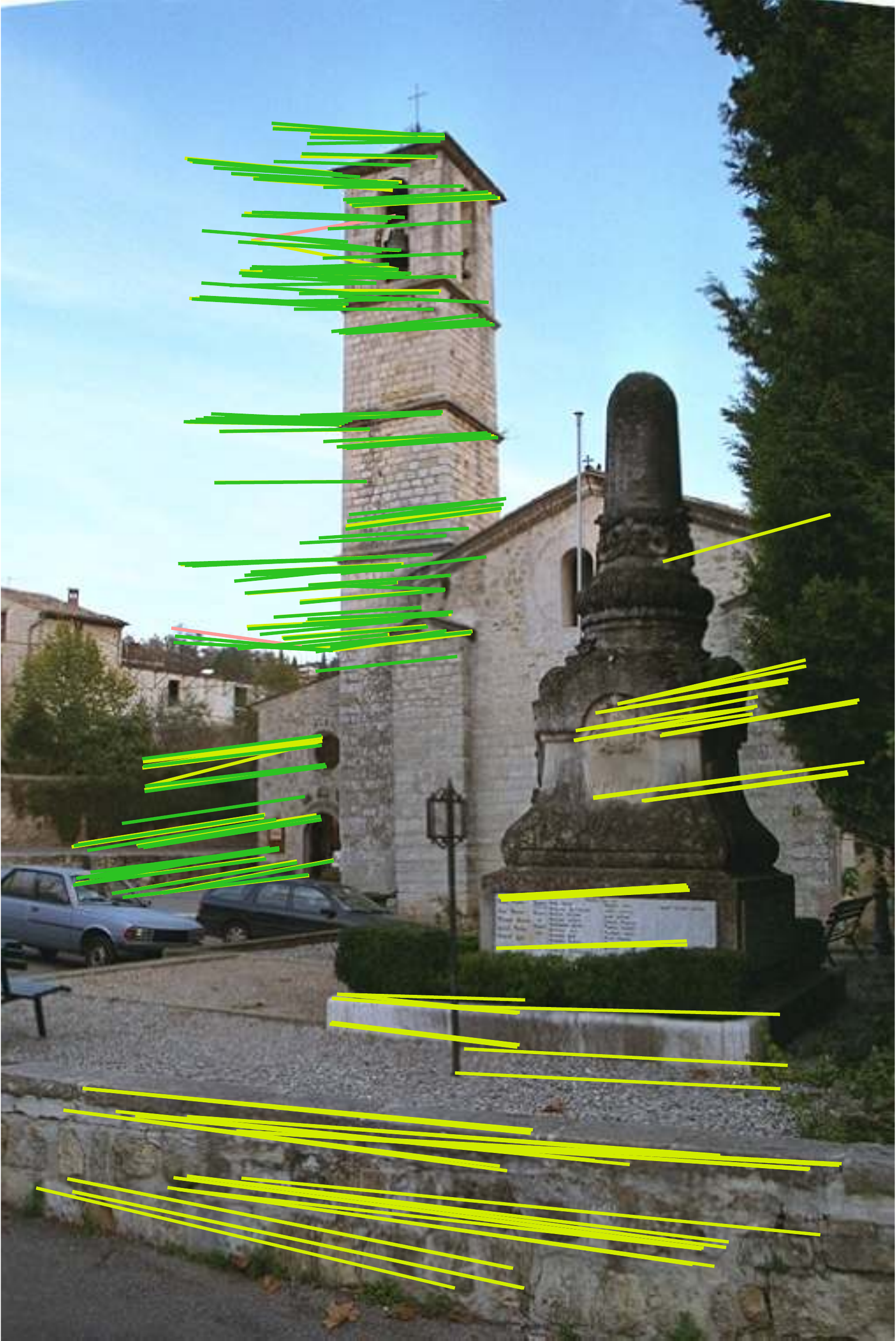}
	\includegraphics[height=7.5em]{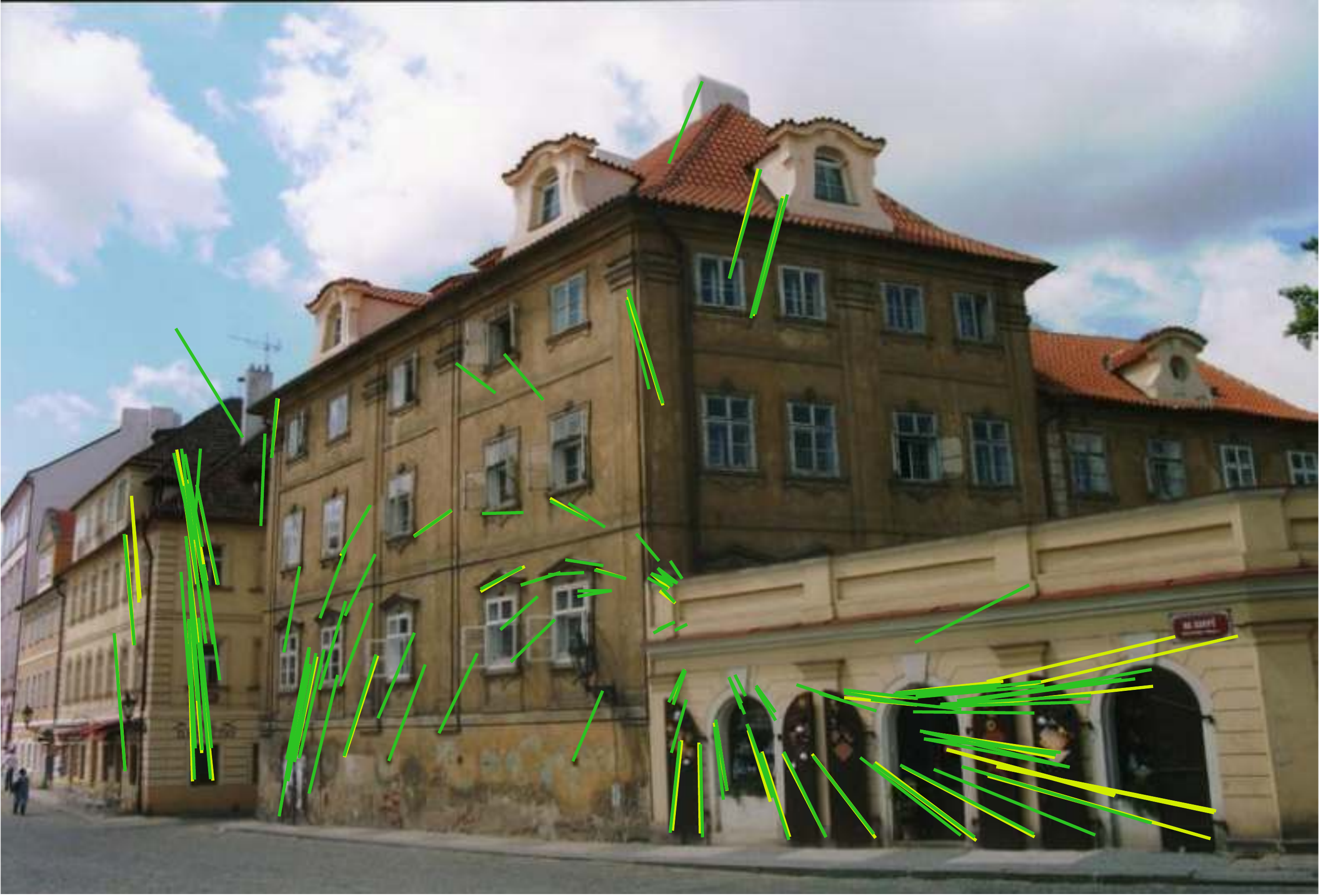}
	\hspace{0.05em}
	\includegraphics[height=7.5em]{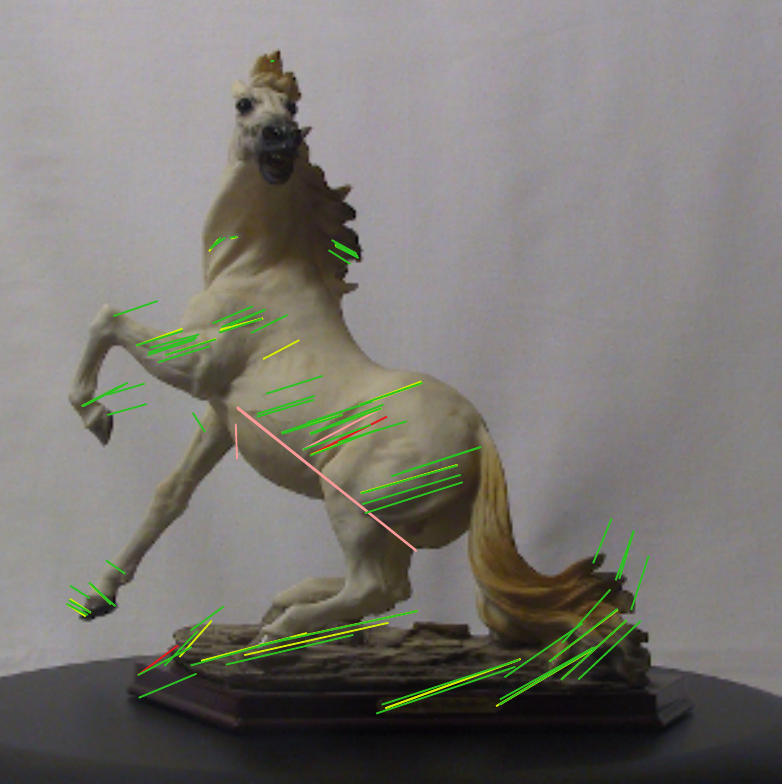}
	\hspace{0.05em}
	\includegraphics[height=7.5em]{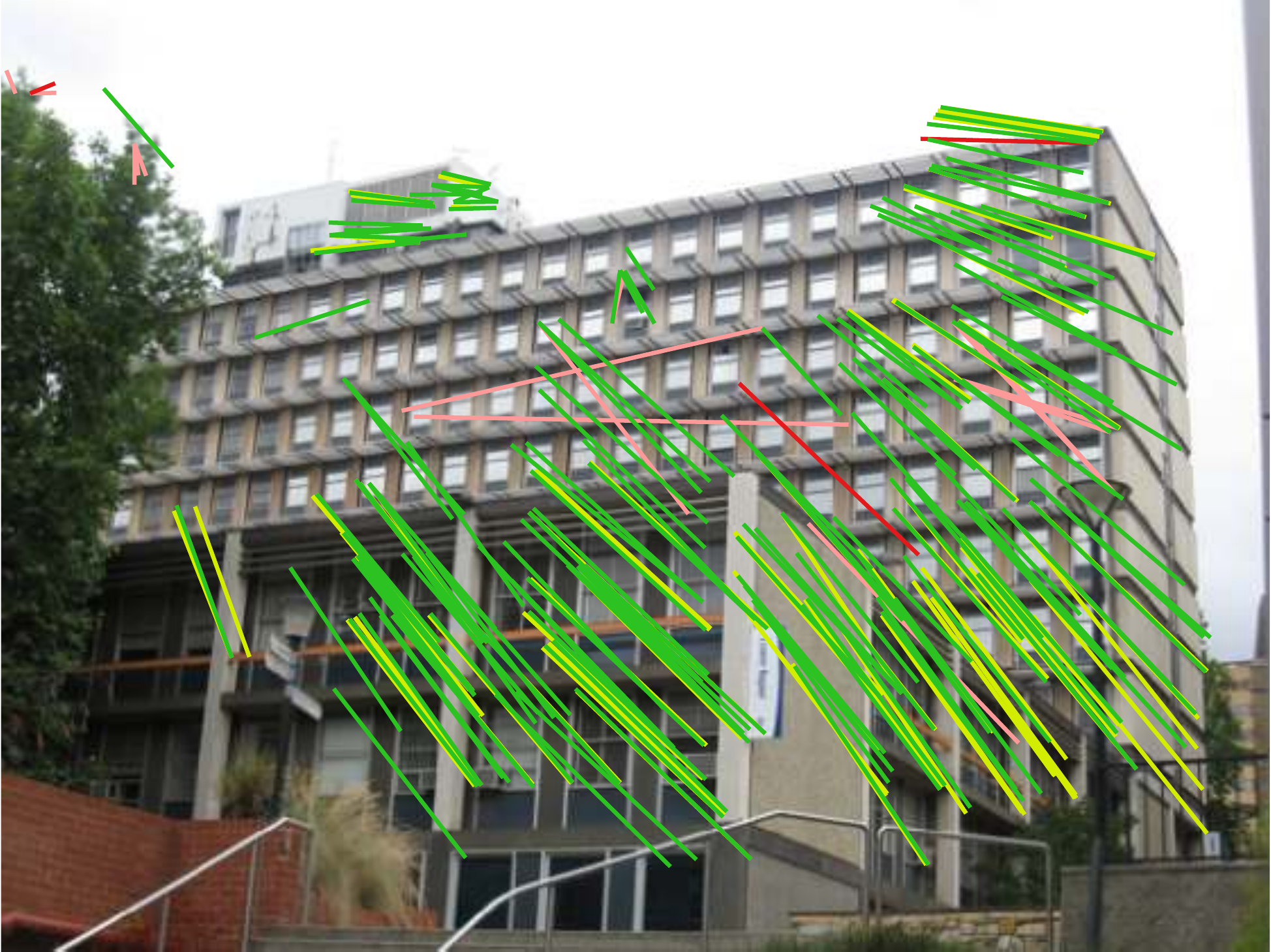}
	\\
	\vspace{0.5em}
	\rotatebox[origin=l]{90}{\mbox{\hspace{2em}DTM}}
	\includegraphics[height=7.5em]{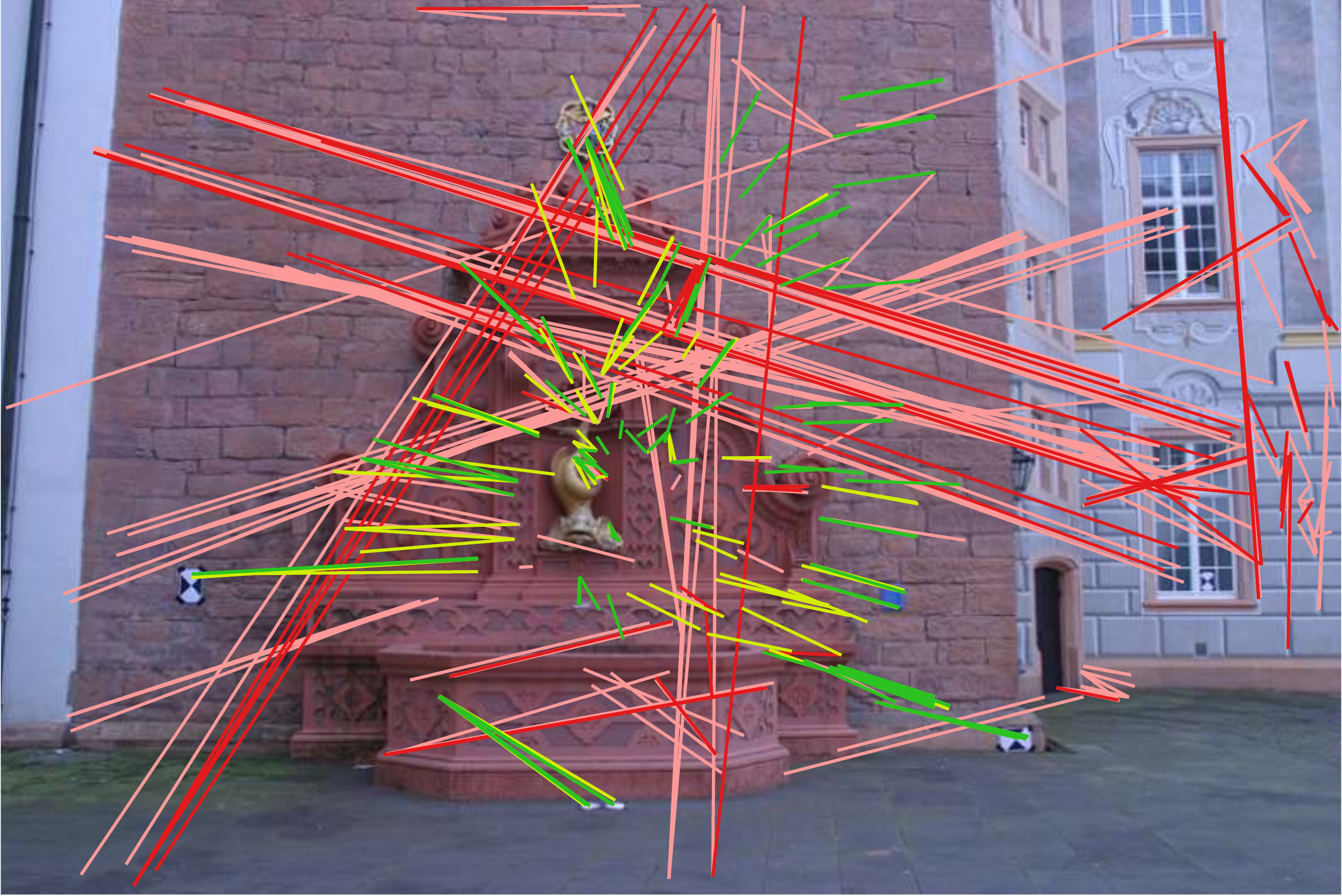}
	\includegraphics[height=7.5em]{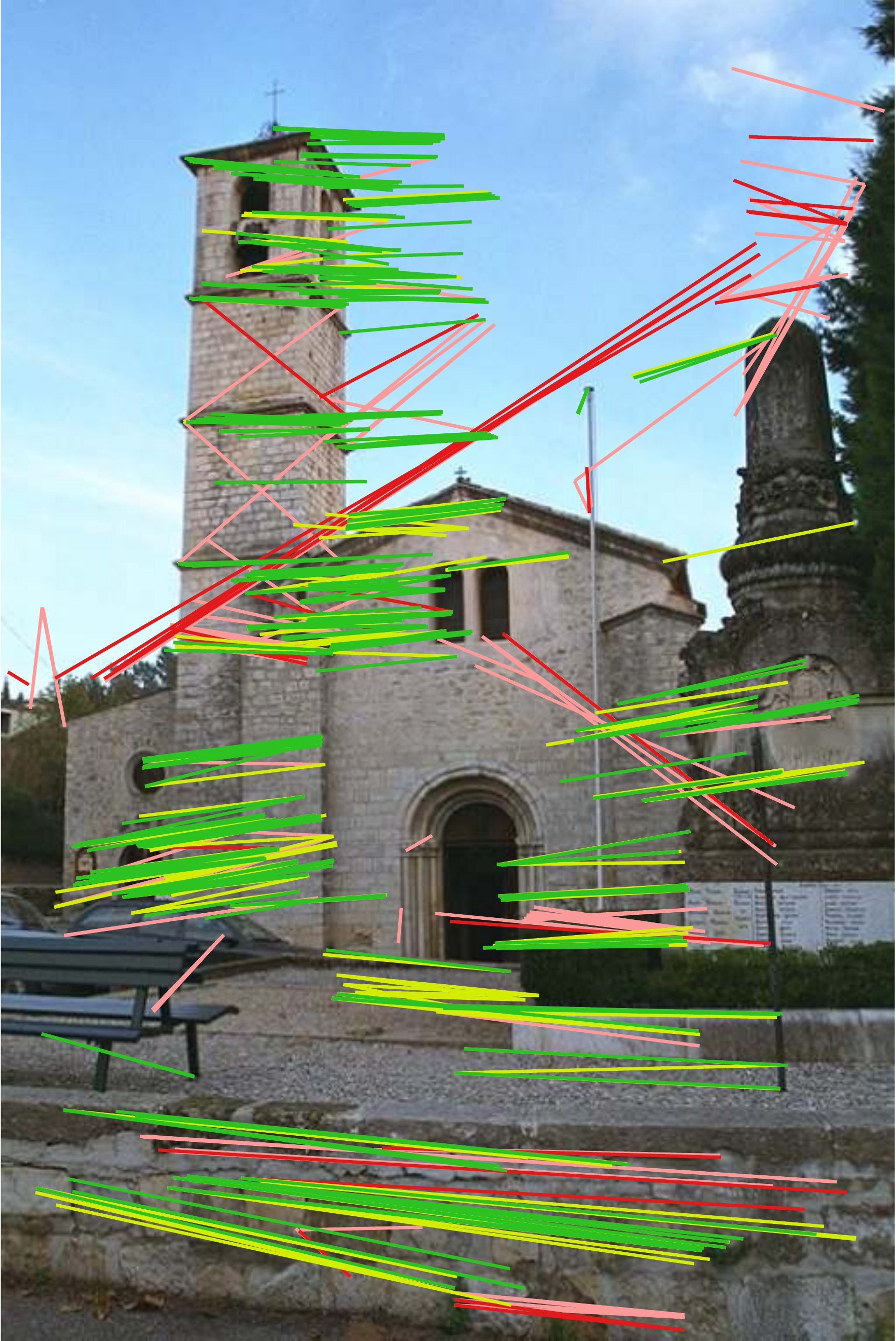}
	\includegraphics[height=7.5em]{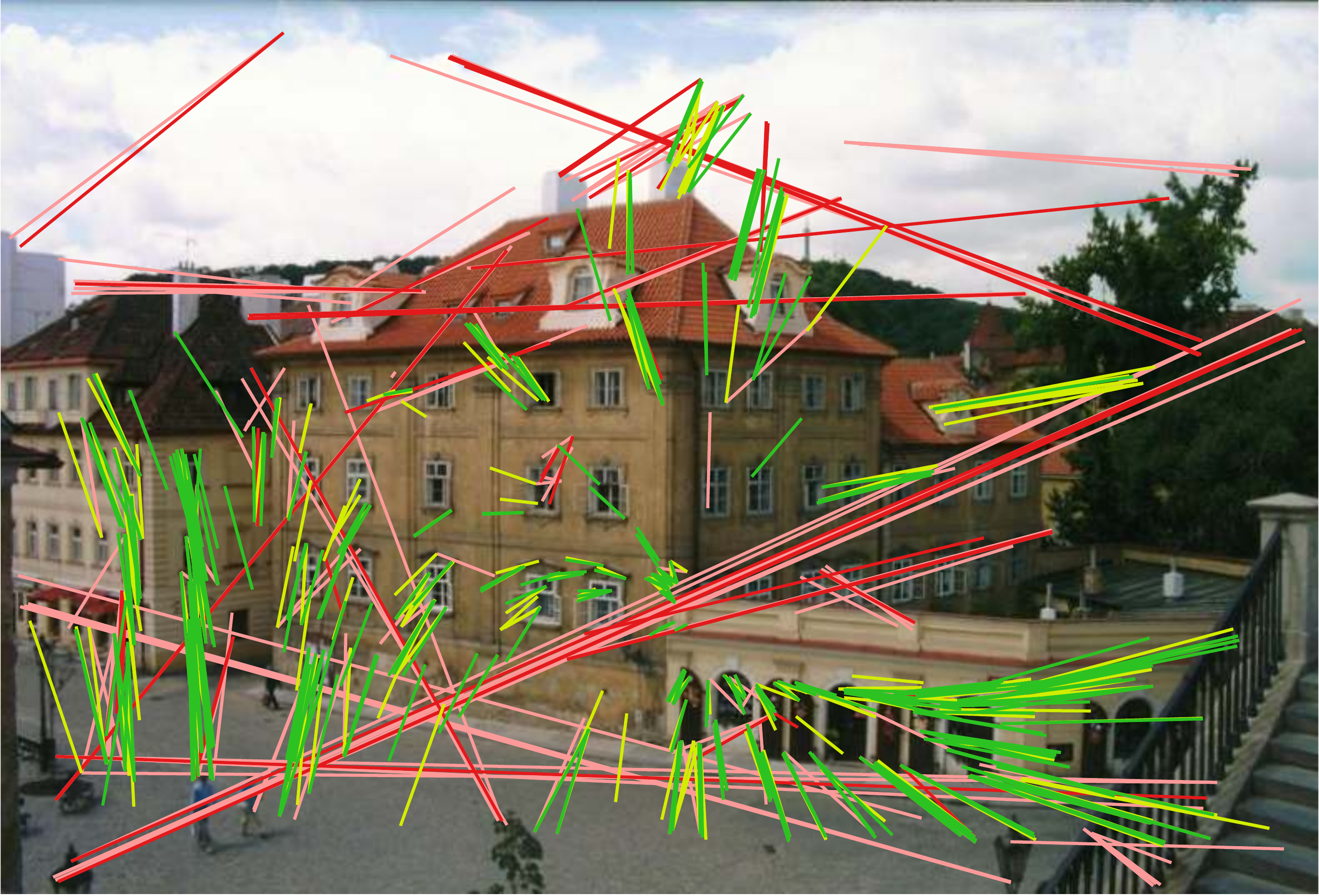}
	\includegraphics[height=7.5em]{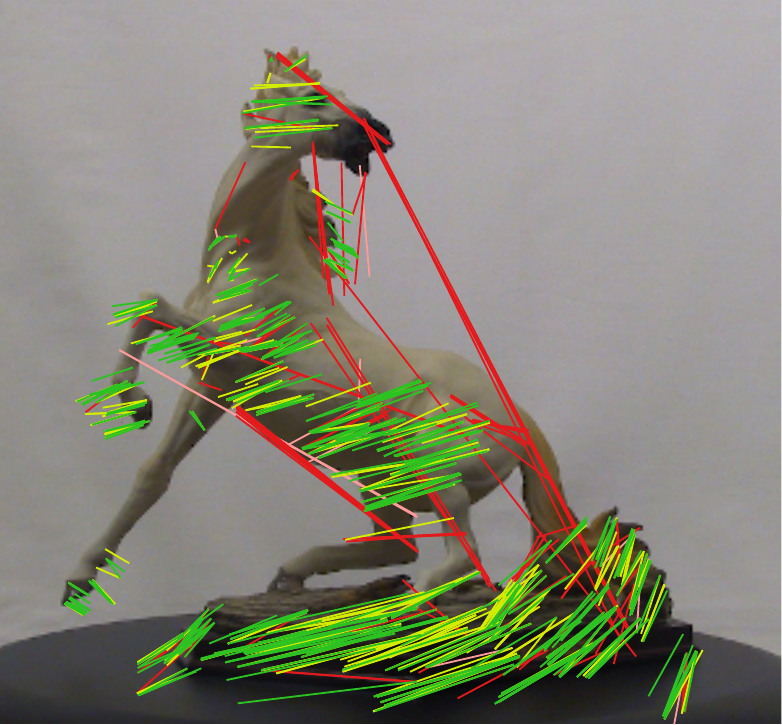}
	\includegraphics[height=7.5em]{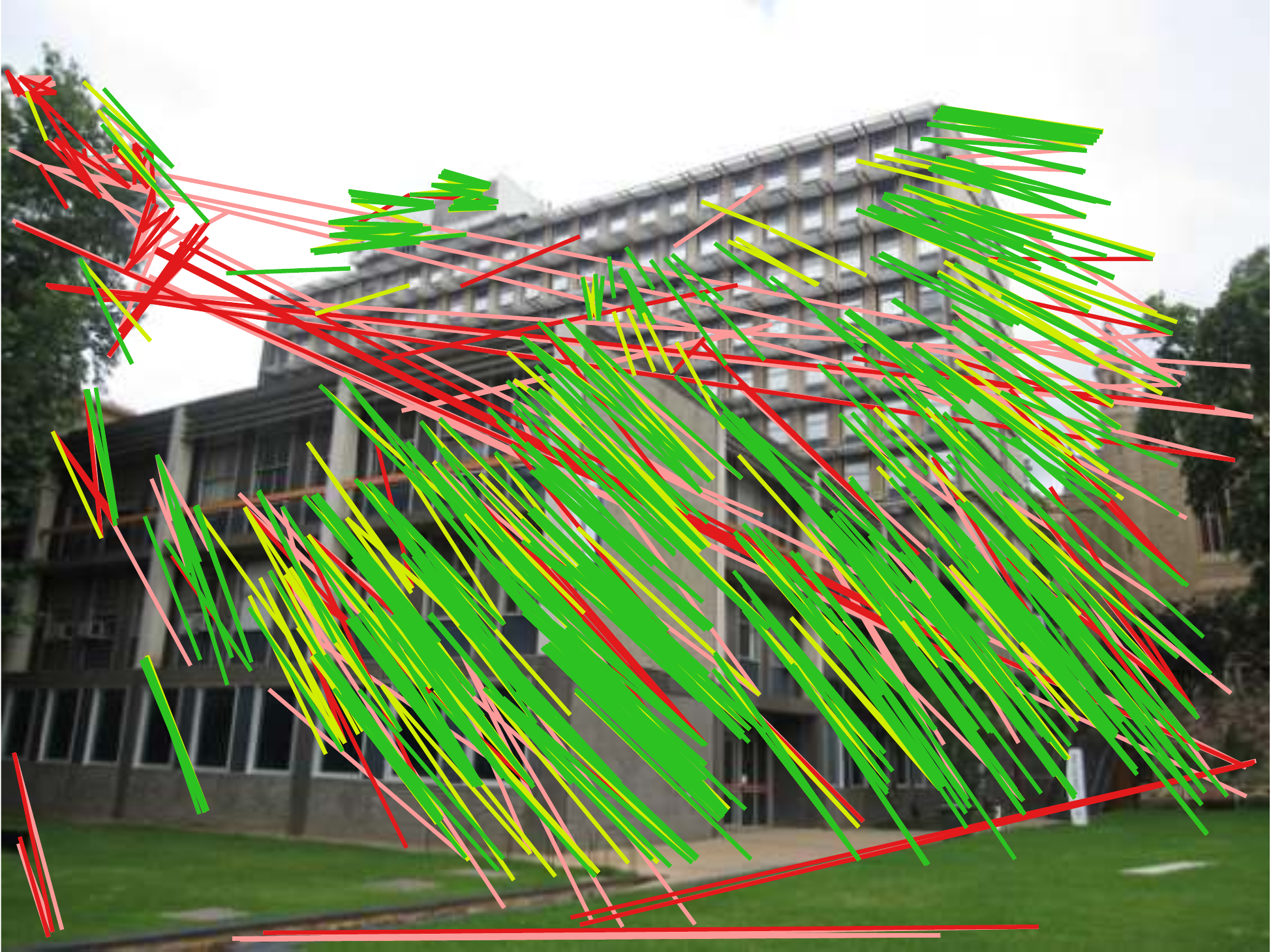}
	\\
	\vspace{0.5em}
	\rotatebox[origin=l]{90}{\mbox{\hspace{2em}LMR}}
	\includegraphics[height=7.5em]{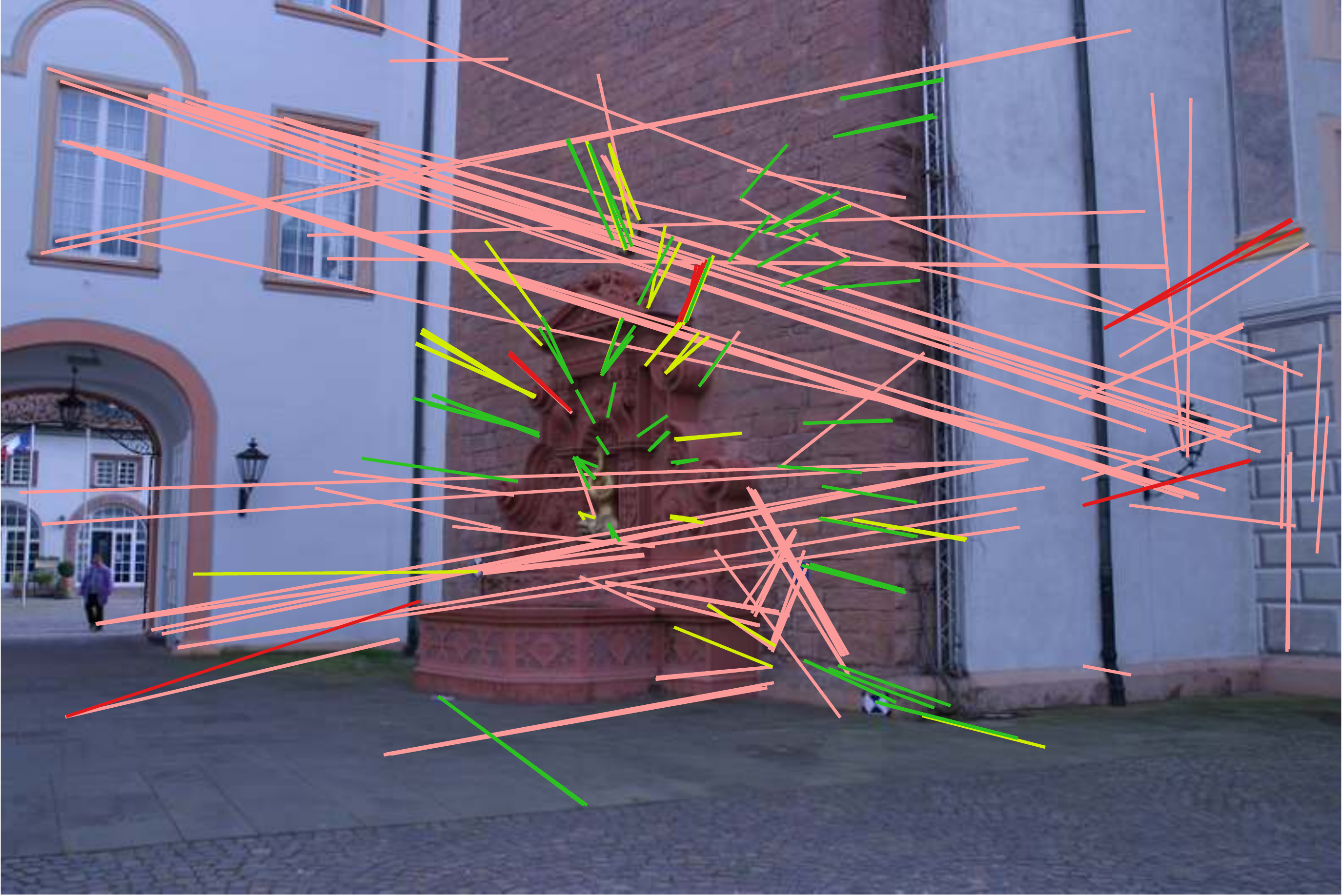}
	\includegraphics[height=7.5em]{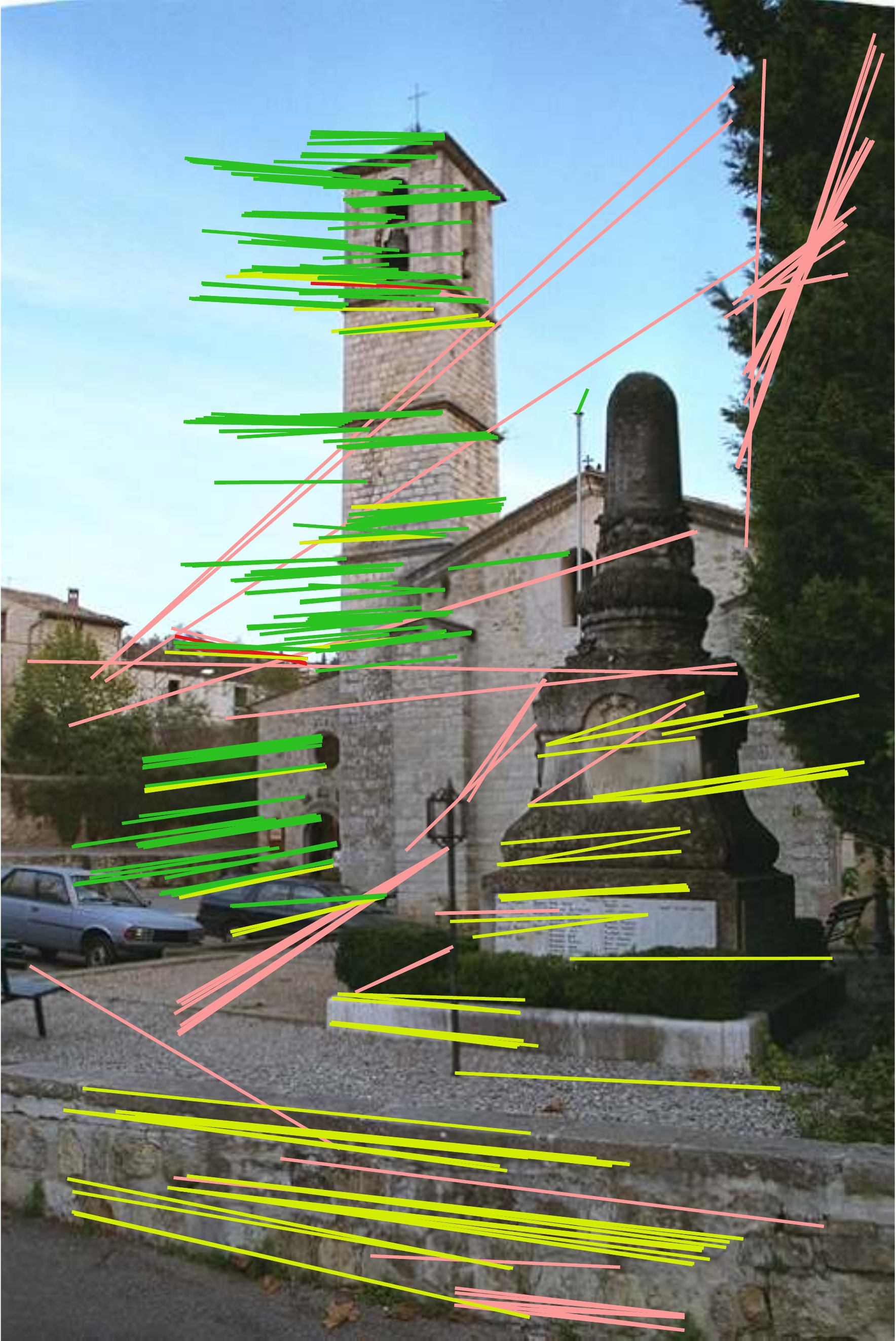}
	\includegraphics[height=7.5em]{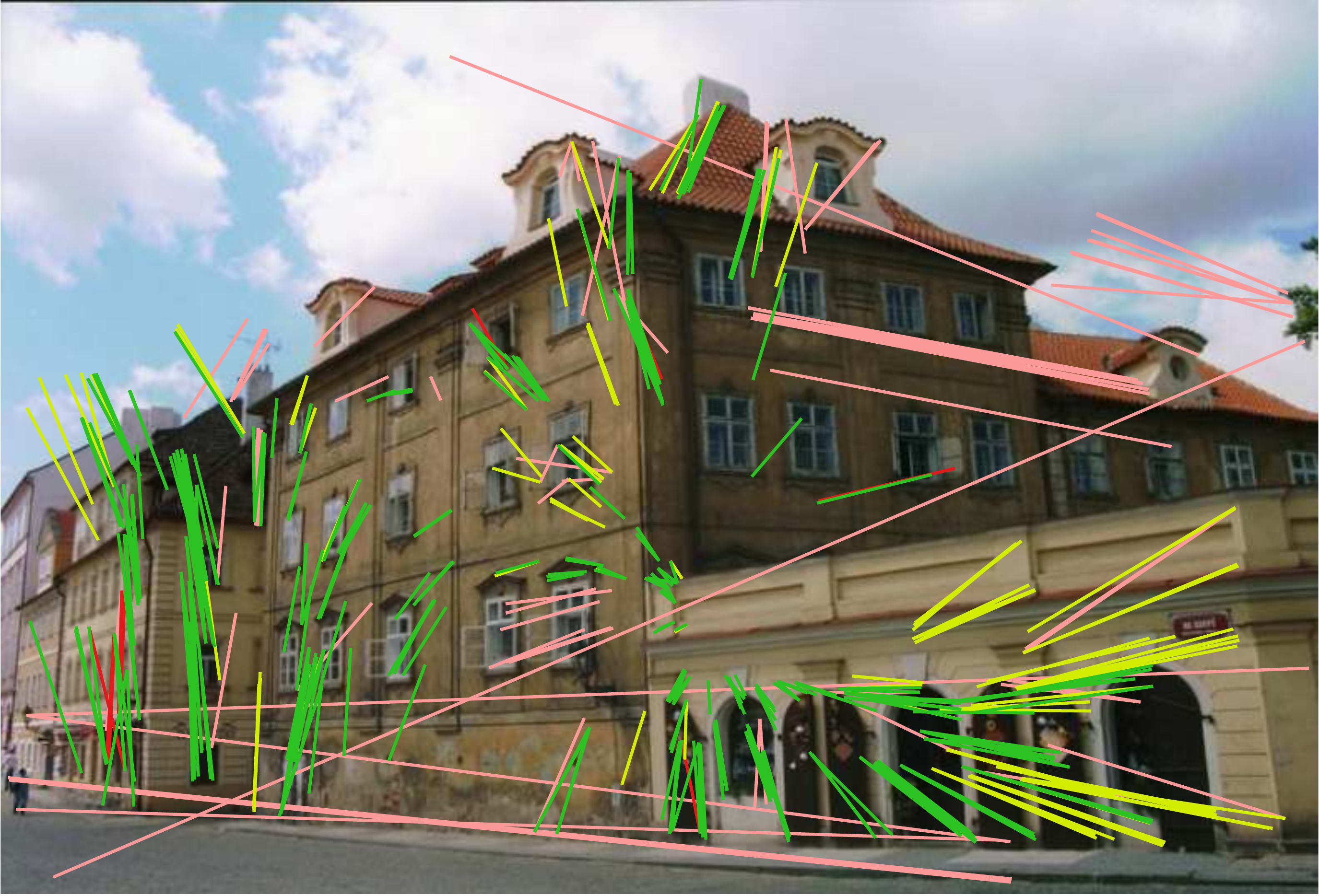}
	\hspace{0.05em}
	\includegraphics[height=7.5em]{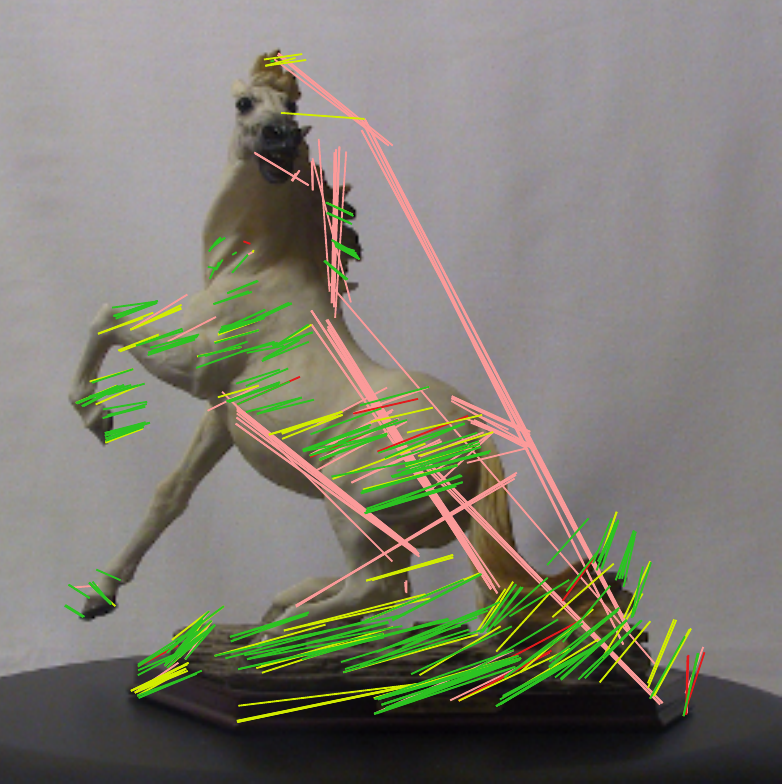}
	\hspace{0.05em}
	\includegraphics[height=7.5em]{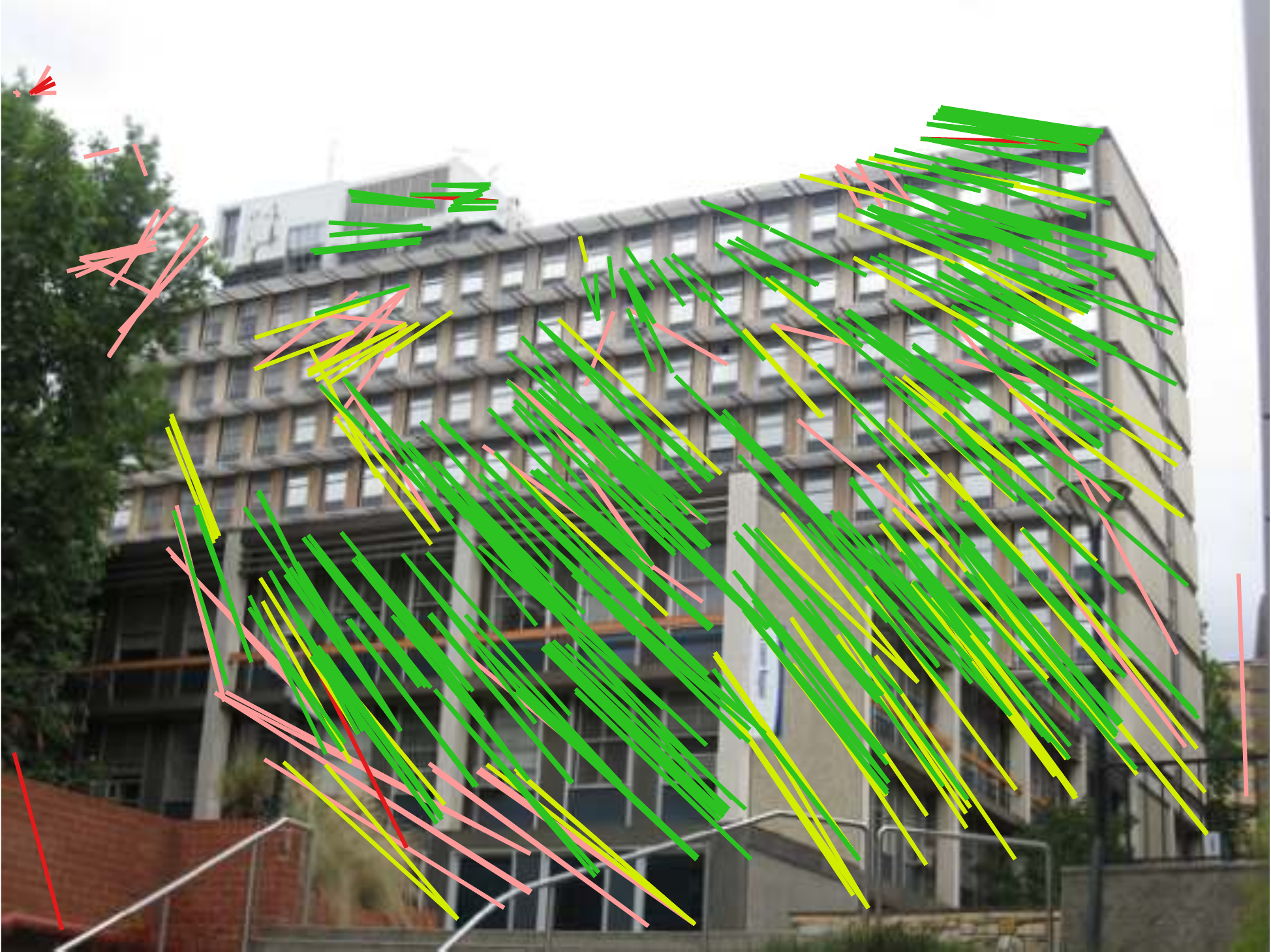}
	\\
	\vspace{0.5em}
	\rotatebox[origin=l]{90}{\mbox{\hspace{2em}LPM}}
	\includegraphics[height=7.5em]{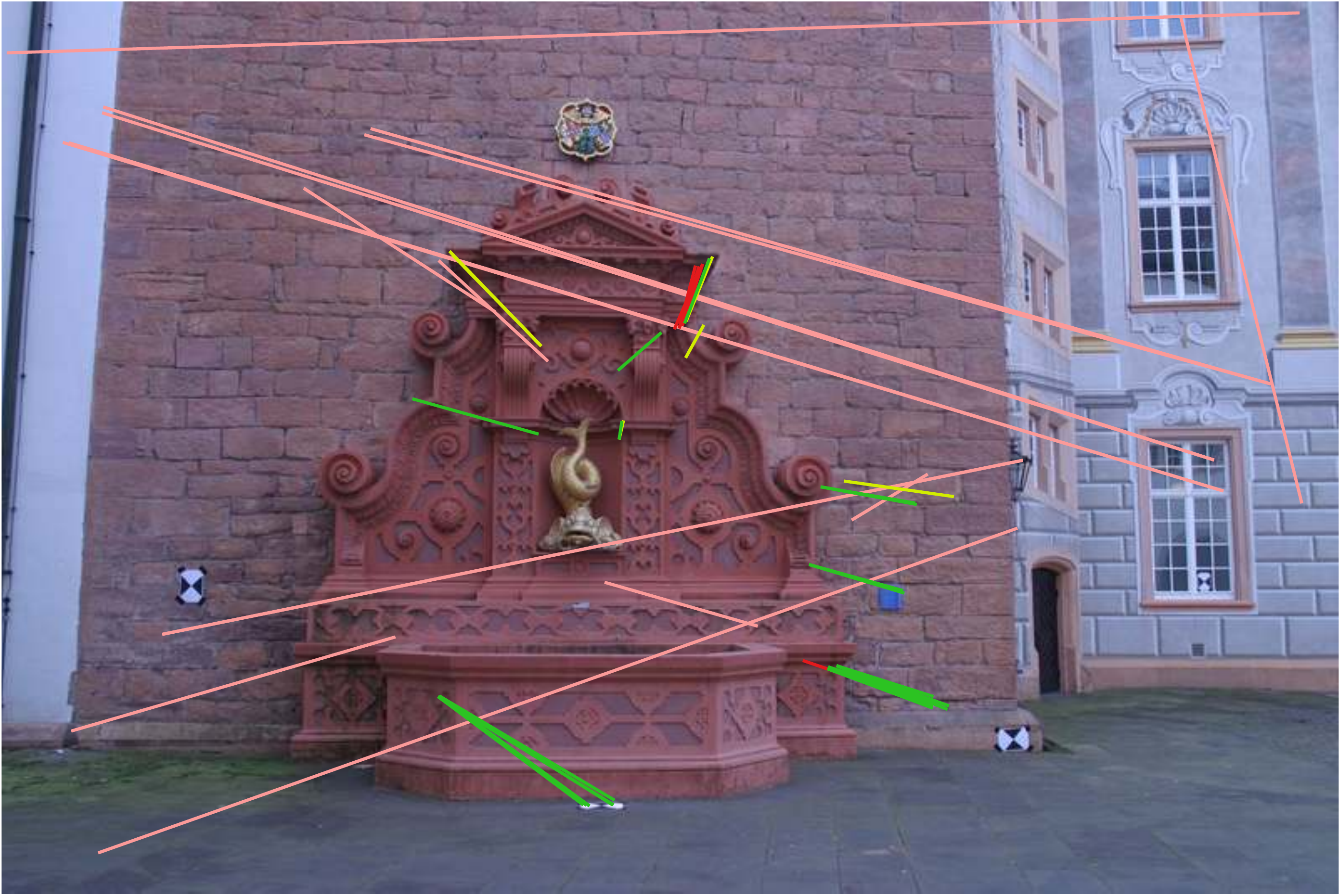}
	\includegraphics[height=7.5em]{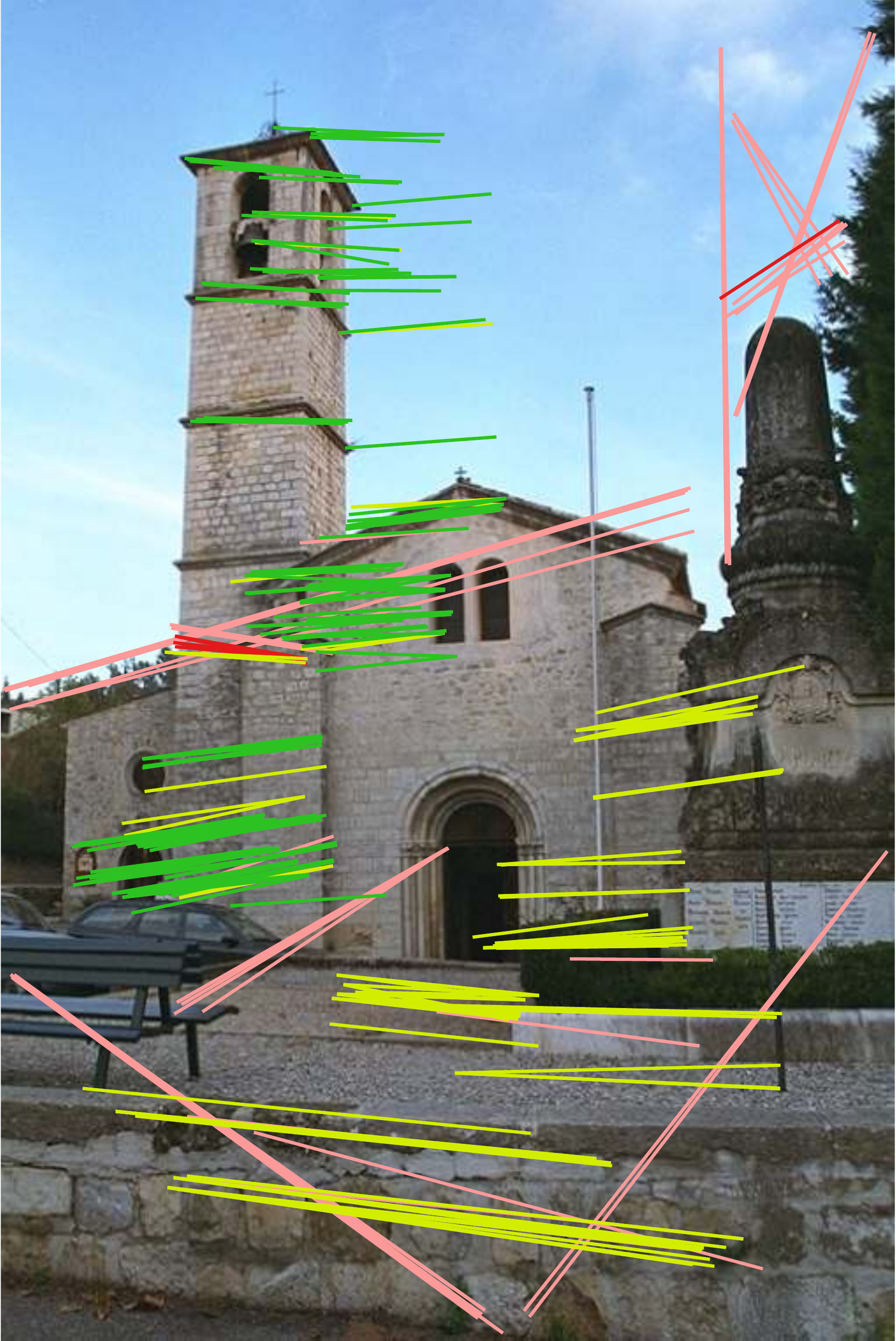}
	\includegraphics[height=7.5em]{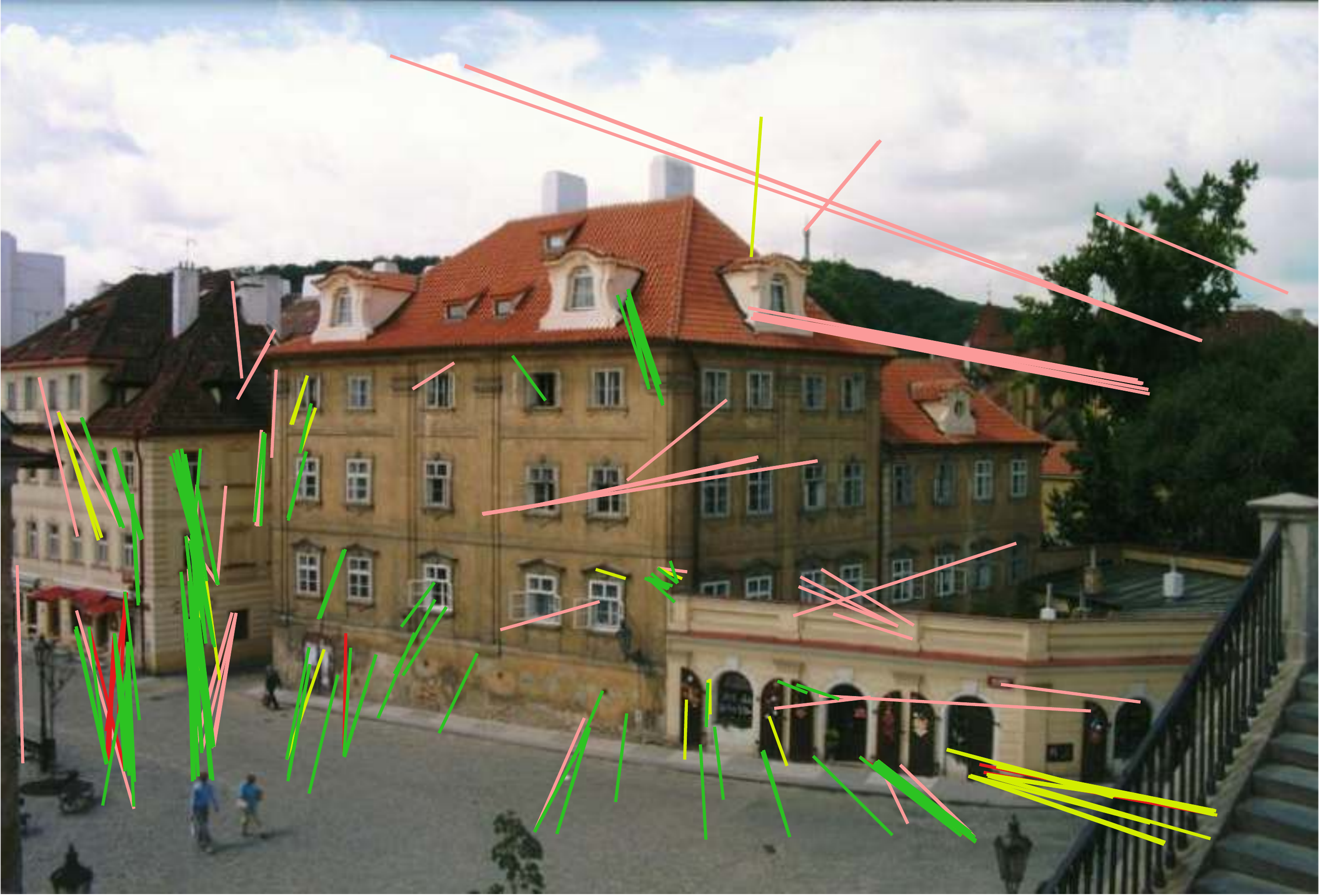}
	\includegraphics[height=7.5em]{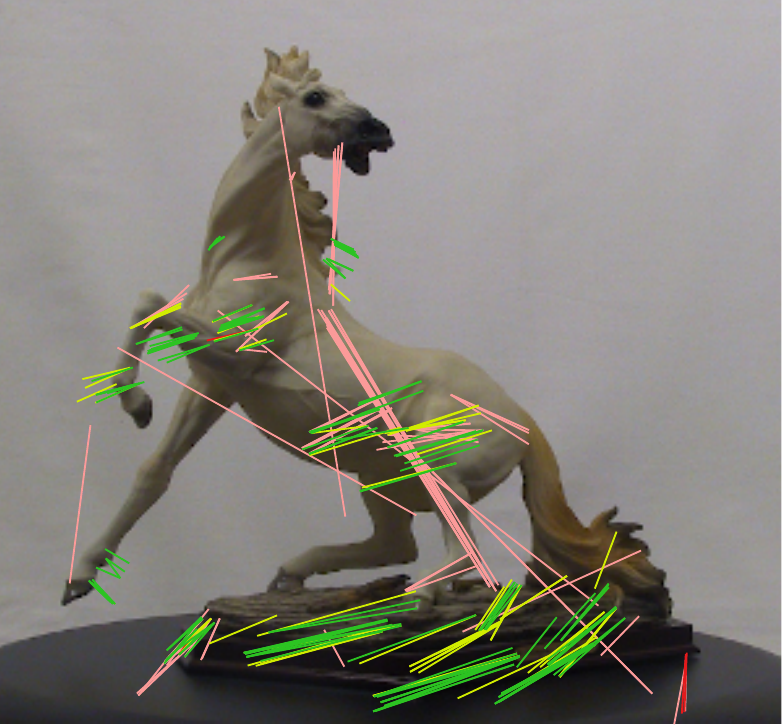}
	\includegraphics[height=7.5em]{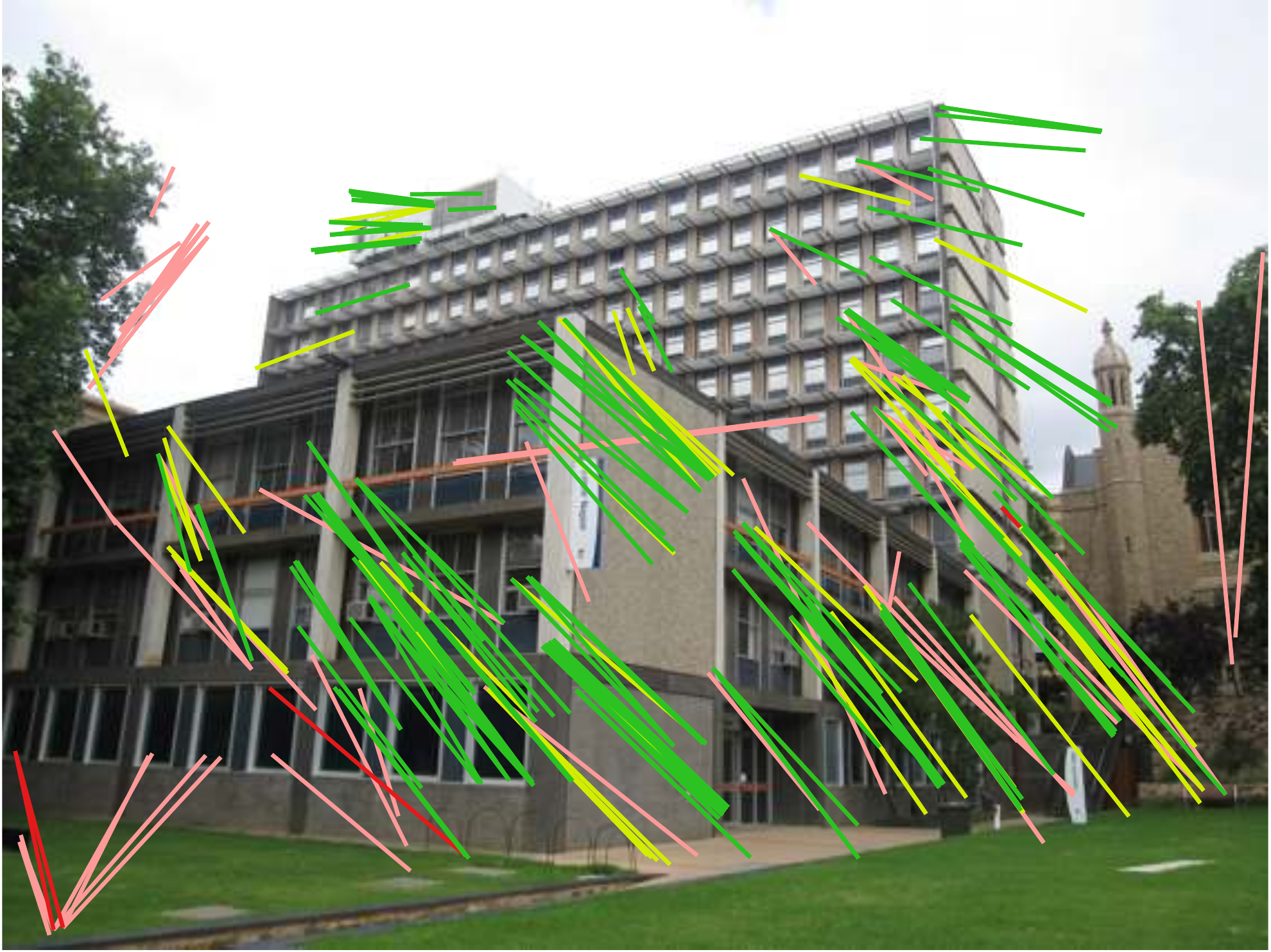}
	\\
	\vspace{0.5em}
	\rotatebox[origin=l]{90}{\mbox{\hspace{2em}GLPM}}
	\includegraphics[height=7.5em]{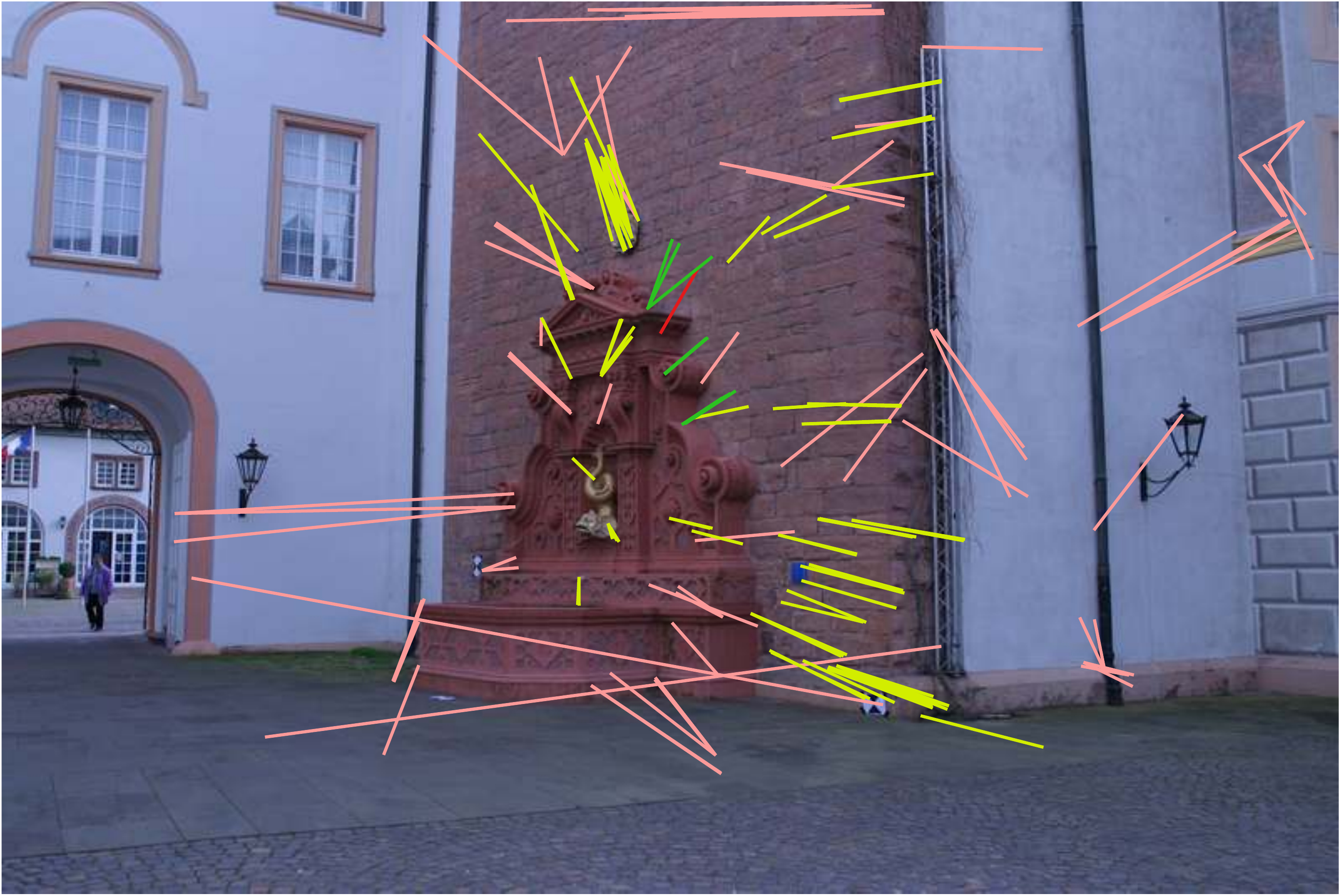}
	\includegraphics[height=7.5em]{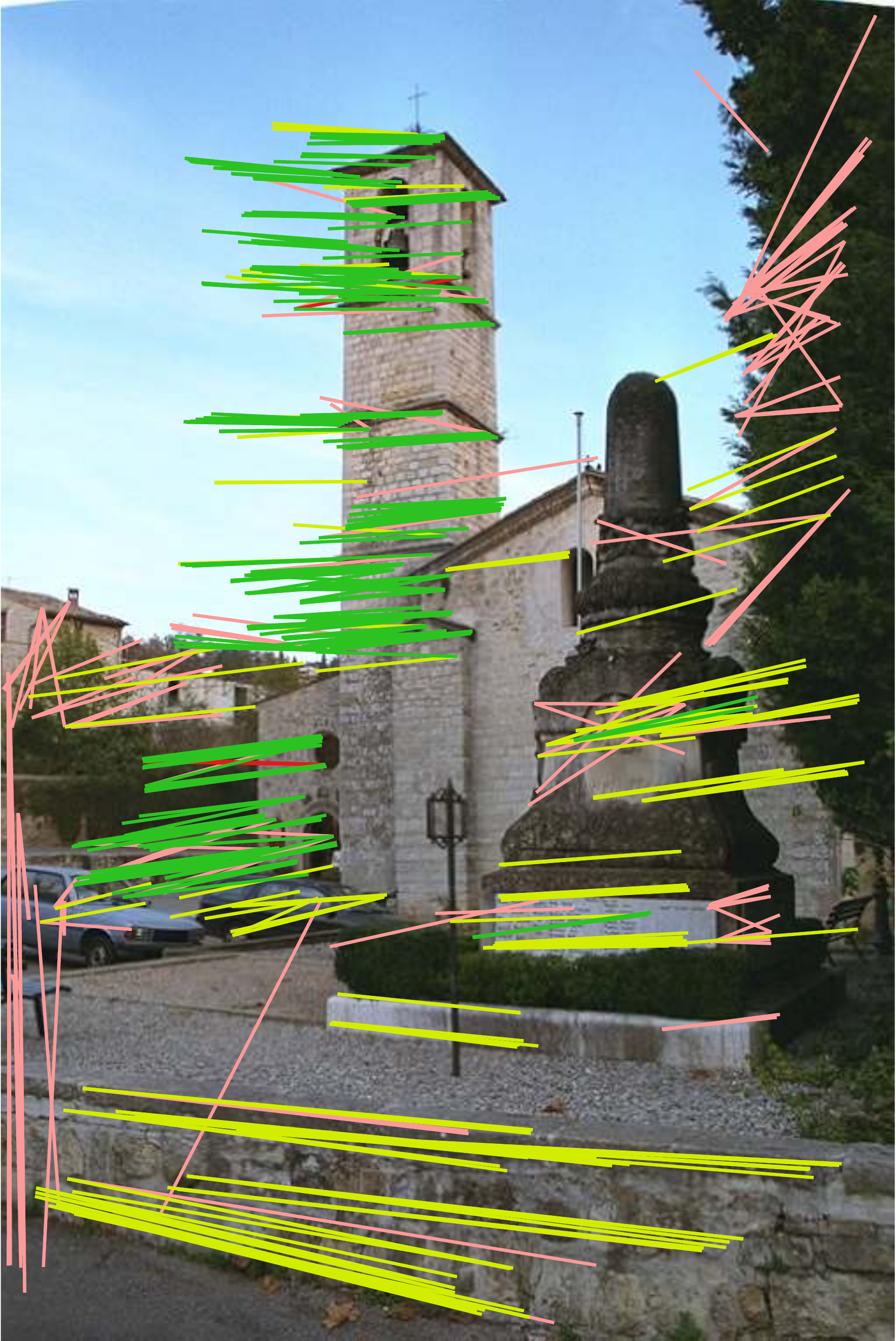}
	\includegraphics[height=7.5em]{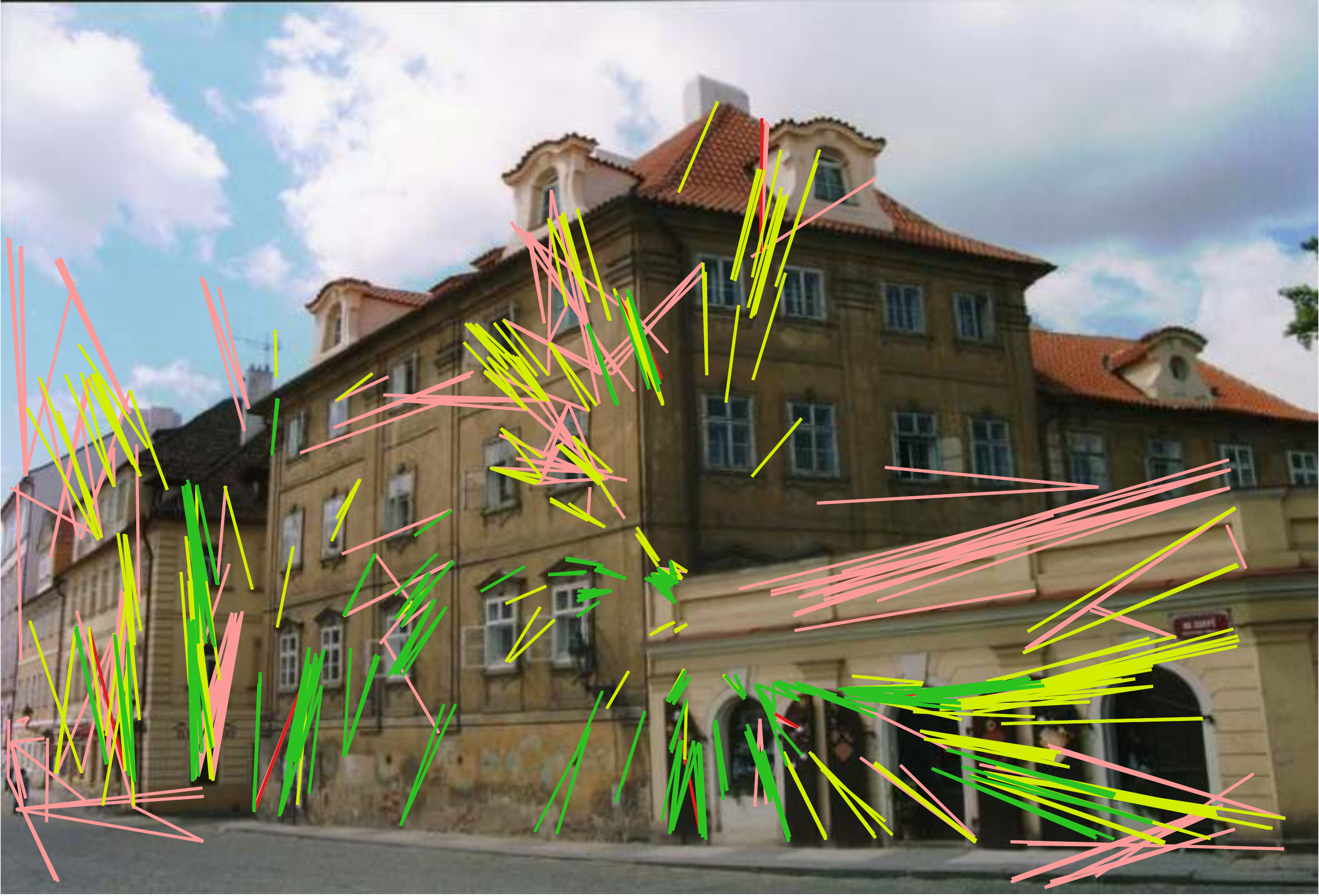}
	\hspace{0.05em}	
	\includegraphics[height=7.5em]{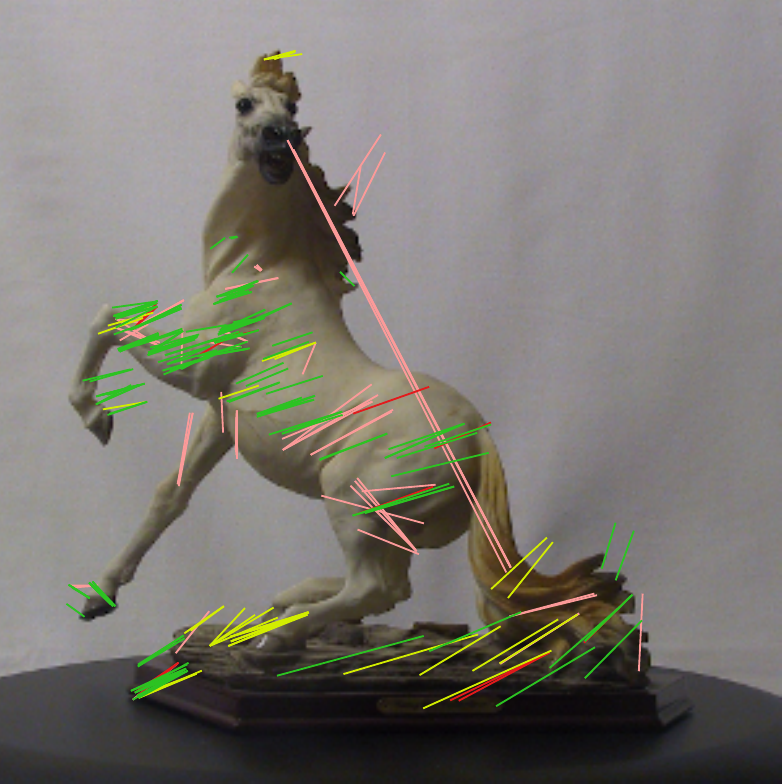}
	\hspace{0.05em}
	\includegraphics[height=7.5em]{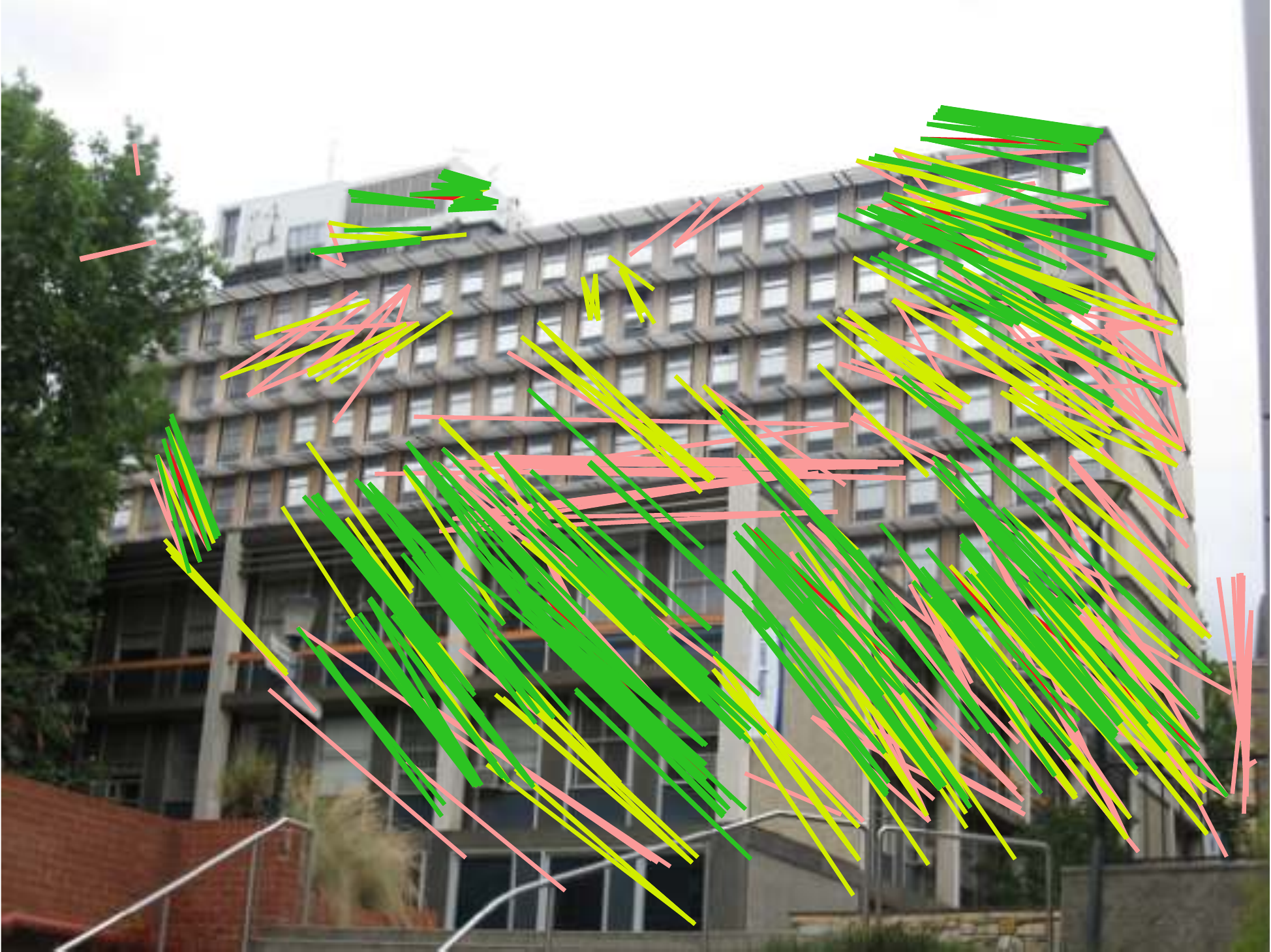}
	\\
	\vspace{0.5em}
	\rotatebox[origin=l]{90}{\mbox{\hspace{2em}GMS}}
	\includegraphics[height=7.5em]{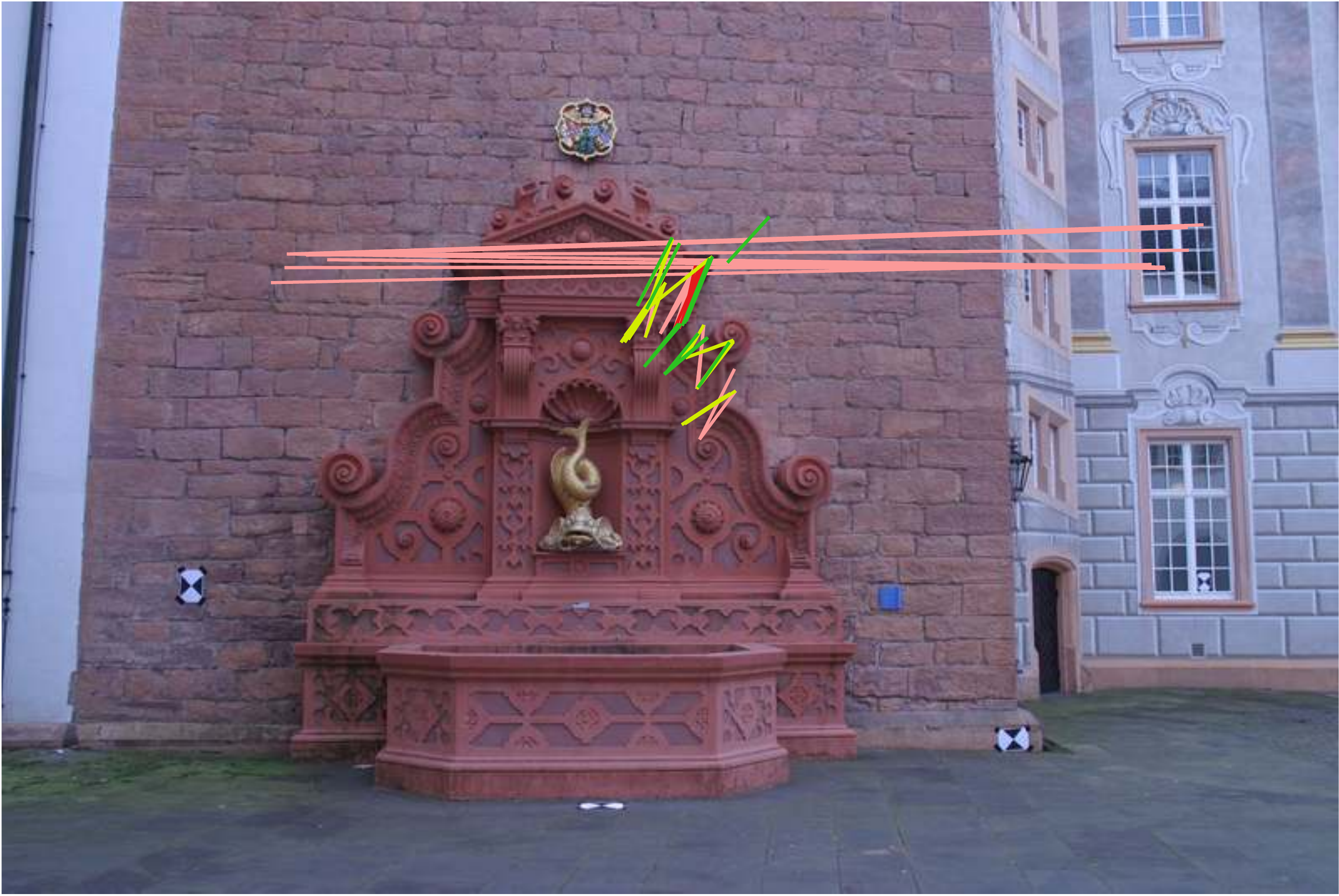}
	\includegraphics[height=7.5em]{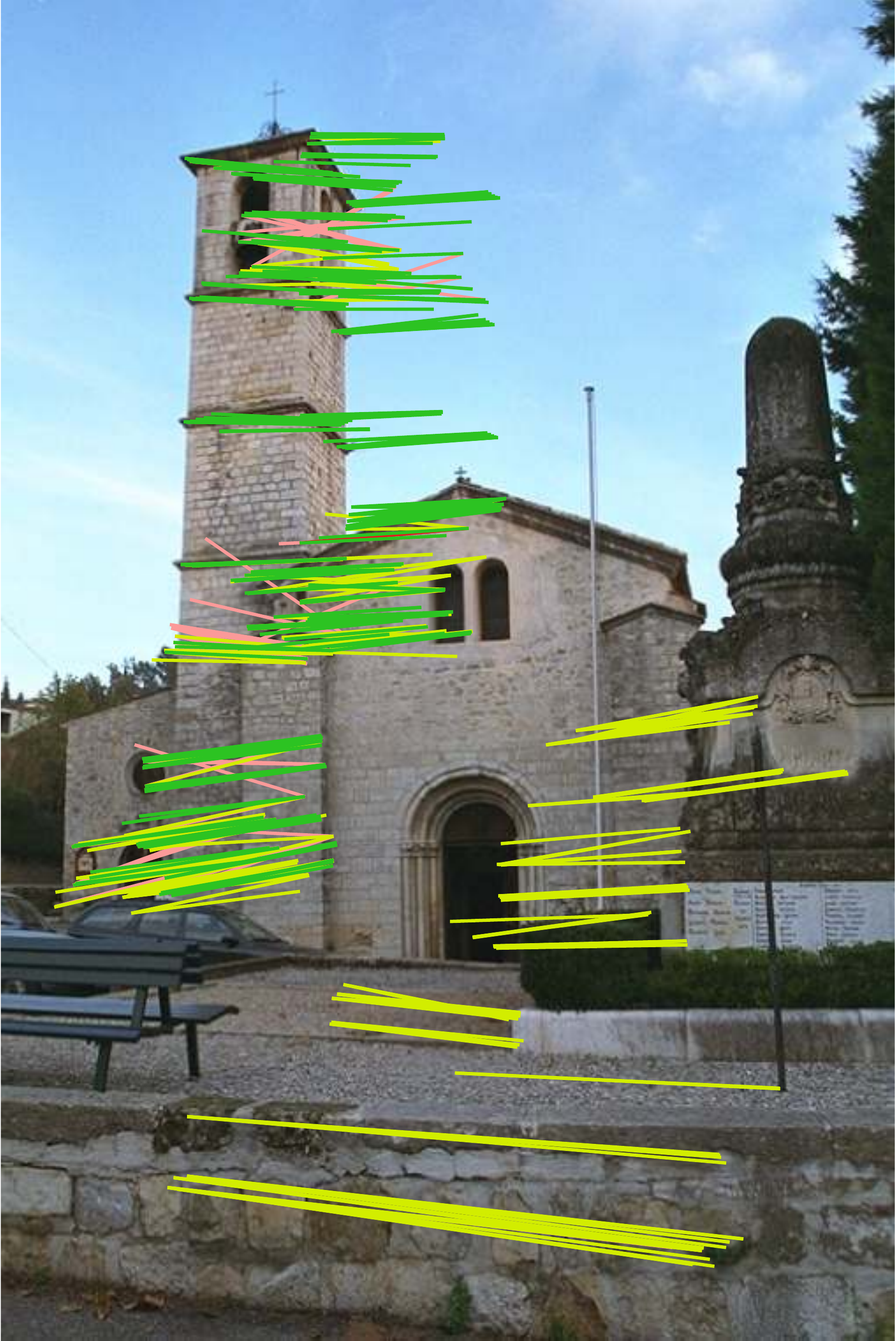}
	\includegraphics[height=7.5em]{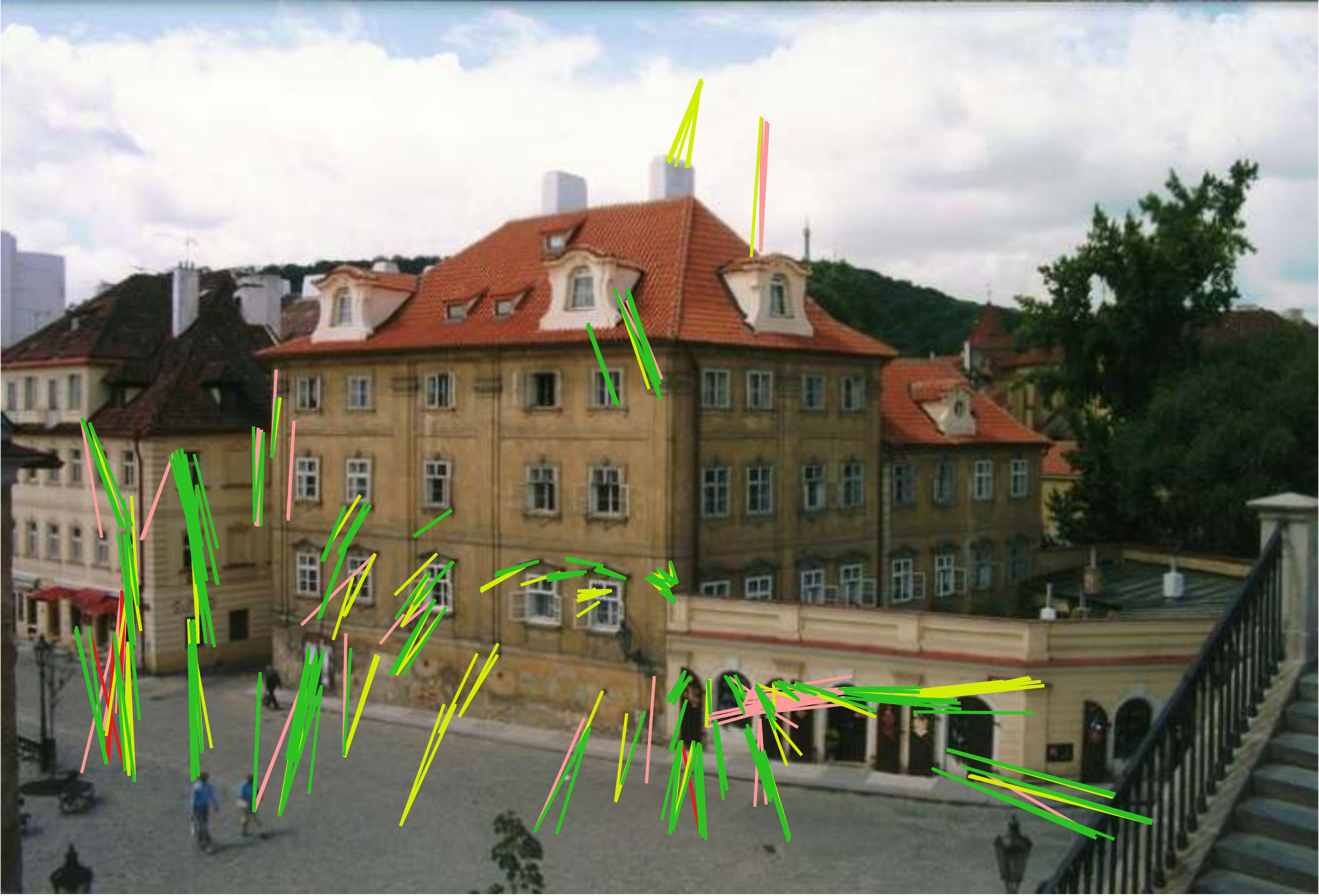}
	\includegraphics[height=7.5em]{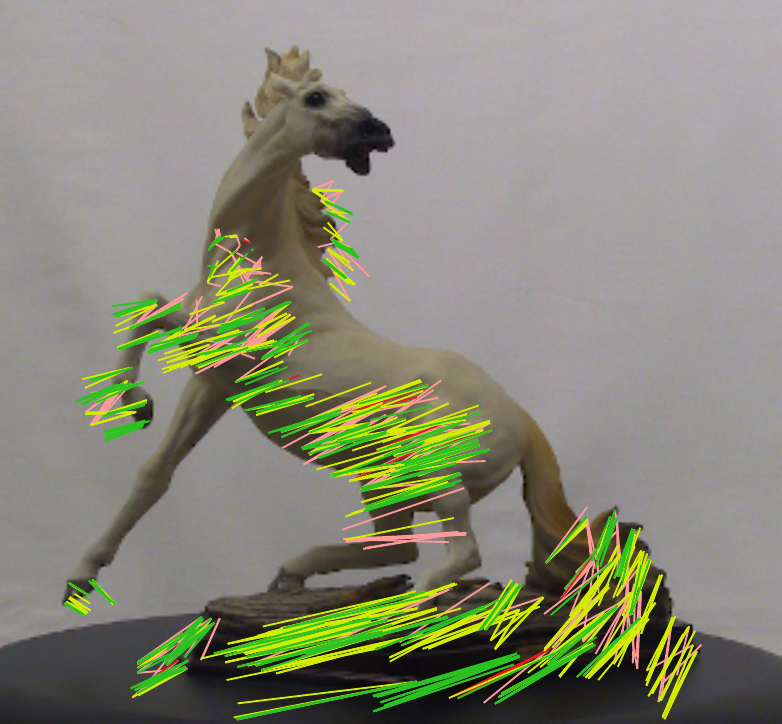}
	\includegraphics[height=7.5em]{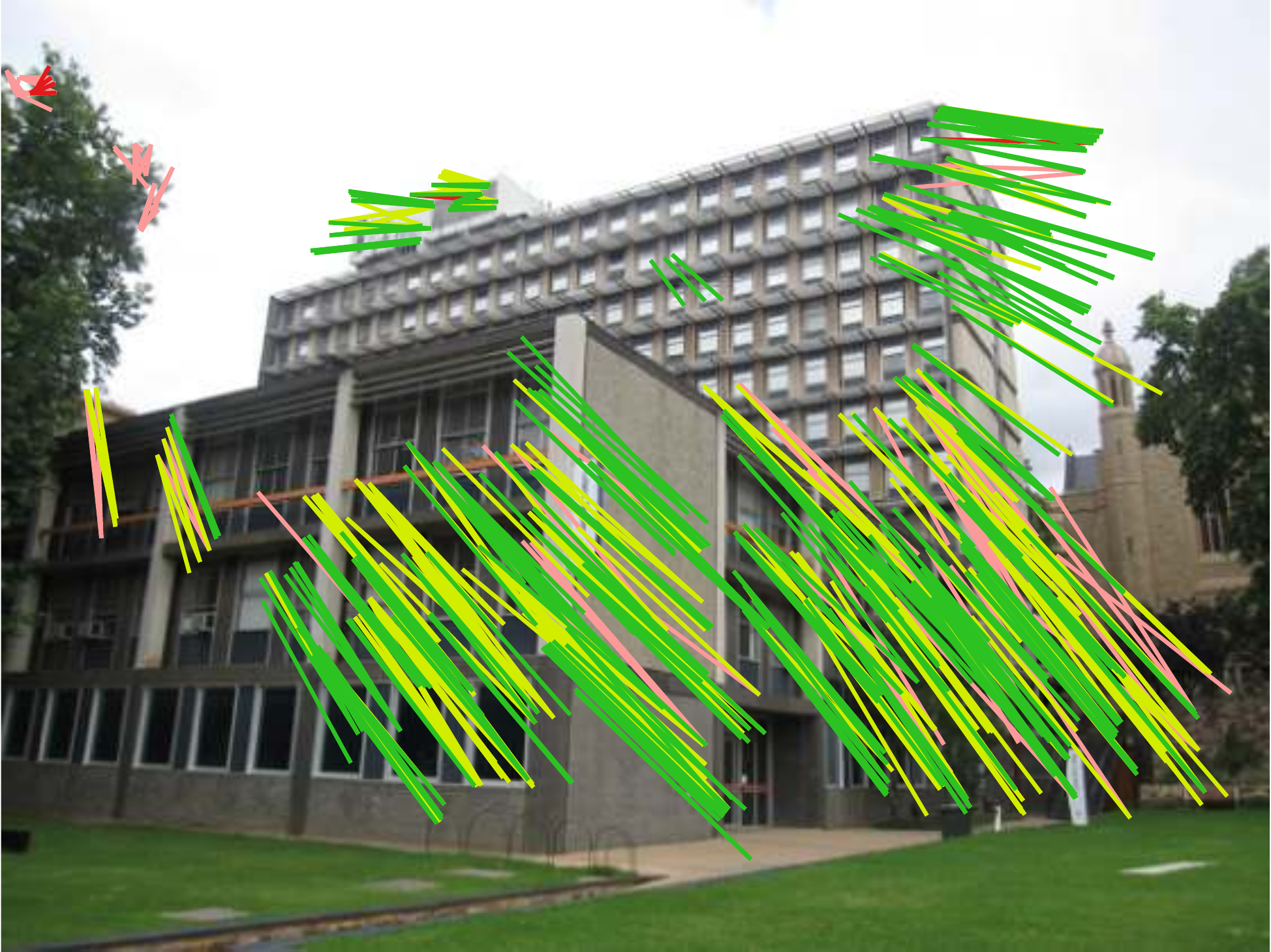}
	\\
	\vspace{0.5em}
	\rotatebox[origin=l]{90}{\mbox{\hspace{2em}VFC}}
	\includegraphics[height=7.5em]{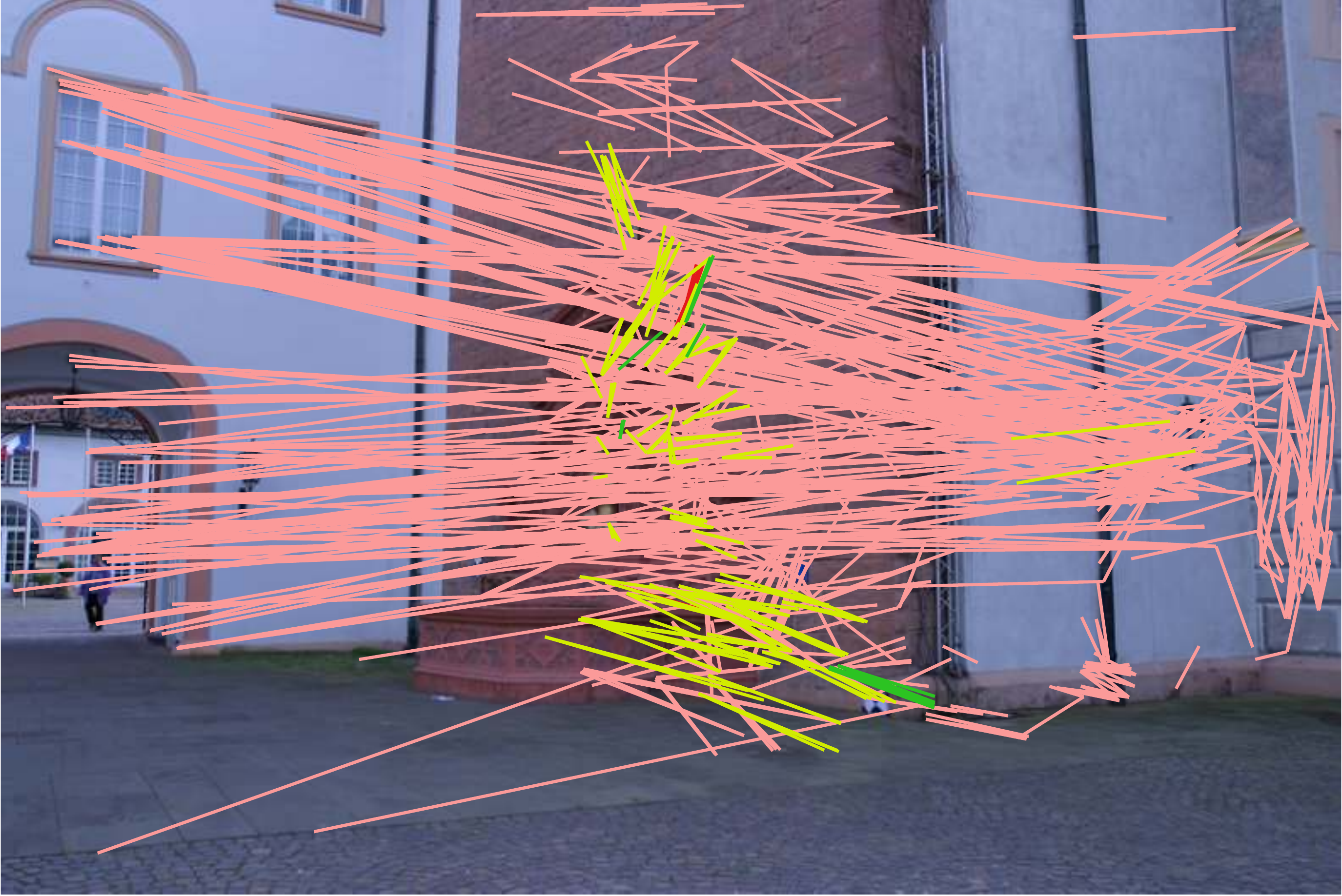}
	\includegraphics[height=7.5em]{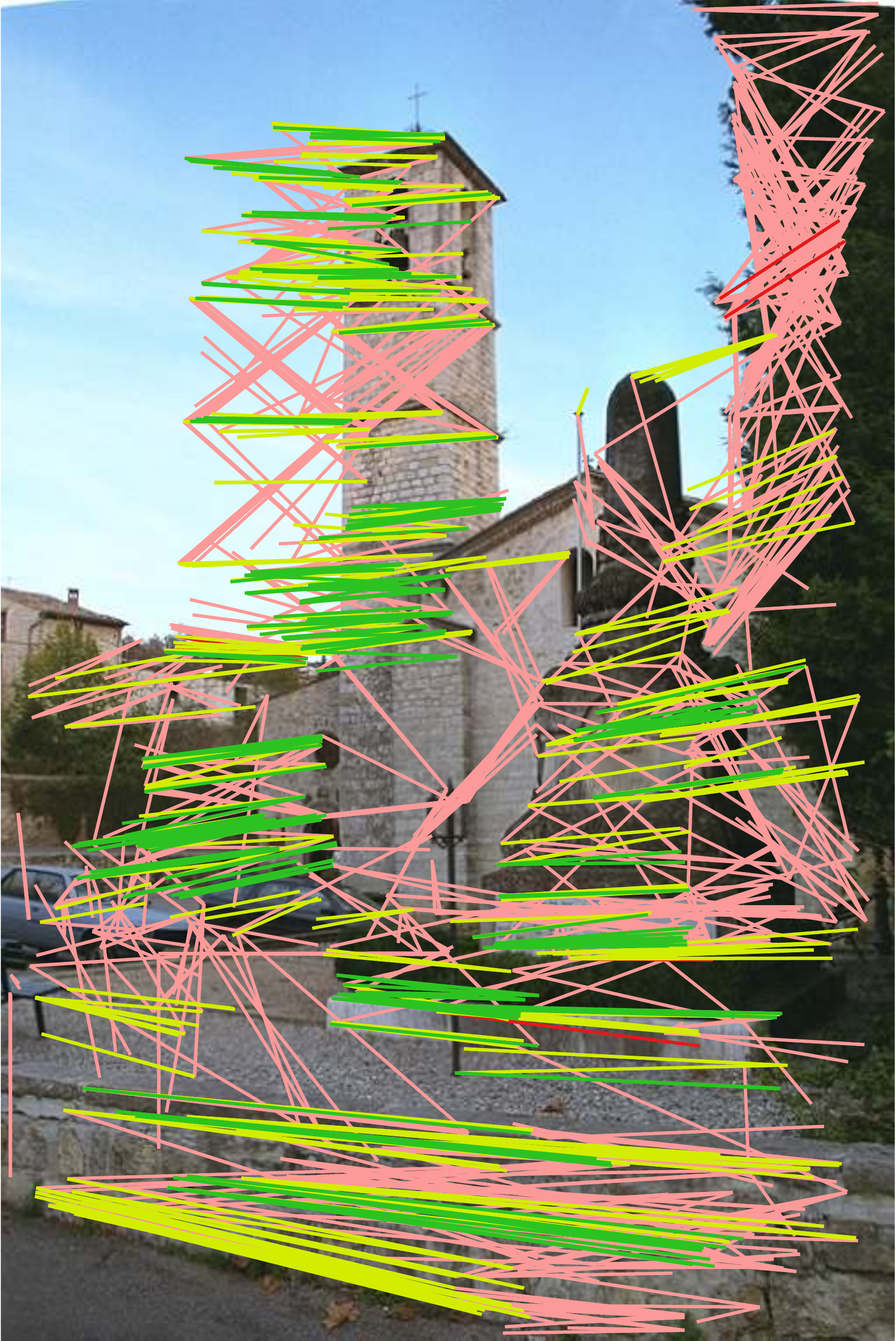}
	\includegraphics[height=7.5em]{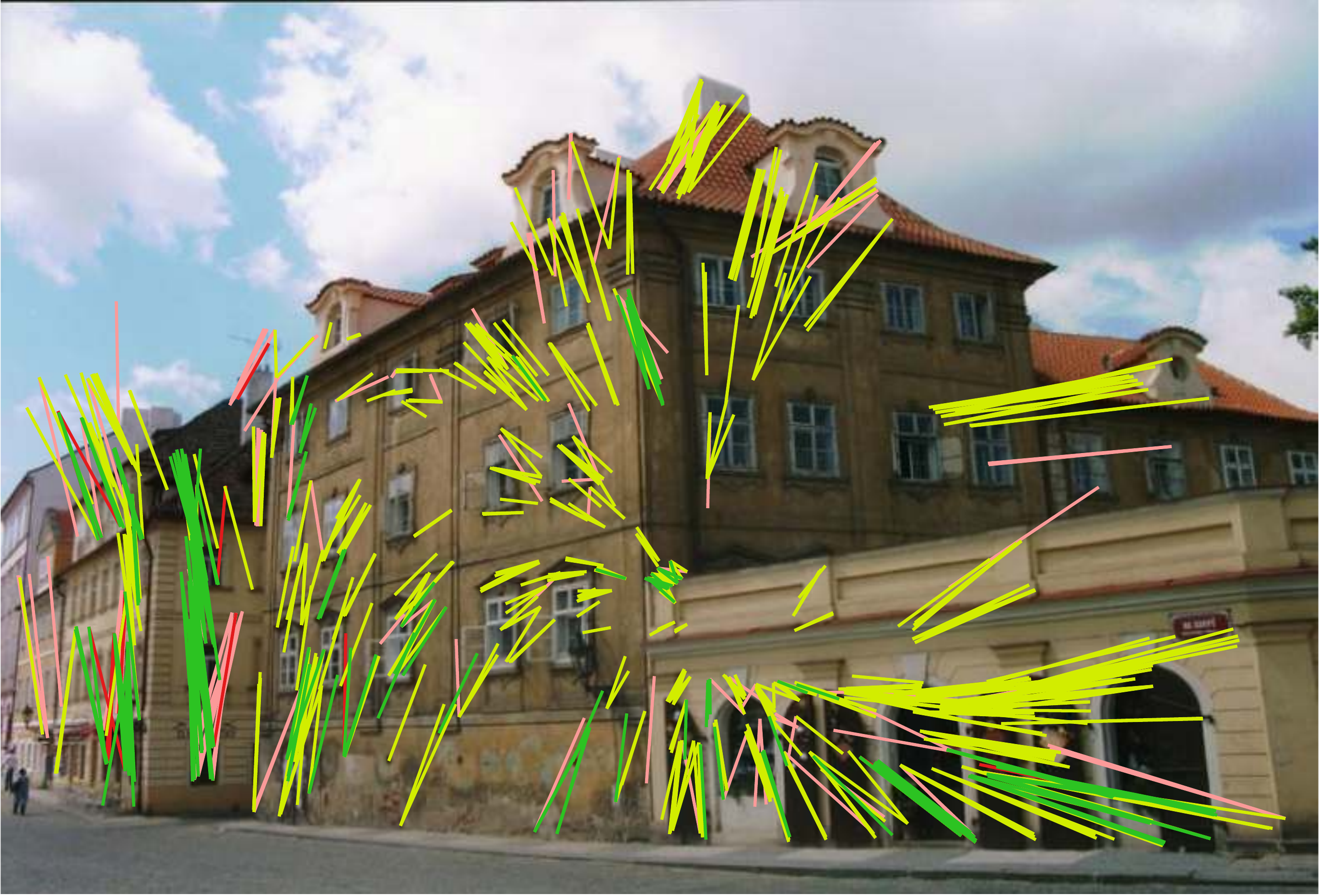}
	\hspace{0.05em}	
	\includegraphics[height=7.5em]{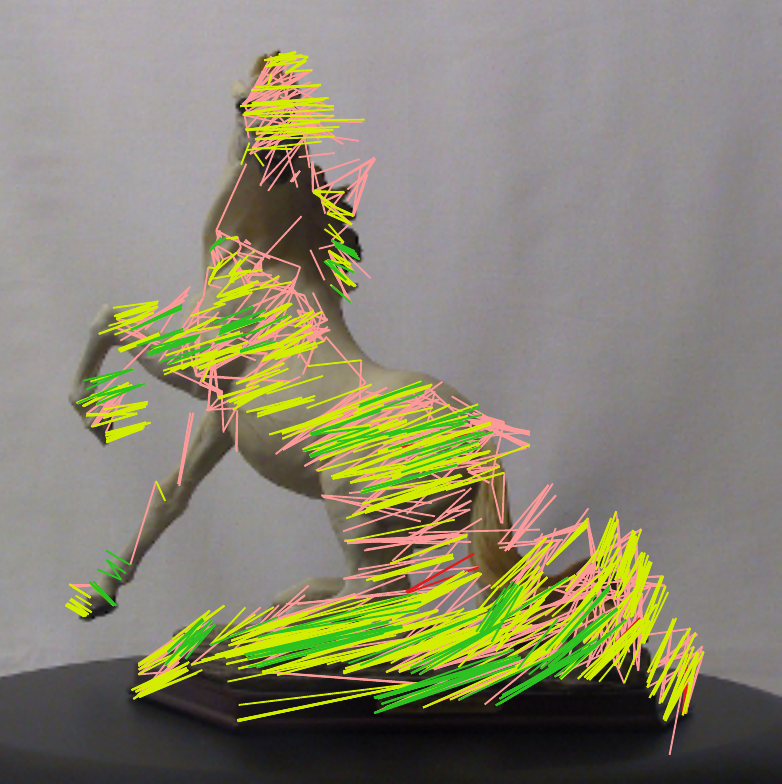}
	\hspace{0.05em}
	\includegraphics[height=7.5em]{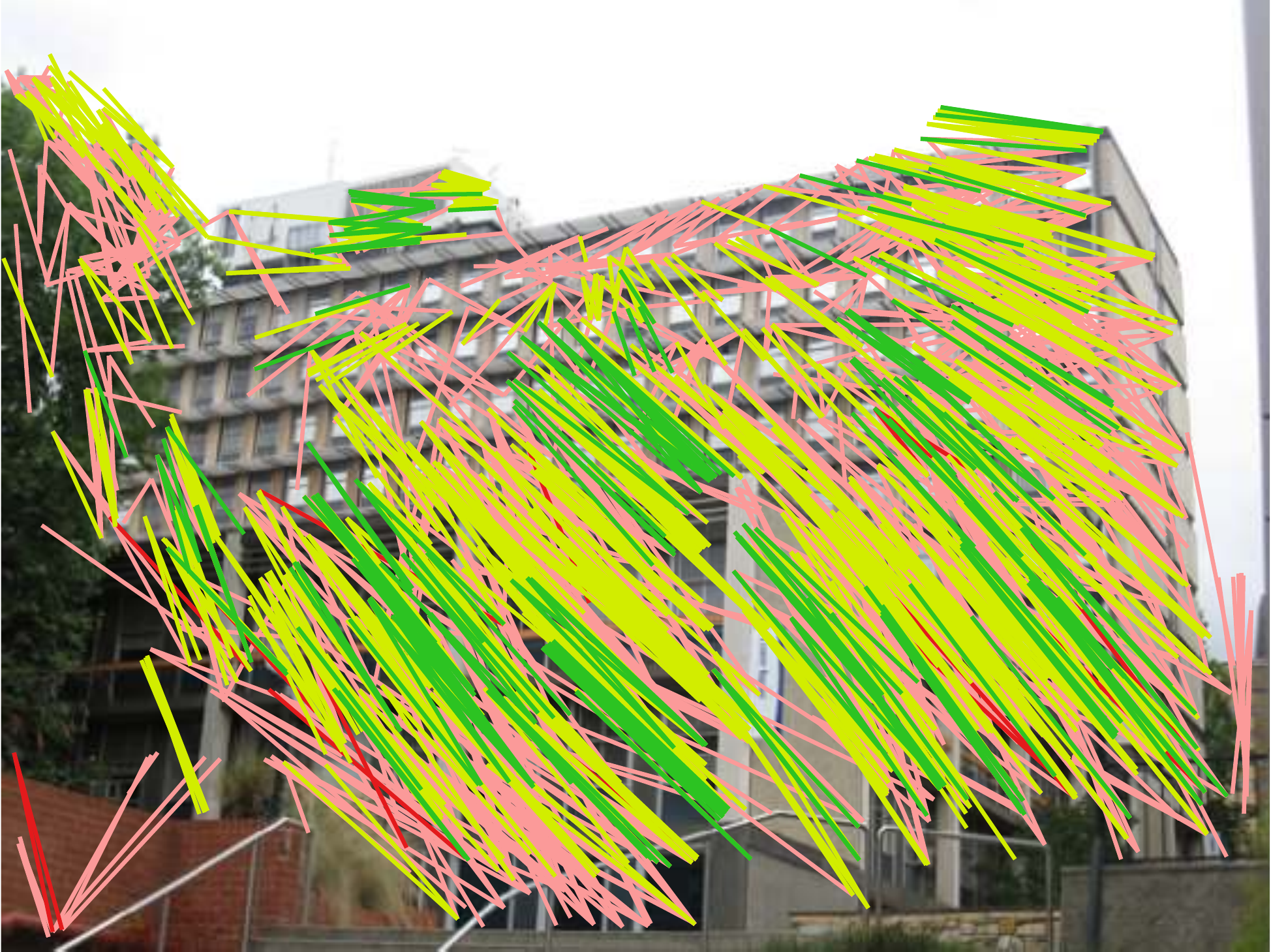}
	\\
	\vspace{0.5em}
	\rotatebox[origin=l]{90}{\mbox{\hspace{2em}LLT}}
	\includegraphics[height=7.5em]{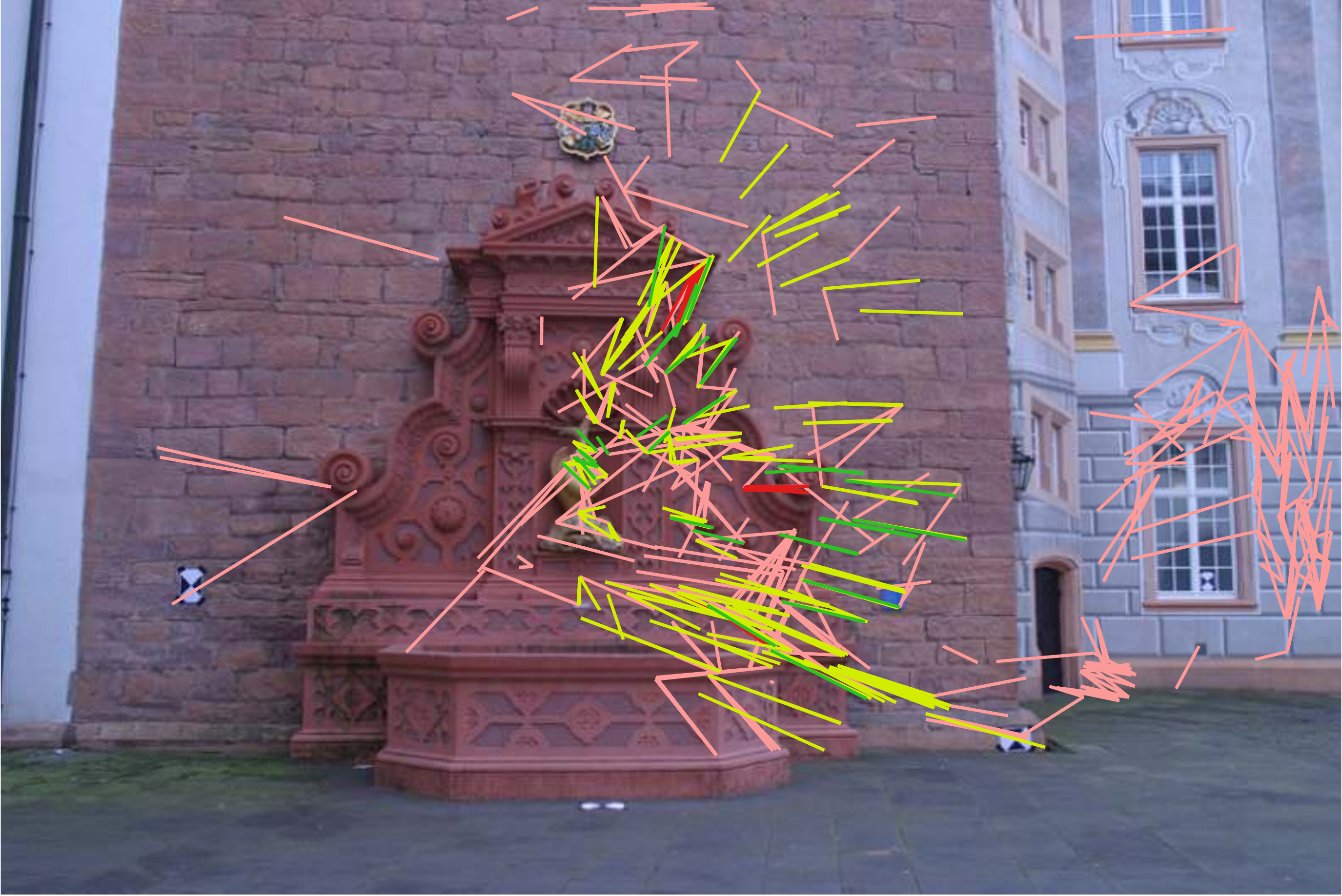}
	\includegraphics[height=7.5em]{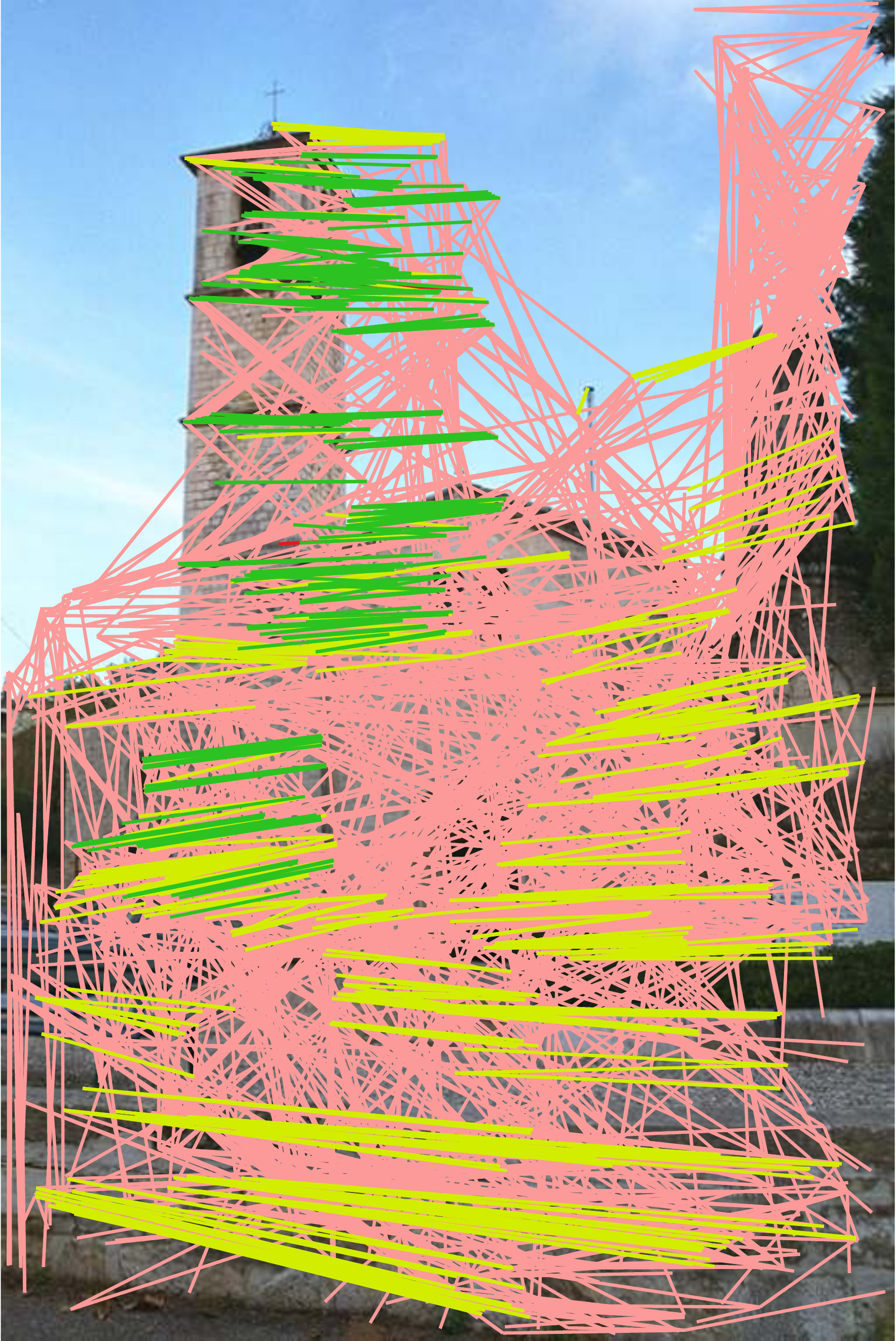}
	\includegraphics[height=7.5em]{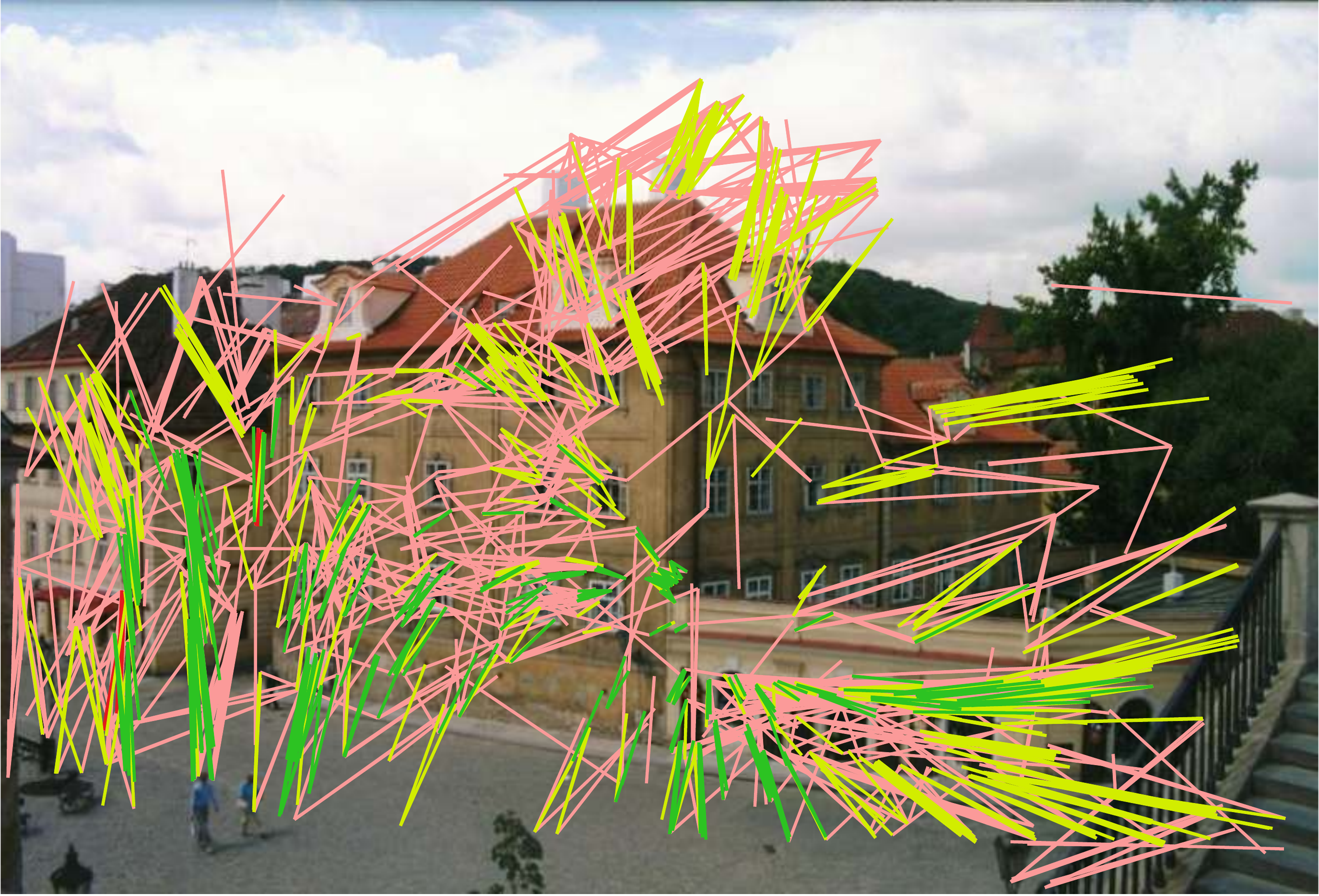}
	\includegraphics[height=7.5em]{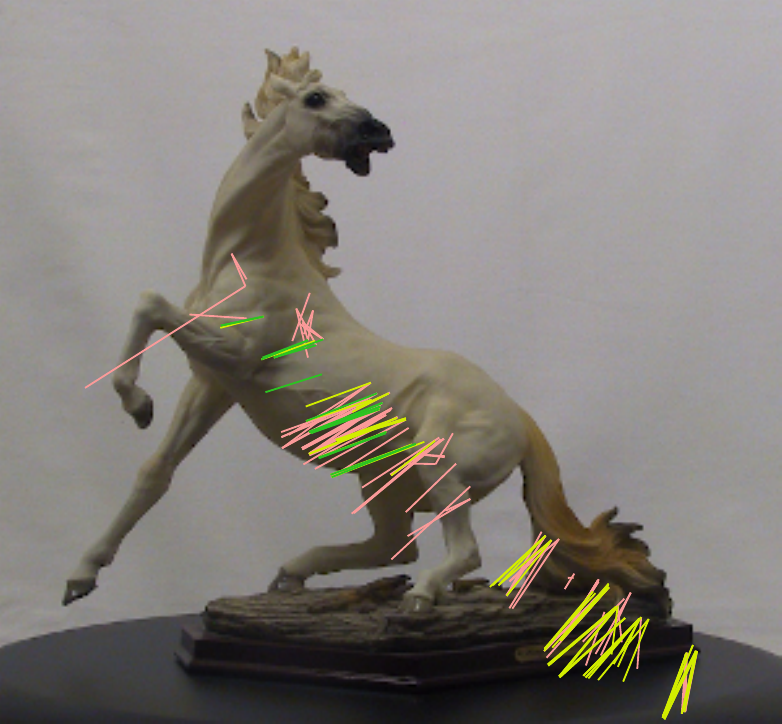}
	\includegraphics[height=7.5em]{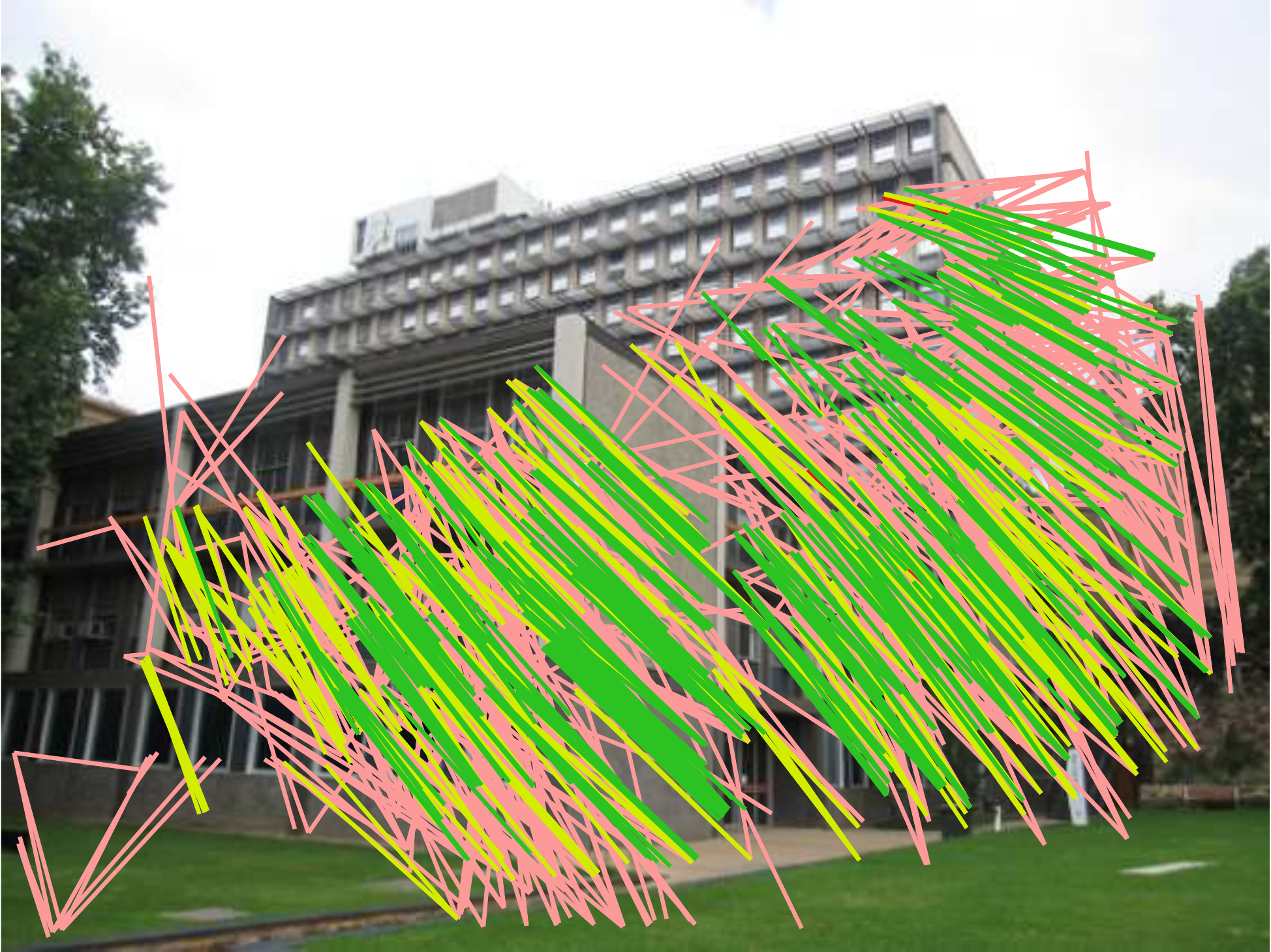}
	\\
	\begin{flushleft}
		\hspace*{7.5em}Fountain01\hspace{3.75em}Valbonne\hspace{4.5em}Kampa\hspace{7em}Horse\hspace{5.75em}NapierB
	\end{flushleft}
	\caption{\label{example_2a}
		Planar and non-planar local spatial filter matches according to the best configuration setup, the images of the input pair alternate among the rows. Image indexes are reported as suffix when the sequence contains more than two images. For each method inlier (yellow, green) and outlier (red and light red) clusters are shown, as well as the 1SAC filtered matches (green, red) (see Sec.~\ref{eval_dt}, best viewed in color and zoomed in).}
\end{figure*}

\begin{figure*}
	\center
	\rotatebox[origin=l]{90}{\mbox{\hspace{0.5em}RFM-SCAN}}
	\includegraphics[height=7.5em]{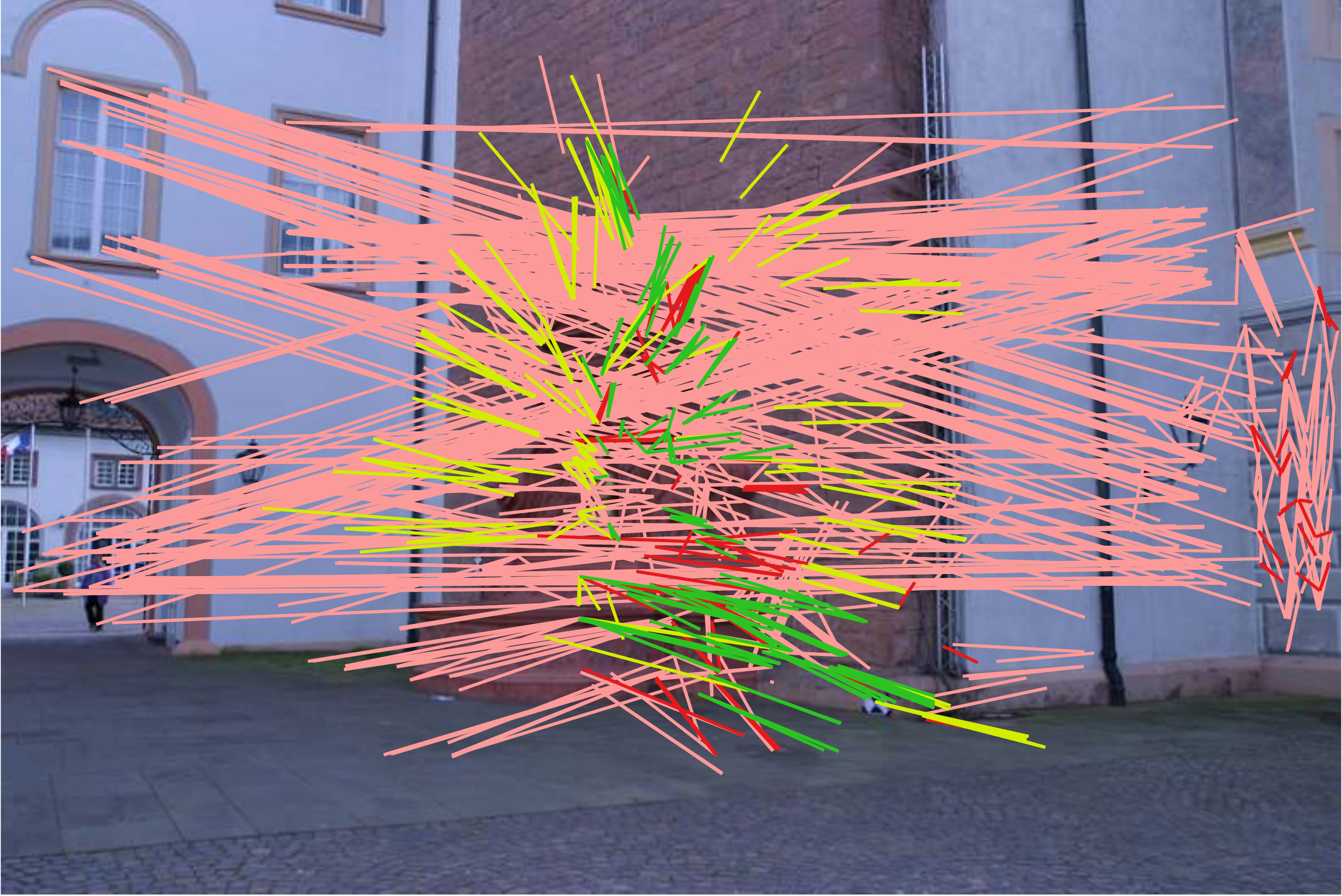}
	\includegraphics[height=7.5em]{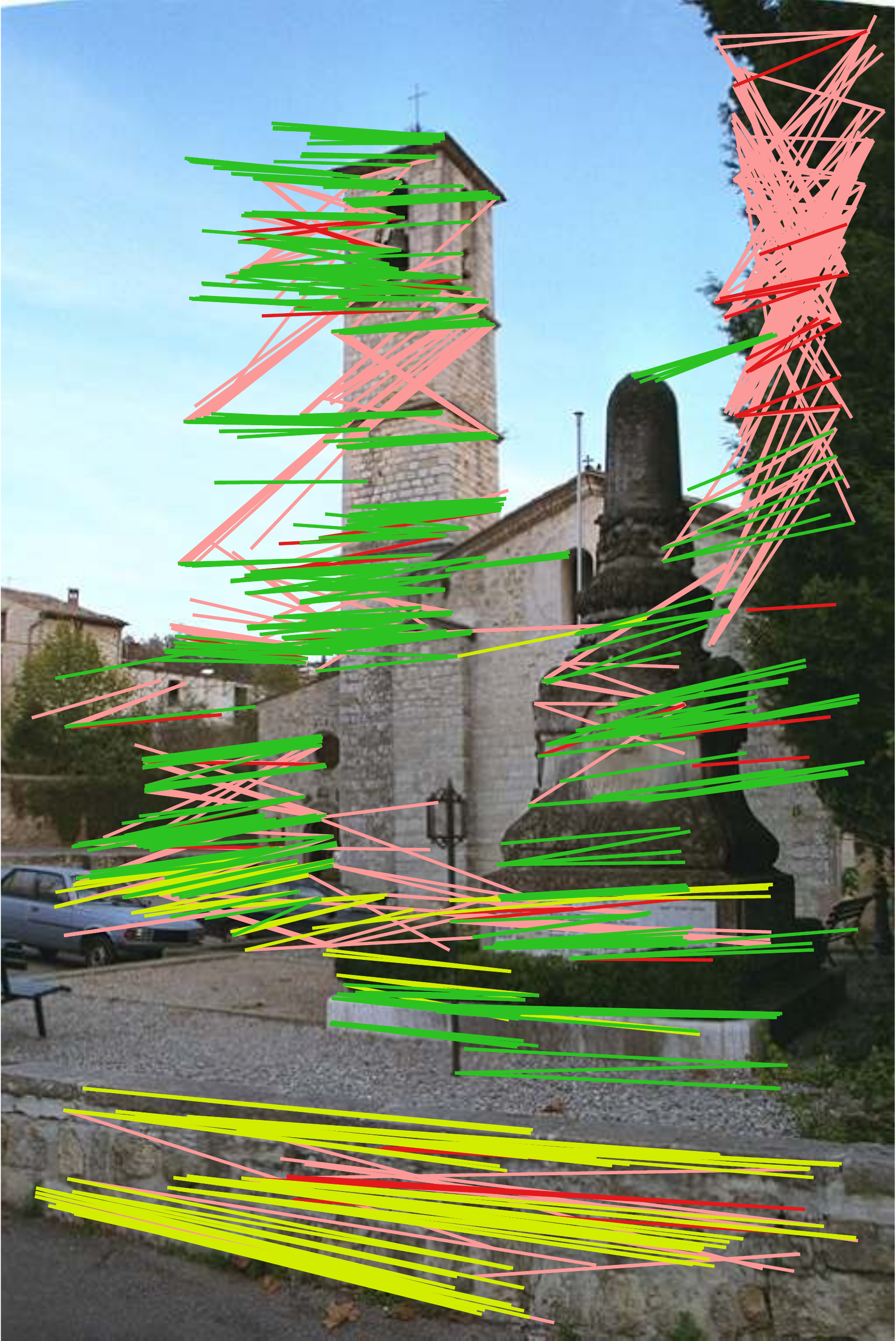}
	\includegraphics[height=7.5em]{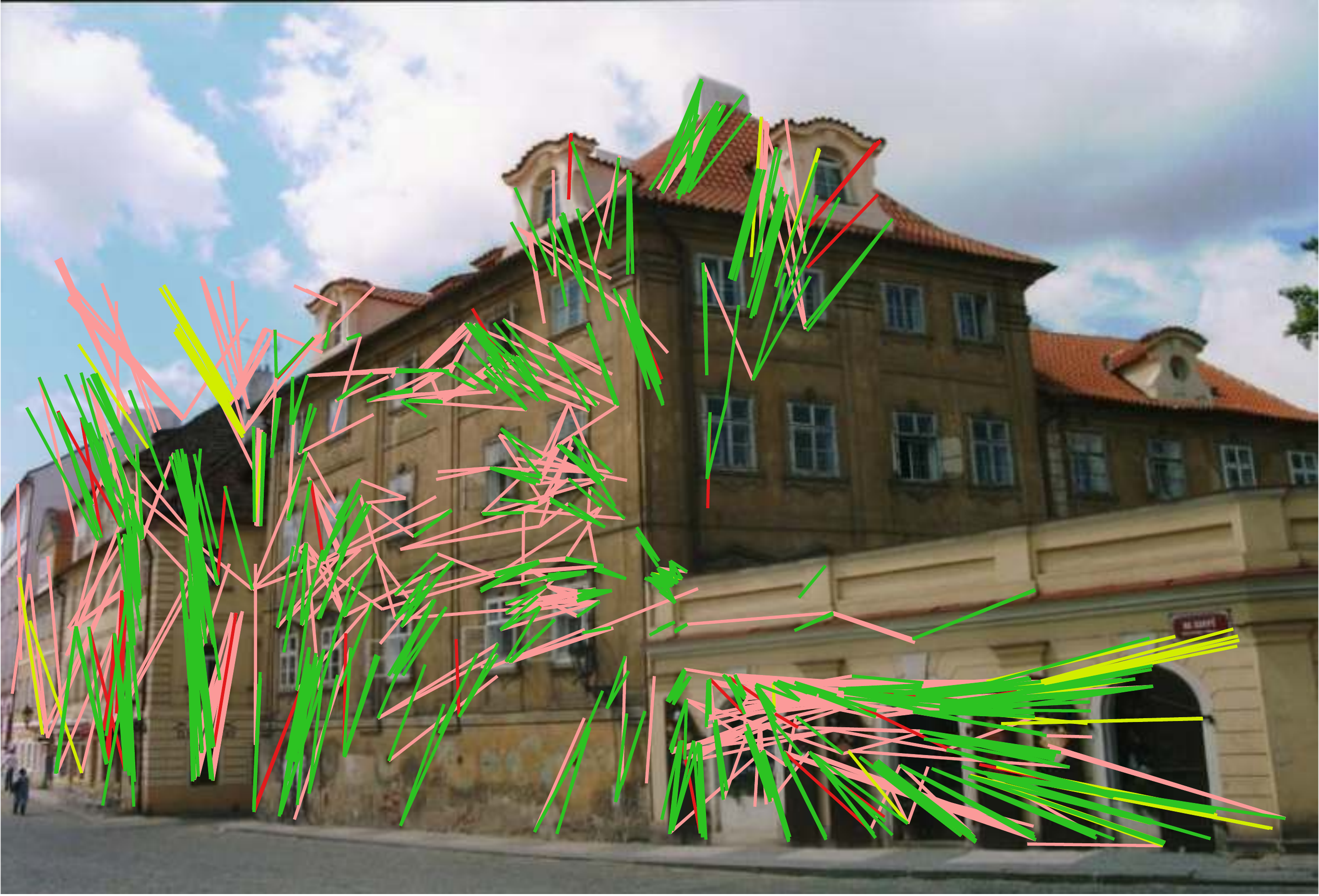}
	\hspace{0.05em}
	\includegraphics[height=7.5em]{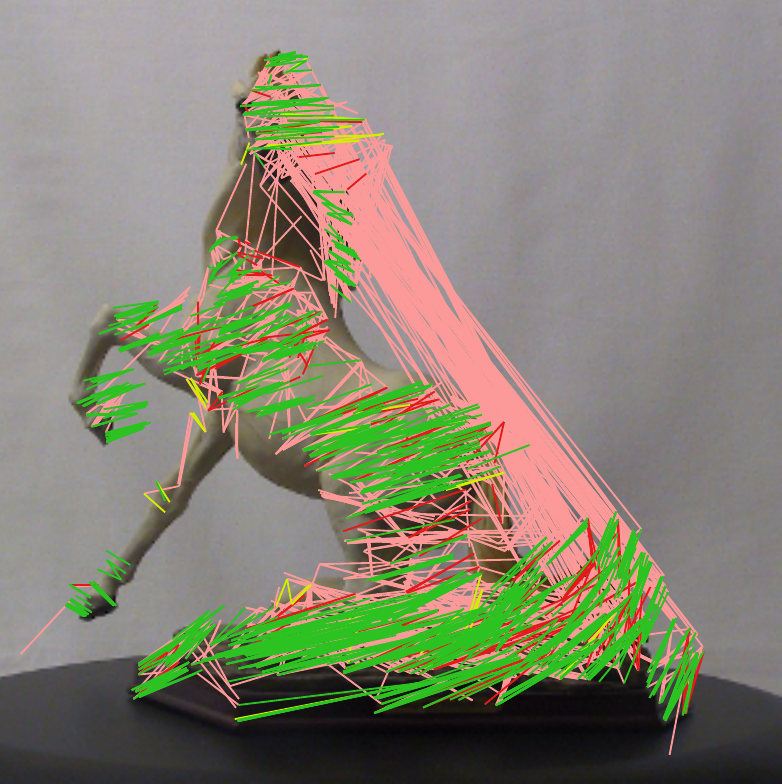}
	\hspace{0.05em}
	\includegraphics[height=7.5em]{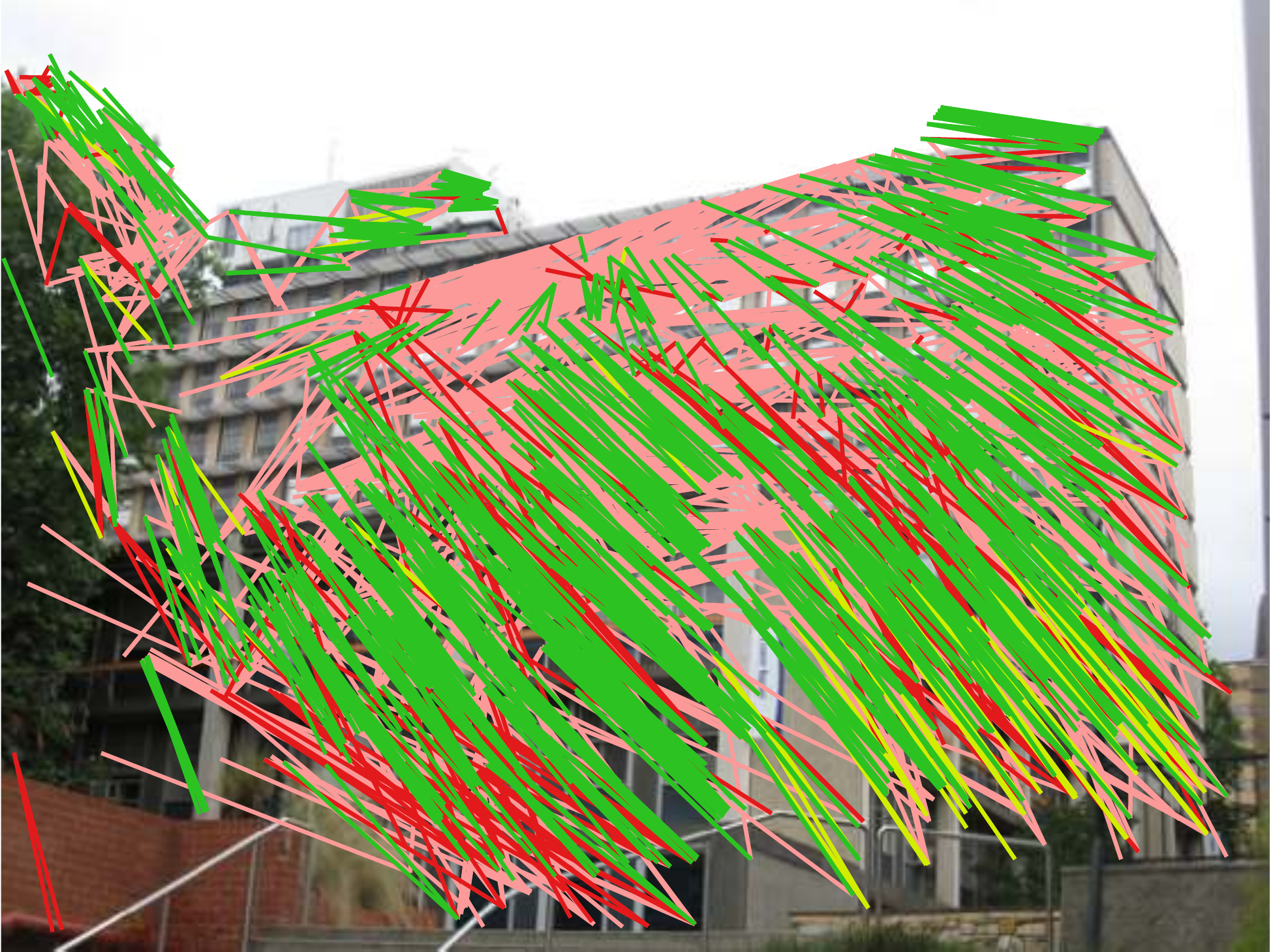}
	\\
	\vspace{0.5em}
	\rotatebox[origin=l]{90}{\mbox{\hspace{1em}AdaLAM}}
	\includegraphics[height=7.5em]{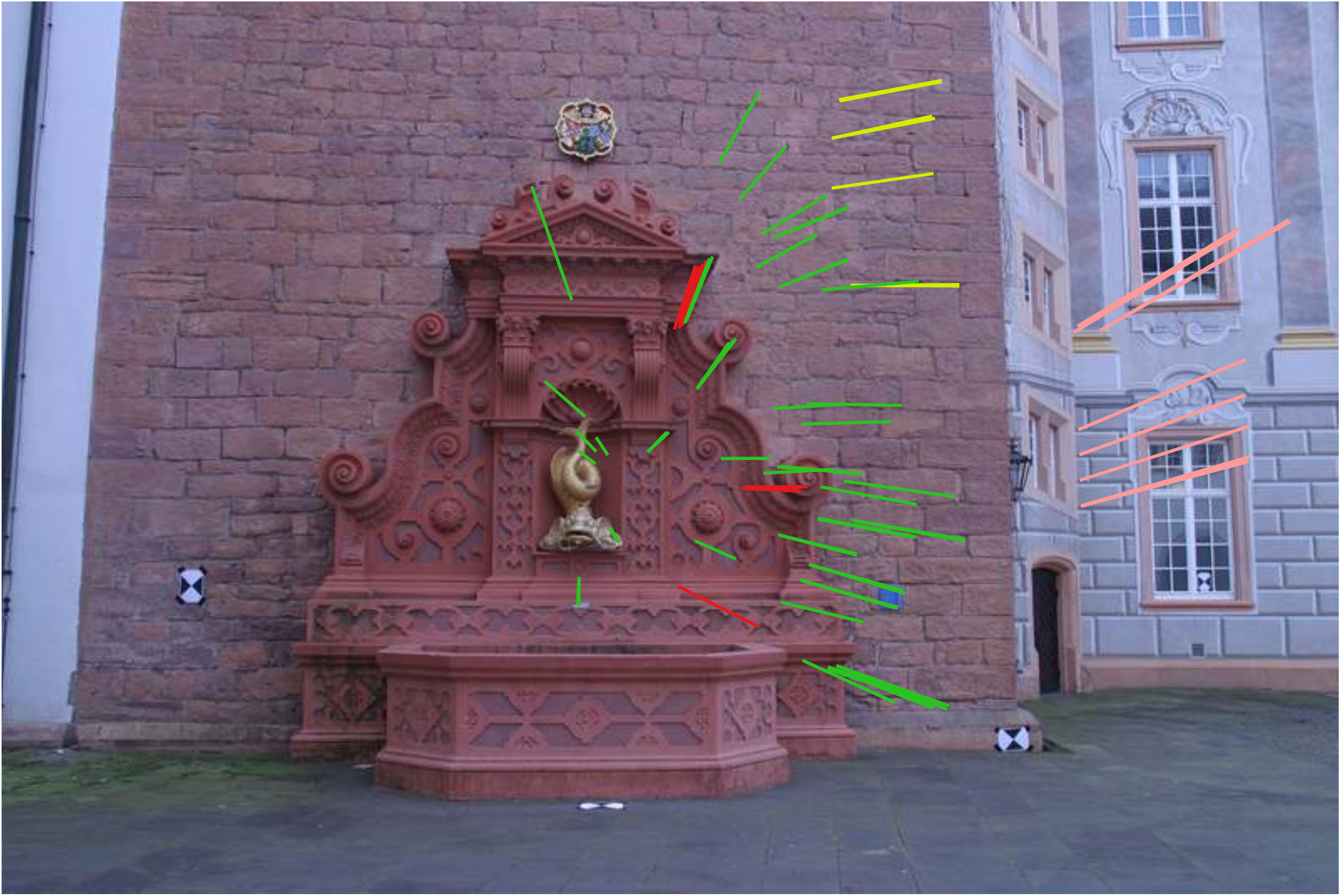}
	\includegraphics[height=7.5em]{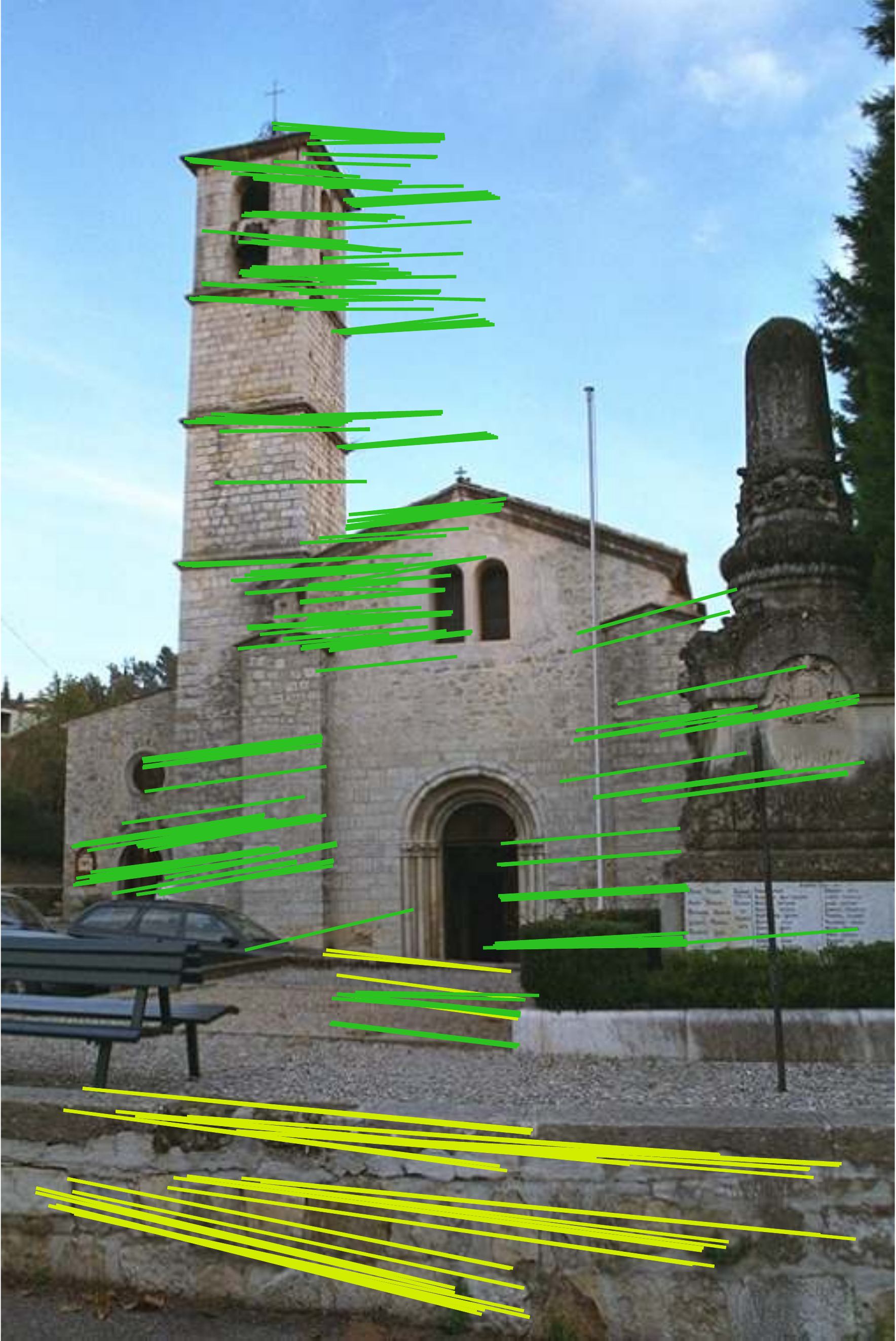}
	\includegraphics[height=7.5em]{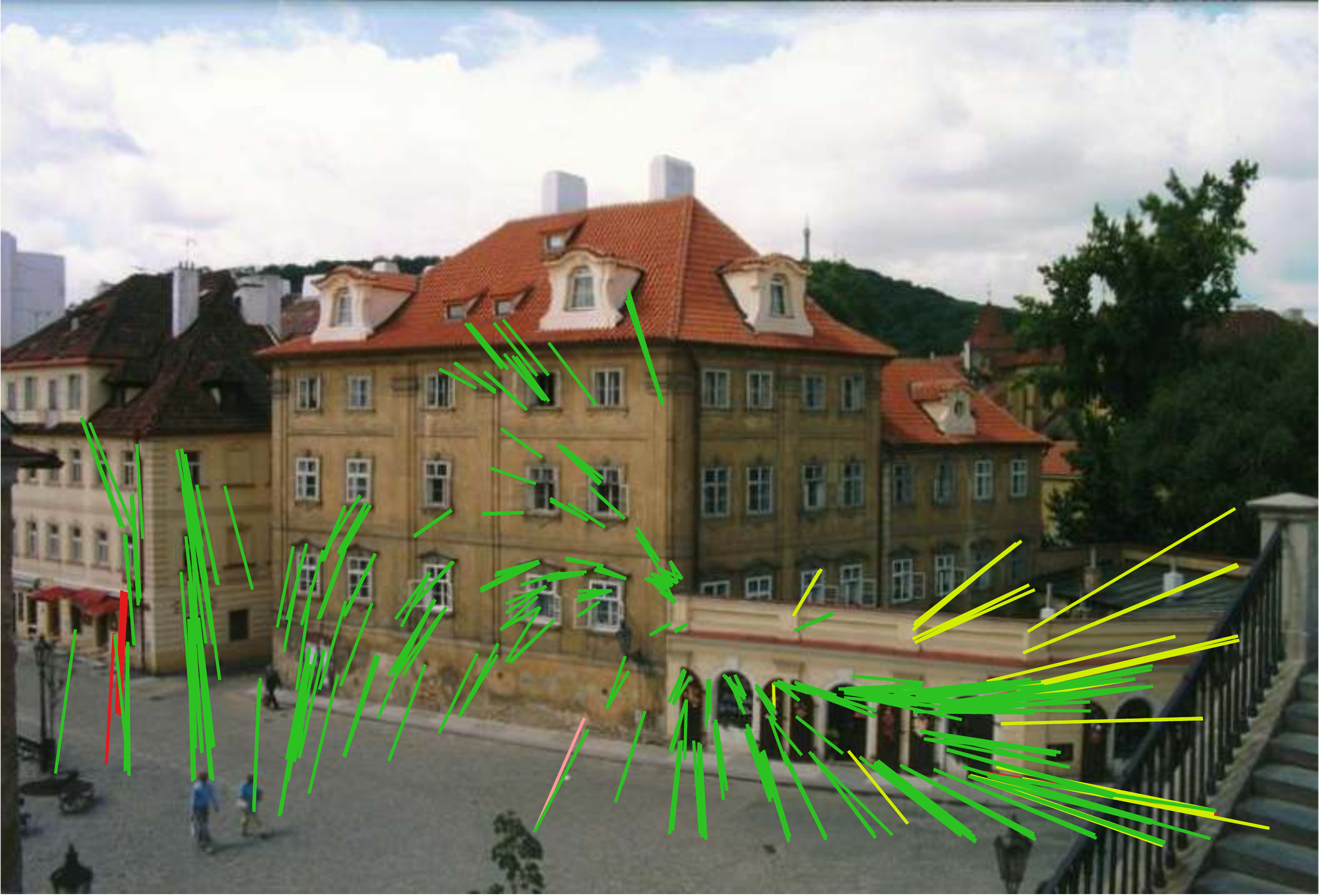}
	\includegraphics[height=7.5em]{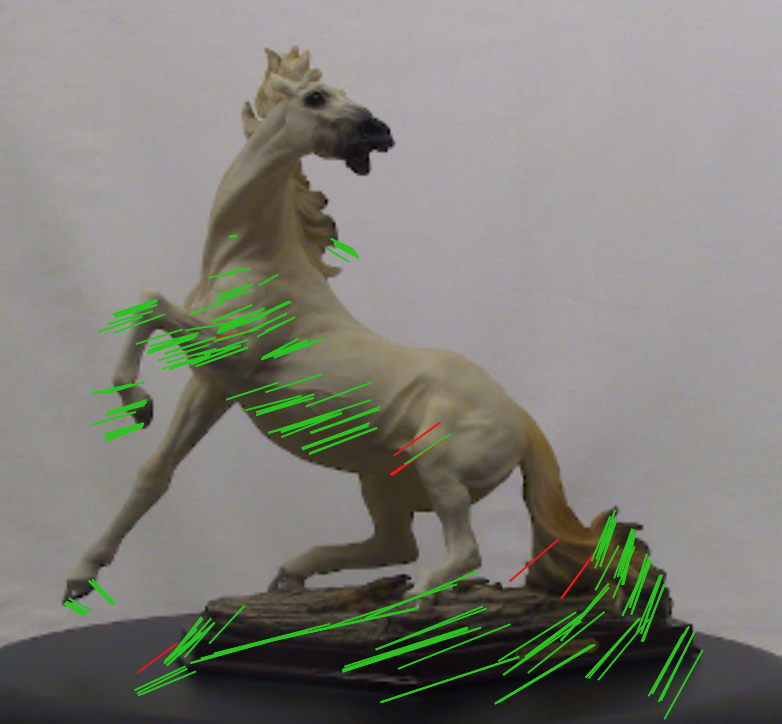}
	\includegraphics[height=7.5em]{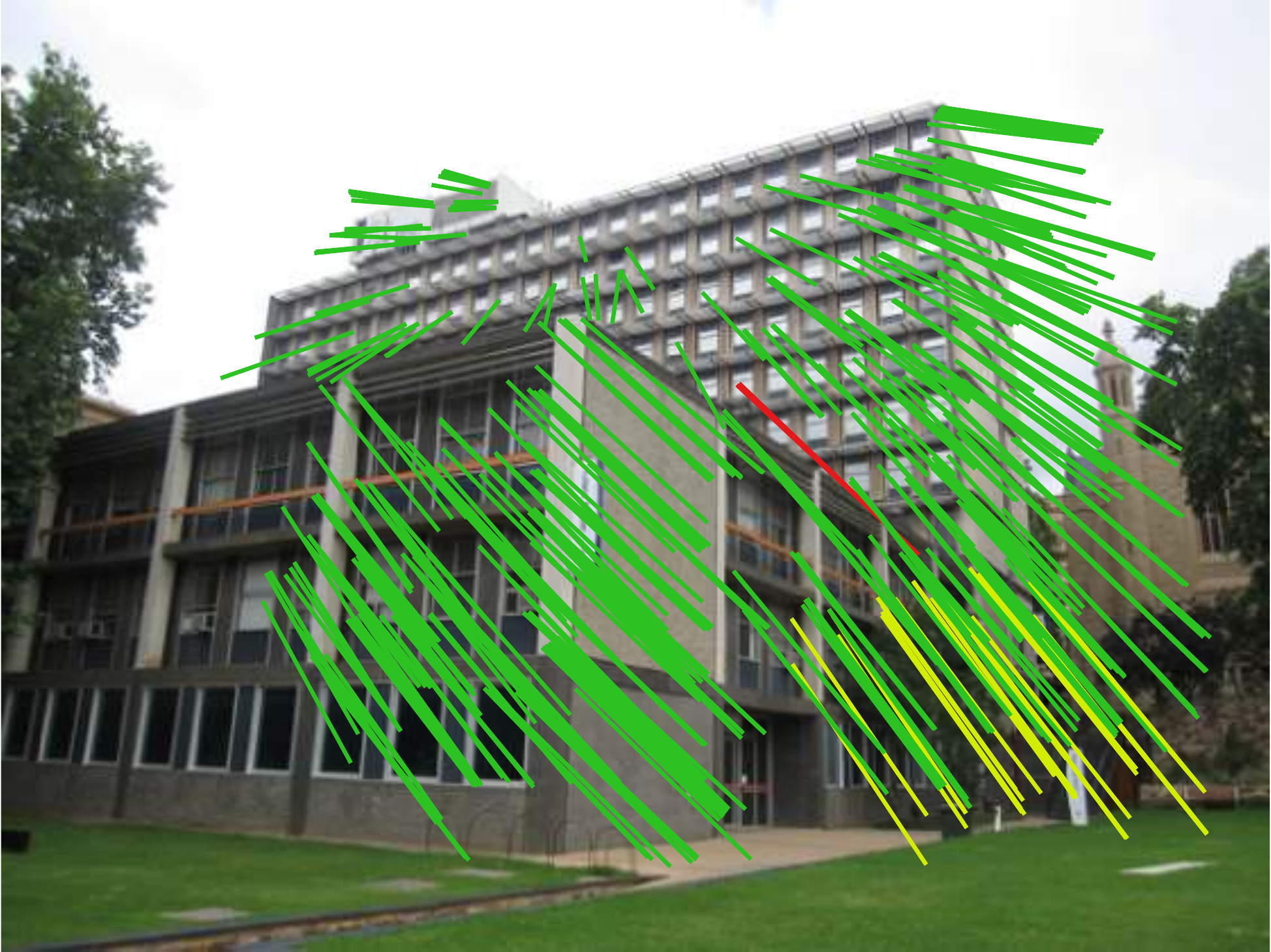}
	\\
	\vspace{0.5em}
	\rotatebox[origin=l]{90}{\mbox{\hspace{2em}OANet}}
	\includegraphics[height=7.5em]{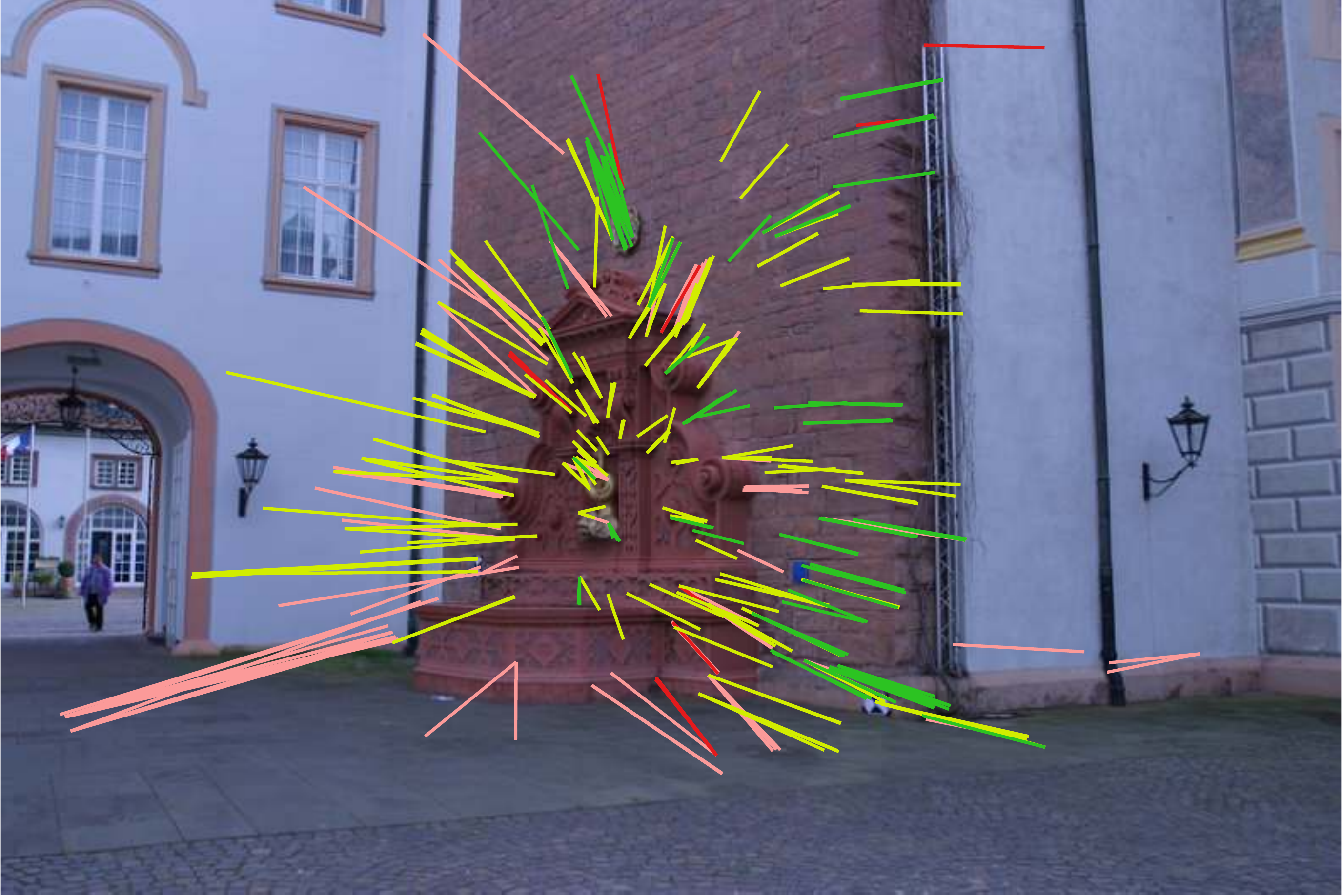}
	\includegraphics[height=7.5em]{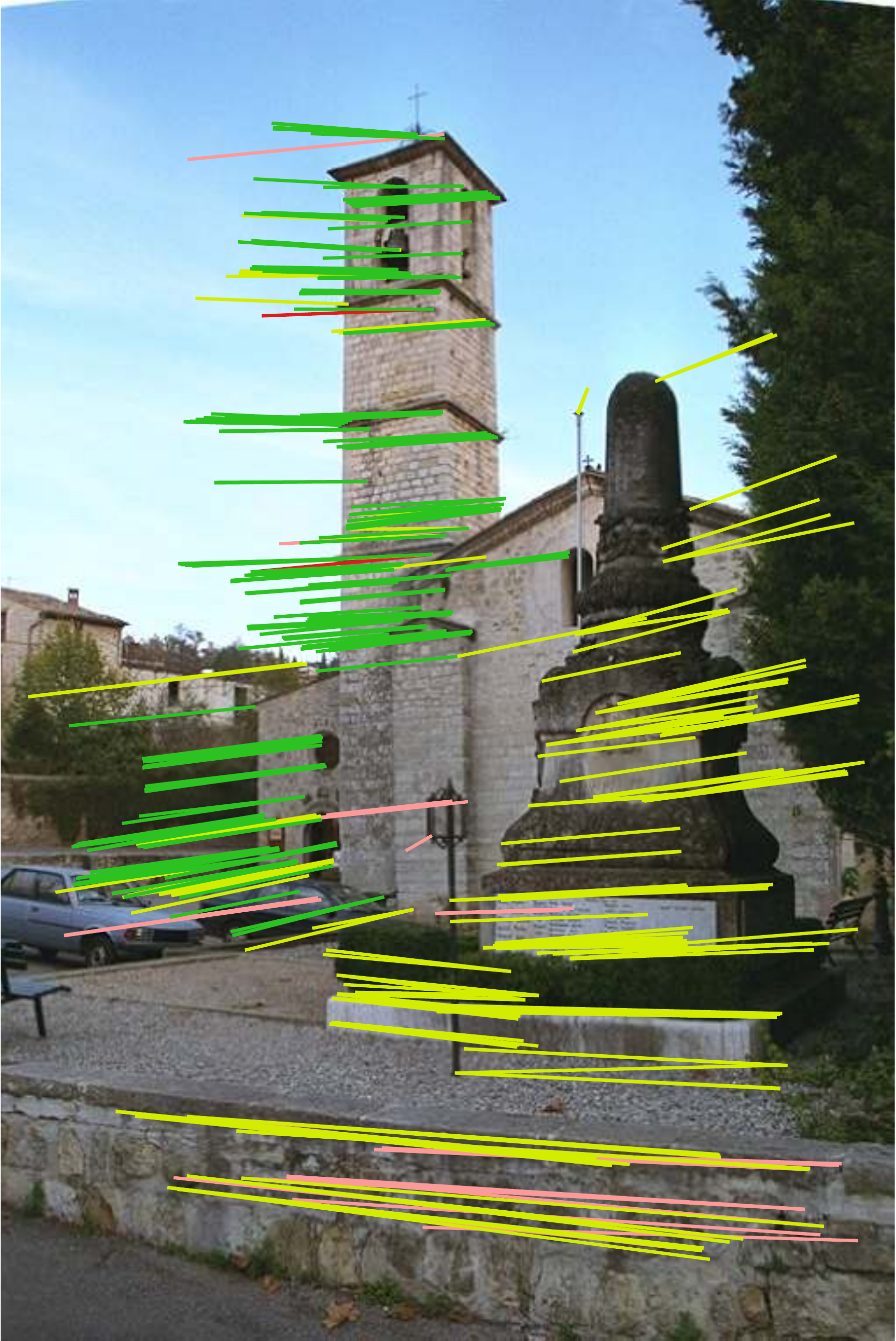}
	\includegraphics[height=7.5em]{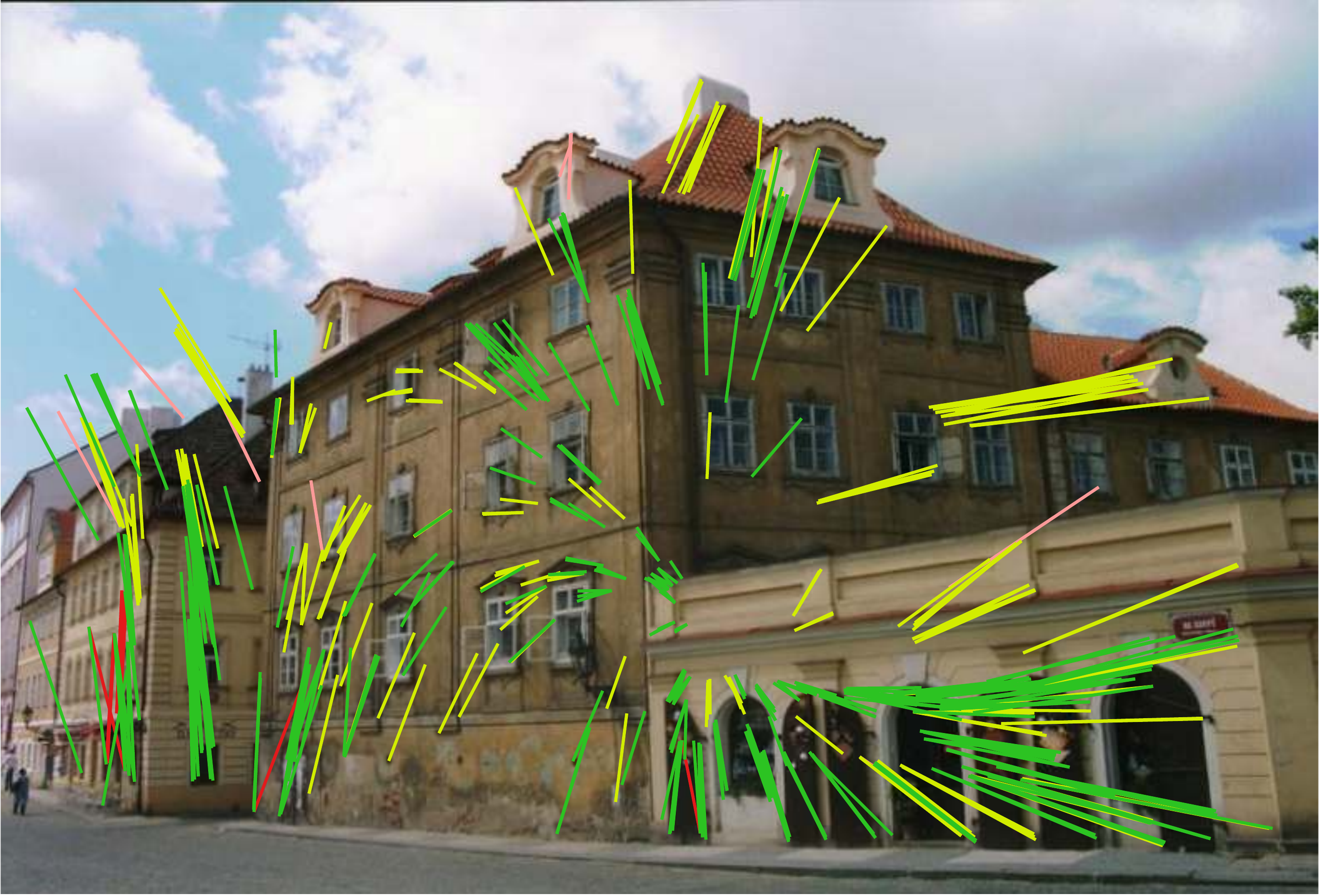}
	\hspace{0.05em}
	\includegraphics[height=7.5em]{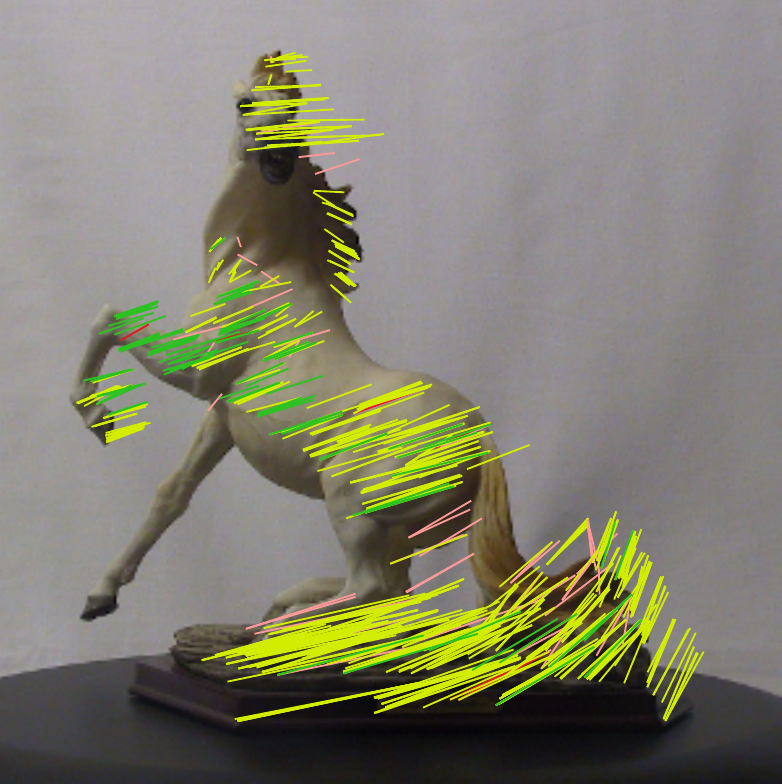}
	\hspace{0.05em}
	\includegraphics[height=7.5em]{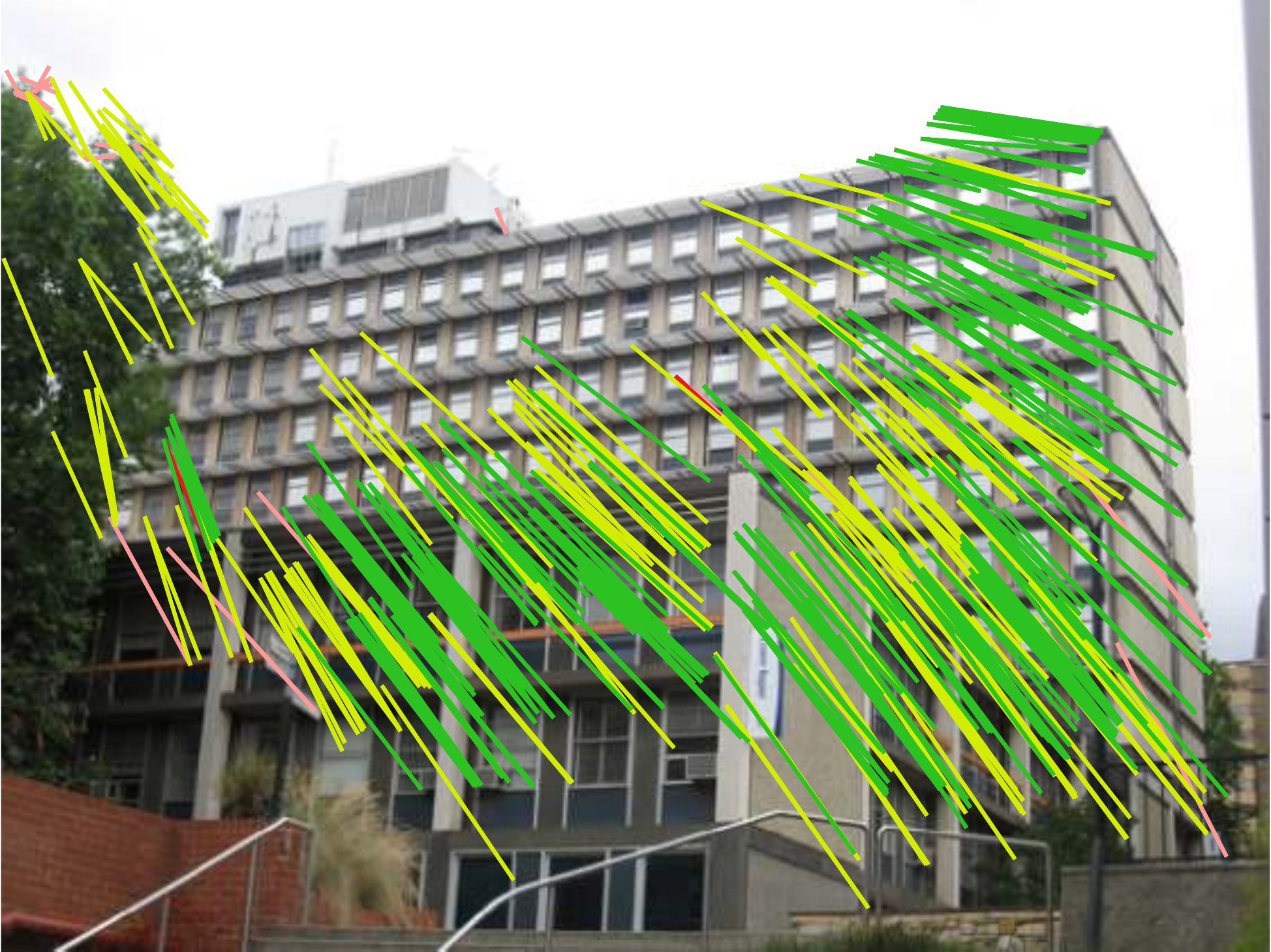}
	\\
	\vspace{0.5em}
	\rotatebox[origin=l]{90}{\mbox{\hspace{2em}ACNe}}
	\includegraphics[height=7.5em]{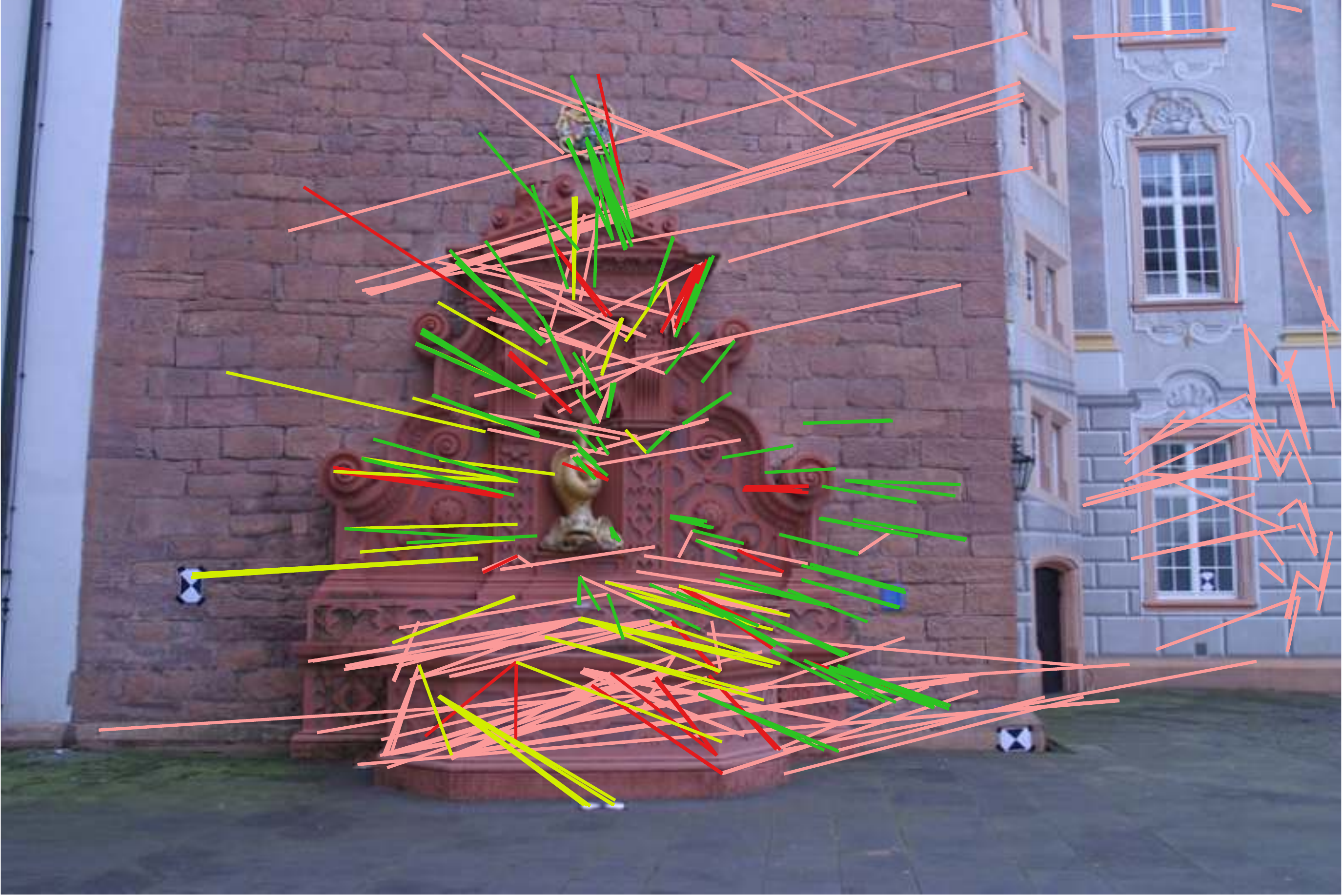}
	\includegraphics[height=7.5em]{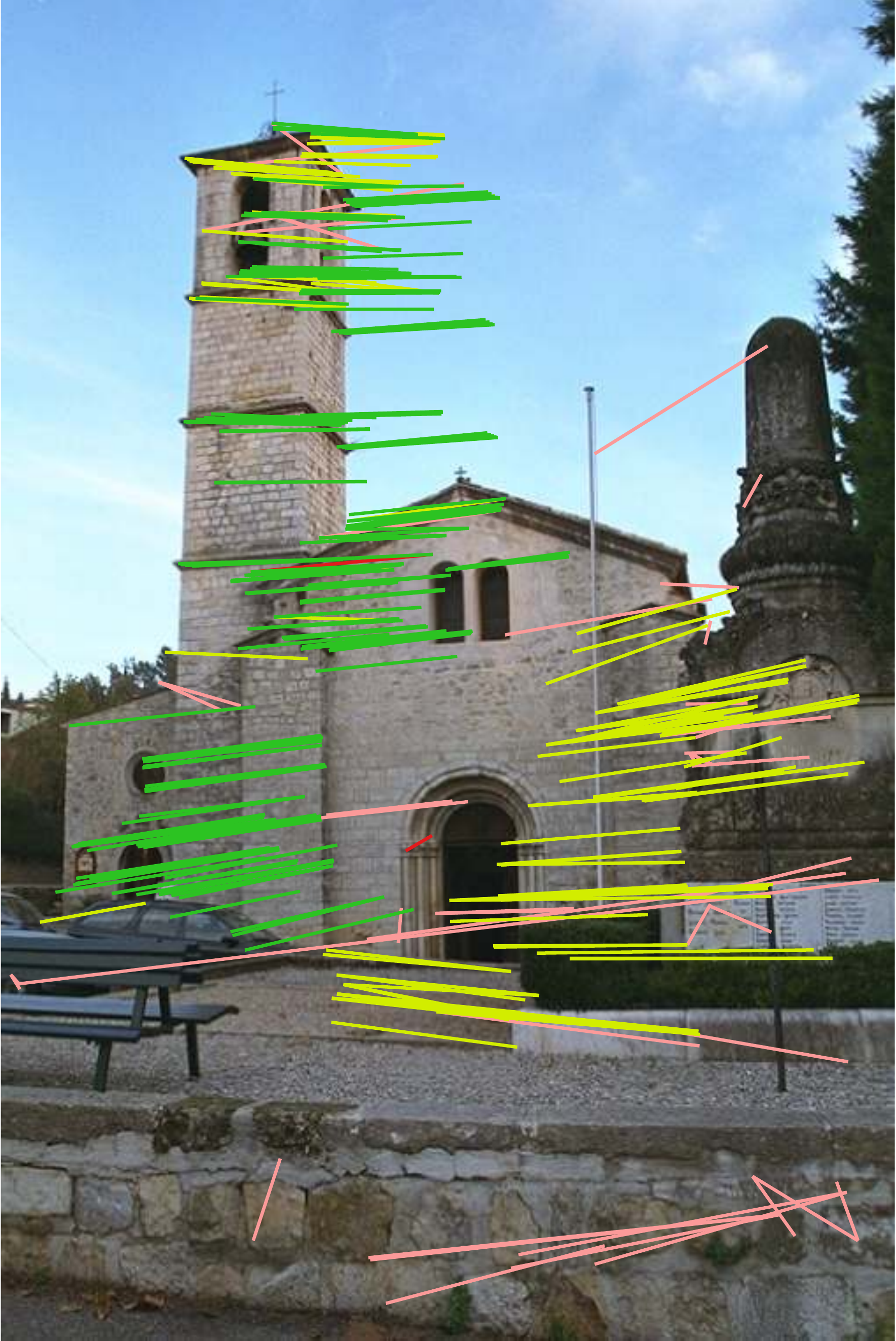}
	\includegraphics[height=7.5em]{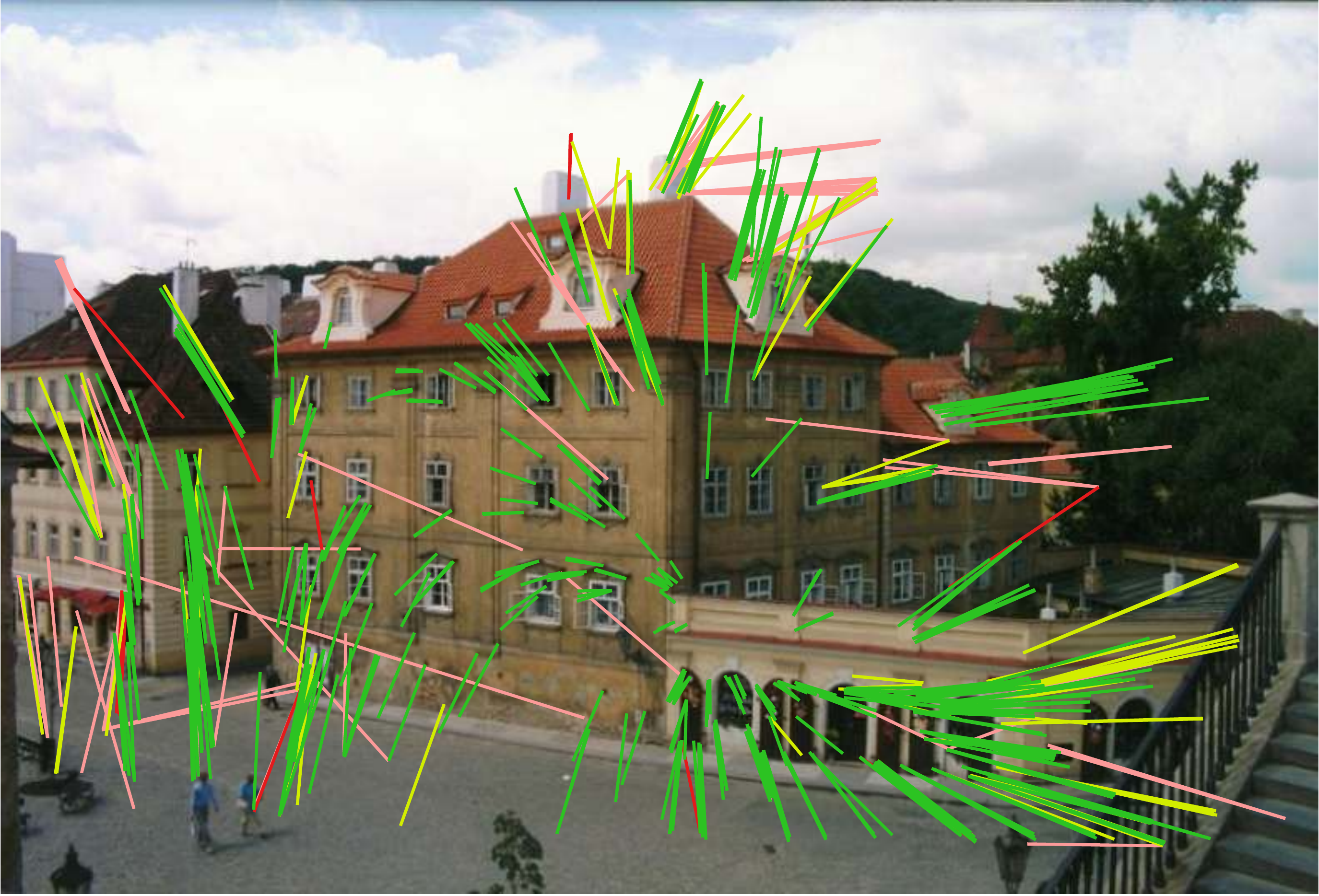}
	\includegraphics[height=7.5em]{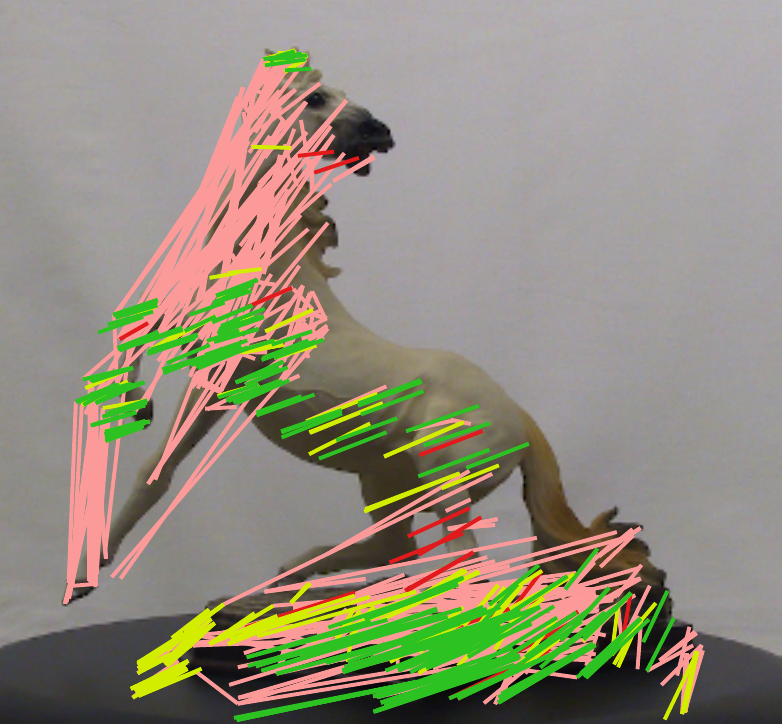}
	\includegraphics[height=7.5em]{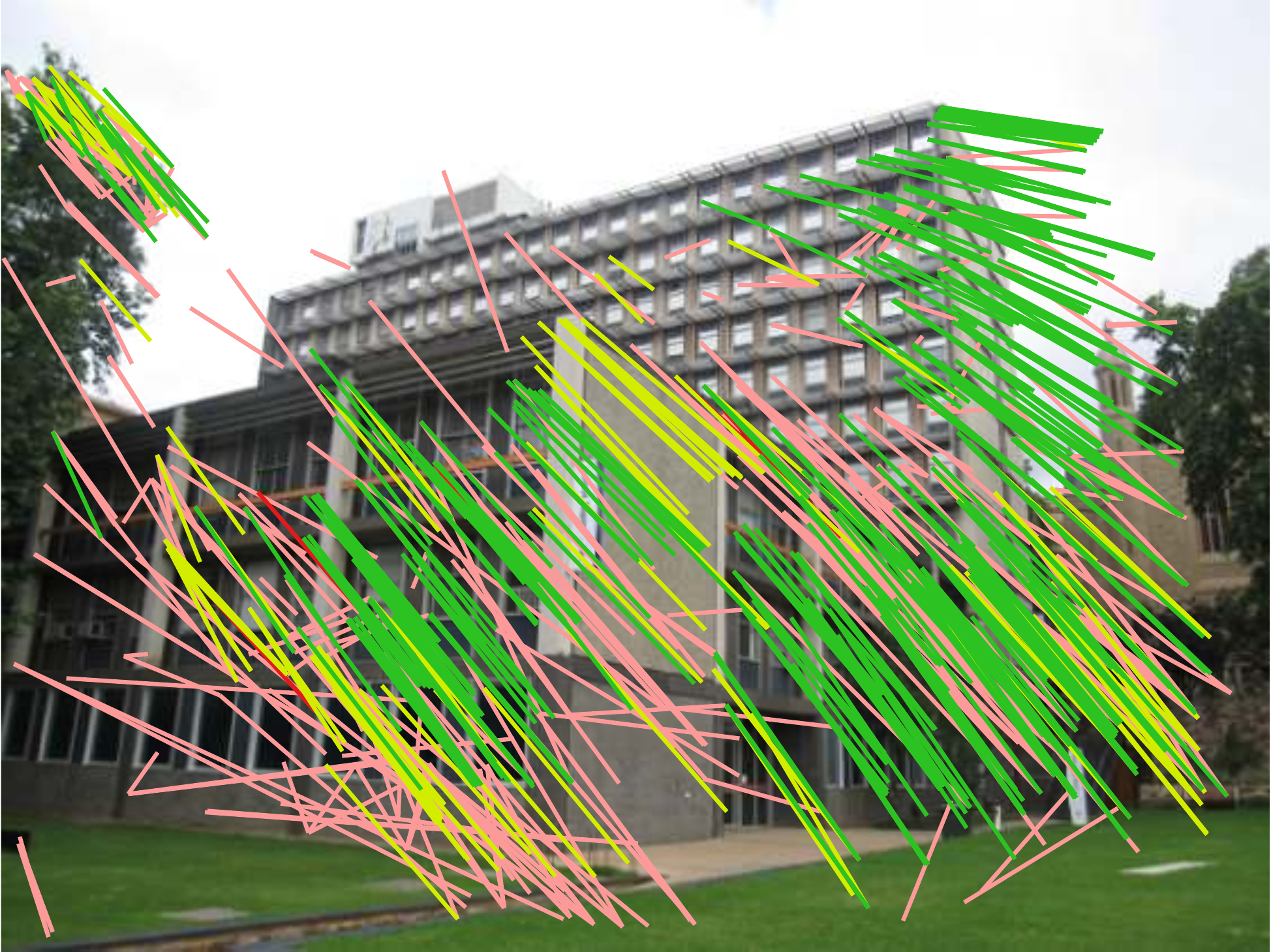}
	\\
	\vspace{0.5em}
	\rotatebox[origin=l]{90}{\mbox{\hspace{2em}PFM}}
	\includegraphics[height=7.5em]{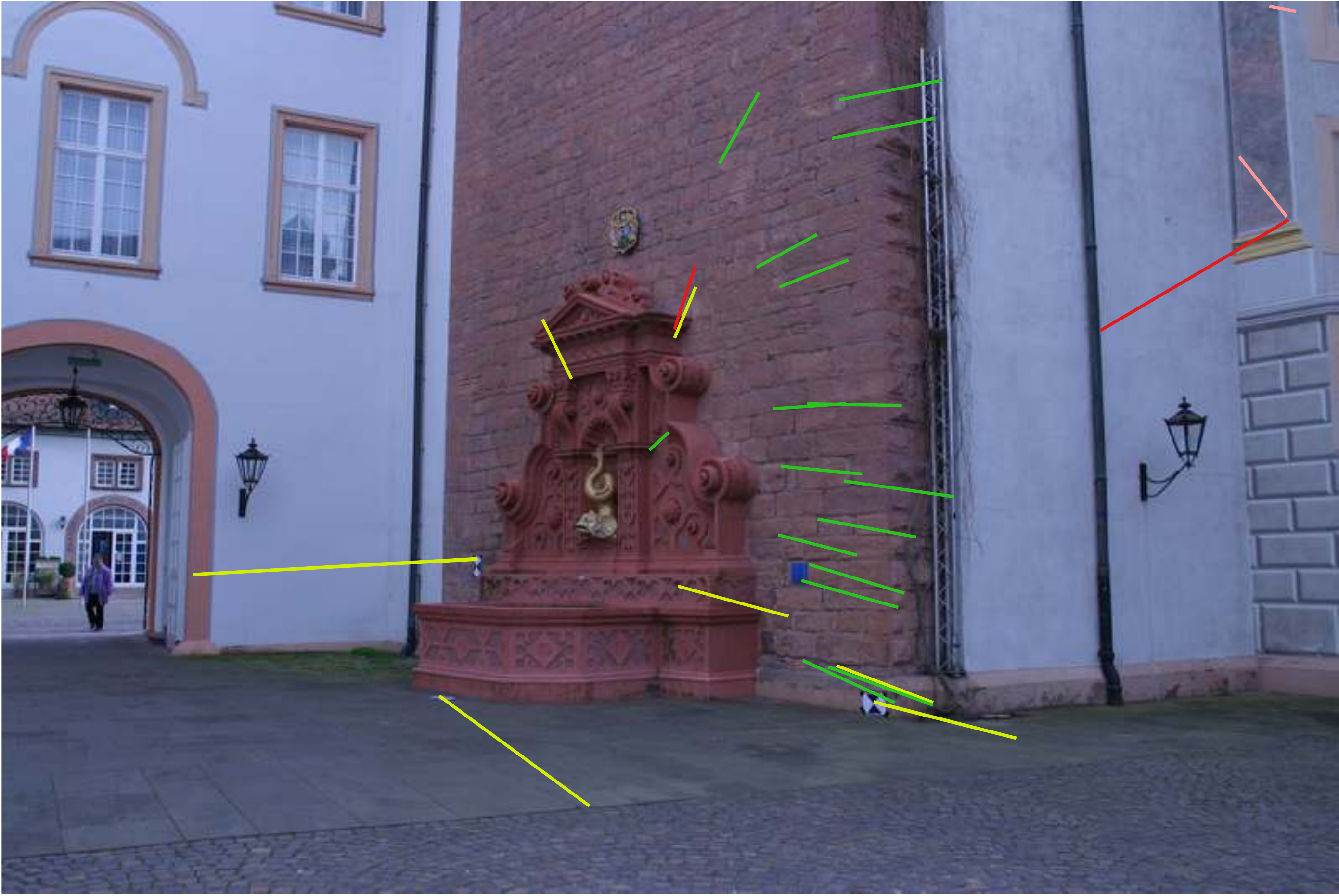}
	\includegraphics[height=7.5em]{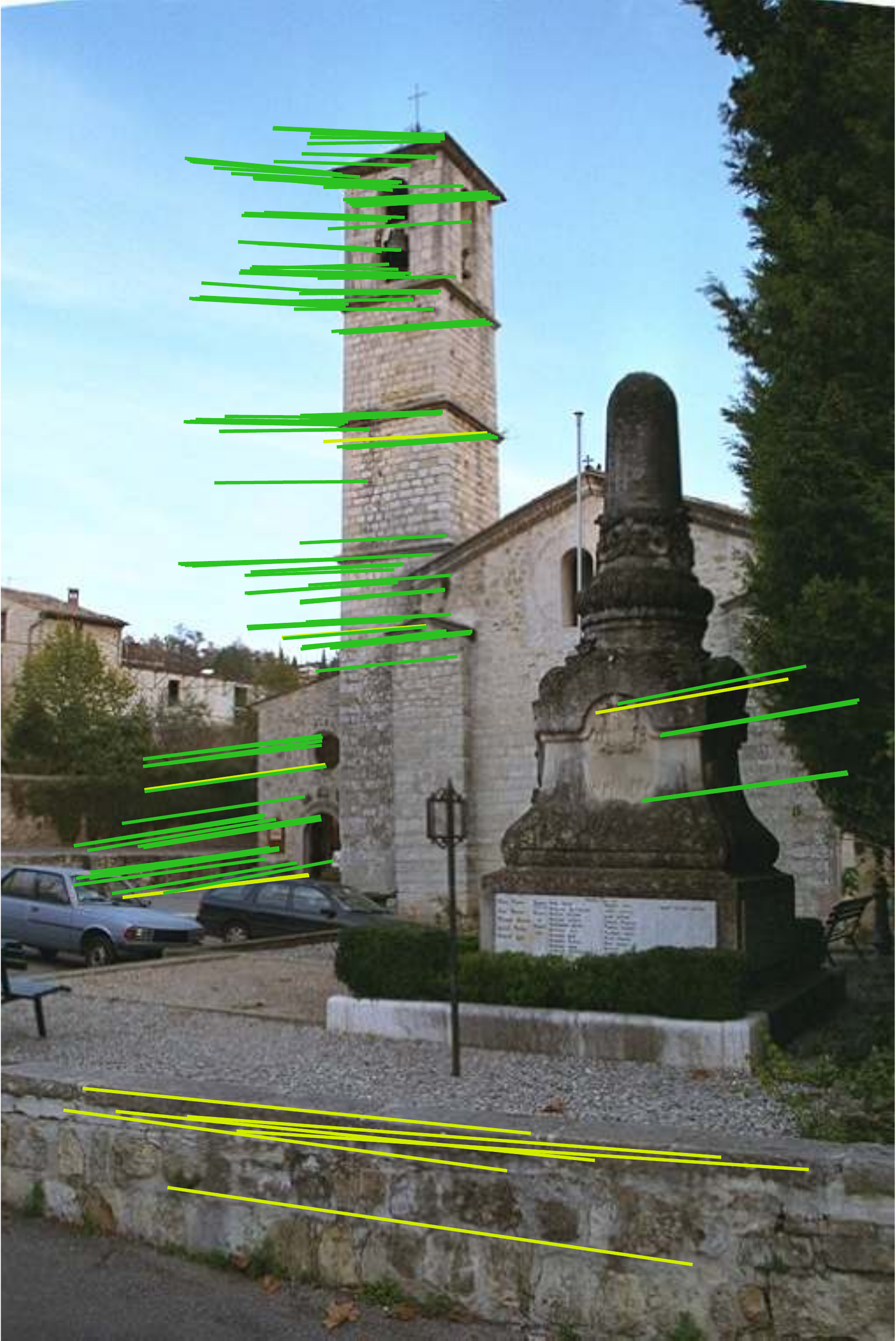}
	\includegraphics[height=7.5em]{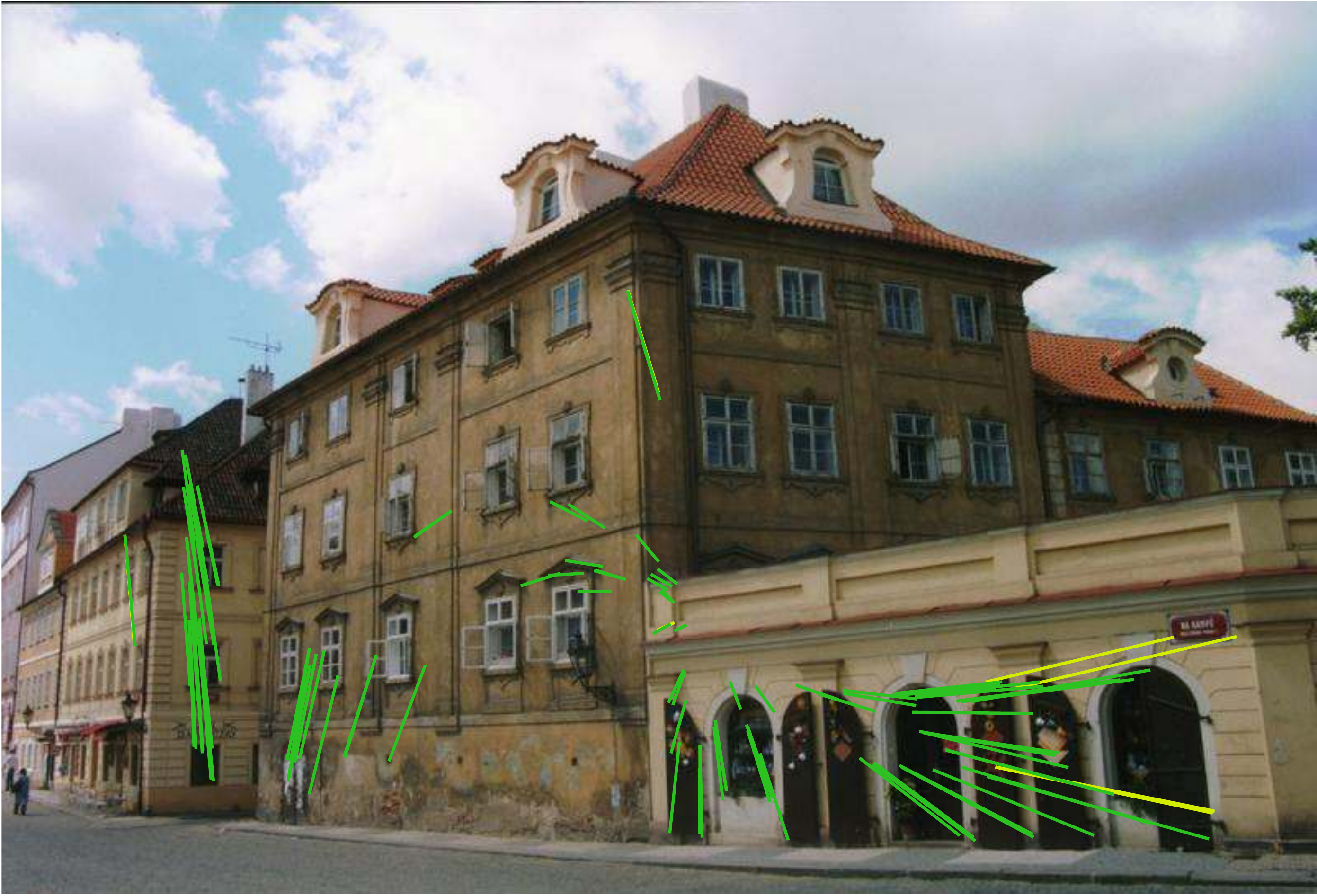}
	\hspace{0.05em}
	\includegraphics[height=7.5em]{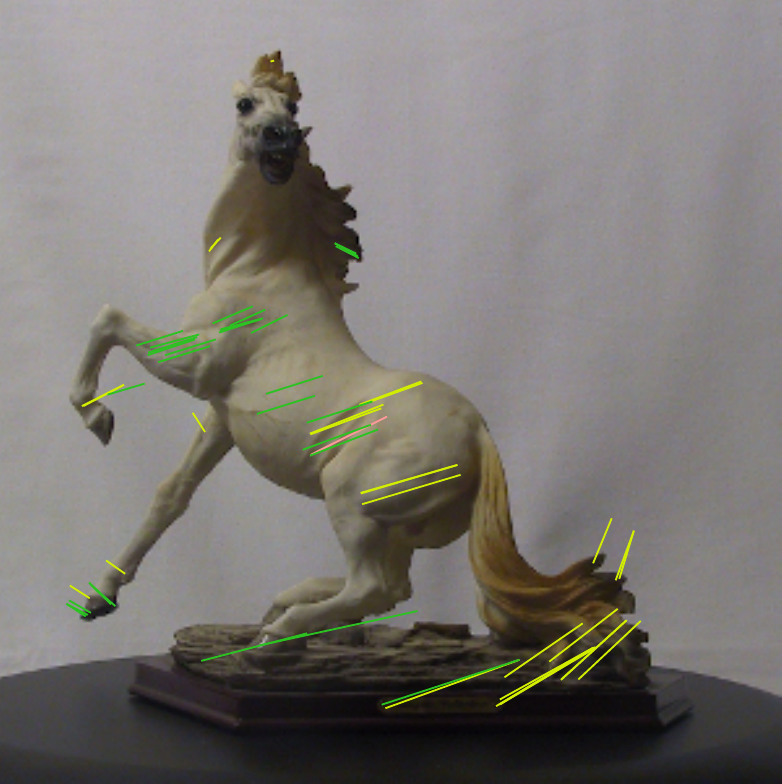}
	\hspace{0.05em}
	\includegraphics[height=7.5em]{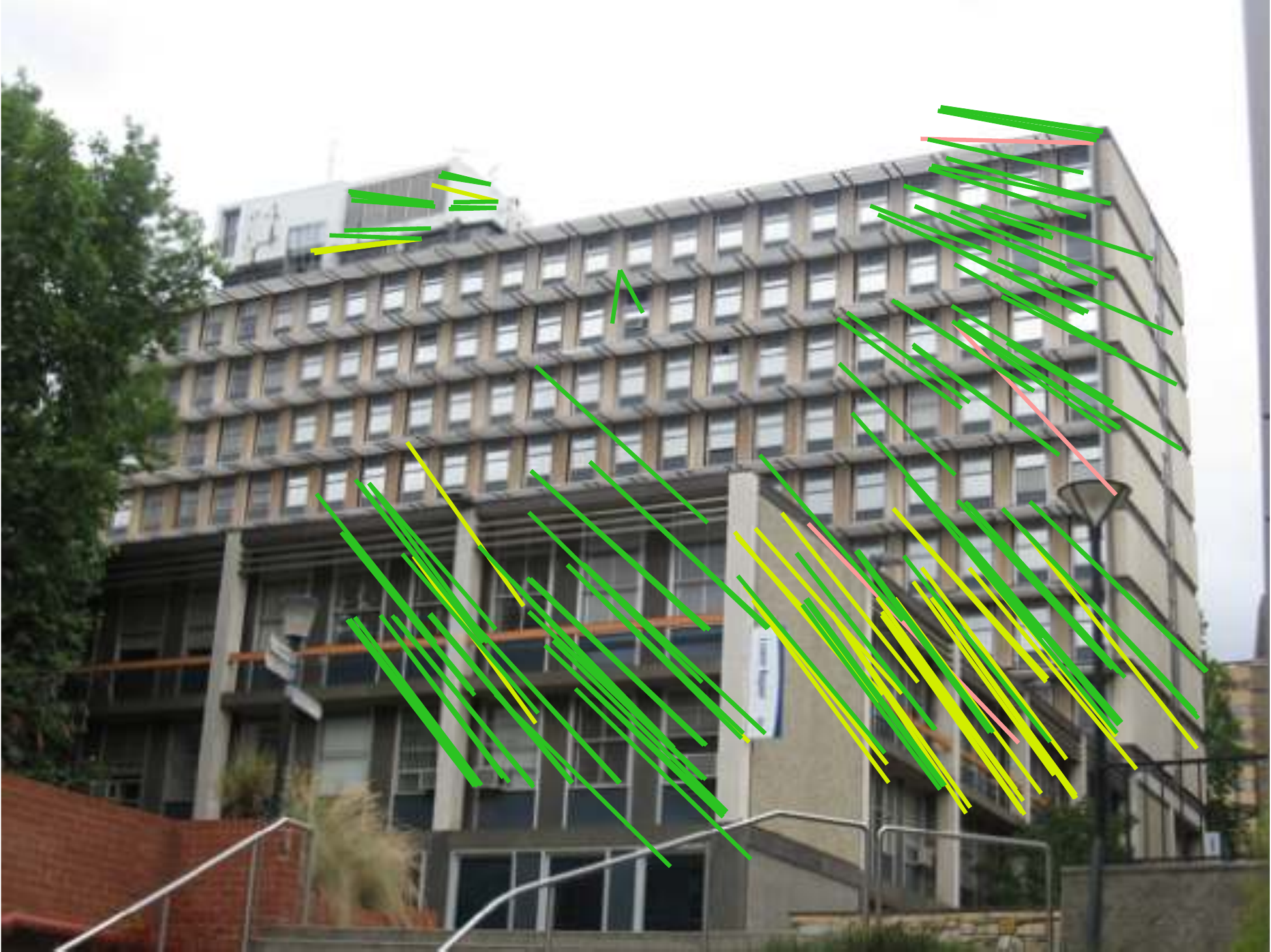}
	\\
	\vspace{0.5em}
	\rotatebox[origin=l]{90}{\mbox{\hspace{2em}PGM}}
	\includegraphics[height=7.5em]{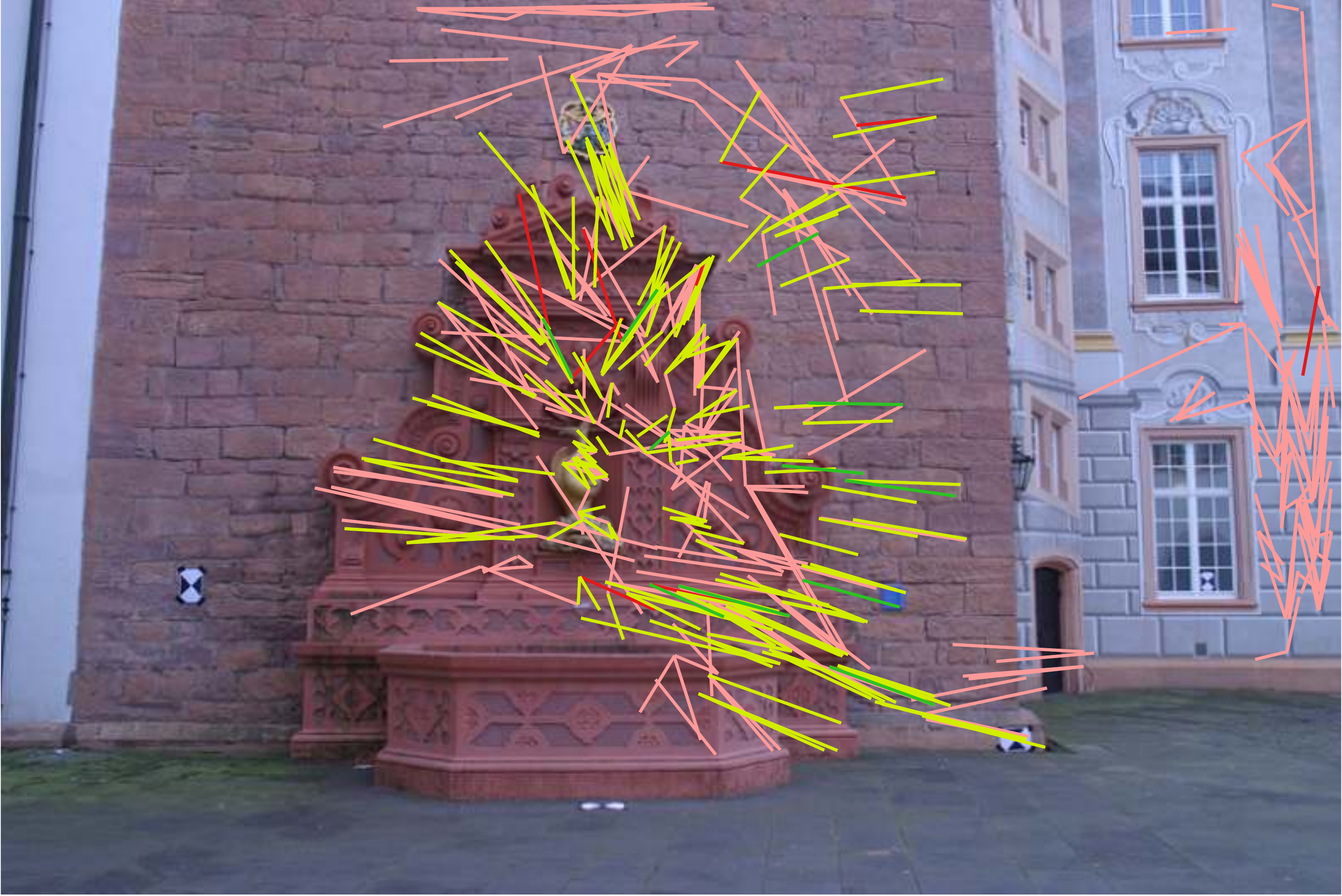}
	\includegraphics[height=7.5em]{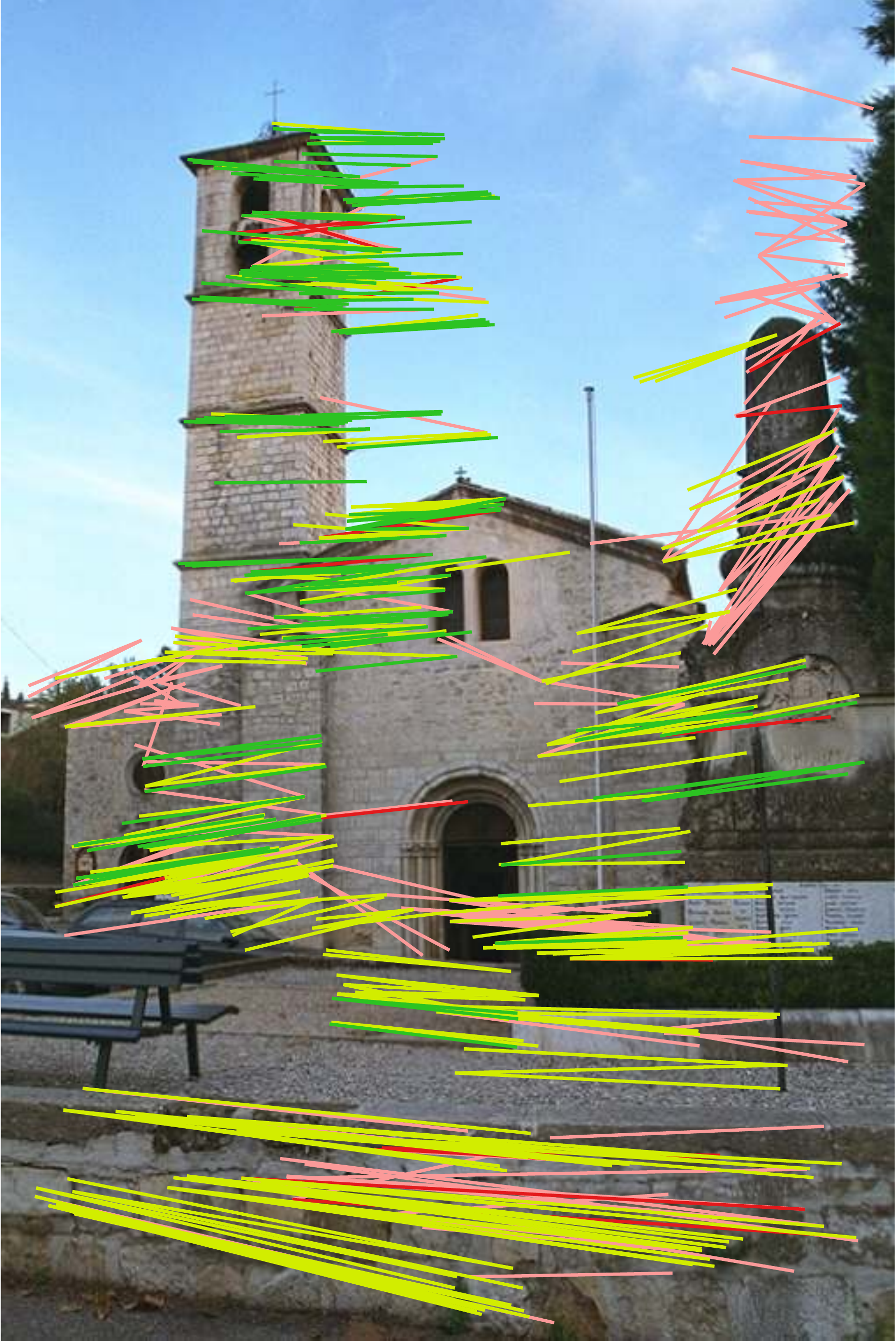}
	\includegraphics[height=7.5em]{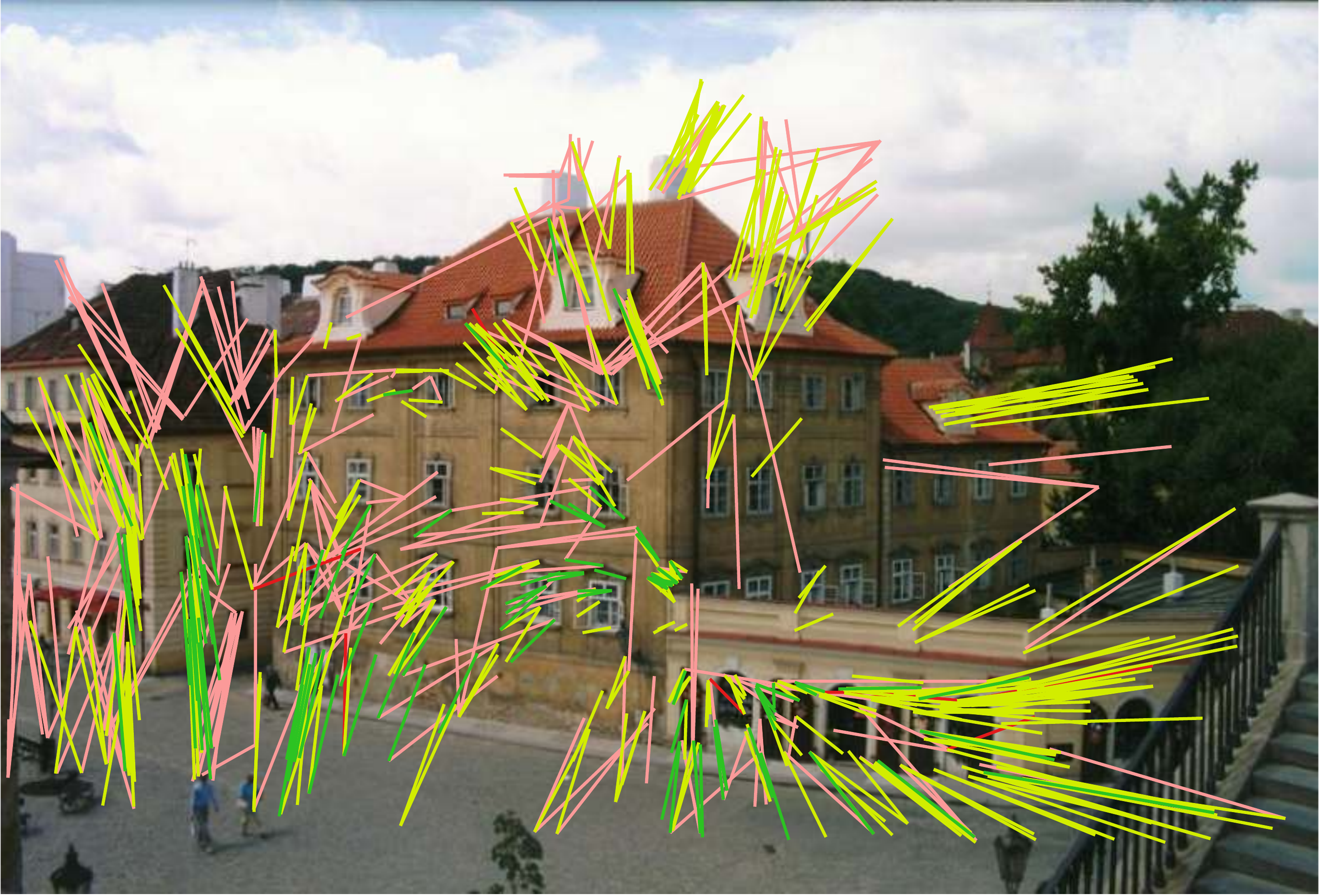}
	\includegraphics[height=7.5em]{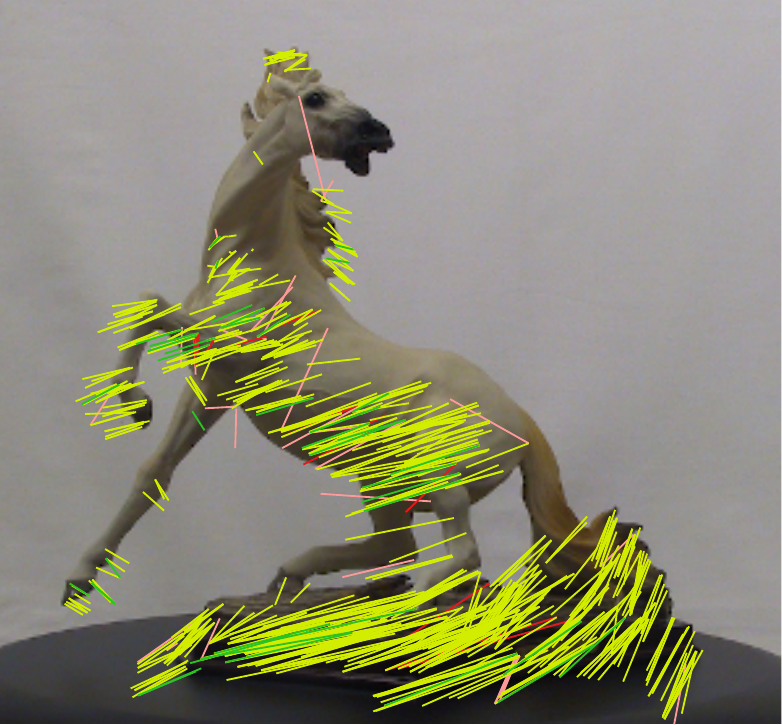}
	\includegraphics[height=7.5em]{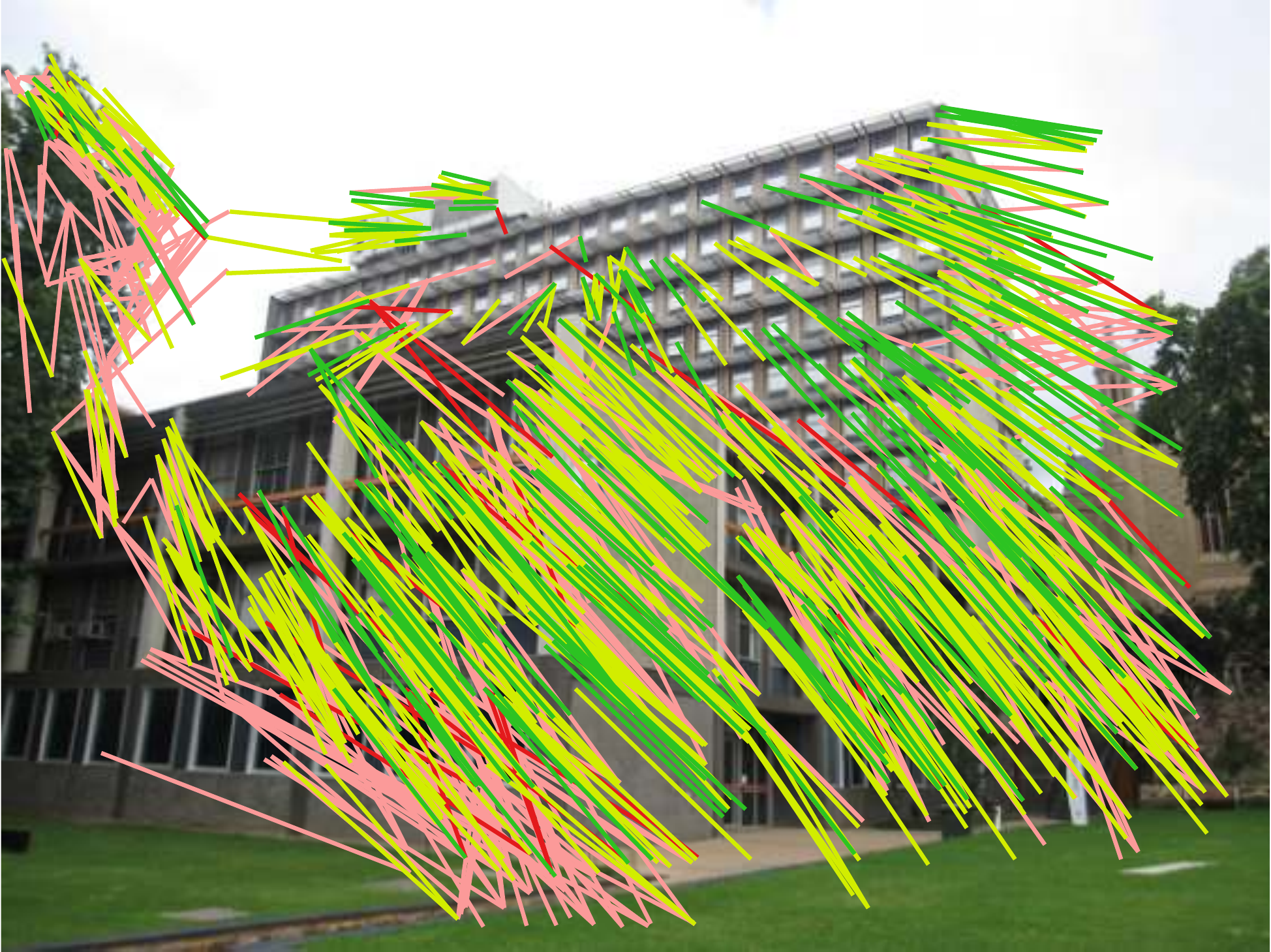}
	\\
	\vspace{0.5em}
	\rotatebox[origin=l]{90}{\mbox{\hspace{2em}SCV}}
	\includegraphics[height=7.5em]{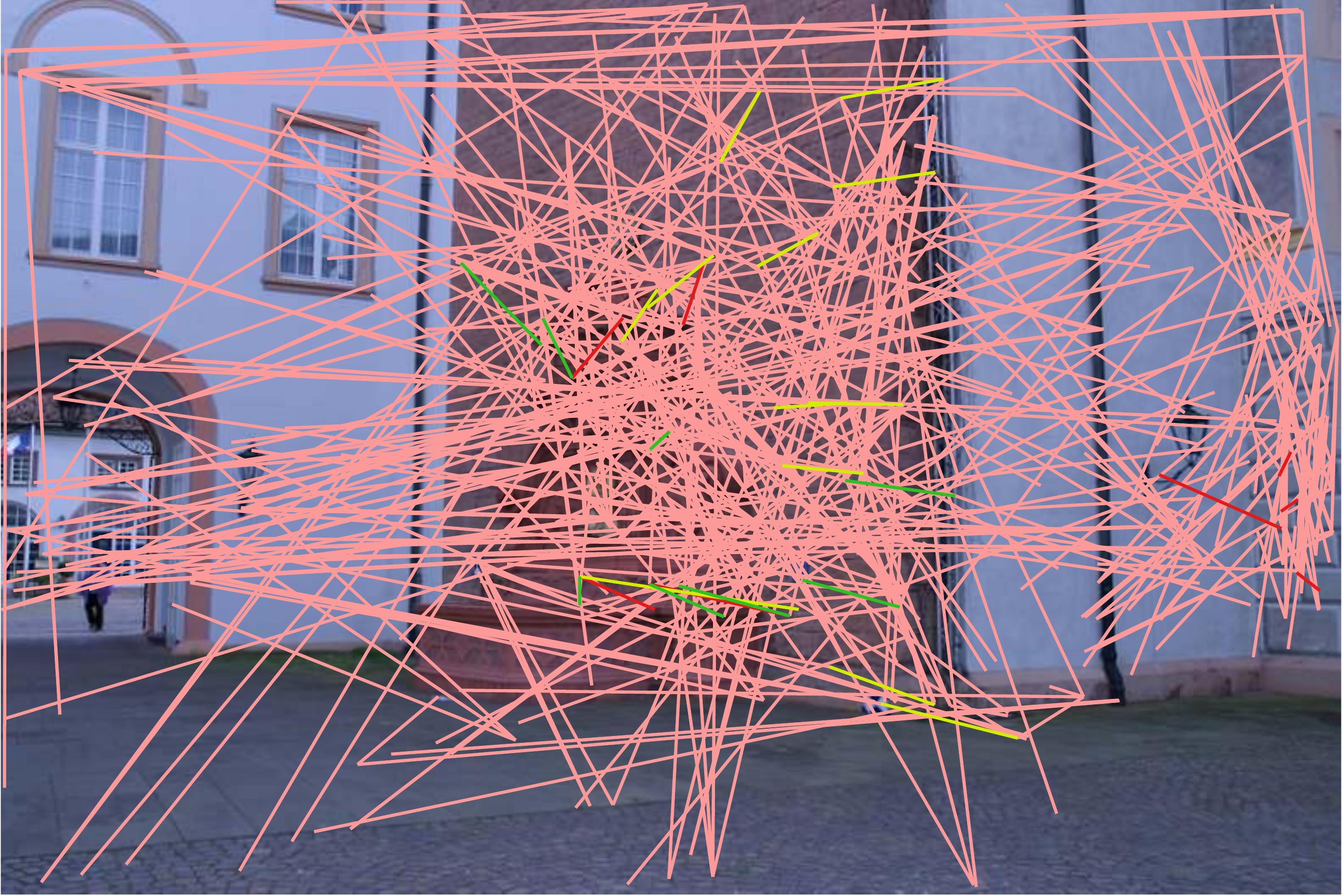}
	\includegraphics[height=7.5em]{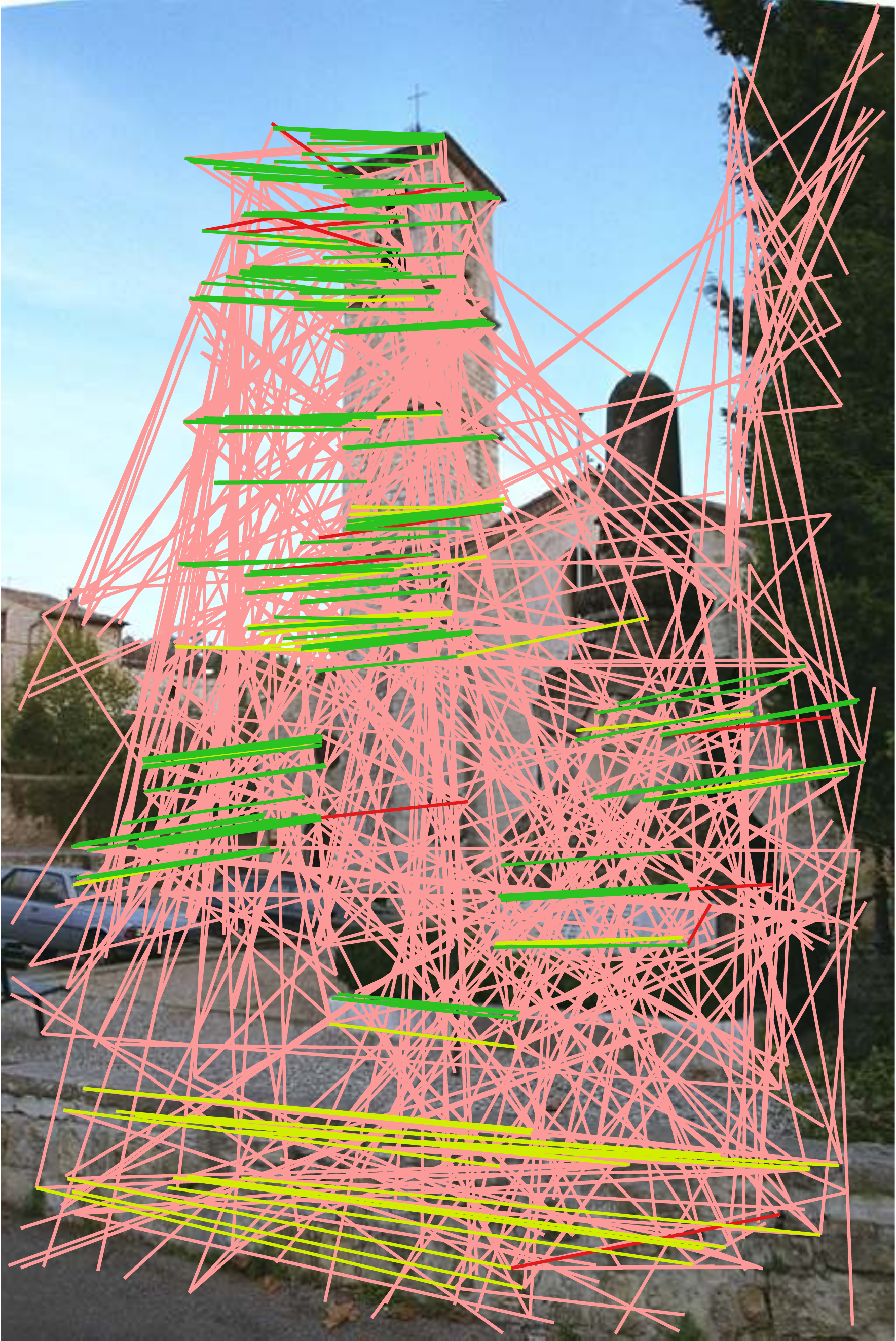}
	\includegraphics[height=7.5em]{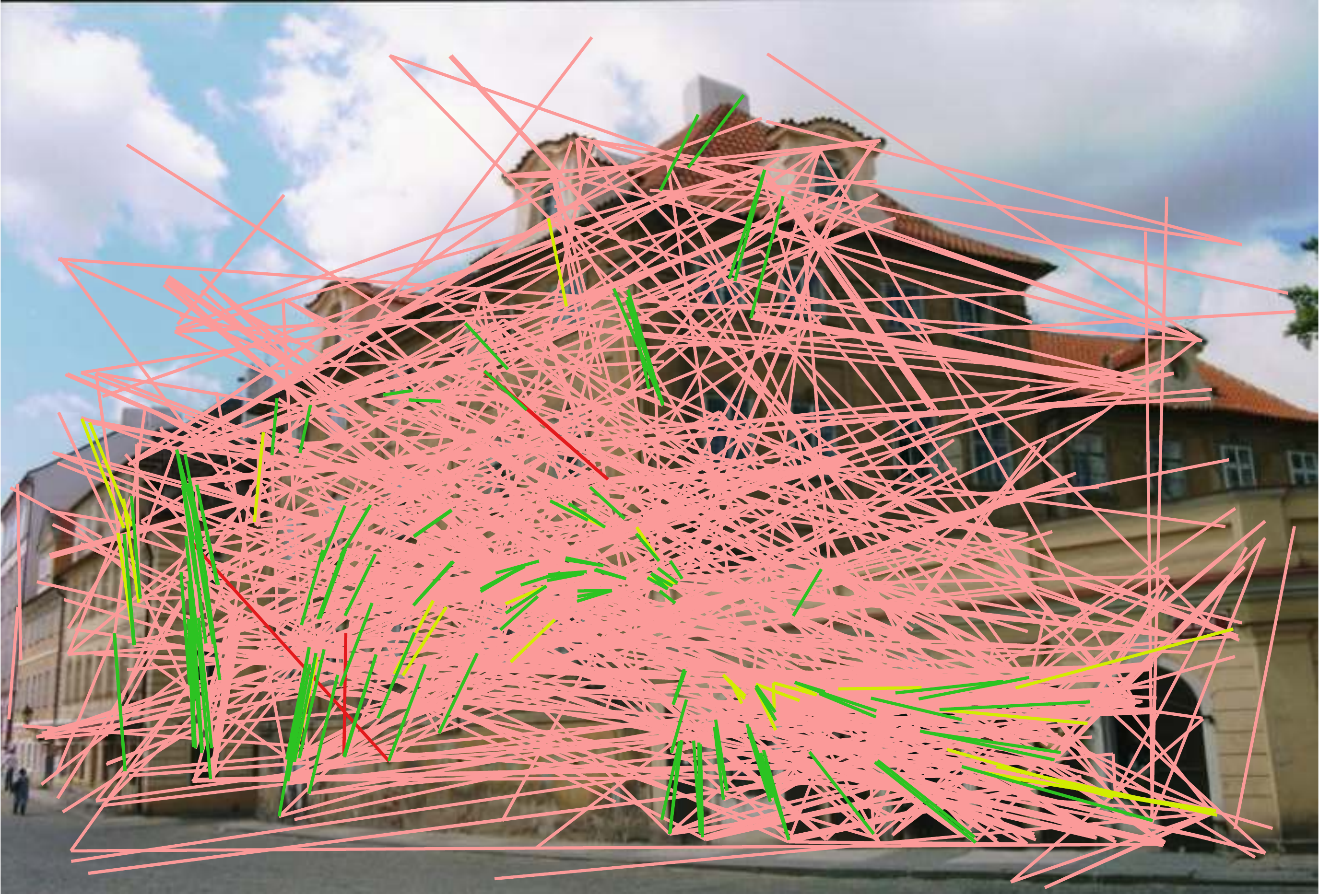}
	\hspace{0.05em}
	\includegraphics[height=7.5em]{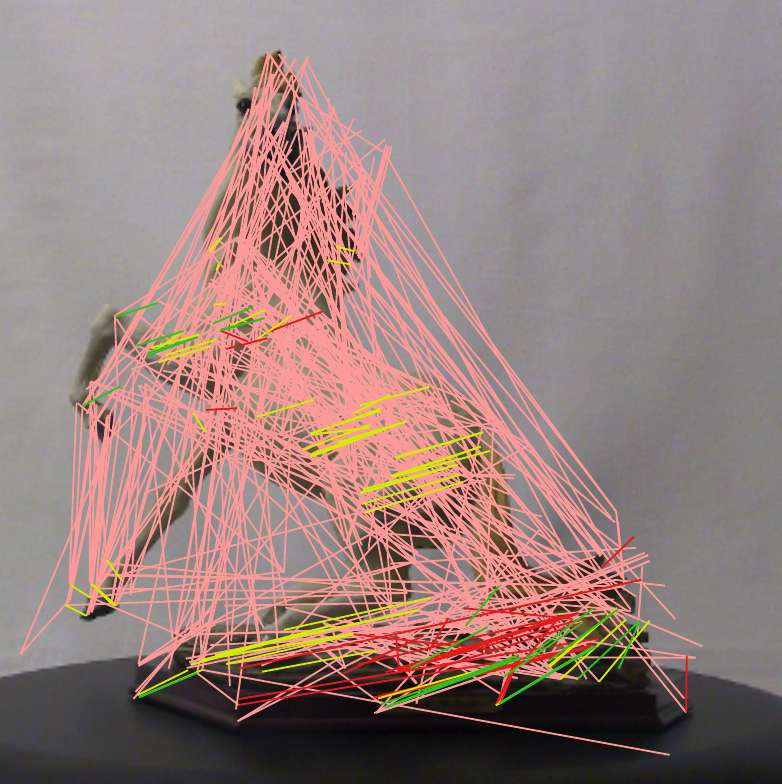}
	\hspace{0.05em}
	\includegraphics[height=7.5em]{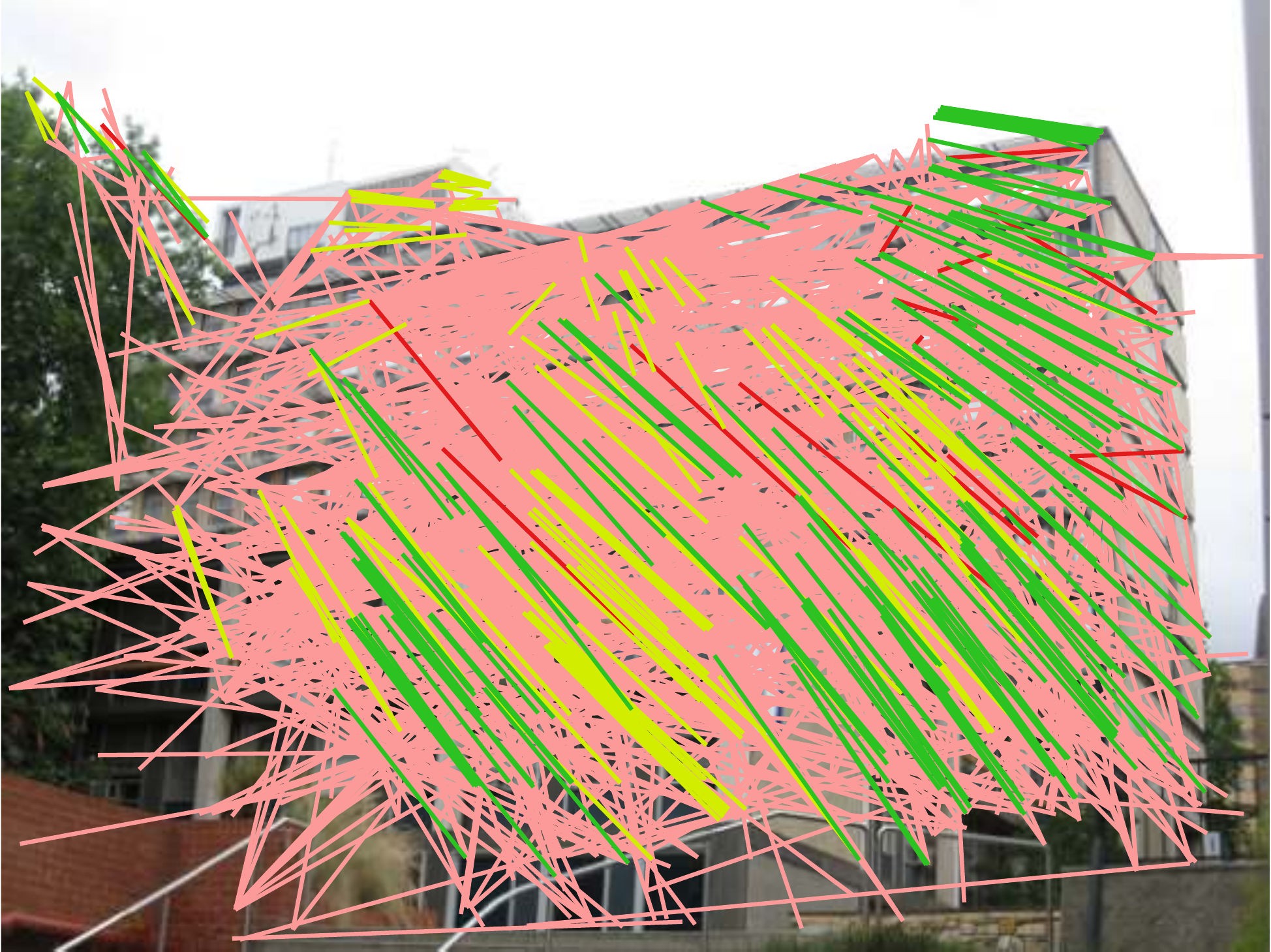}
	\\
	\vspace{0.5em}
	\rotatebox[origin=l]{90}{\mbox{\hspace{2em}BM}}
	\includegraphics[height=7.5em]{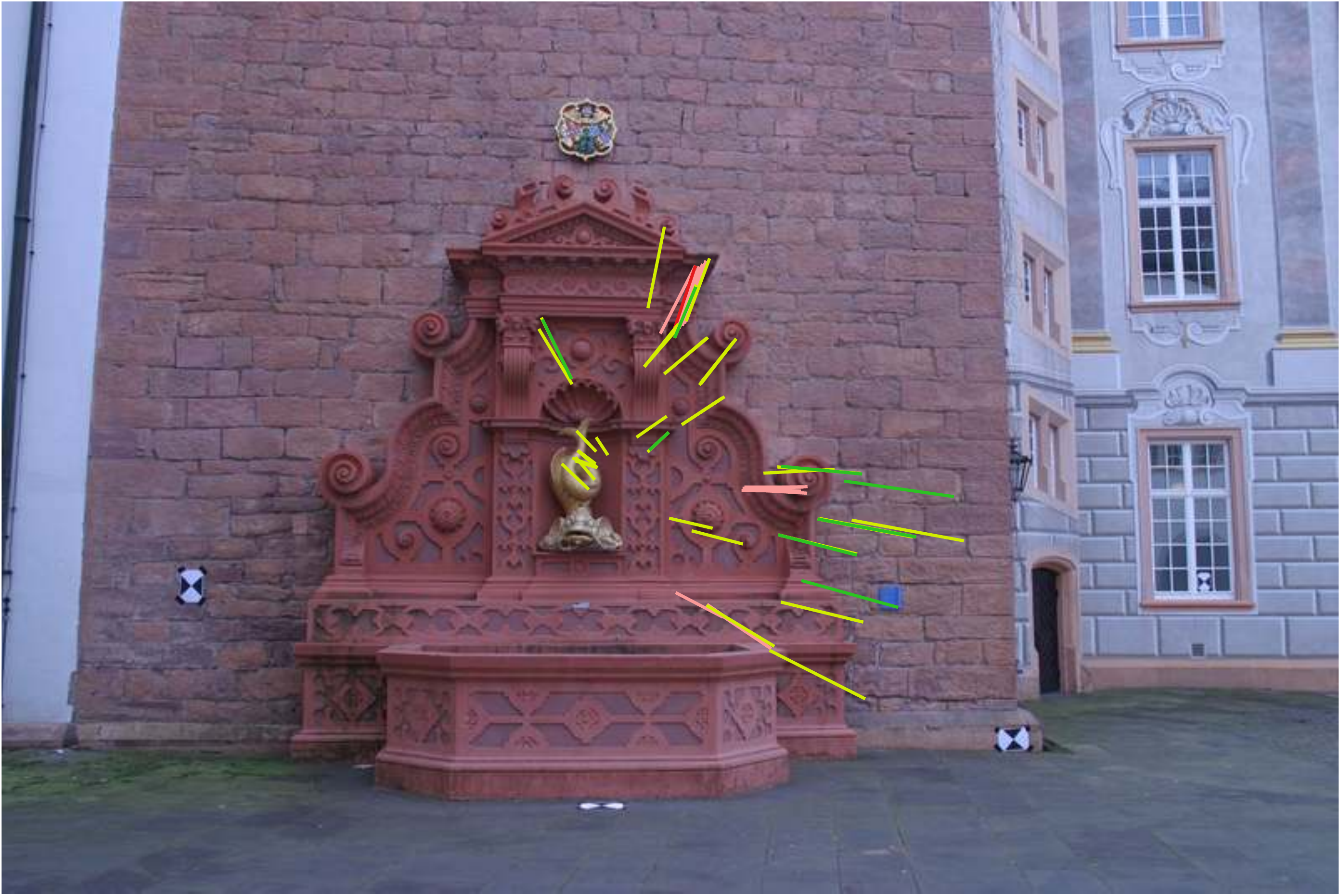}
	\includegraphics[height=7.5em]{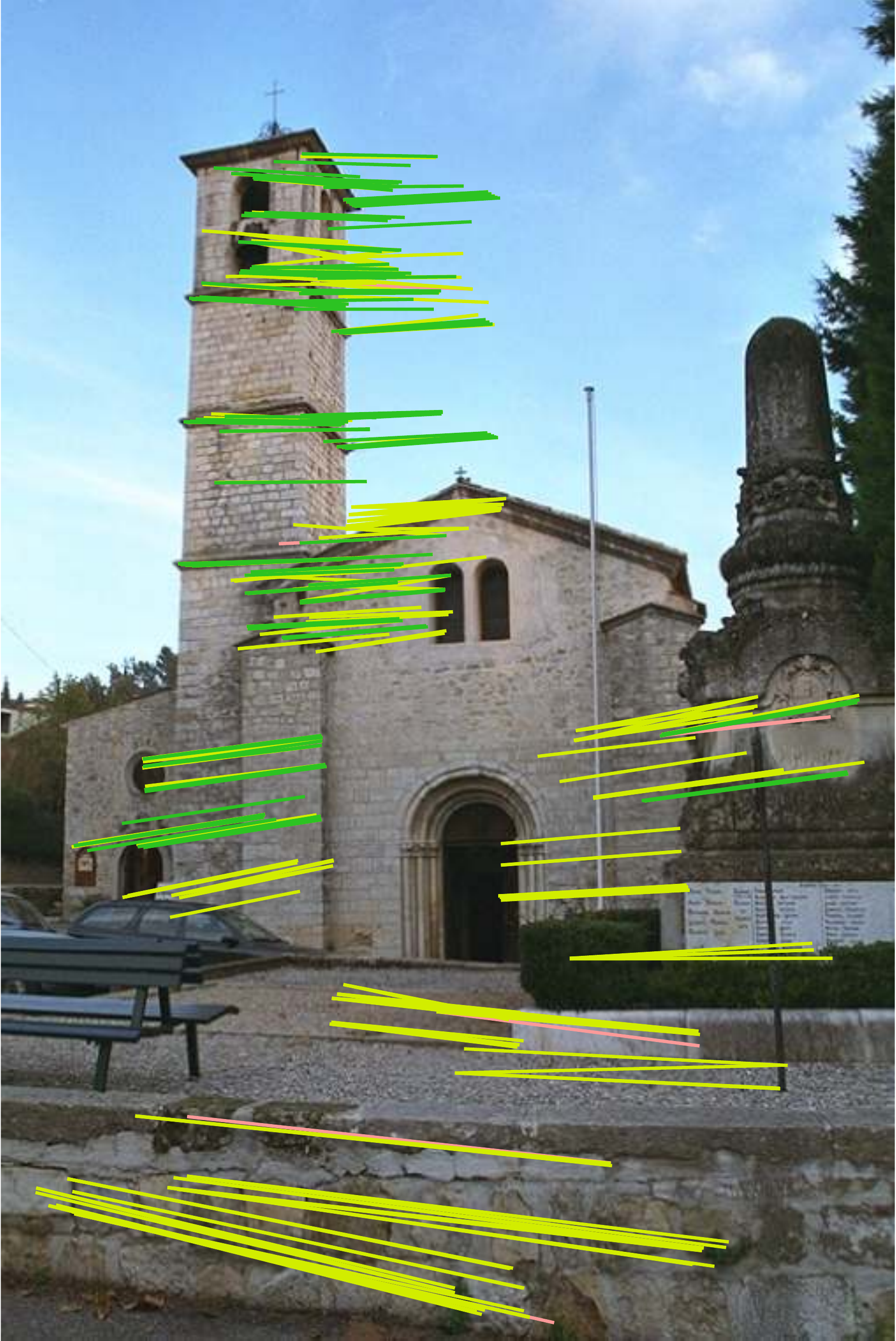}
	\includegraphics[height=7.5em]{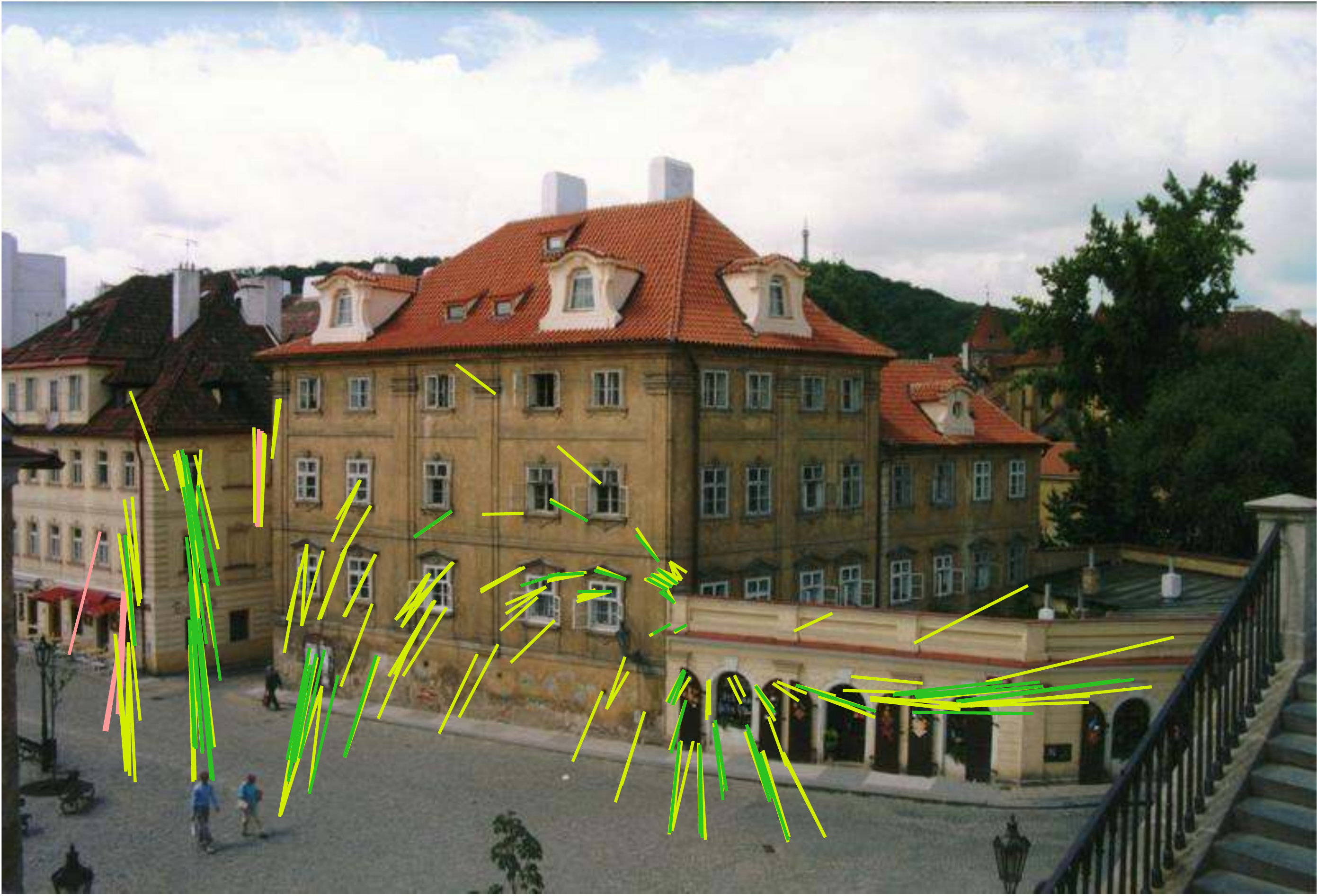}
	\includegraphics[height=7.5em]{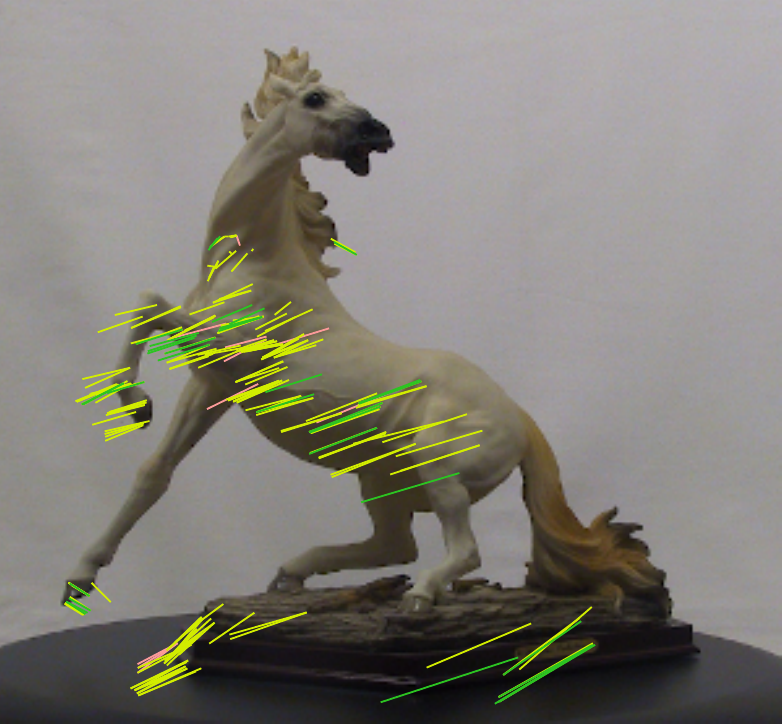}
	\includegraphics[height=7.5em]{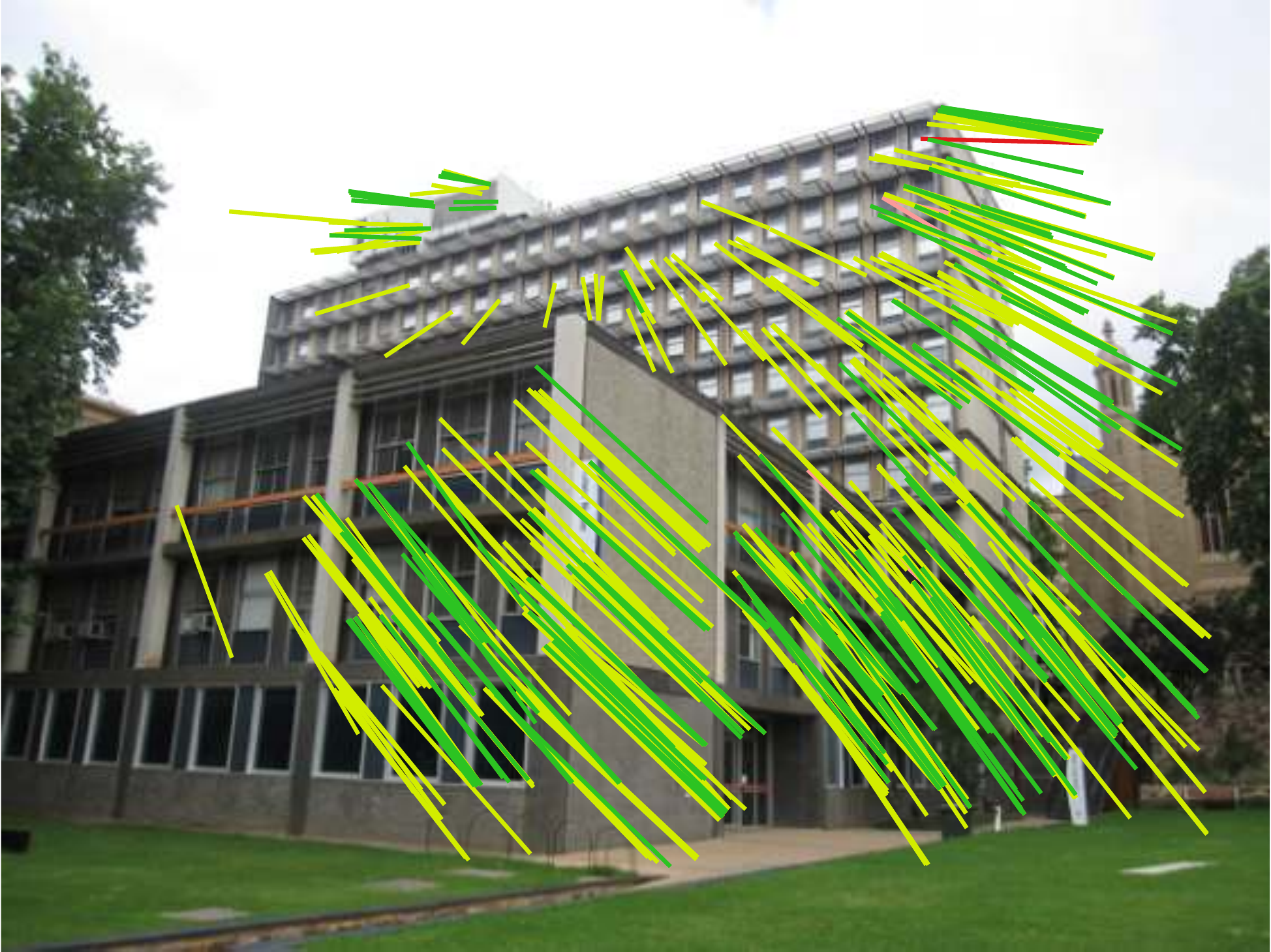}
	\\
	\begin{flushleft}
		\hspace*{7.5em}Fountain01\hspace{3.75em}Valbonne\hspace{4.5em}Kampa\hspace{7em}Horse\hspace{5.75em}NapierB
	\end{flushleft}
	\caption{\label{example_2b}
		Planar and non-planar local spatial filter matches according to the best configuration setup, the images of the input pair alternate among the rows. Image indexes are reported as suffix when the sequence contains more than two images. For each method inlier (yellow, green) and outlier (red and light red) clusters are shown, as well as the 1SAC filtered matches (green, red) (see Sec.~\ref{eval_dt}, best viewed in color and zoomed in).}
\end{figure*}

\begin{figure*}
	\center
	\rotatebox[origin=l]{90}{\mbox{\hspace{3em}th}}
	\includegraphics[height=7.5em]{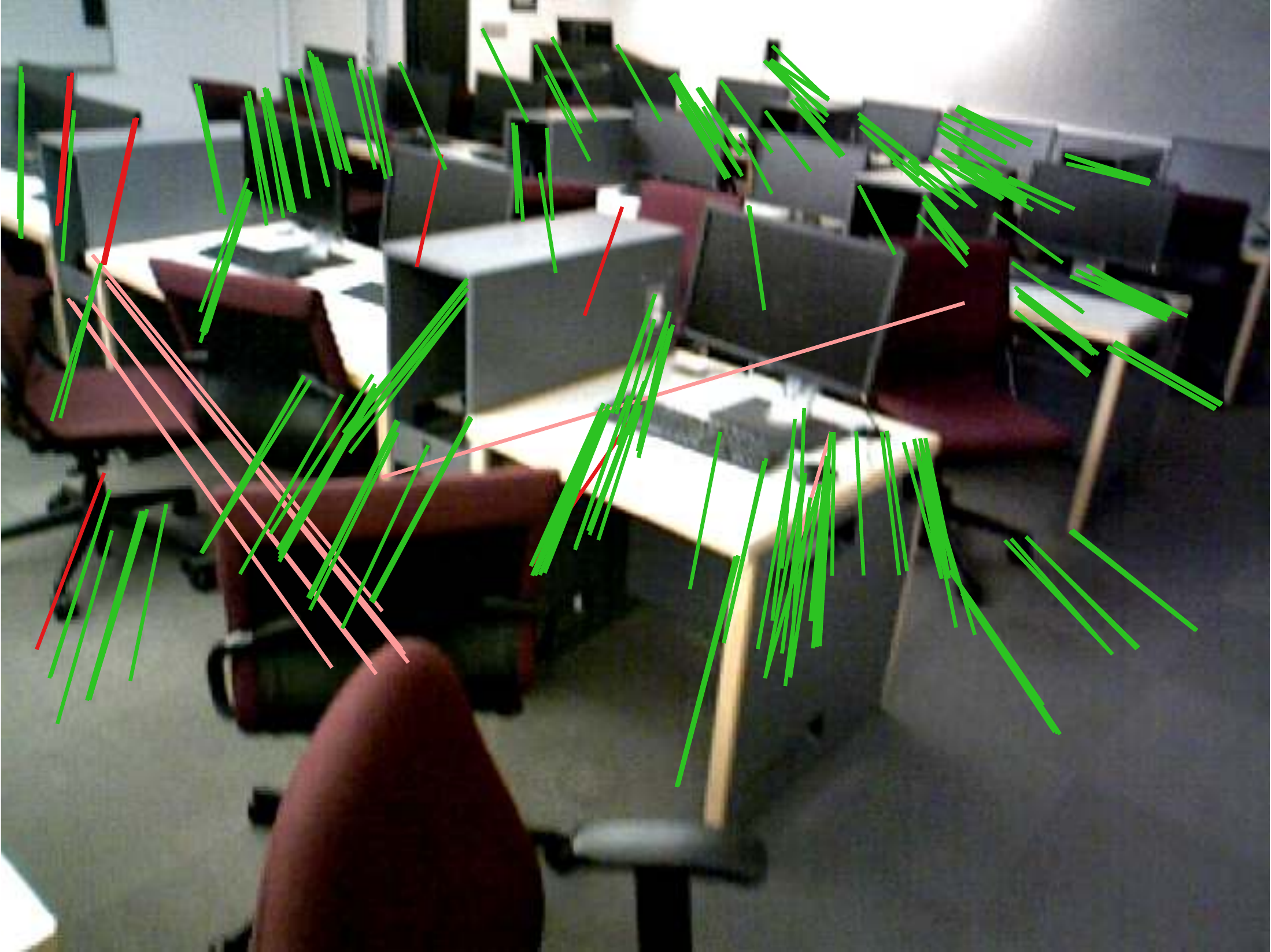}
	\includegraphics[height=7.5em]{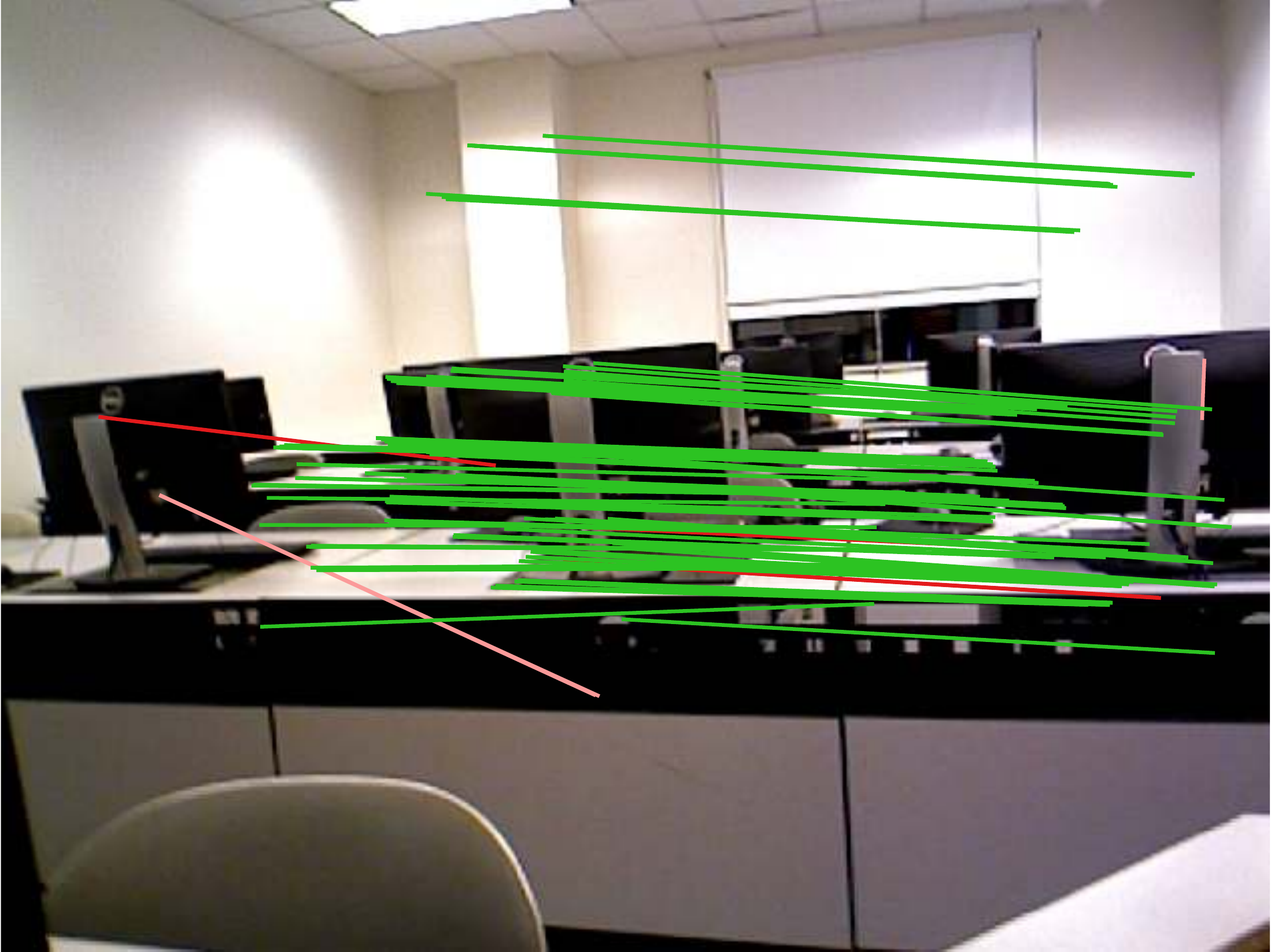}
	\includegraphics[height=7.5em]{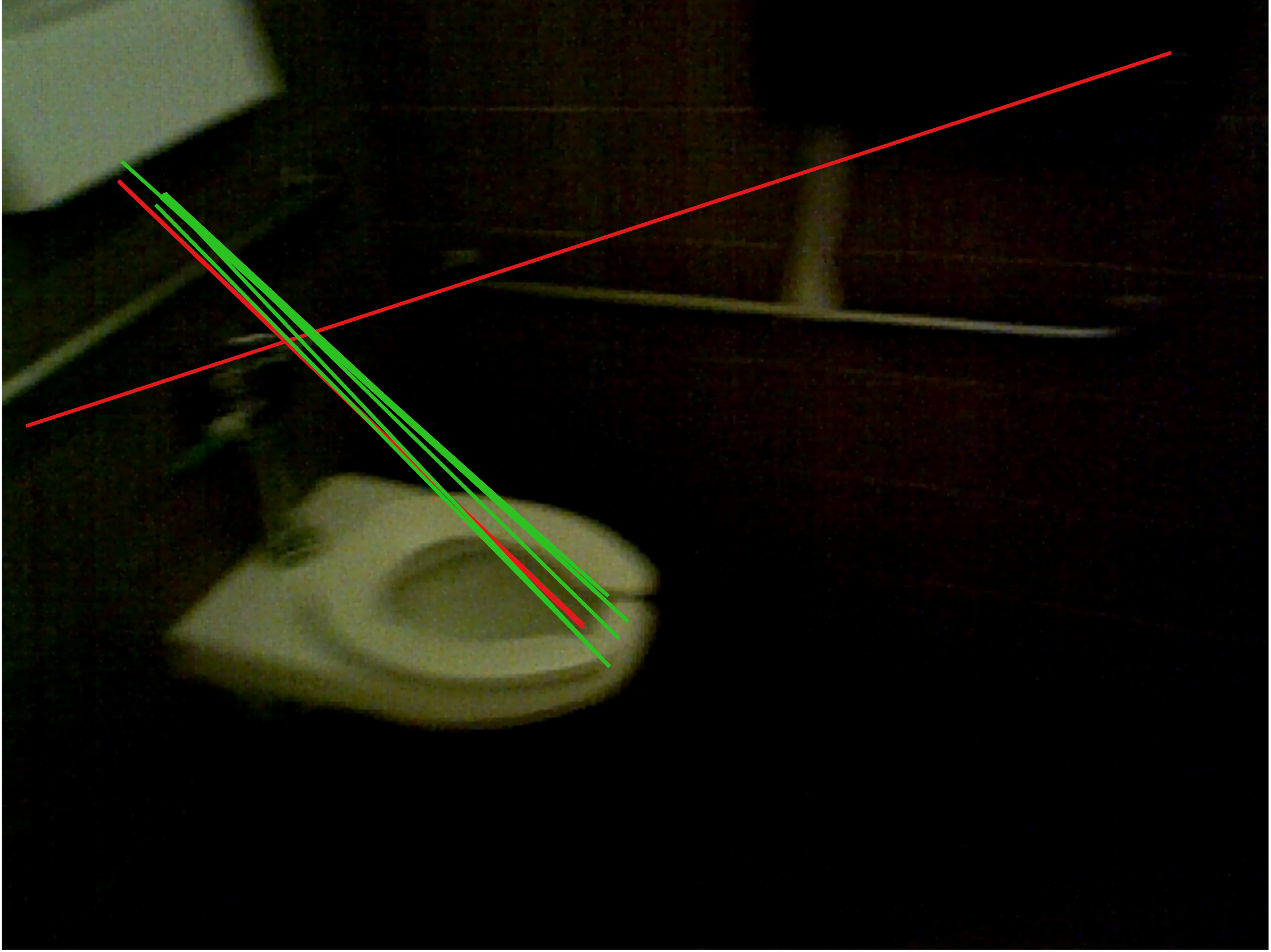}
	\includegraphics[height=7.5em]{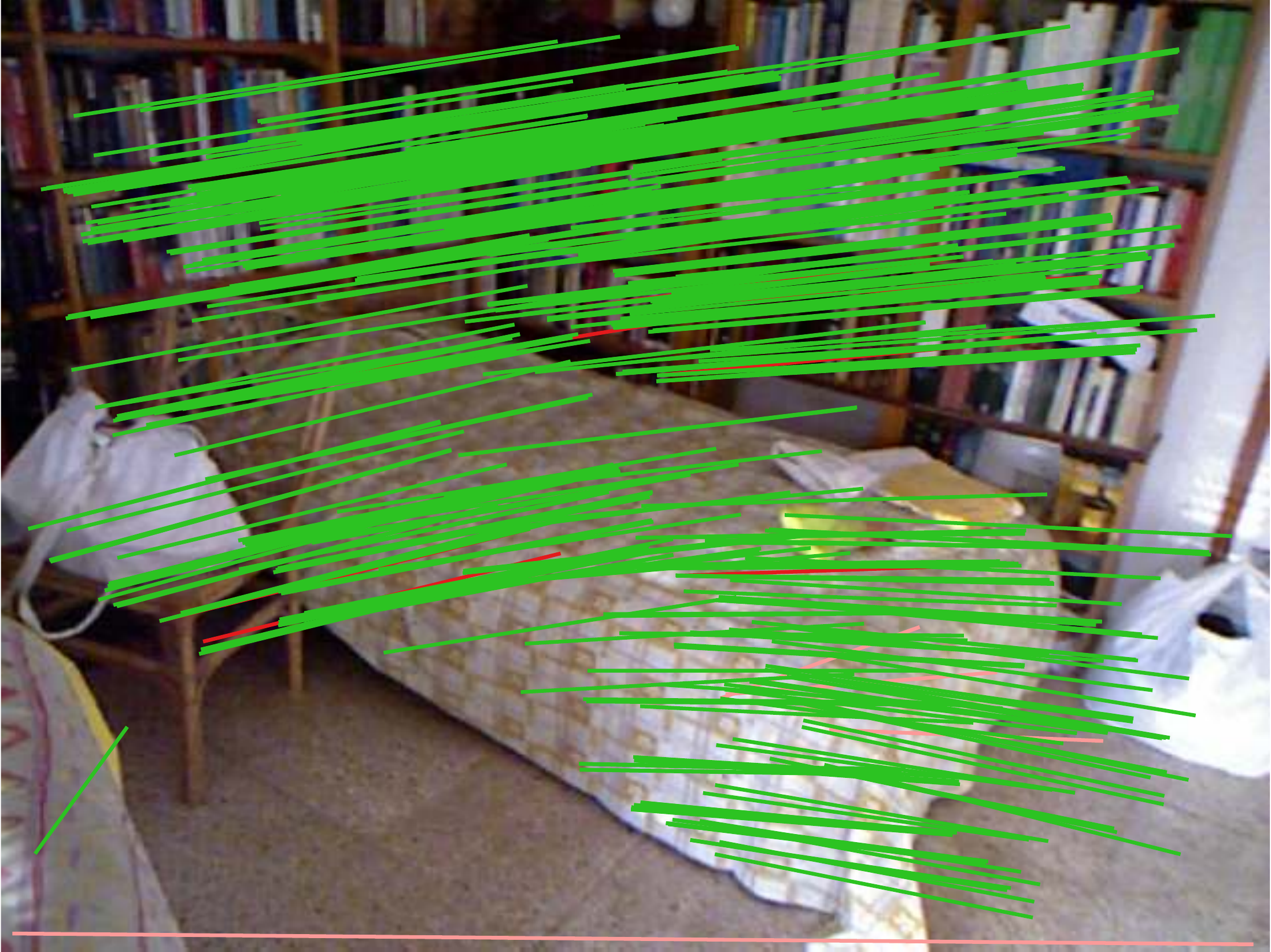}
	\includegraphics[height=7.5em]{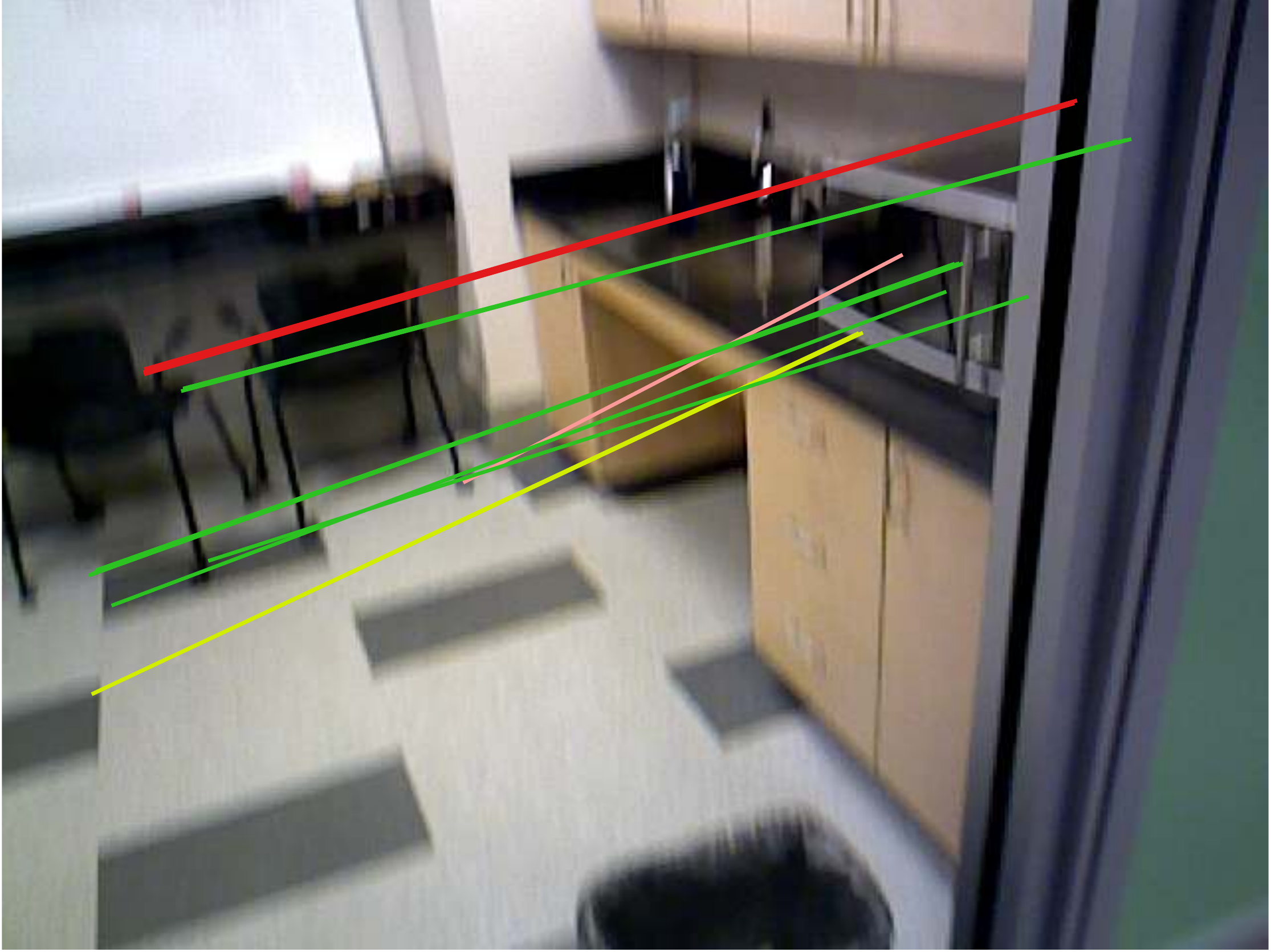}
	\\
	\vspace{0.5em}
	\rotatebox[origin=l]{90}{\mbox{\hspace{2em}DTM}}
	\includegraphics[height=7.5em]{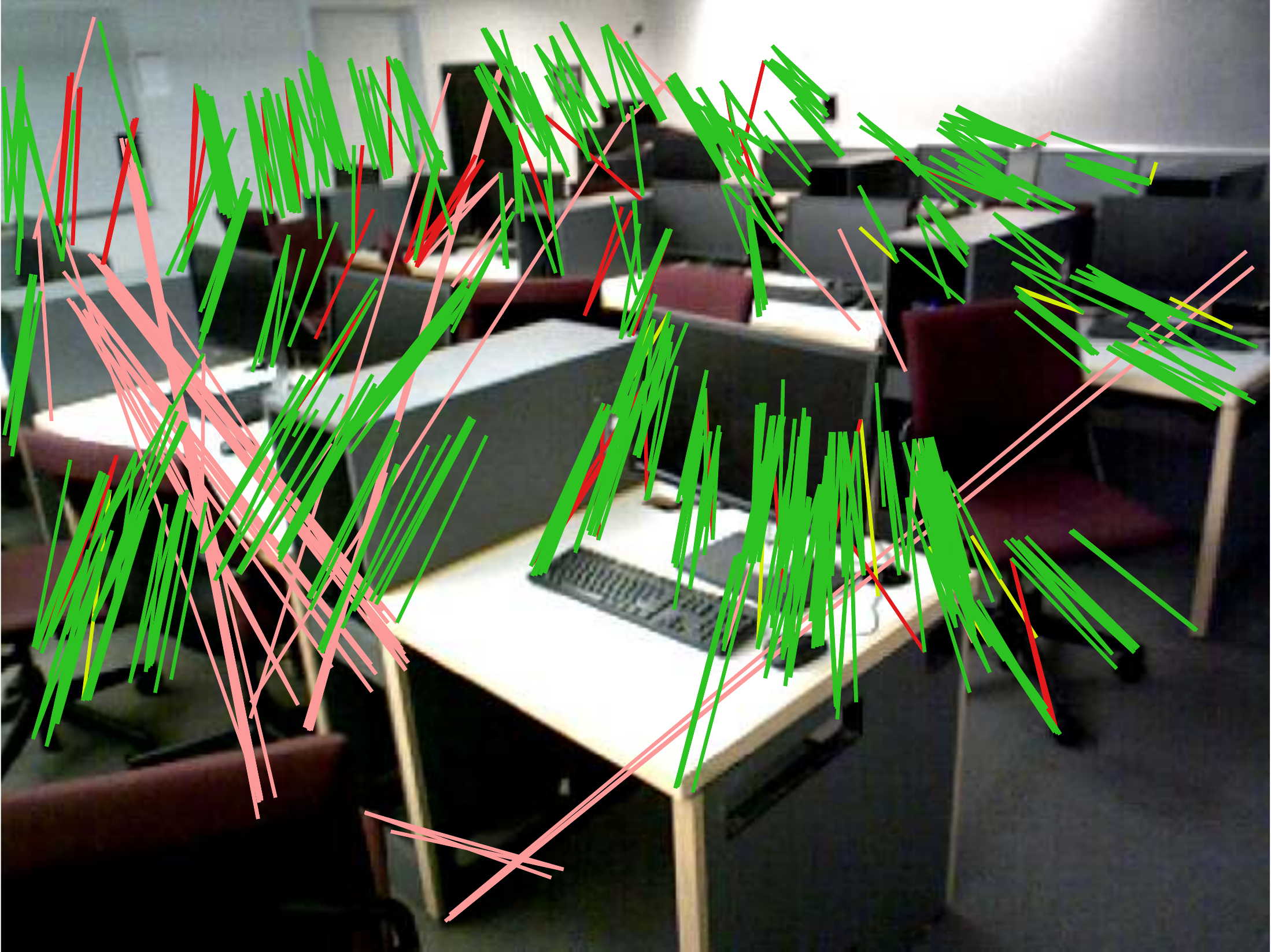}
	\includegraphics[height=7.5em]{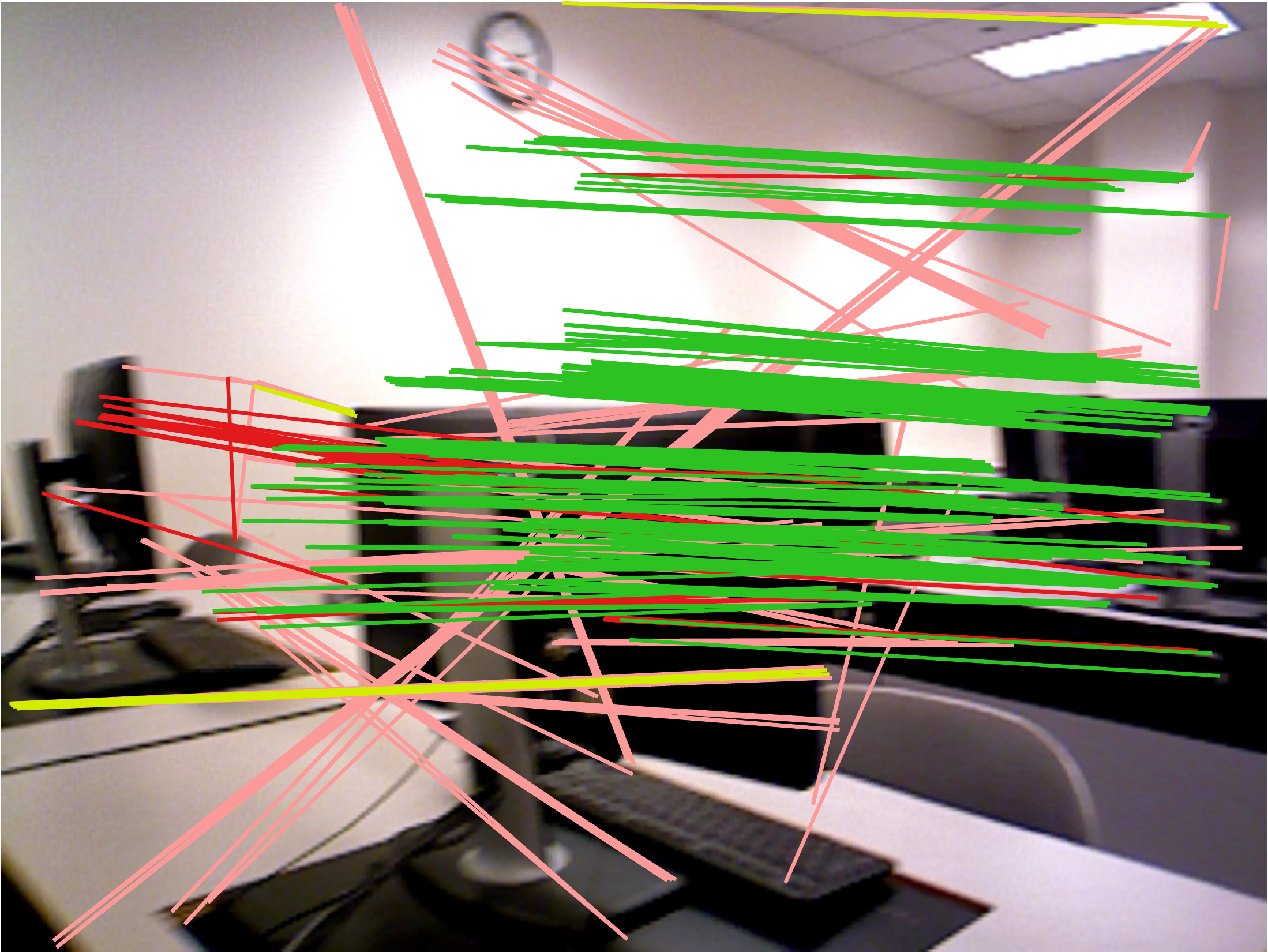}
	\includegraphics[height=7.5em]{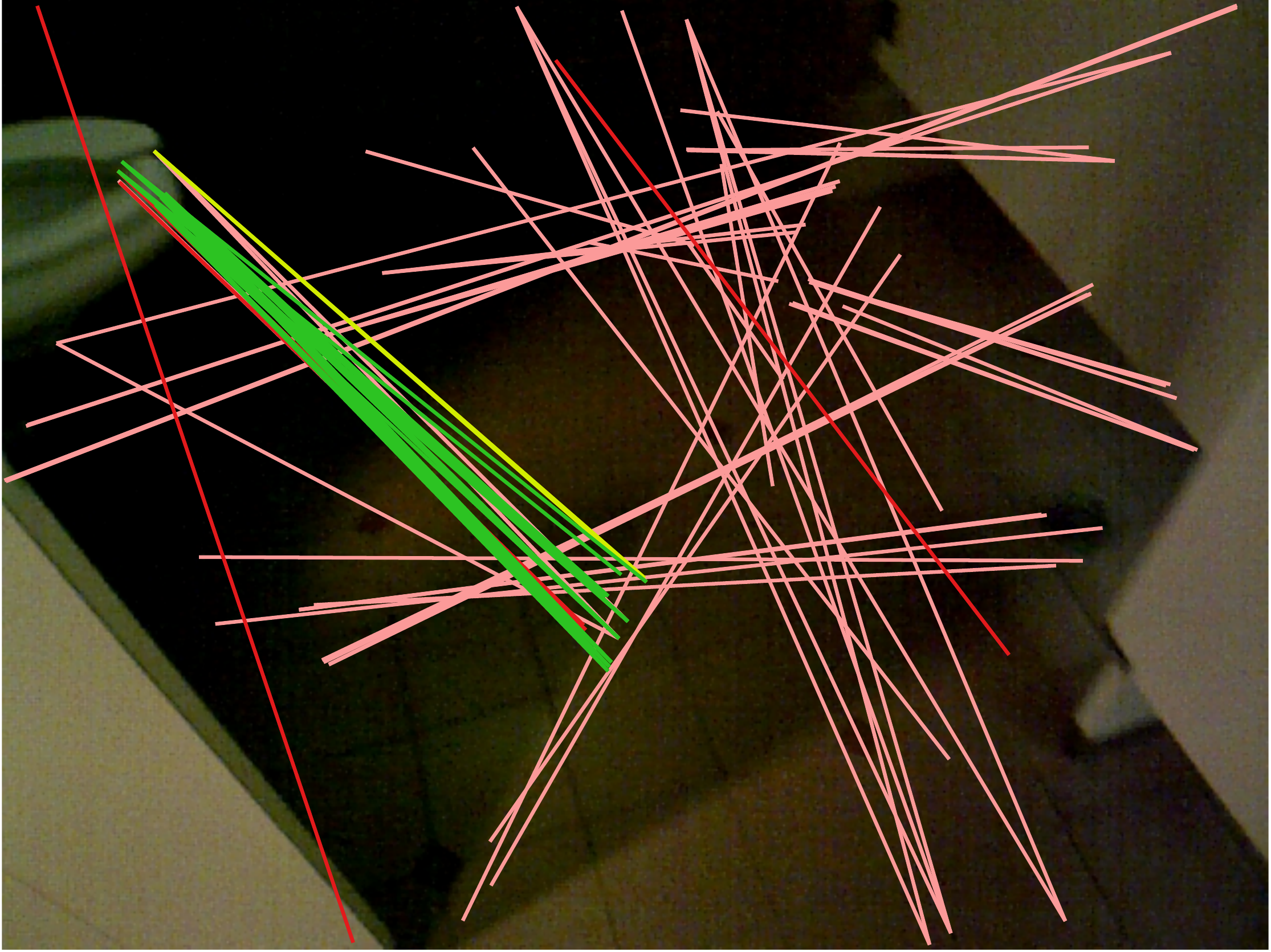}
	\includegraphics[height=7.5em]{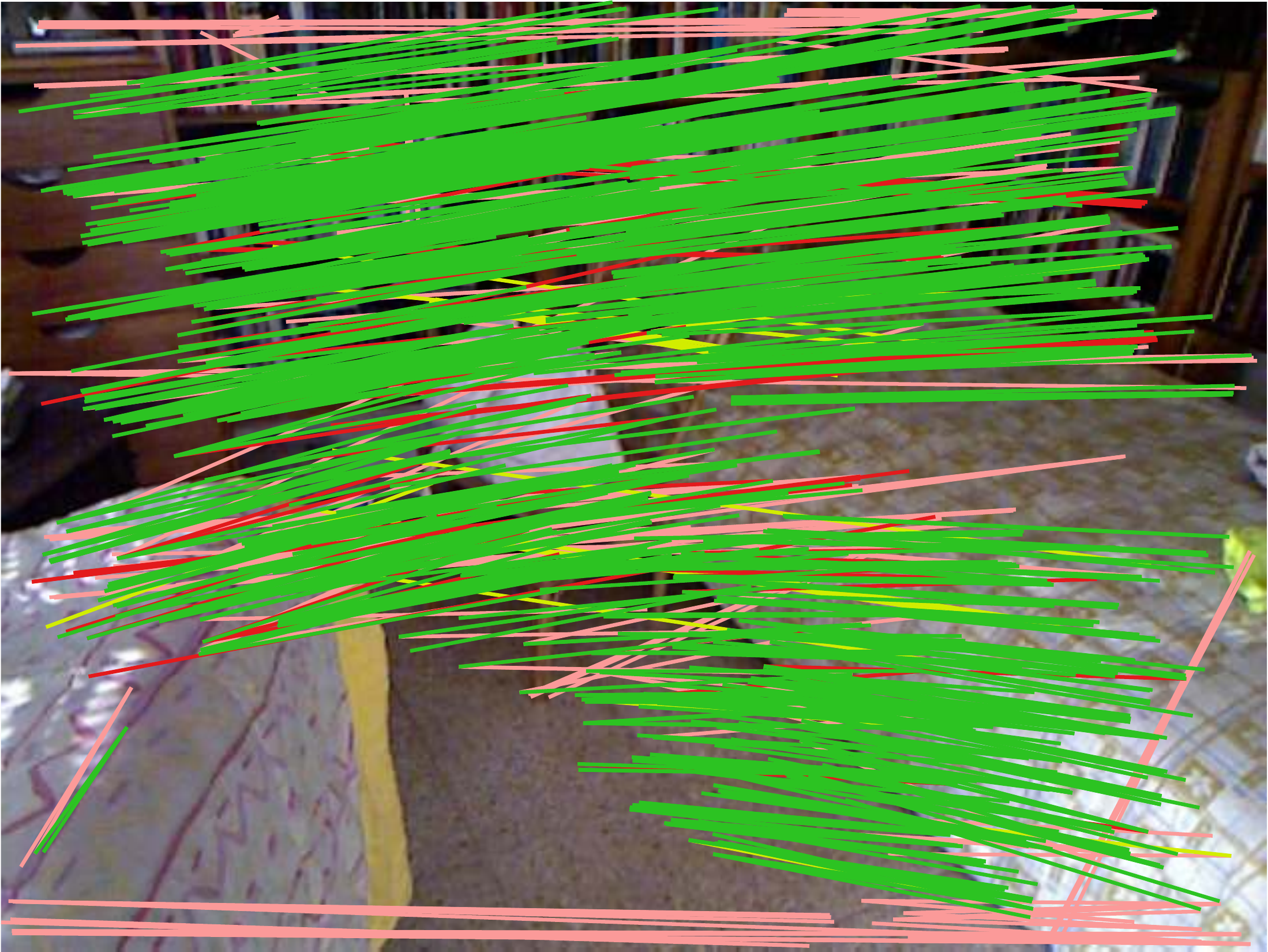}
	\includegraphics[height=7.5em]{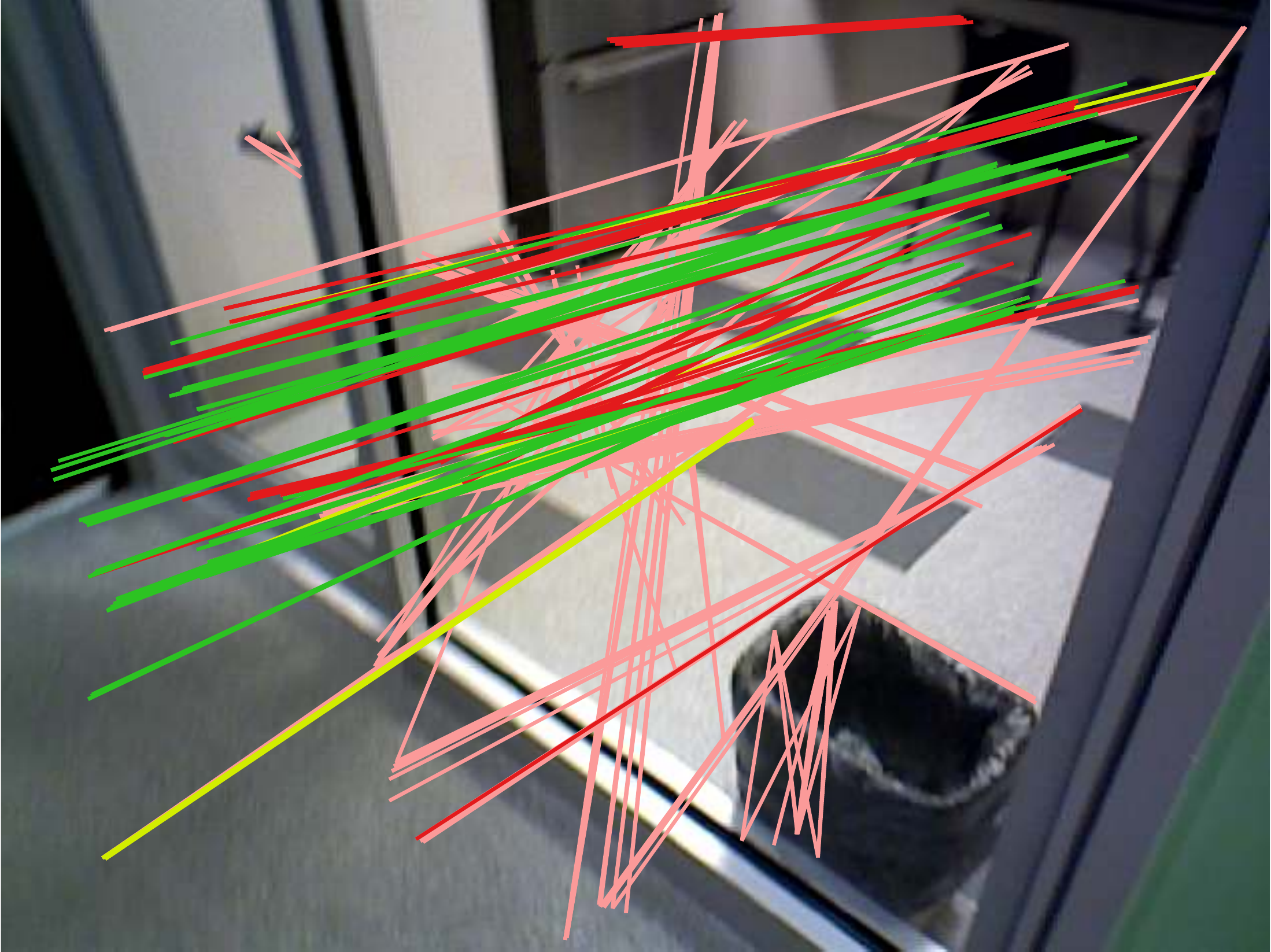}
	\\
	\vspace{0.5em}
	\rotatebox[origin=l]{90}{\mbox{\hspace{2em}LMR}}
	\includegraphics[height=7.5em]{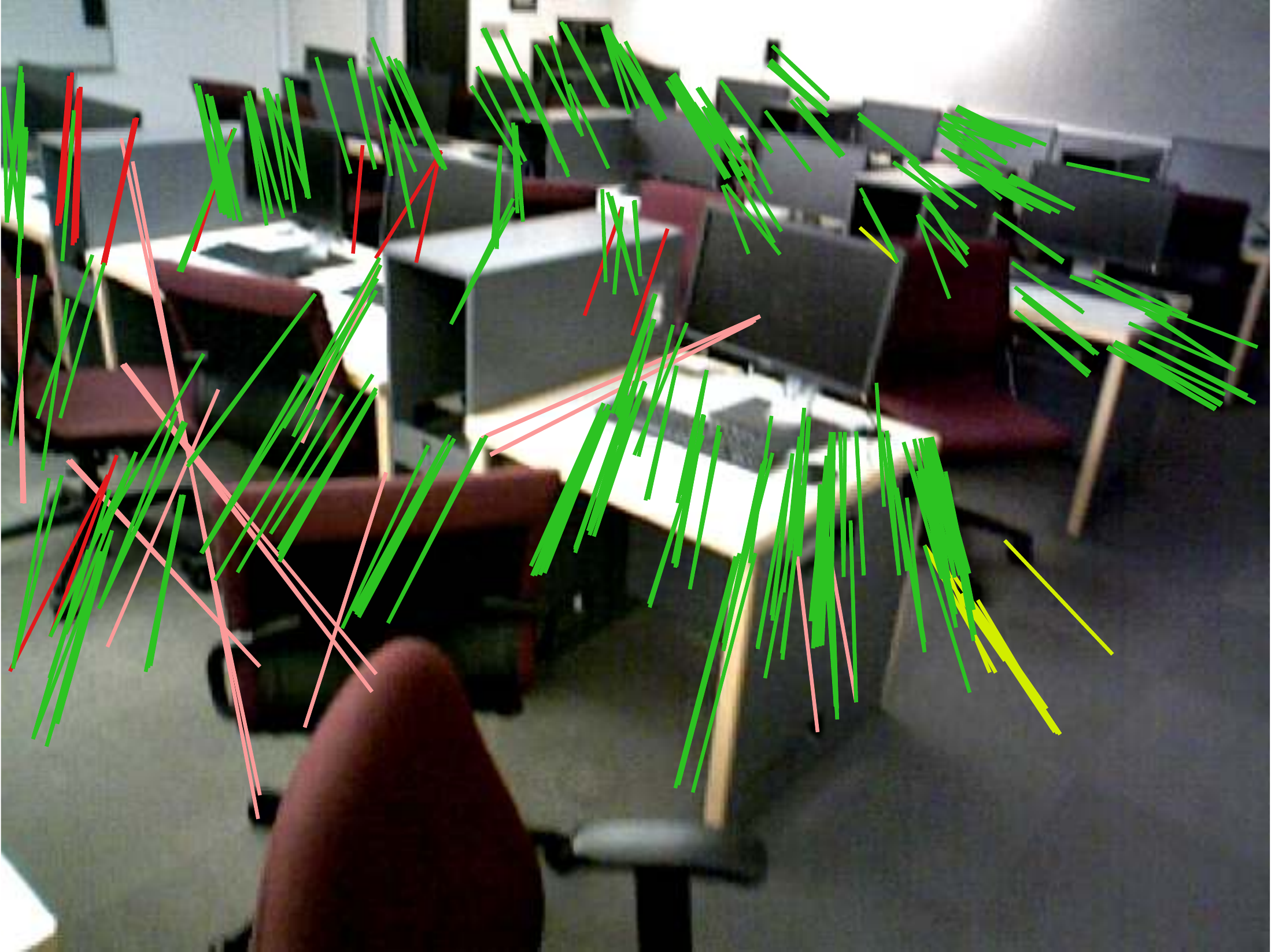}
	\includegraphics[height=7.5em]{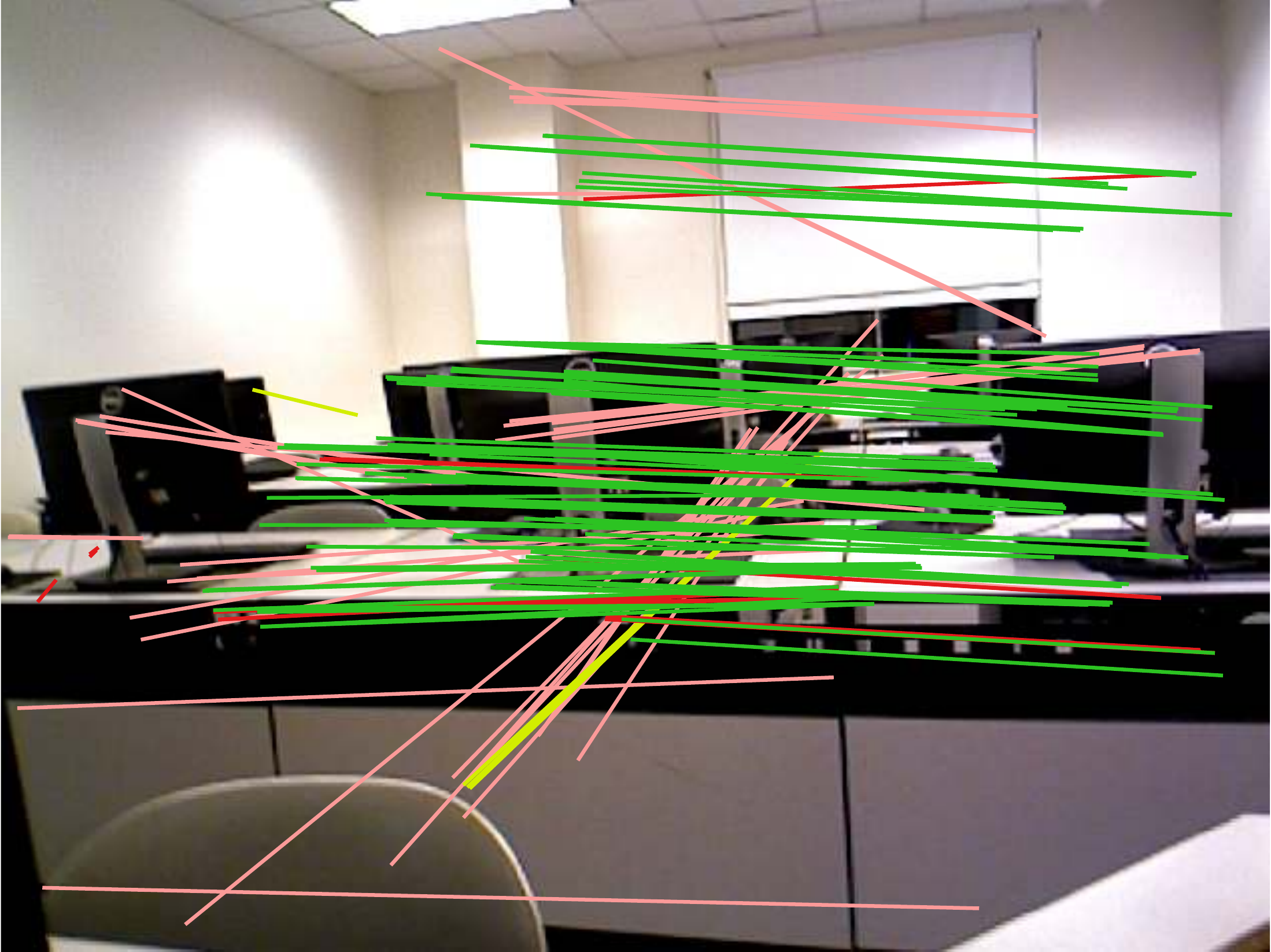}
	\includegraphics[height=7.5em]{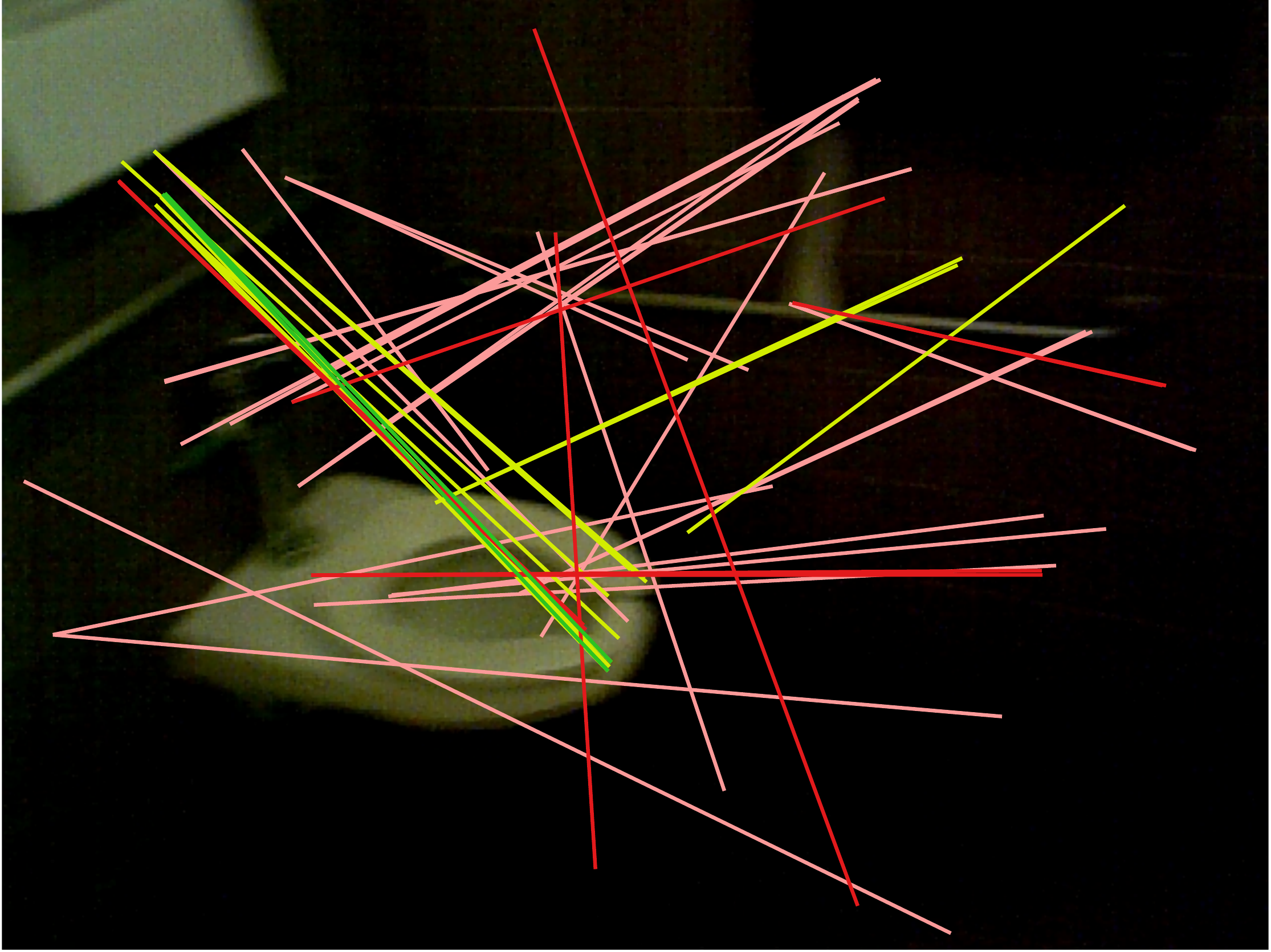}
	\includegraphics[height=7.5em]{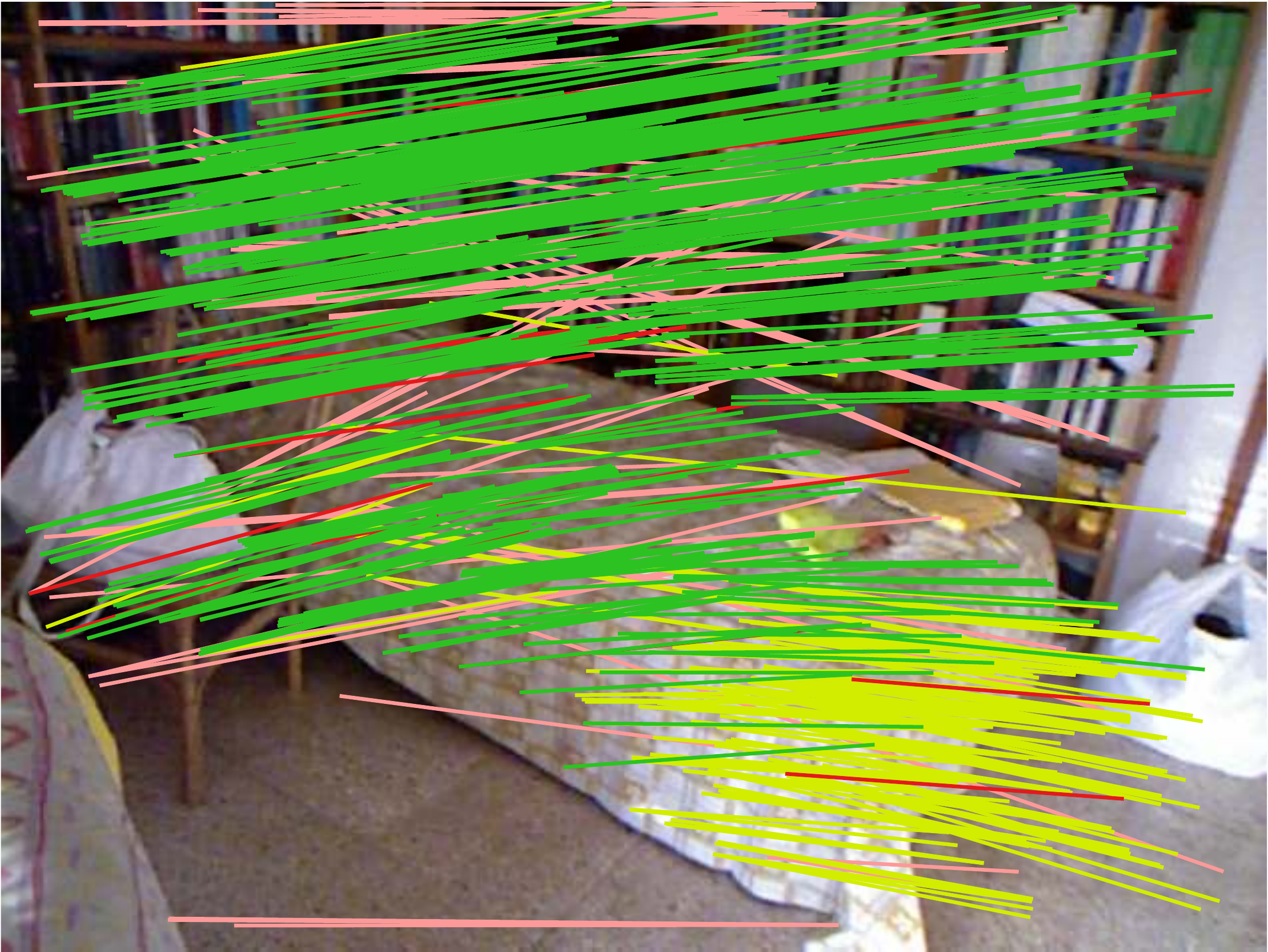}
	\includegraphics[height=7.5em]{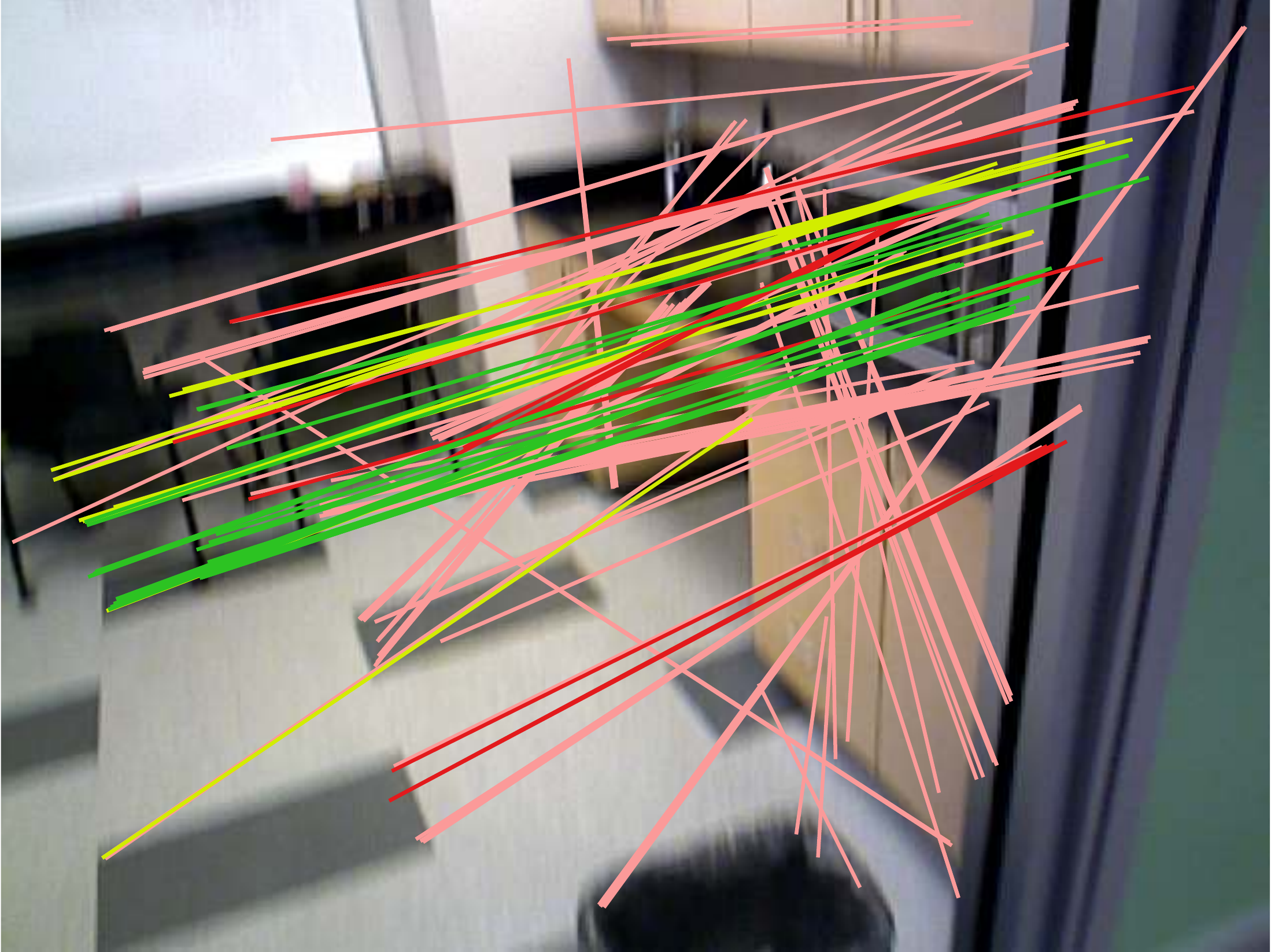}
	\\
	\vspace{0.5em}
	\rotatebox[origin=l]{90}{\mbox{\hspace{2em}LPM}}
	\includegraphics[height=7.5em]{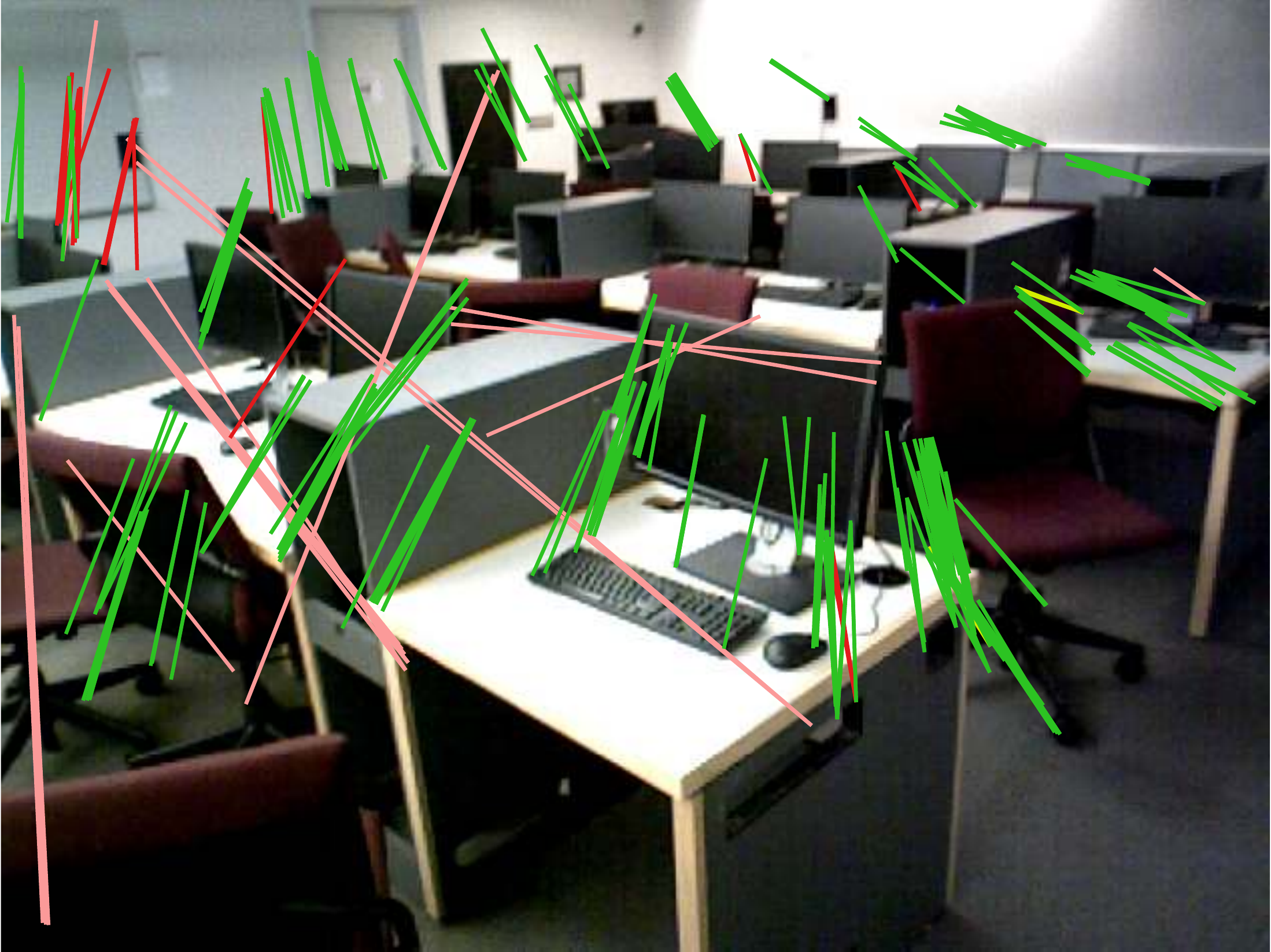}
	\includegraphics[height=7.5em]{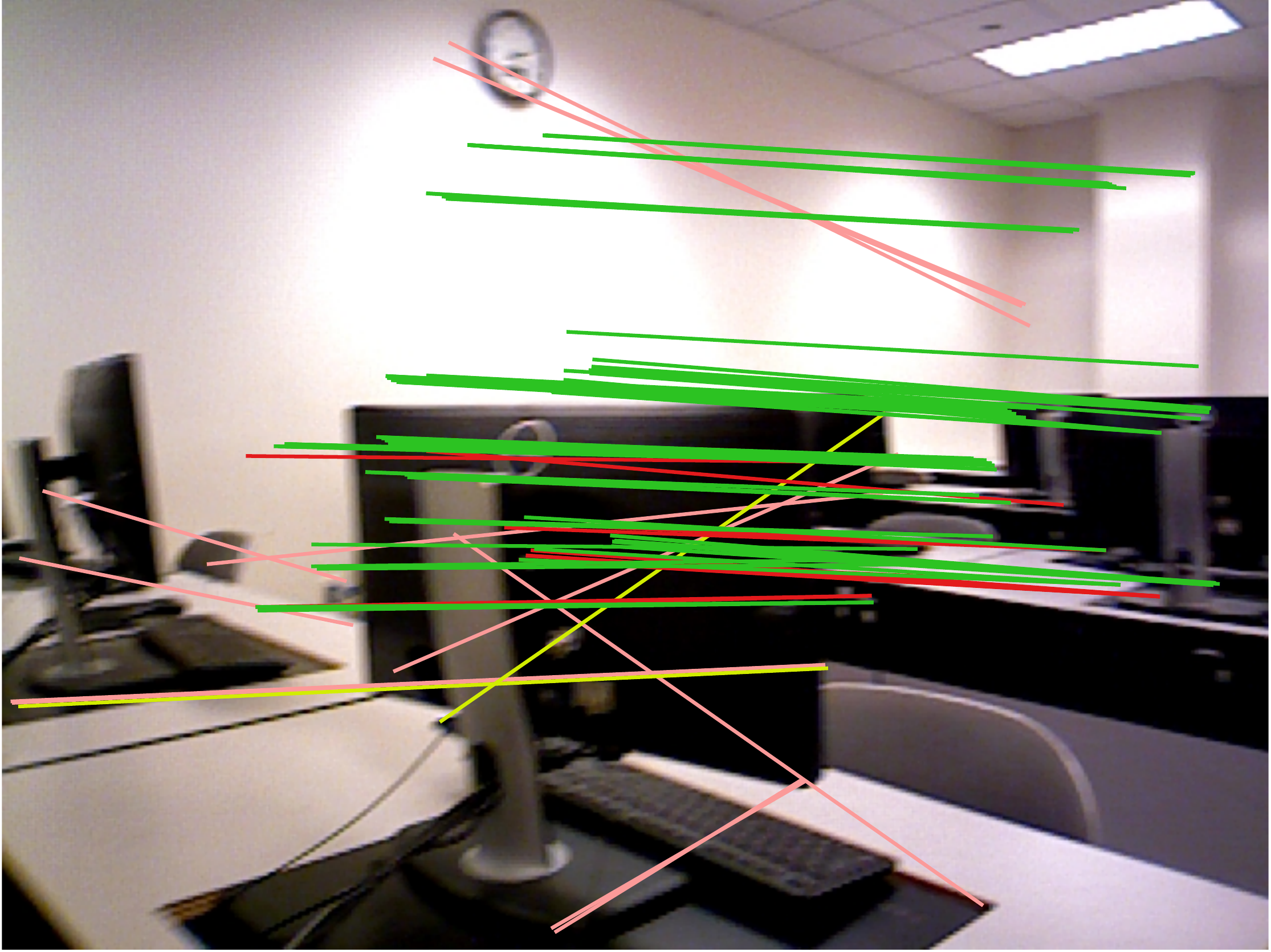}
	\includegraphics[height=7.5em]{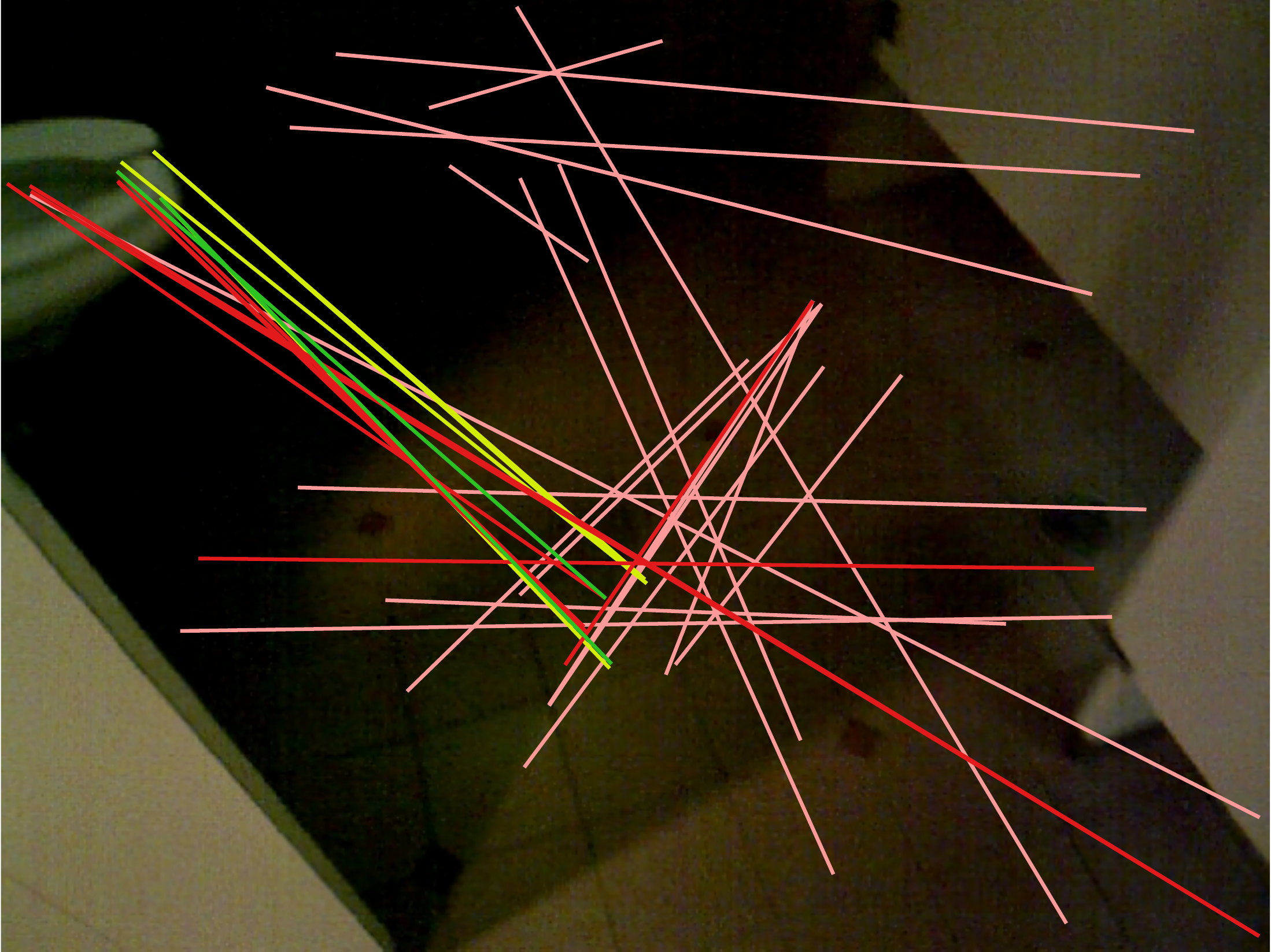}
	\includegraphics[height=7.5em]{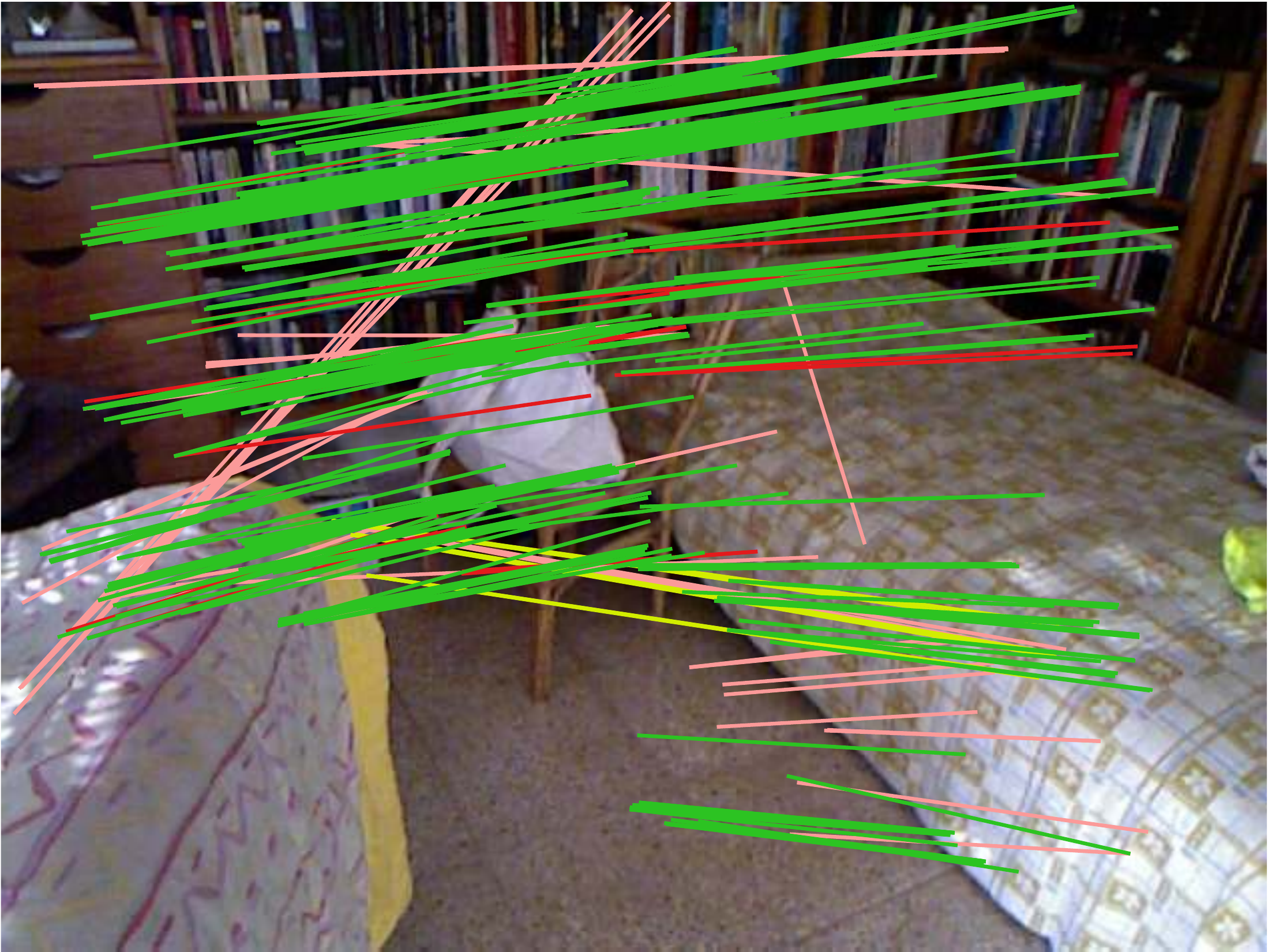}
	\includegraphics[height=7.5em]{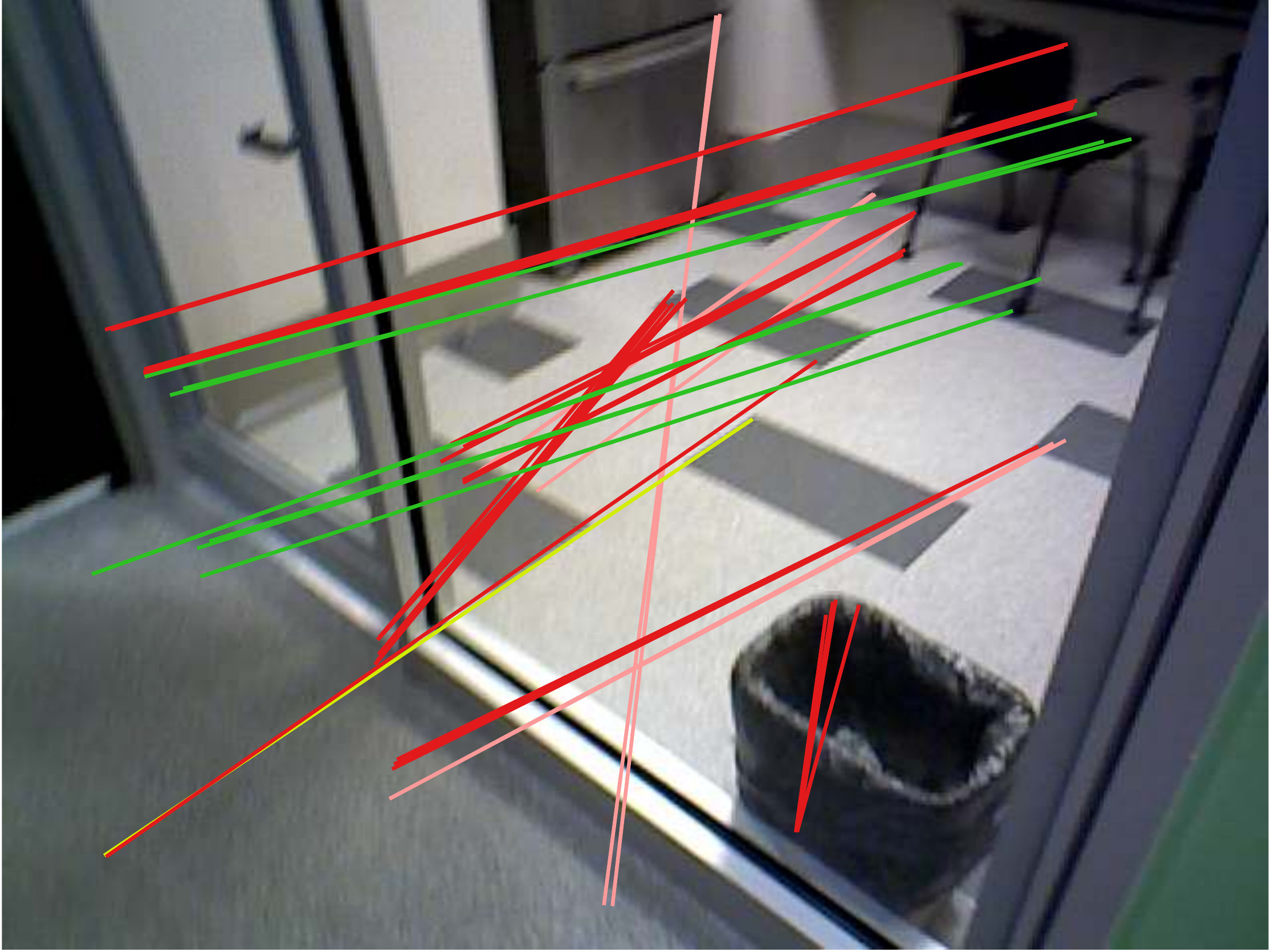}
	\\
	\vspace{0.5em}
	\rotatebox[origin=l]{90}{\mbox{\hspace{2em}GLPM}}
	\includegraphics[height=7.5em]{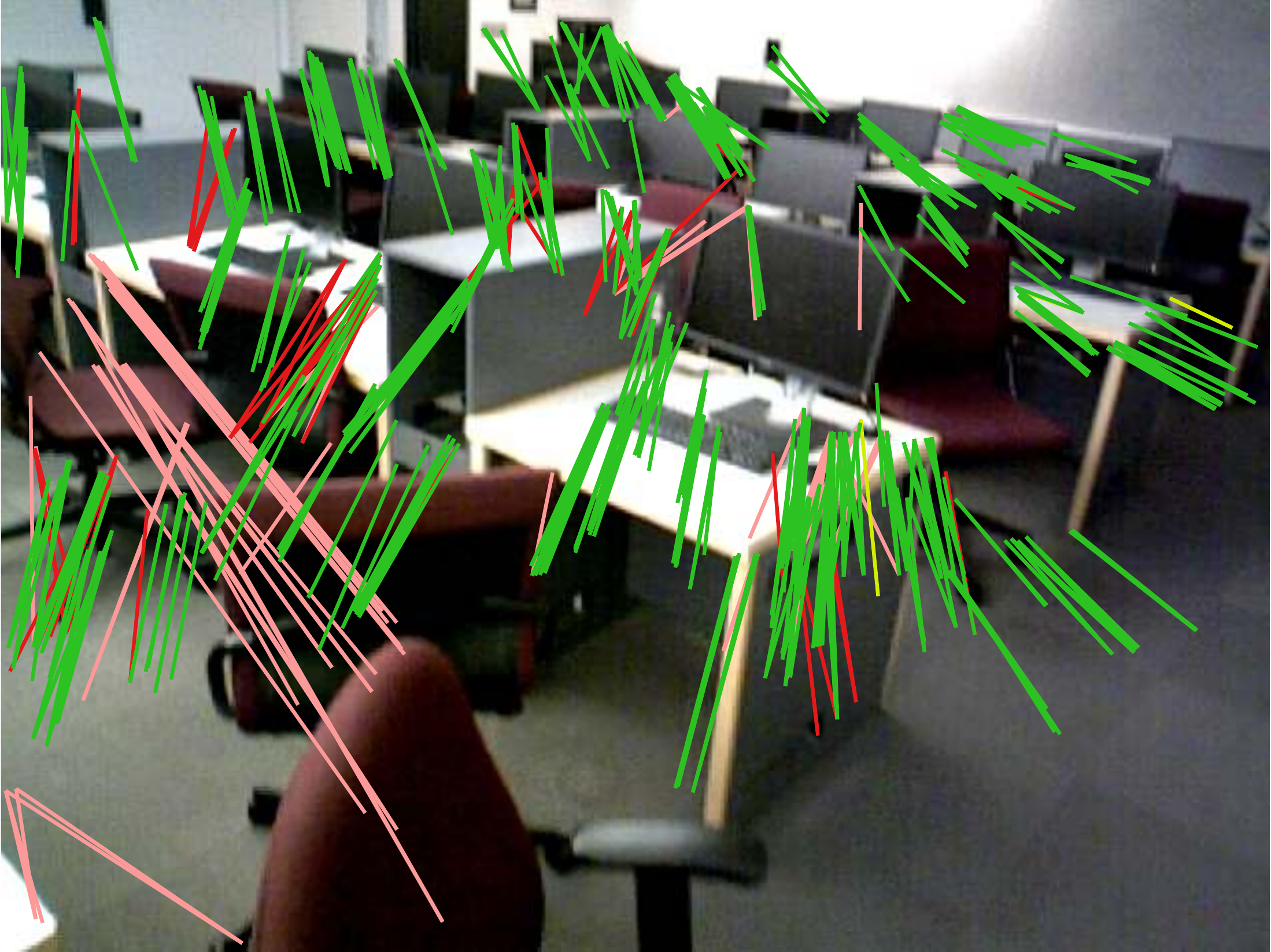}
	\includegraphics[height=7.5em]{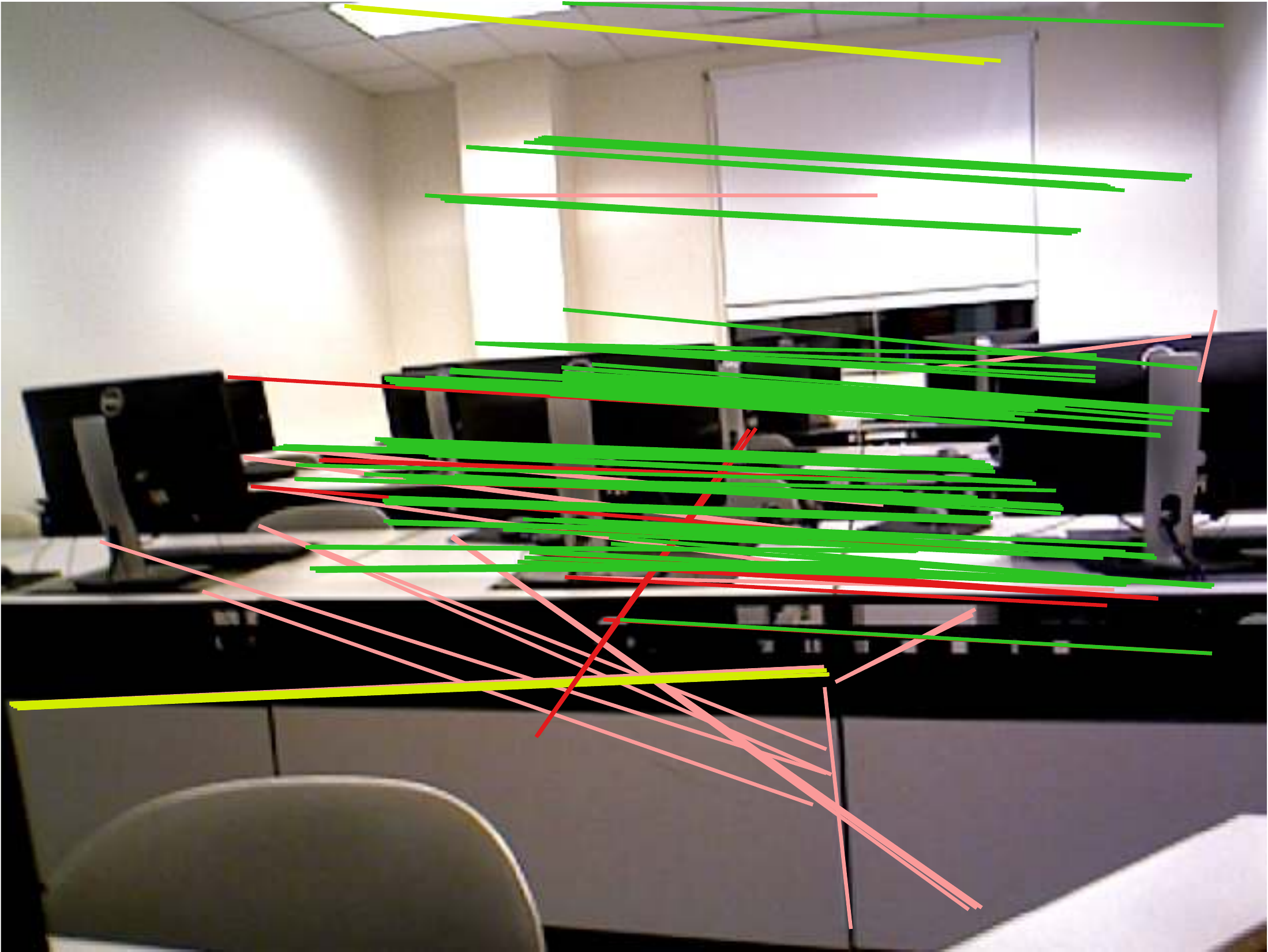}
	\includegraphics[height=7.5em]{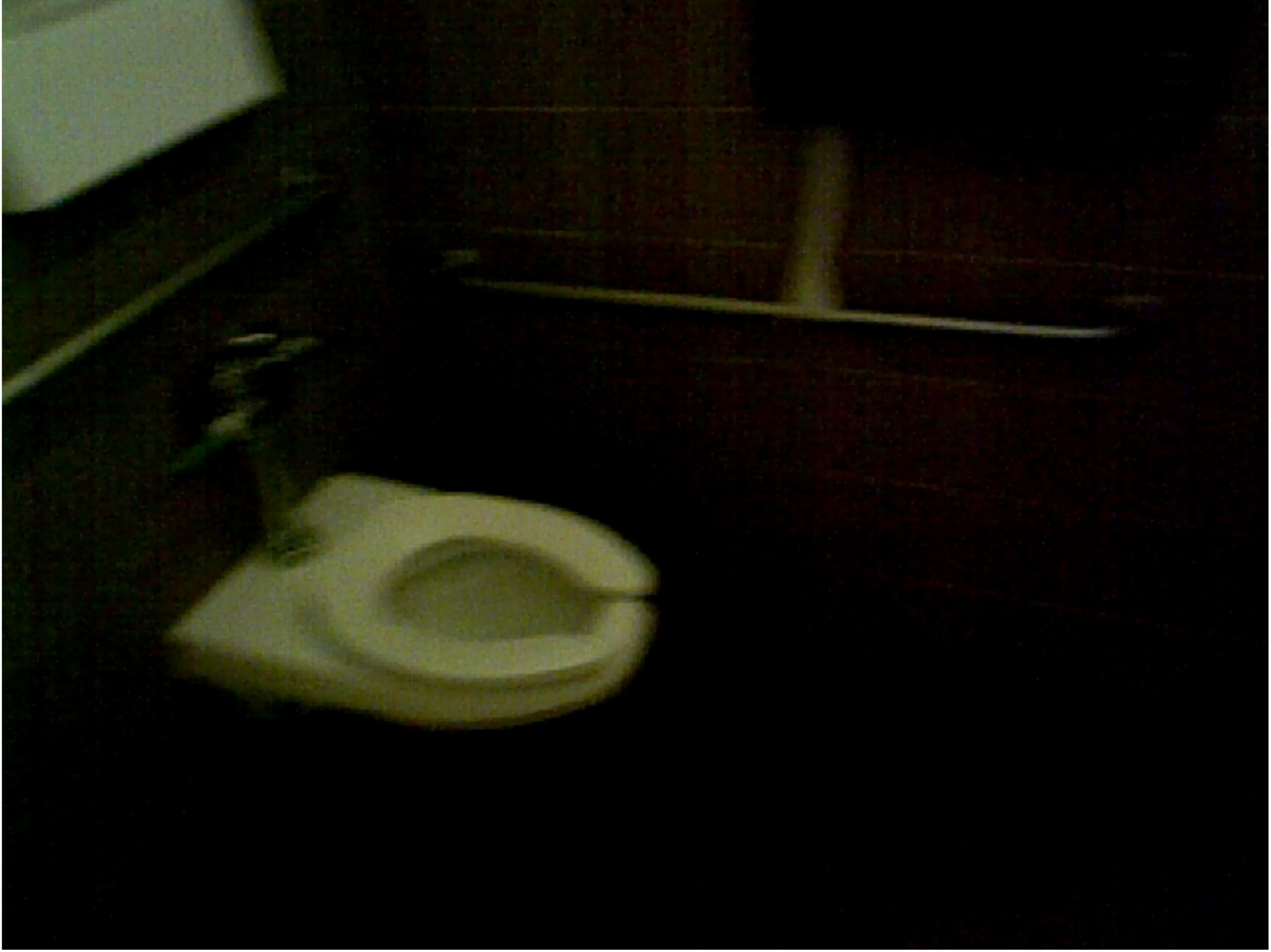}
	\includegraphics[height=7.5em]{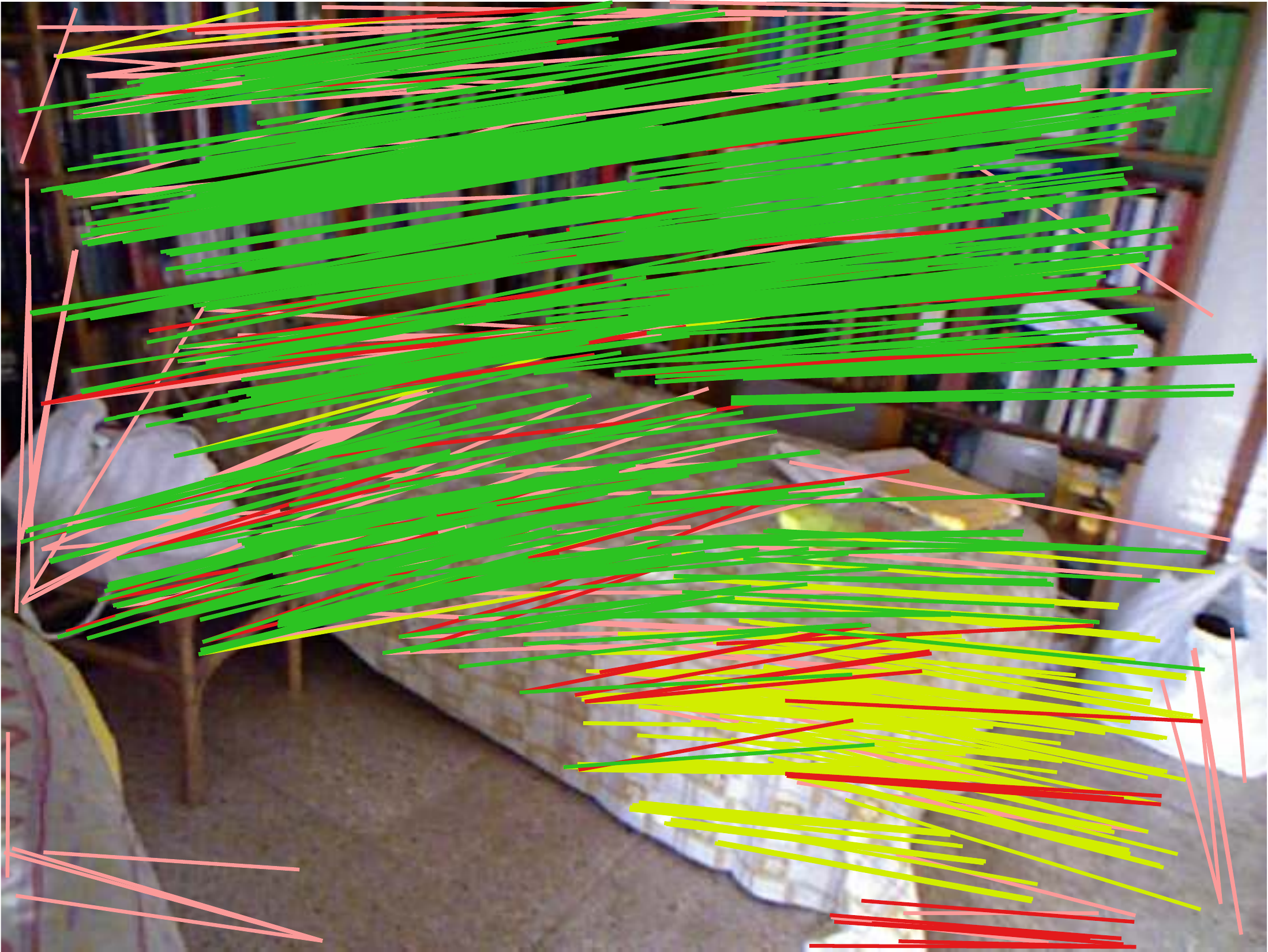}
	\includegraphics[height=7.5em]{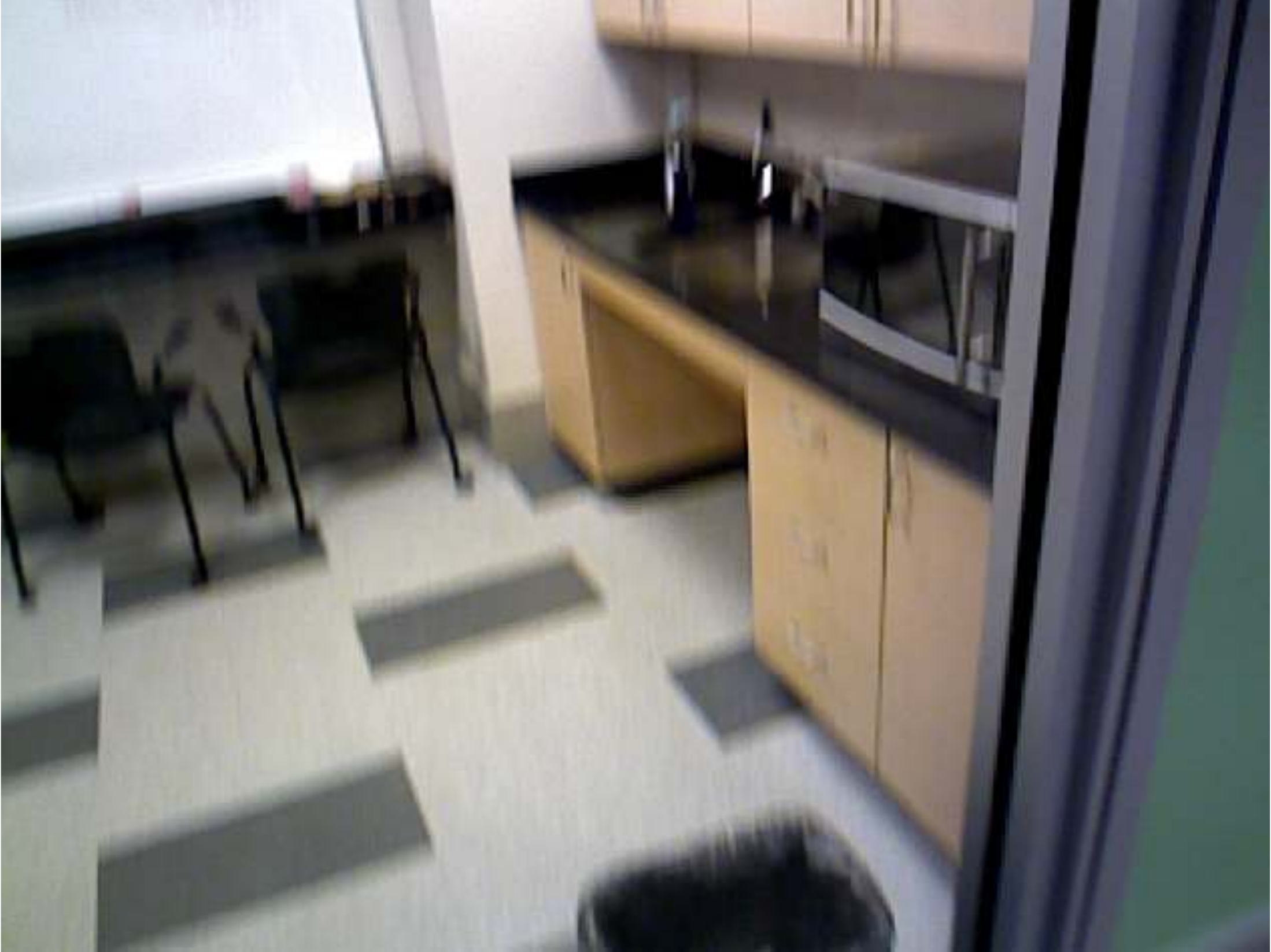}
	\\
	\vspace{0.5em}
	\rotatebox[origin=l]{90}{\mbox{\hspace{2em}GMS}}
	\includegraphics[height=7.5em]{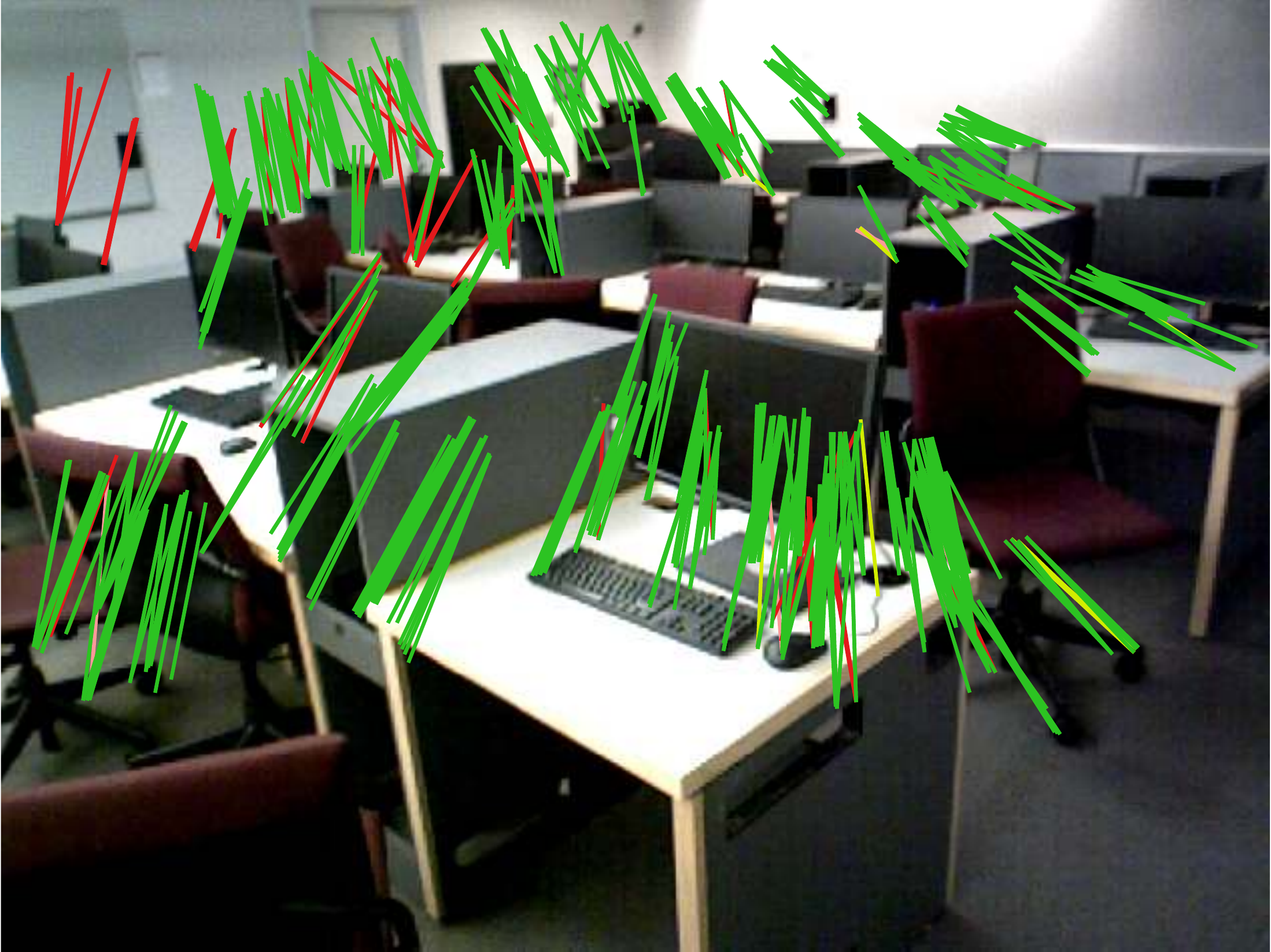}
	\includegraphics[height=7.5em]{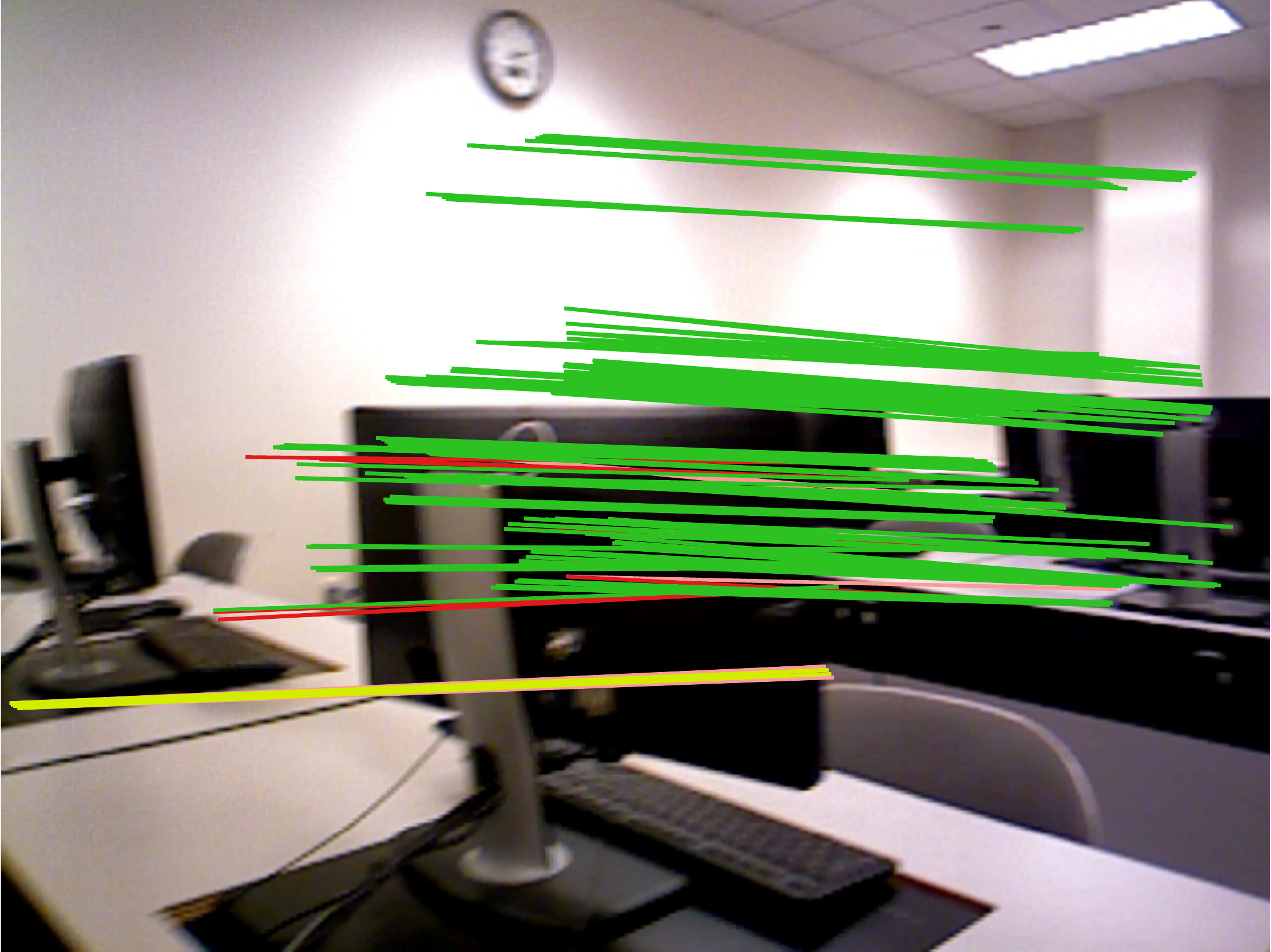}
	\includegraphics[height=7.5em]{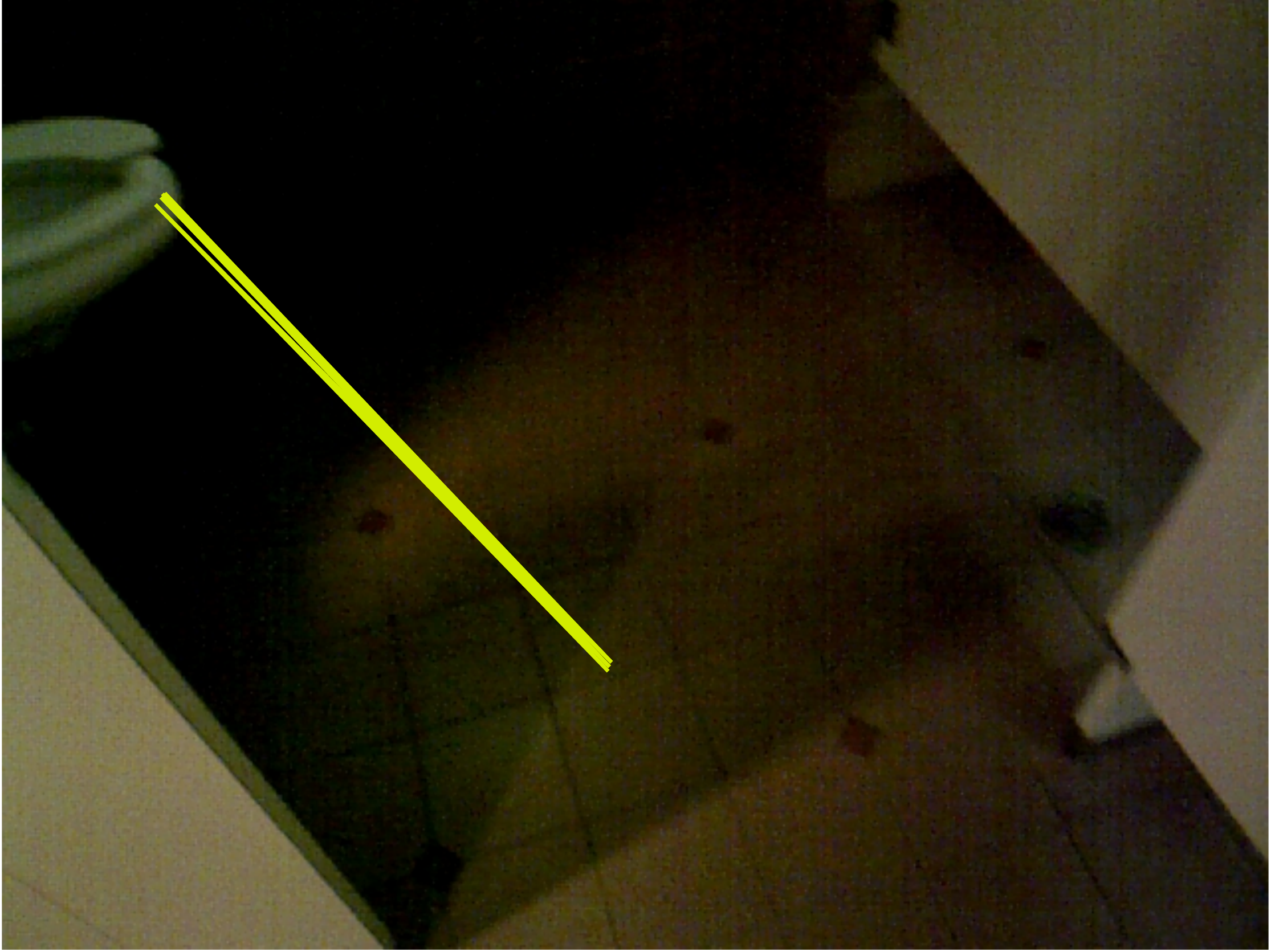}
	\includegraphics[height=7.5em]{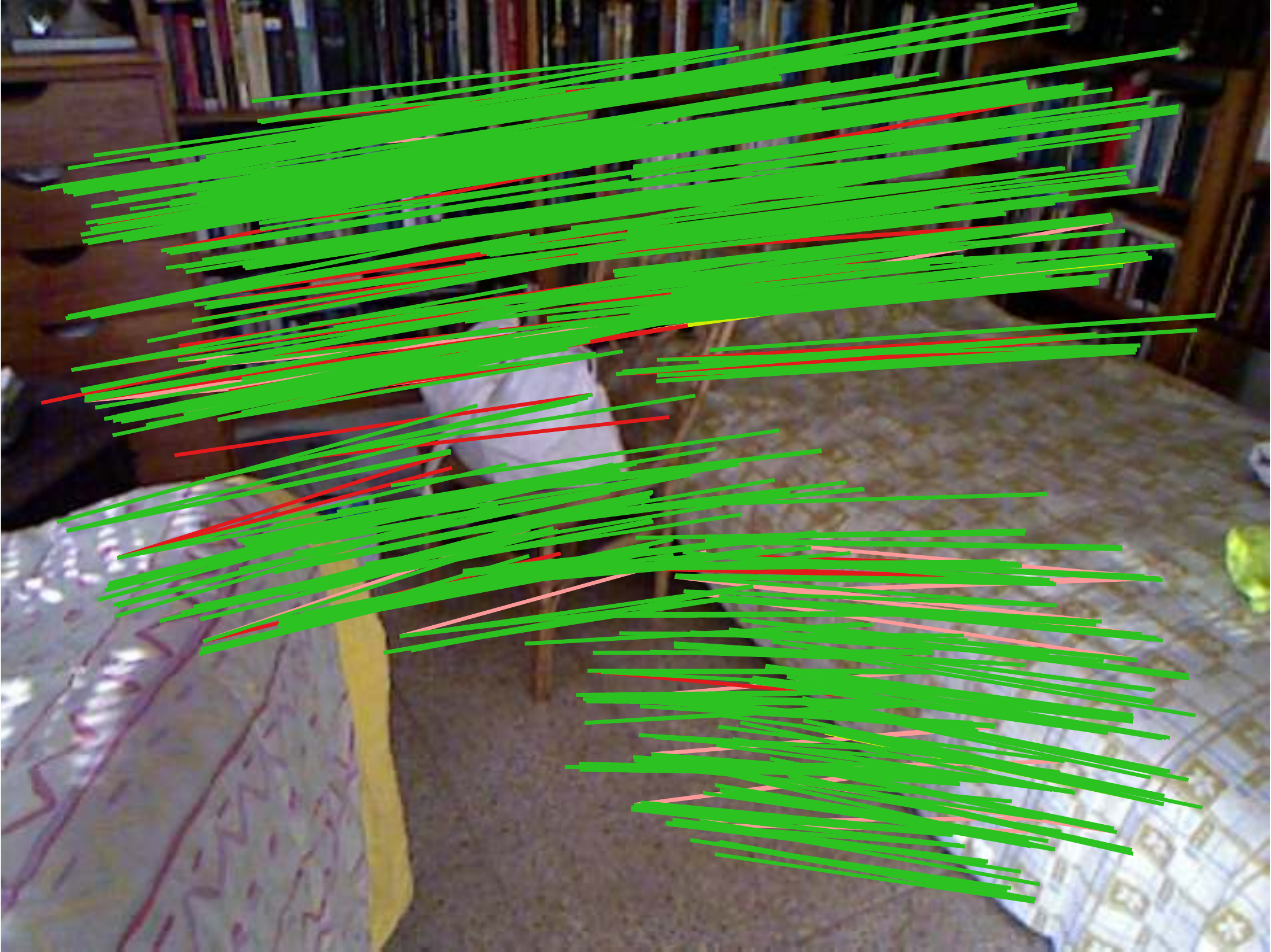}
	\includegraphics[height=7.5em]{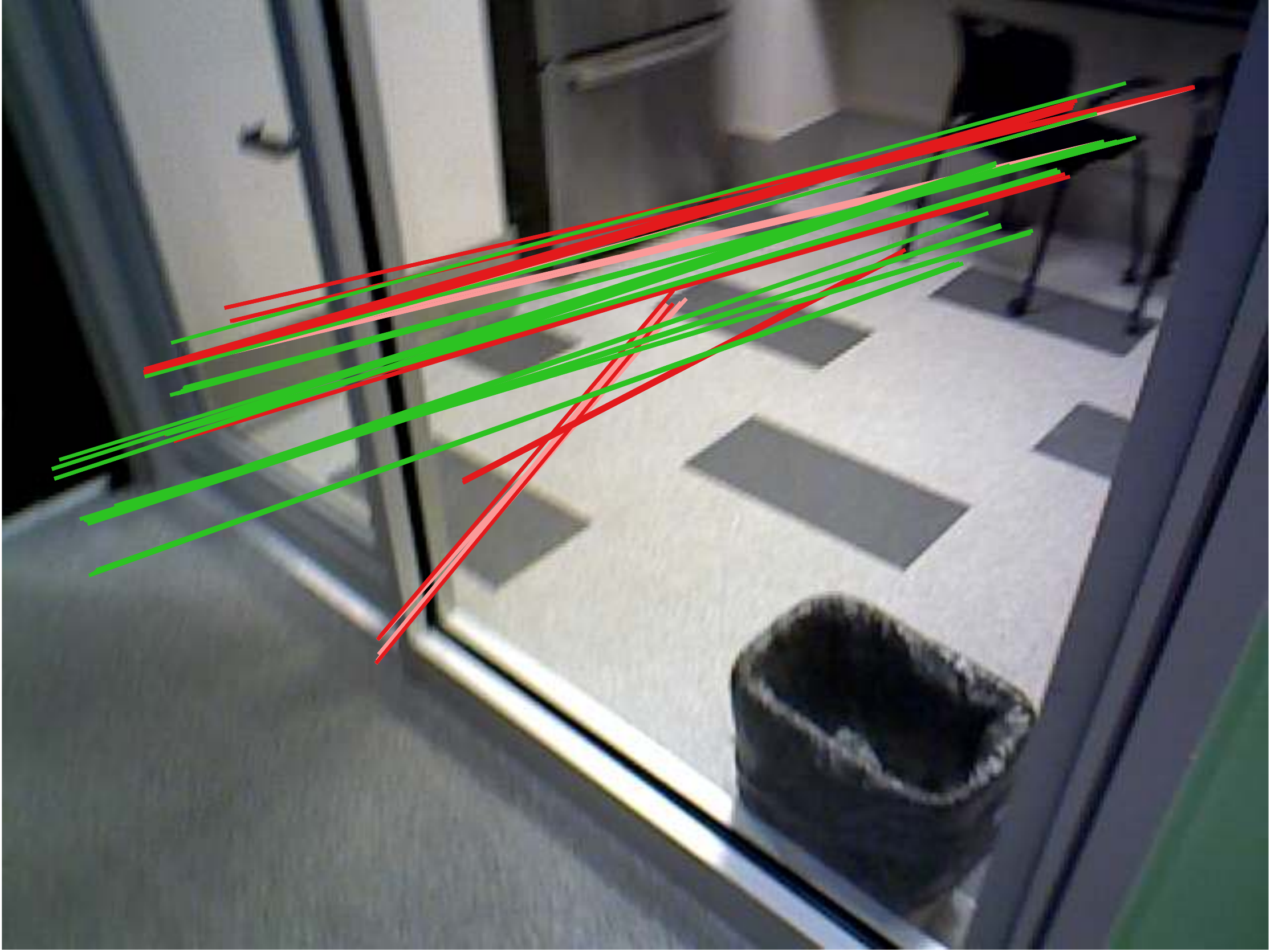}
	\\
	\vspace{0.5em}
	\rotatebox[origin=l]{90}{\mbox{\hspace{2em}VFC}}
	\includegraphics[height=7.5em]{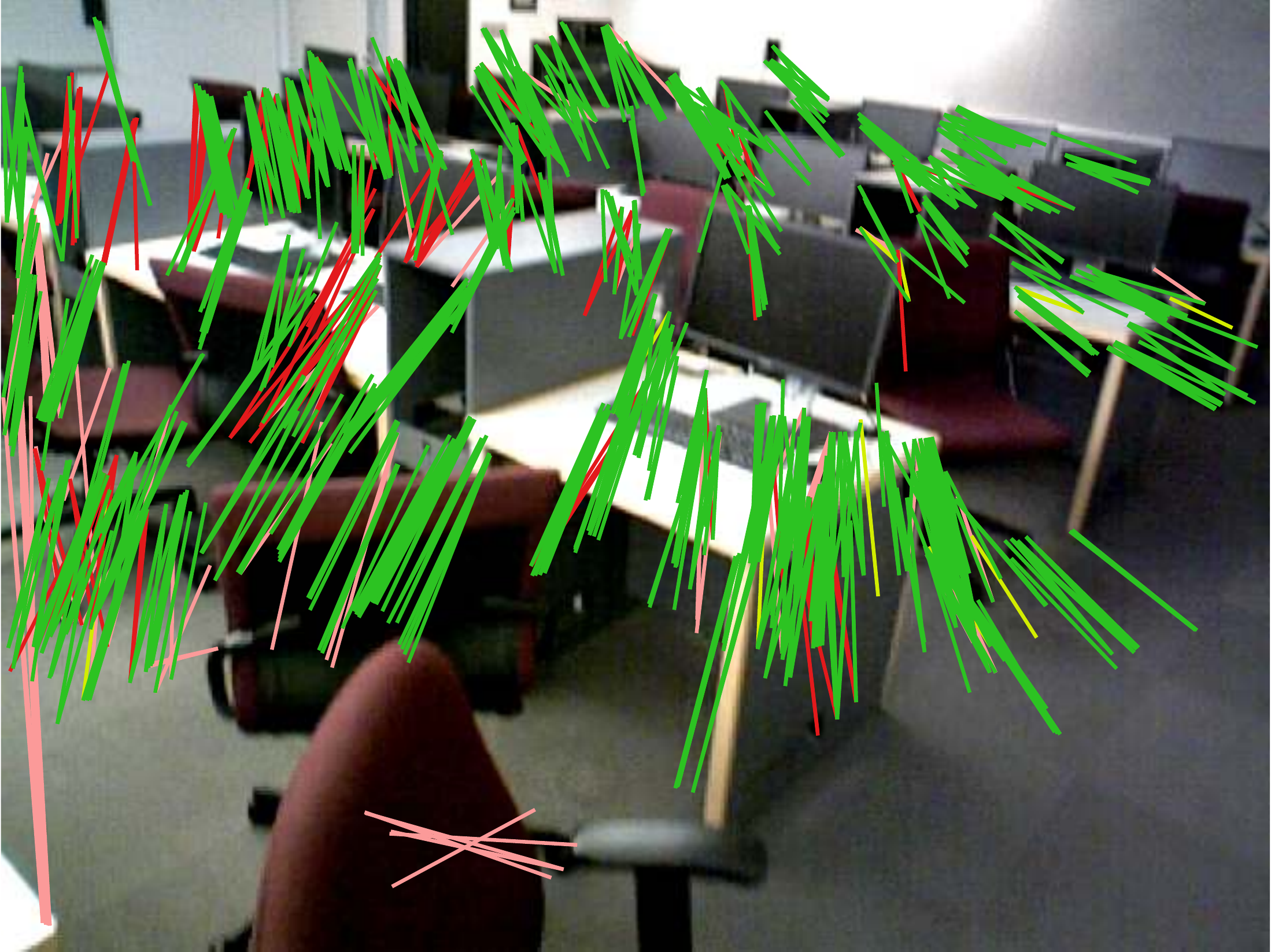}
	\includegraphics[height=7.5em]{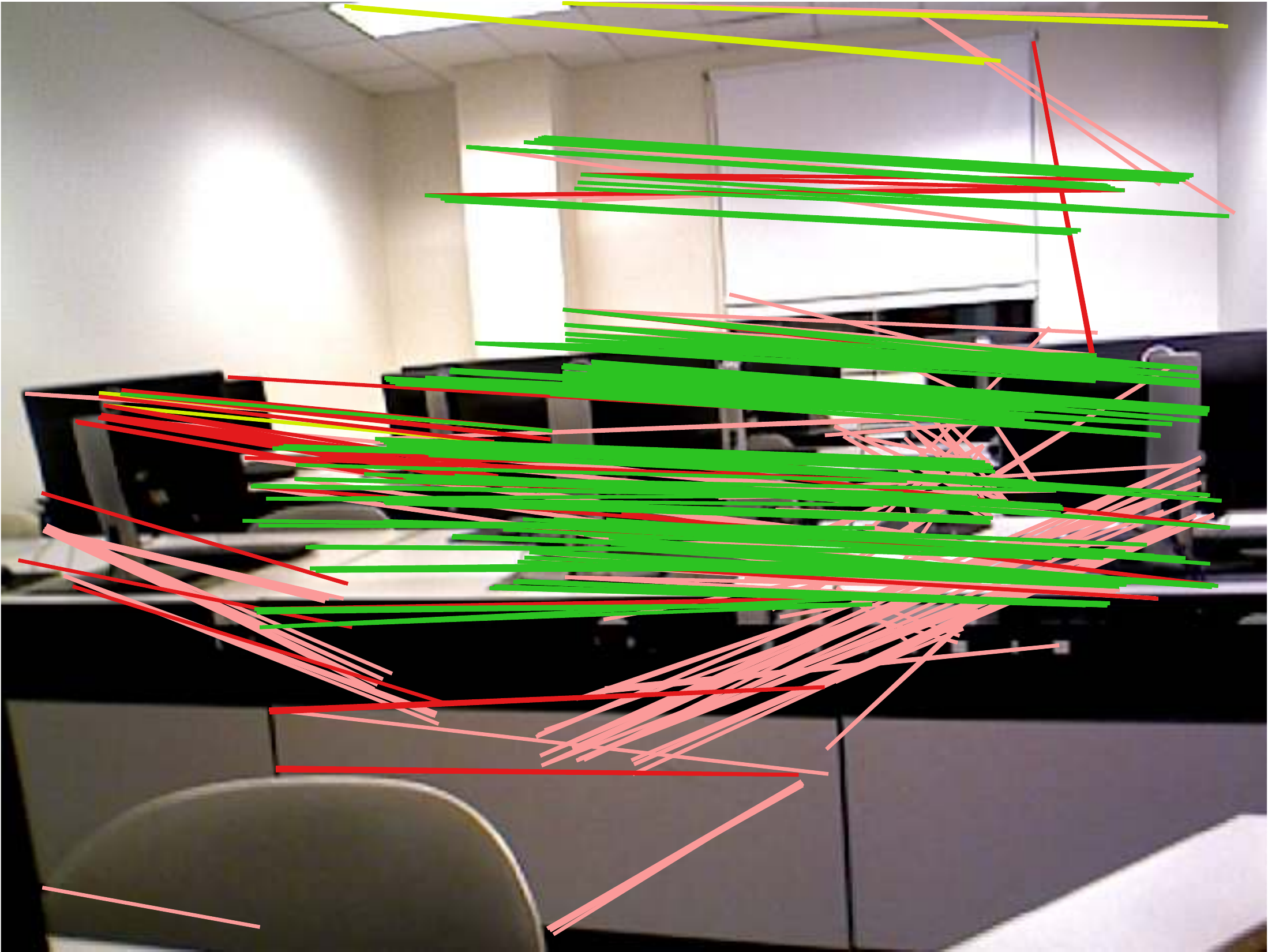}
	\includegraphics[height=7.5em]{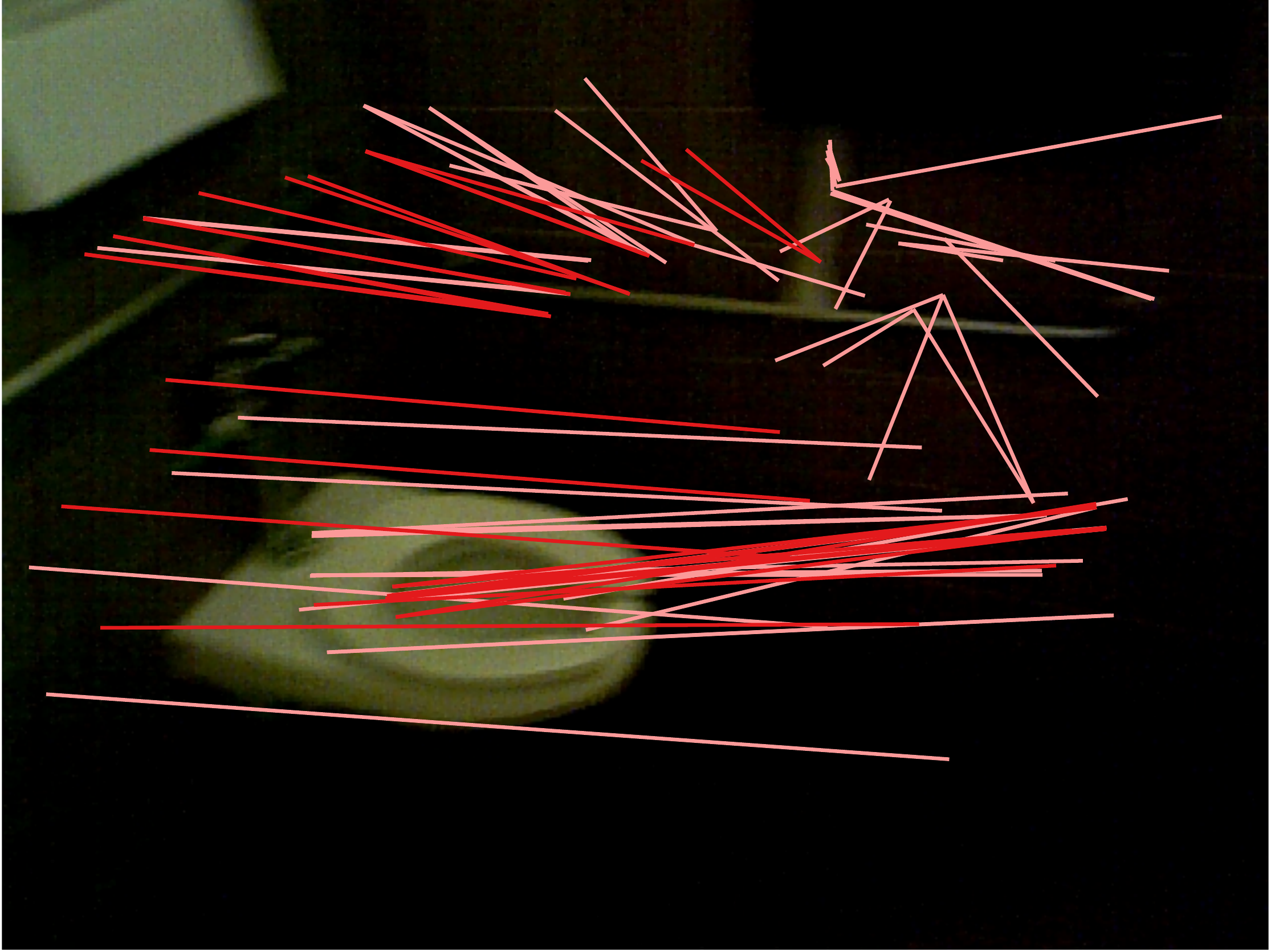}
	\includegraphics[height=7.5em]{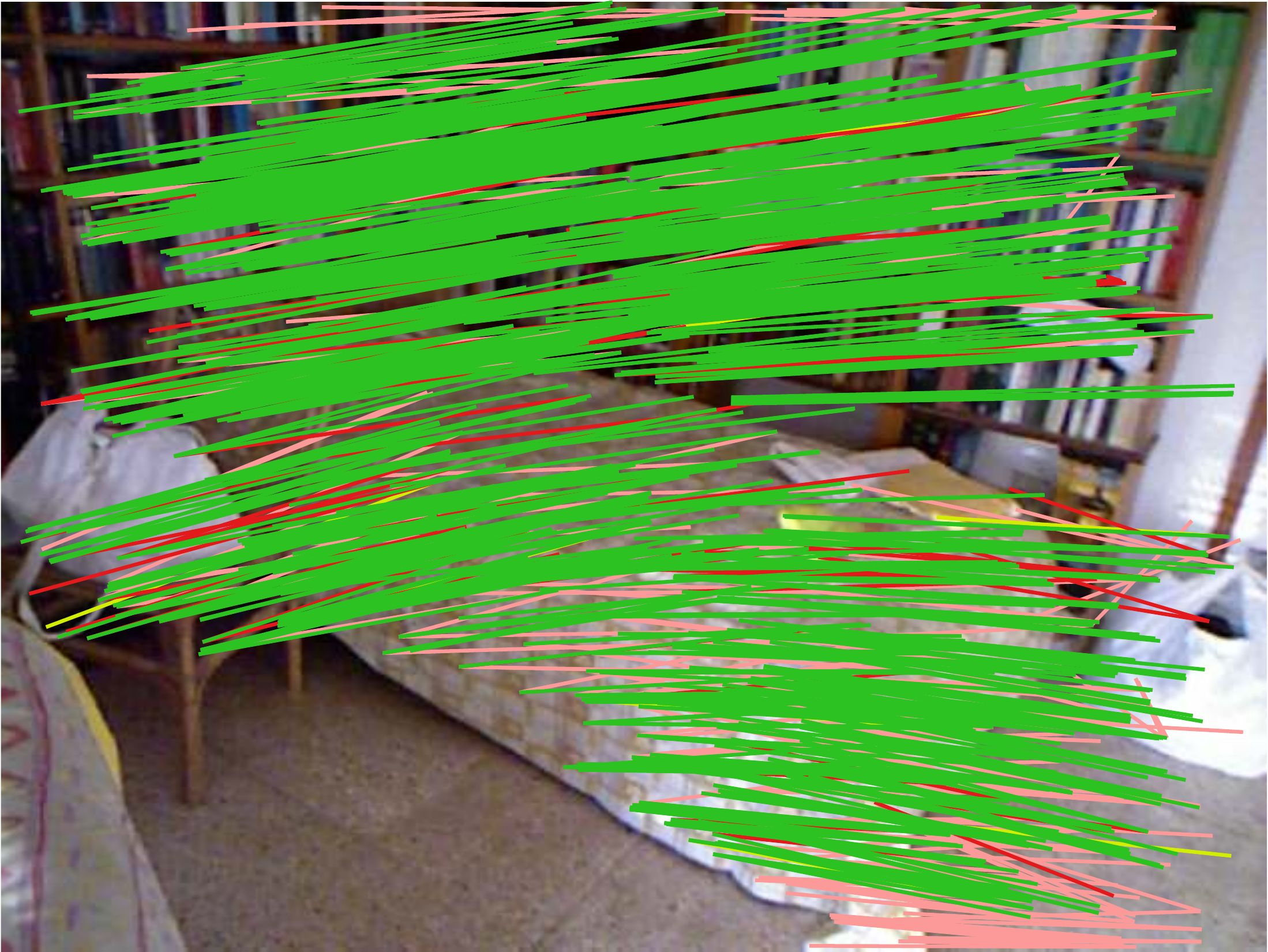}
	\includegraphics[height=7.5em]{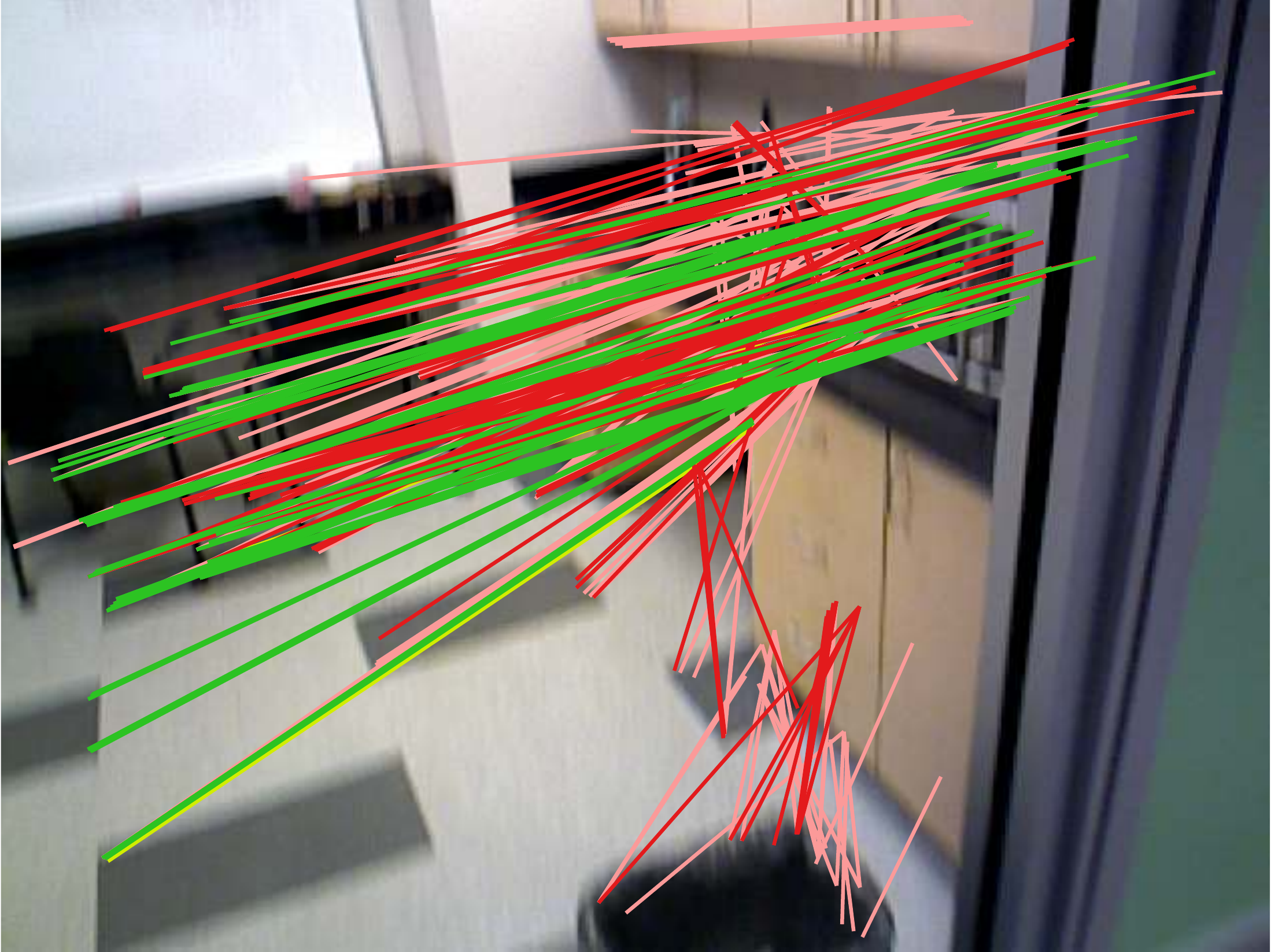}
	\\
	\vspace{0.5em}
	\rotatebox[origin=l]{90}{\mbox{\hspace{2em}LLT}}
	\includegraphics[height=7.5em]{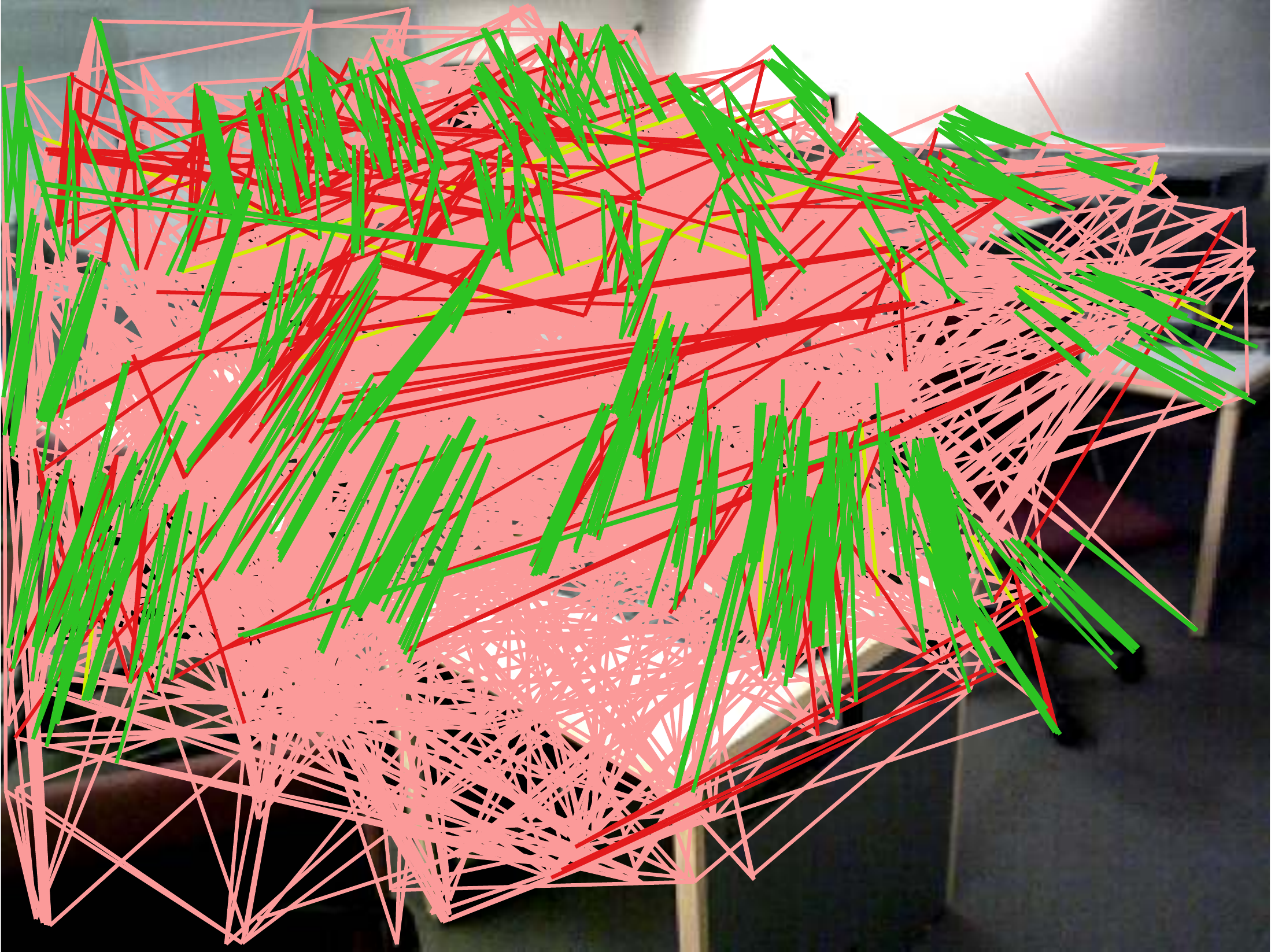}
	\includegraphics[height=7.5em]{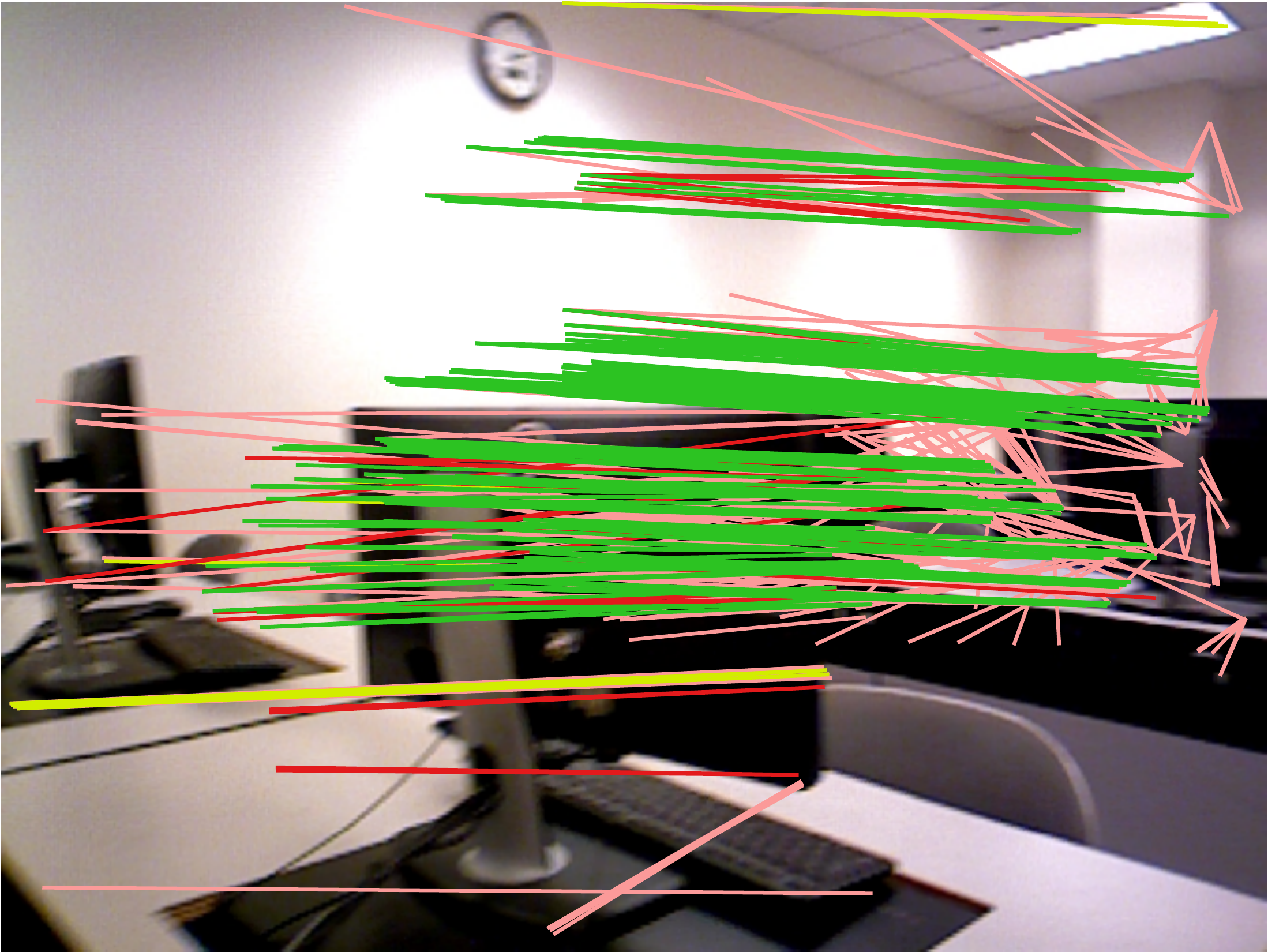}
	\includegraphics[height=7.5em]{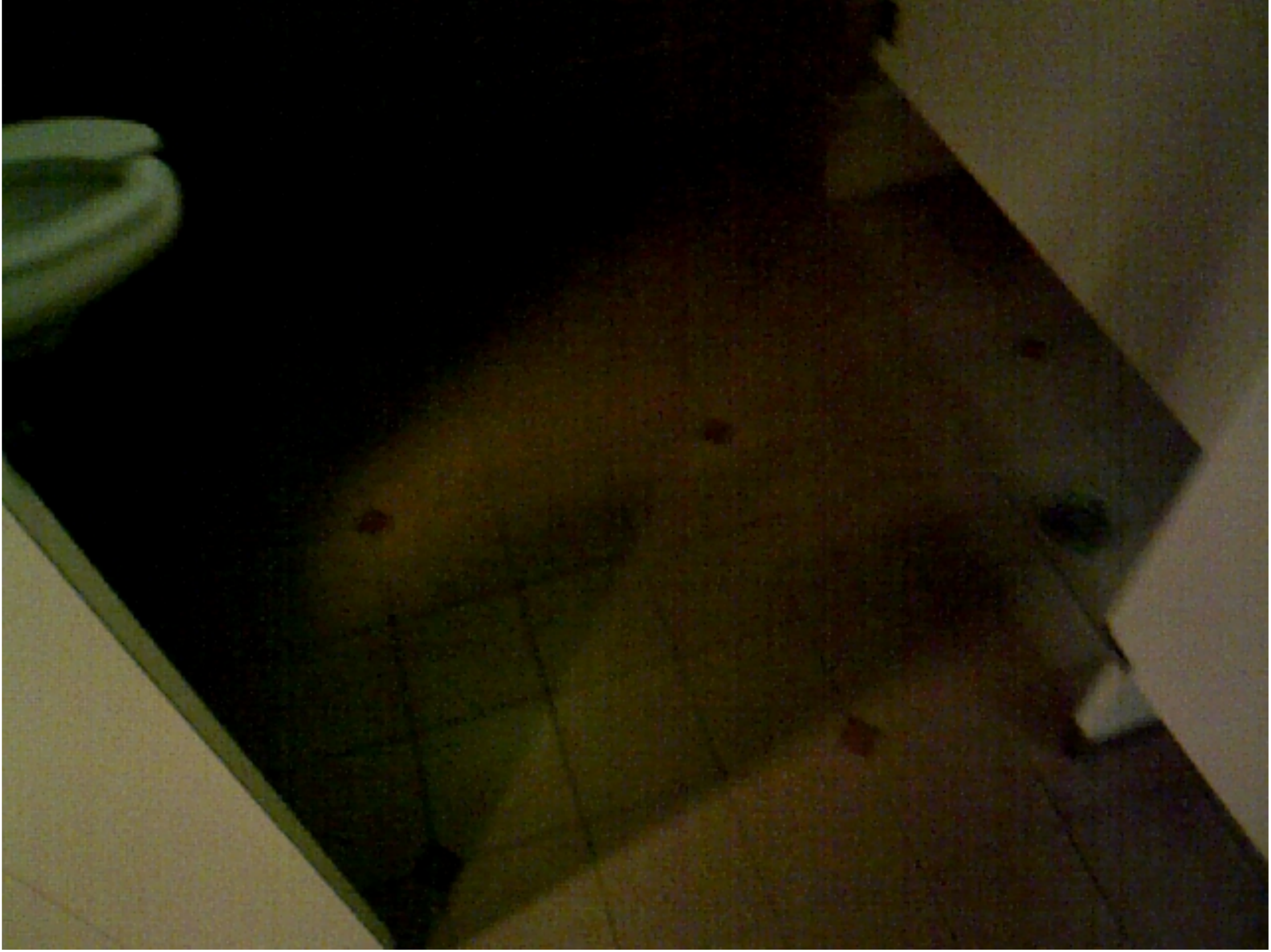}
	\includegraphics[height=7.5em]{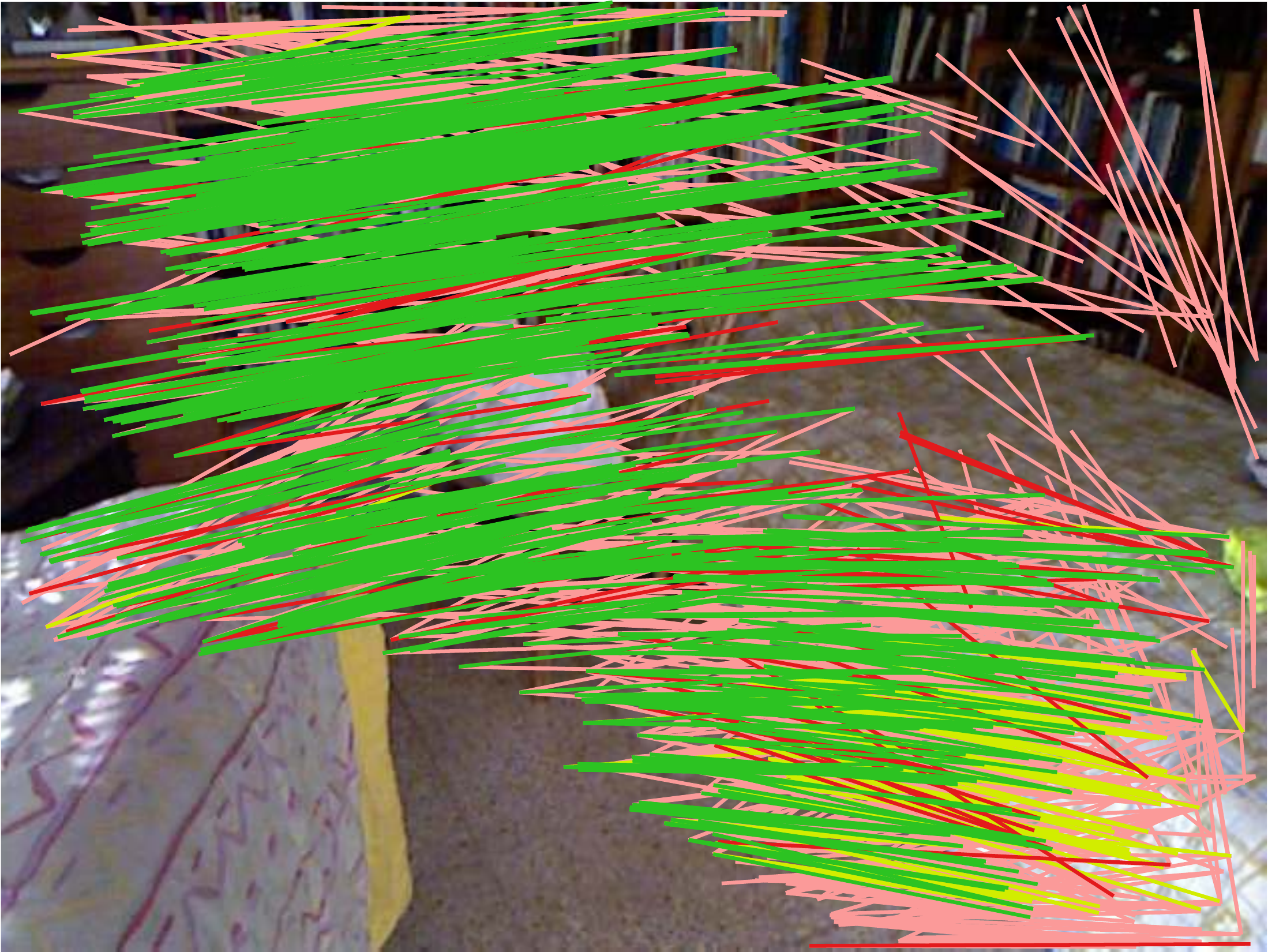}
	\includegraphics[height=7.5em]{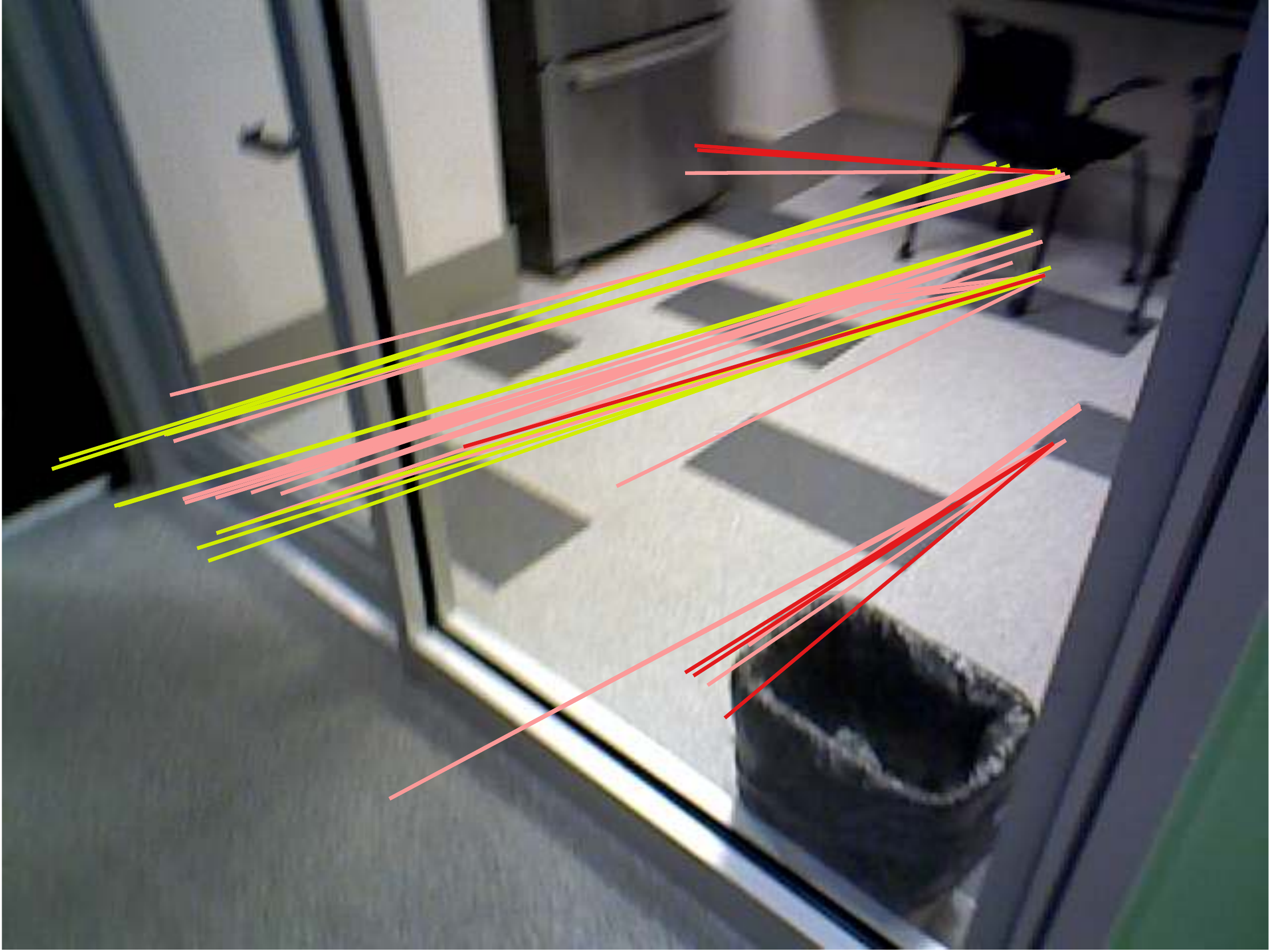}
	\\
	\begin{flushleft}
		\hspace{1.75em}Harvard Computer Lab\hspace{2.5em}Brown CS\hspace{4em}Harvard Restroom\hspace{2em}Home Puigpunyen\hspace{2.5em}MIT Office
	\end{flushleft}
	\caption{\label{example_3a}
		SUN3D local spatial filter matches according to the best configuration setup, the images of the input pair alternate among the rows. For each method inlier (yellow, green) and outlier (red and light red) clusters are shown, as well as the 1SAC filtered matches (green, red) (see Sec.~\ref{eval_dt}, best viewed in color and zoomed in).}
\end{figure*}

\begin{figure*}
	\center
	\rotatebox[origin=l]{90}{\mbox{\hspace{0.5em}RFM-SCAN}}
	\includegraphics[height=7.5em]{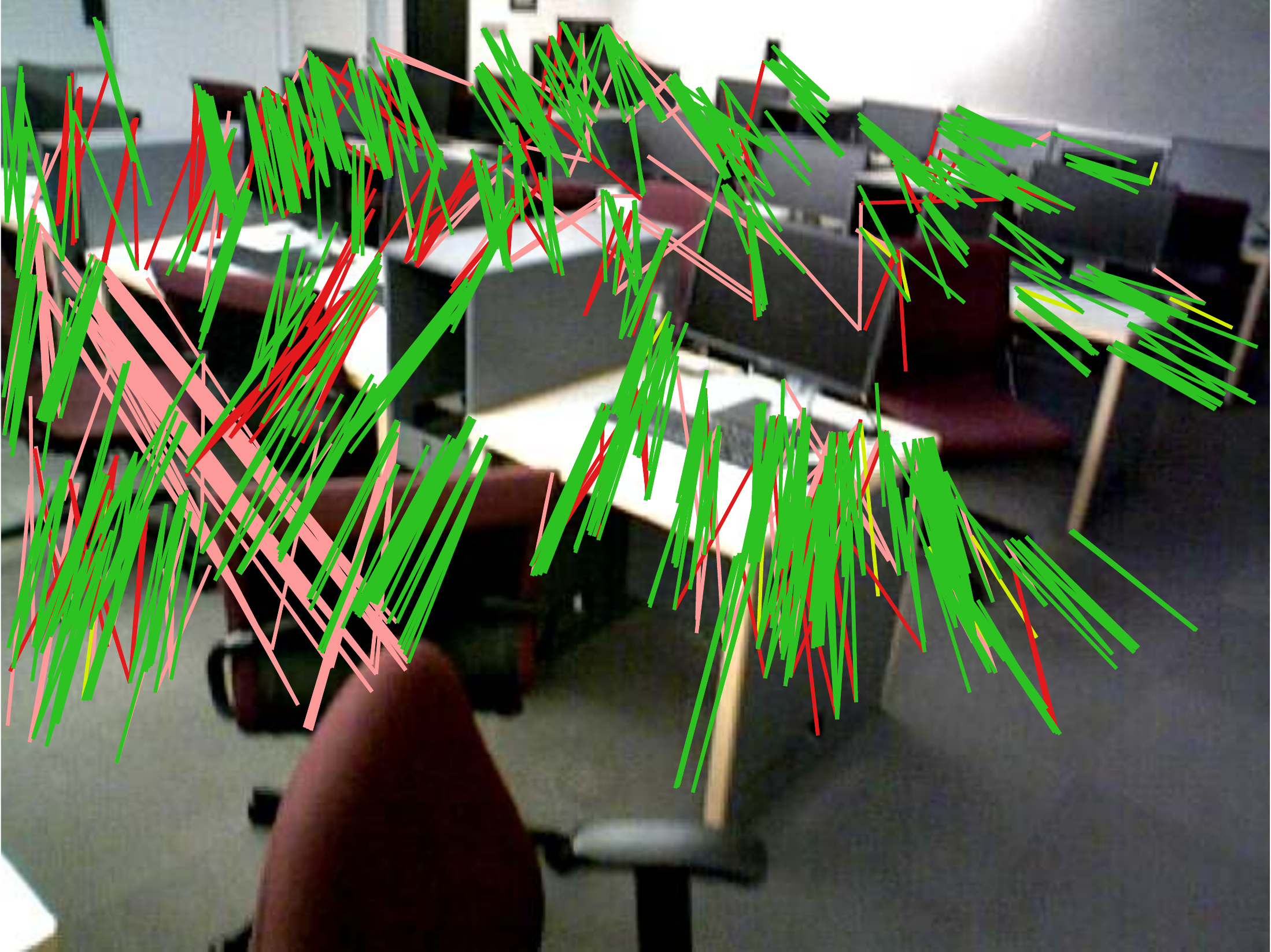}
	\includegraphics[height=7.5em]{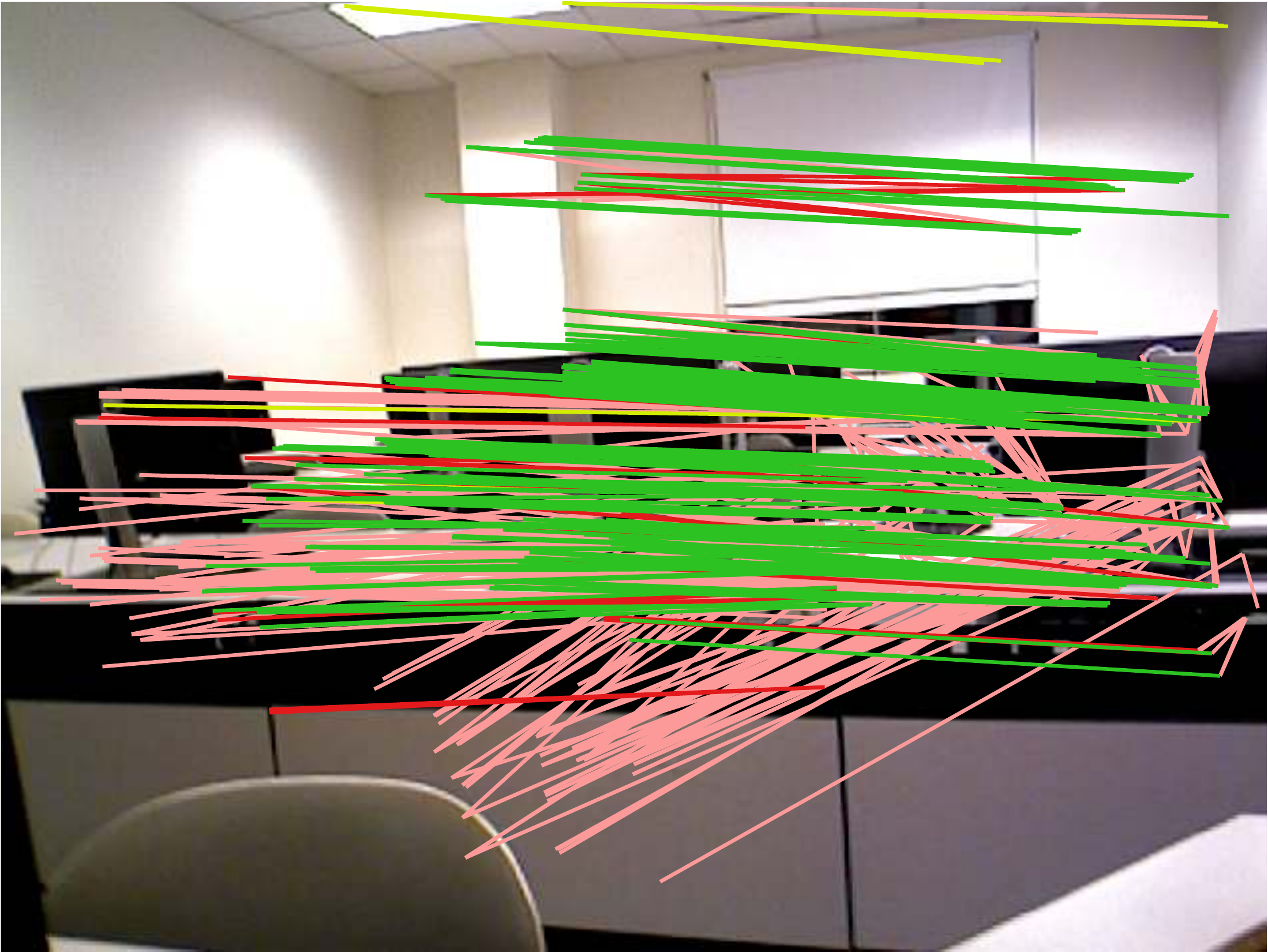}
	\includegraphics[height=7.5em]{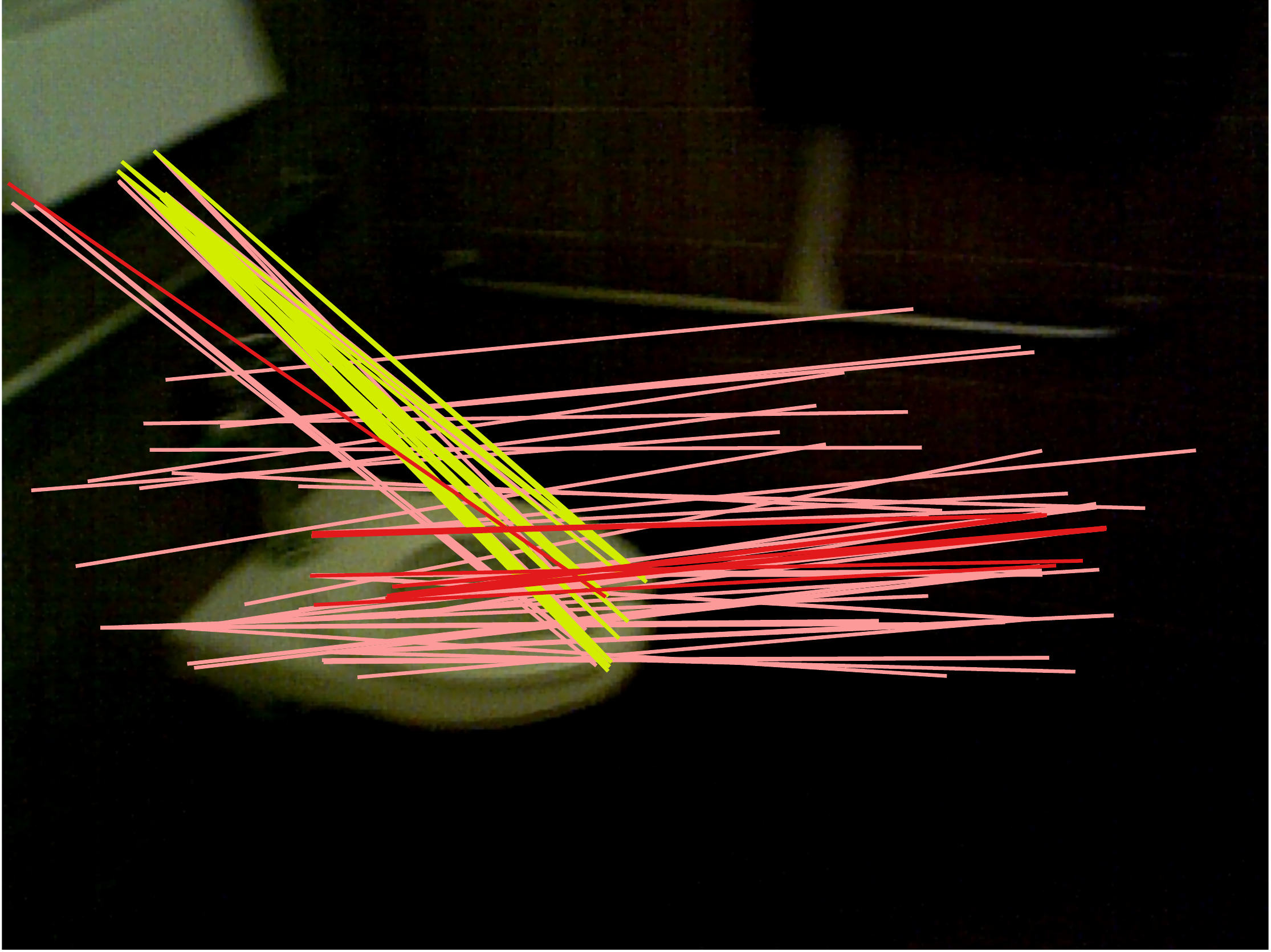}
	\includegraphics[height=7.5em]{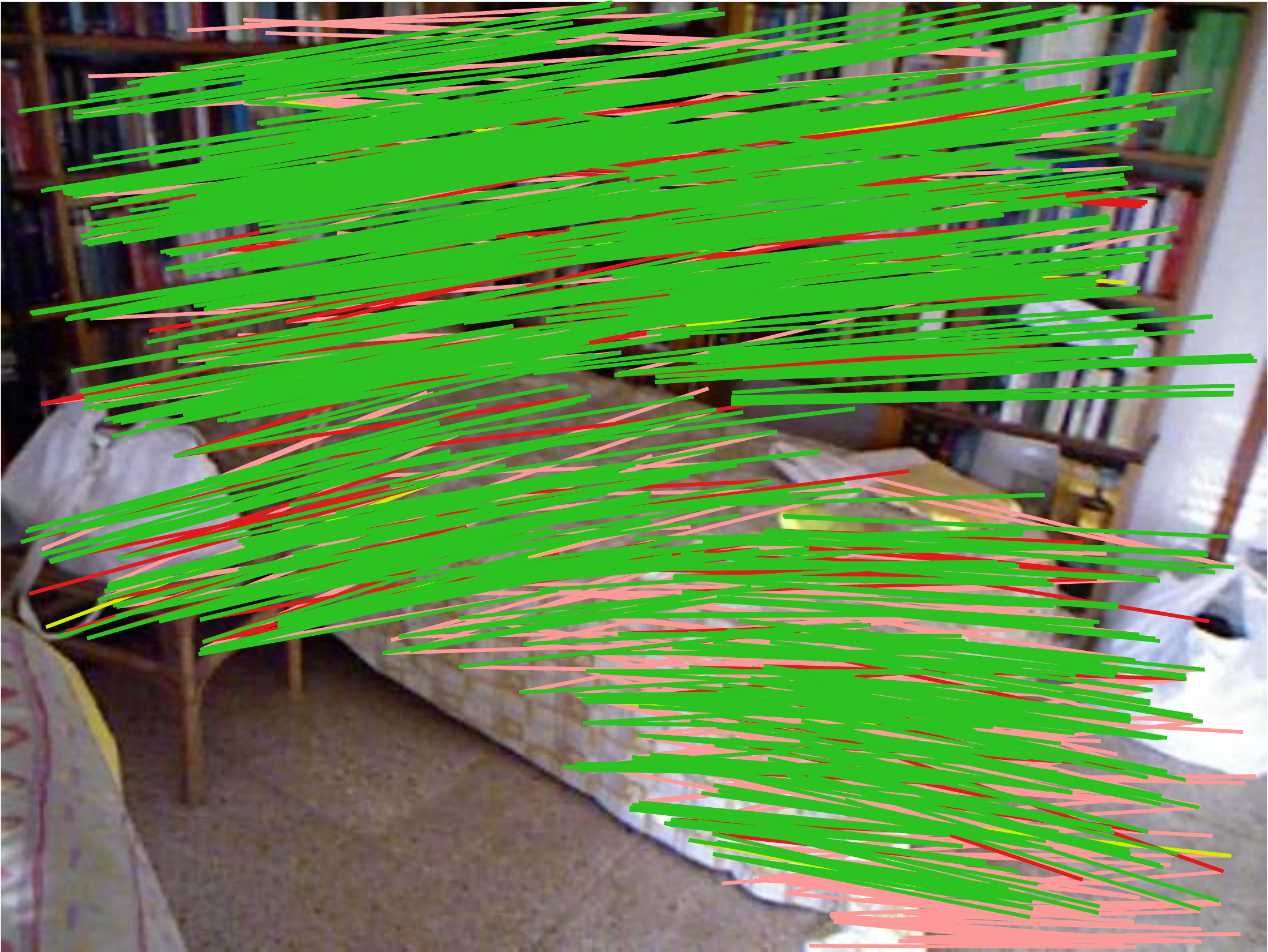}
	\includegraphics[height=7.5em]{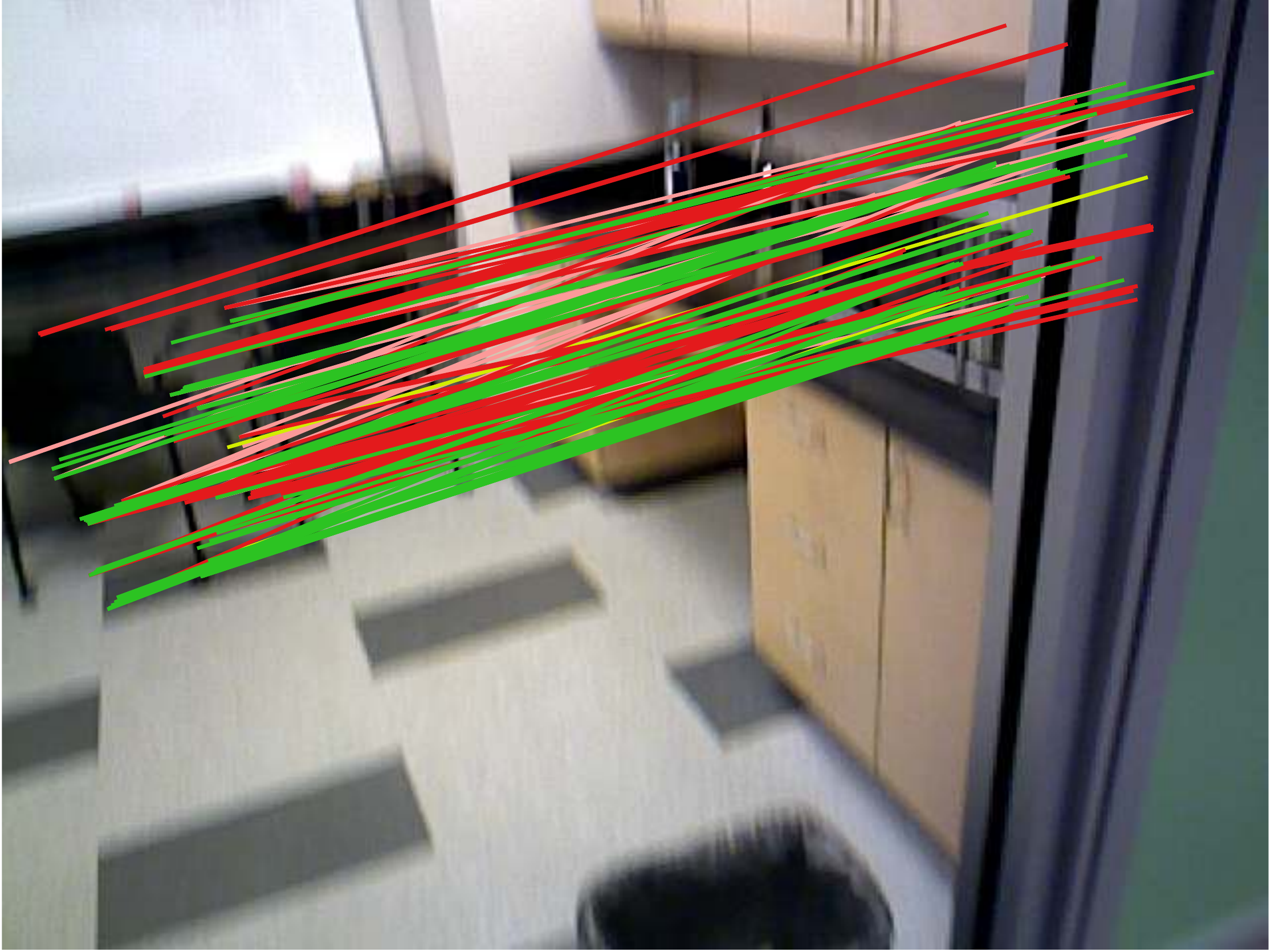}
	\\
	\vspace{0.5em}
	\rotatebox[origin=l]{90}{\mbox{\hspace{1em}AdaLAM}}
	\includegraphics[height=7.5em]{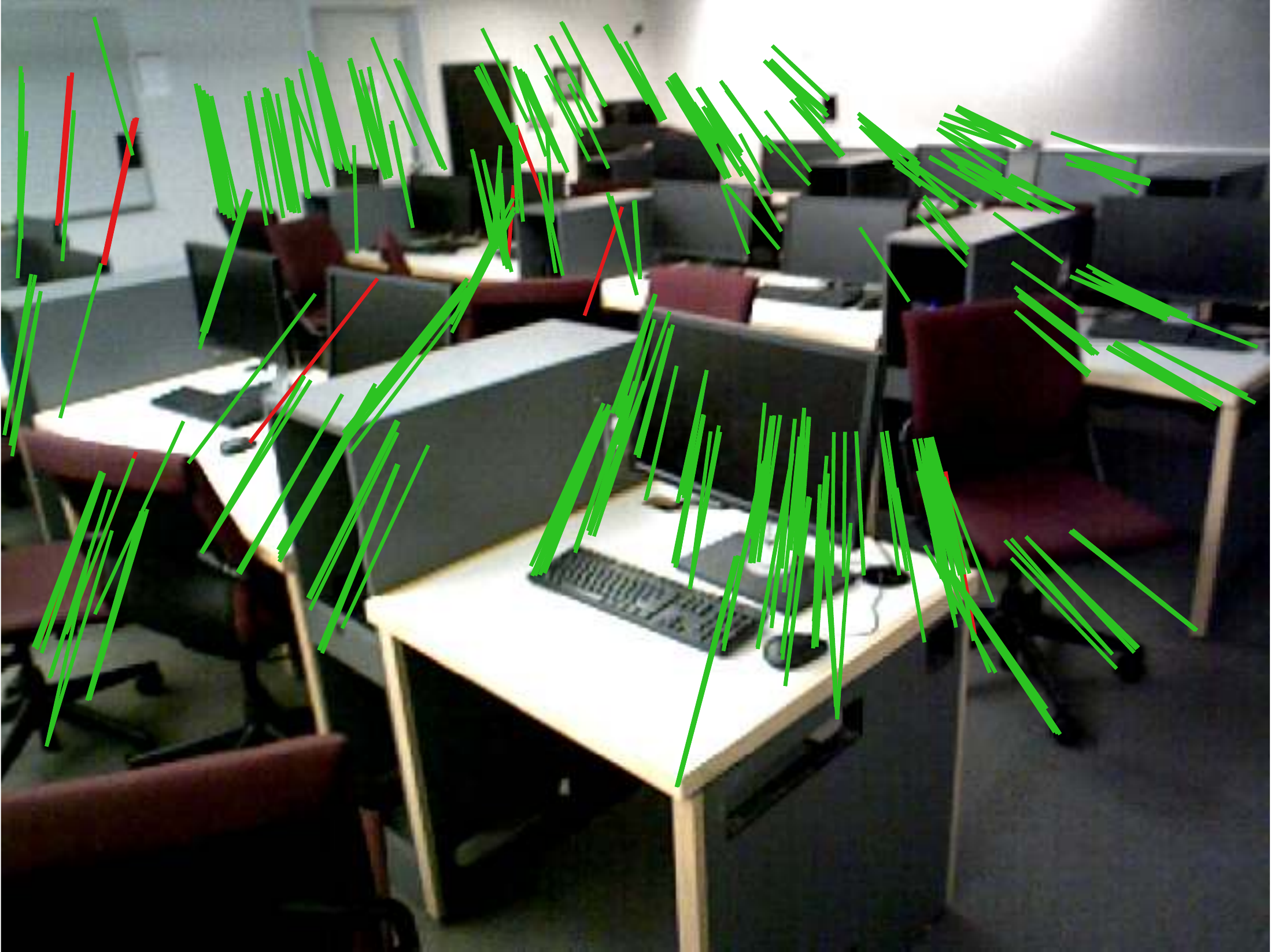}
	\includegraphics[height=7.5em]{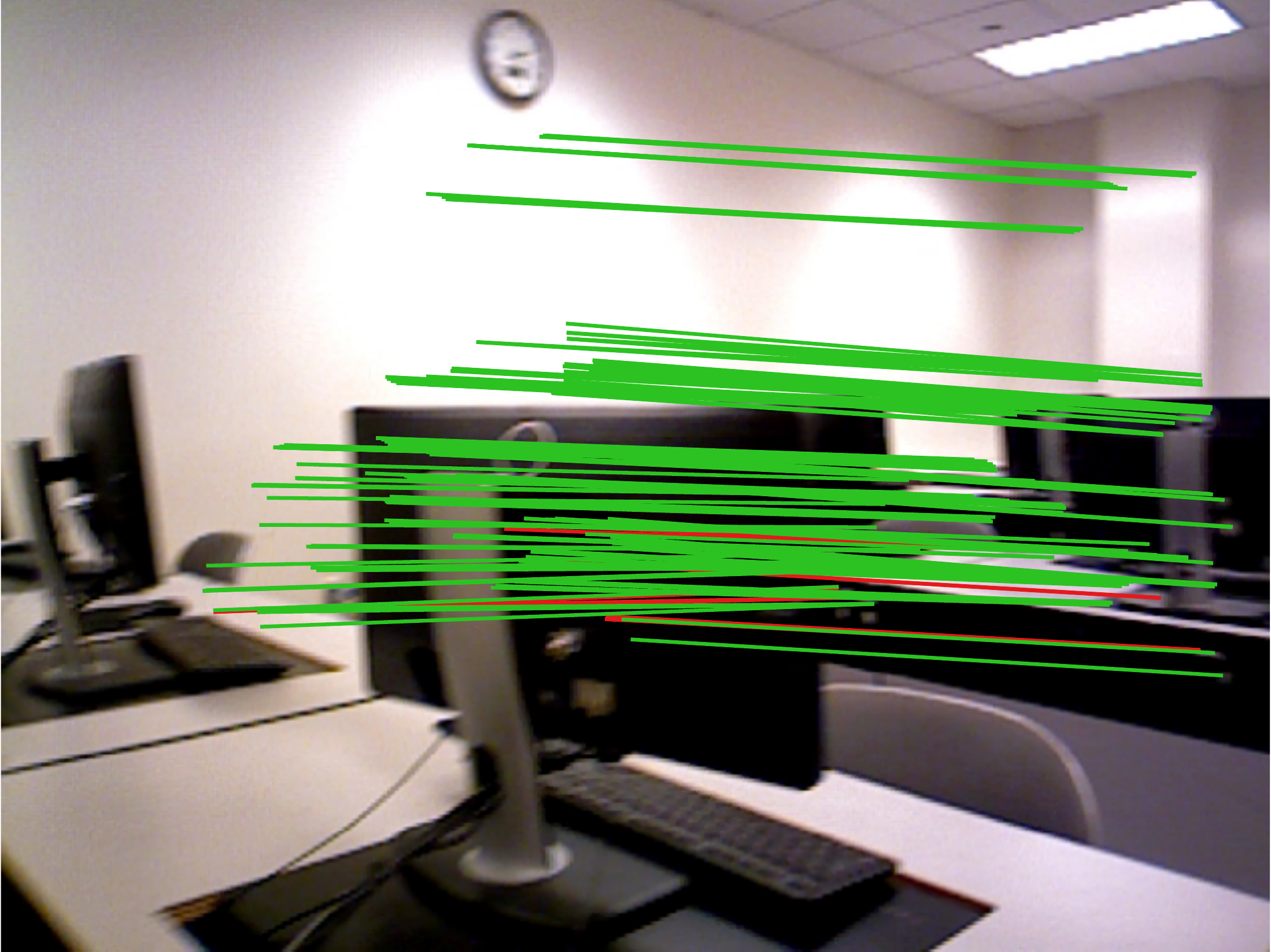}
	\includegraphics[height=7.5em]{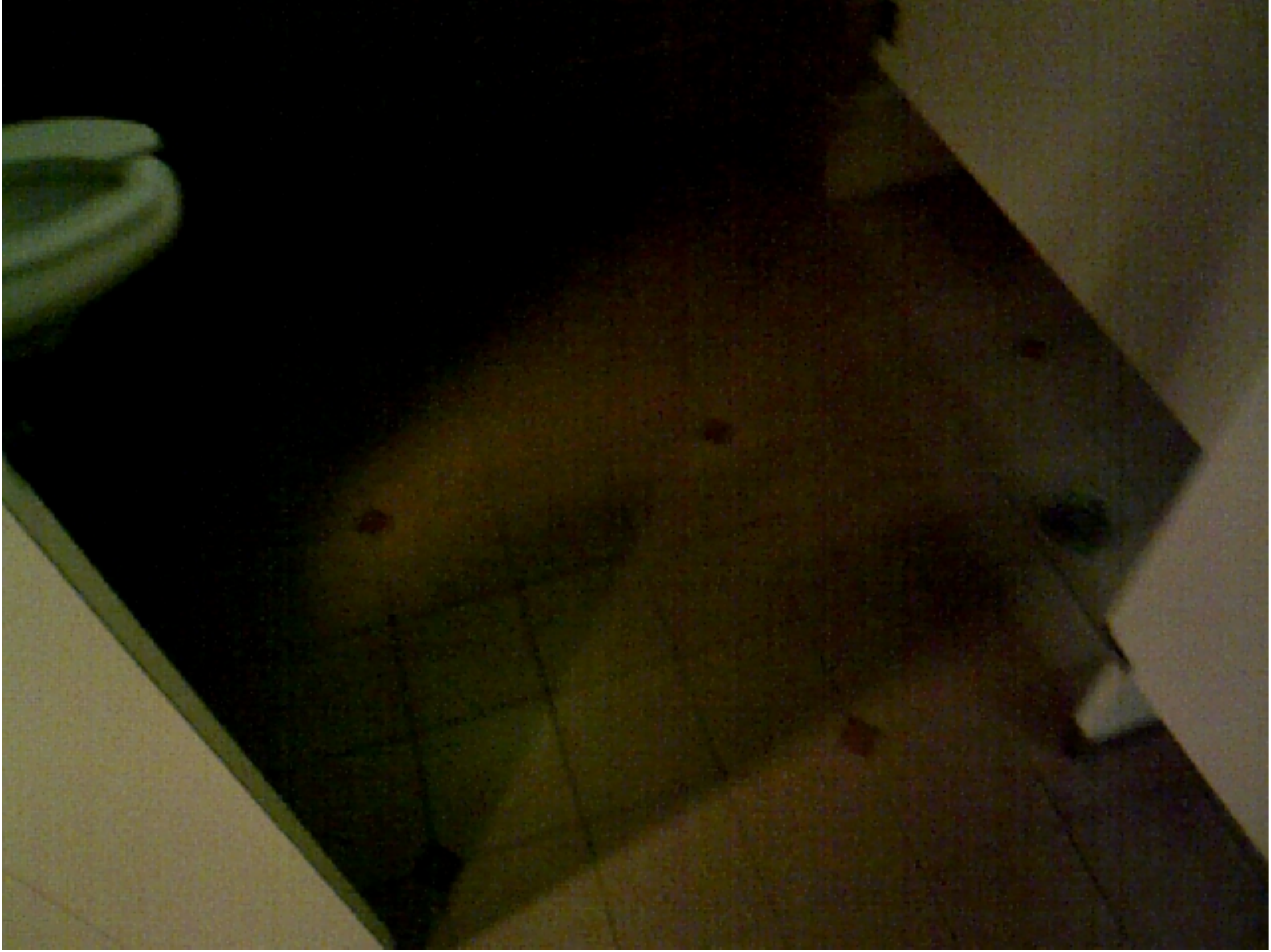}
	\includegraphics[height=7.5em]{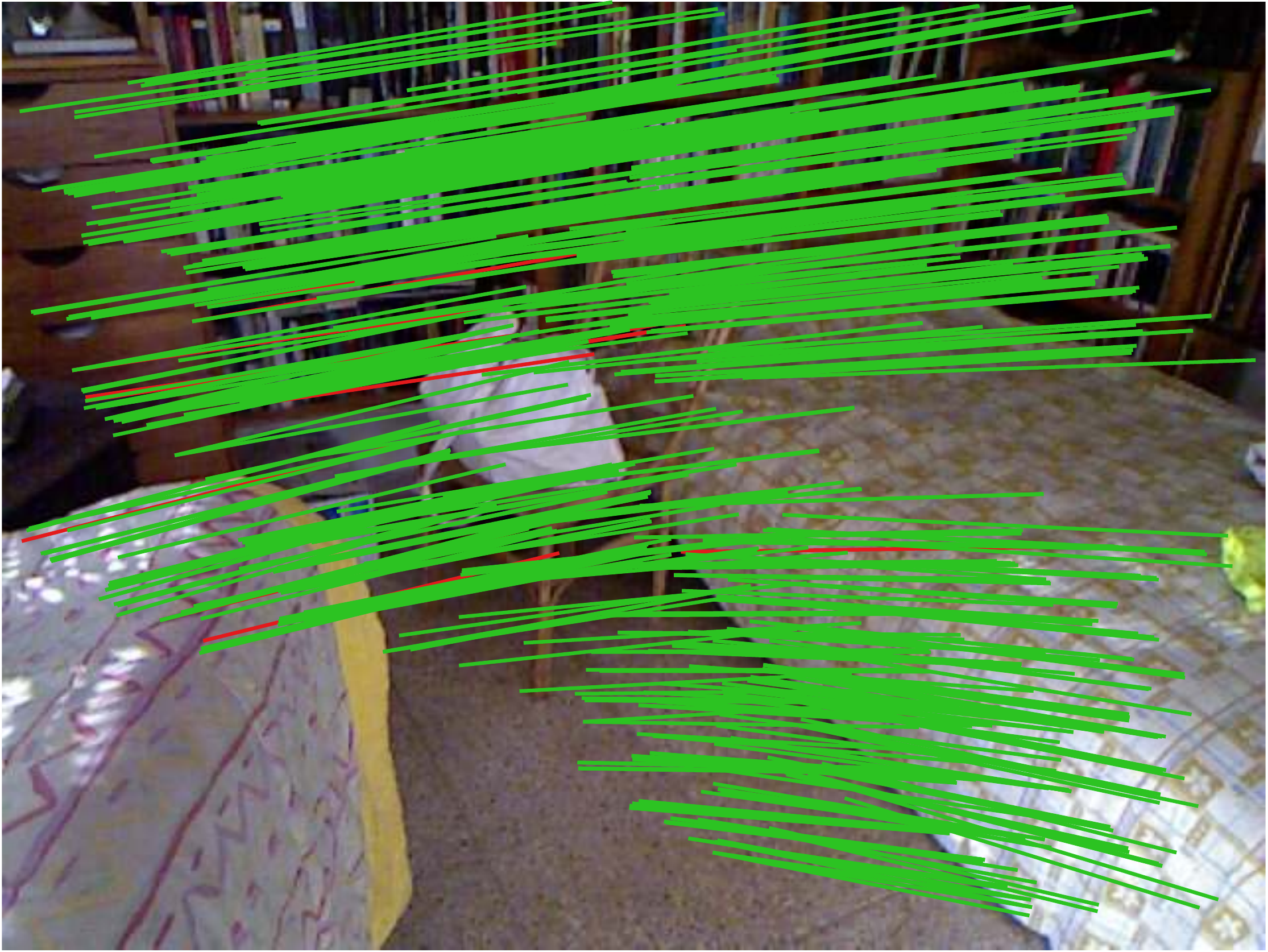}
	\includegraphics[height=7.5em]{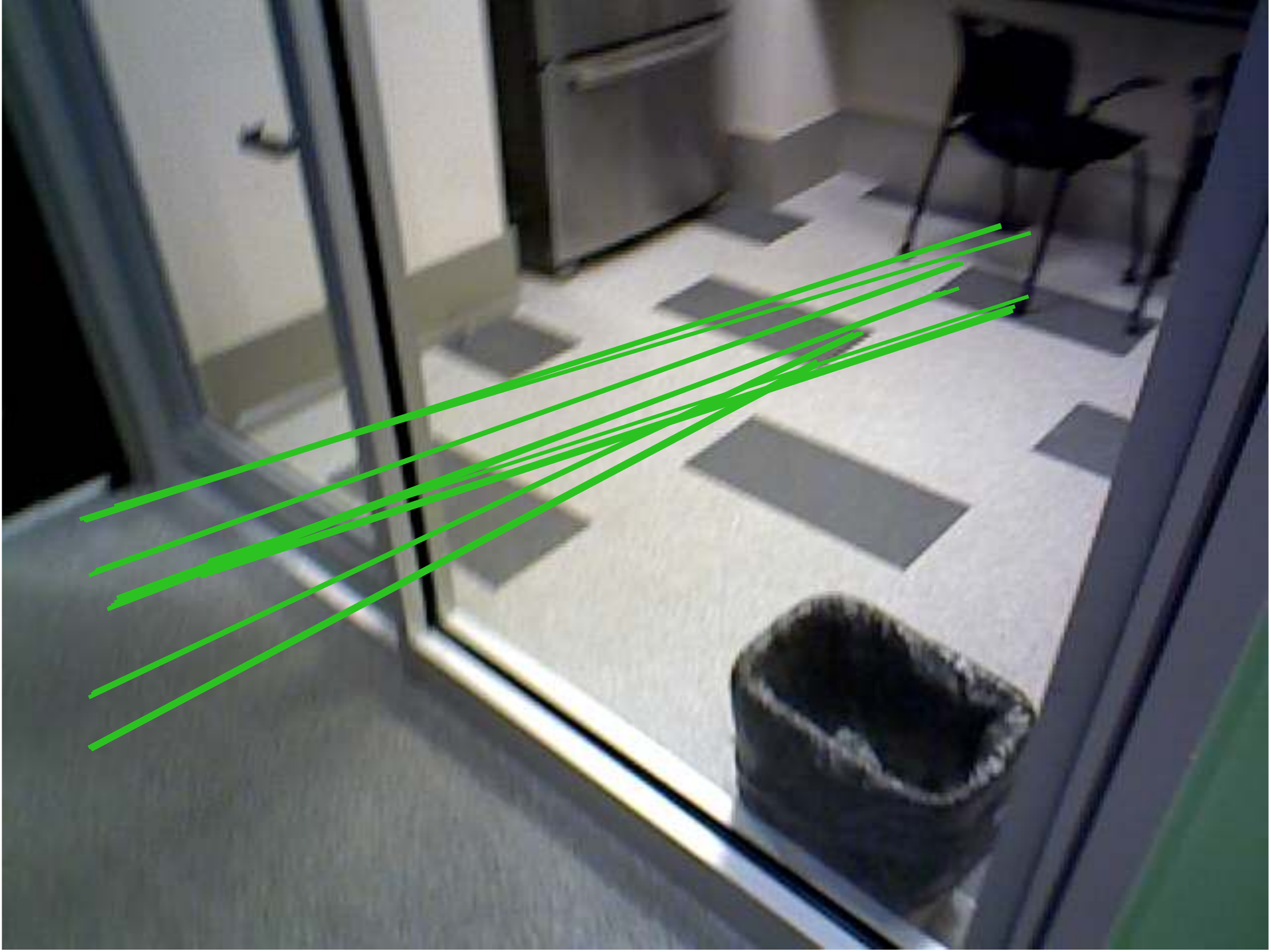}
	\\
	\vspace{0.5em}
	\rotatebox[origin=l]{90}{\mbox{\hspace{2em}OANet}}
	\includegraphics[height=7.5em]{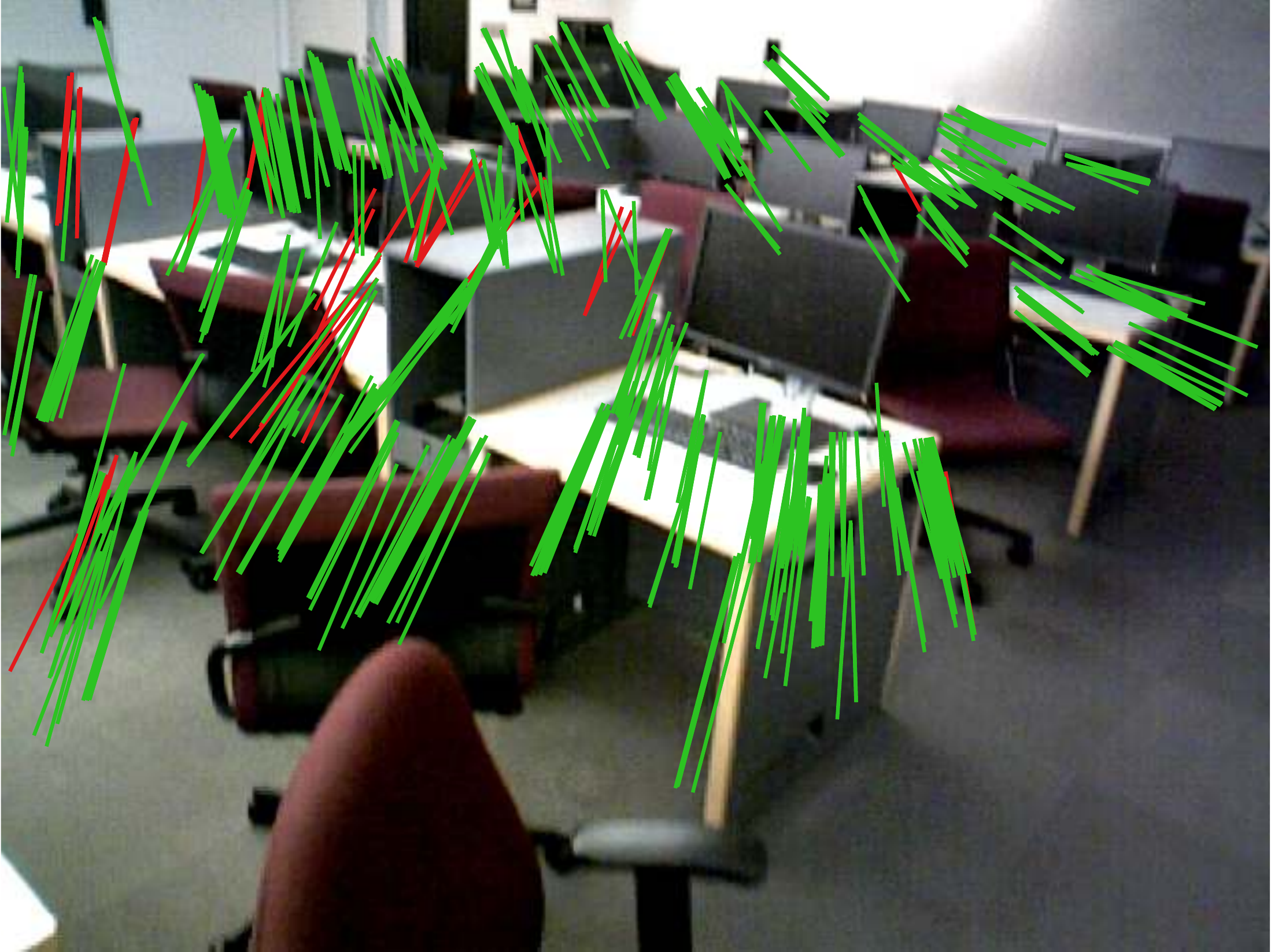}
	\includegraphics[height=7.5em]{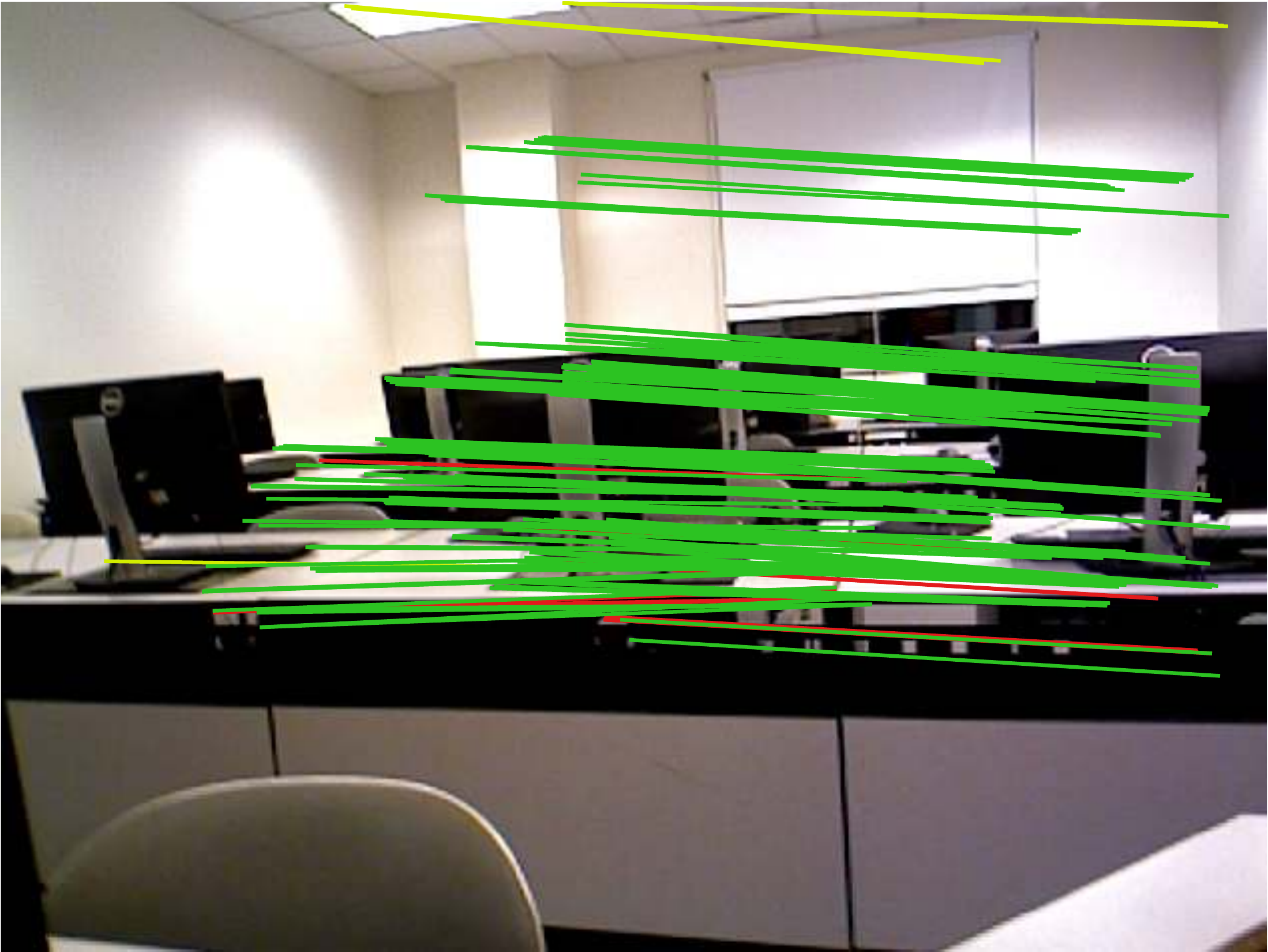}
	\includegraphics[height=7.5em]{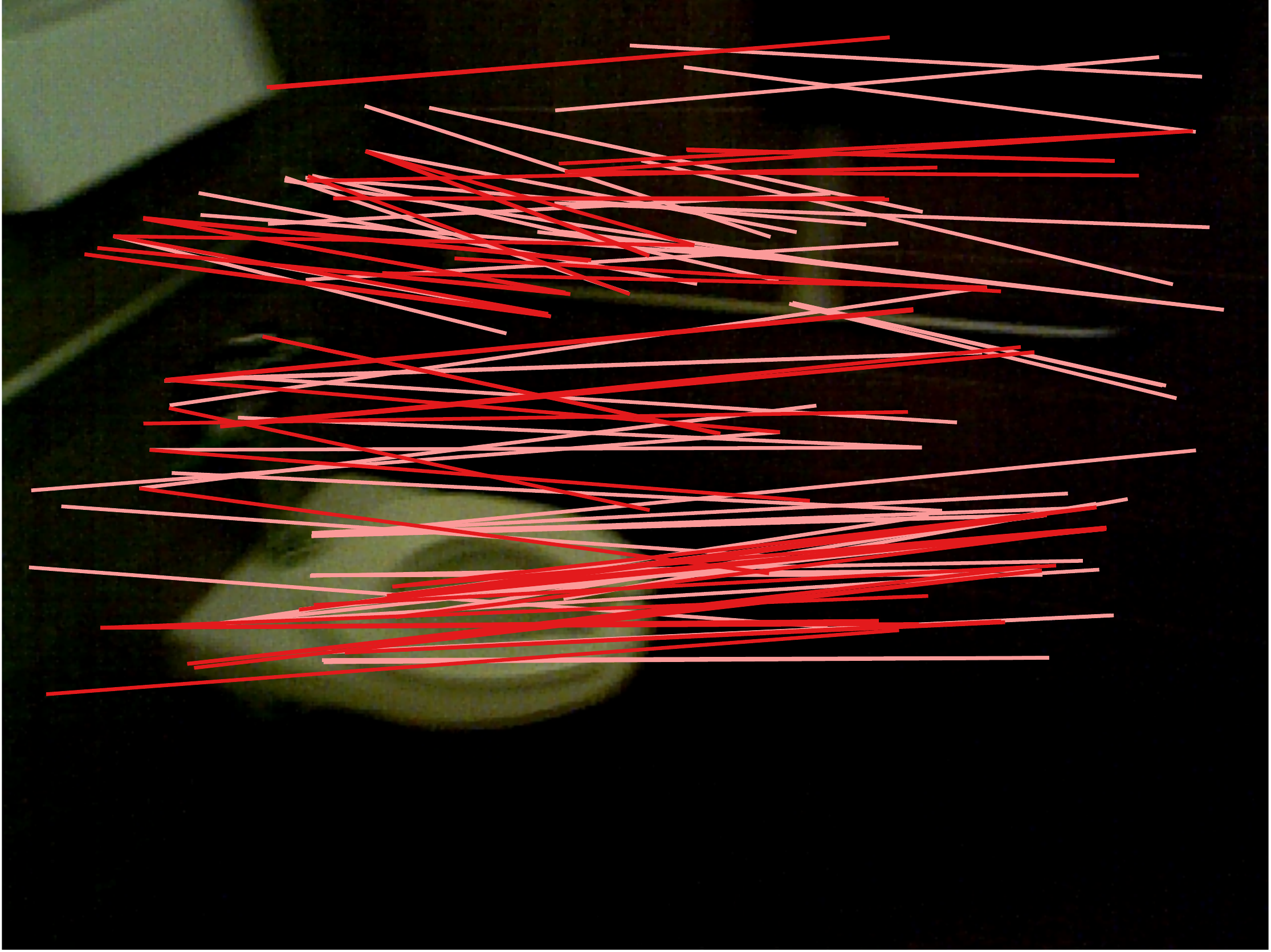}
	\includegraphics[height=7.5em]{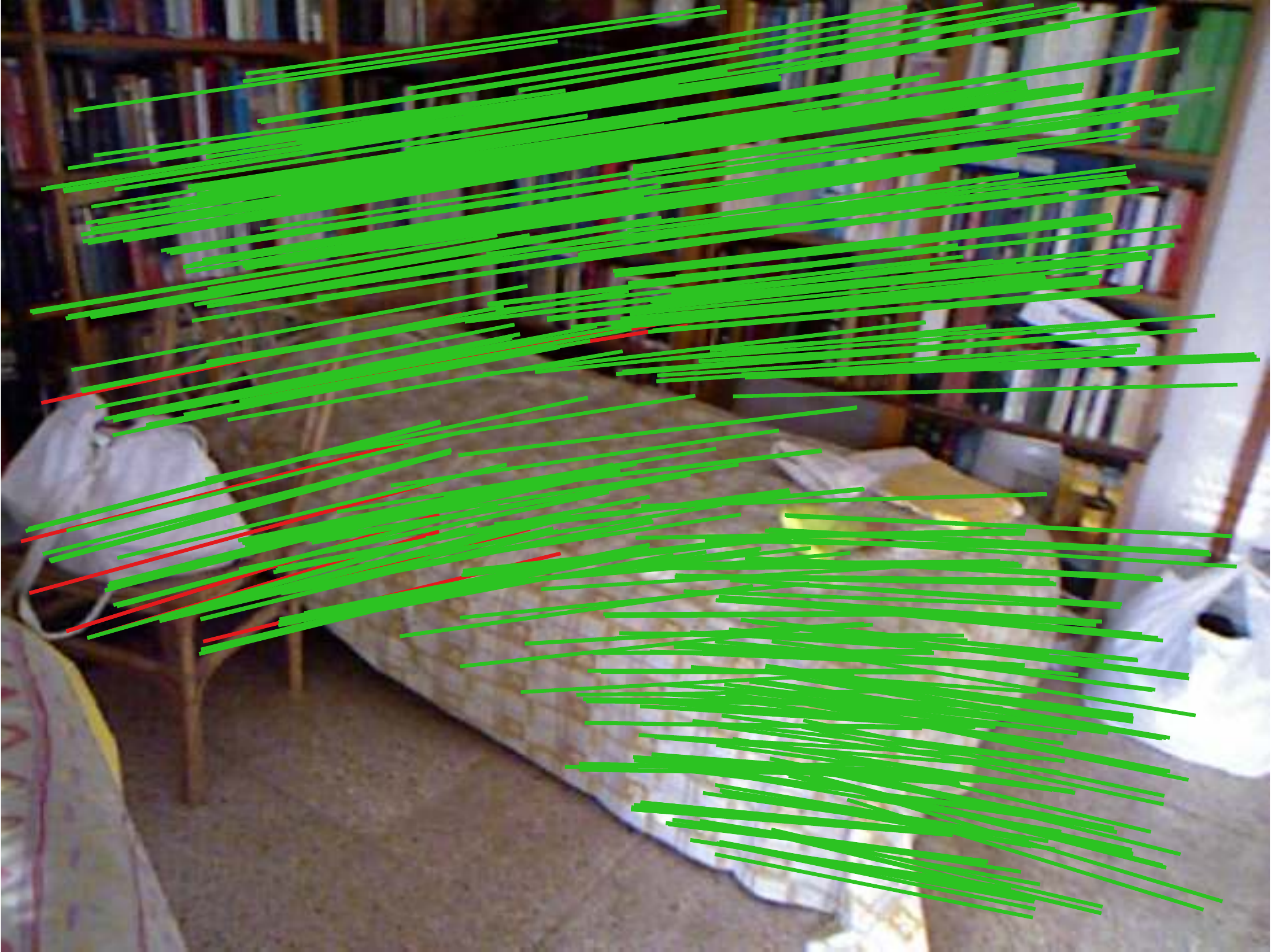}
	\includegraphics[height=7.5em]{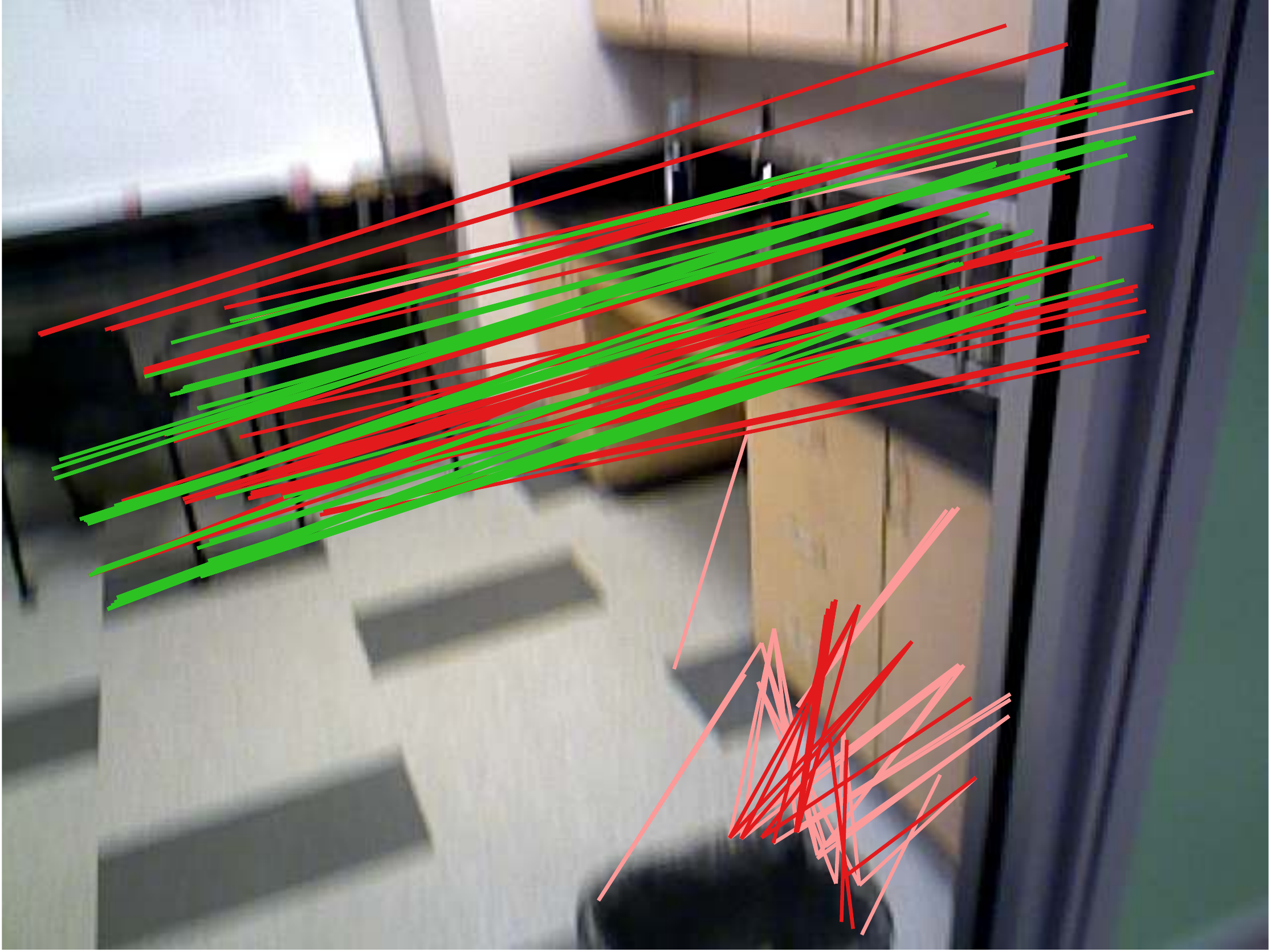}
	\\
	\vspace{0.5em}
	\rotatebox[origin=l]{90}{\mbox{\hspace{2em}ACNe$^\star$}}
	\includegraphics[height=7.5em]{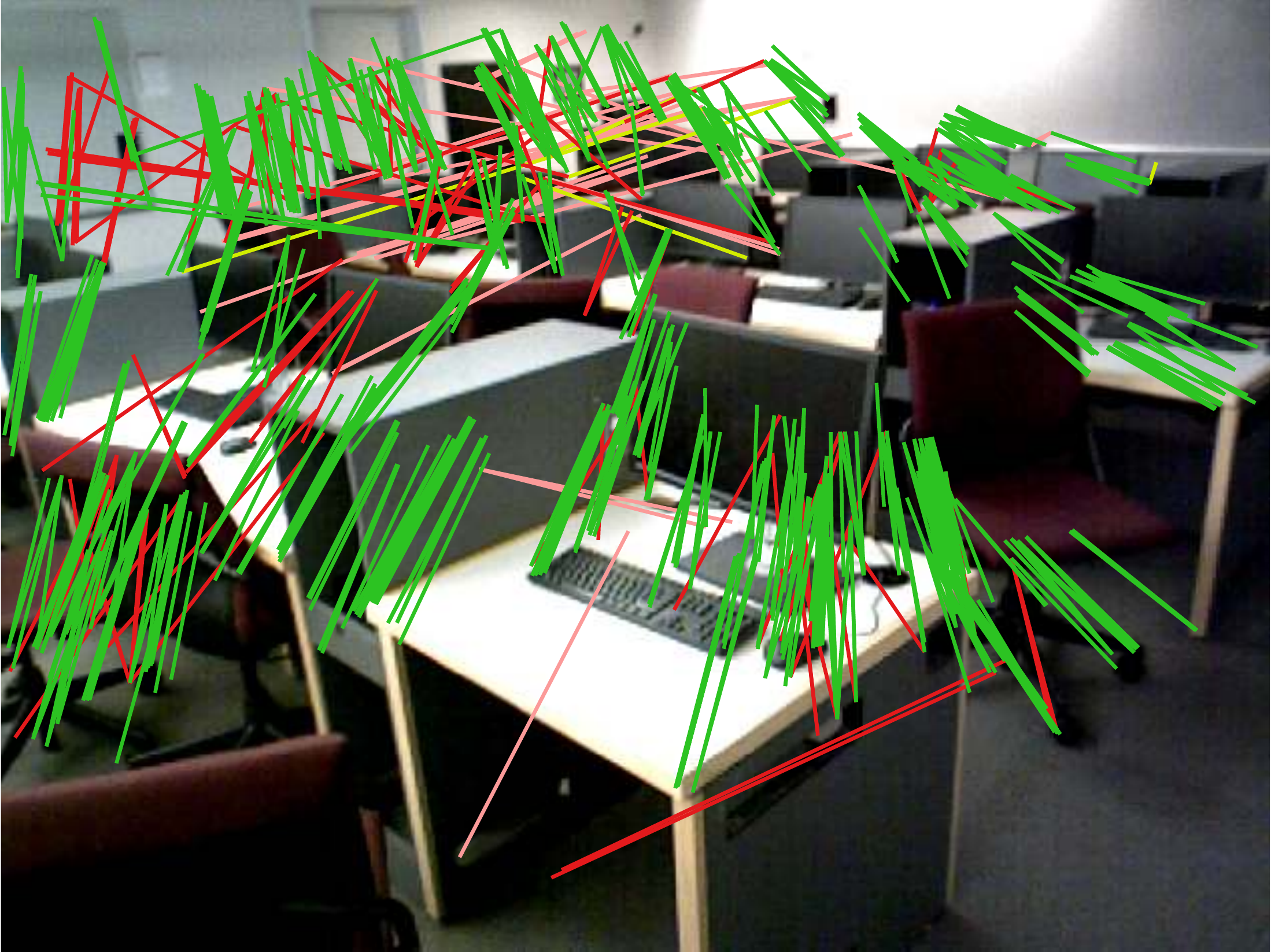}
	\includegraphics[height=7.5em]{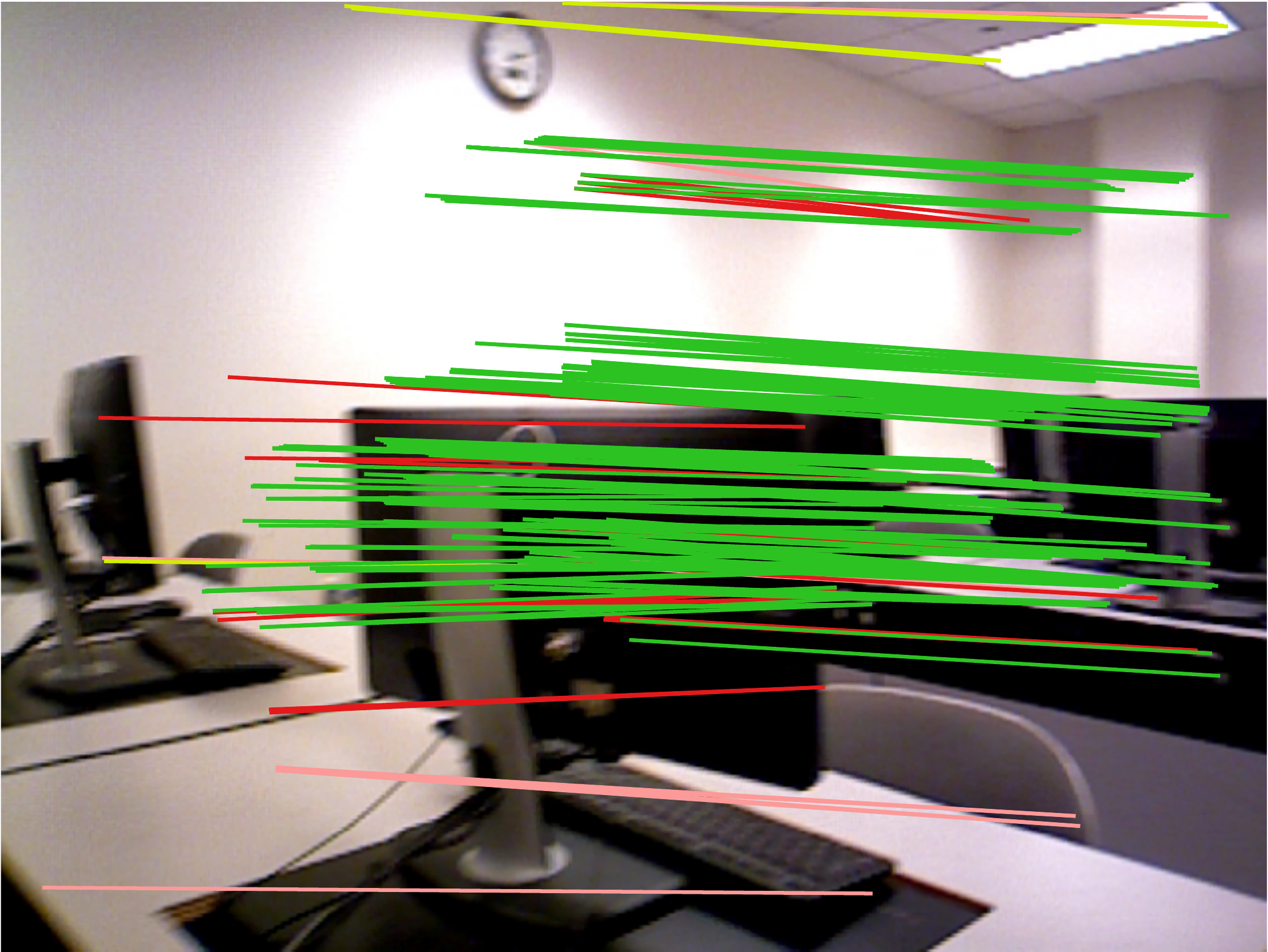}
	\includegraphics[height=7.5em]{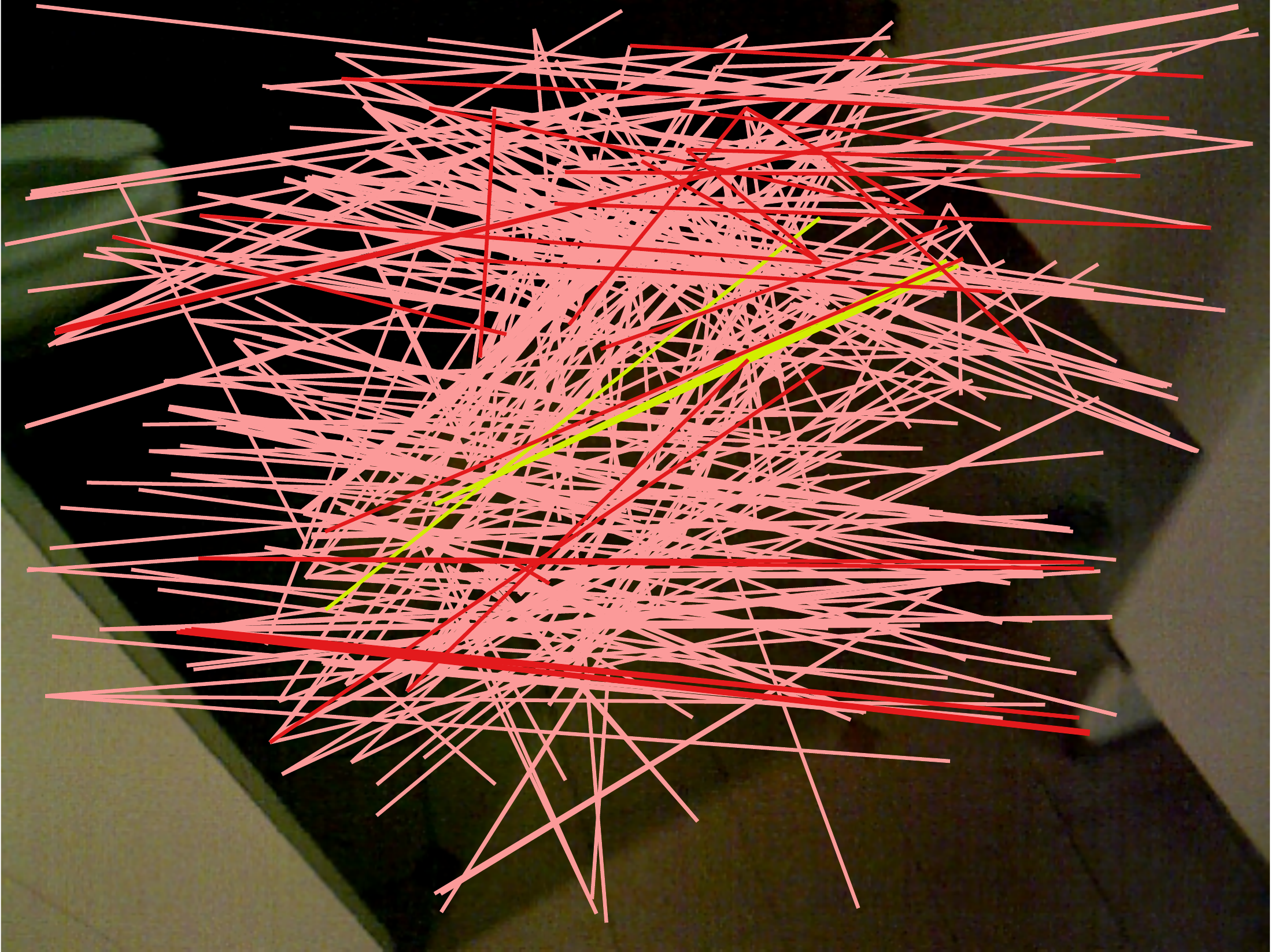}
	\includegraphics[height=7.5em]{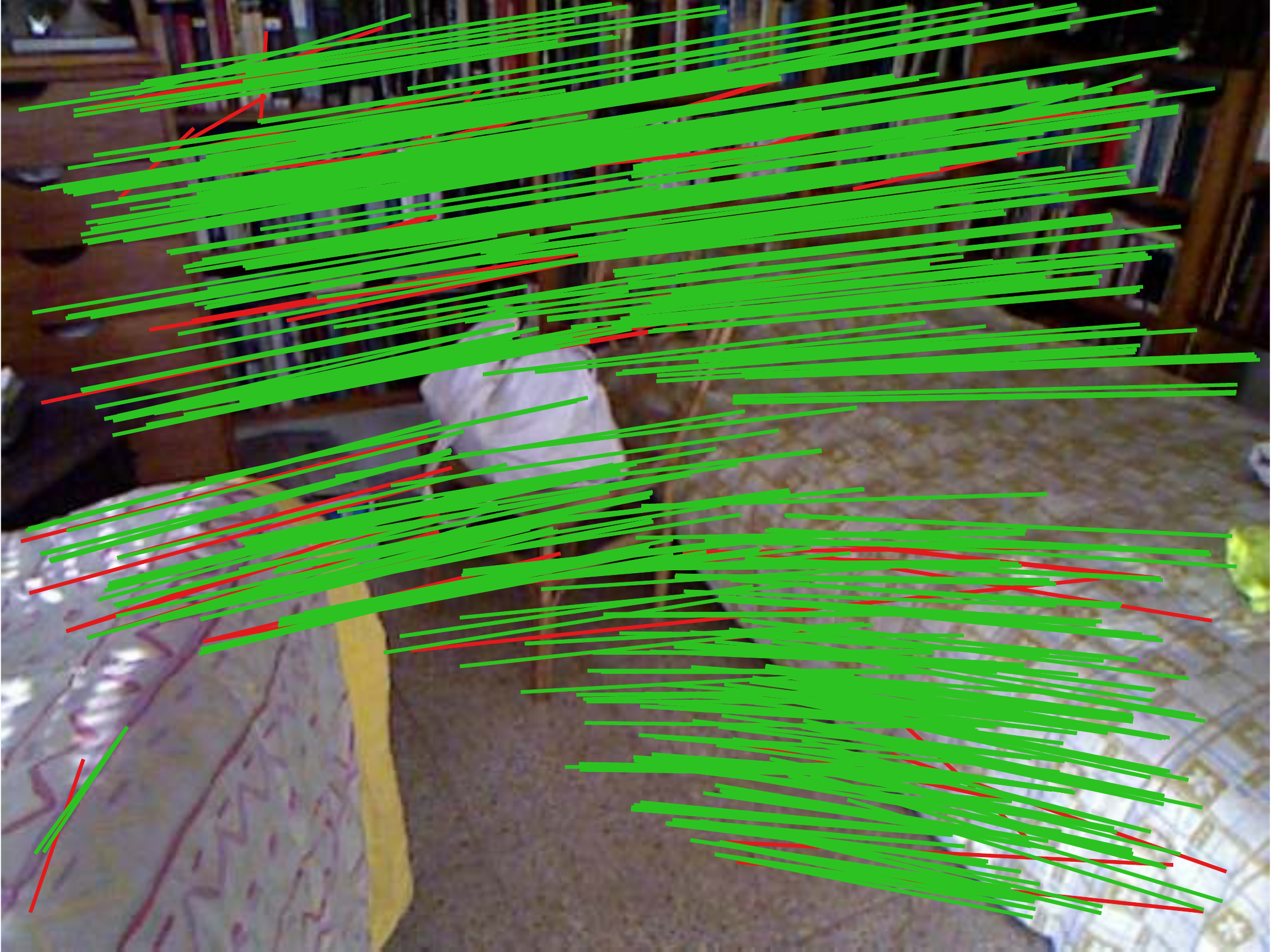}
	\includegraphics[height=7.5em]{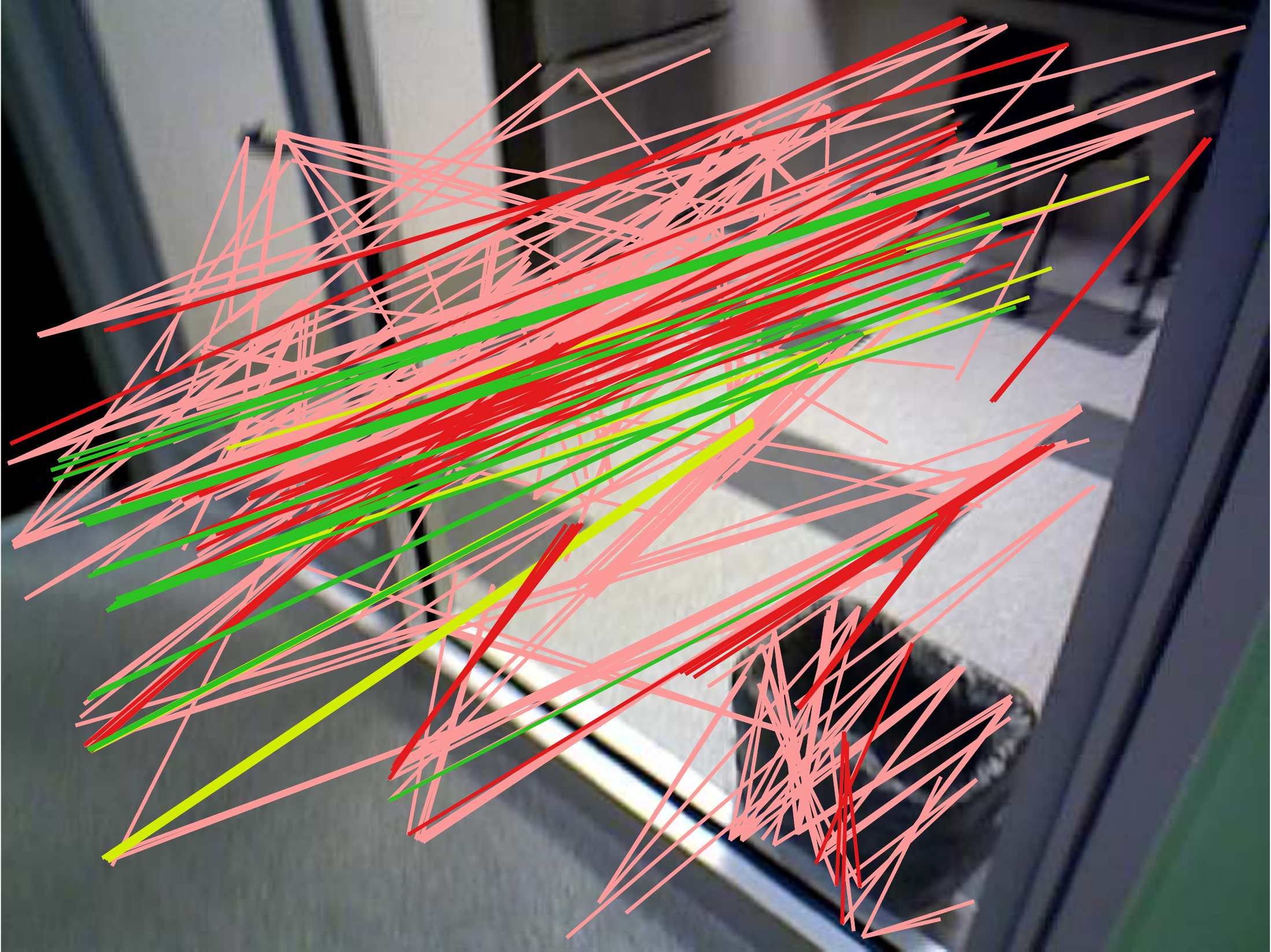}
	\\
	\vspace{0.5em}
	\rotatebox[origin=l]{90}{\mbox{\hspace{2em}PFM}}
	\includegraphics[height=7.5em]{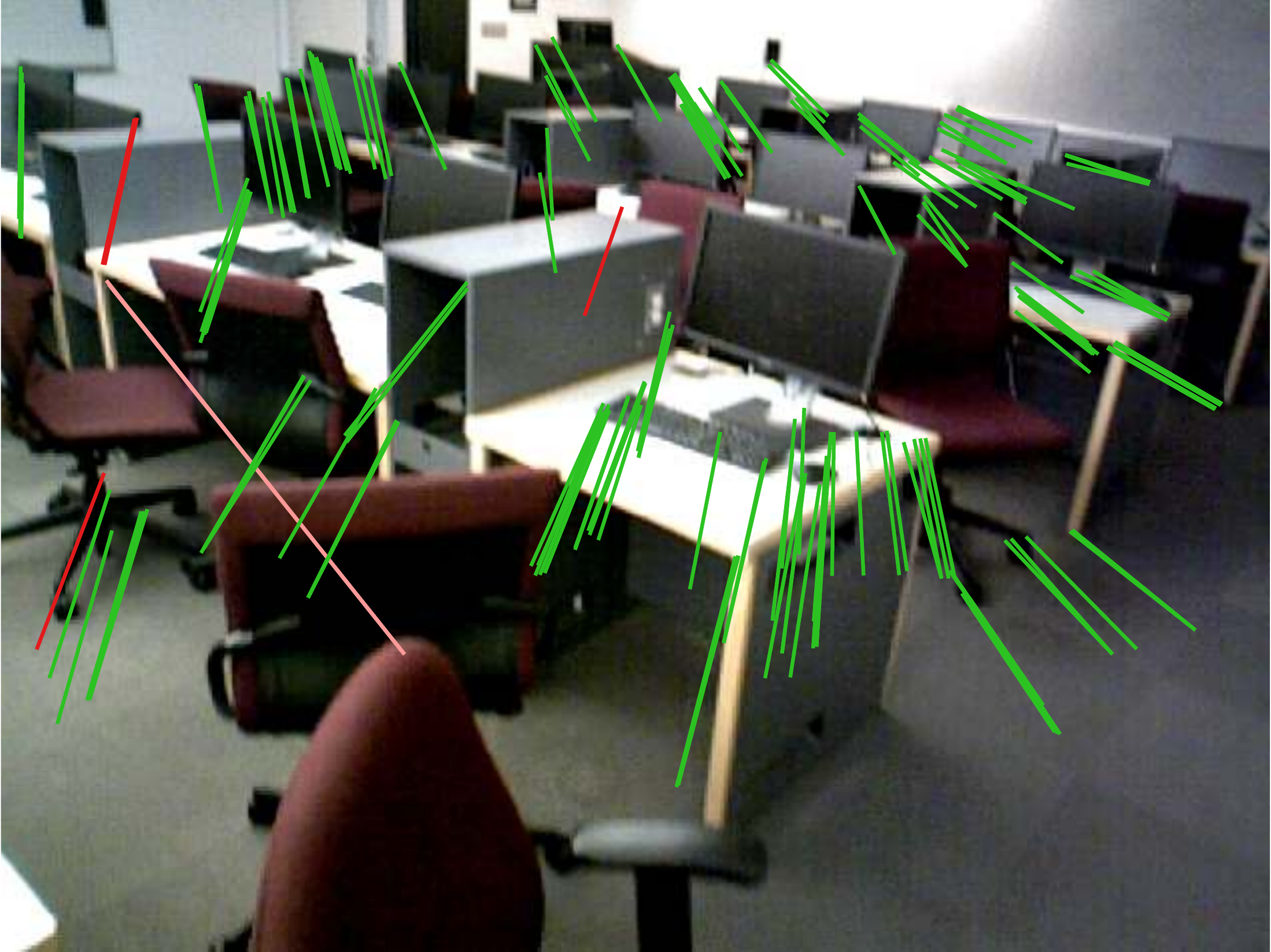}
	\includegraphics[height=7.5em]{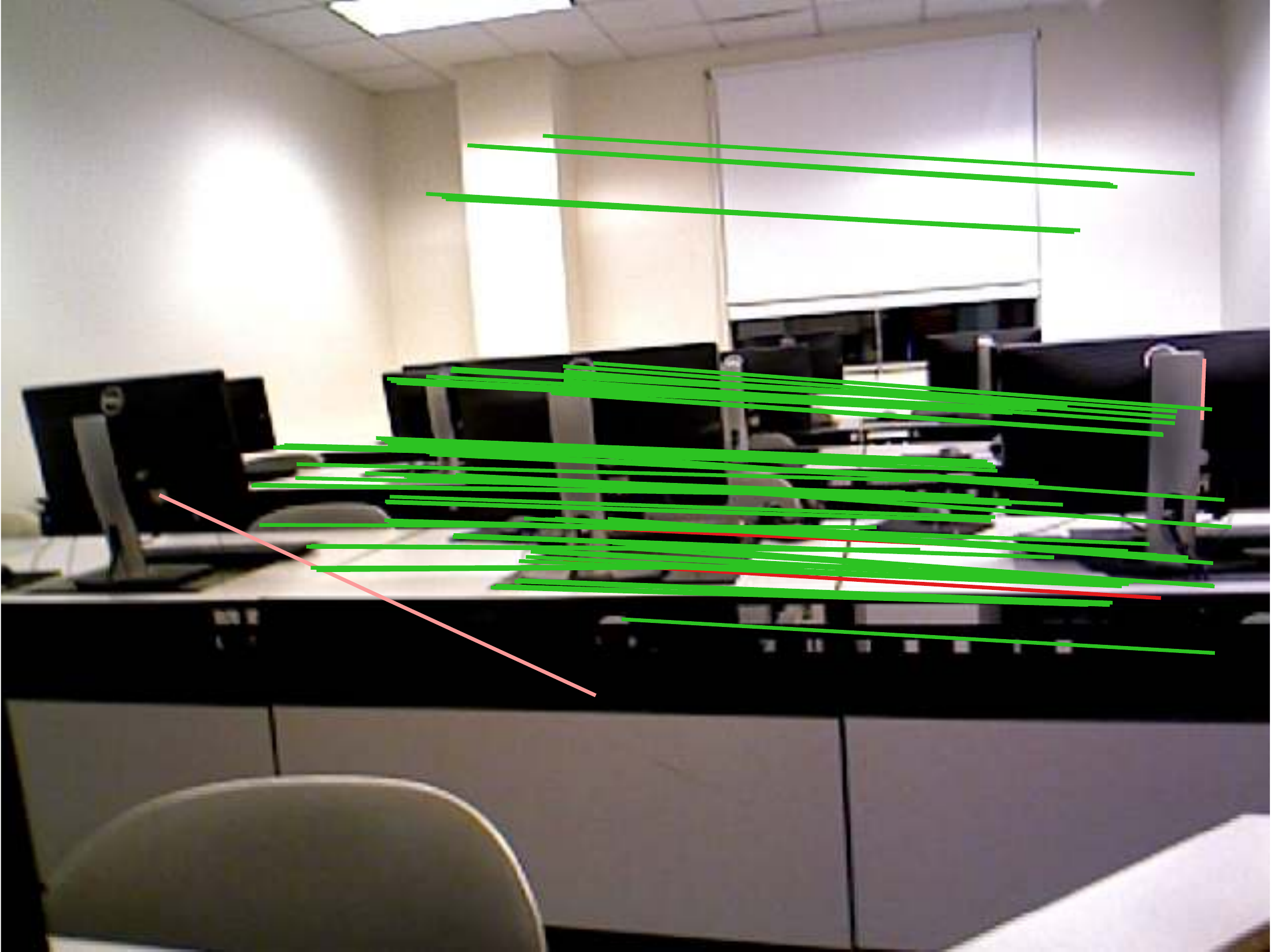}
	\includegraphics[height=7.5em]{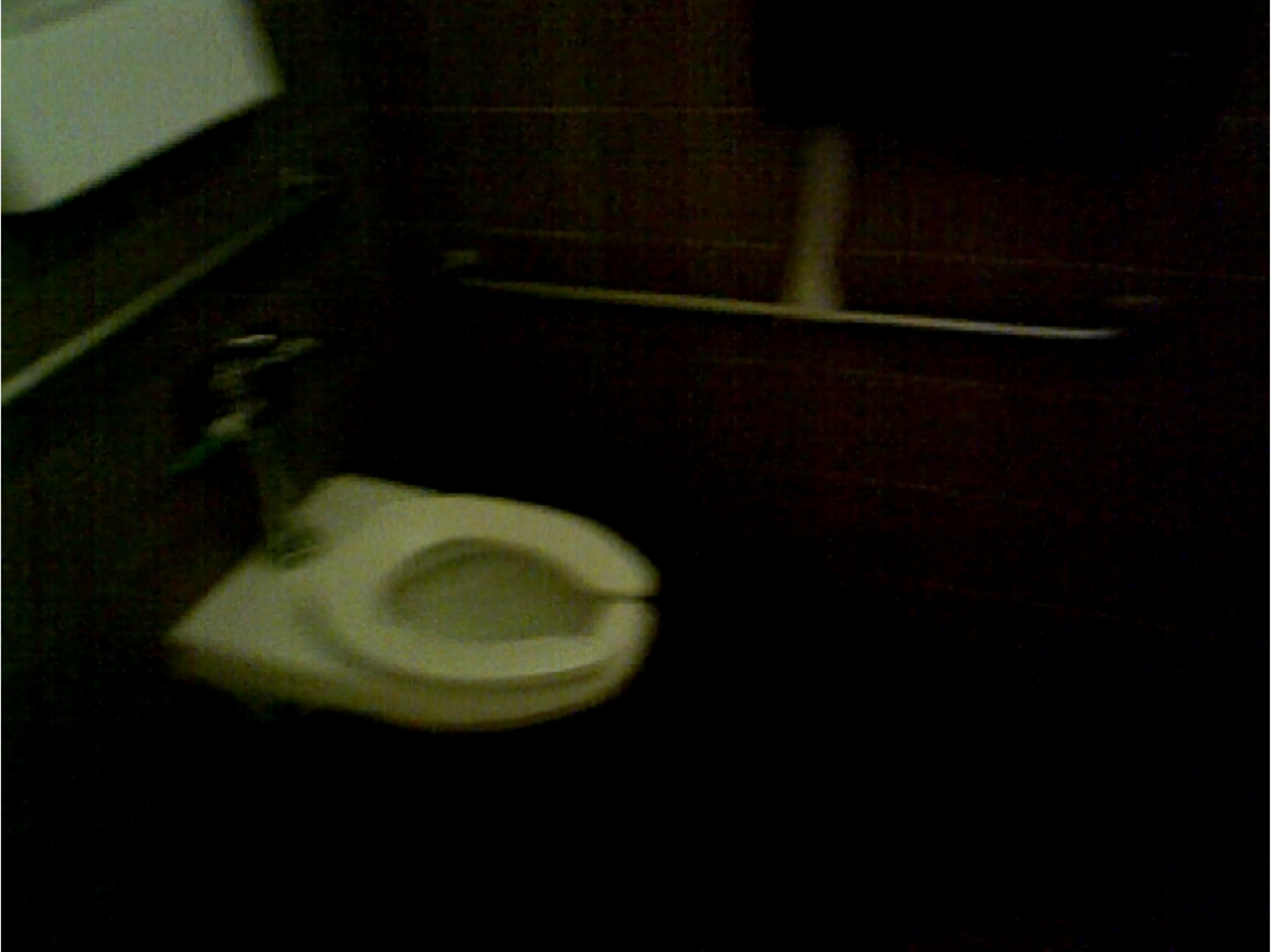}
	\includegraphics[height=7.5em]{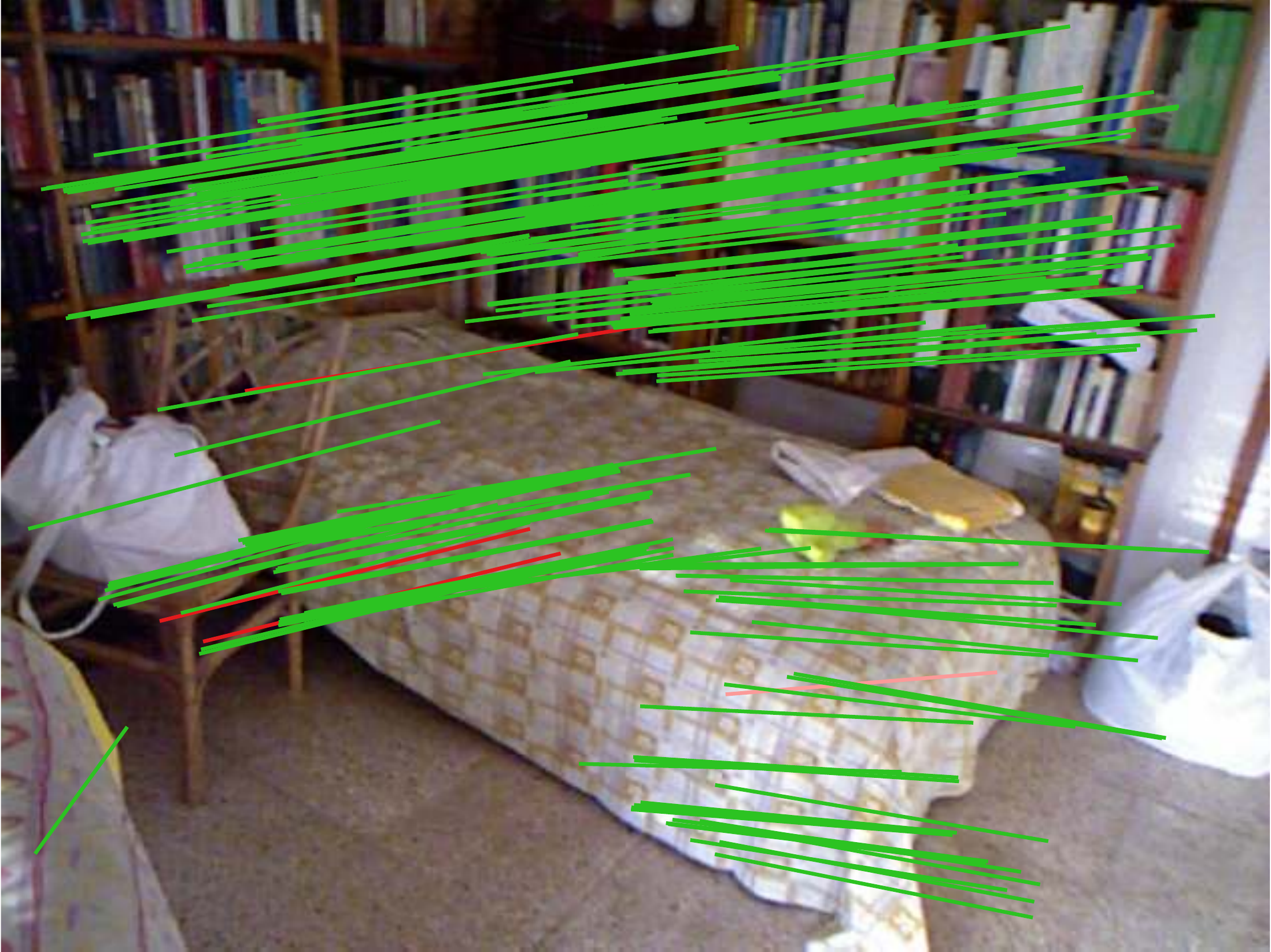}
	\includegraphics[height=7.5em]{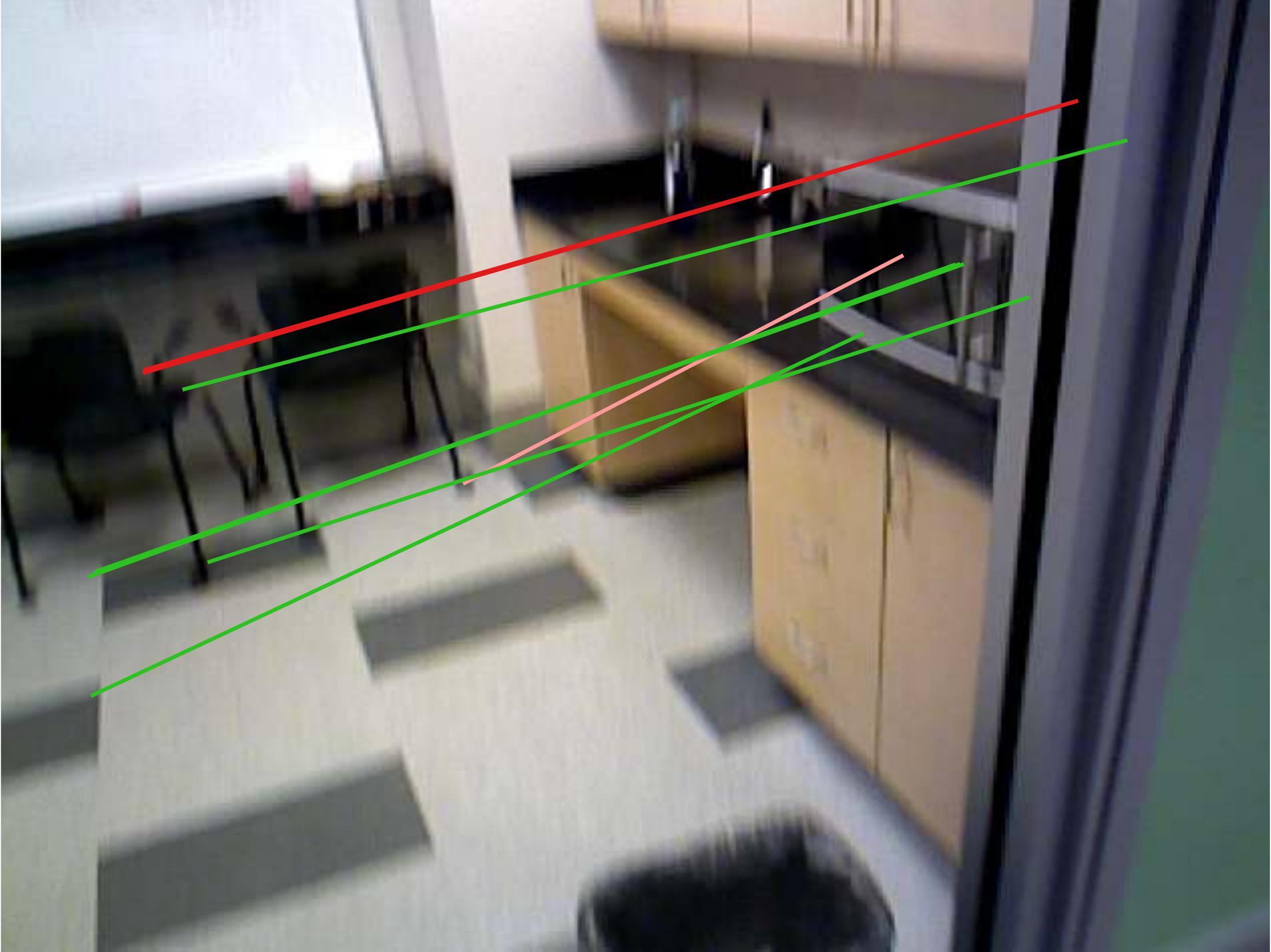}
	\\
	\vspace{0.5em}
	\rotatebox[origin=l]{90}{\mbox{\hspace{2em}PGM}}
	\includegraphics[height=7.5em]{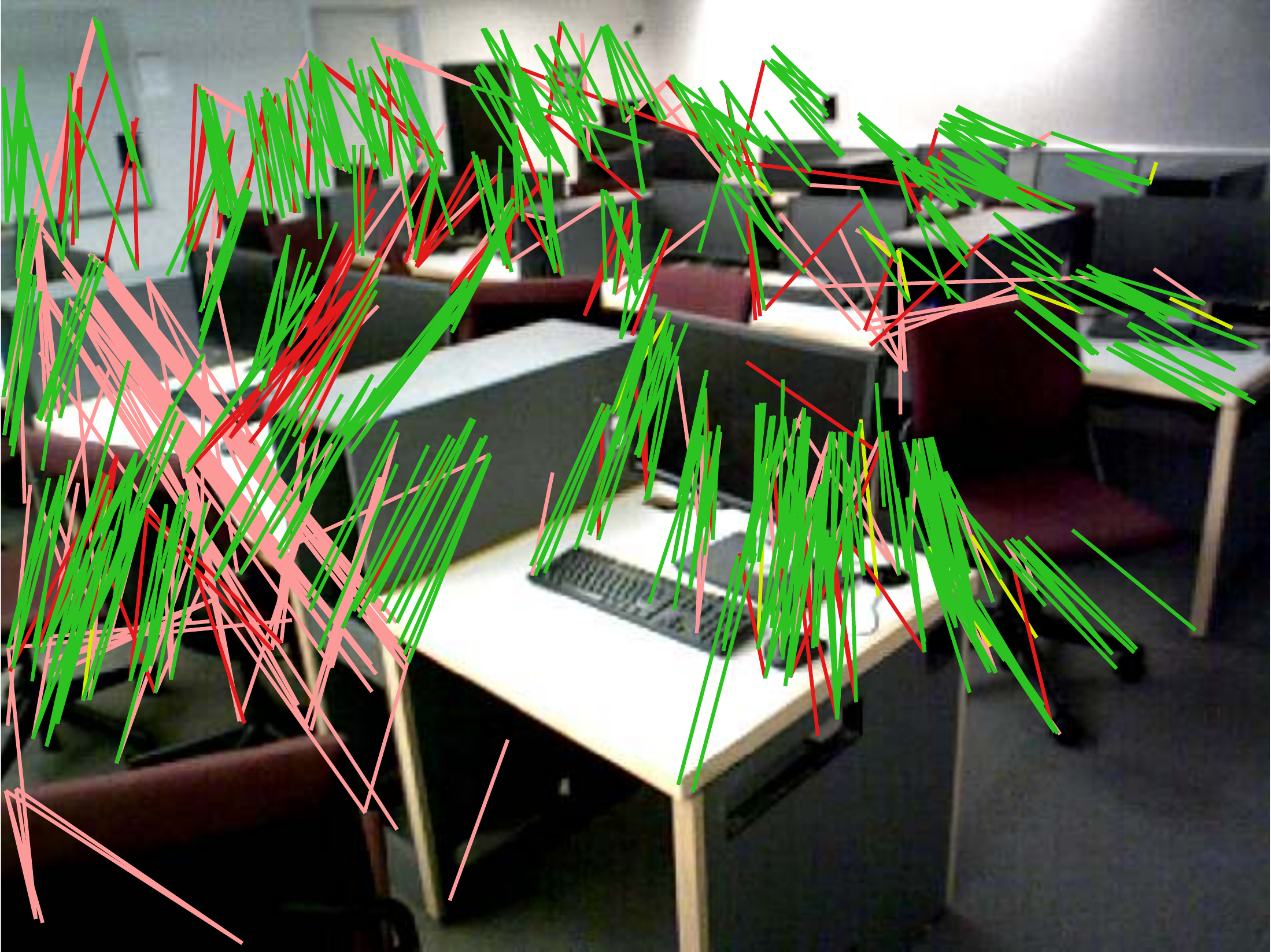}
	\includegraphics[height=7.5em]{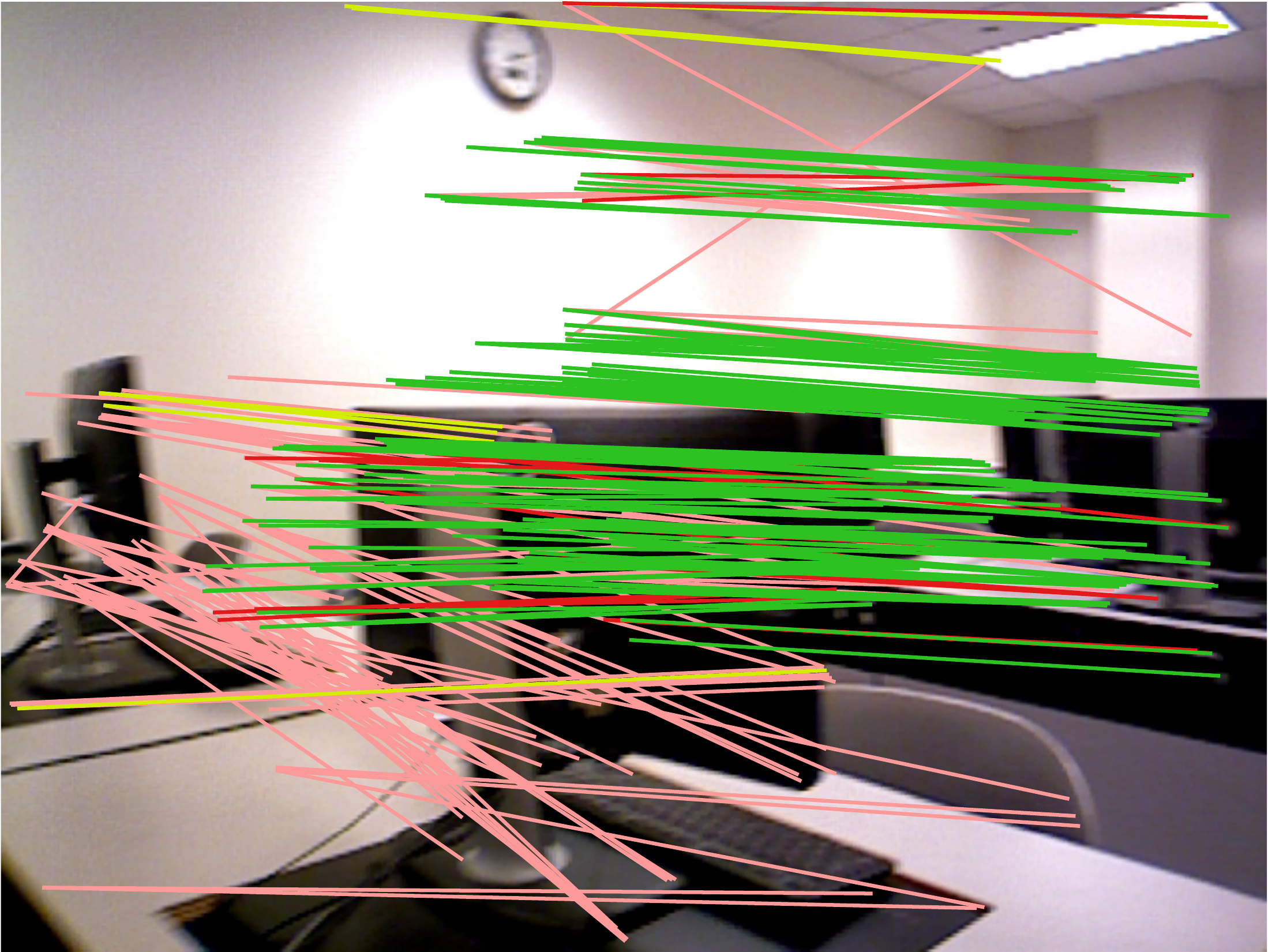}
	\includegraphics[height=7.5em]{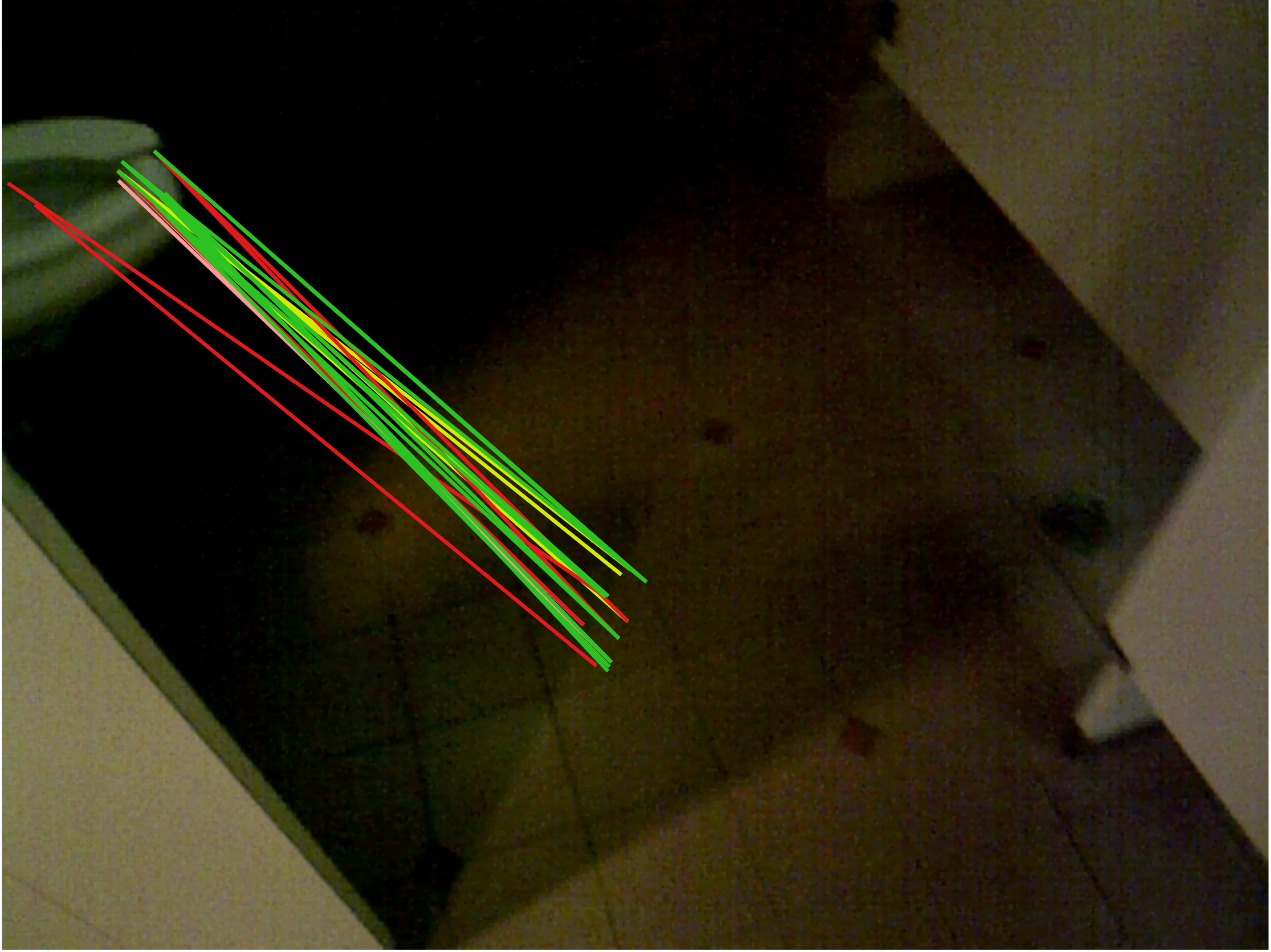}
	\includegraphics[height=7.5em]{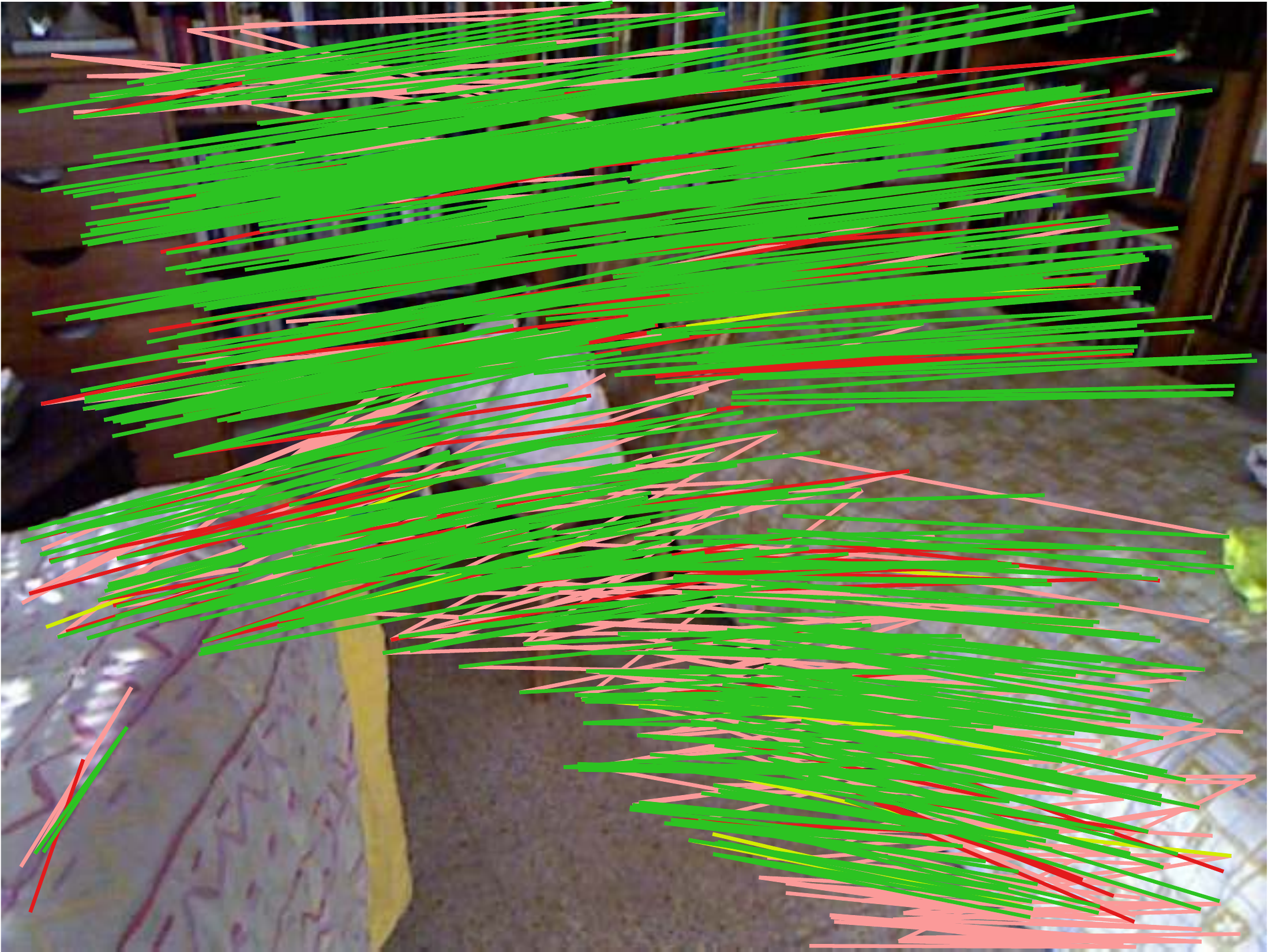}
	\includegraphics[height=7.5em]{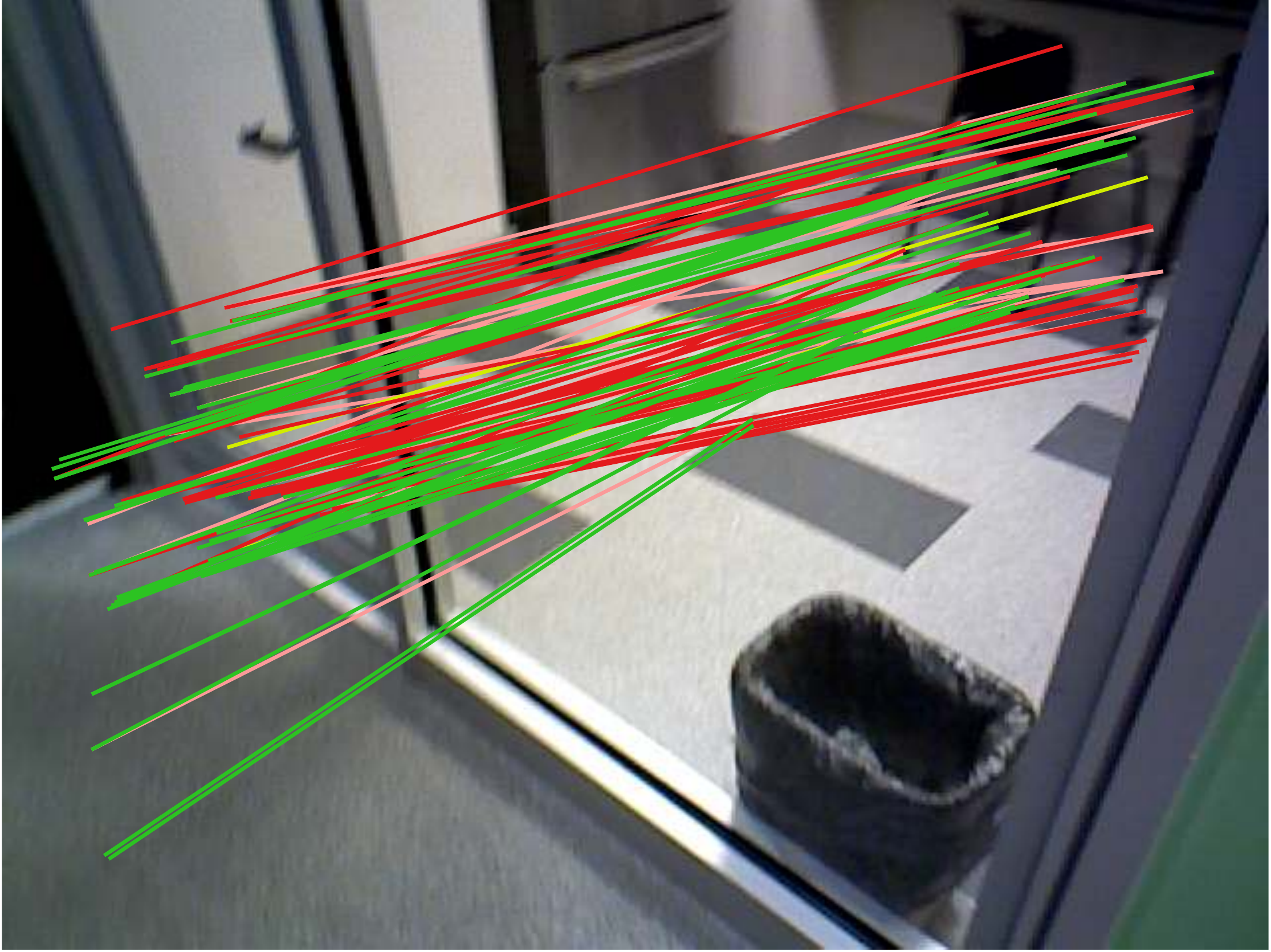}
	\\
	\vspace{0.5em}
	\rotatebox[origin=l]{90}{\mbox{\hspace{2em}SCV}}
	\includegraphics[height=7.5em]{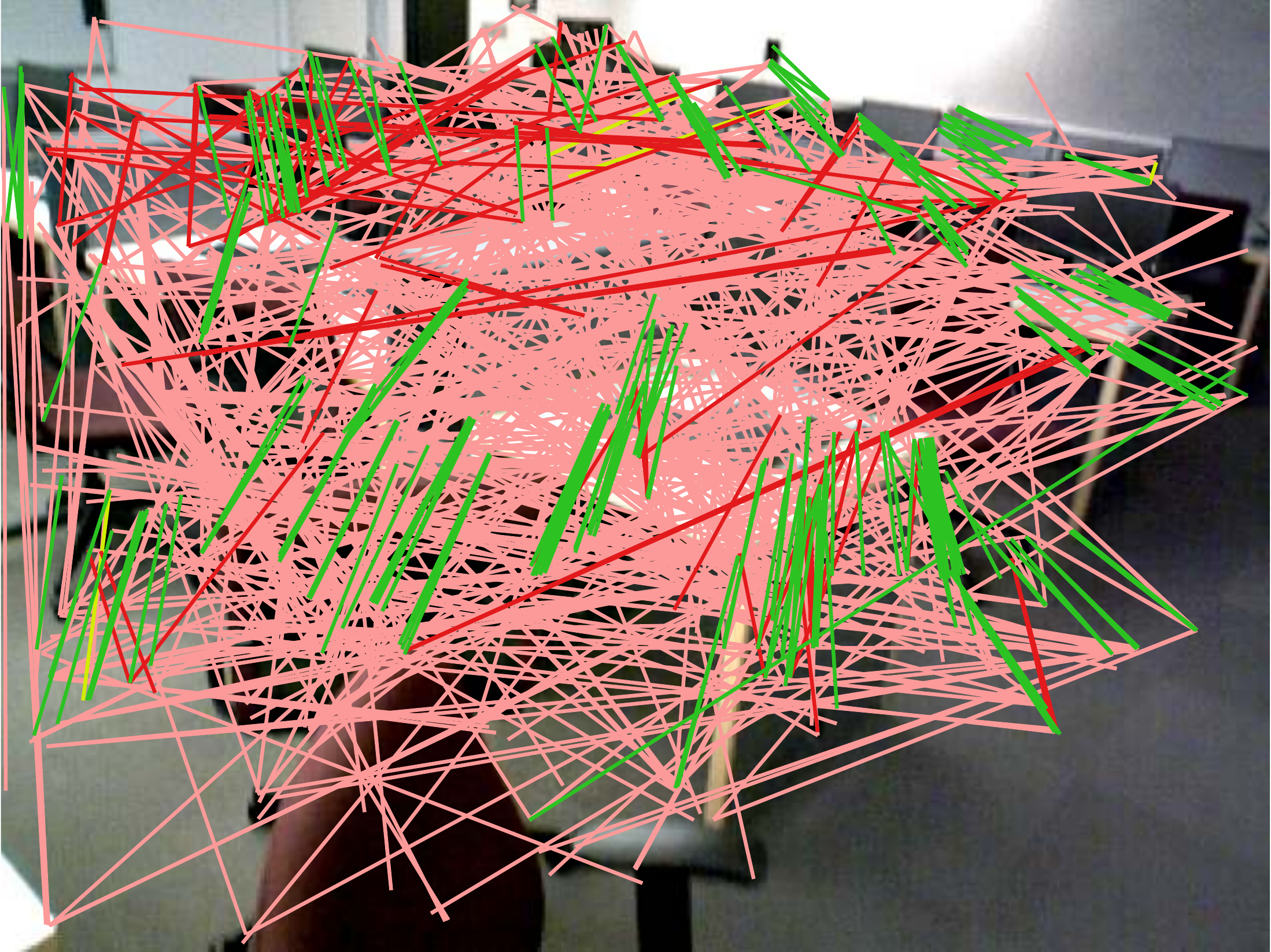}
	\includegraphics[height=7.5em]{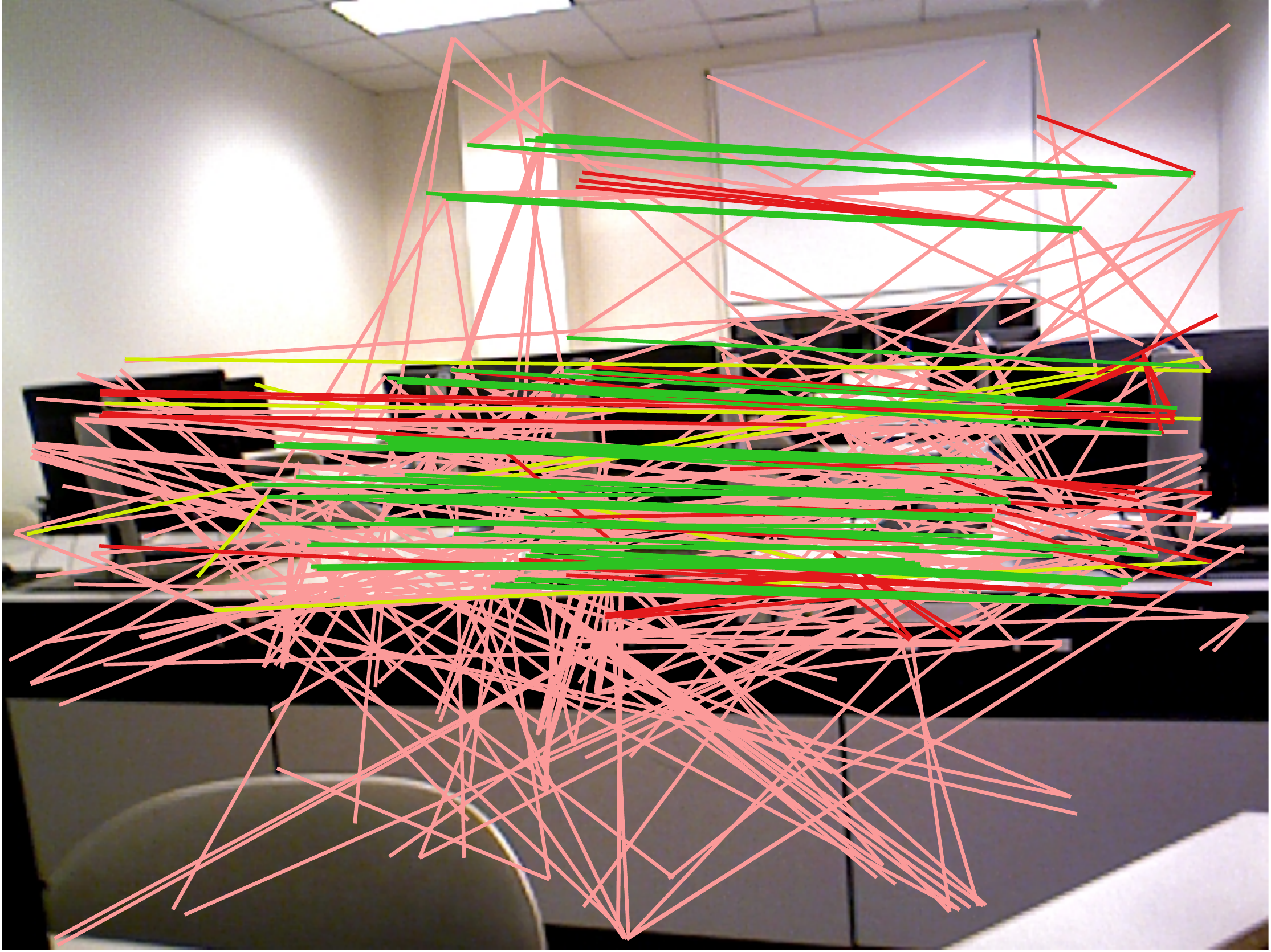}
	\includegraphics[height=7.5em]{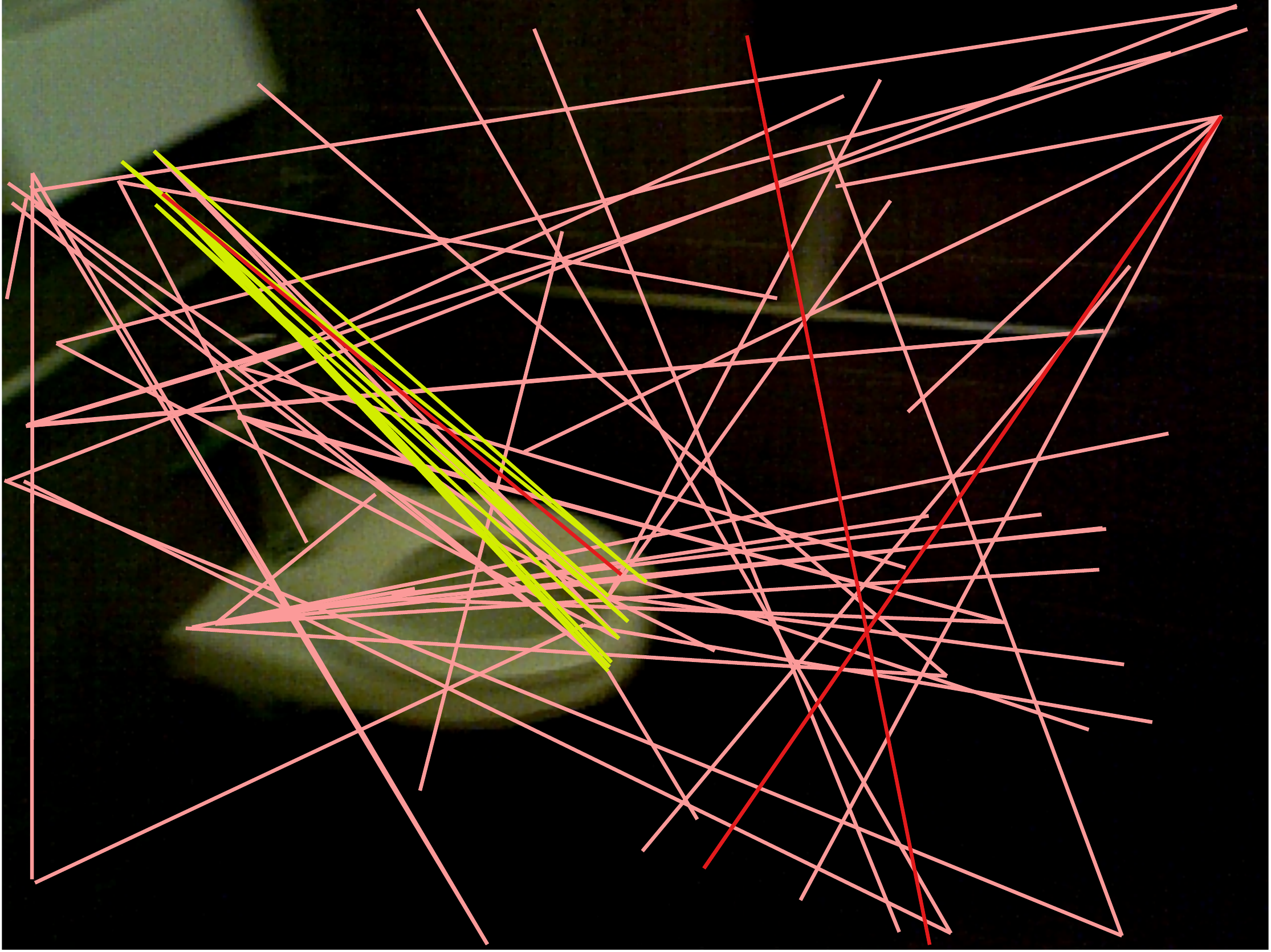}
	\includegraphics[height=7.5em]{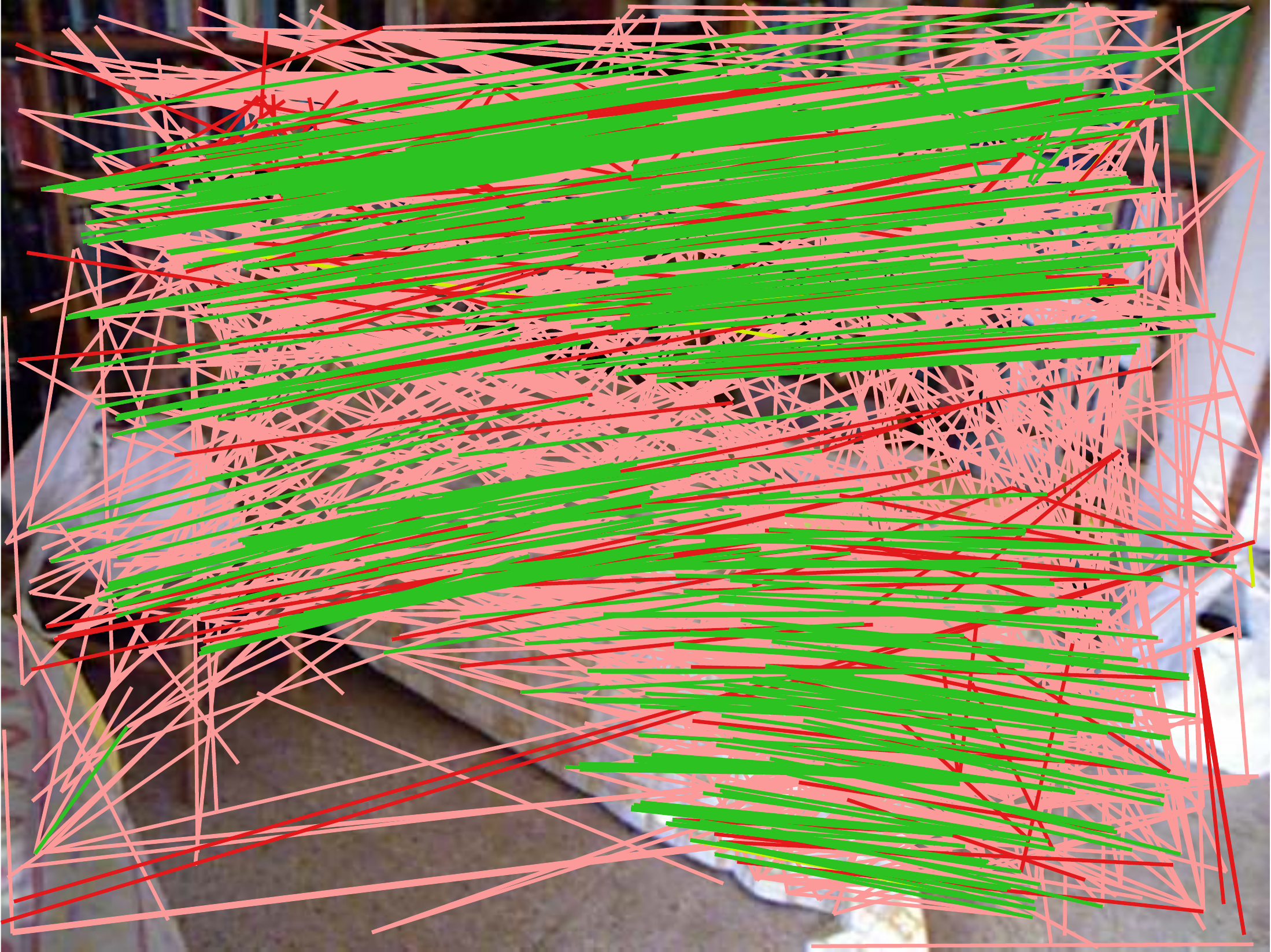}
	\includegraphics[height=7.5em]{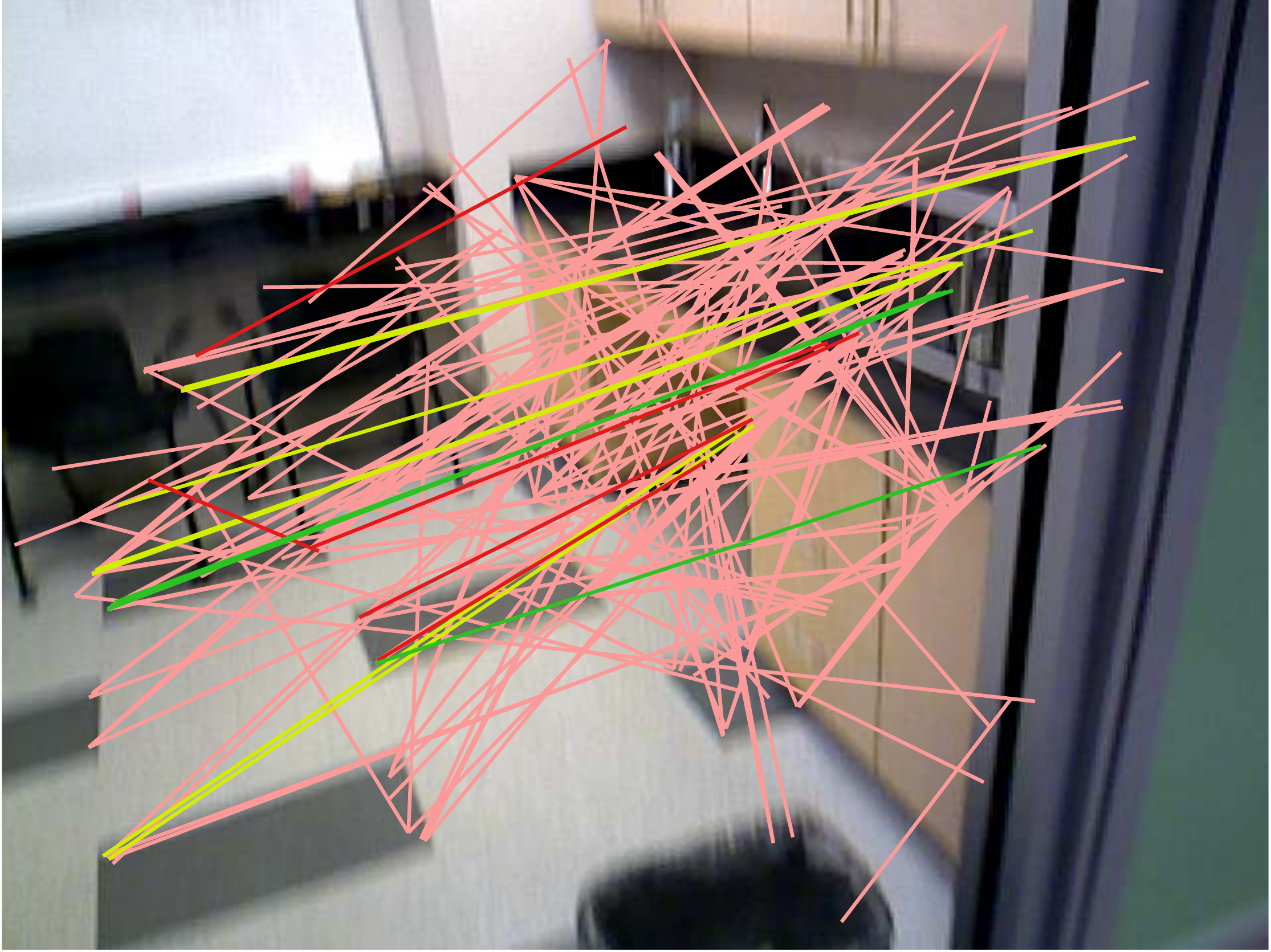}
	\\
	\vspace{0.5em}
	\rotatebox[origin=l]{90}{\mbox{\hspace{2em}BM}}
	\includegraphics[height=7.5em]{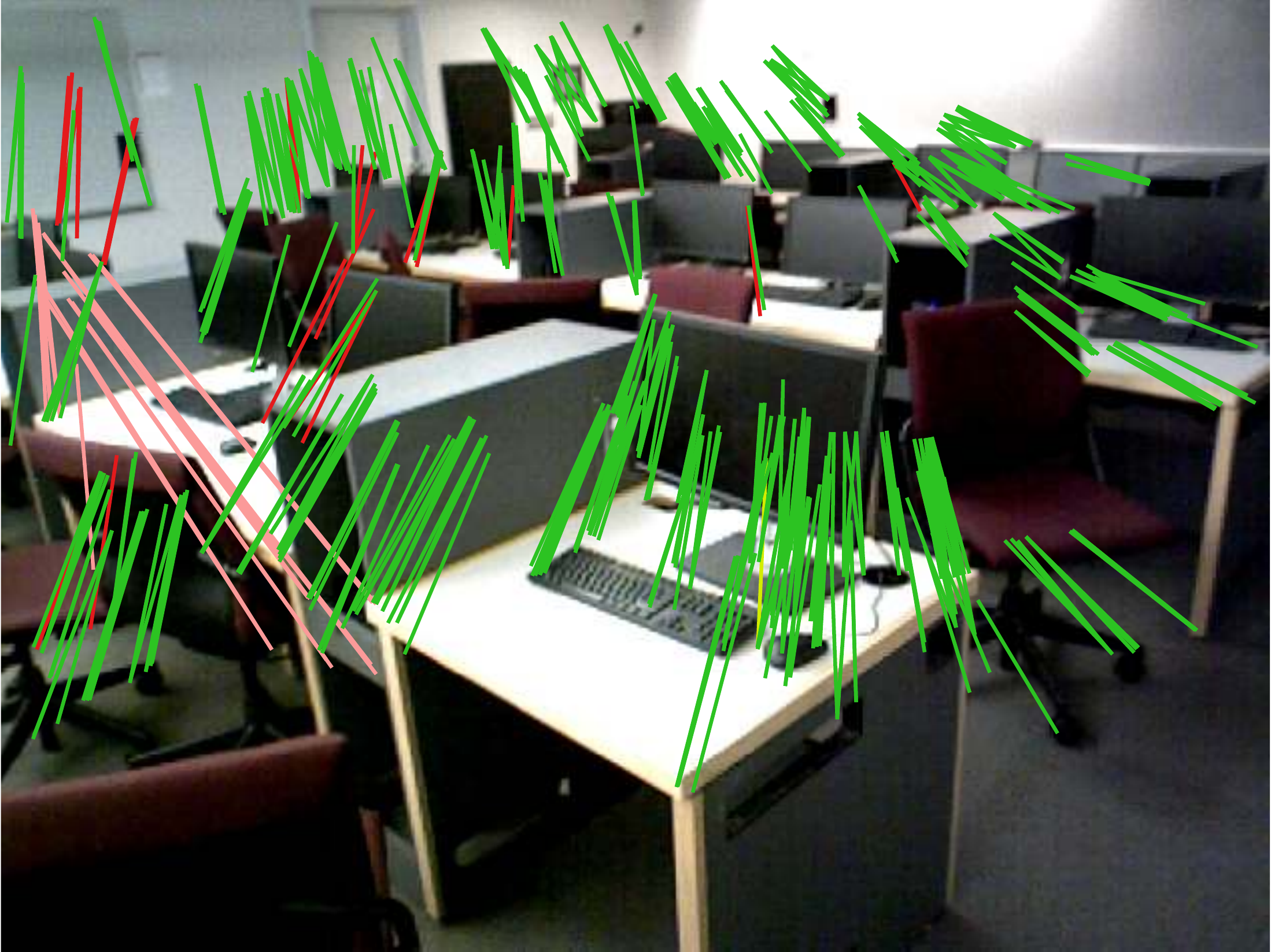}
	\includegraphics[height=7.5em]{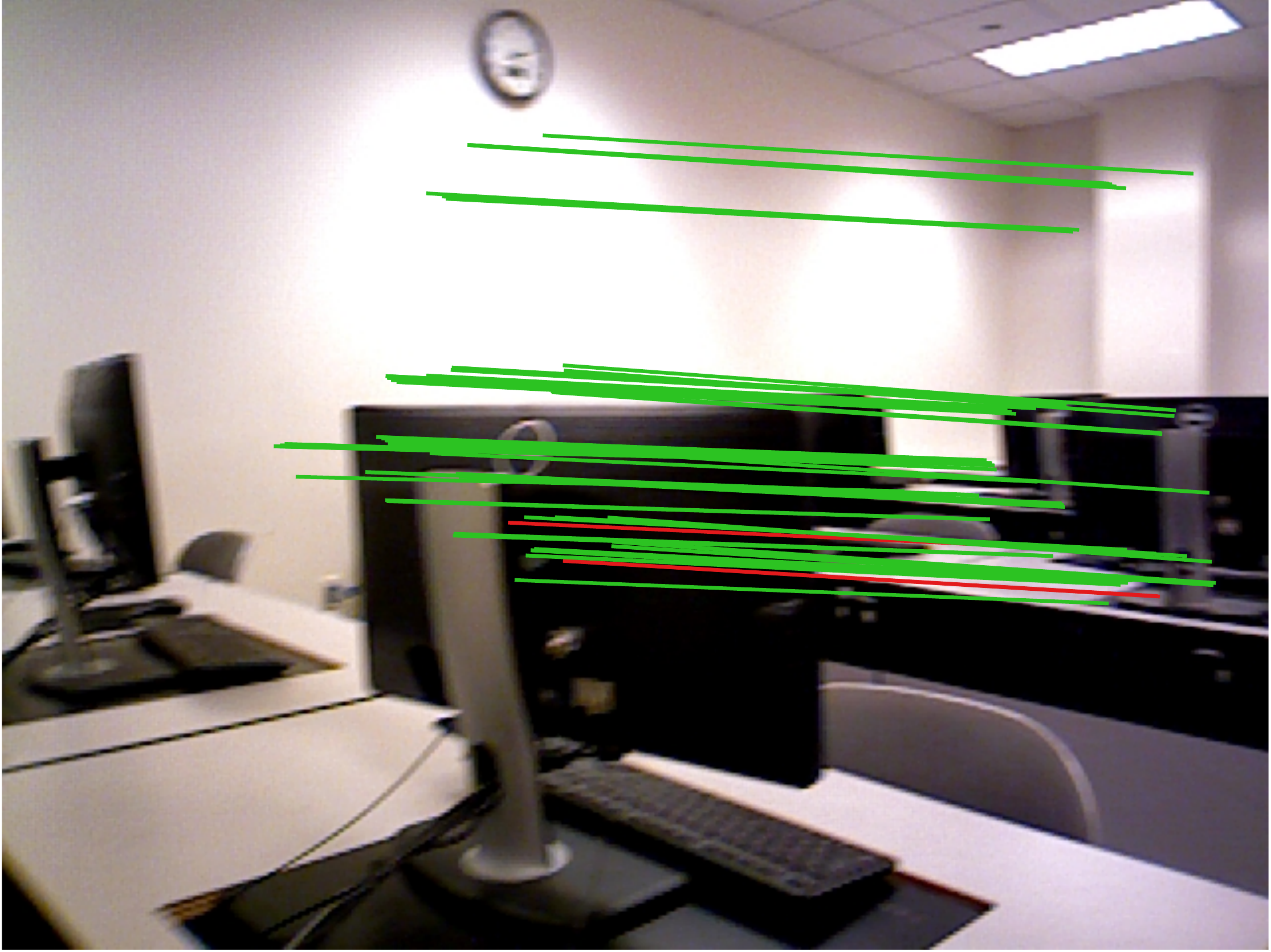}
	\includegraphics[height=7.5em]{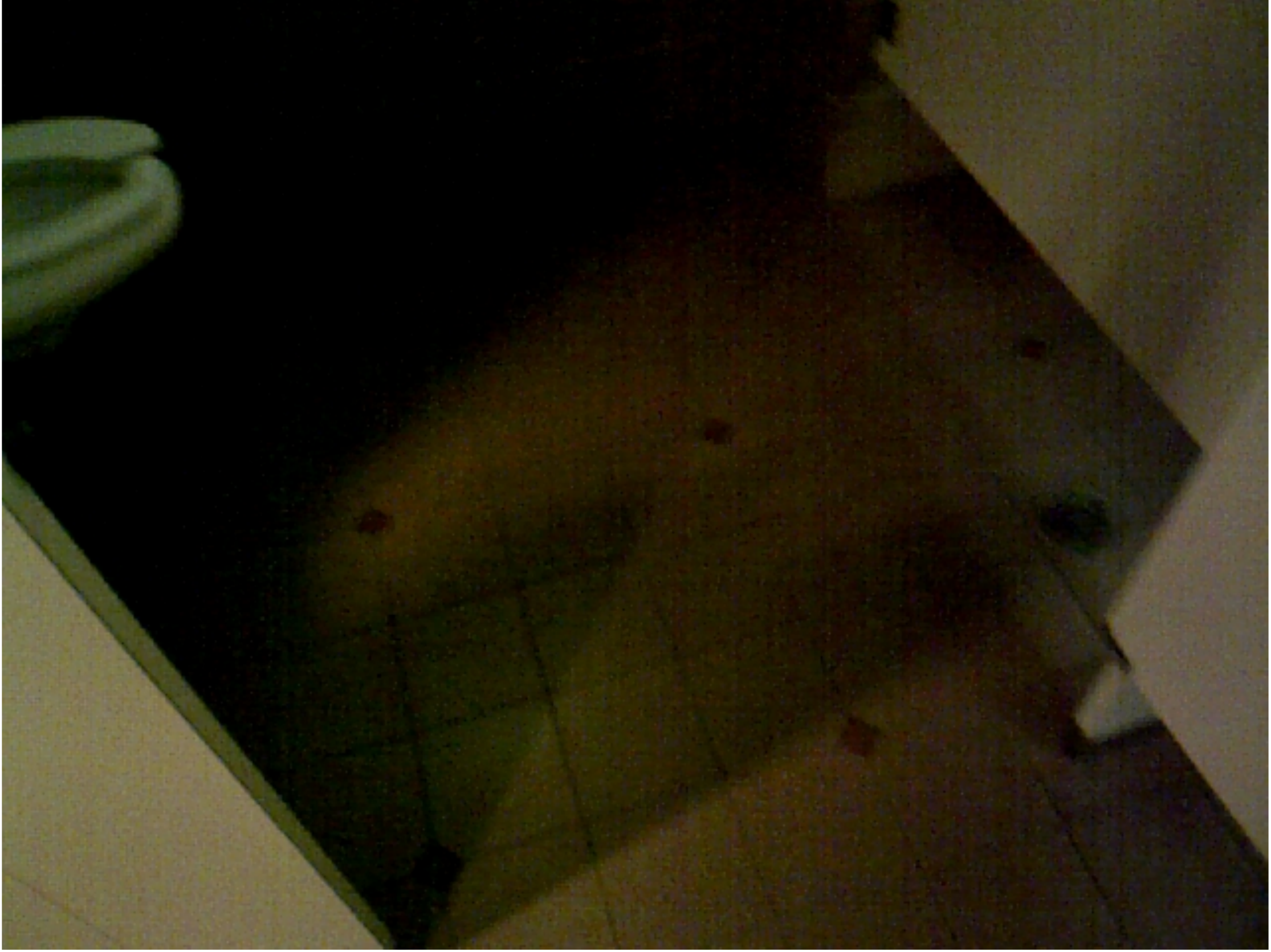}
	\includegraphics[height=7.5em]{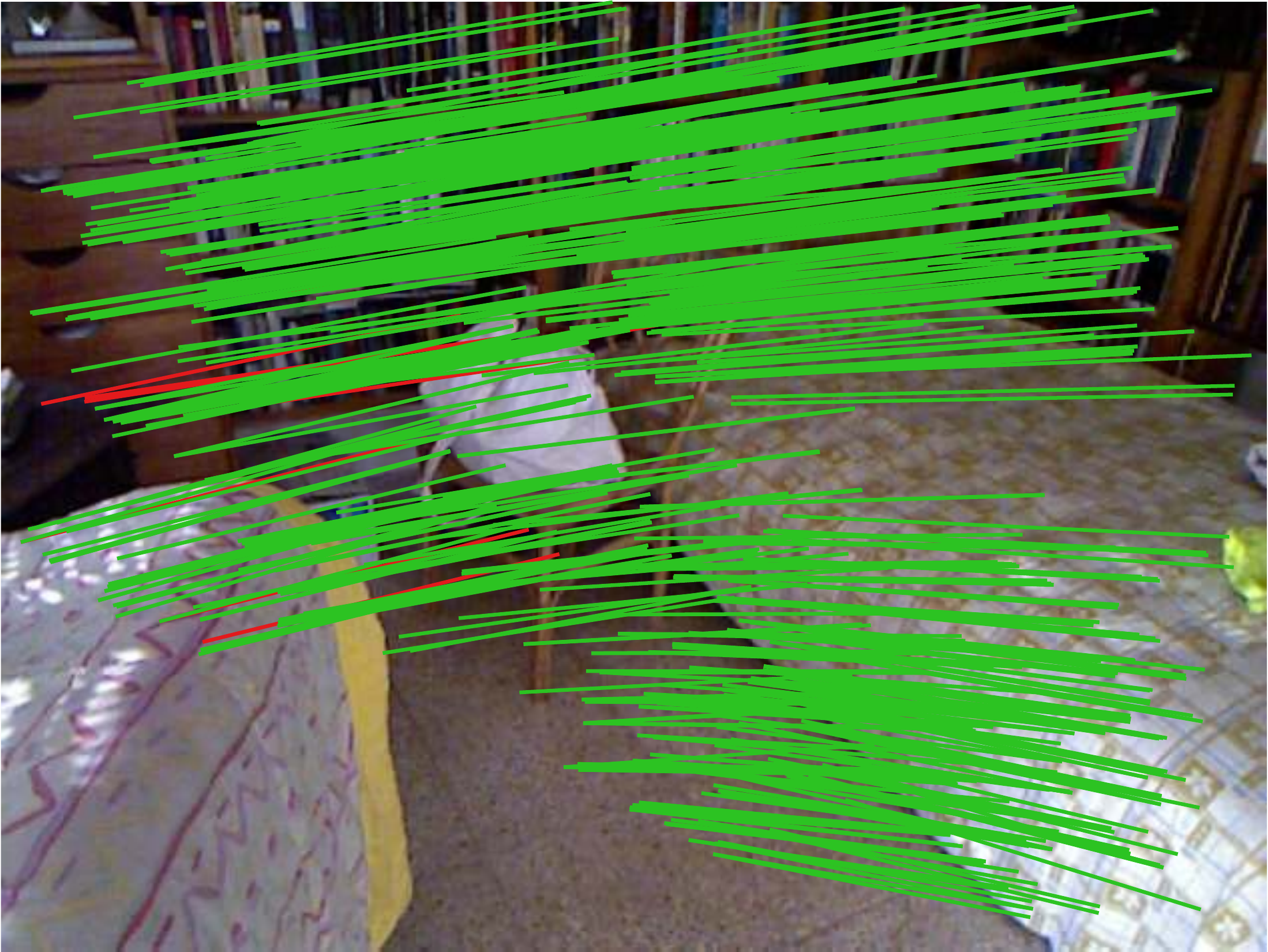}
	\includegraphics[height=7.5em]{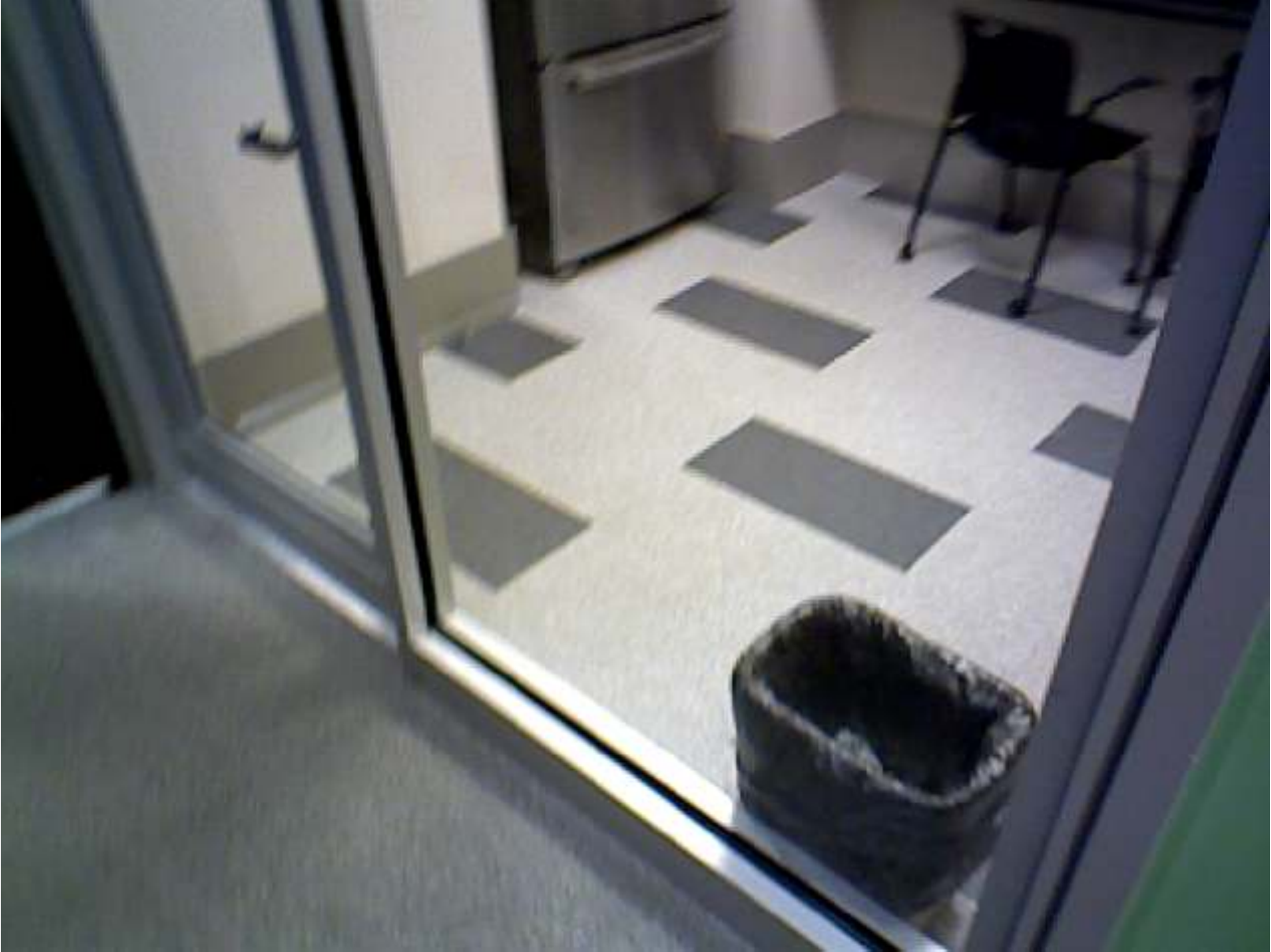}
	\\
	\begin{flushleft}
		\hspace{1.75em}Harvard Computer Lab\hspace{2.5em}Brown CS\hspace{4em}Harvard Restroom\hspace{2em}Home Puigpunyen\hspace{2.5em}MIT Office
	\end{flushleft}
	\caption{\label{example_3b}
		SUN3D local spatial filter matches according to the best configuration setup, the images of the input pair alternate among the rows. For each method inlier (yellow, green) and outlier (red and light red) clusters are shown, as well as the 1SAC filtered matches (green, red) (see Sec.~\ref{eval_dt}, best viewed in color and zoomed in).}
\end{figure*}

\begin{figure*}
	\center
	\rotatebox[origin=l]{90}{\mbox{\hspace{3em}th}}
	\includegraphics[height=7.5em]{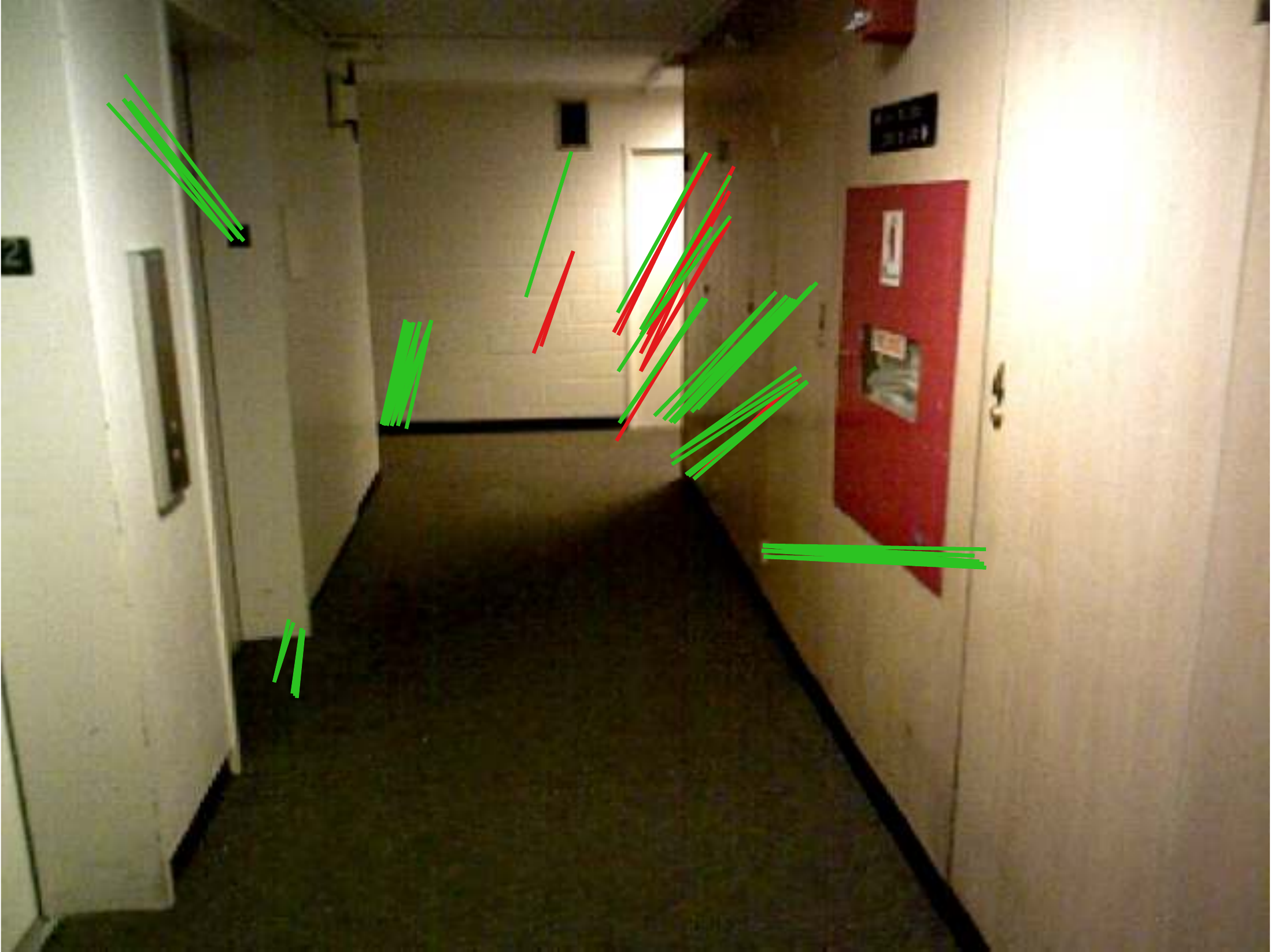}
	\includegraphics[height=7.5em]{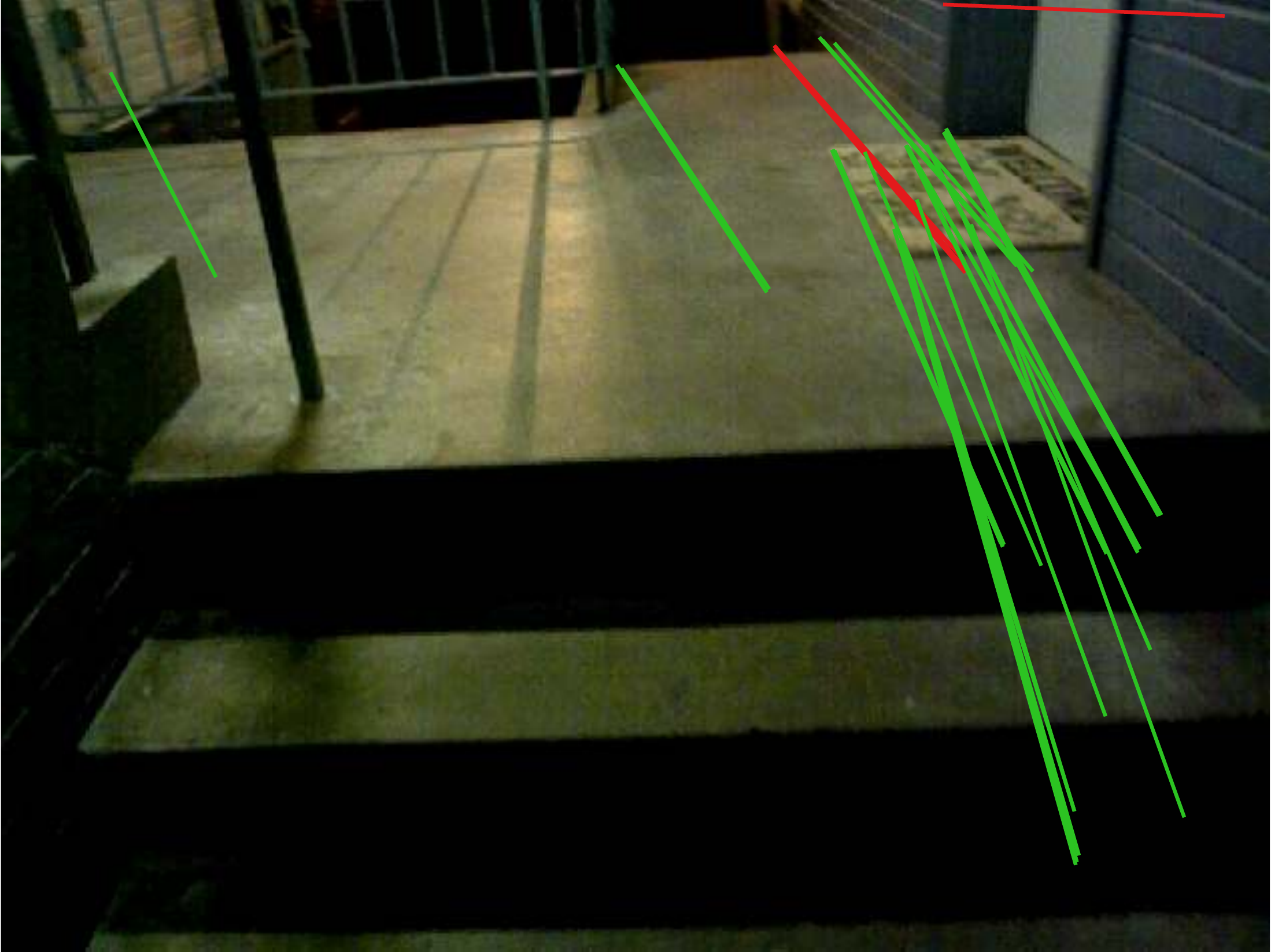}
	\includegraphics[height=7.5em]{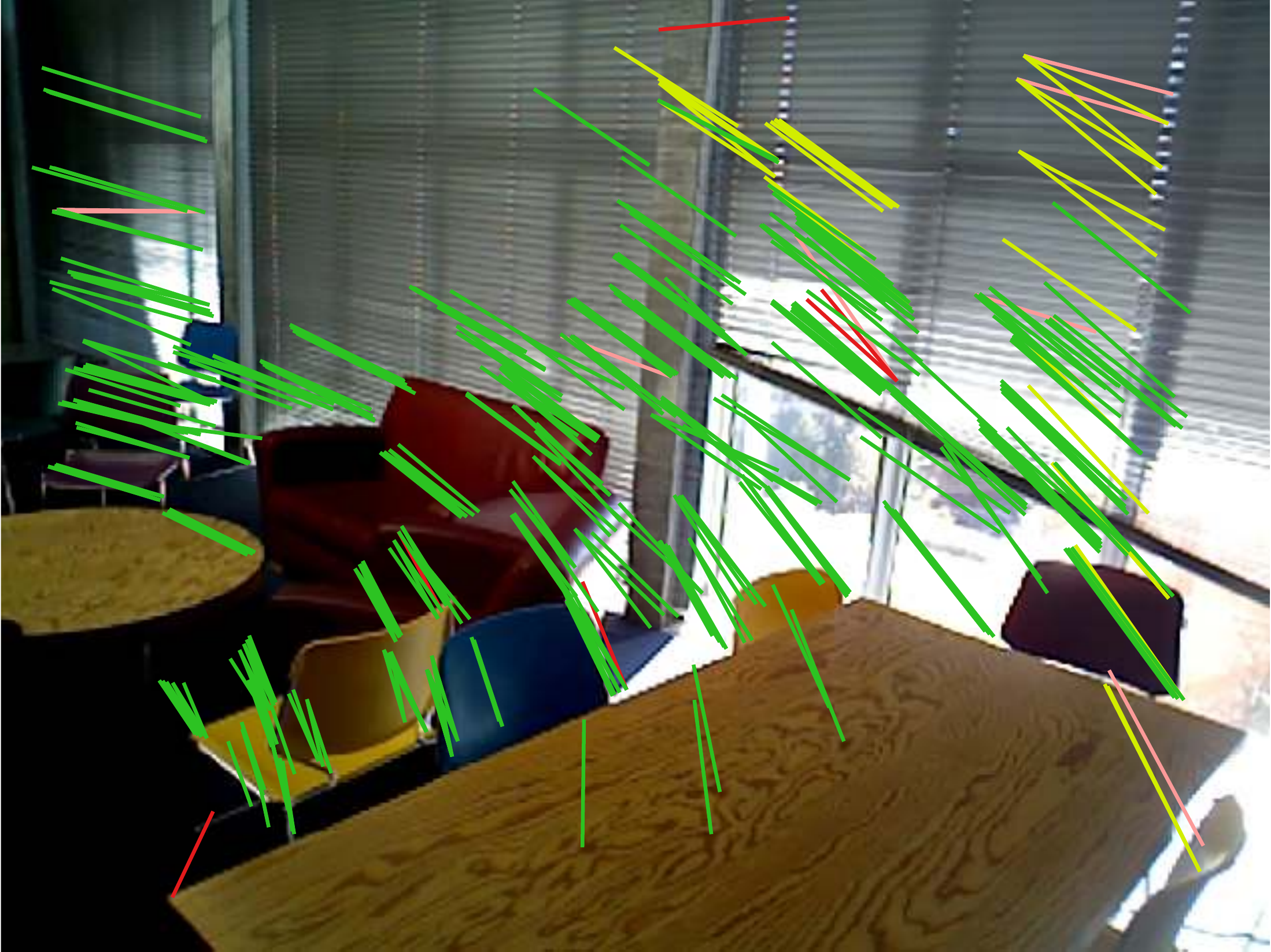}
	\includegraphics[height=7.5em]{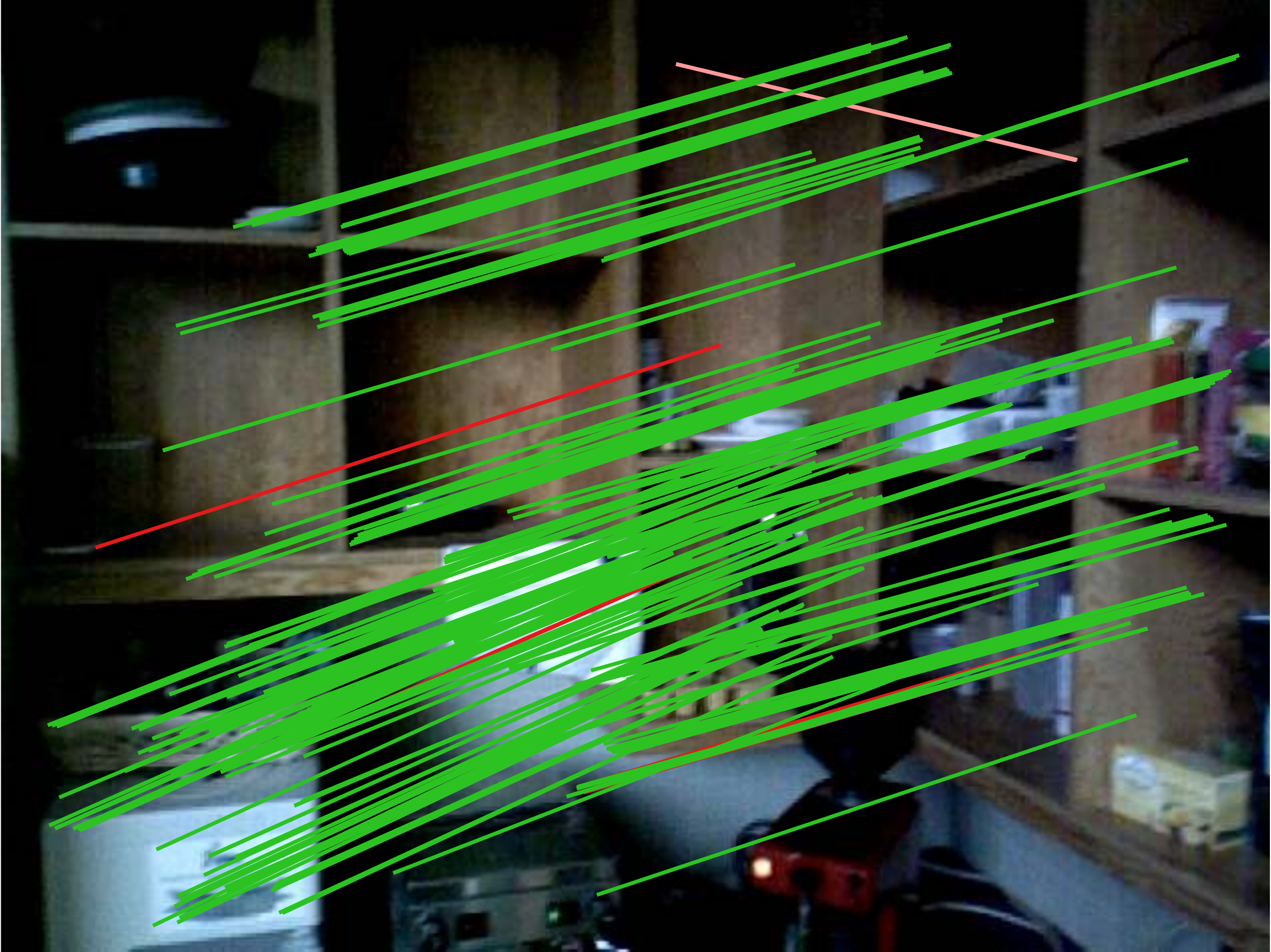}
	\includegraphics[height=7.5em]{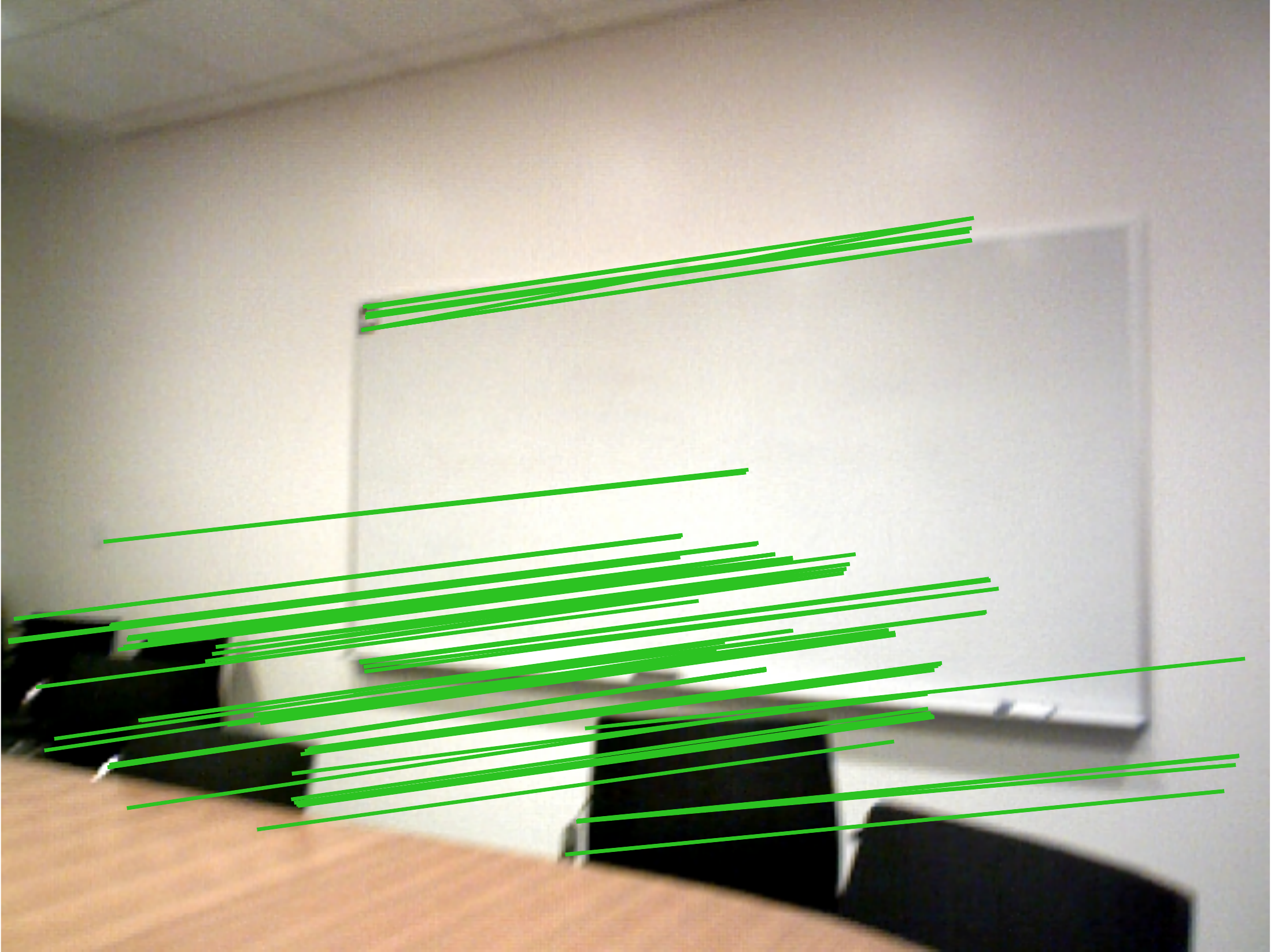}
	\\
	\vspace{0.5em}
	\rotatebox[origin=l]{90}{\mbox{\hspace{2em}DTM}}
	\includegraphics[height=7.5em]{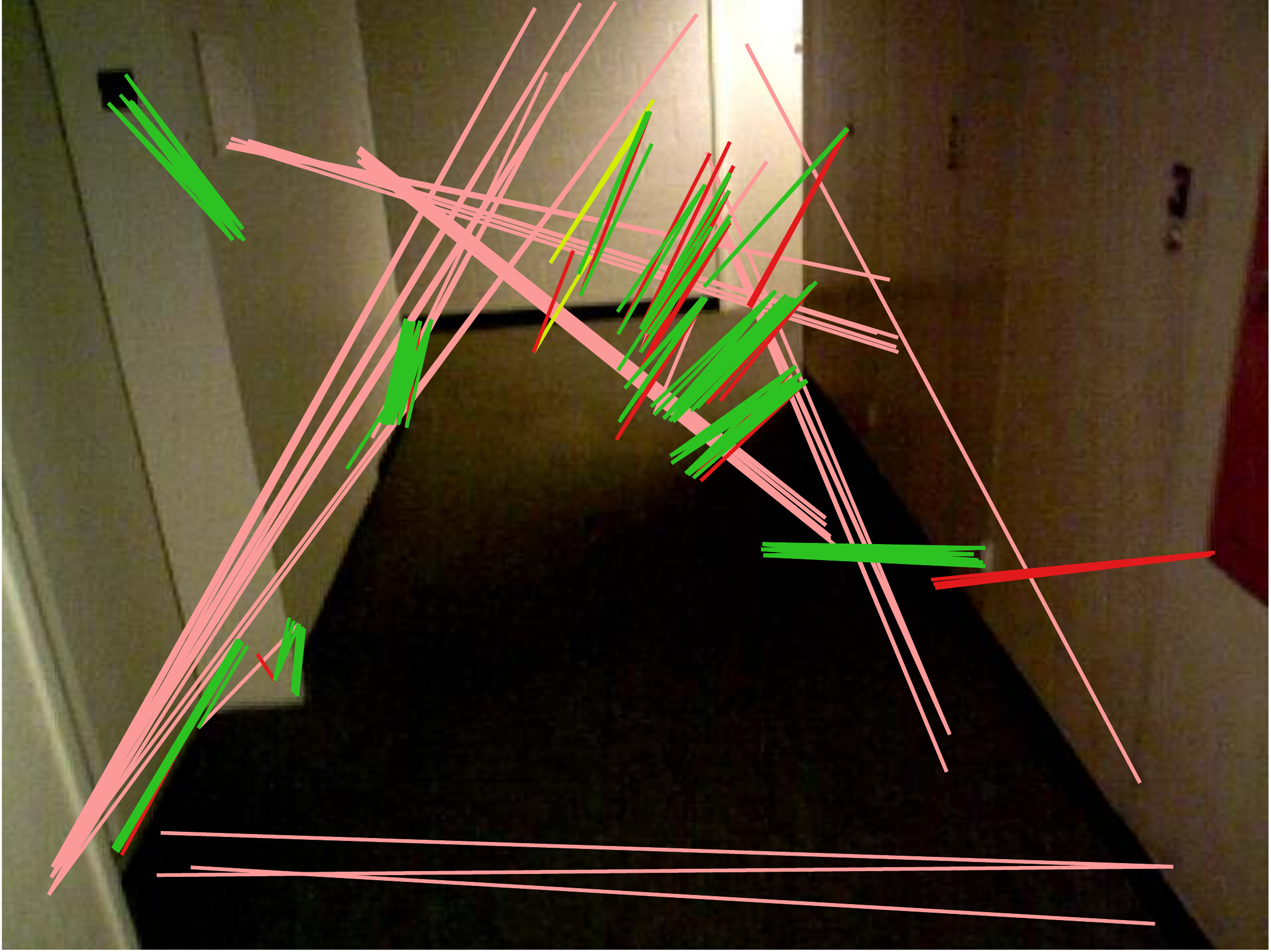}
	\includegraphics[height=7.5em]{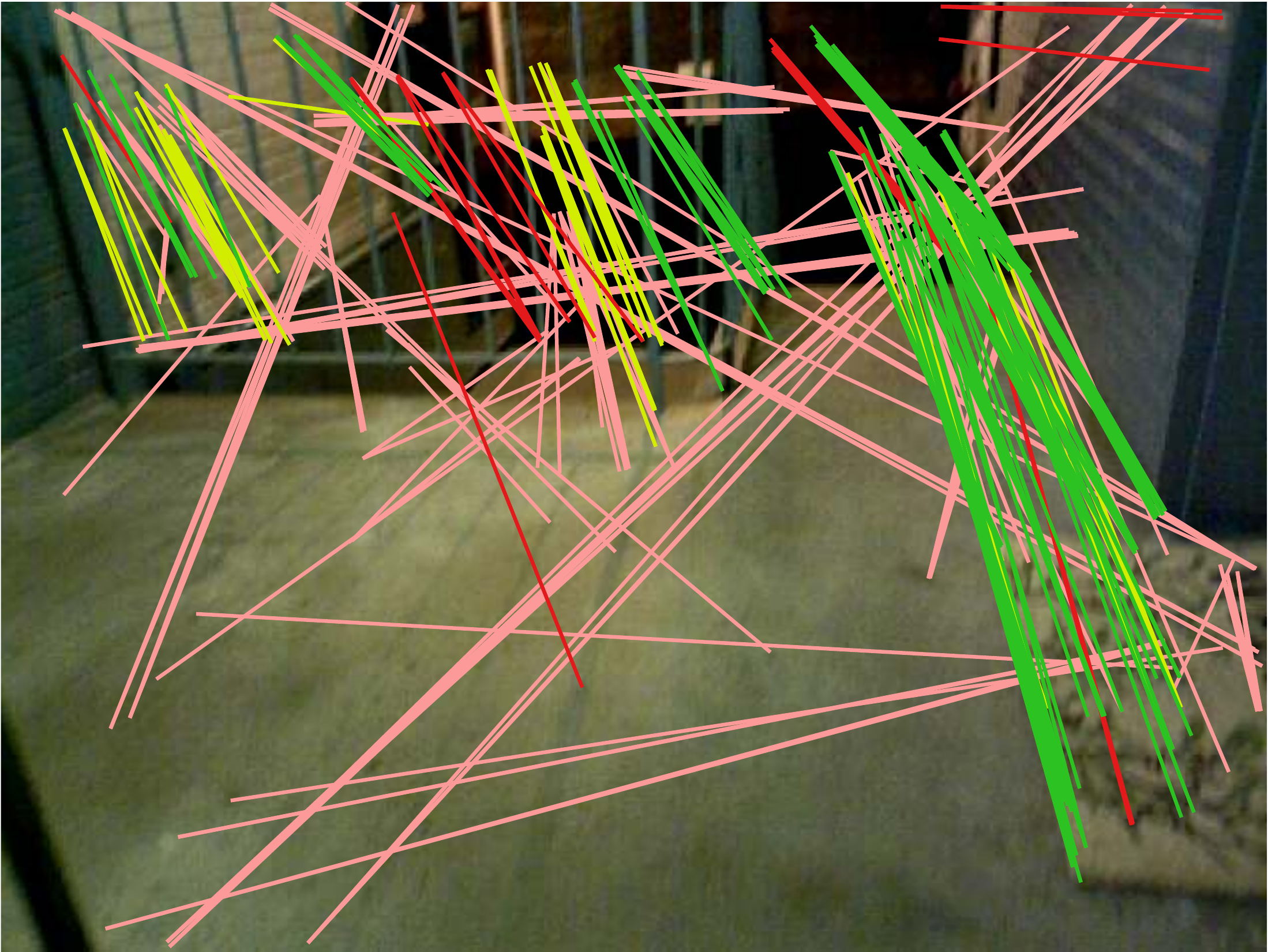}
	\includegraphics[height=7.5em]{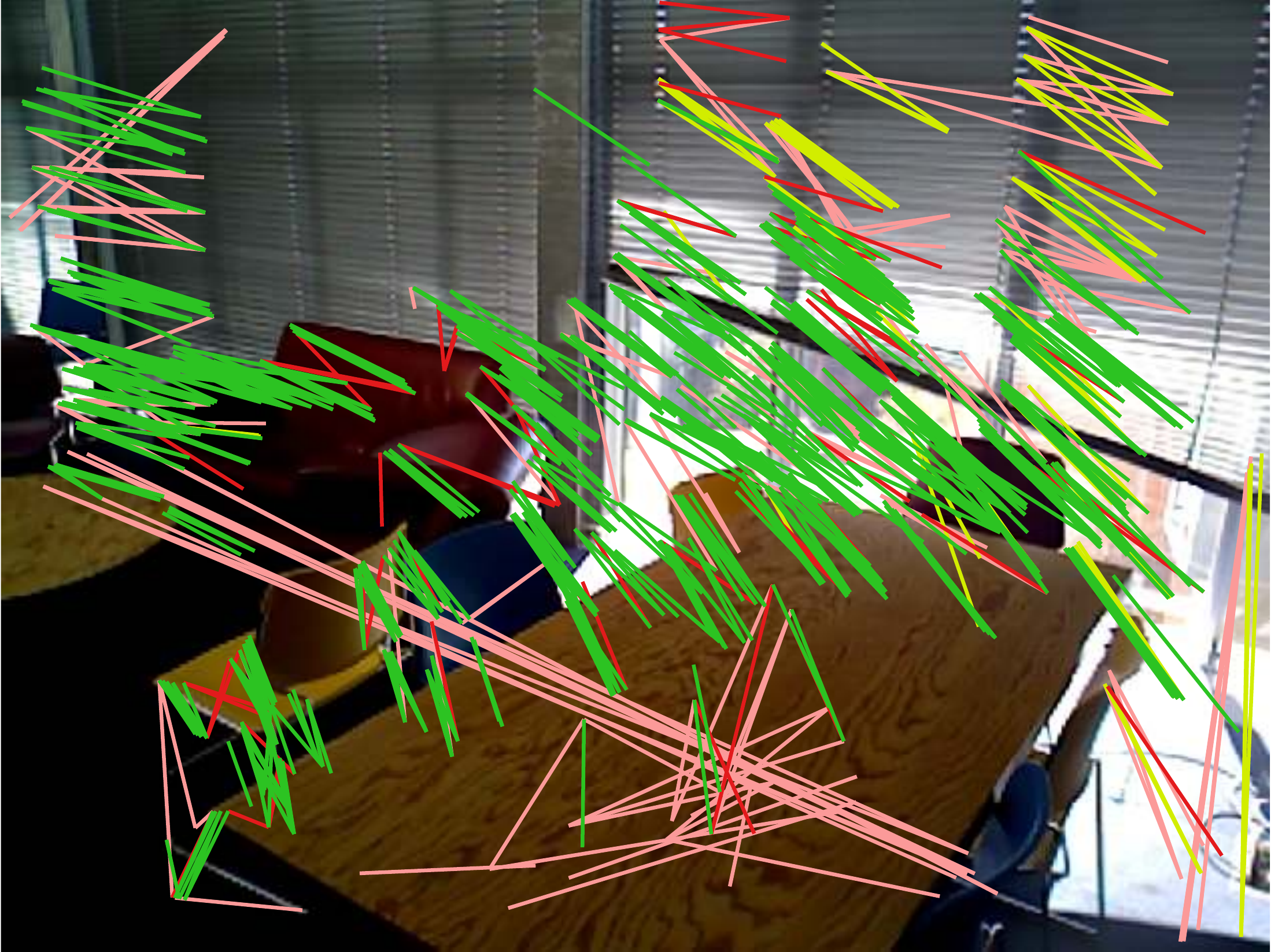}
	\includegraphics[height=7.5em]{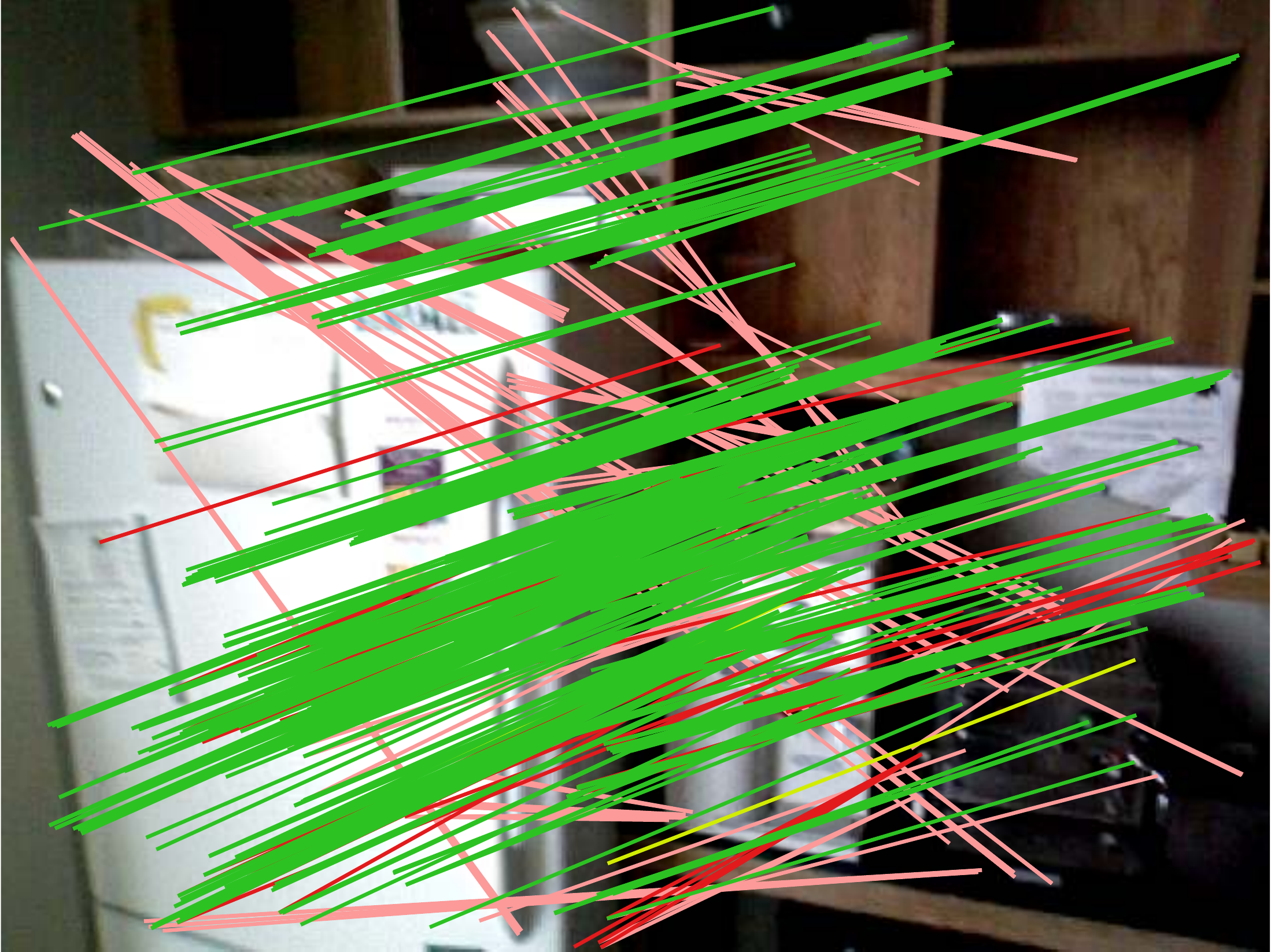}
	\includegraphics[height=7.5em]{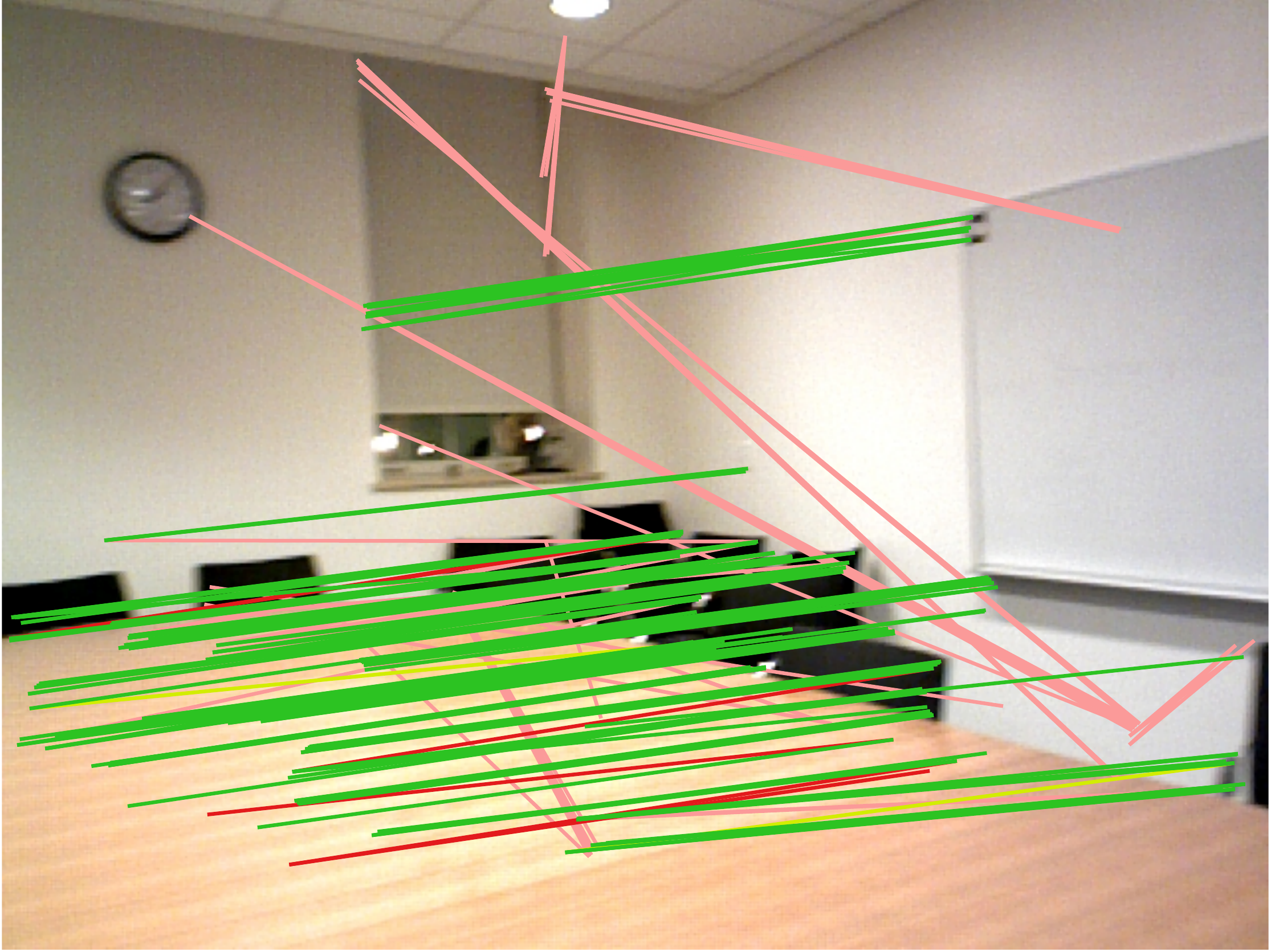}
	\\
	\vspace{0.5em}
	\rotatebox[origin=l]{90}{\mbox{\hspace{2em}LMR}}
	\includegraphics[height=7.5em]{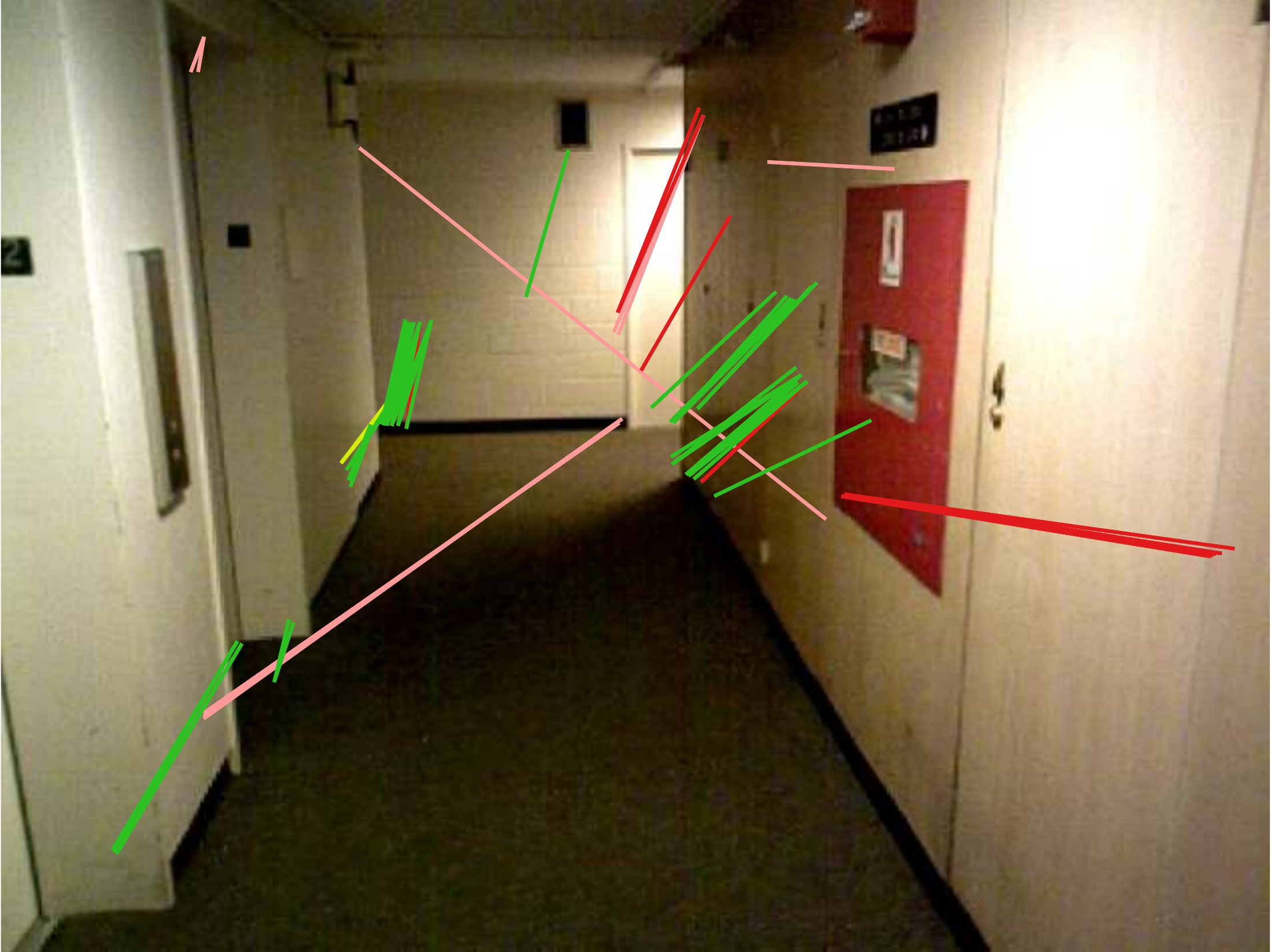}
	\includegraphics[height=7.5em]{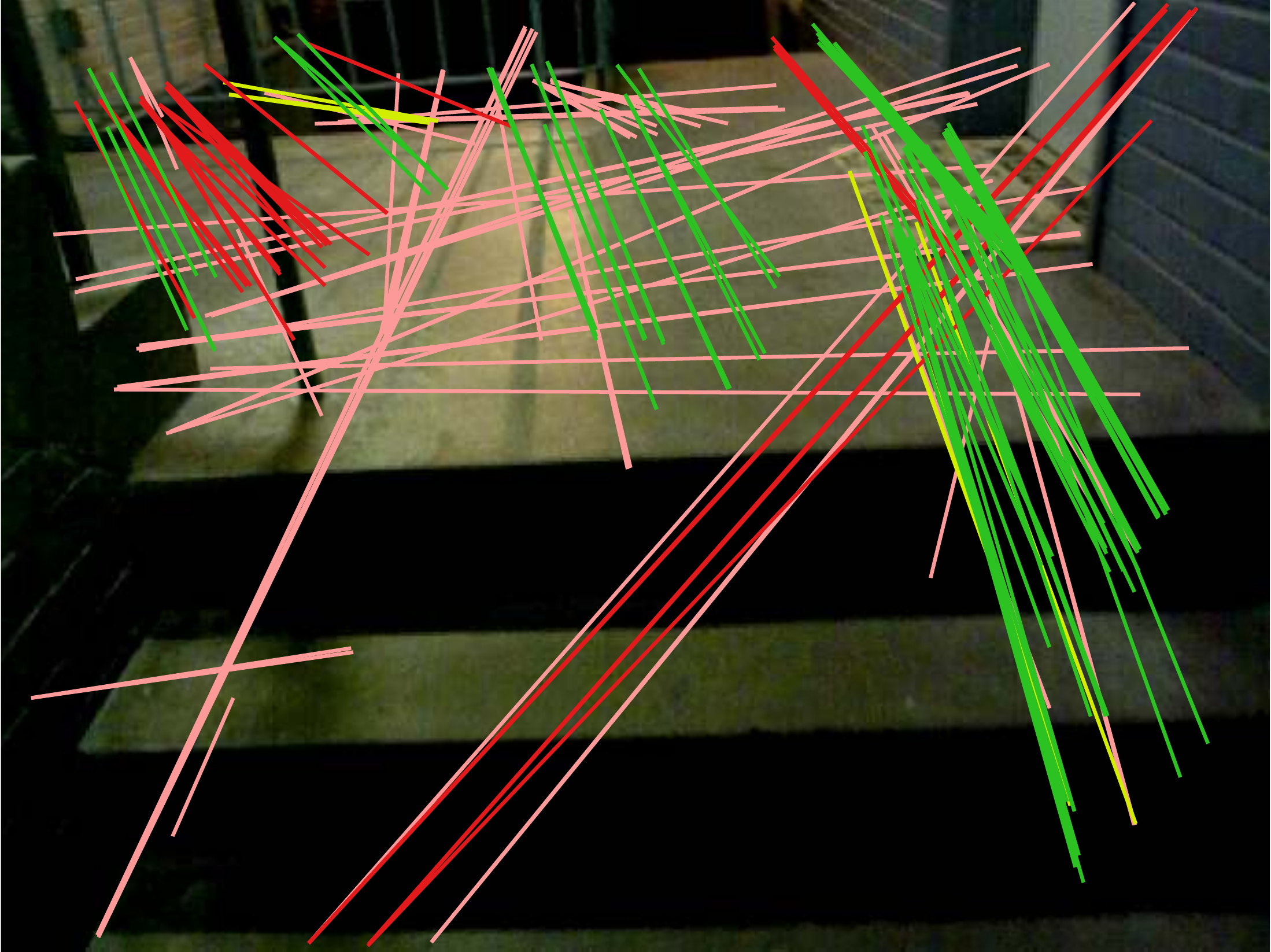}
	\includegraphics[height=7.5em]{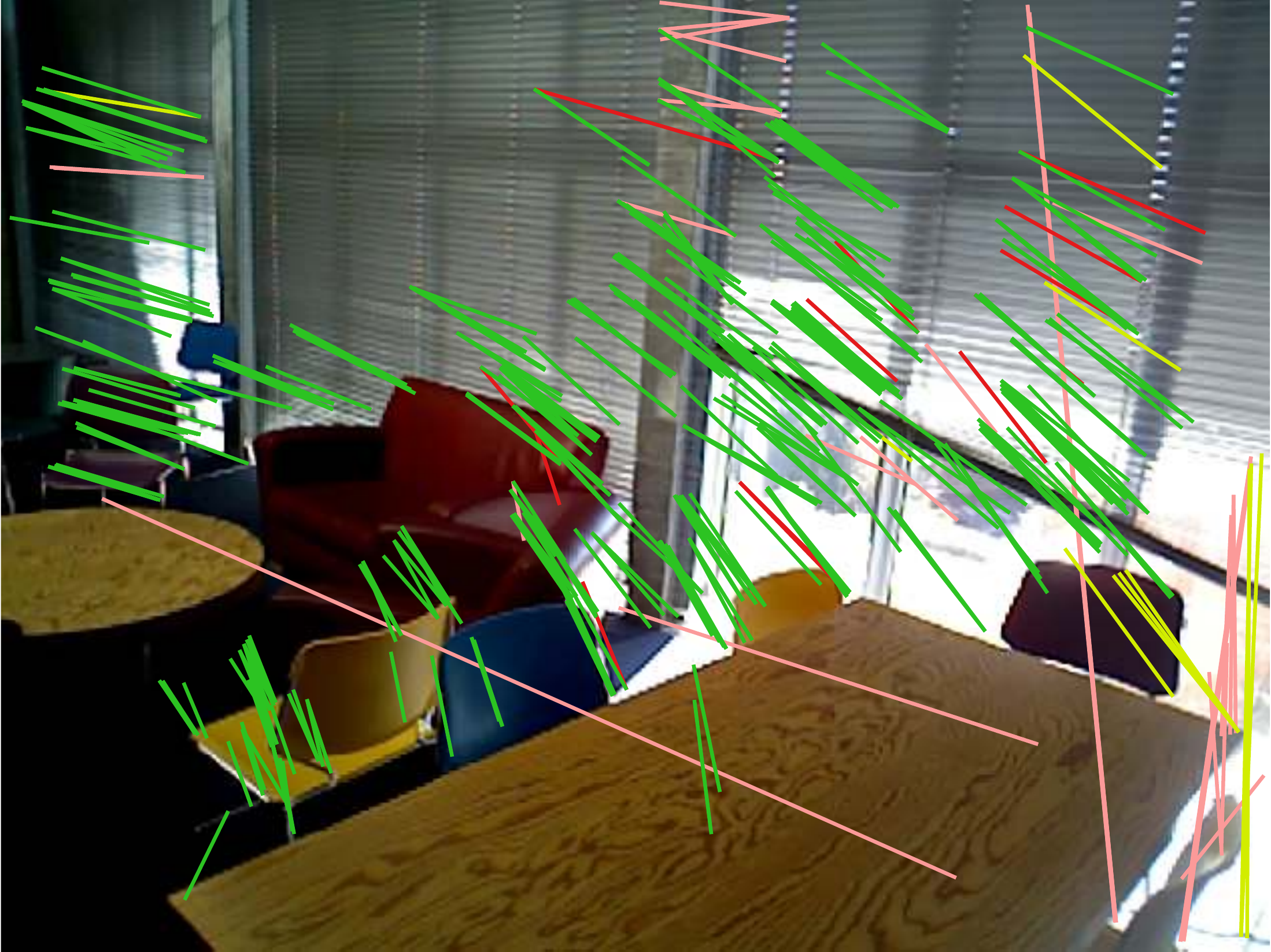}
	\includegraphics[height=7.5em]{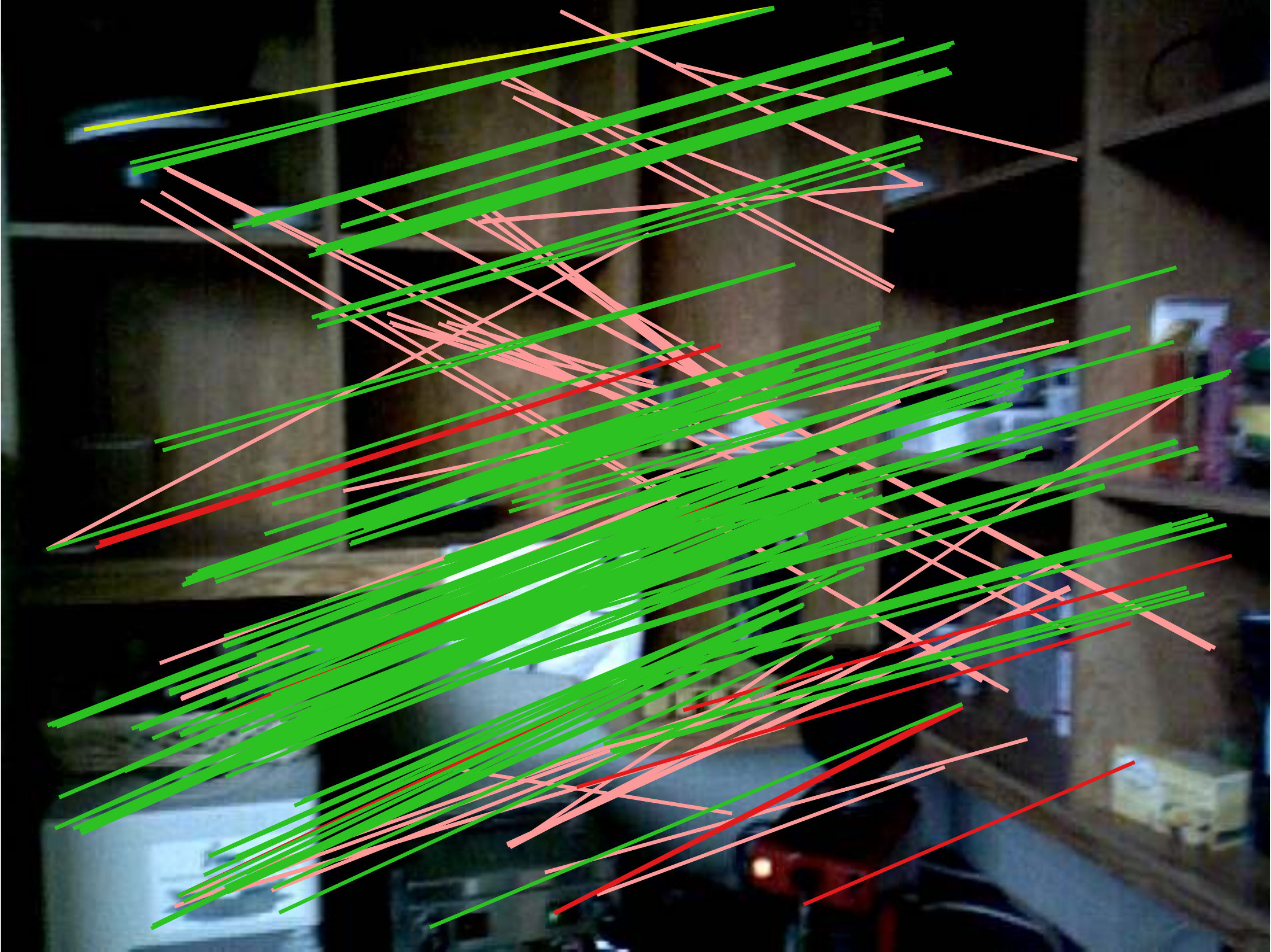}
	\includegraphics[height=7.5em]{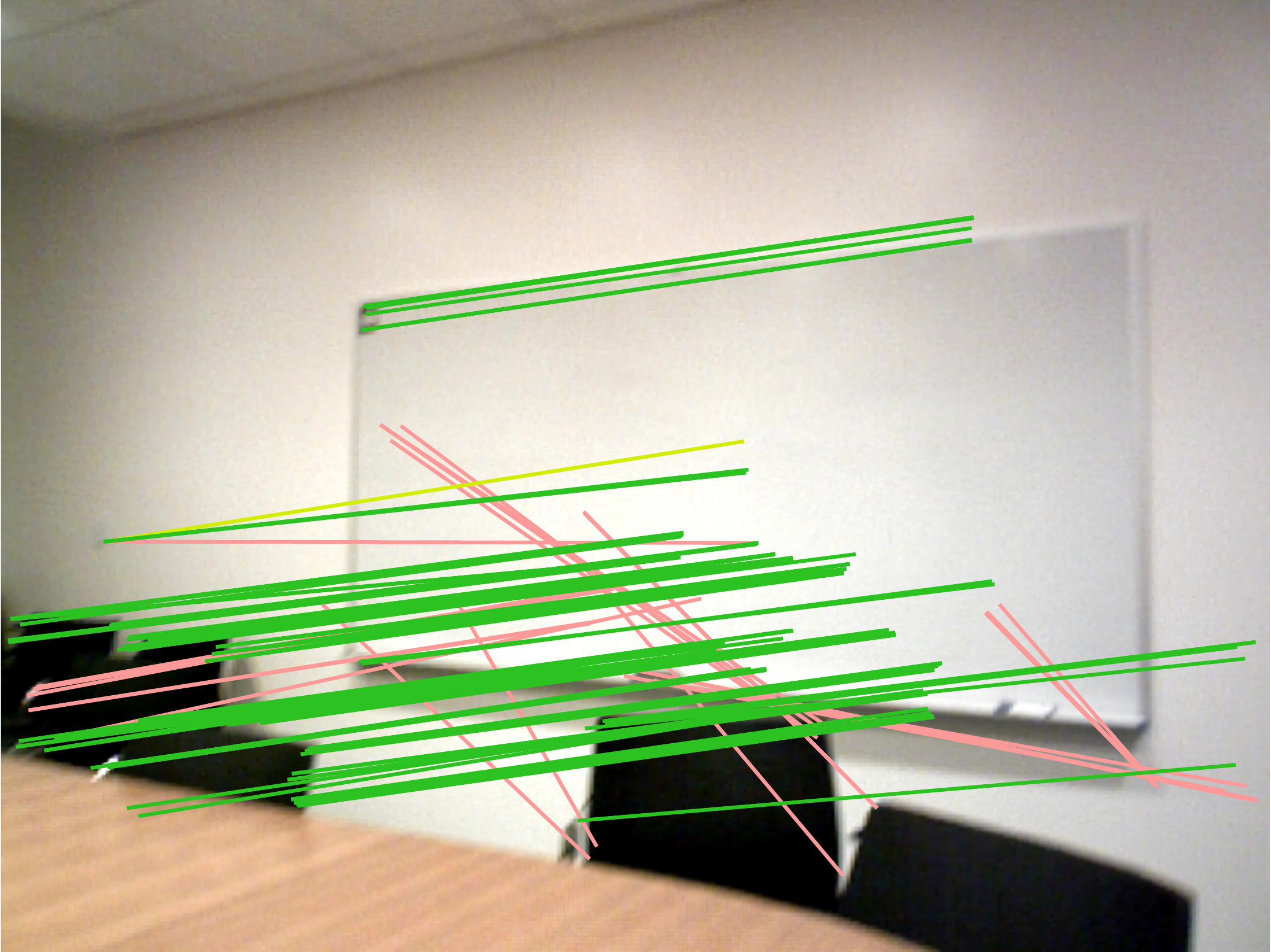}
	\\
	\vspace{0.5em}
	\rotatebox[origin=l]{90}{\mbox{\hspace{2em}LPM}}
	\includegraphics[height=7.5em]{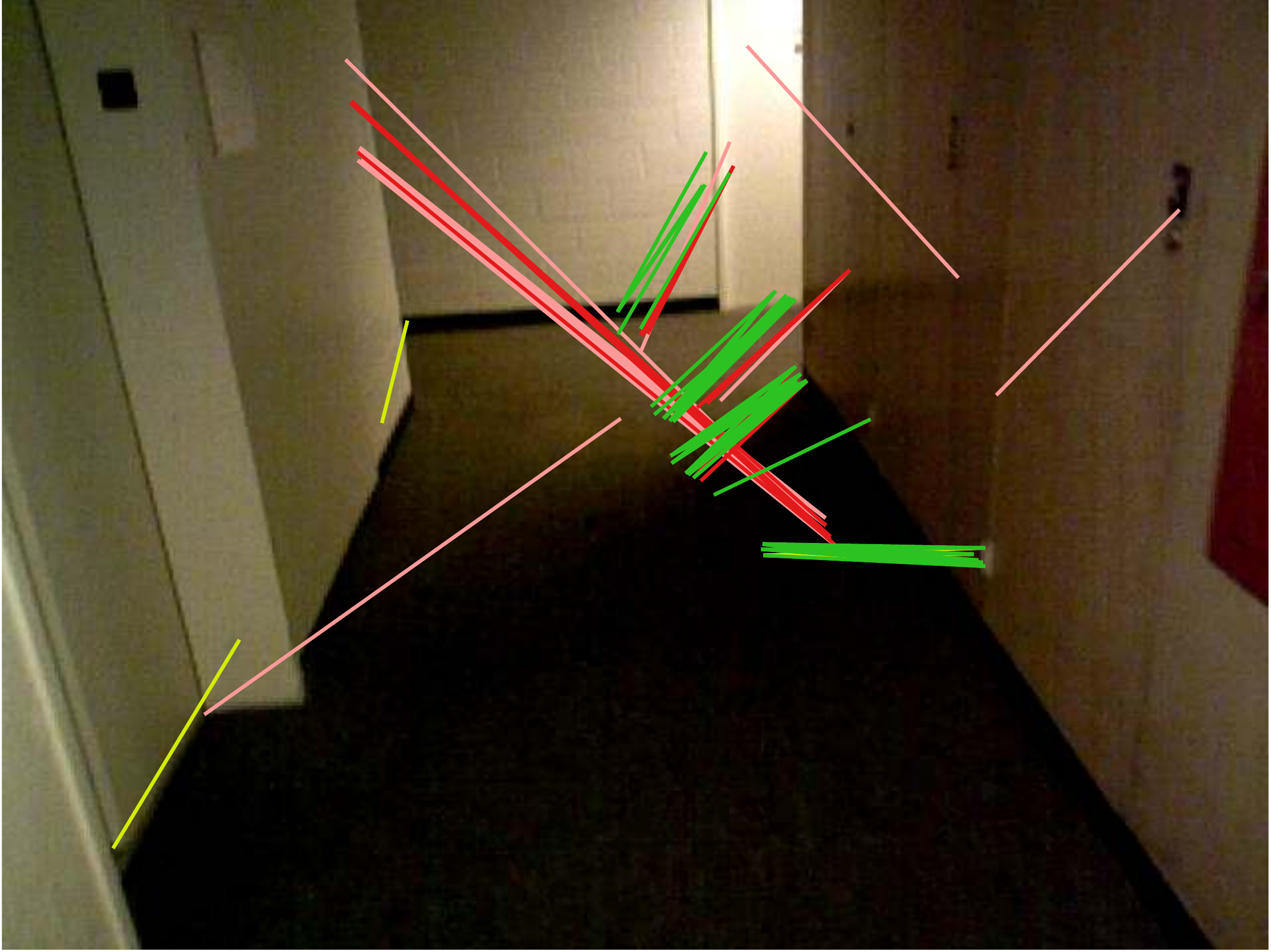}
	\includegraphics[height=7.5em]{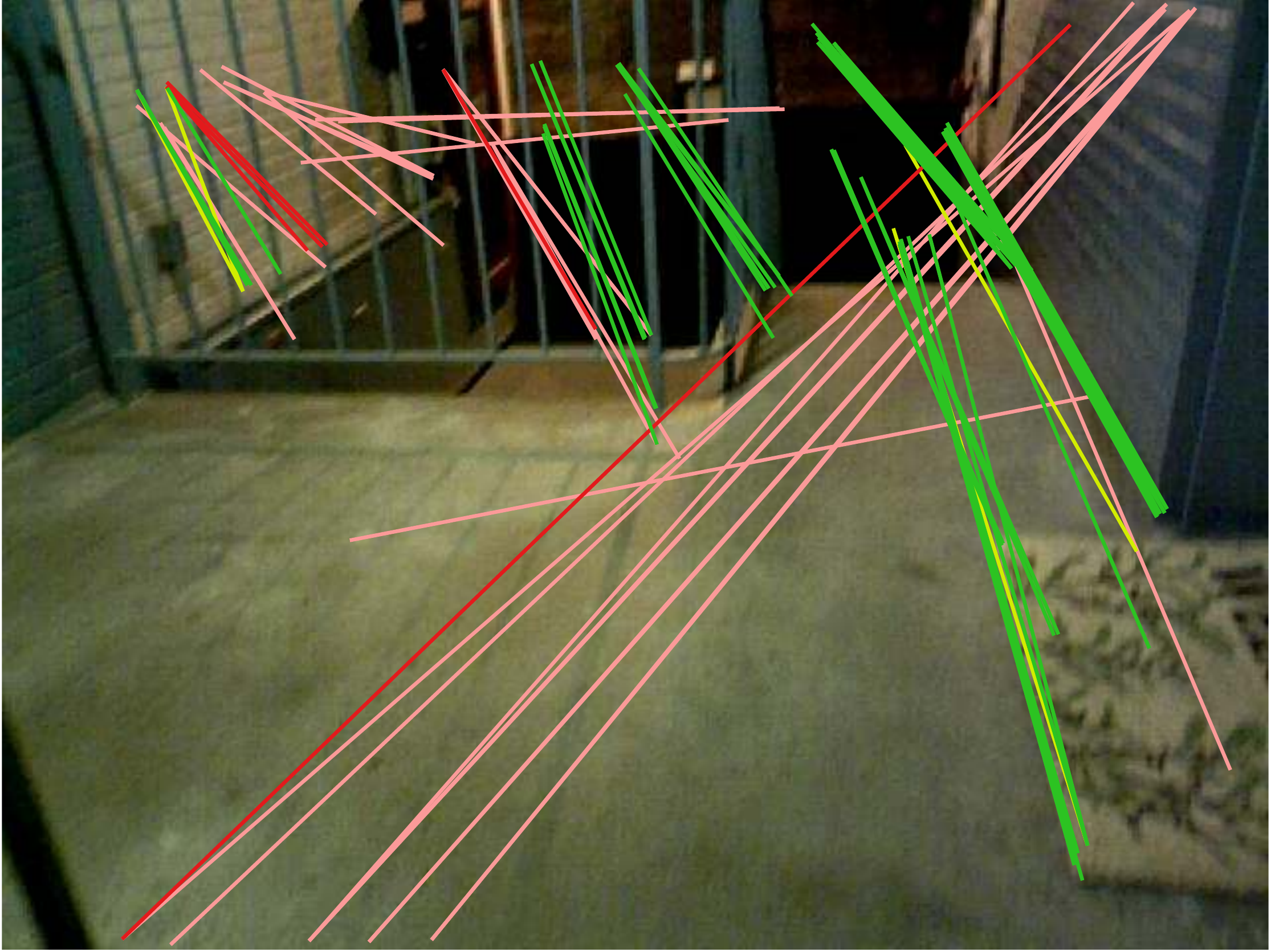}
	\includegraphics[height=7.5em]{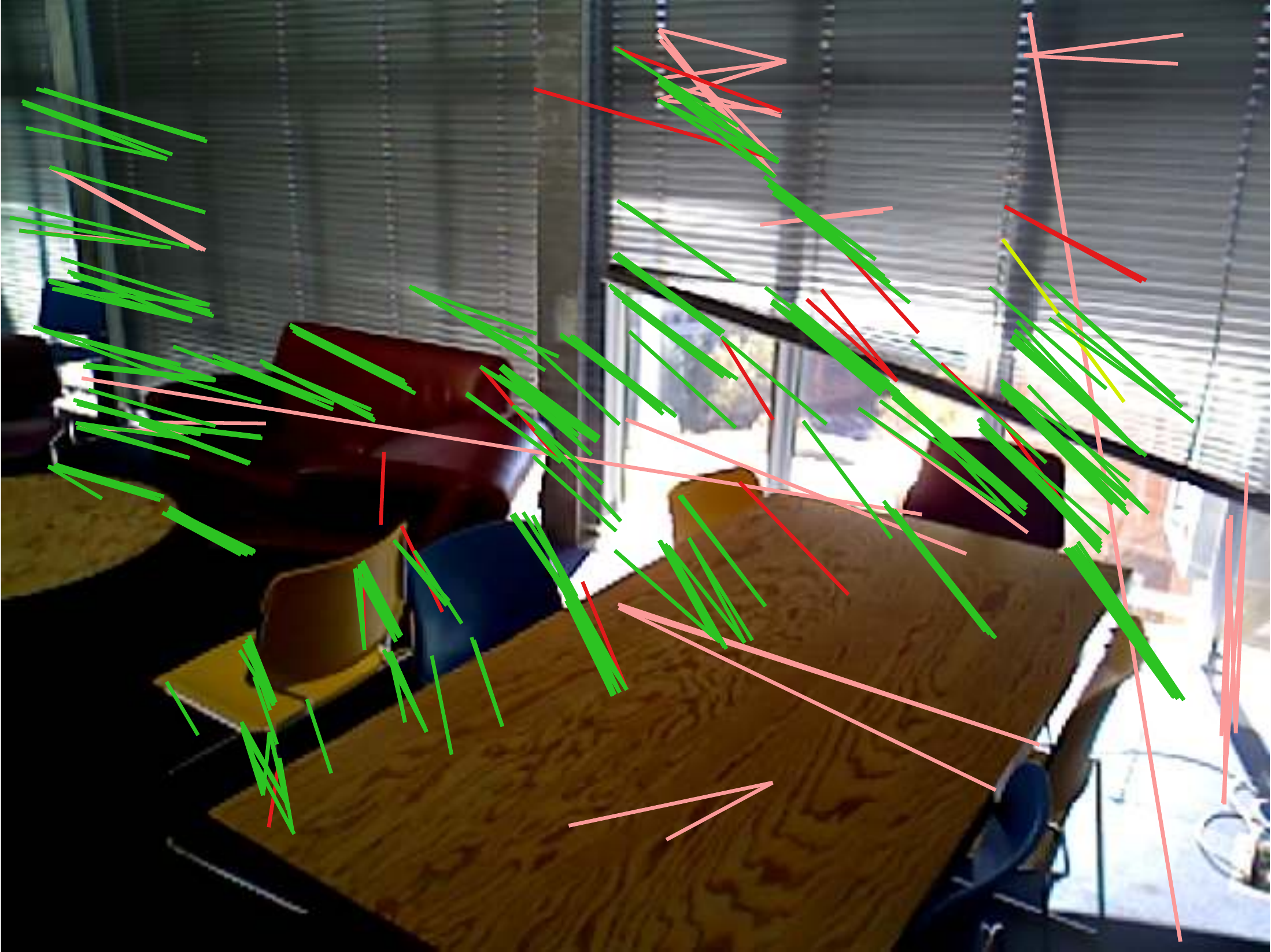}
	\includegraphics[height=7.5em]{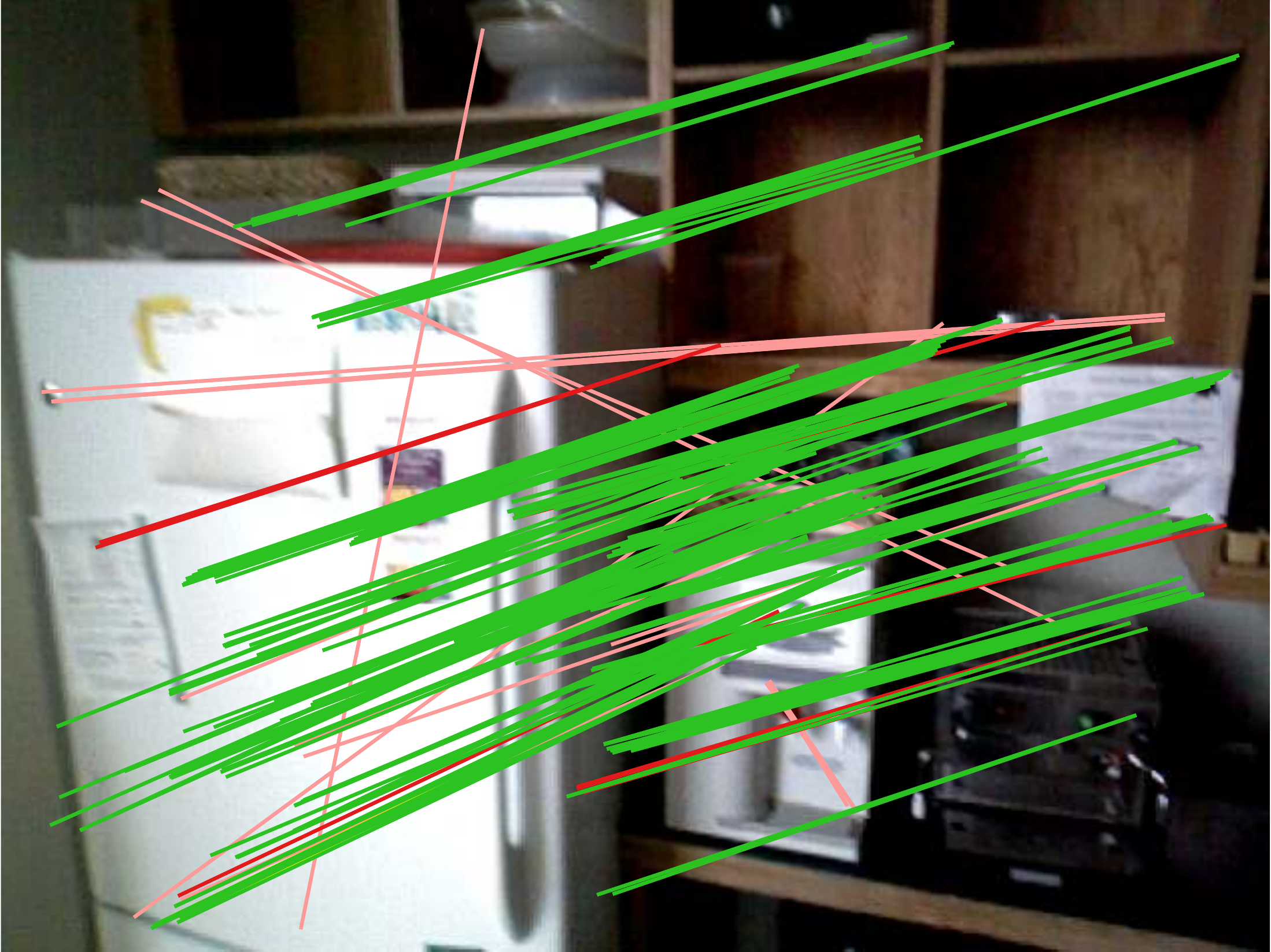}
	\includegraphics[height=7.5em]{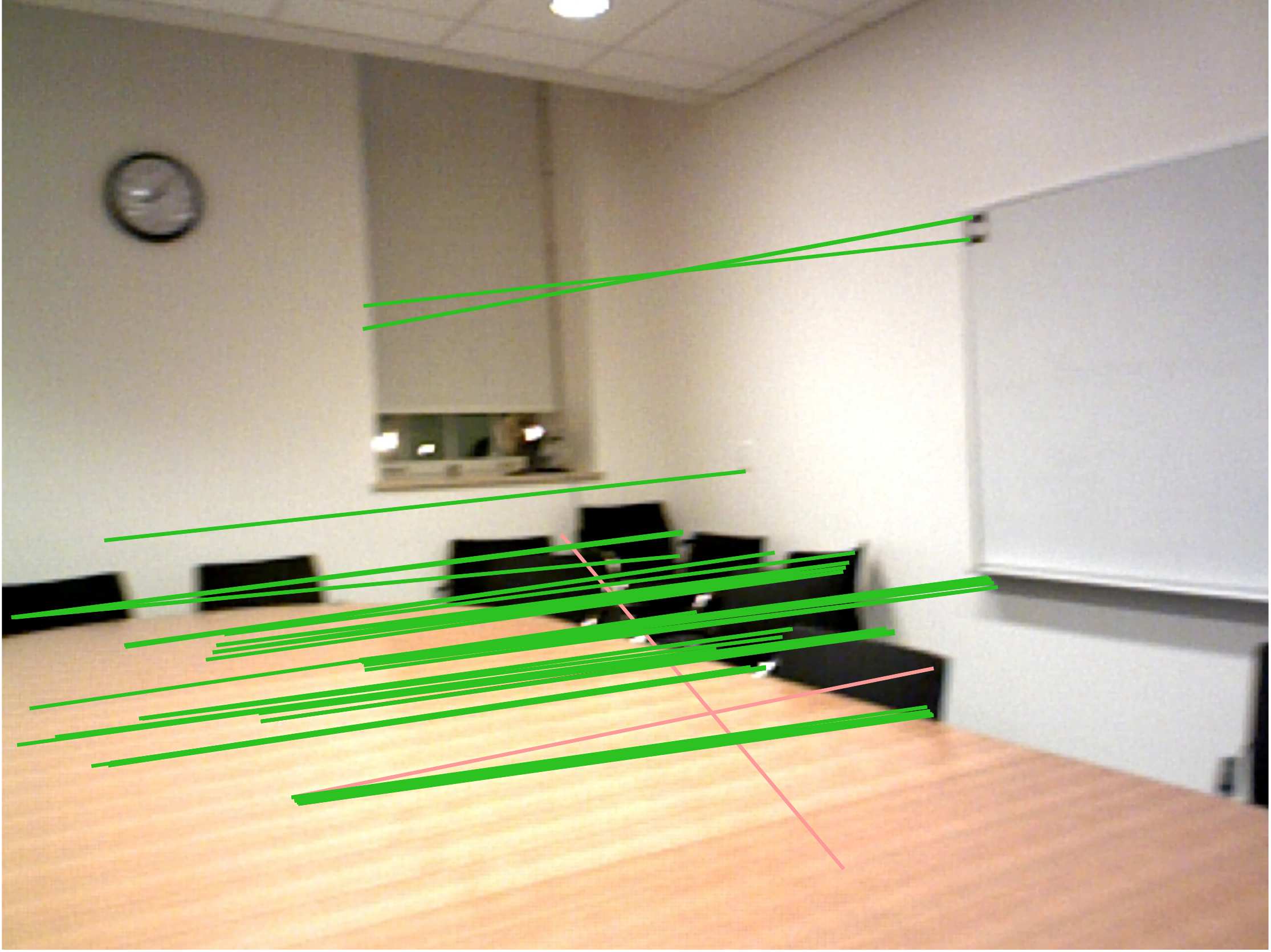}
	\\
	\vspace{0.5em}
	\rotatebox[origin=l]{90}{\mbox{\hspace{2em}GLPM}}
	\includegraphics[height=7.5em]{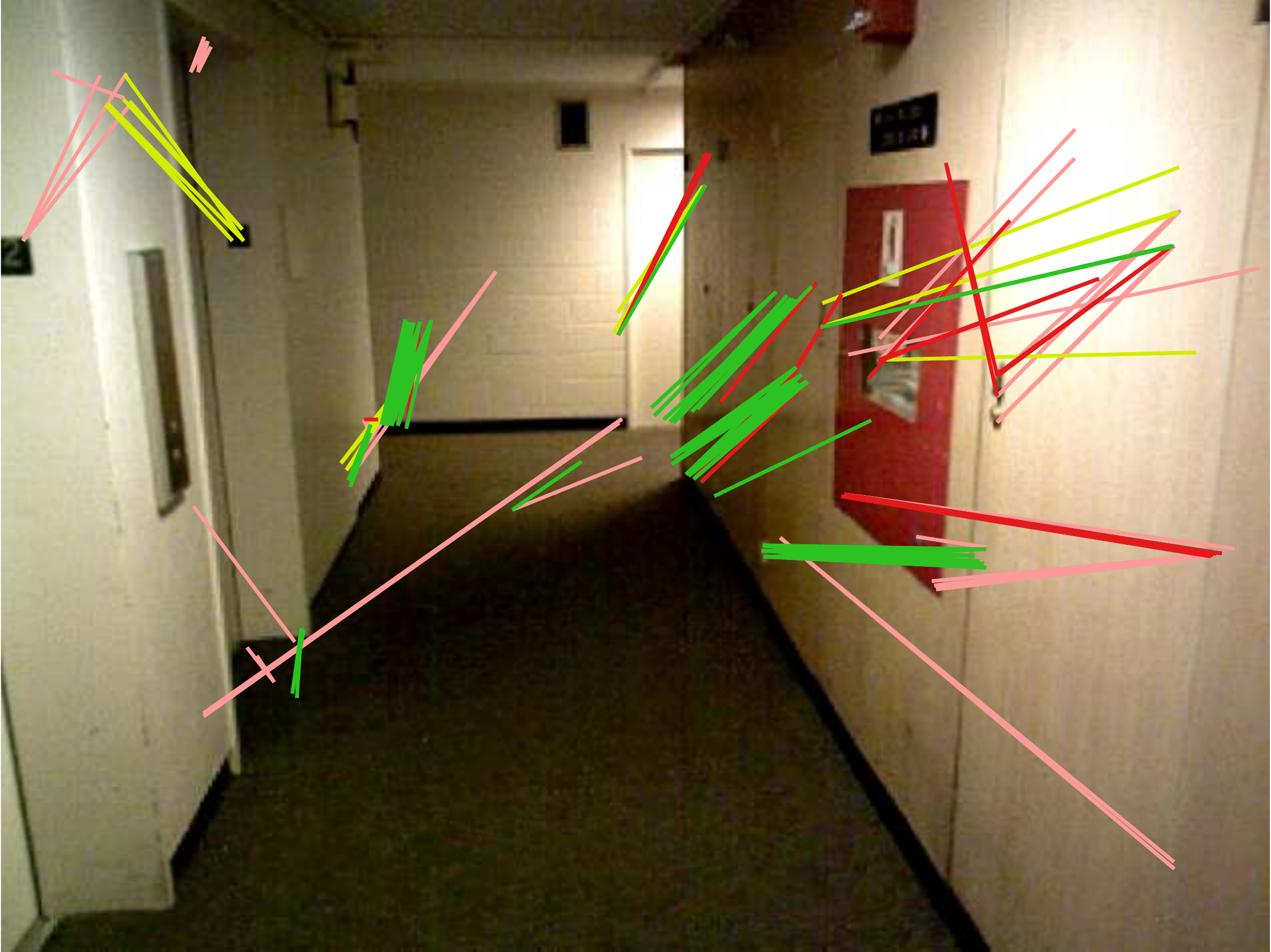}
	\includegraphics[height=7.5em]{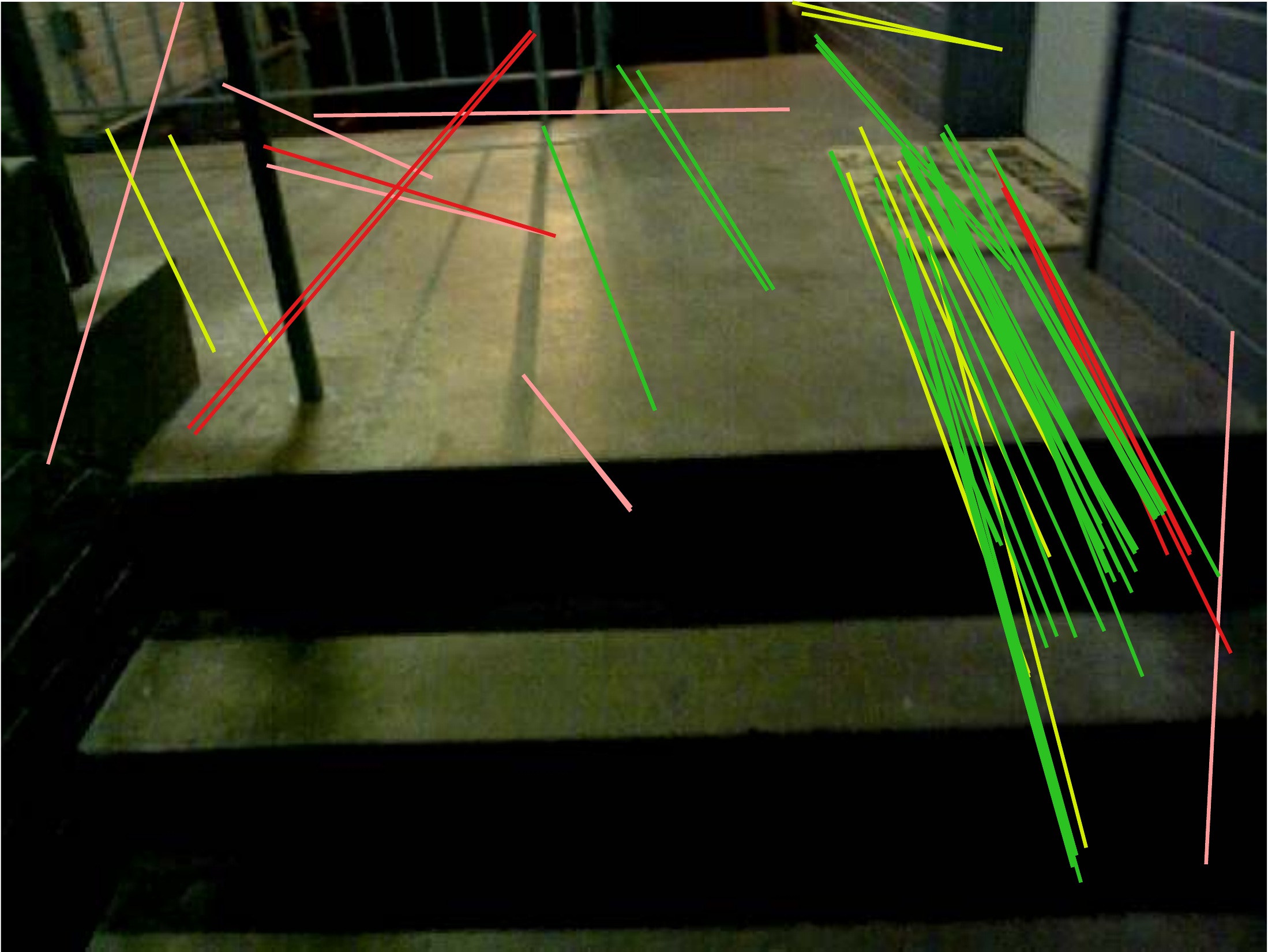}
	\includegraphics[height=7.5em]{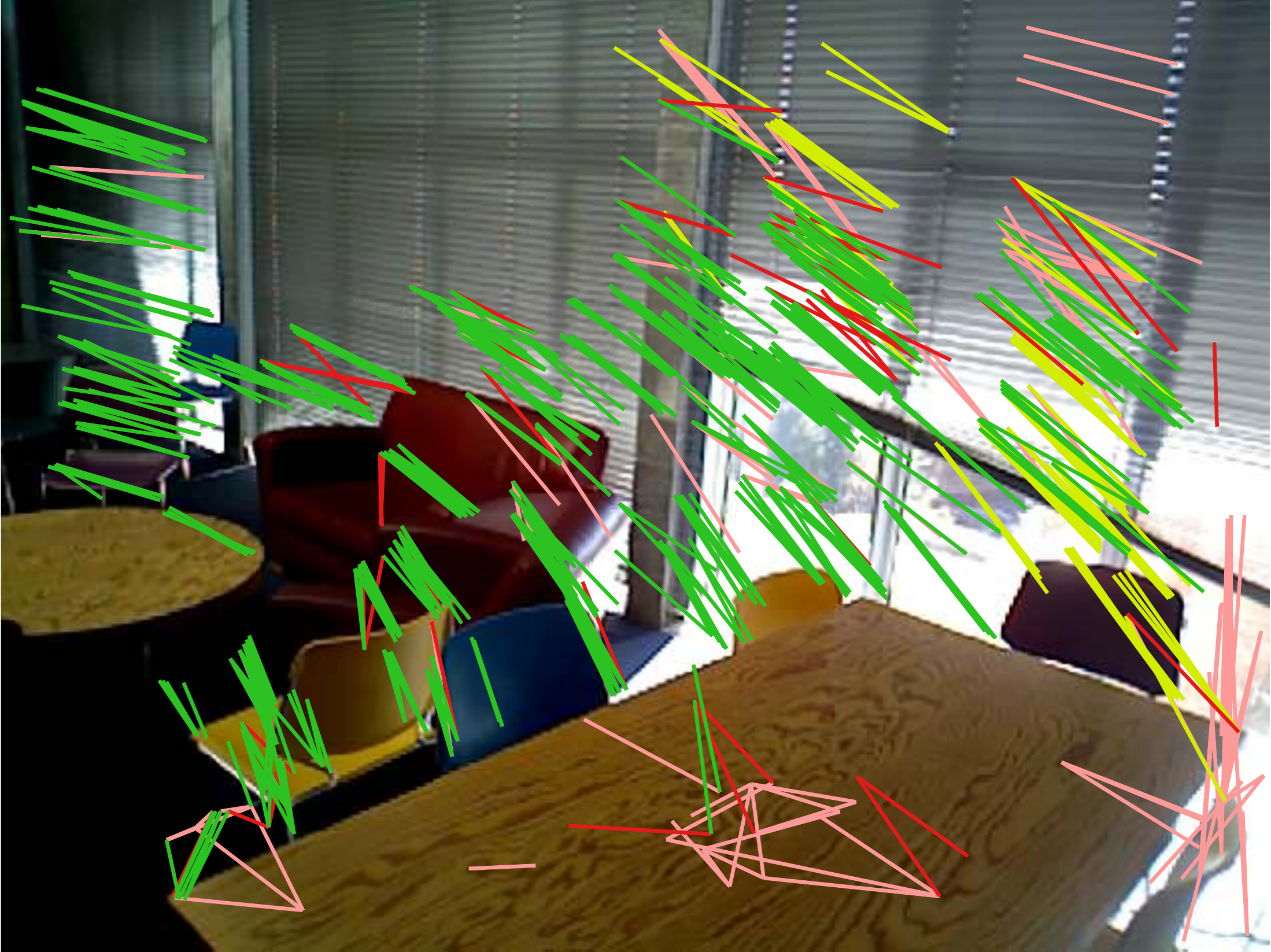}
	\includegraphics[height=7.5em]{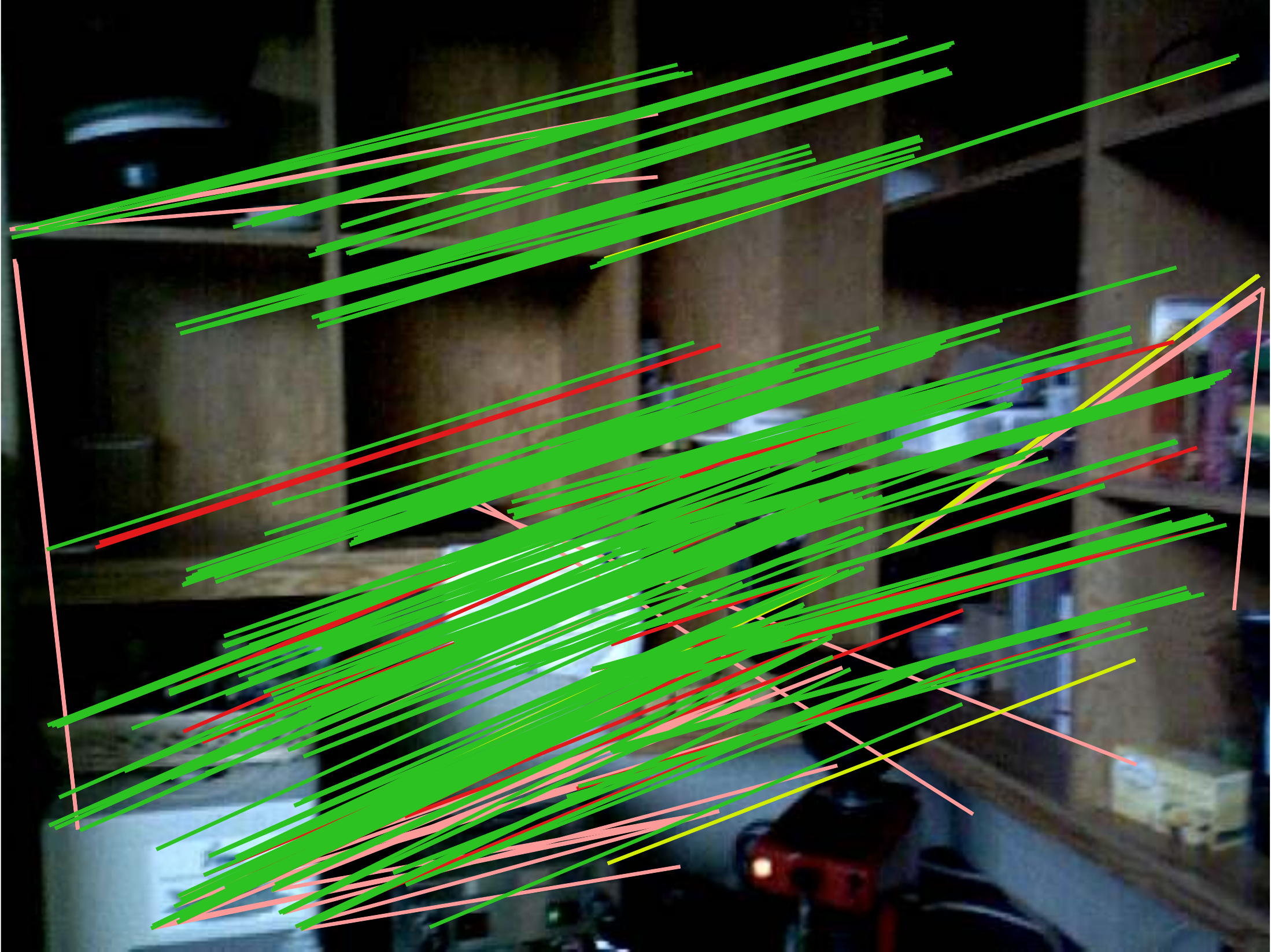}
	\includegraphics[height=7.5em]{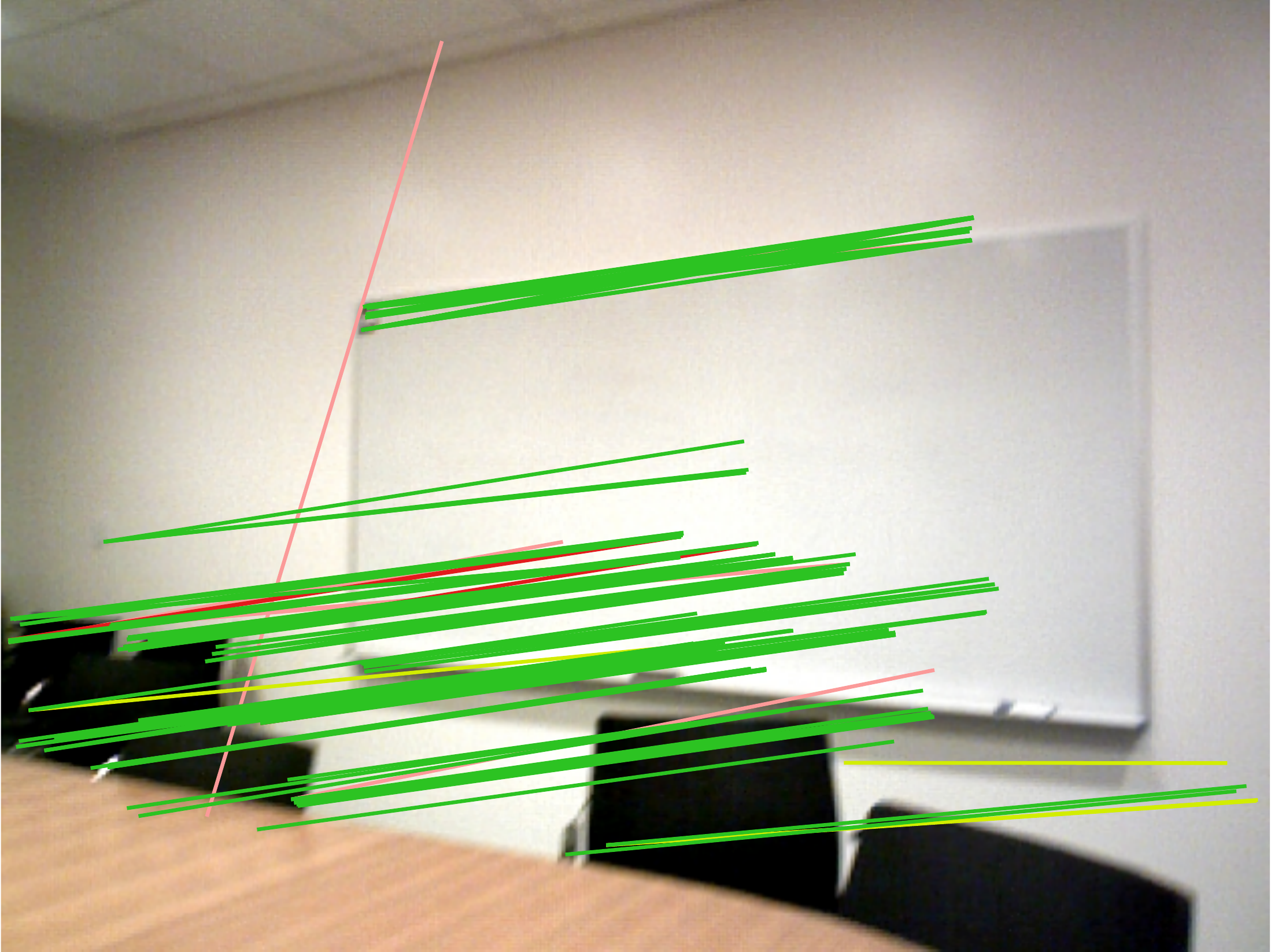}
	\\
	\vspace{0.5em}
	\rotatebox[origin=l]{90}{\mbox{\hspace{2em}GMS}}
	\includegraphics[height=7.5em]{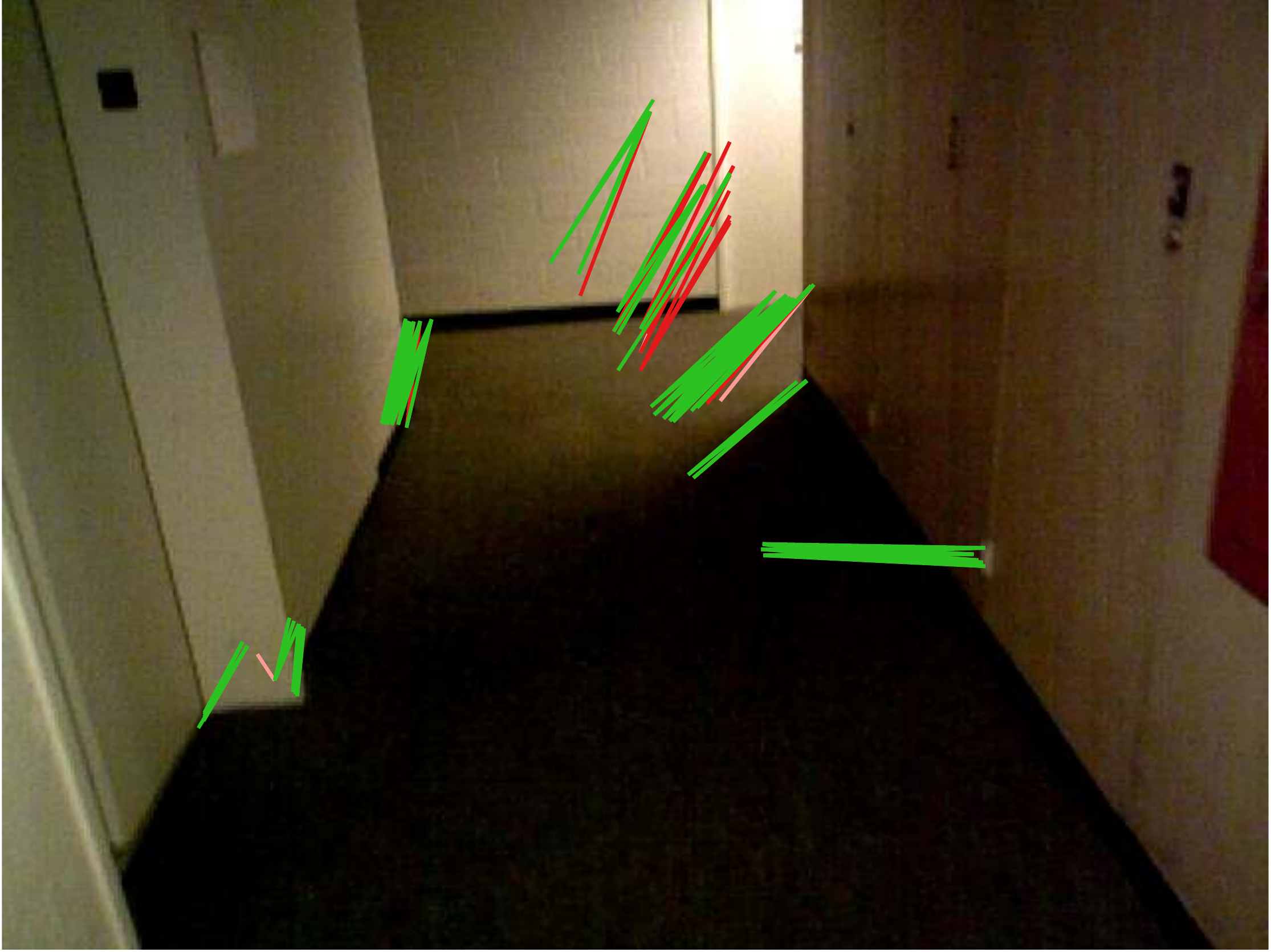}
	\includegraphics[height=7.5em]{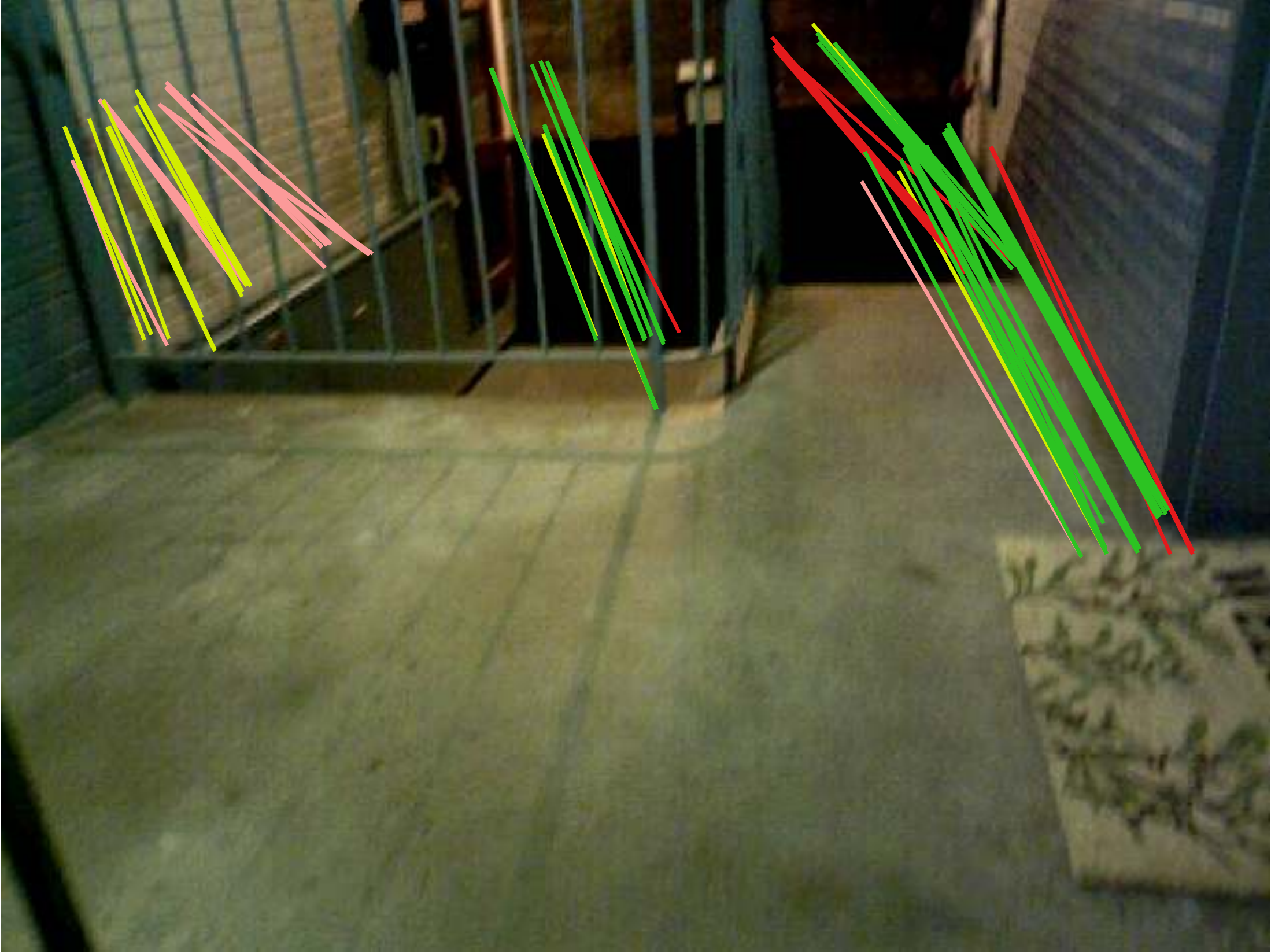}
	\includegraphics[height=7.5em]{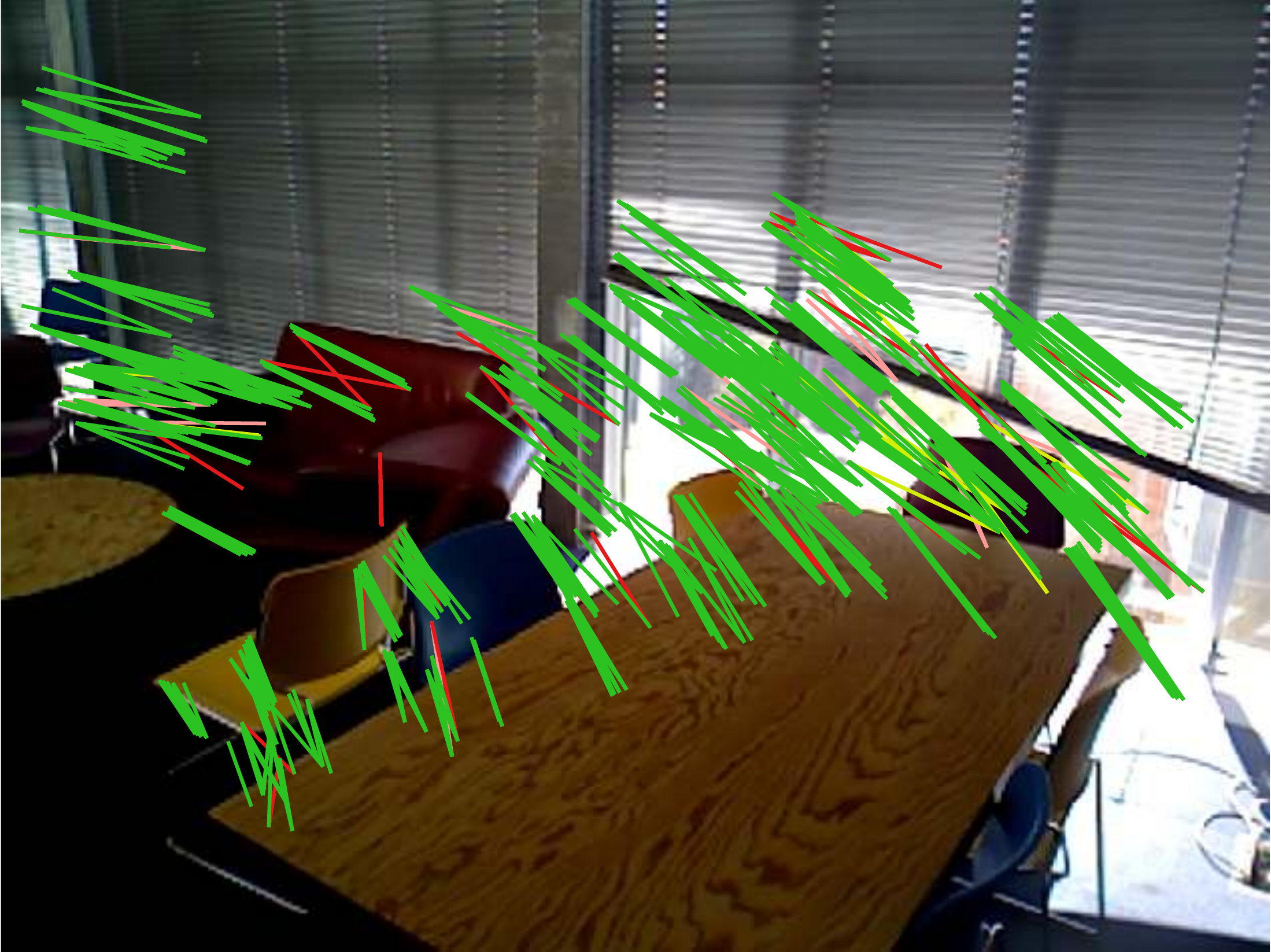}
	\includegraphics[height=7.5em]{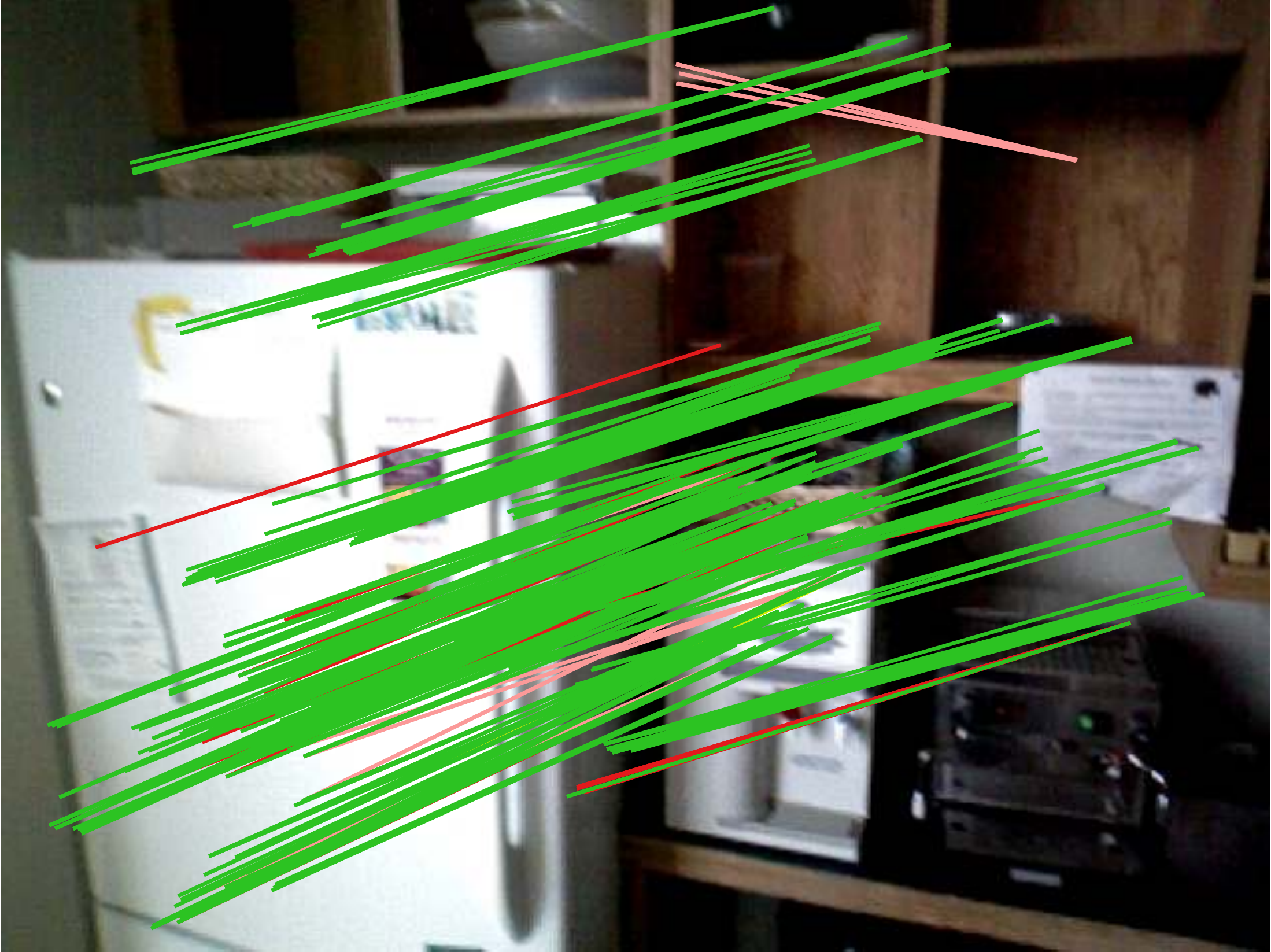}
	\includegraphics[height=7.5em]{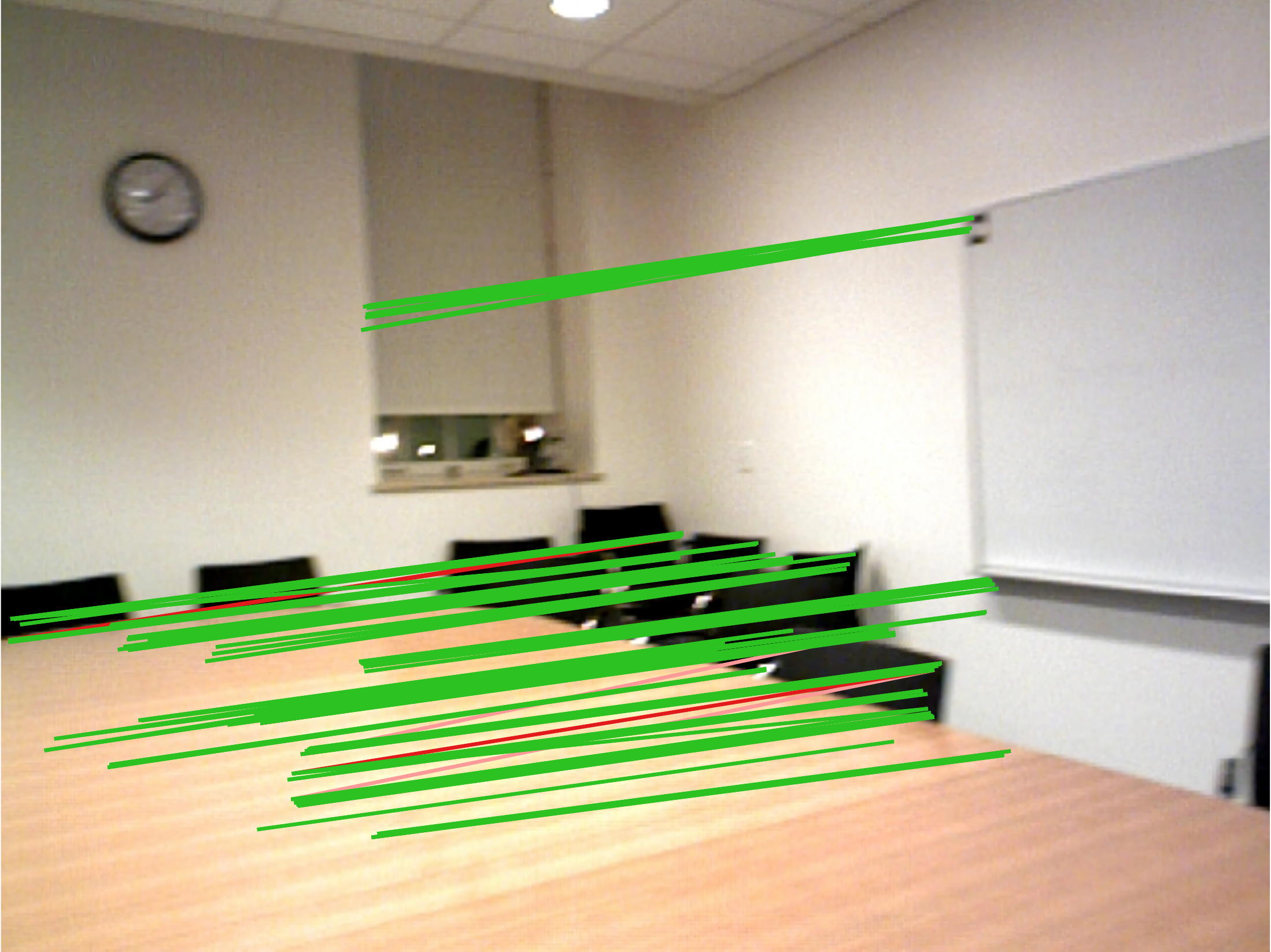}
	\\
	\vspace{0.5em}
	\rotatebox[origin=l]{90}{\mbox{\hspace{2em}VFC}}
	\includegraphics[height=7.5em]{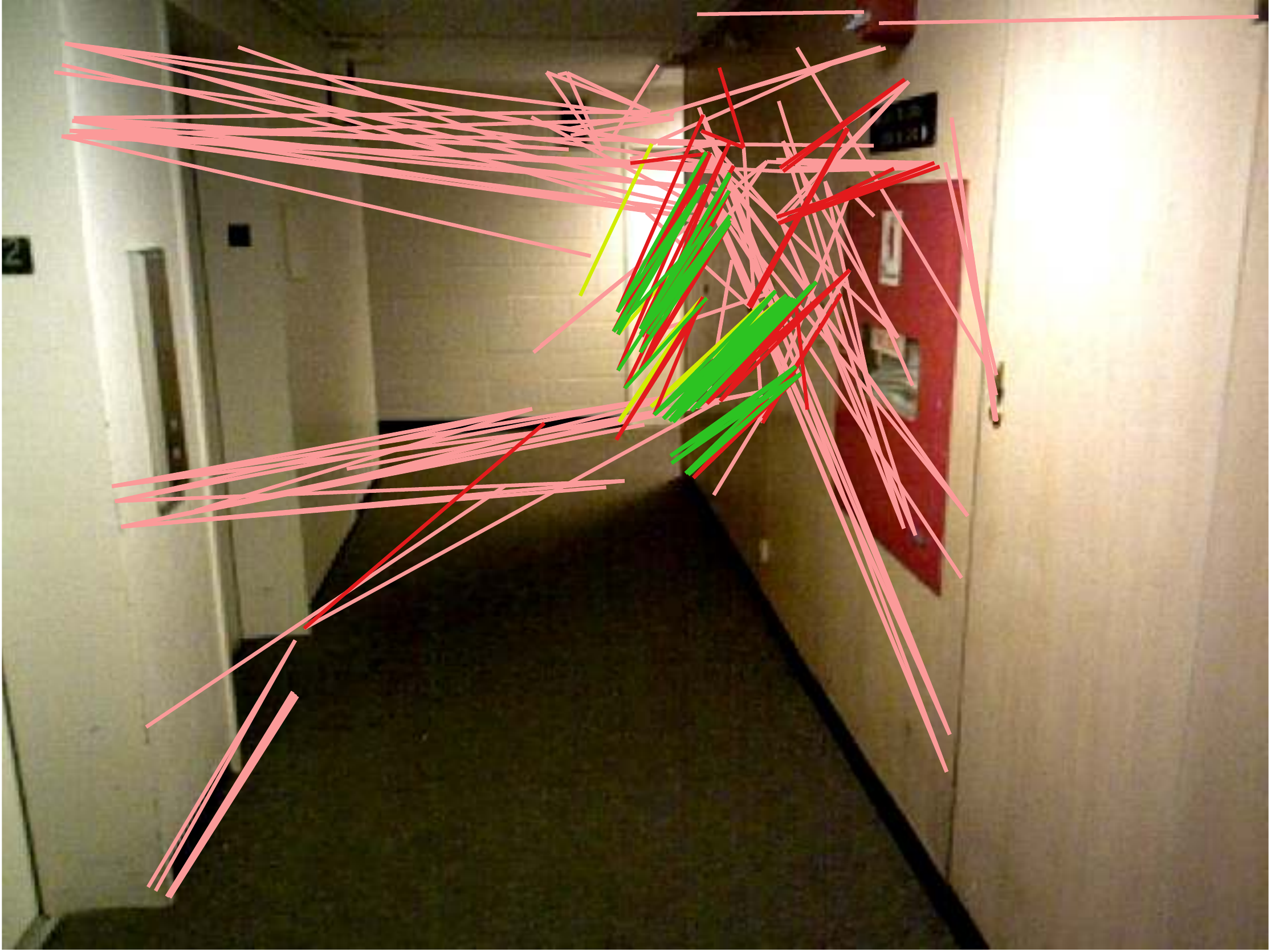}
	\includegraphics[height=7.5em]{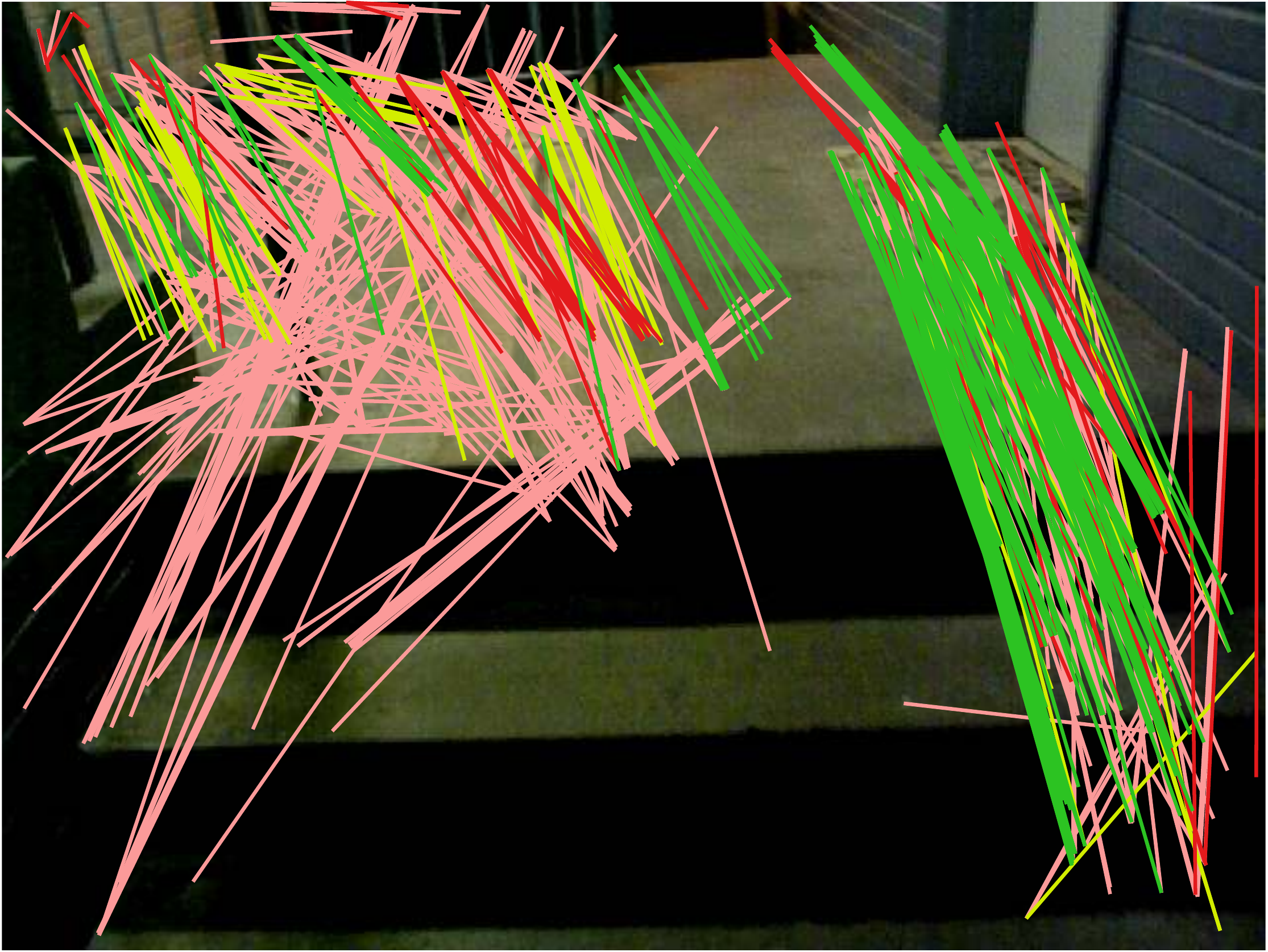}
	\includegraphics[height=7.5em]{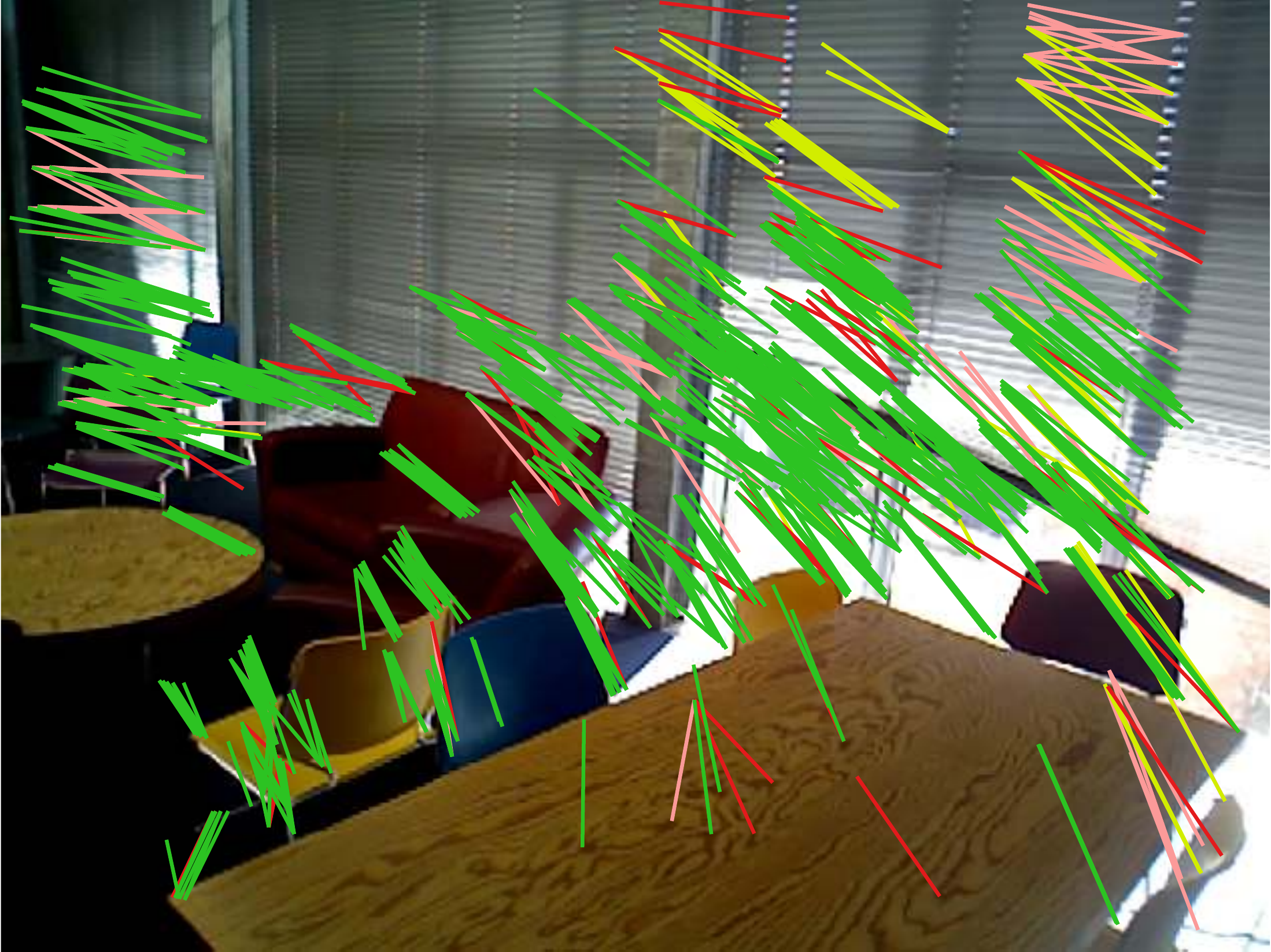}
	\includegraphics[height=7.5em]{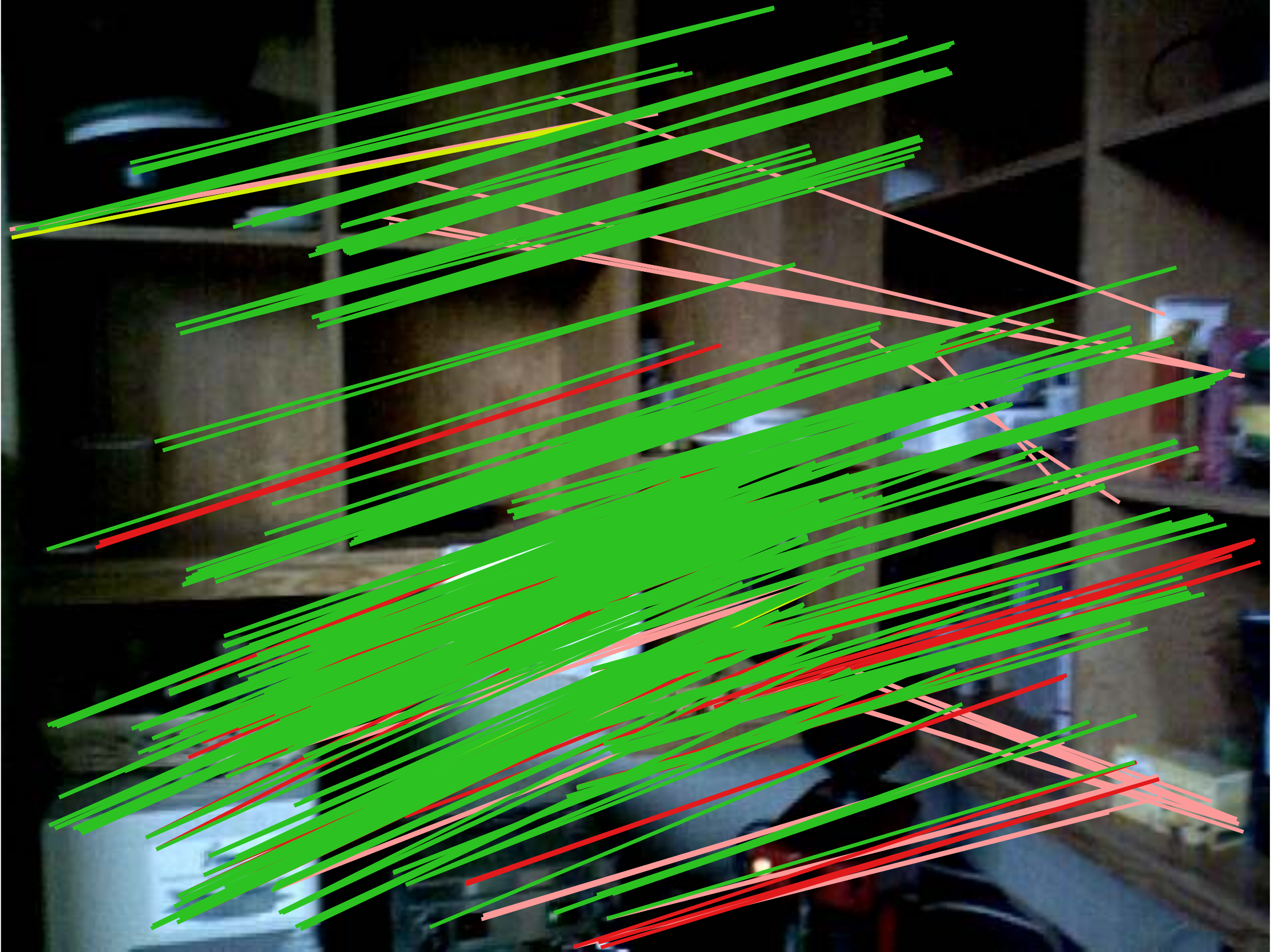}
	\includegraphics[height=7.5em]{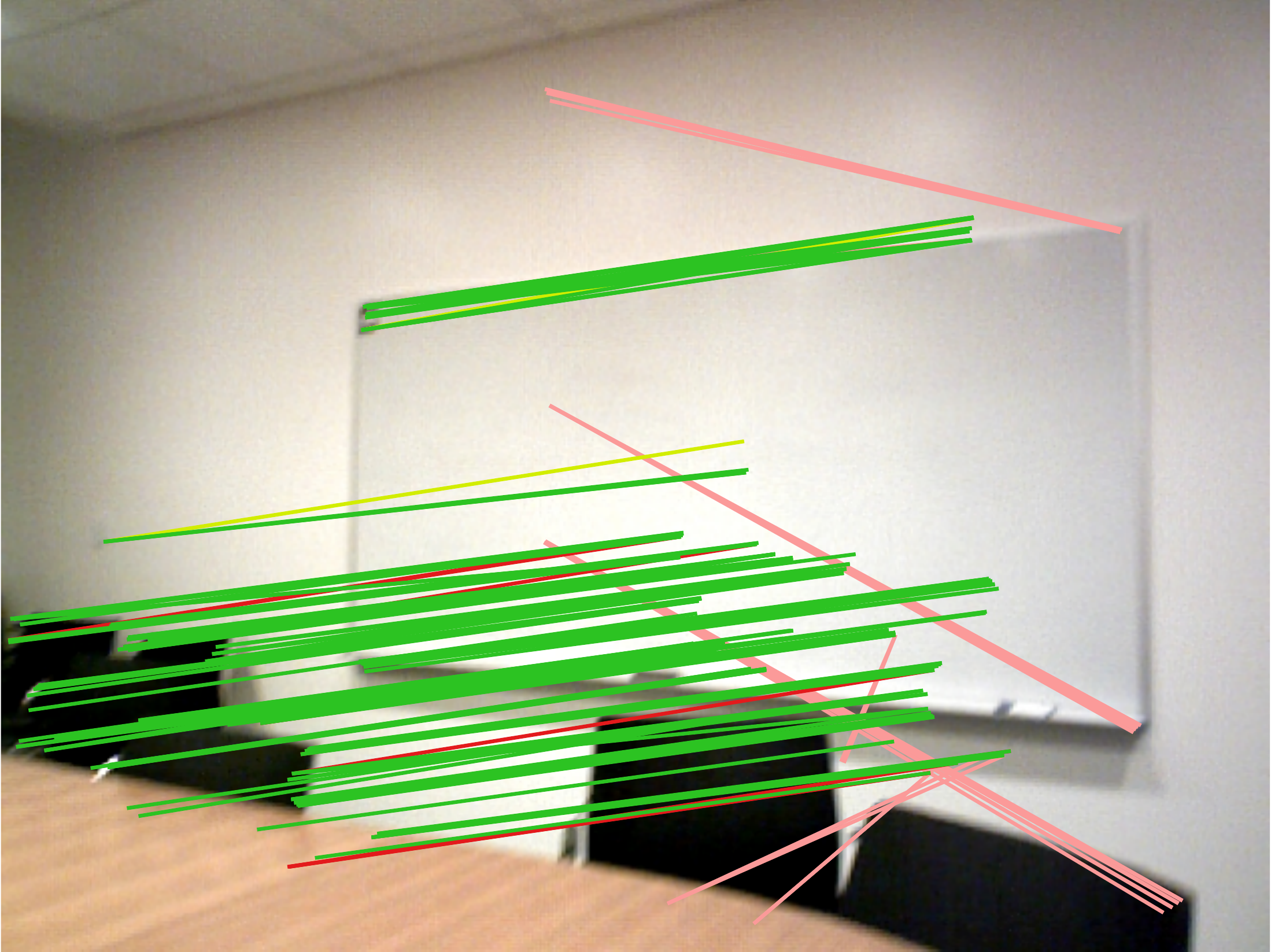}
	\\
	\vspace{0.5em}
	\rotatebox[origin=l]{90}{\mbox{\hspace{2em}LLT}}
	\includegraphics[height=7.5em]{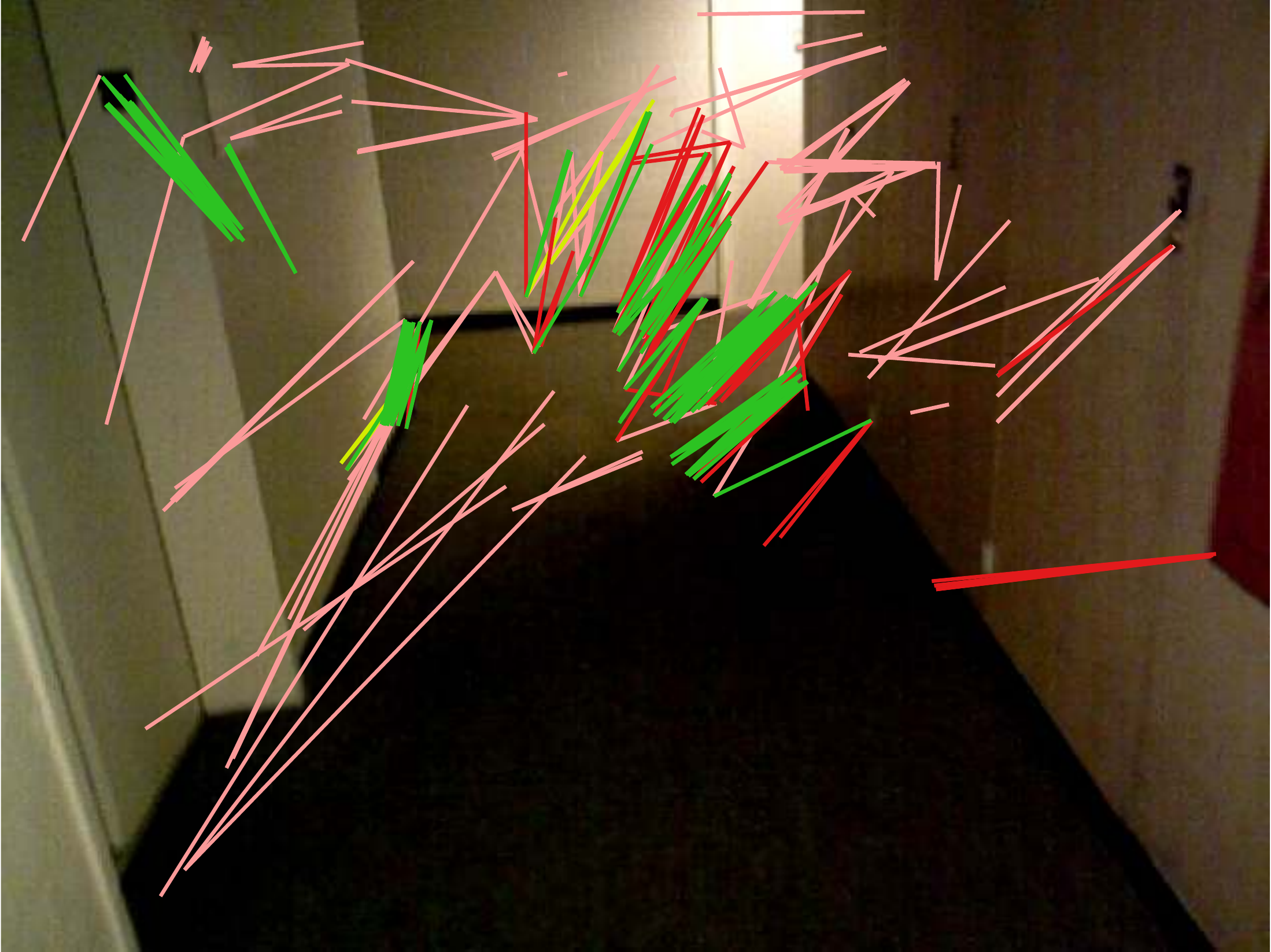}
	\includegraphics[height=7.5em]{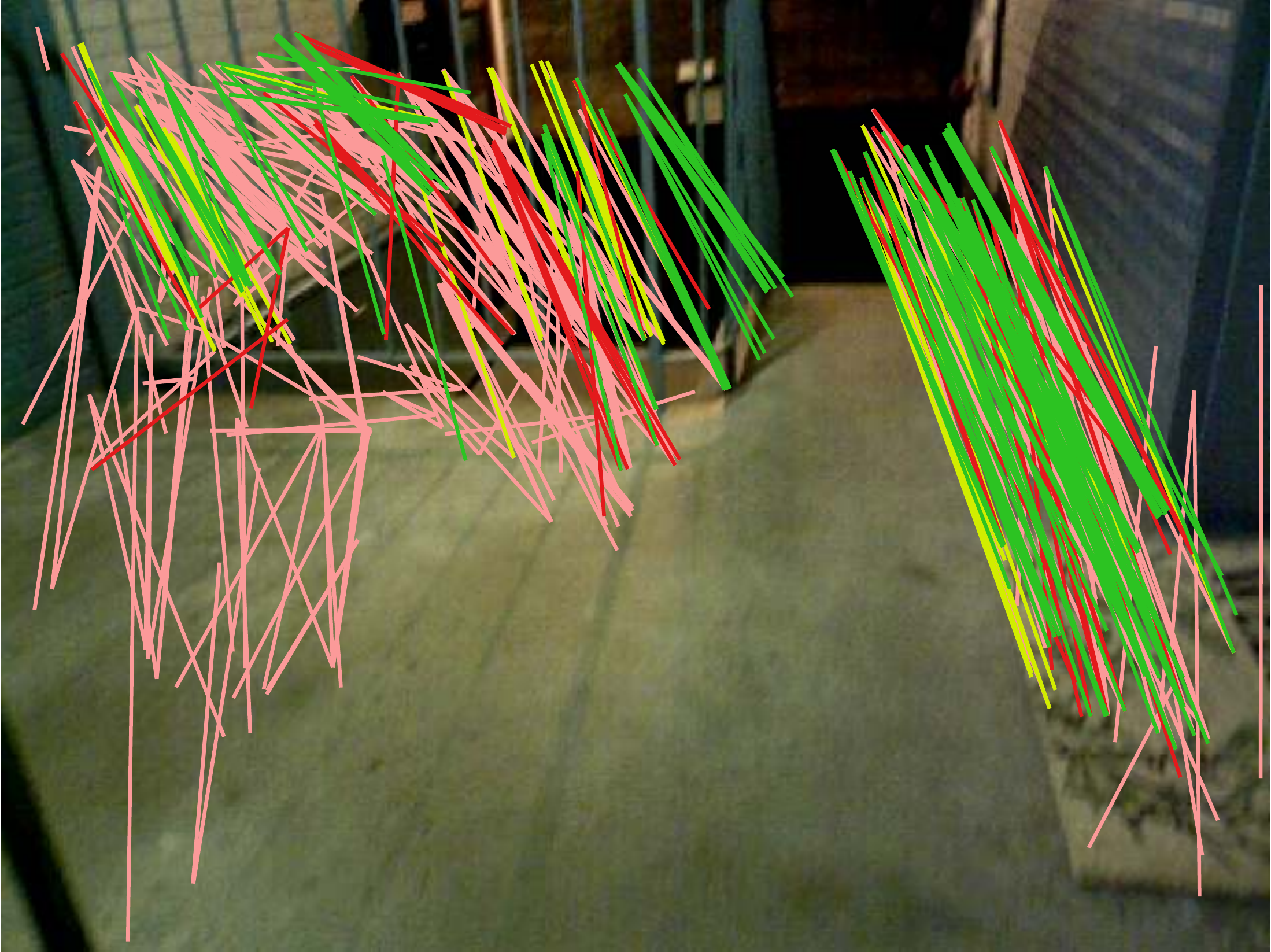}
	\includegraphics[height=7.5em]{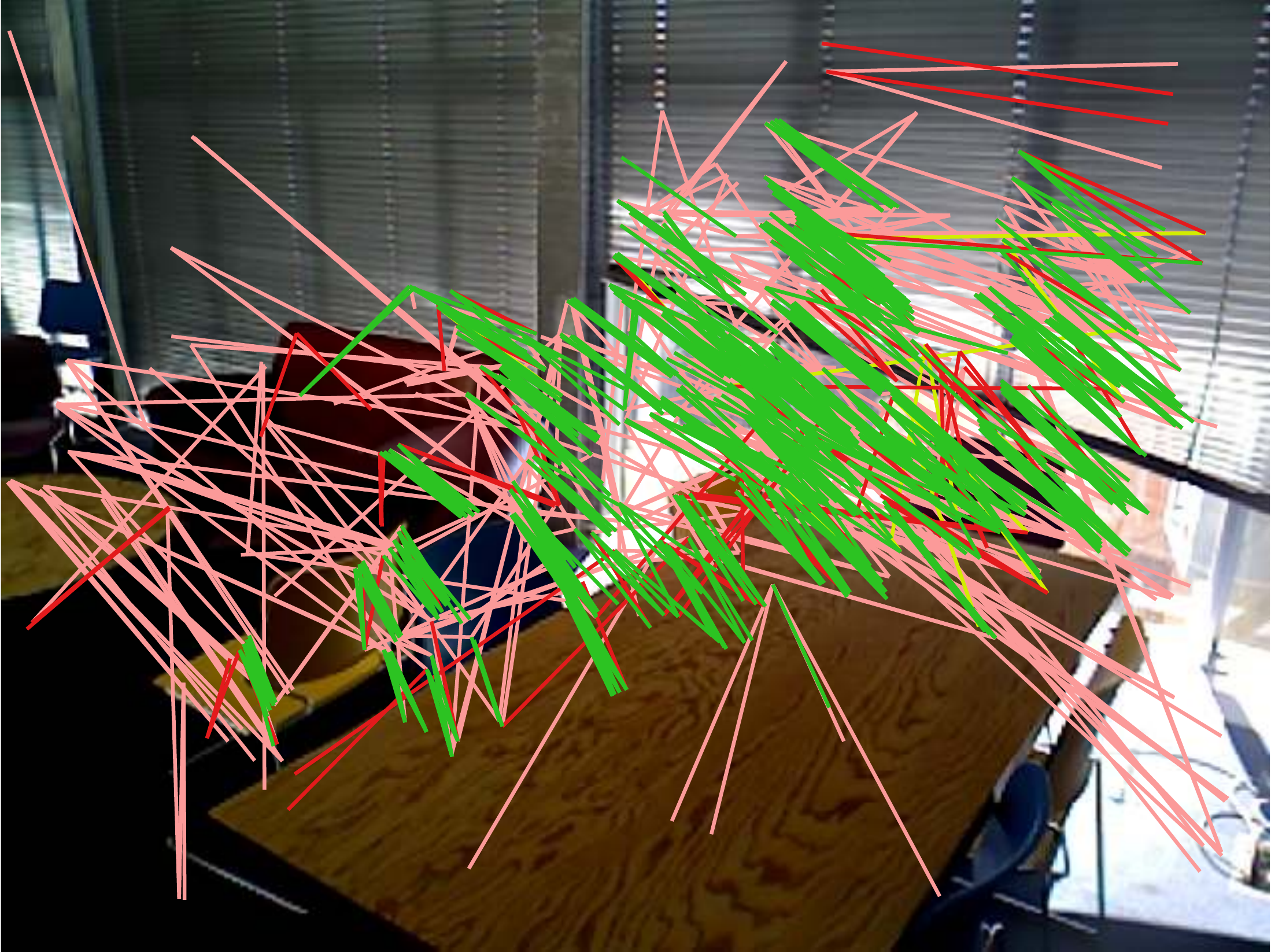}
	\includegraphics[height=7.5em]{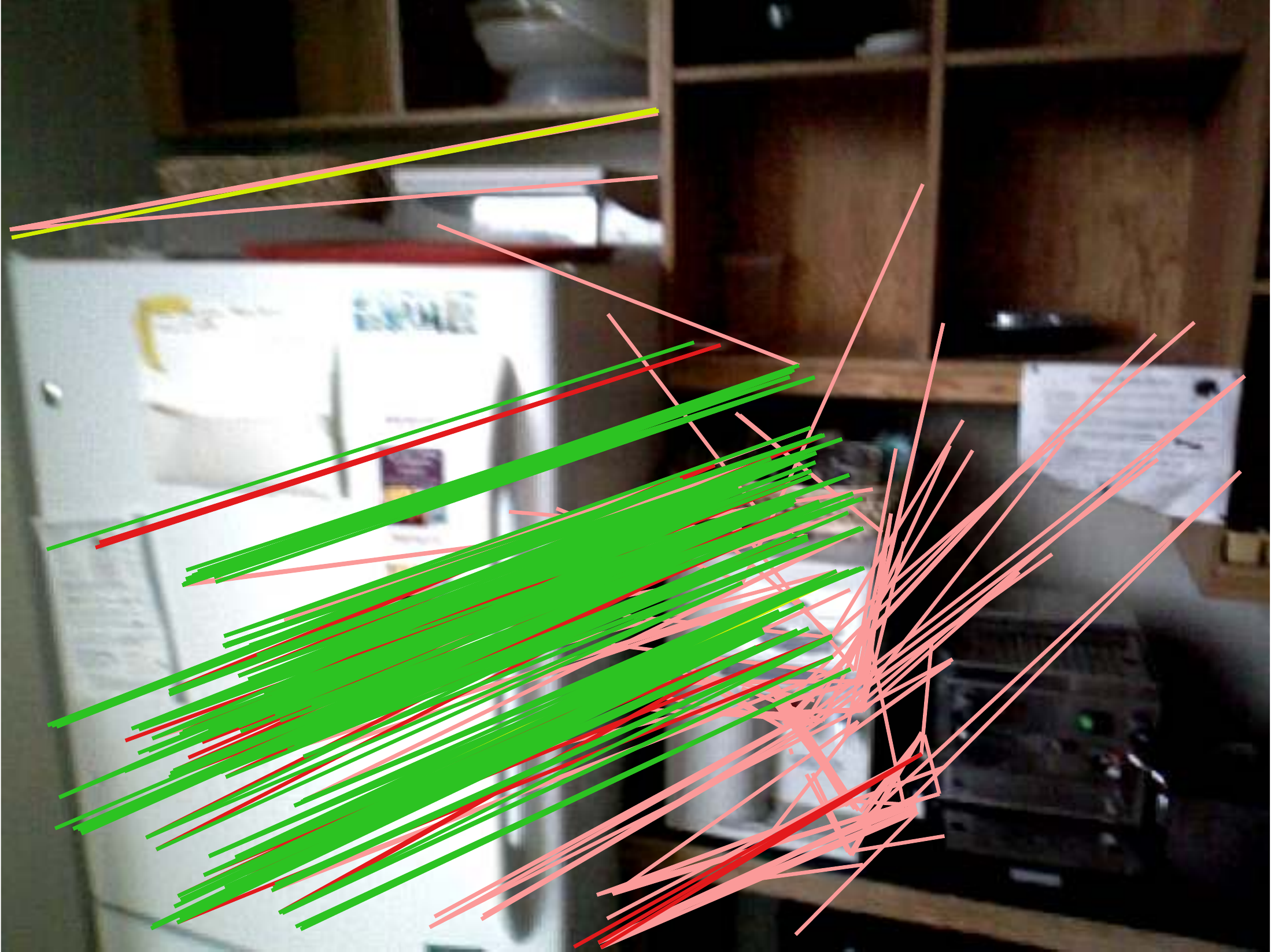}
	\includegraphics[height=7.5em]{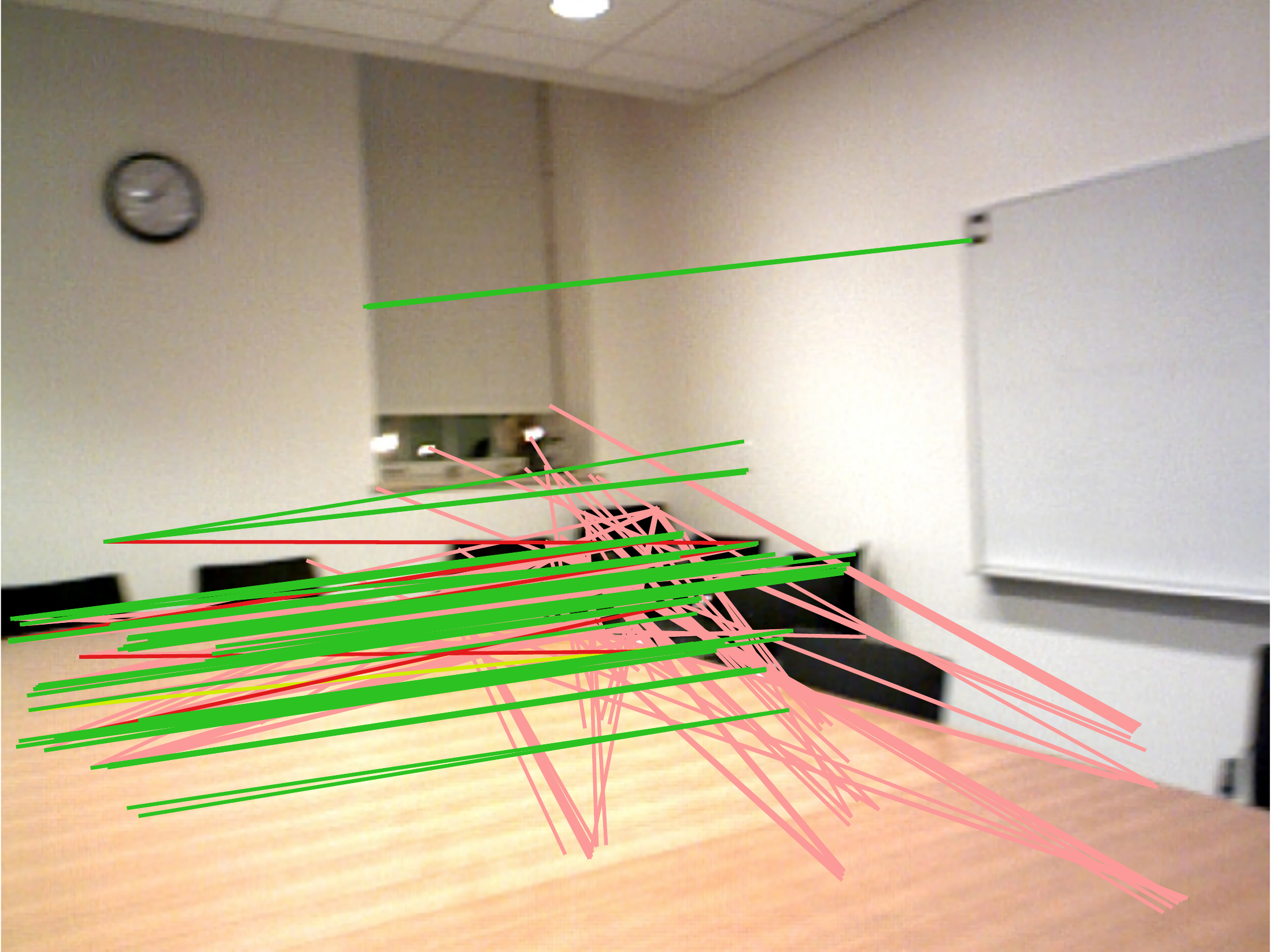}
	\\
	\begin{flushleft}
		\hspace*{4em}MIT Corridor\hspace{4.75em}MIT Stairs\hspace{4.75em}MIT LoungeA \hspace{4em}MIT LoungeB\hspace{1.75em}Brown Cognitive Science
	\end{flushleft}
	\caption{\label{example_4a}
		SUN3D local spatial filter matches according to the best configuration setup, the images of the input pair alternate among the rows. For each method inlier (yellow, green) and outlier (red and light red) clusters are shown, as well as the 1SAC filtered matches (green, red) (see Sec.~\ref{eval_dt}, best viewed in color and zoomed in).}
\end{figure*}

\begin{figure*}
	\center
	\rotatebox[origin=l]{90}{\mbox{\hspace{0.5em}RFM-SCAN}}
	\includegraphics[height=7.5em]{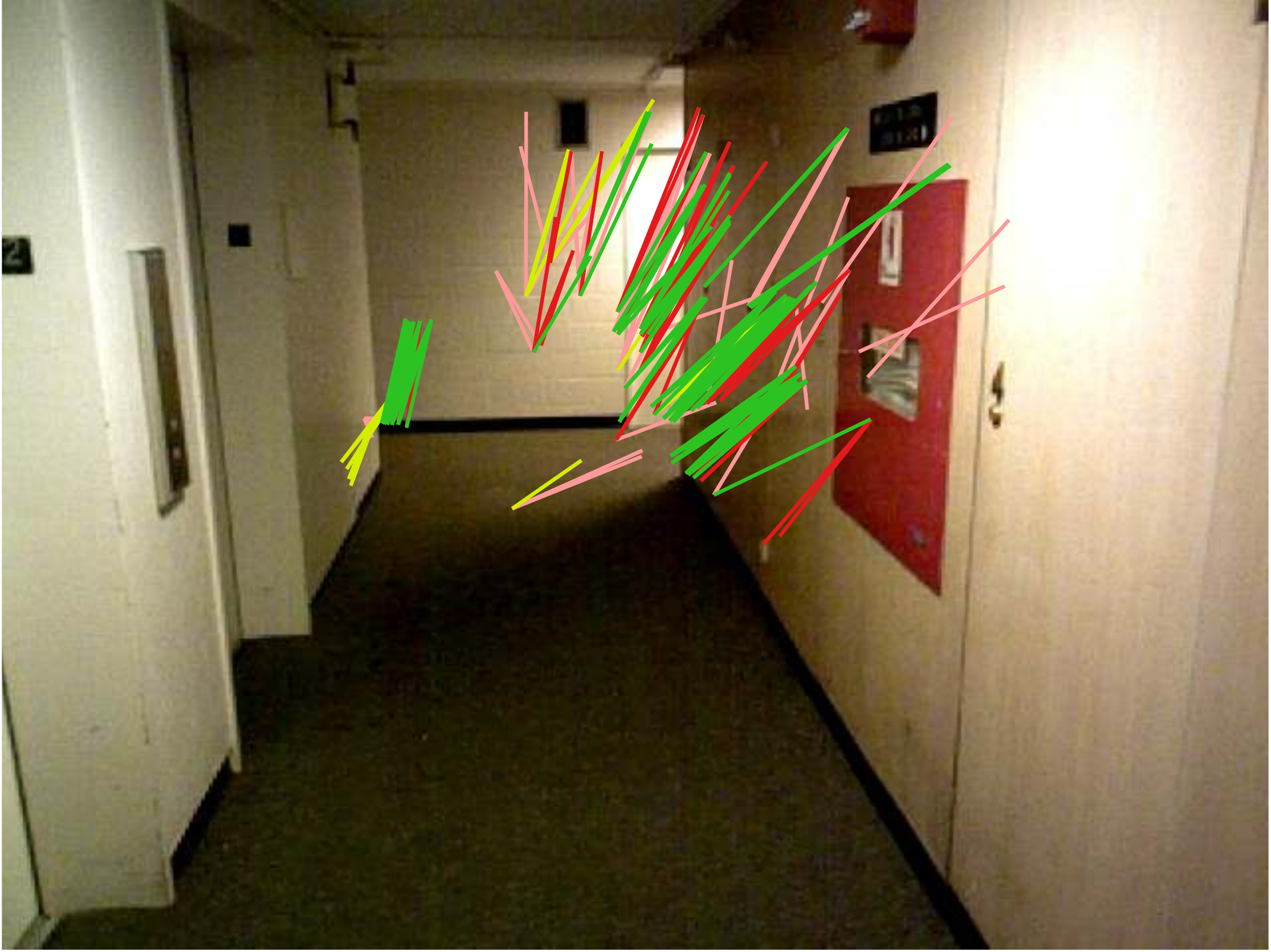}
	\includegraphics[height=7.5em]{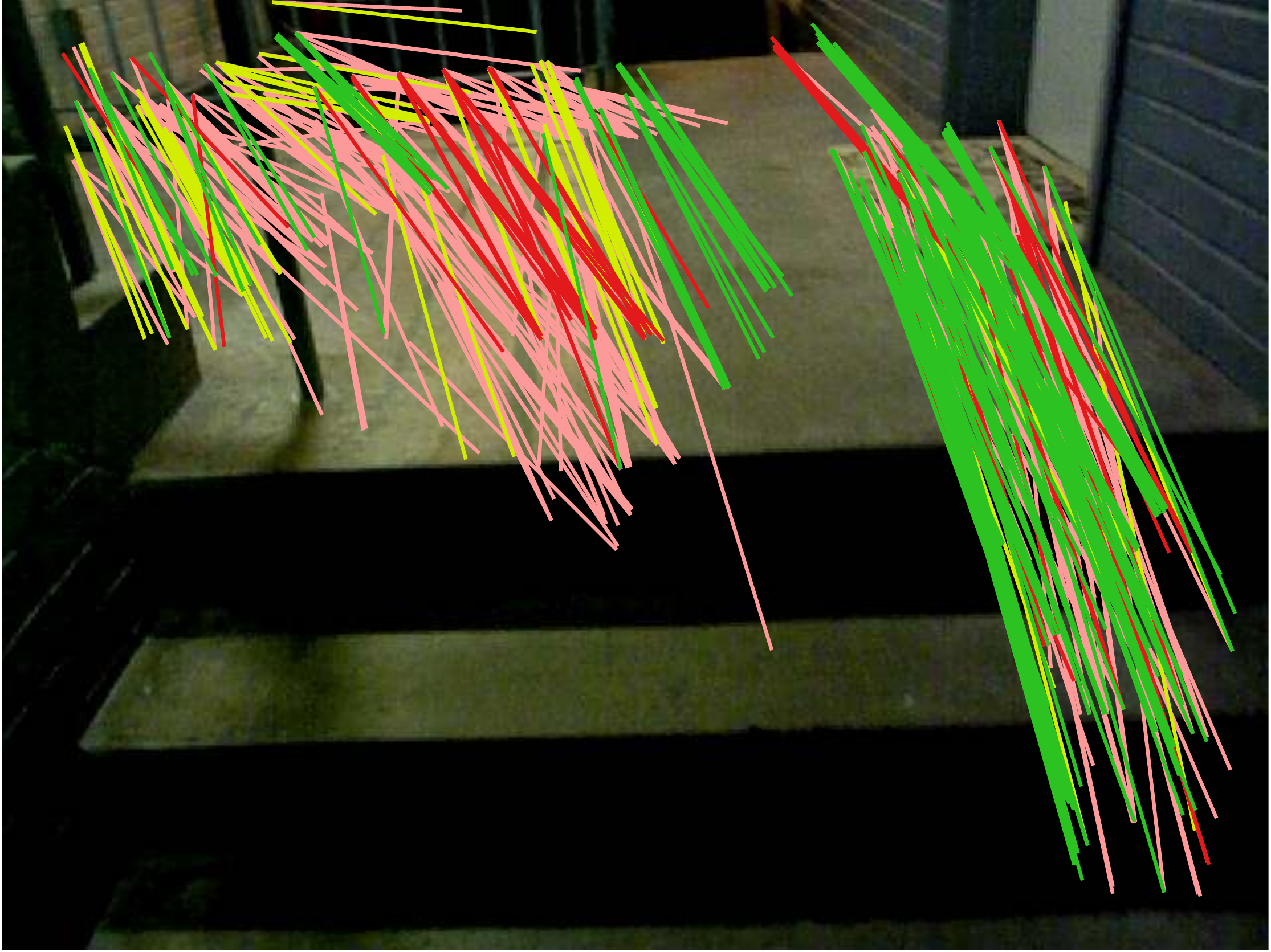}
	\includegraphics[height=7.5em]{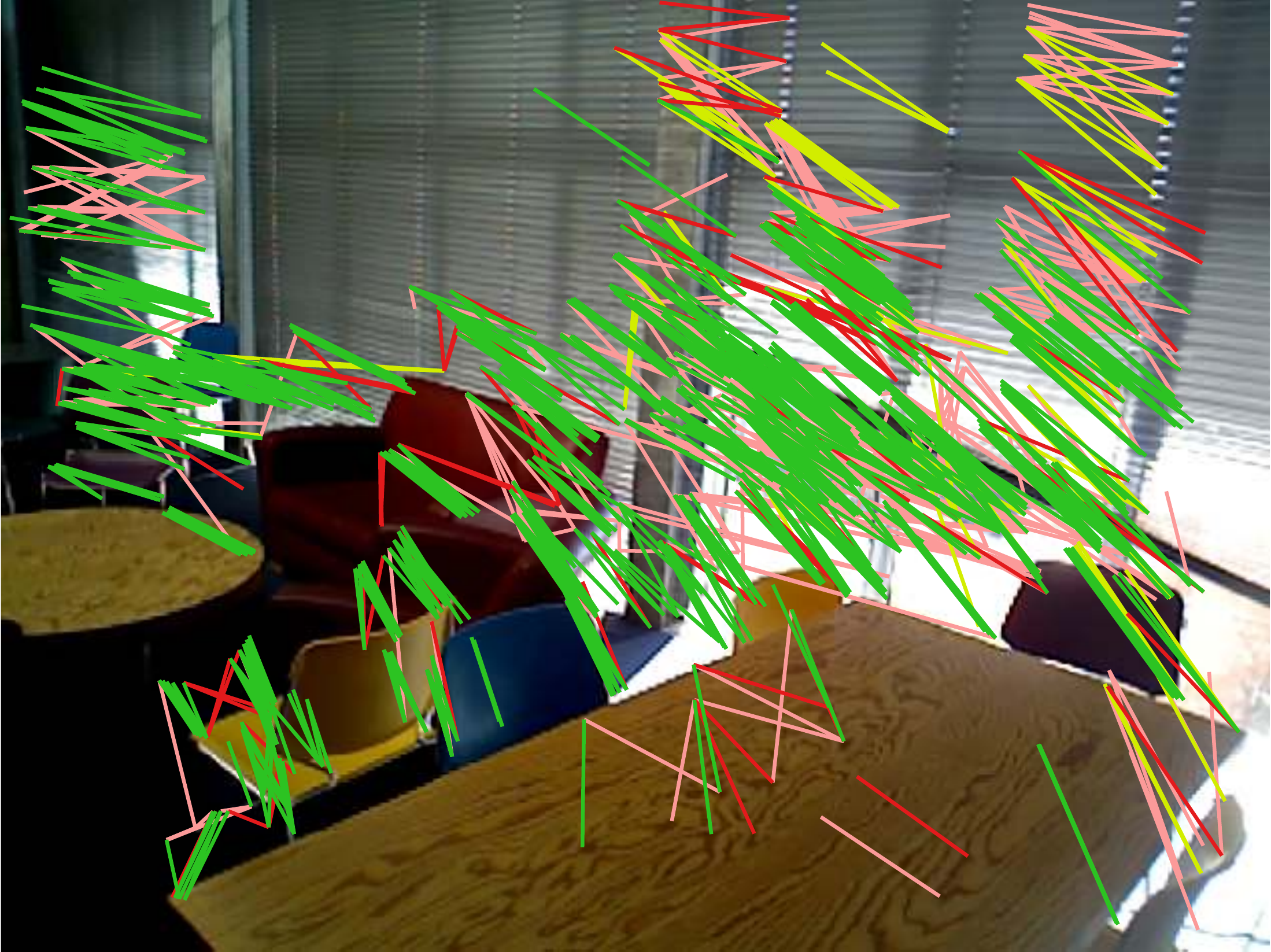}
	\includegraphics[height=7.5em]{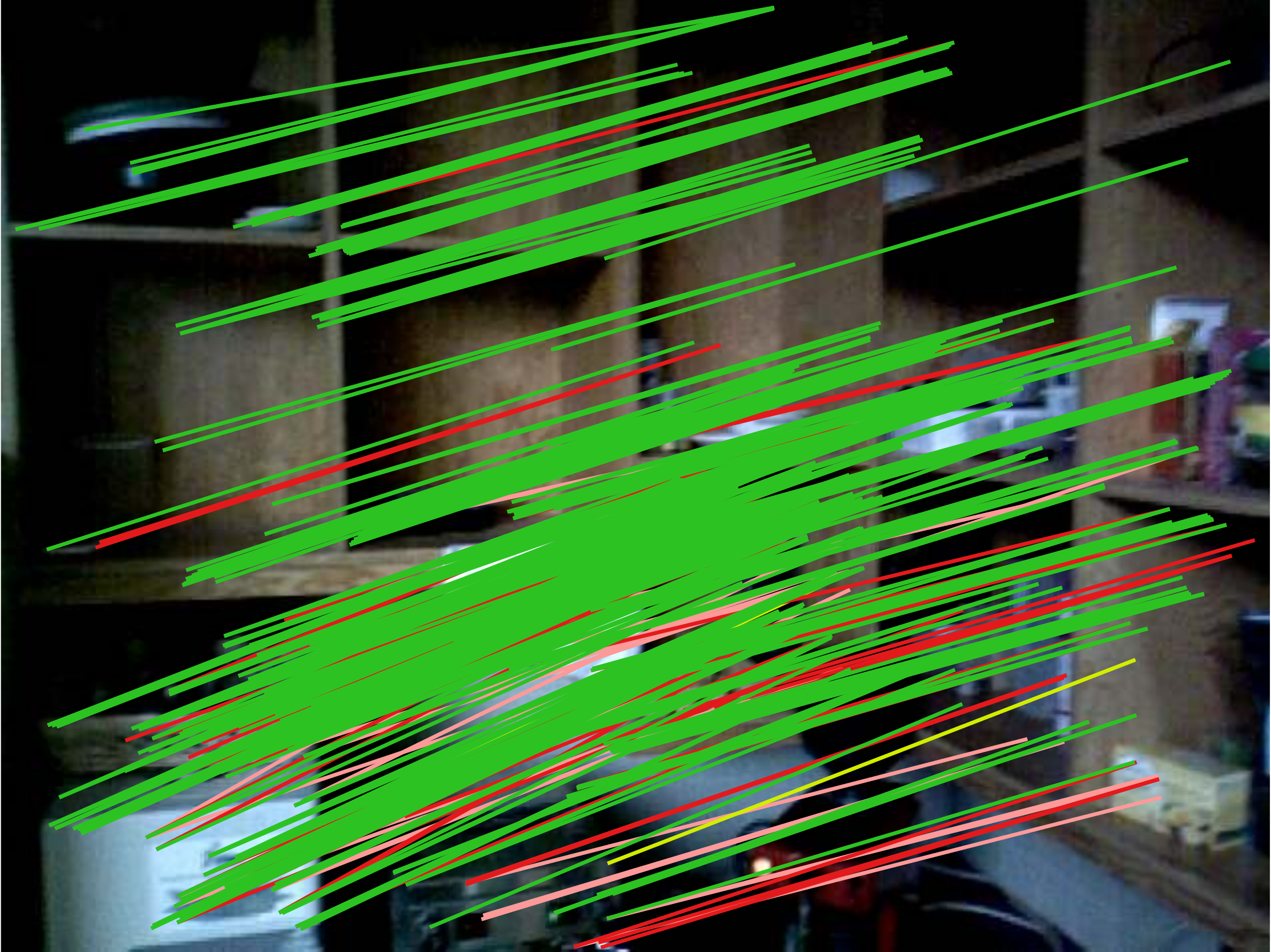}
	\includegraphics[height=7.5em]{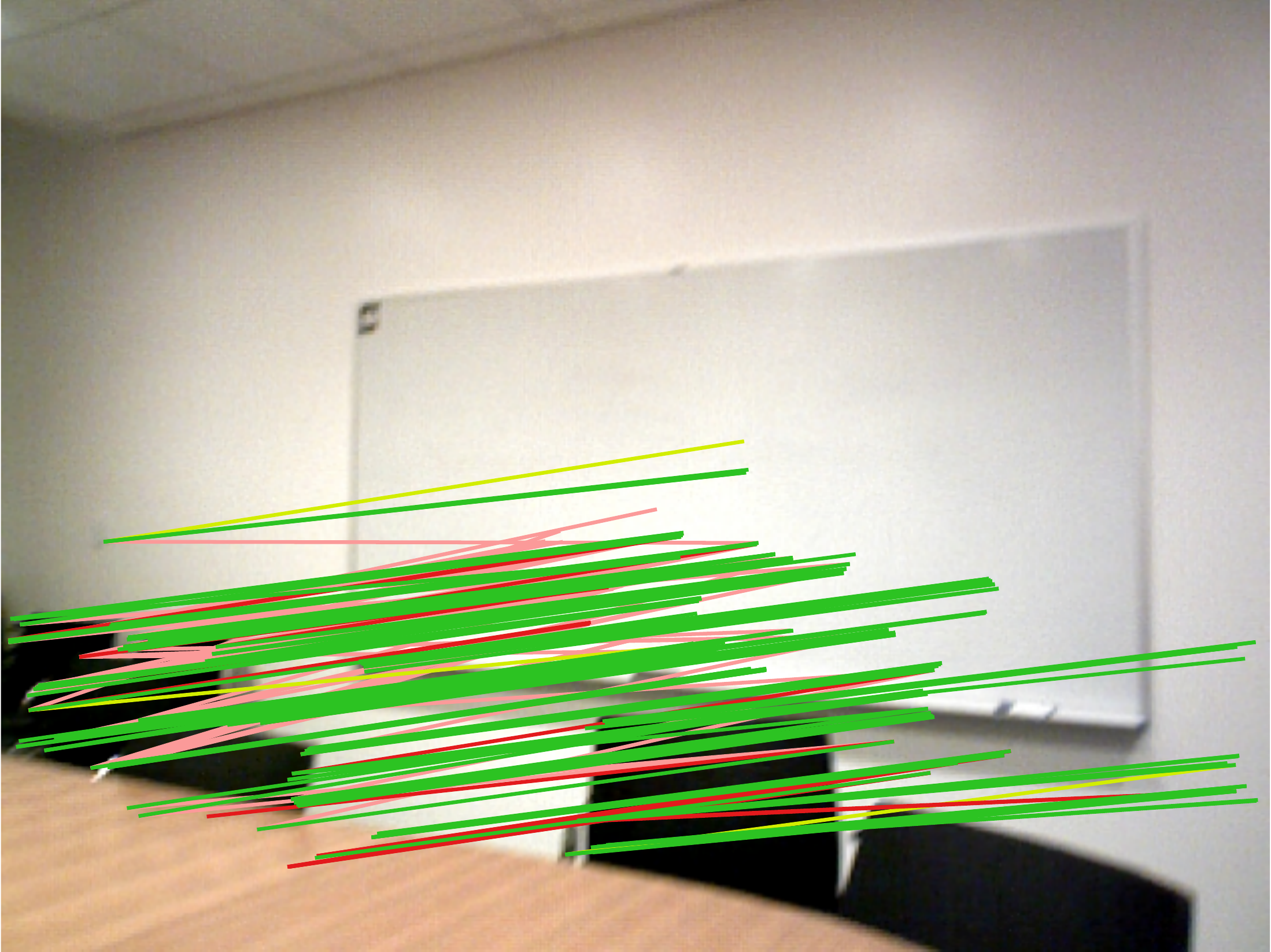}
	\\
	\vspace{0.5em}
	\rotatebox[origin=l]{90}{\mbox{\hspace{1em}AdaLAM}}
	\includegraphics[height=7.5em]{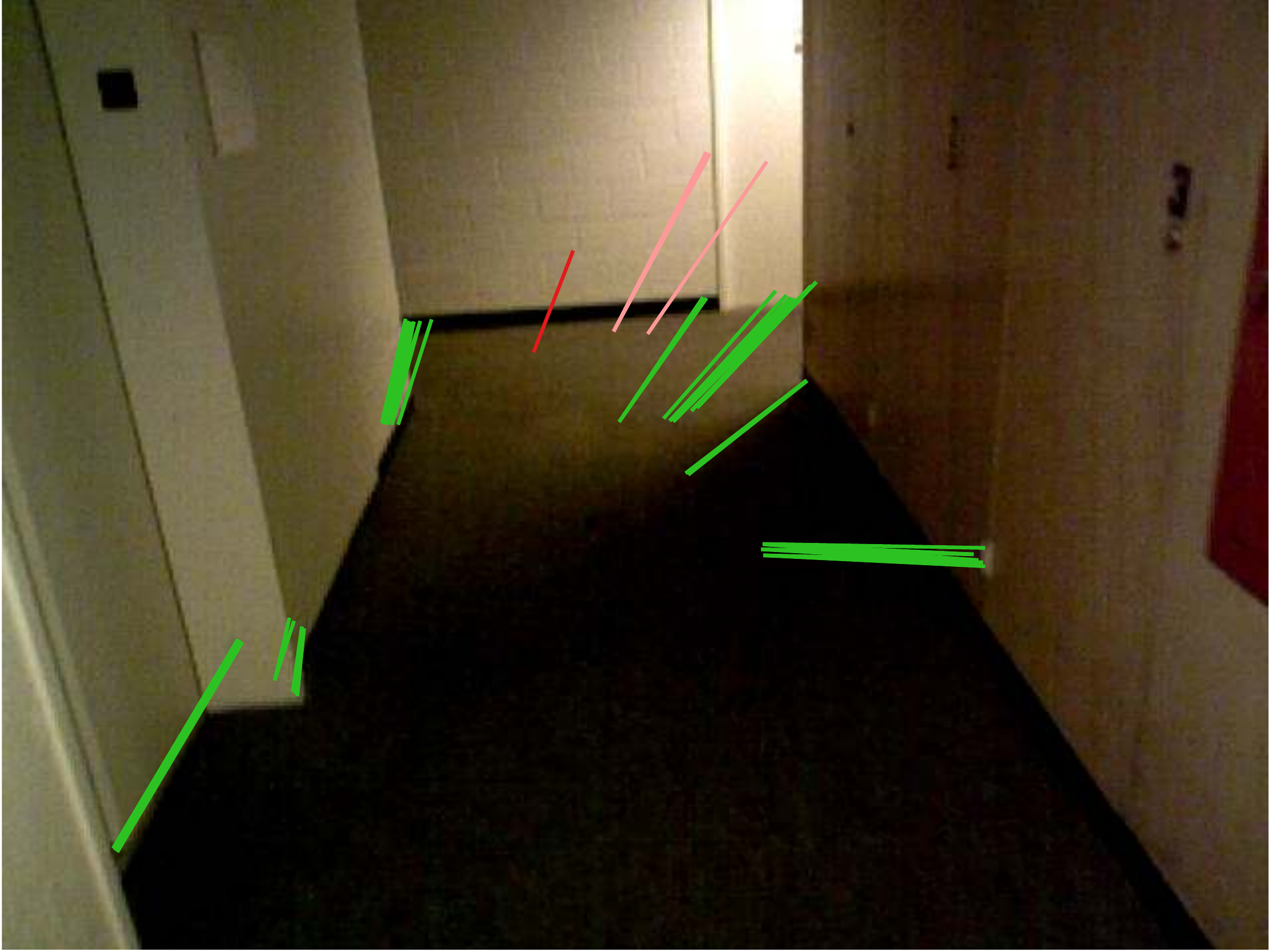}
	\includegraphics[height=7.5em]{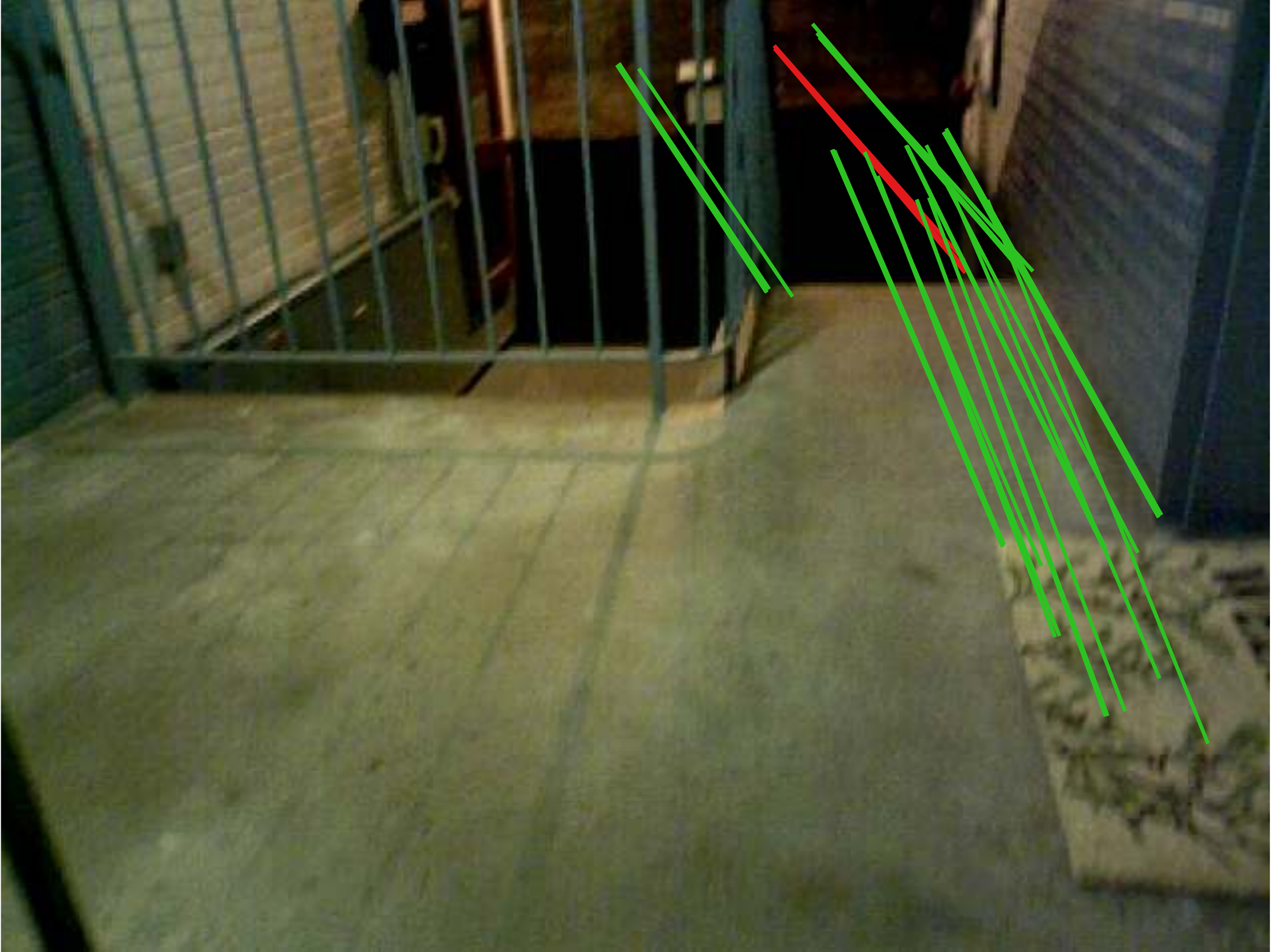}
	\includegraphics[height=7.5em]{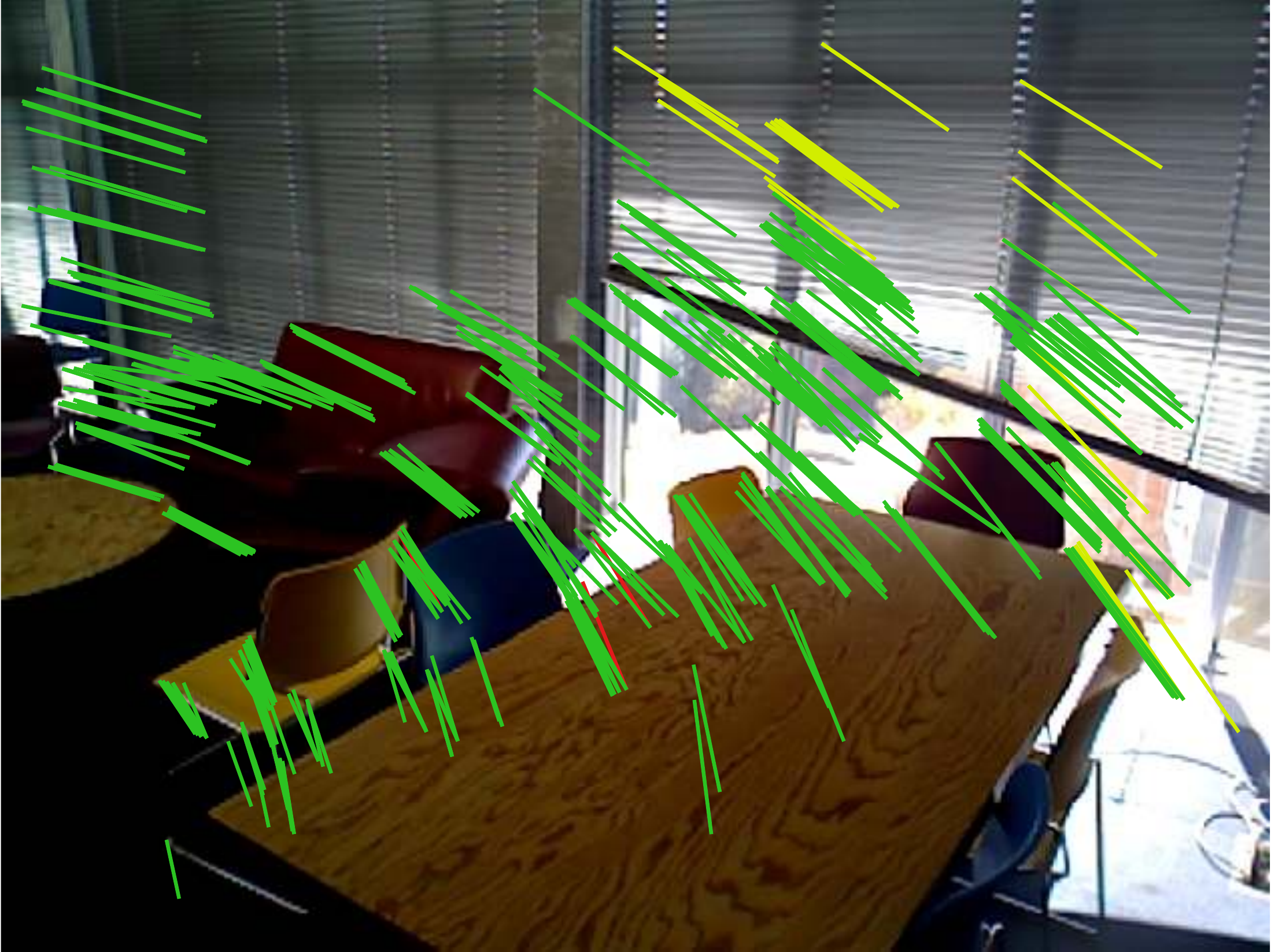}
	\includegraphics[height=7.5em]{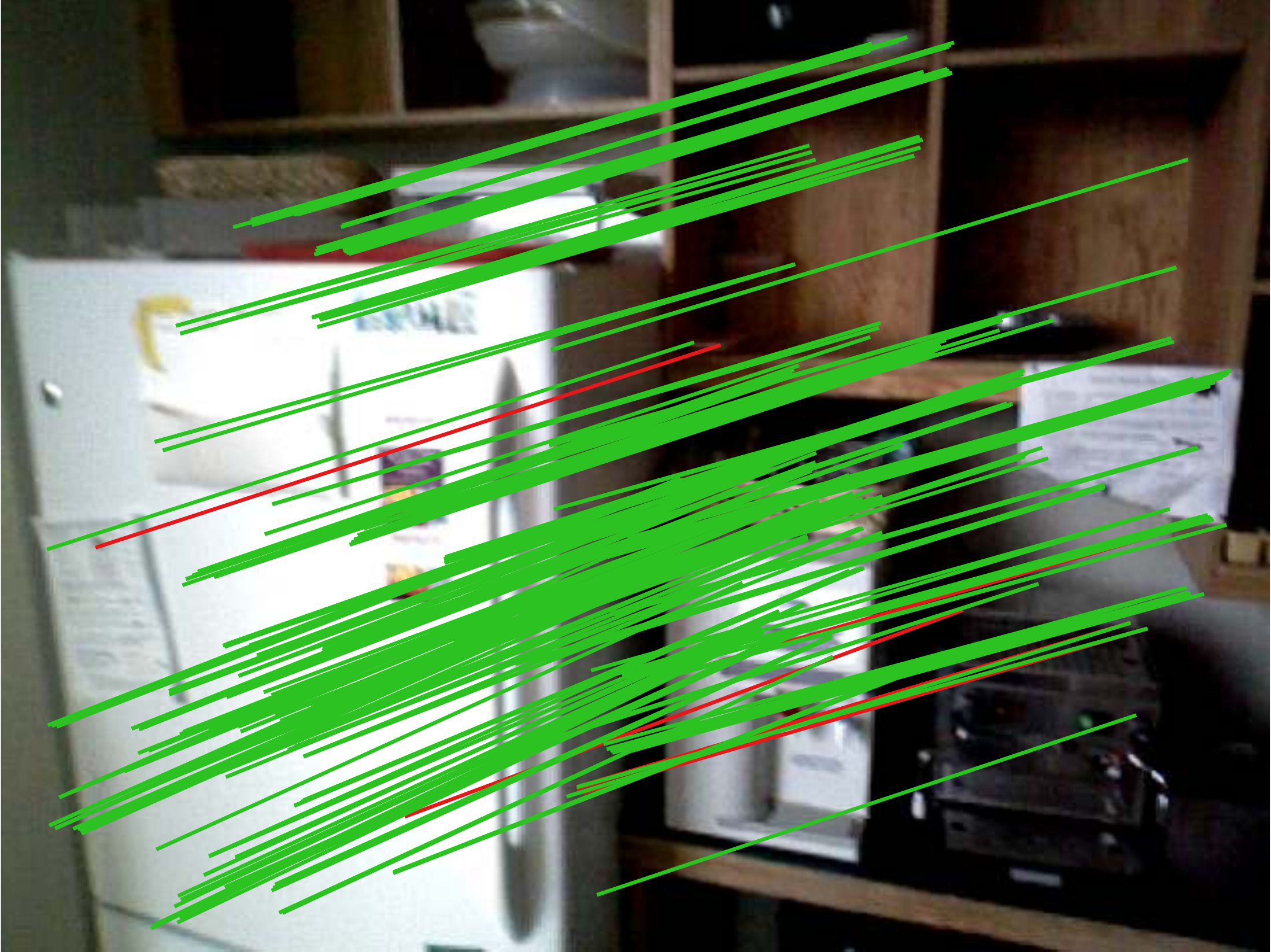}
	\includegraphics[height=7.5em]{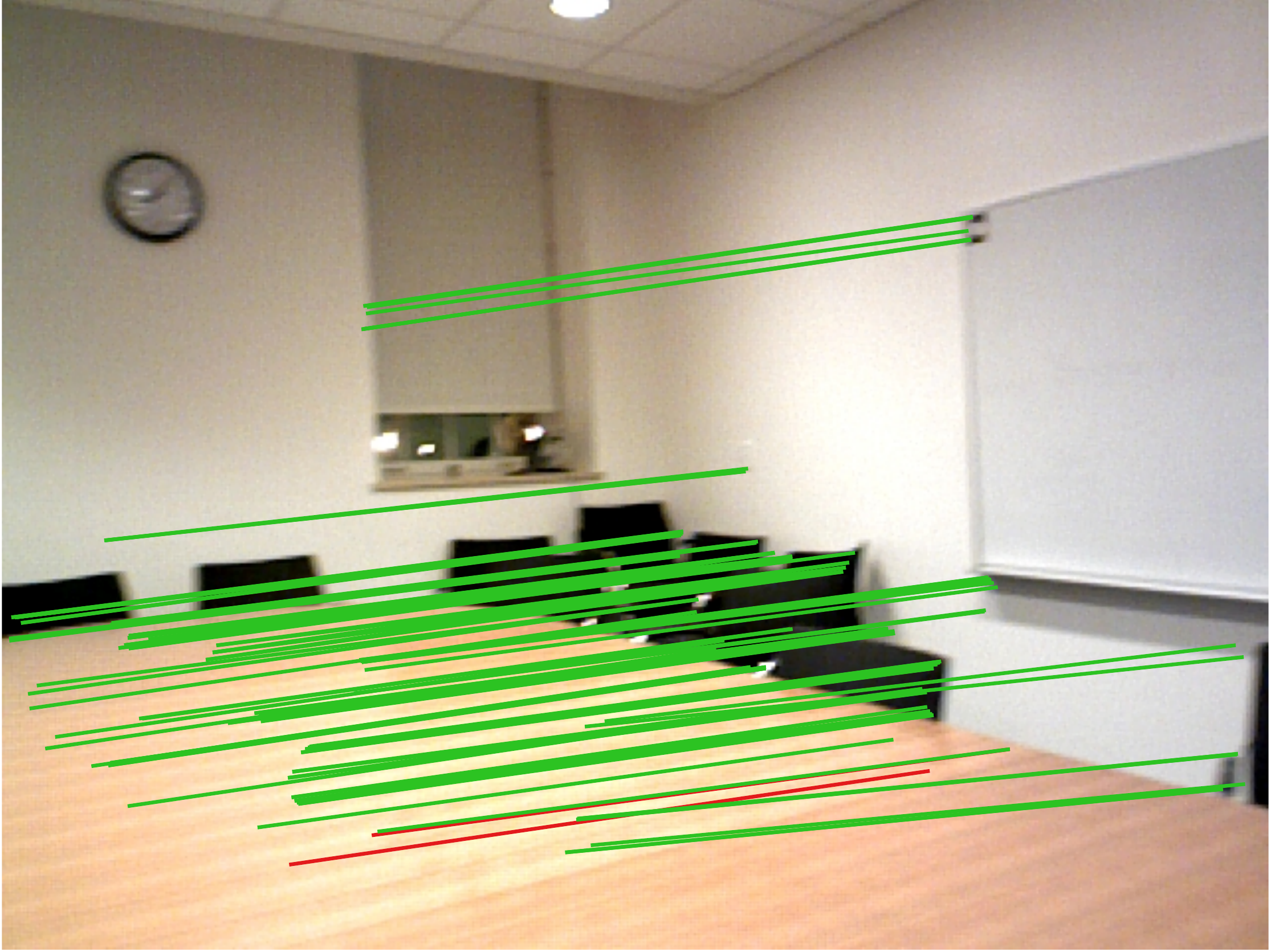}
	\\
	\vspace{0.5em}
	\rotatebox[origin=l]{90}{\mbox{\hspace{2em}OANet}}
	\includegraphics[height=7.5em]{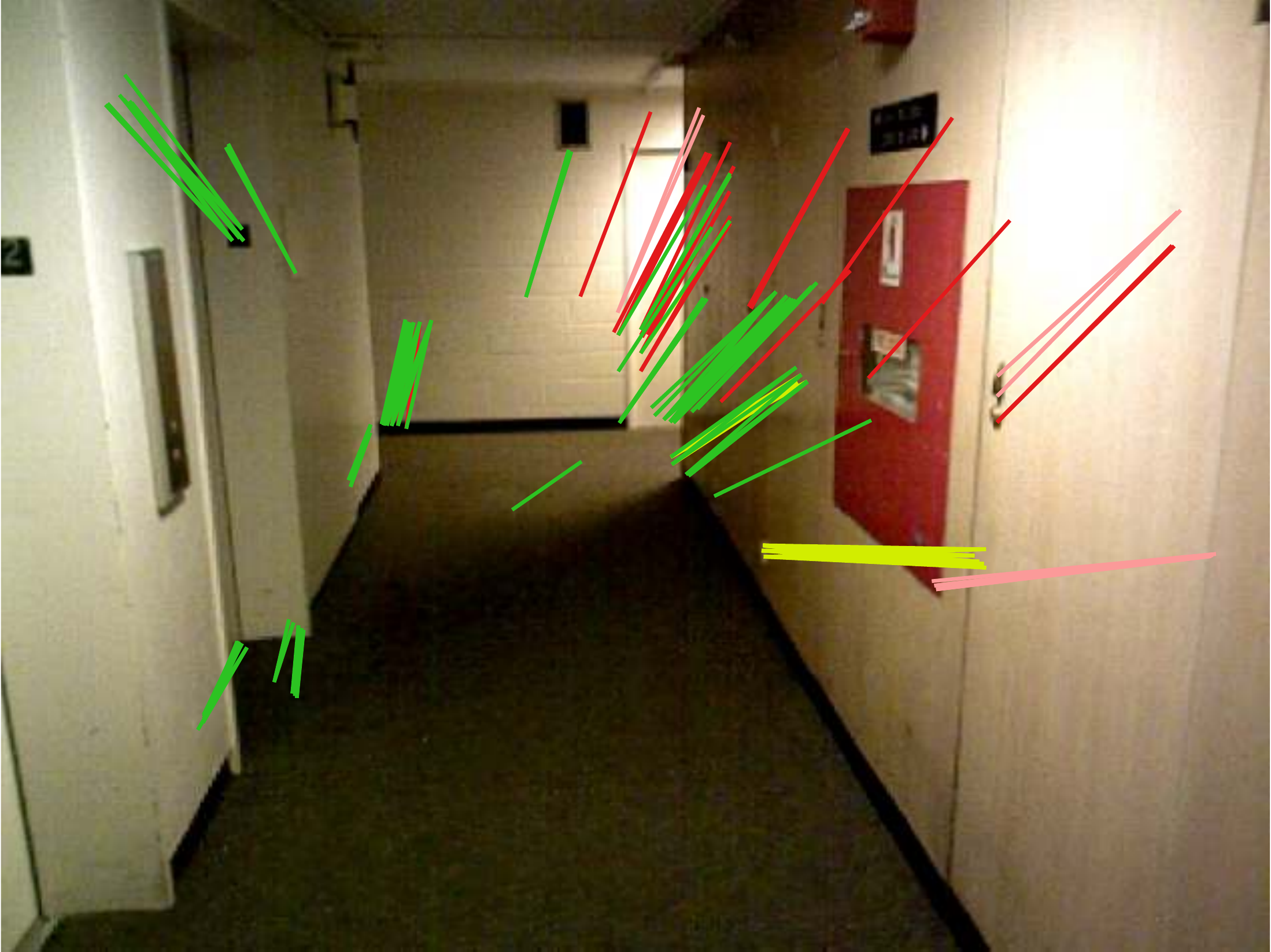}
	\includegraphics[height=7.5em]{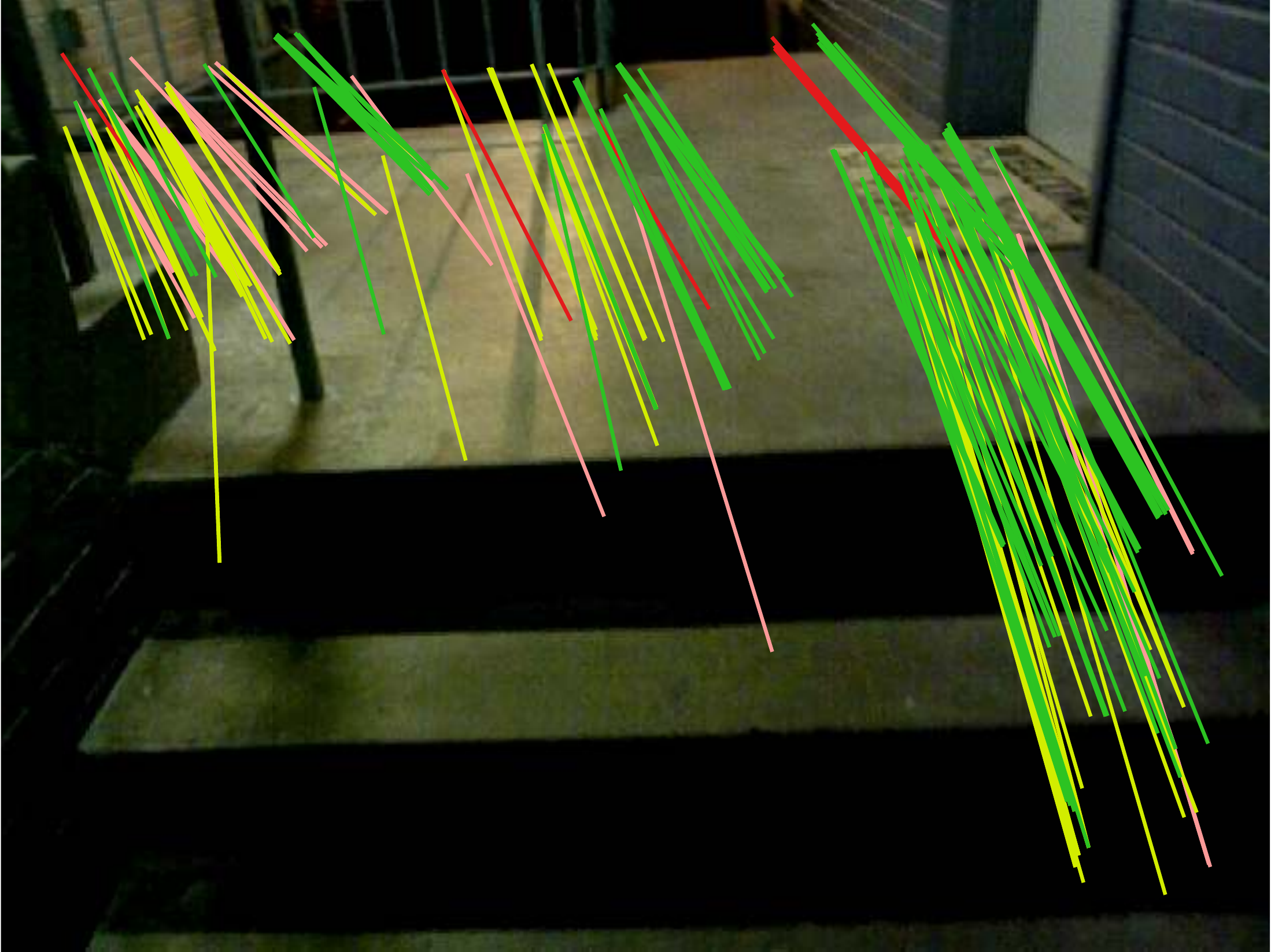}
	\includegraphics[height=7.5em]{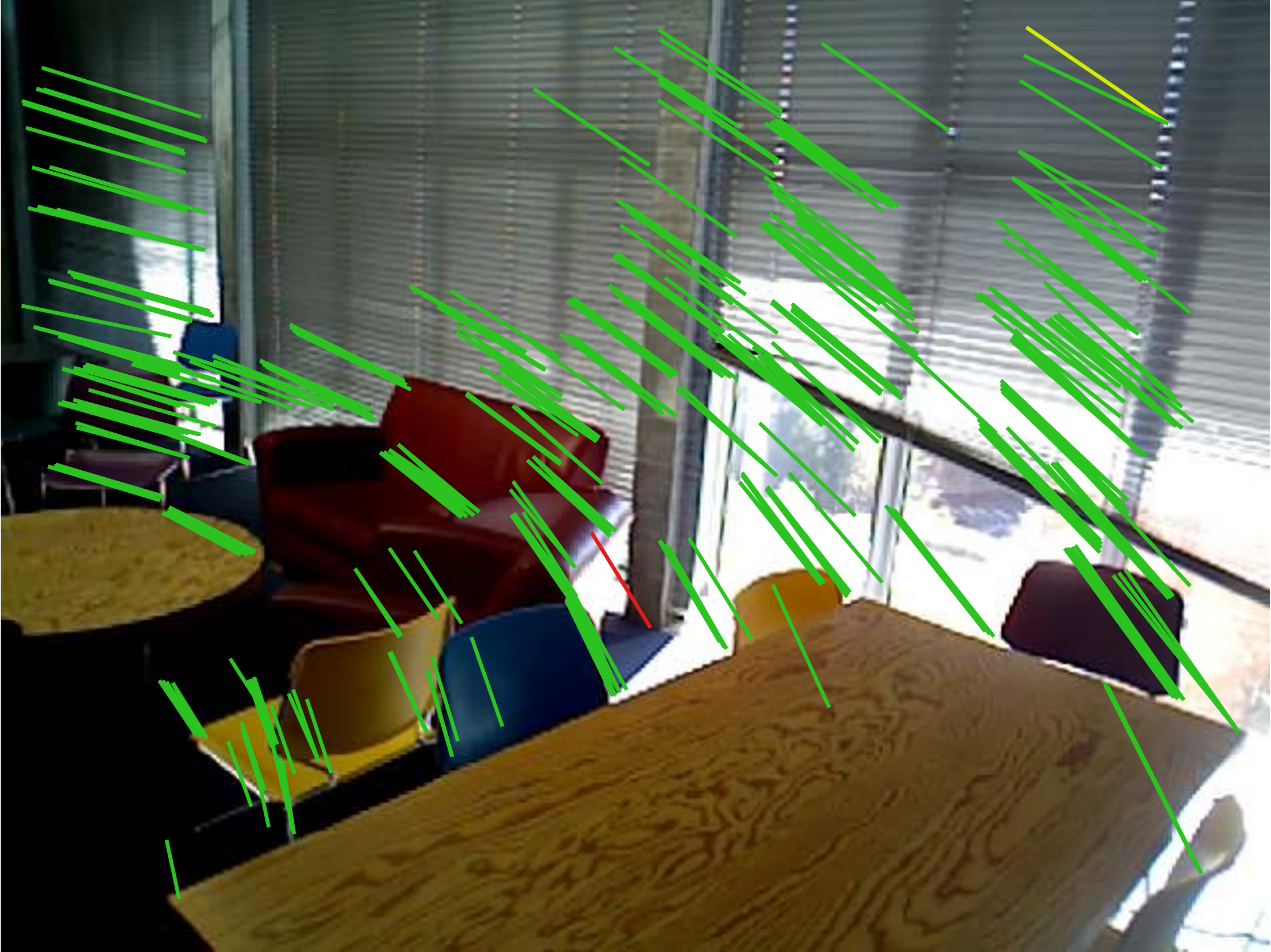}
	\includegraphics[height=7.5em]{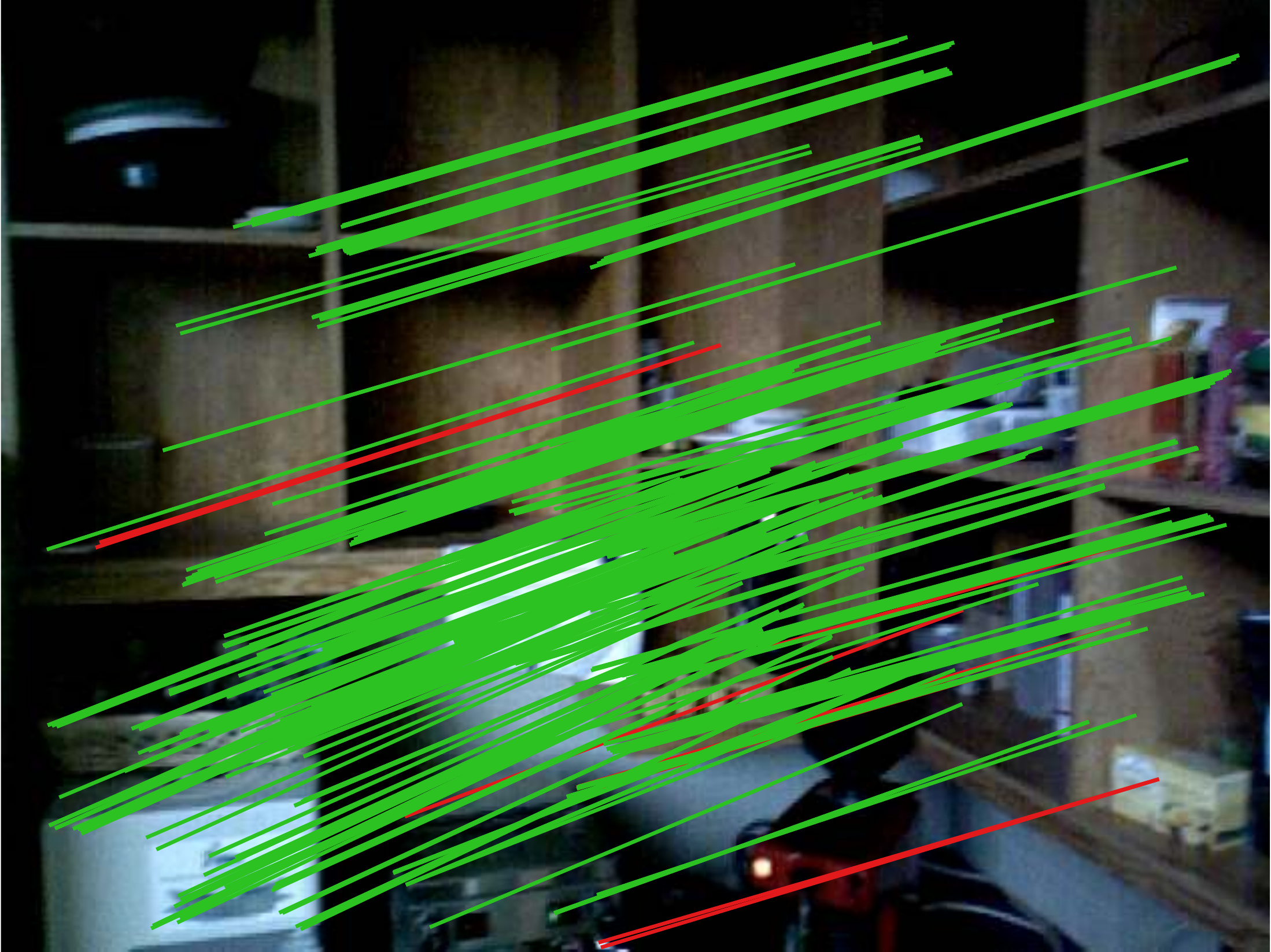}
	\includegraphics[height=7.5em]{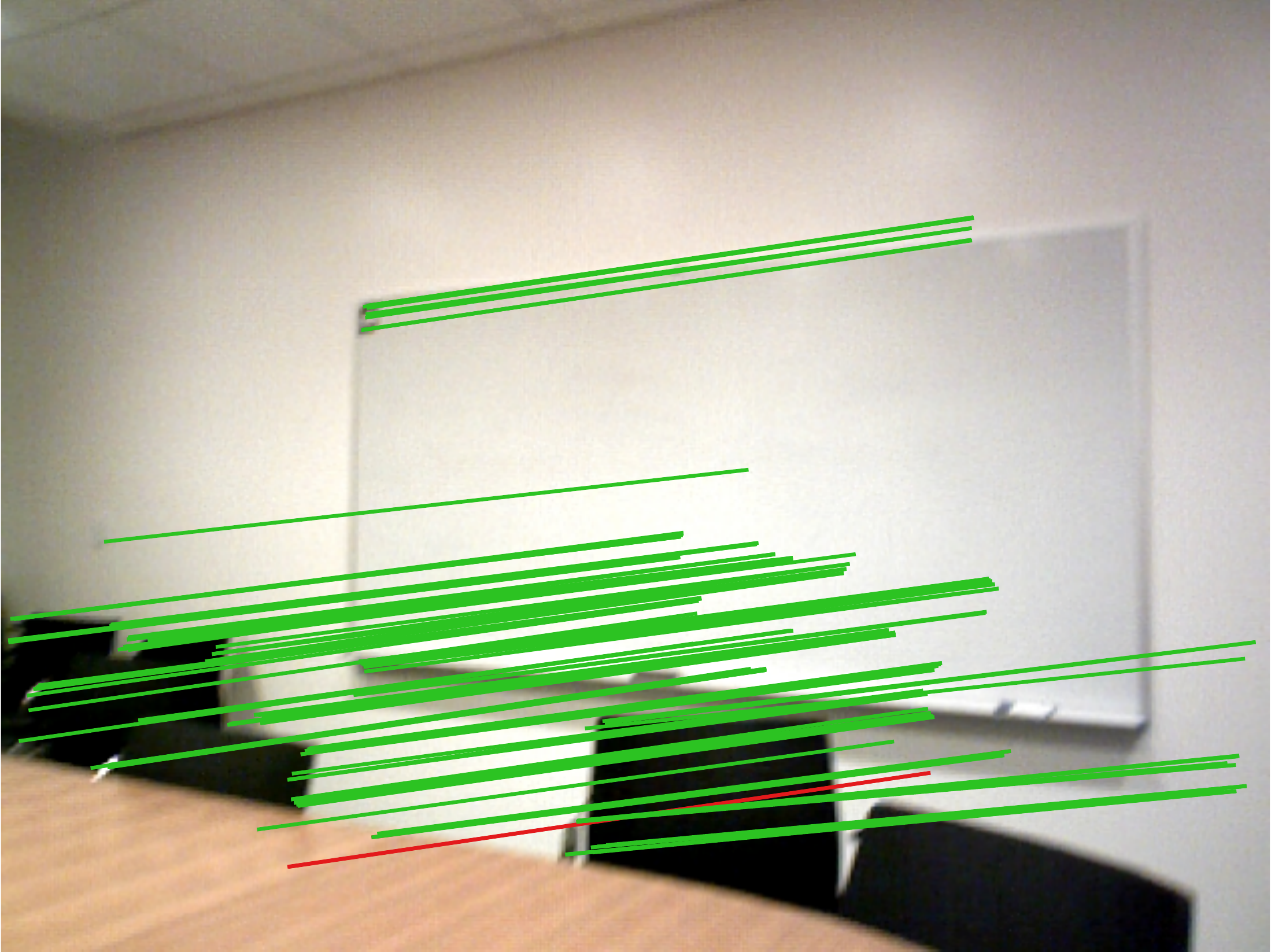}
	\\
	\vspace{0.5em}
	\rotatebox[origin=l]{90}{\mbox{\hspace{2em}ACNe$^\star$}}
	\includegraphics[height=7.5em]{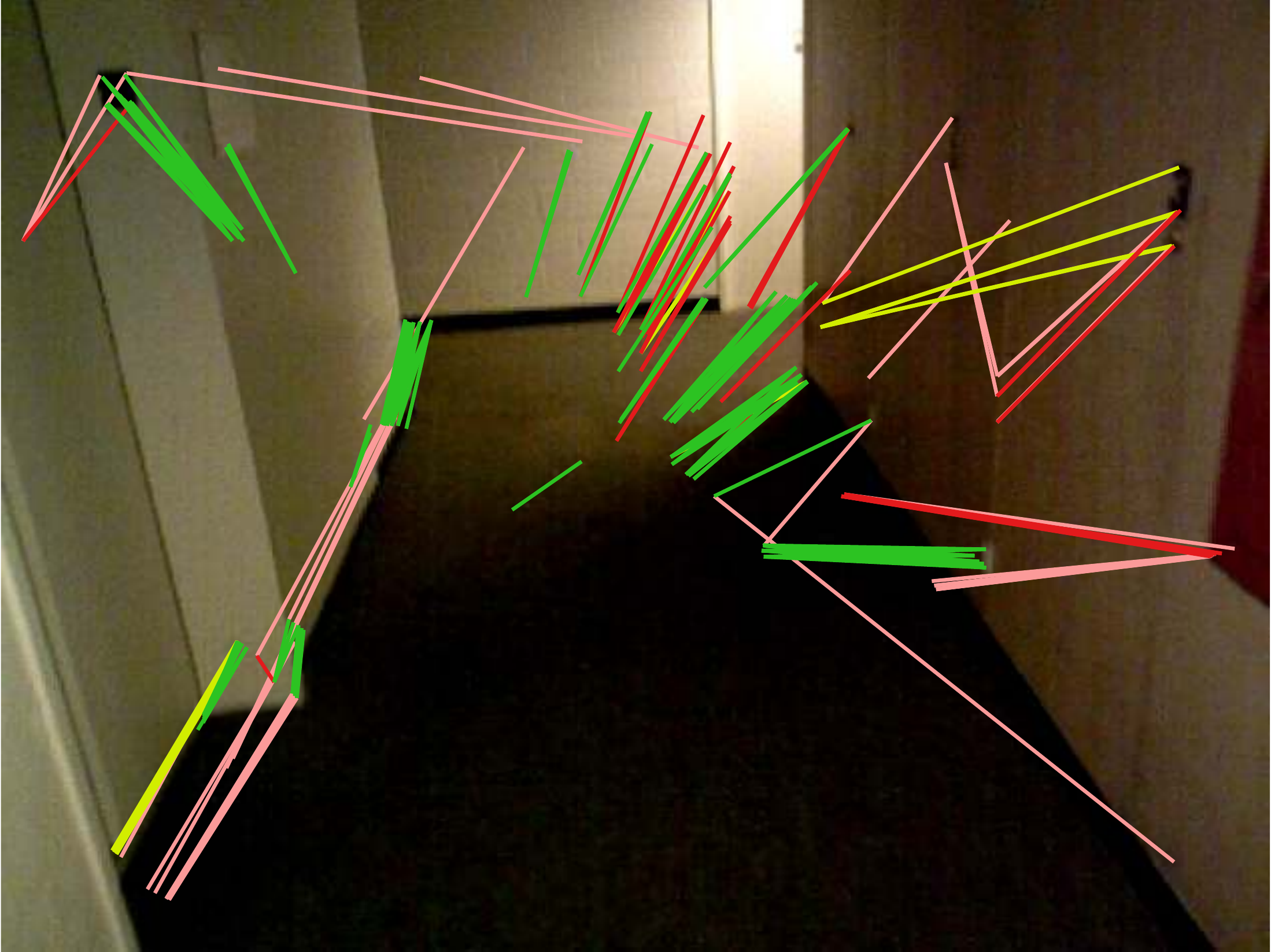}
	\includegraphics[height=7.5em]{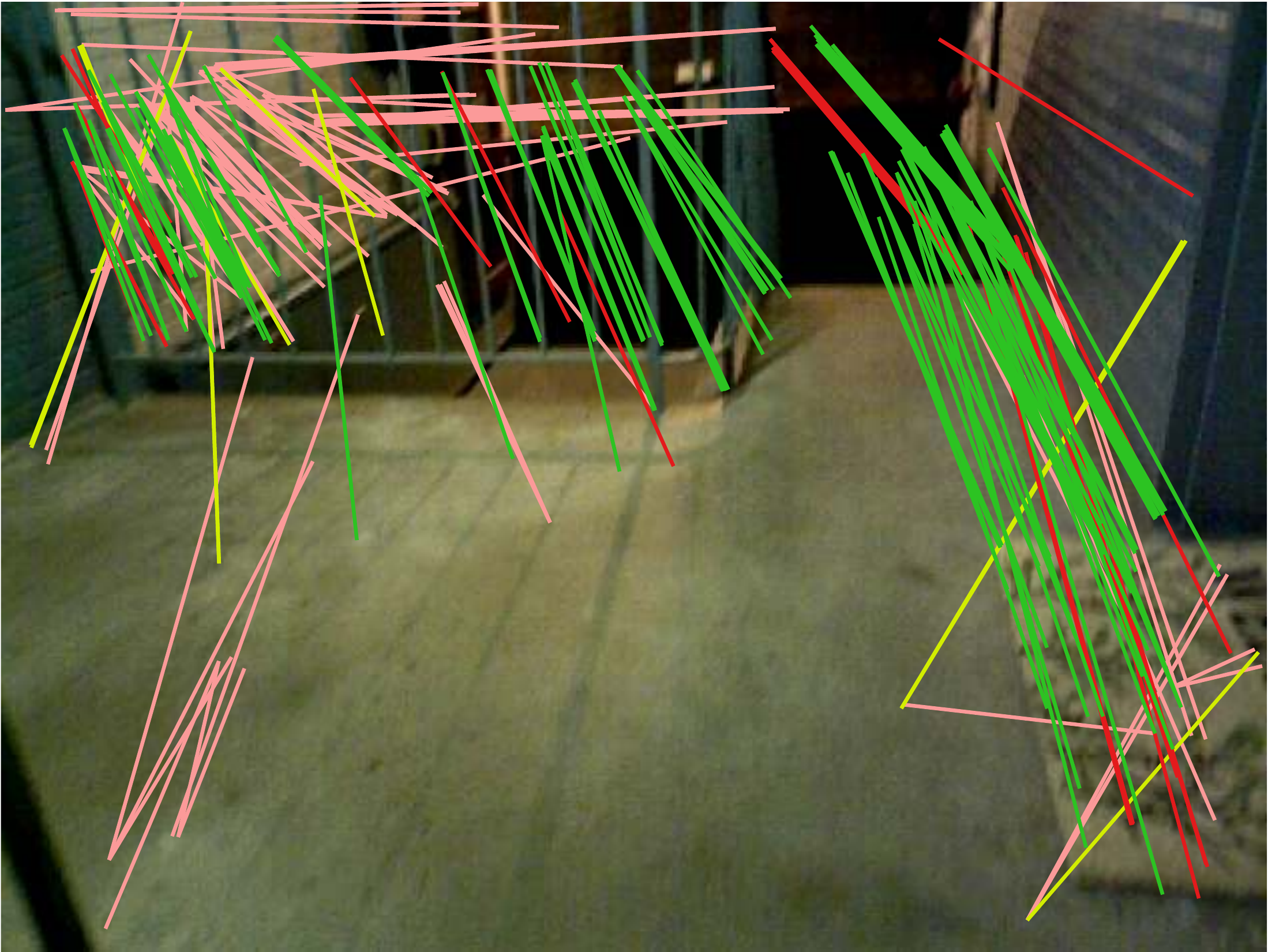}
	\includegraphics[height=7.5em]{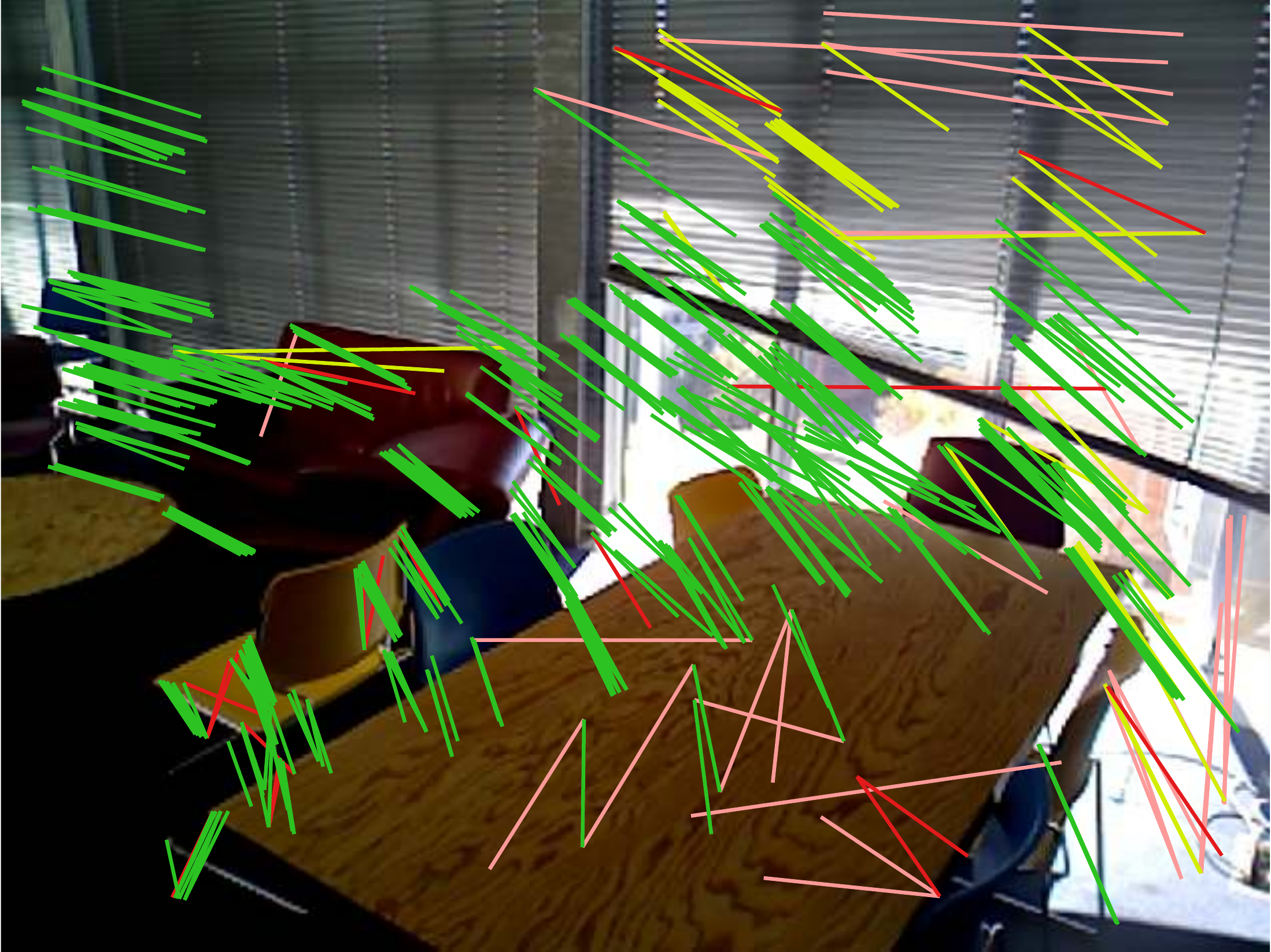}
	\includegraphics[height=7.5em]{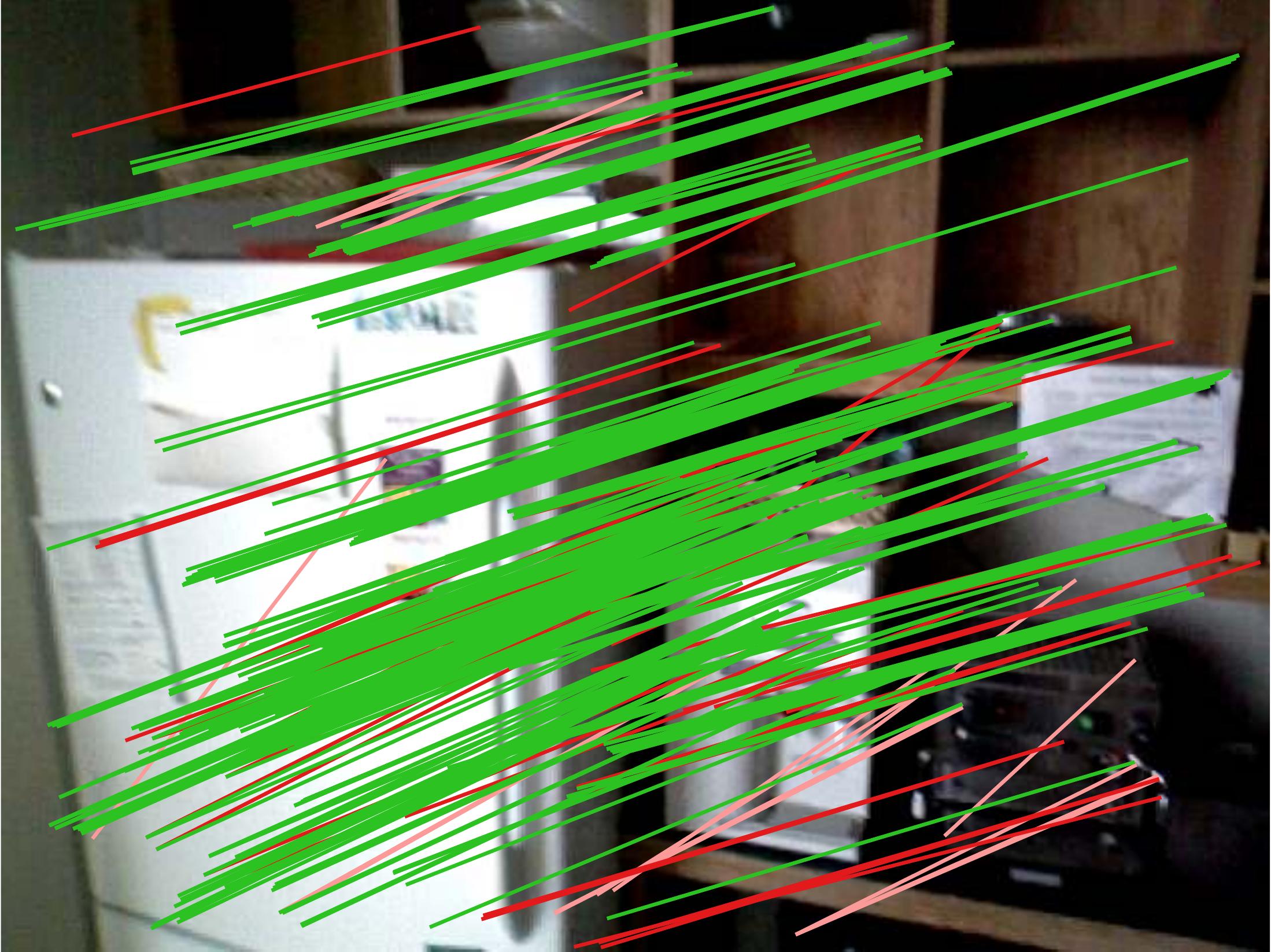}
	\includegraphics[height=7.5em]{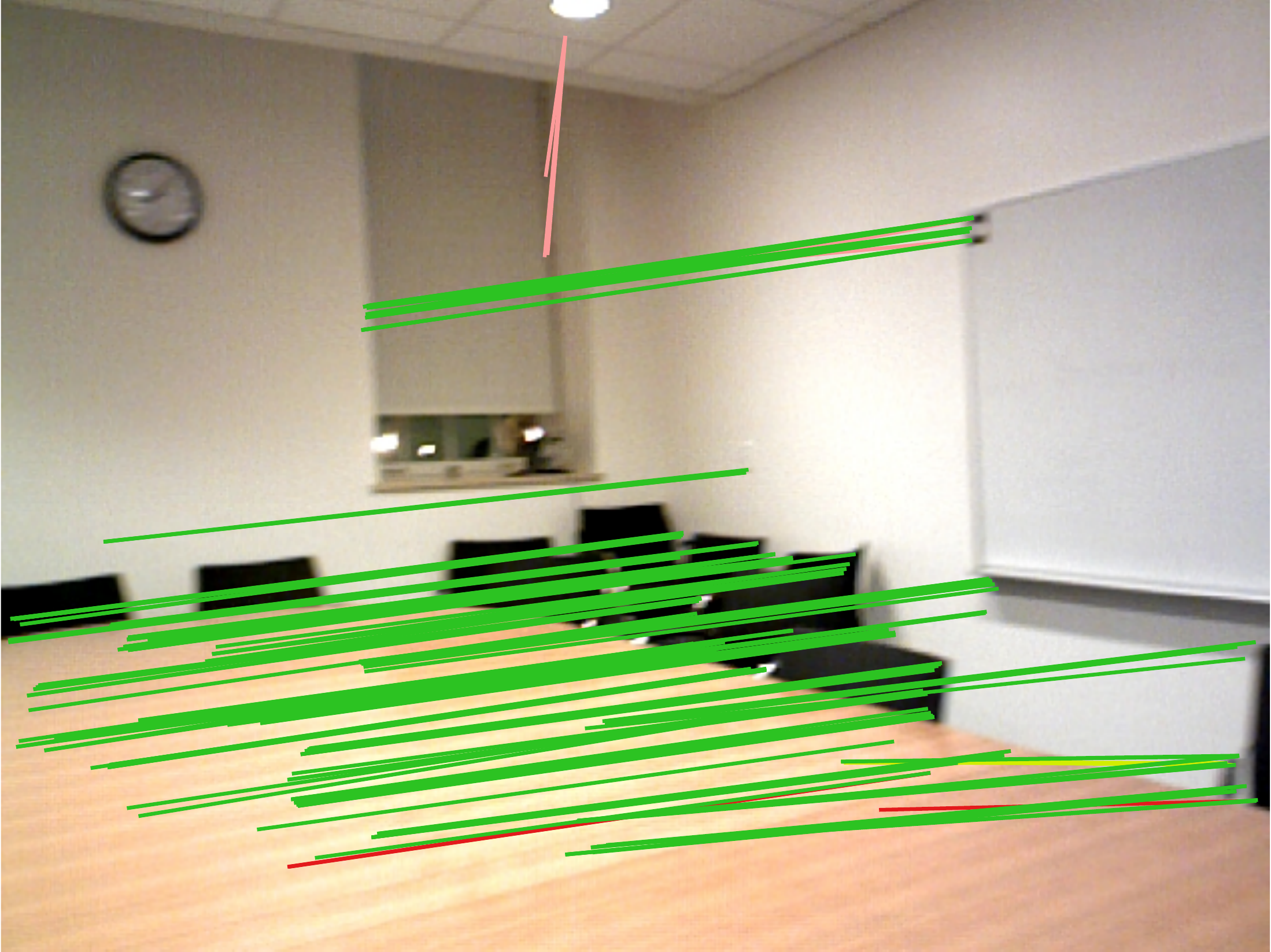}
	\\
	\vspace{0.5em}
	\rotatebox[origin=l]{90}{\mbox{\hspace{2em}PFM}}
	\includegraphics[height=7.5em]{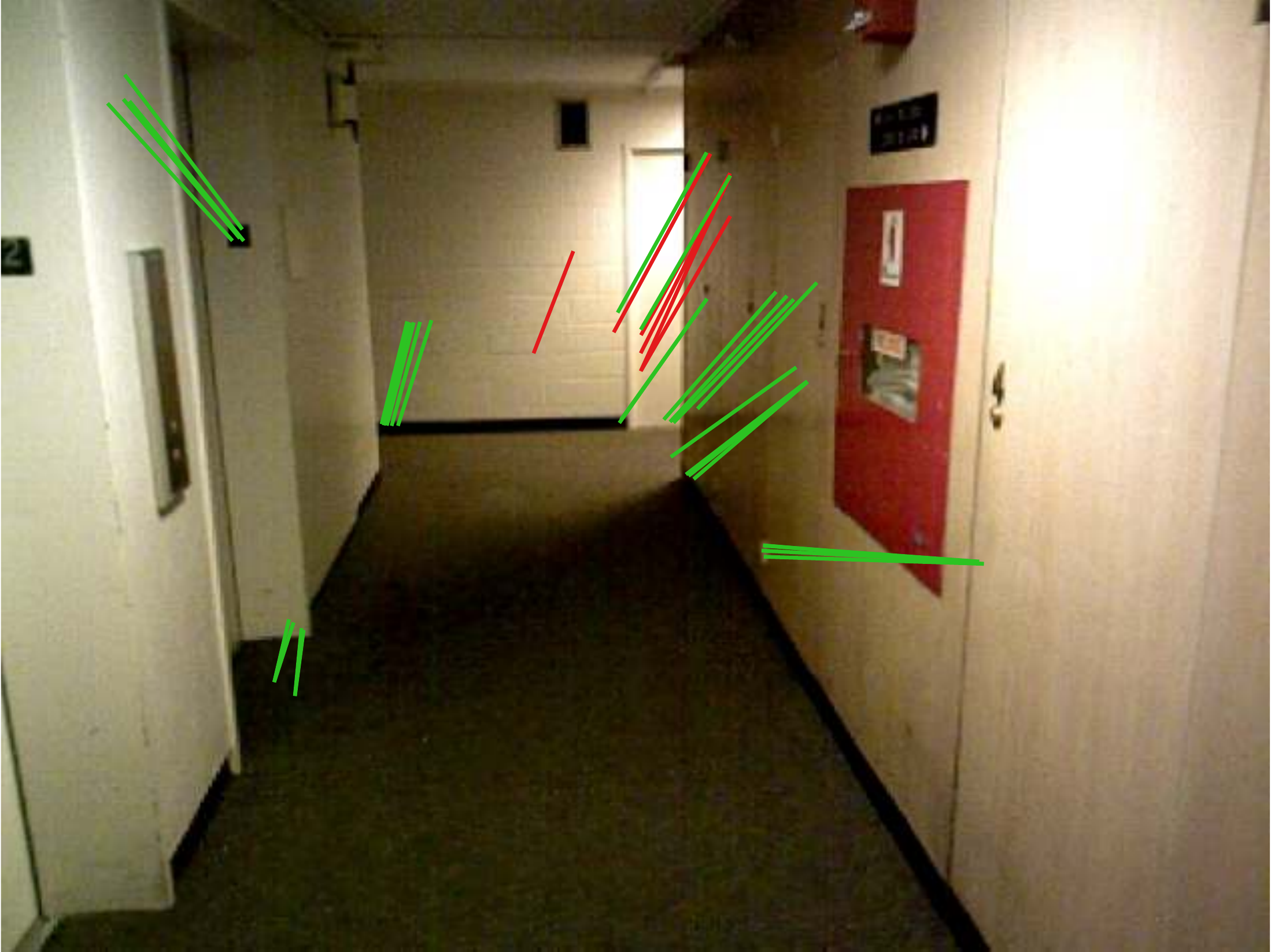}
	\includegraphics[height=7.5em]{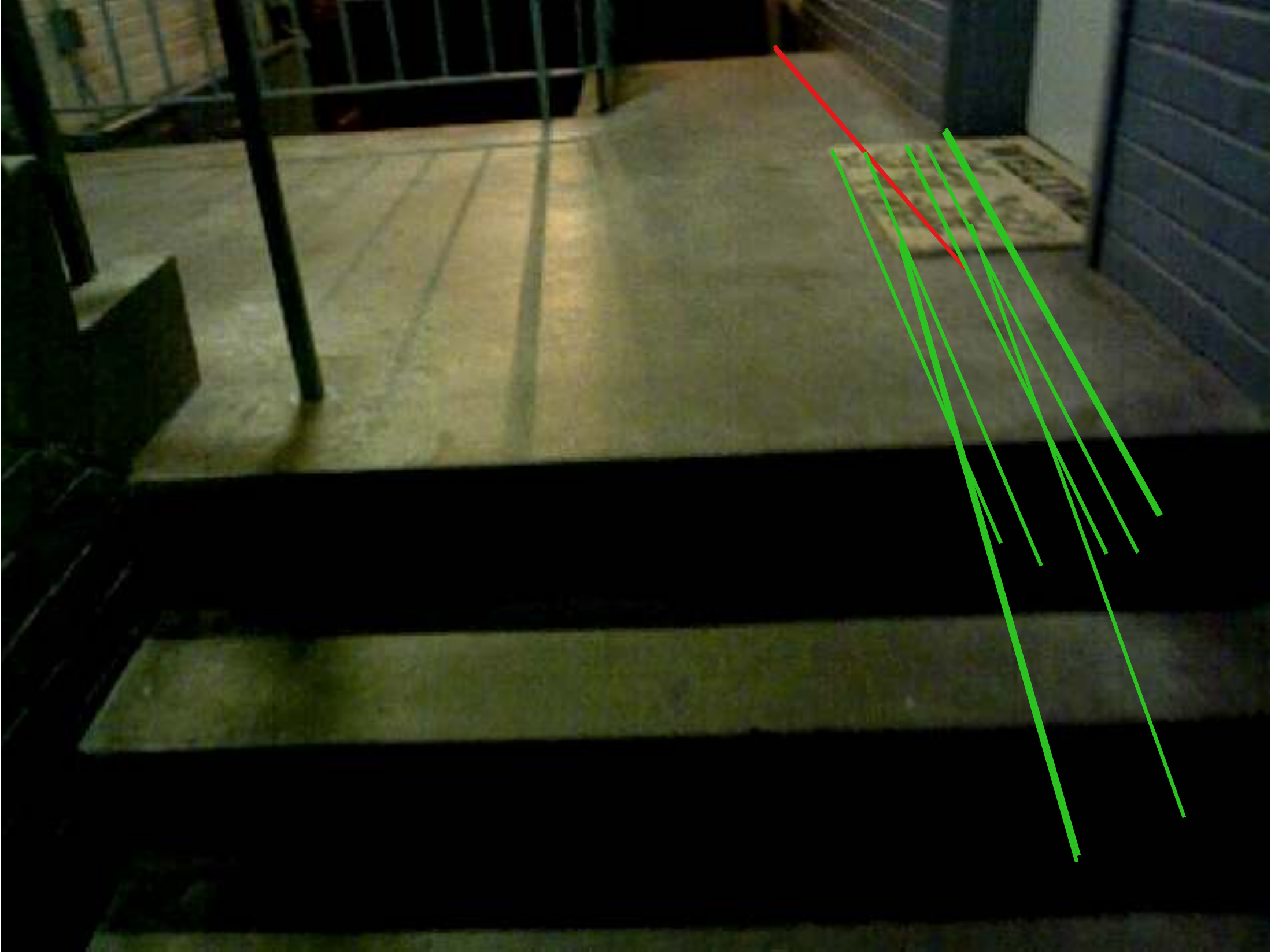}
	\includegraphics[height=7.5em]{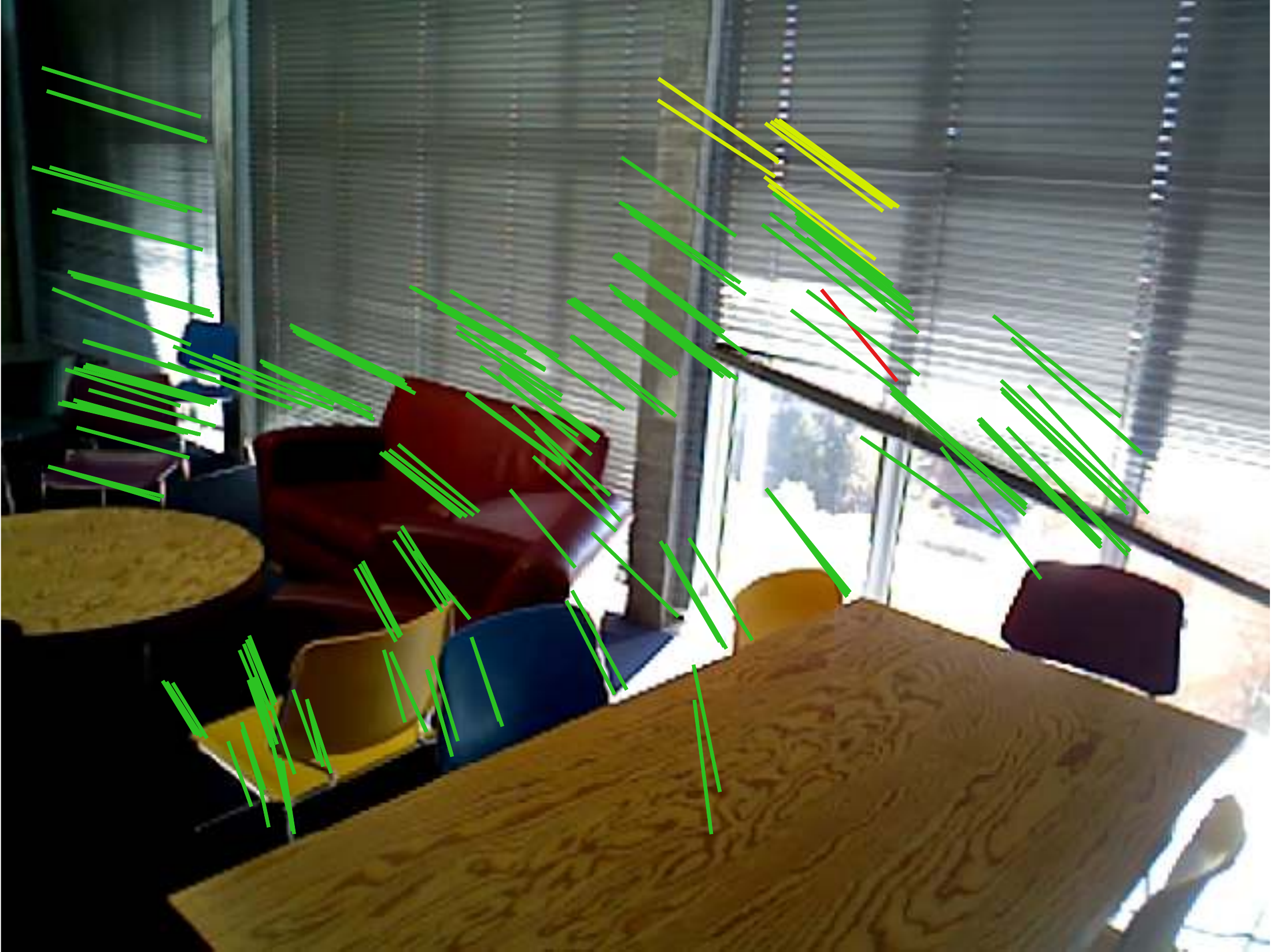}
	\includegraphics[height=7.5em]{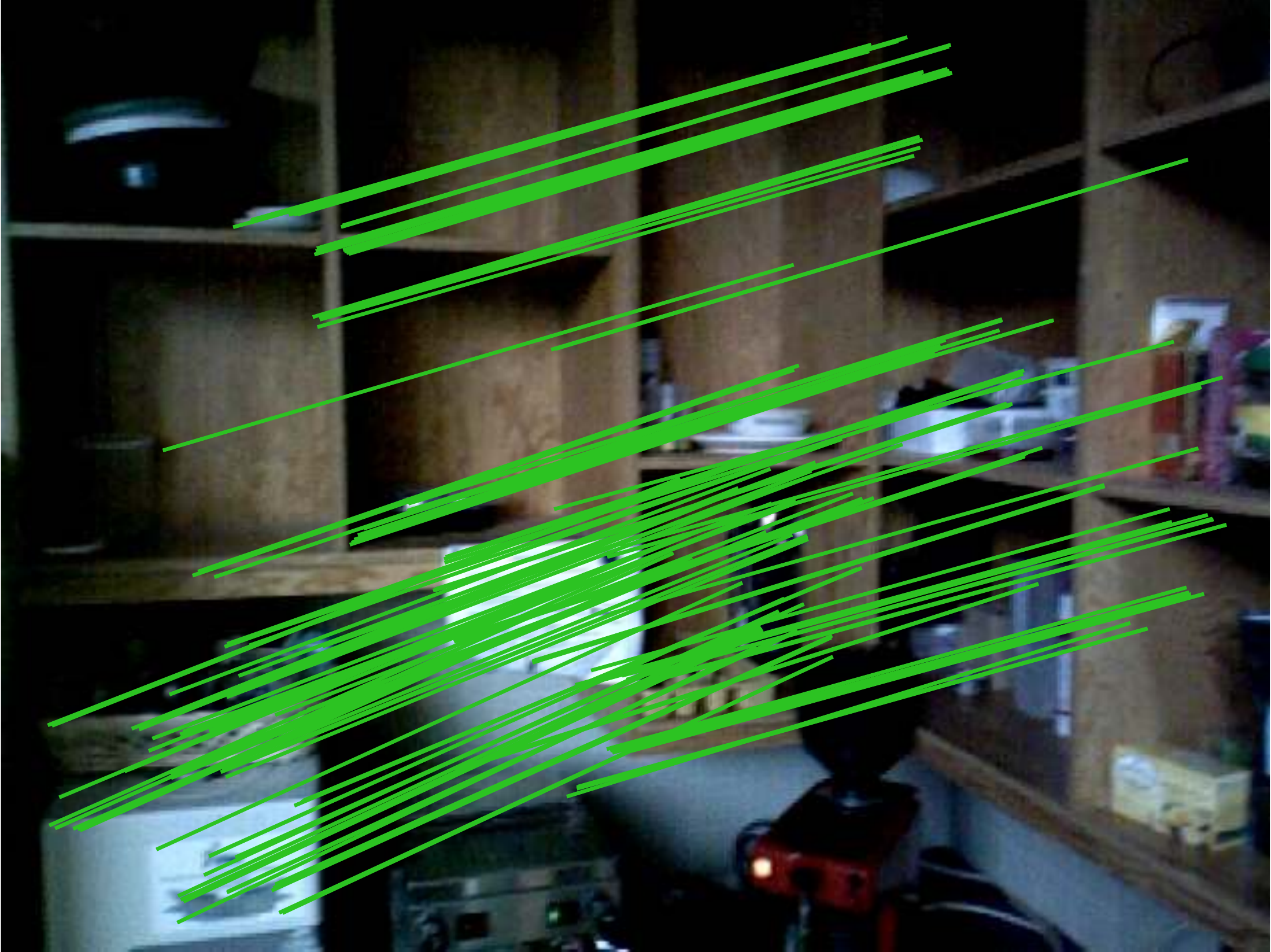}
	\includegraphics[height=7.5em]{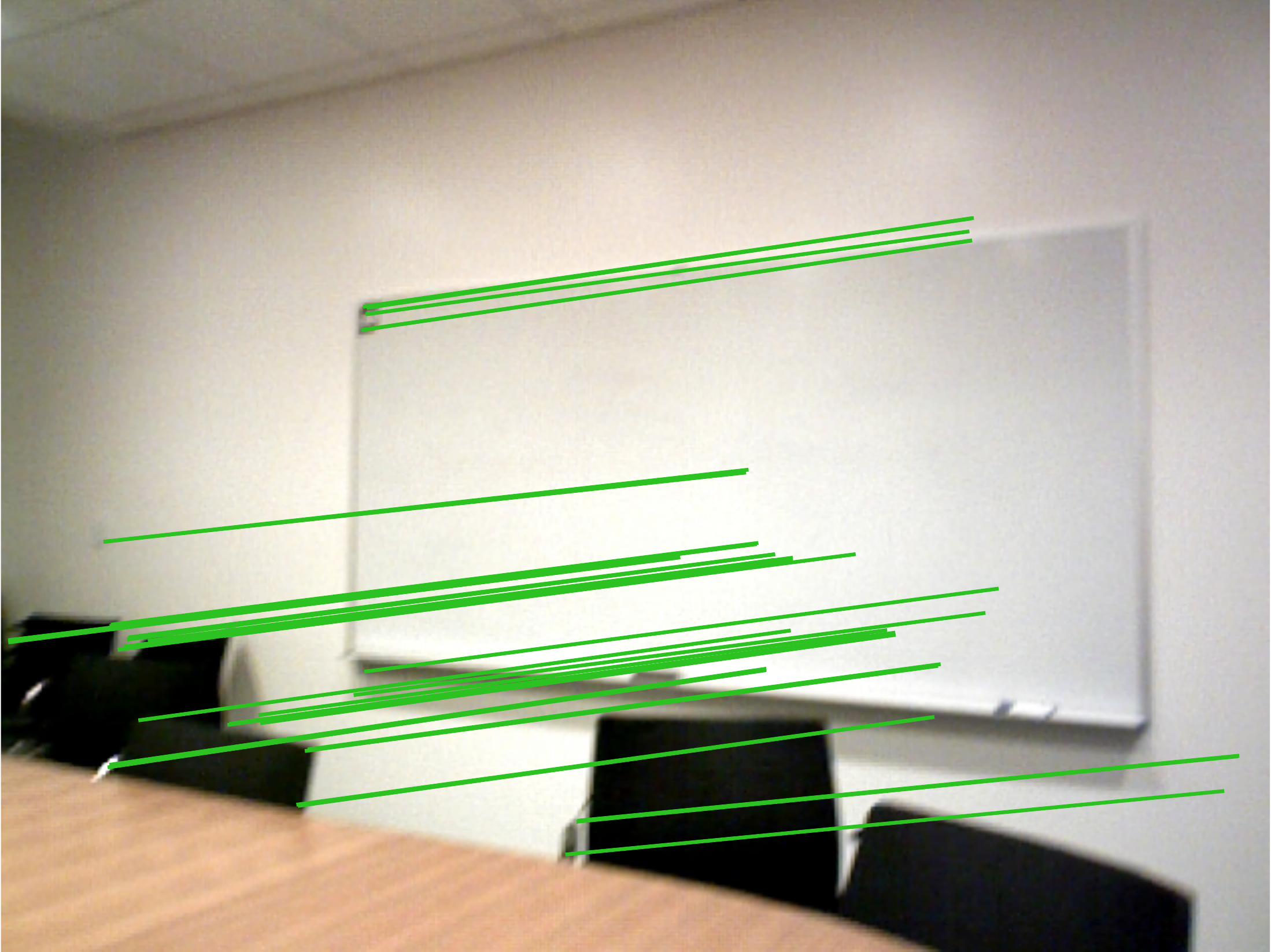}
	\\
	\vspace{0.5em}
	\rotatebox[origin=l]{90}{\mbox{\hspace{2em}PGM}}
	\includegraphics[height=7.5em]{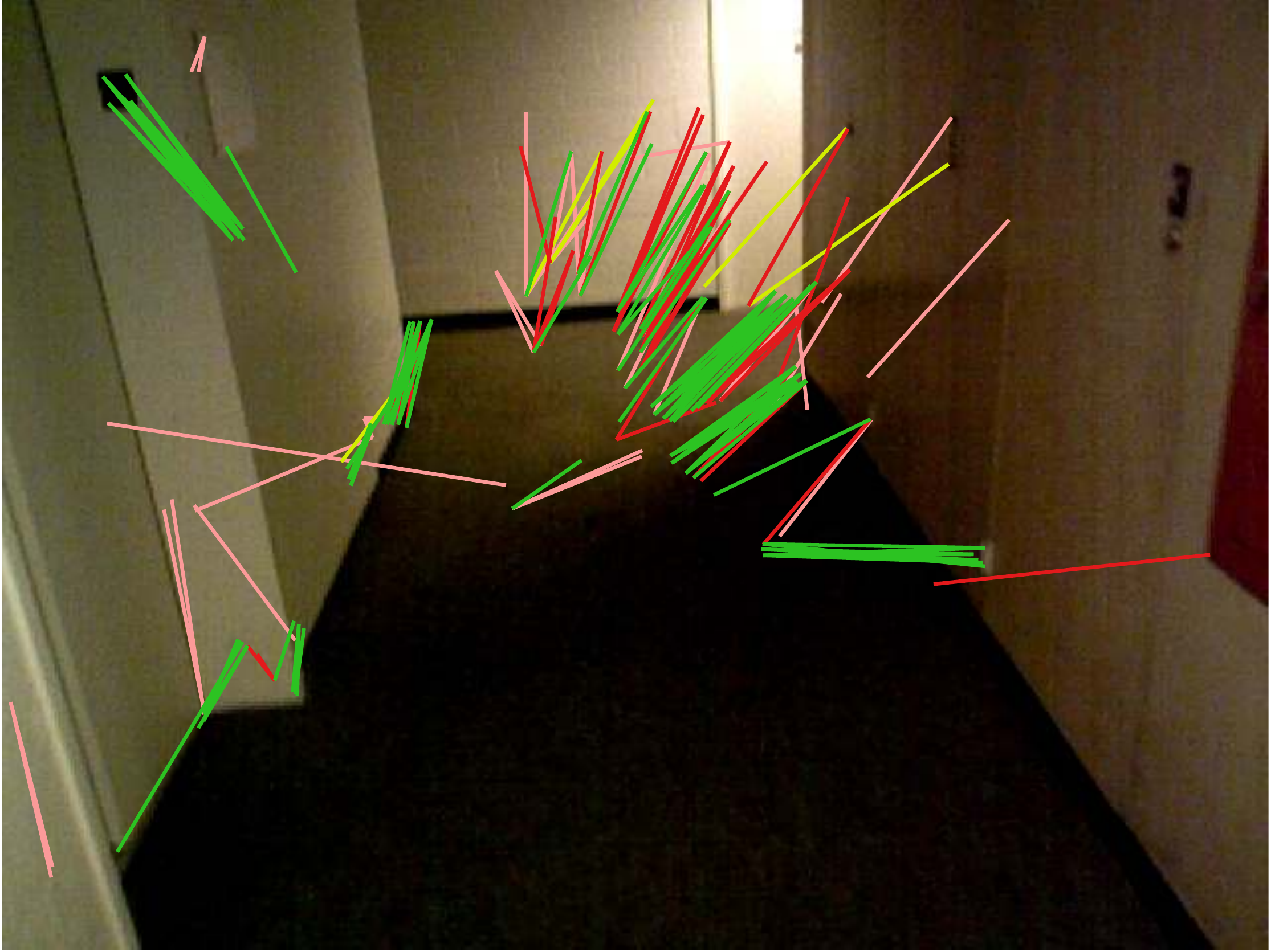}
	\includegraphics[height=7.5em]{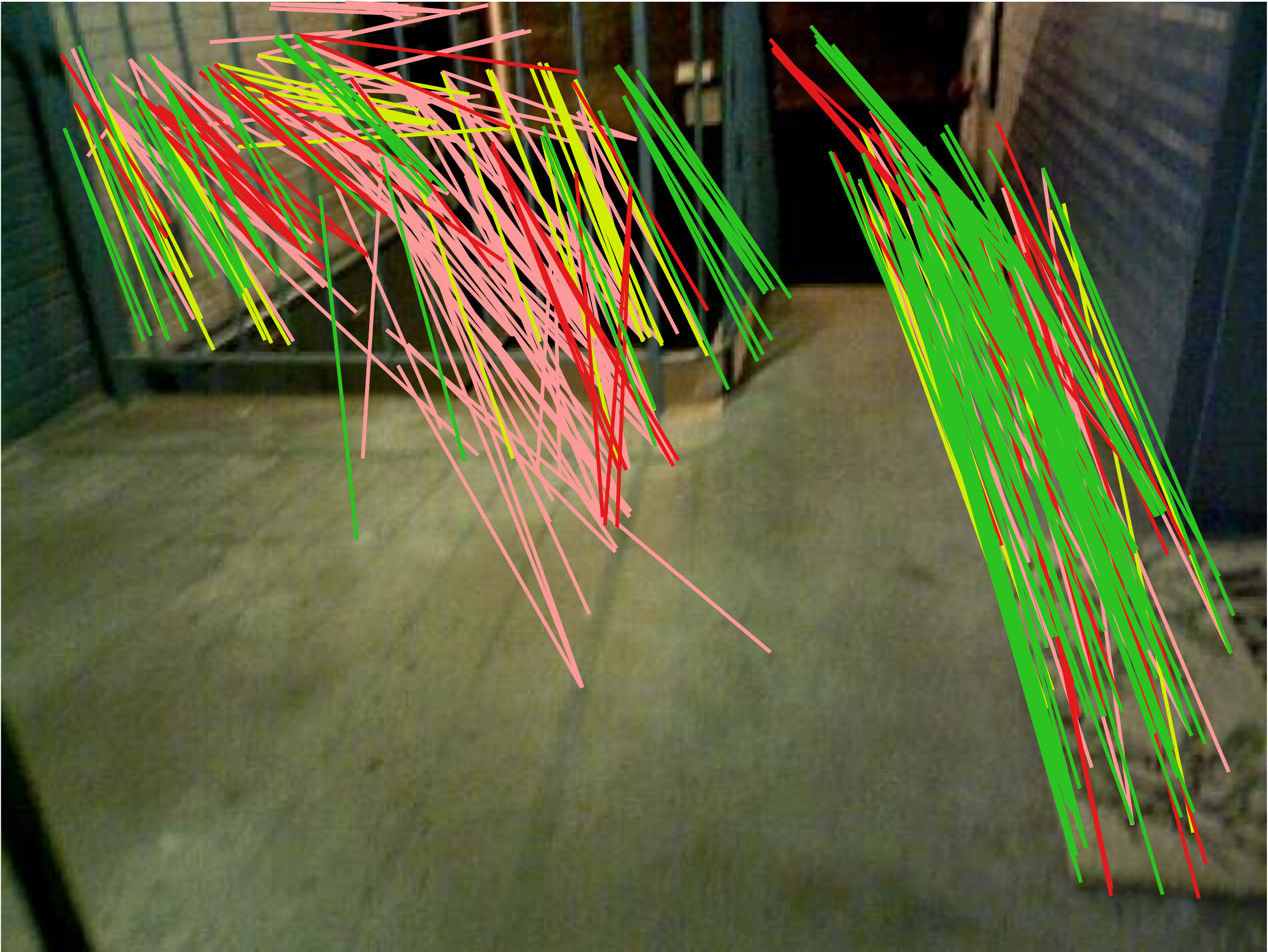}
	\includegraphics[height=7.5em]{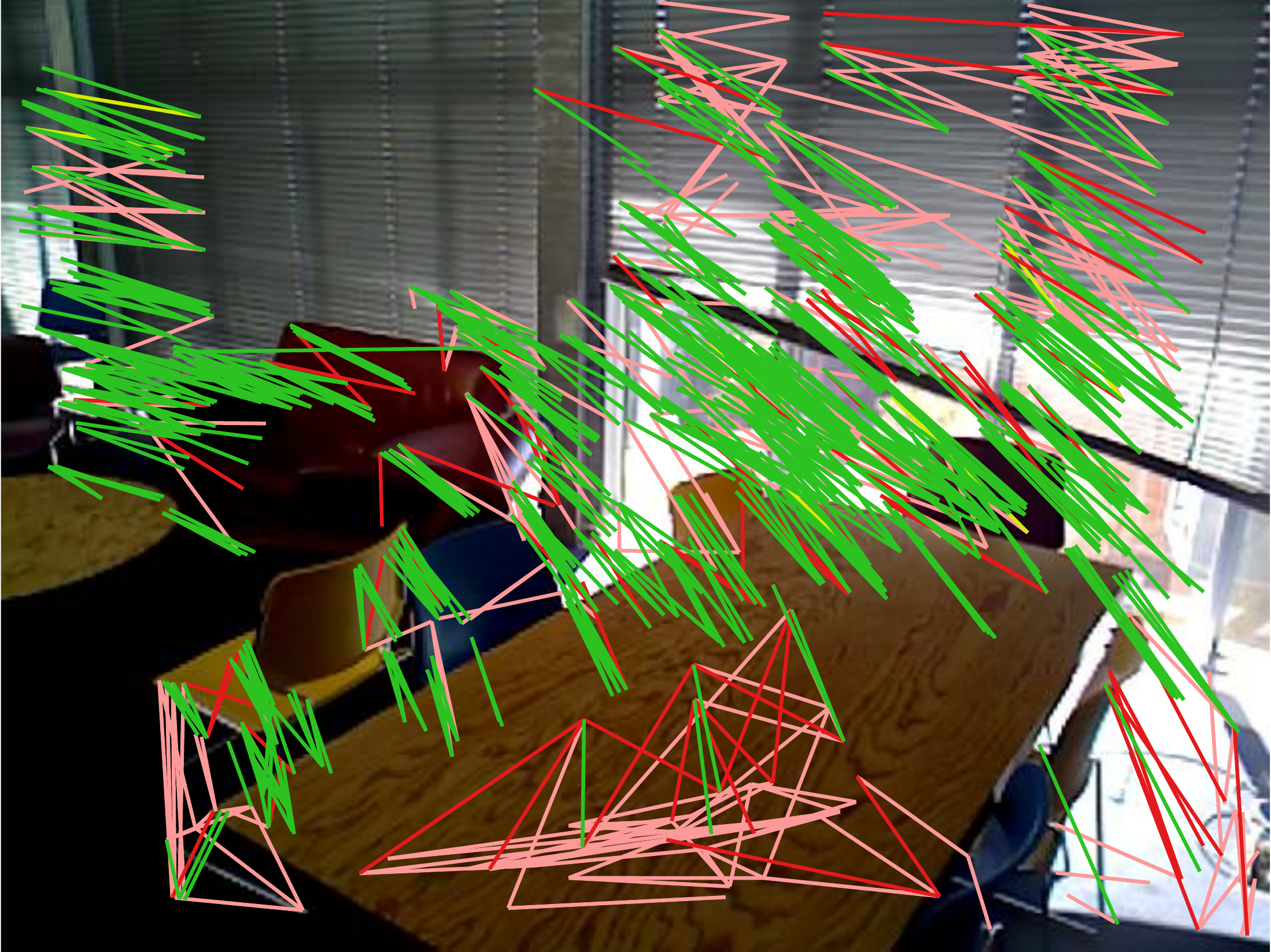}
	\includegraphics[height=7.5em]{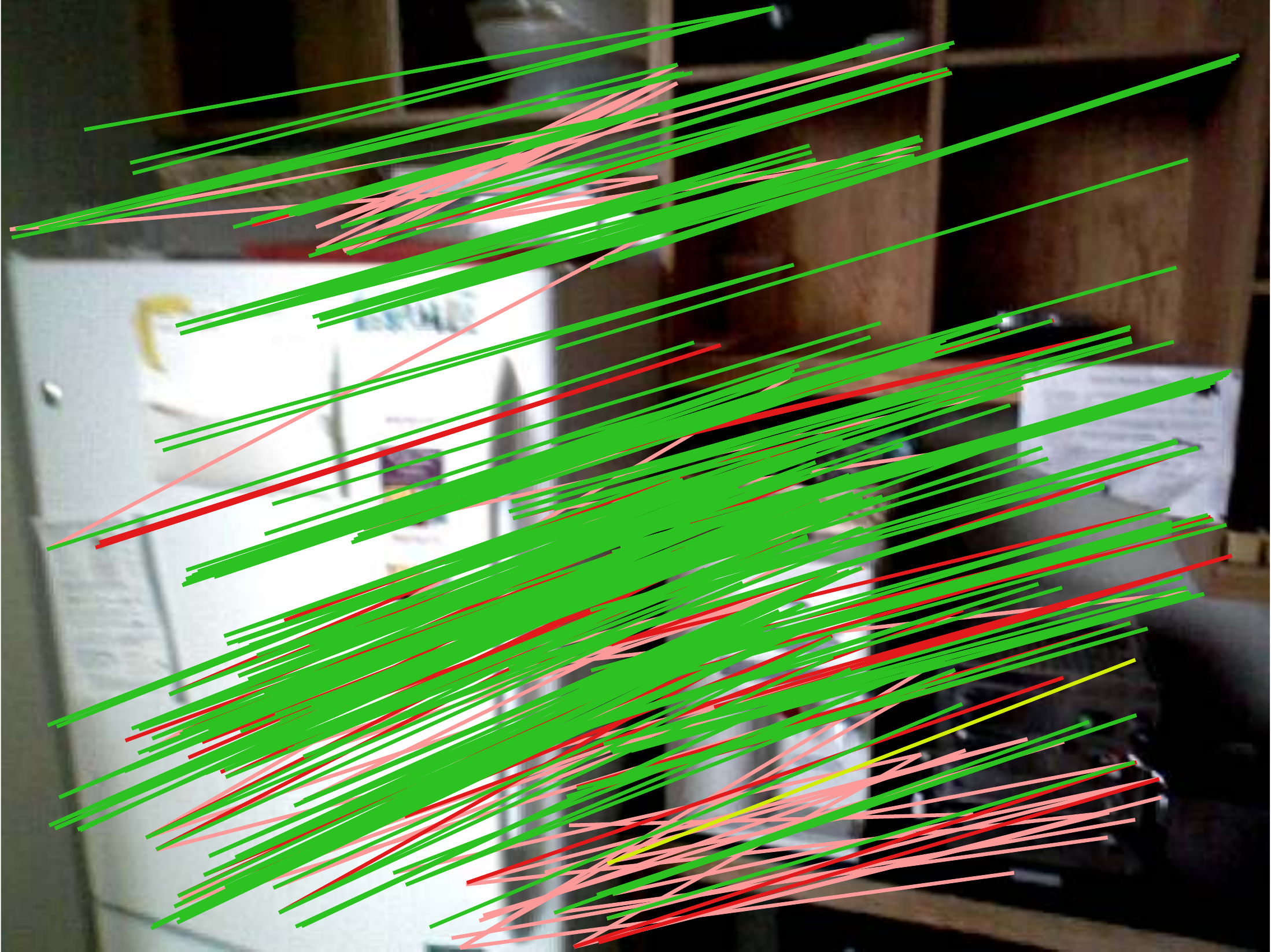}
	\includegraphics[height=7.5em]{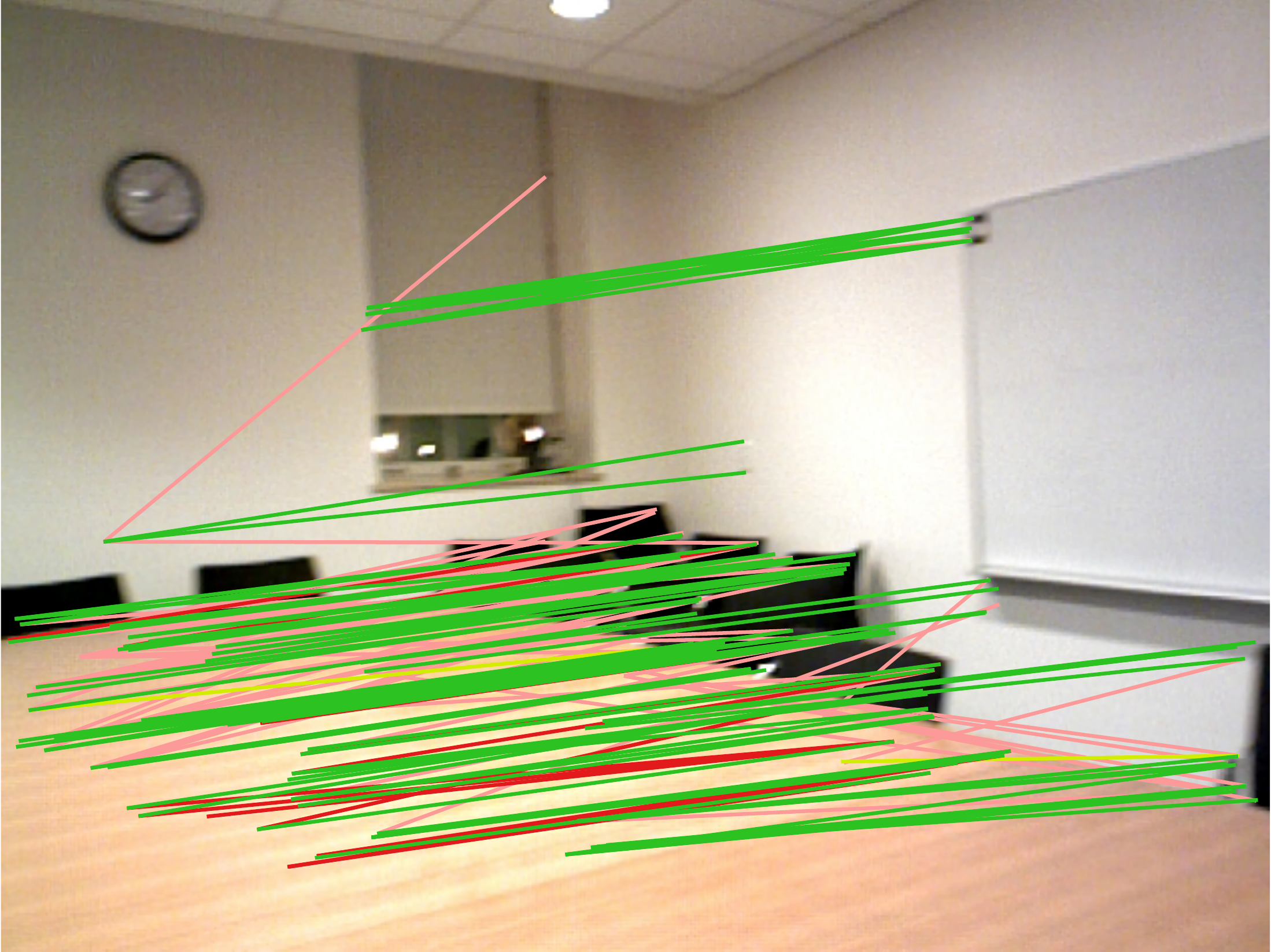}
	\\
	\vspace{0.5em}
	\rotatebox[origin=l]{90}{\mbox{\hspace{2em}SCV}}
	\includegraphics[height=7.5em]{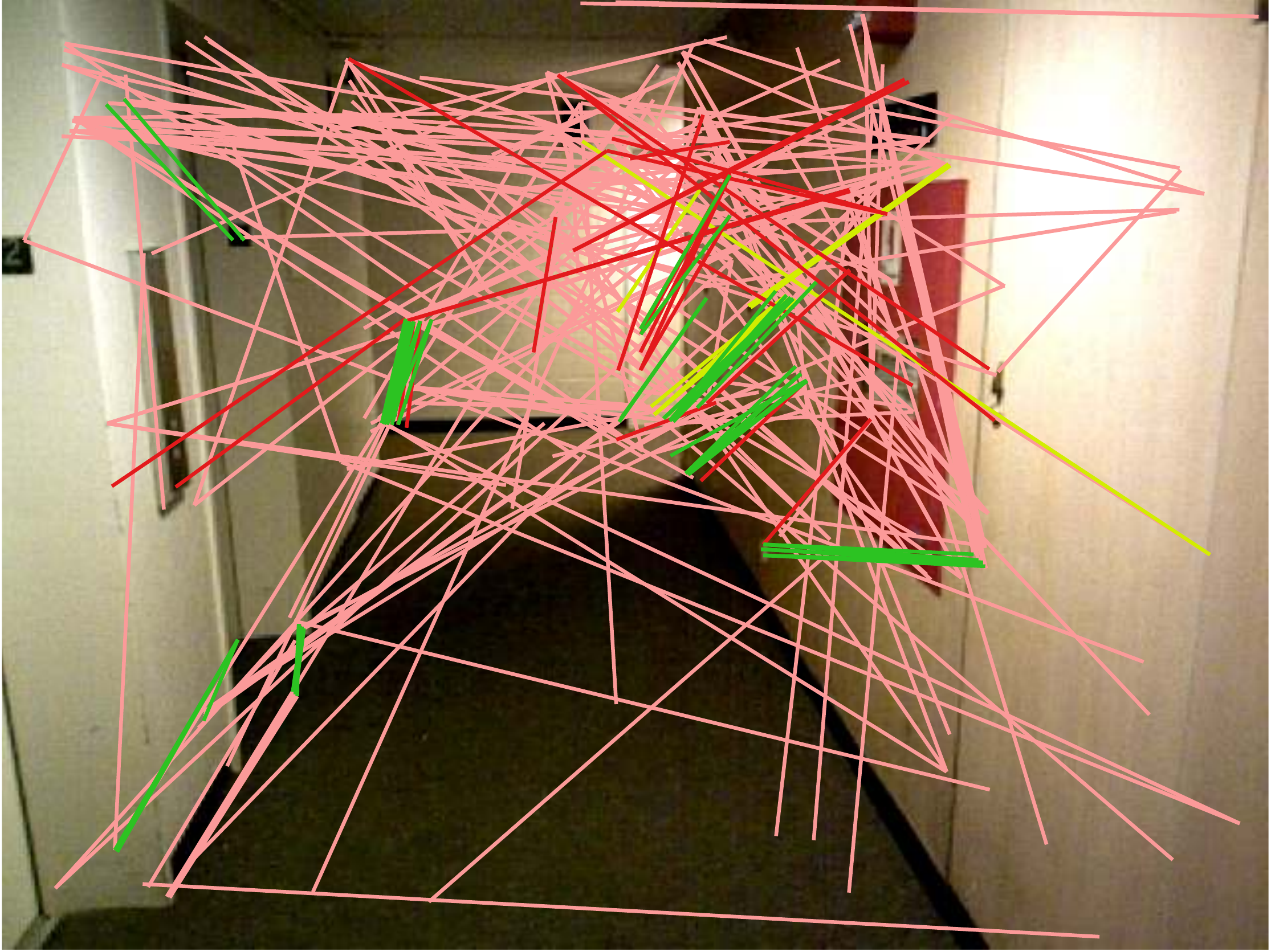}
	\includegraphics[height=7.5em]{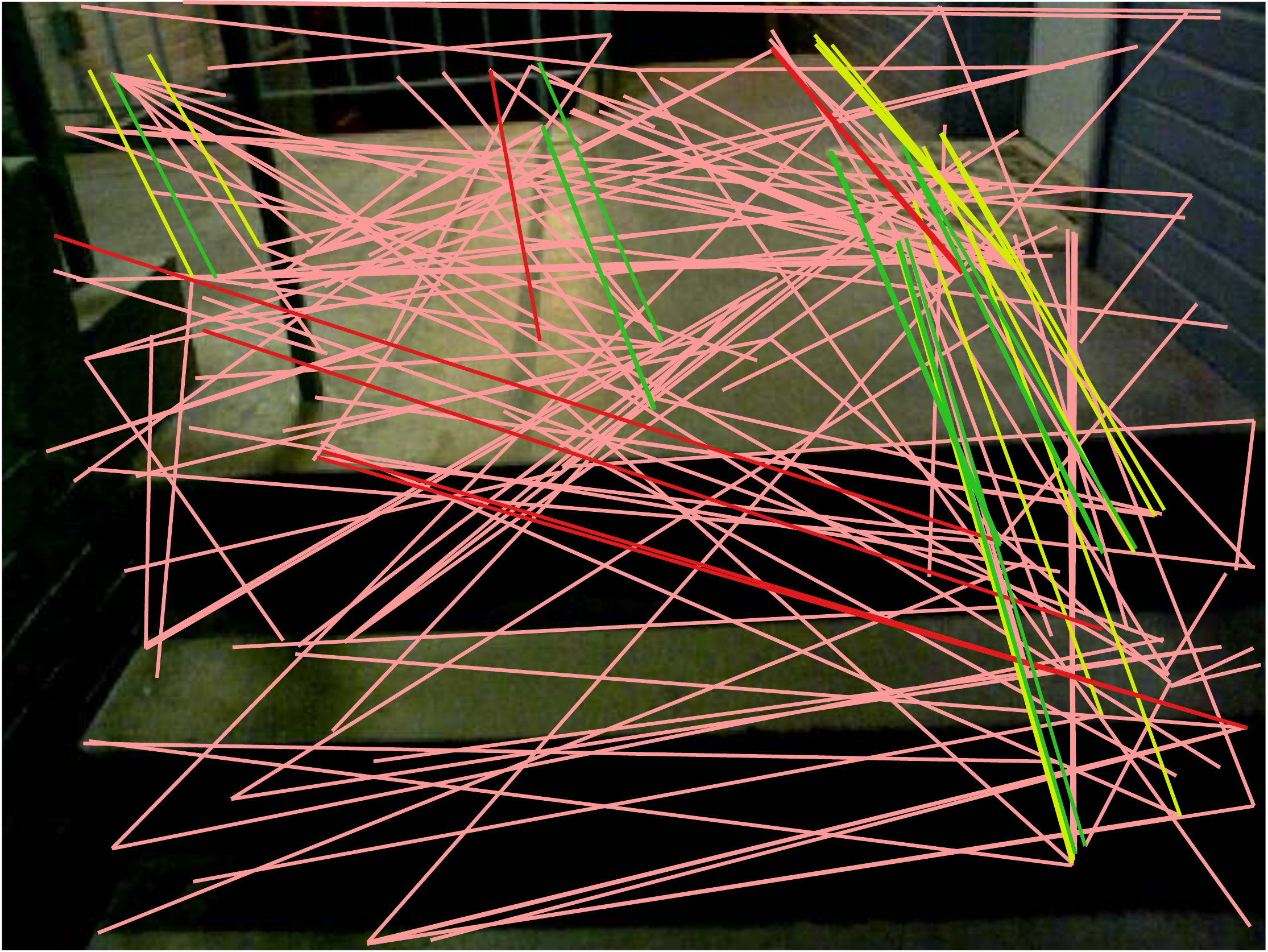}
	\includegraphics[height=7.5em]{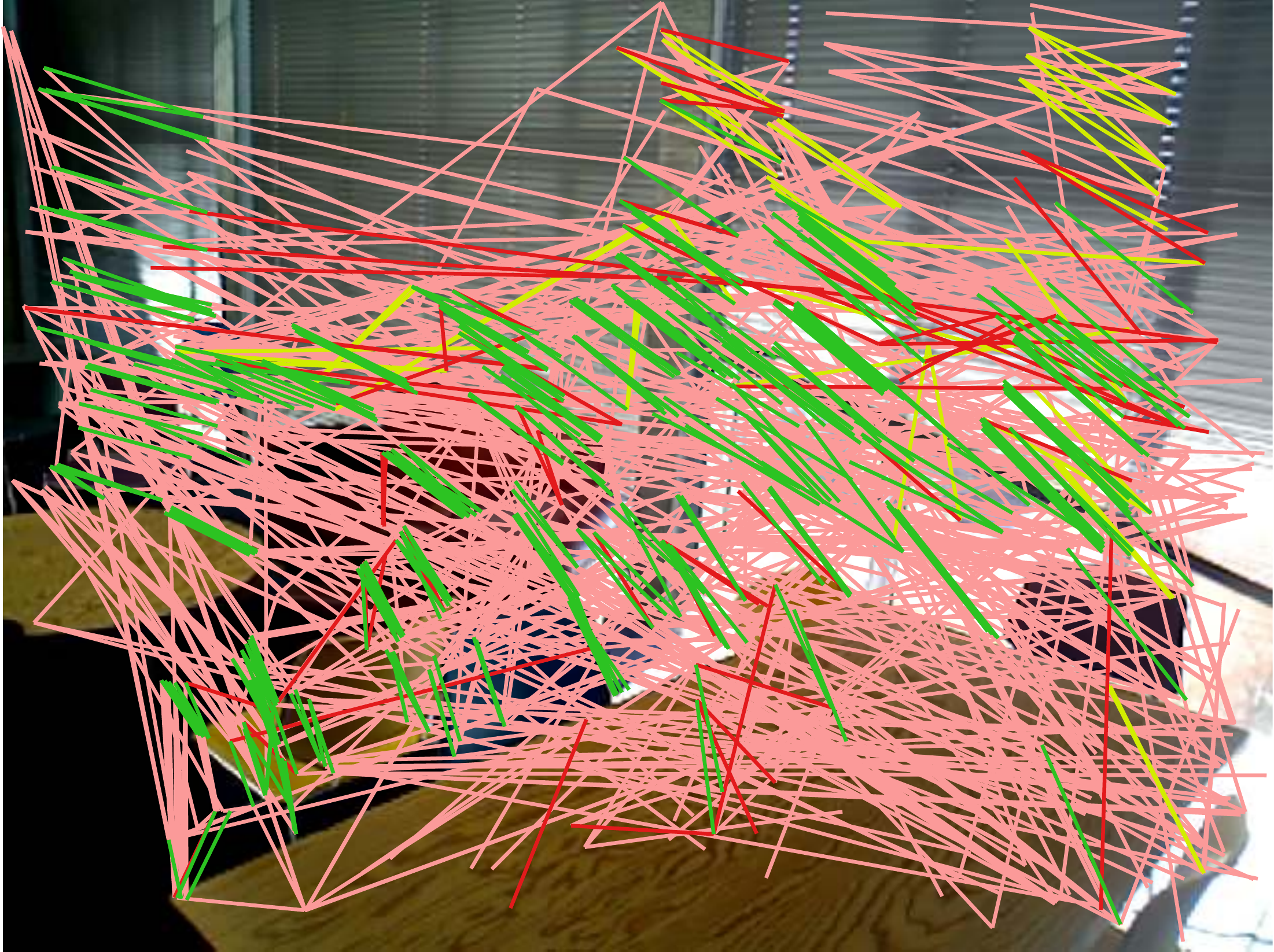}
	\includegraphics[height=7.5em]{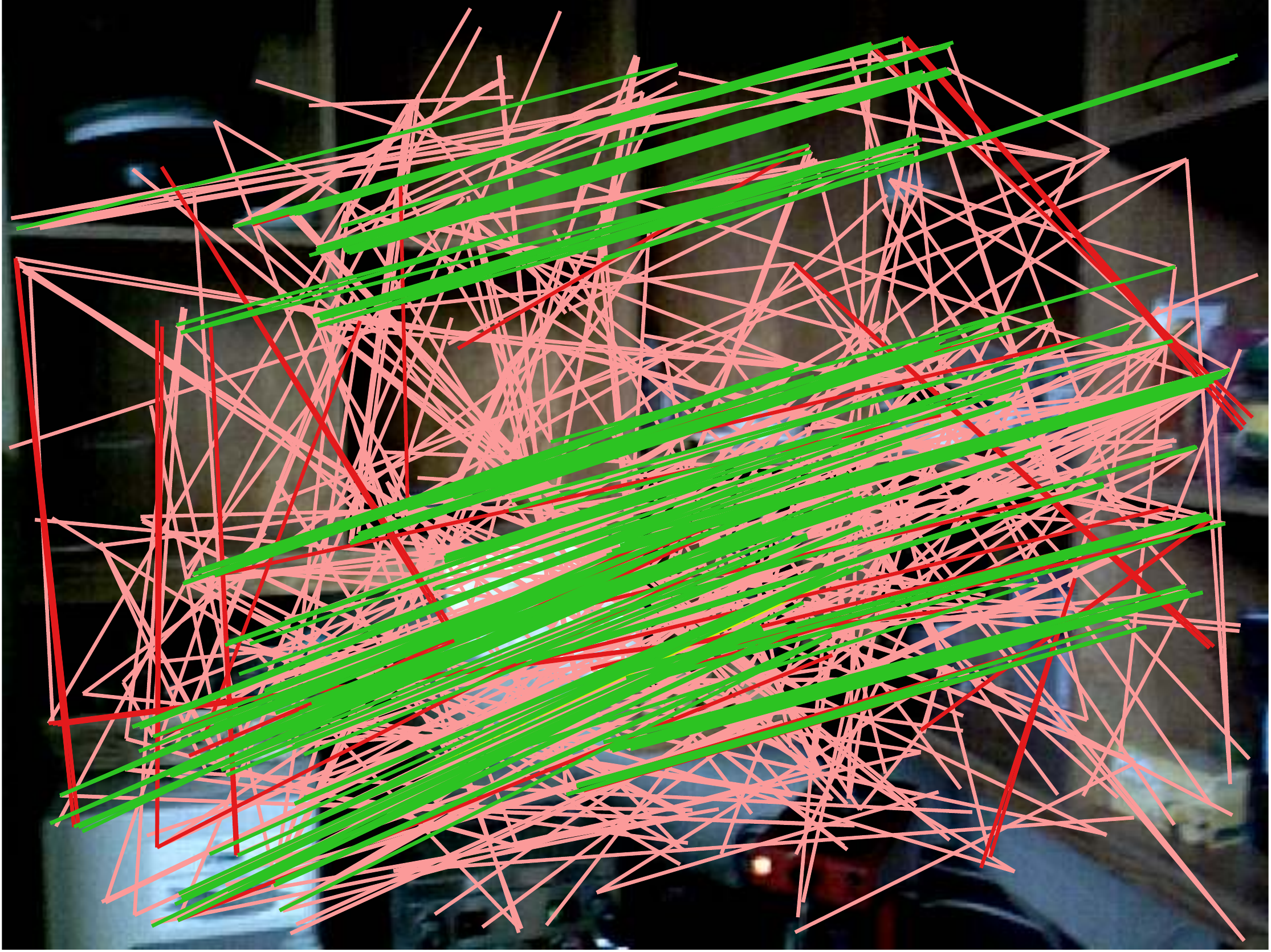}
	\includegraphics[height=7.5em]{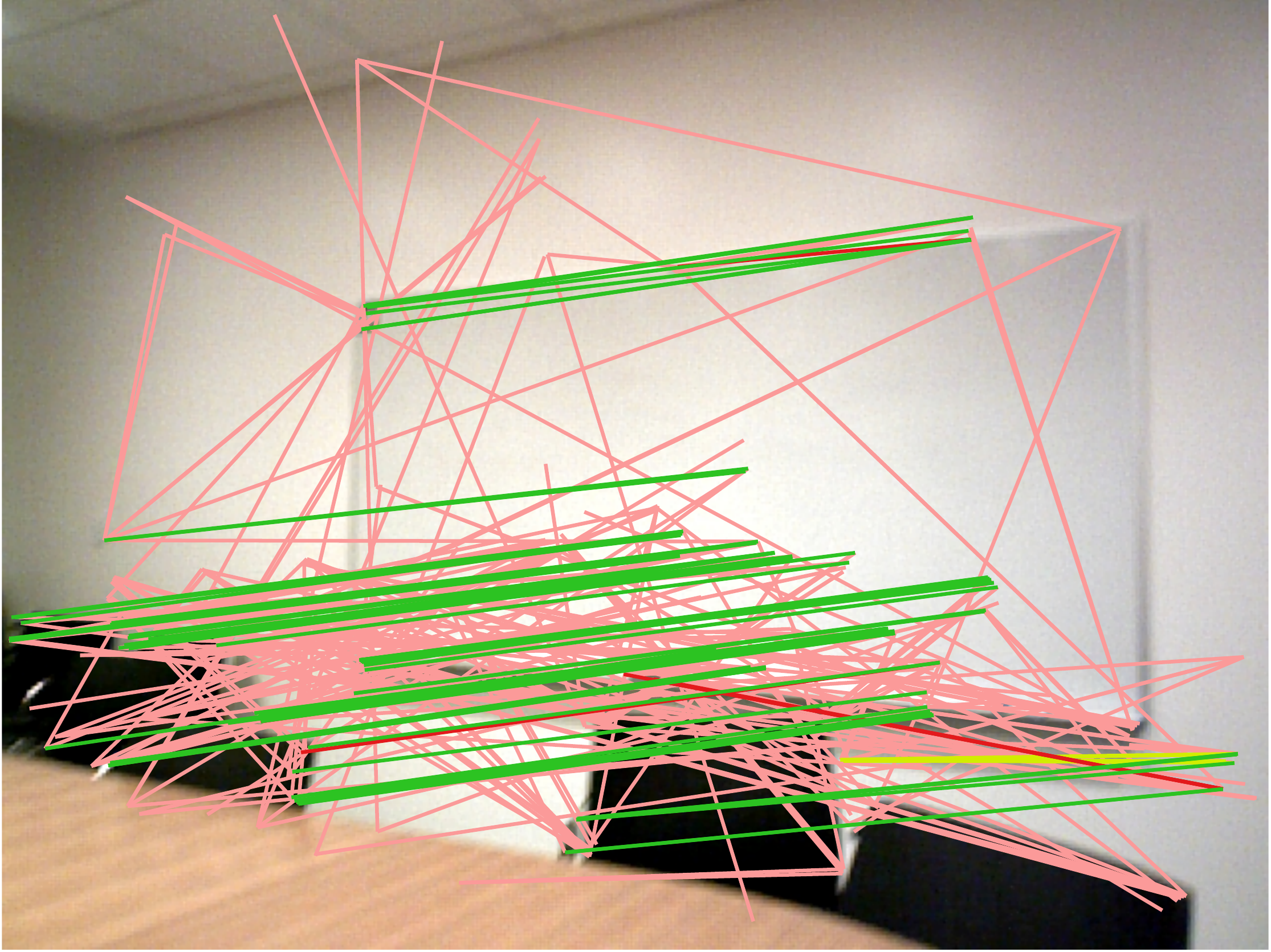}
	\\
	\vspace{0.5em}
	\rotatebox[origin=l]{90}{\mbox{\hspace{2em}BM}}
	\includegraphics[height=7.5em]{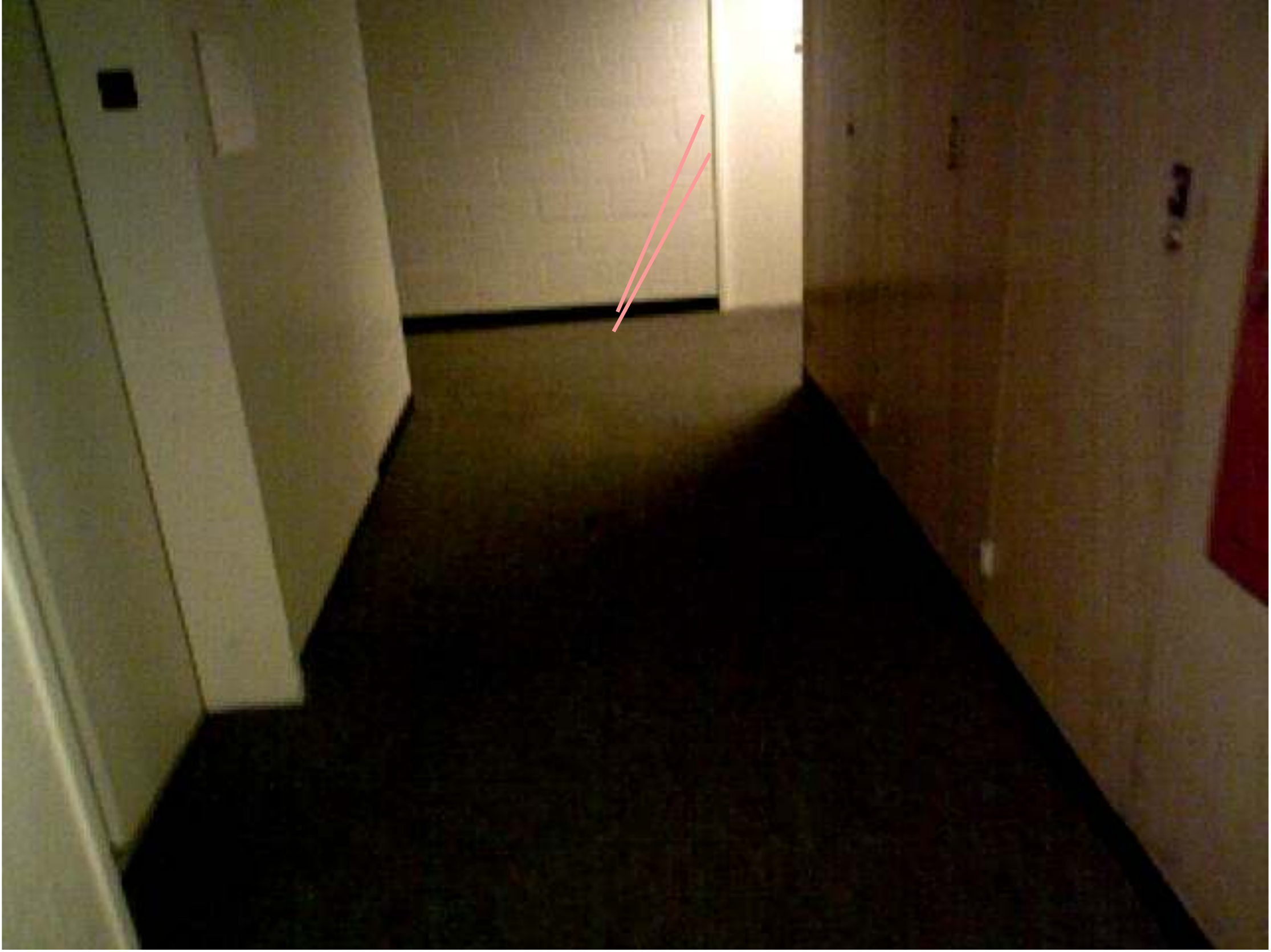}
	\includegraphics[height=7.5em]{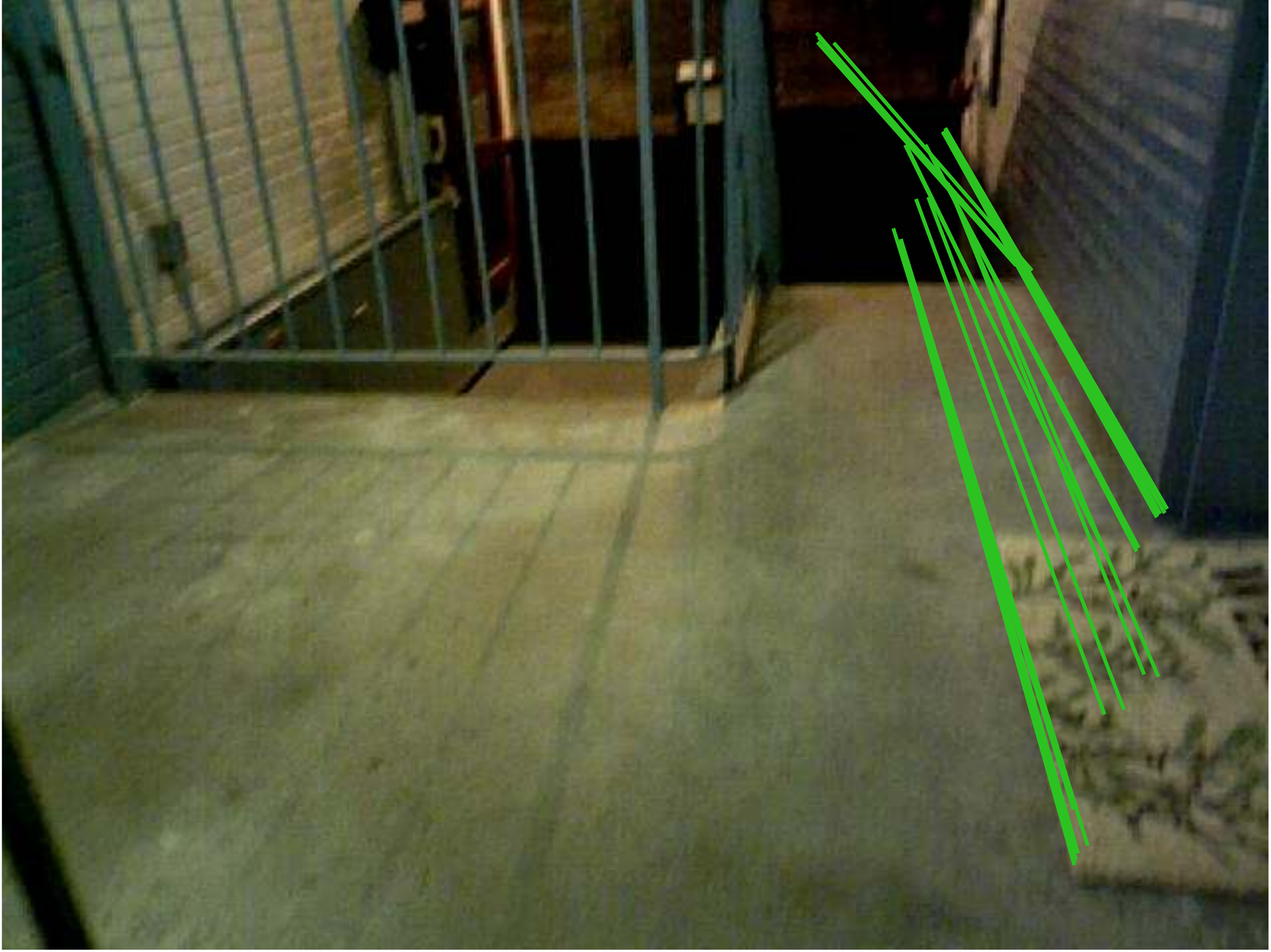}
	\includegraphics[height=7.5em]{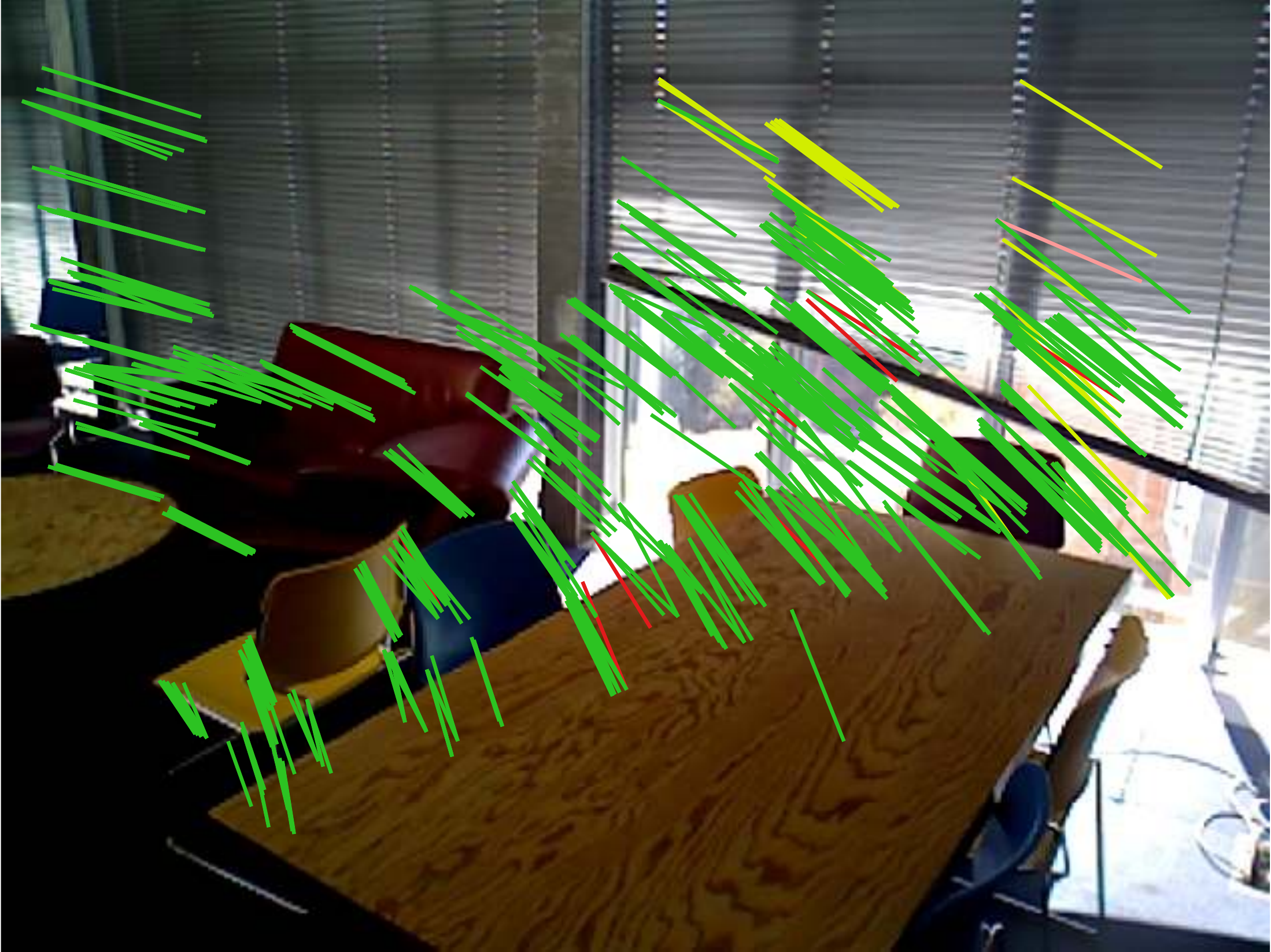}
	\includegraphics[height=7.5em]{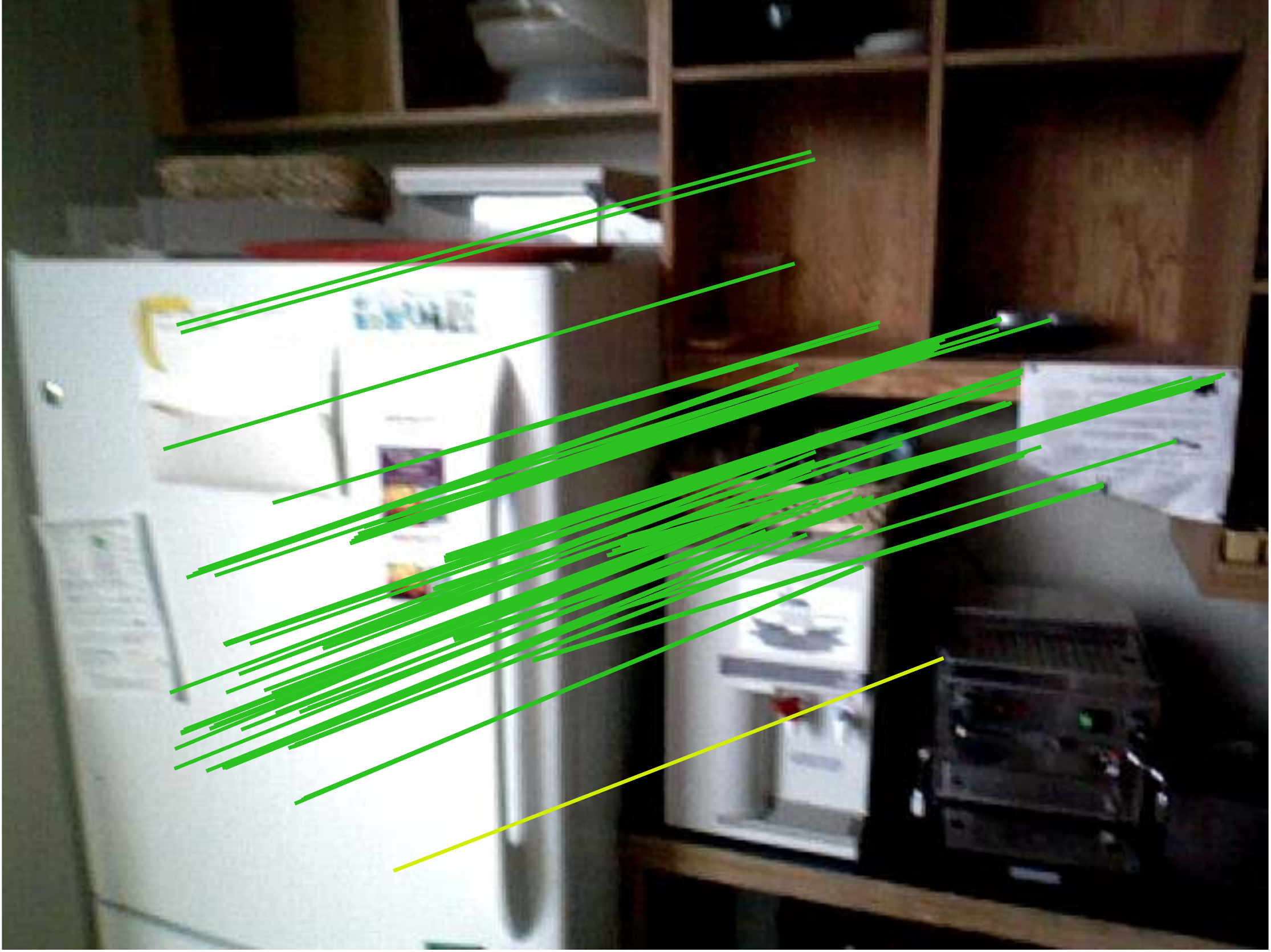}
	\includegraphics[height=7.5em]{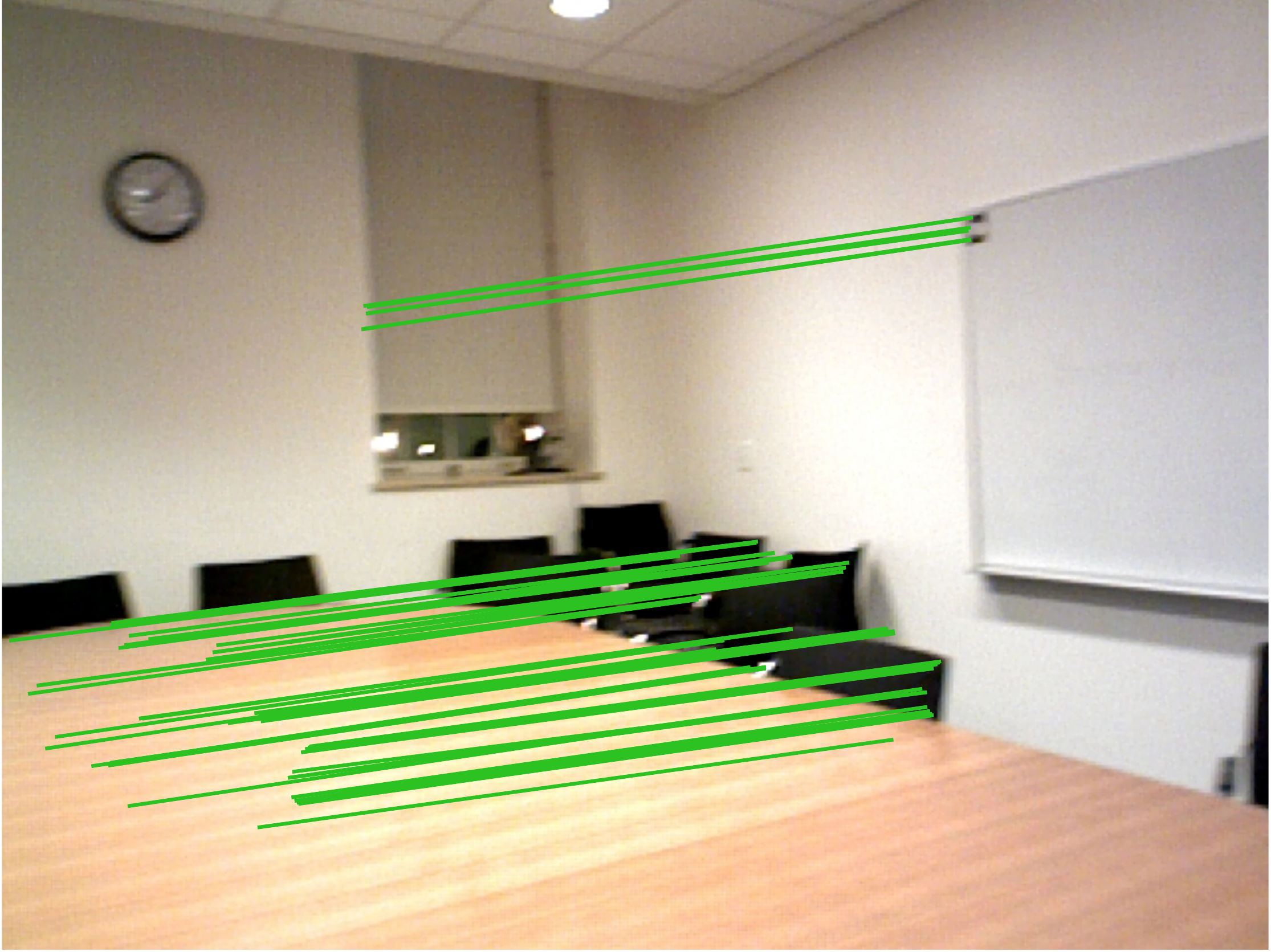}
	\\
	\begin{flushleft}
		\hspace*{4em}MIT Corridor\hspace{4.75em}MIT Stairs\hspace{4.75em}MIT LoungeA \hspace{4em}MIT LoungeB\hspace{1.75em}Brown Cognitive Science
	\end{flushleft}
	\caption{\label{example_4b}
		SUN3D local spatial filter matches according to the best configuration setup, the images of the input pair alternate among the rows. For each method inlier (yellow, green) and outlier (red and light red) clusters are shown, as well as the 1SAC filtered matches (green, red) (see Sec.~\ref{eval_dt}, best viewed in color and zoomed in).}
\end{figure*}

Figures~\ref{example_1a}-\ref{example_4b} provide a visual qualitative comparison of the local spatial filters evaluated in Sec.~\ref{eval_dt}, which agrees with the quantitative analysis presented in the main text. Clearly, there is no method that gives always, and with respect to all the evaluated criteria, the best results. Anyways, DTM appears to be quite robust and generally provides good results, comparable with the best spatial filters both in terms of precision and recall. Notice also that, as reported in Sec.~\ref{ks}, DTM does not require to set parameters to define the neighborhoods or the criteria for the selection of the initial and output matches. From the visual comparison, with respect to other spatial filters such as ACNe and PGM, DTM wrong matches (red and light red) often concern image areas near the triangulation boundary, which lacks a neighborhood covering in all directions. These wrong matches are generally removed by 1SAC (light red), that unlike the case of the other spatial filters, for DTM seems to remove less inliers (yellow), maybe due to the kind of outliers. In order to be robust, inliers should be able to survive to 1SAC (green) and to be well distributed on the image area. These observations can better appreciated for instance in the LeuvenB, Fountain01, Valbonne, Harvard Restroom and MIT Office image pairs.

\begin{figure*}
	\center
	\vspace{-40em}
	\rotatebox[origin=c]{90}{\hspace{1.5em} non-planar\hspace{8.5em}planar}
	\subfloat[global behaviour]{\label{plot_all_inl}
		\begin{minipage}[c]{0.21\textwidth}
			\includegraphics[height=\textwidth]{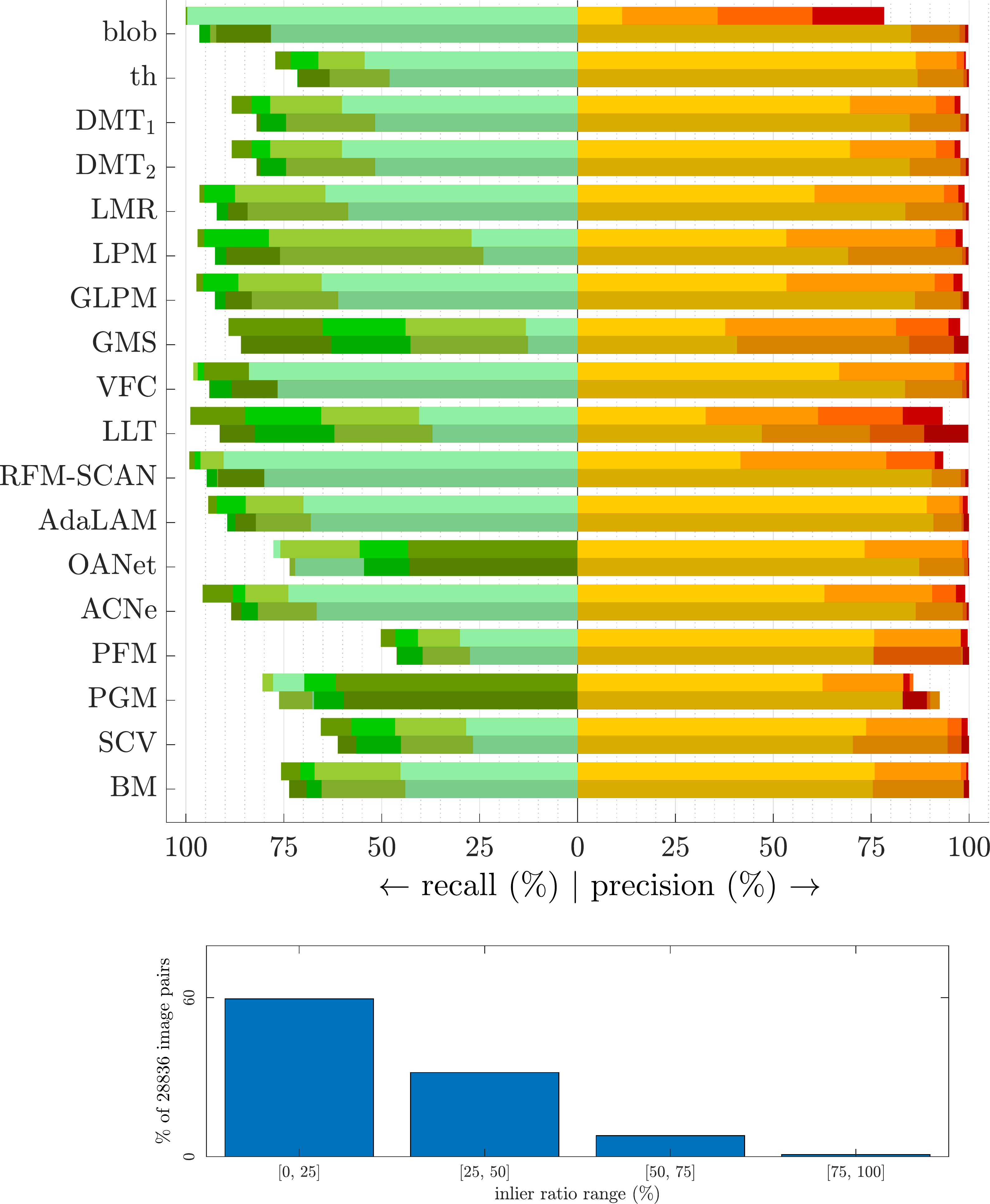}
			\\[1em]
			\includegraphics[height=\textwidth]{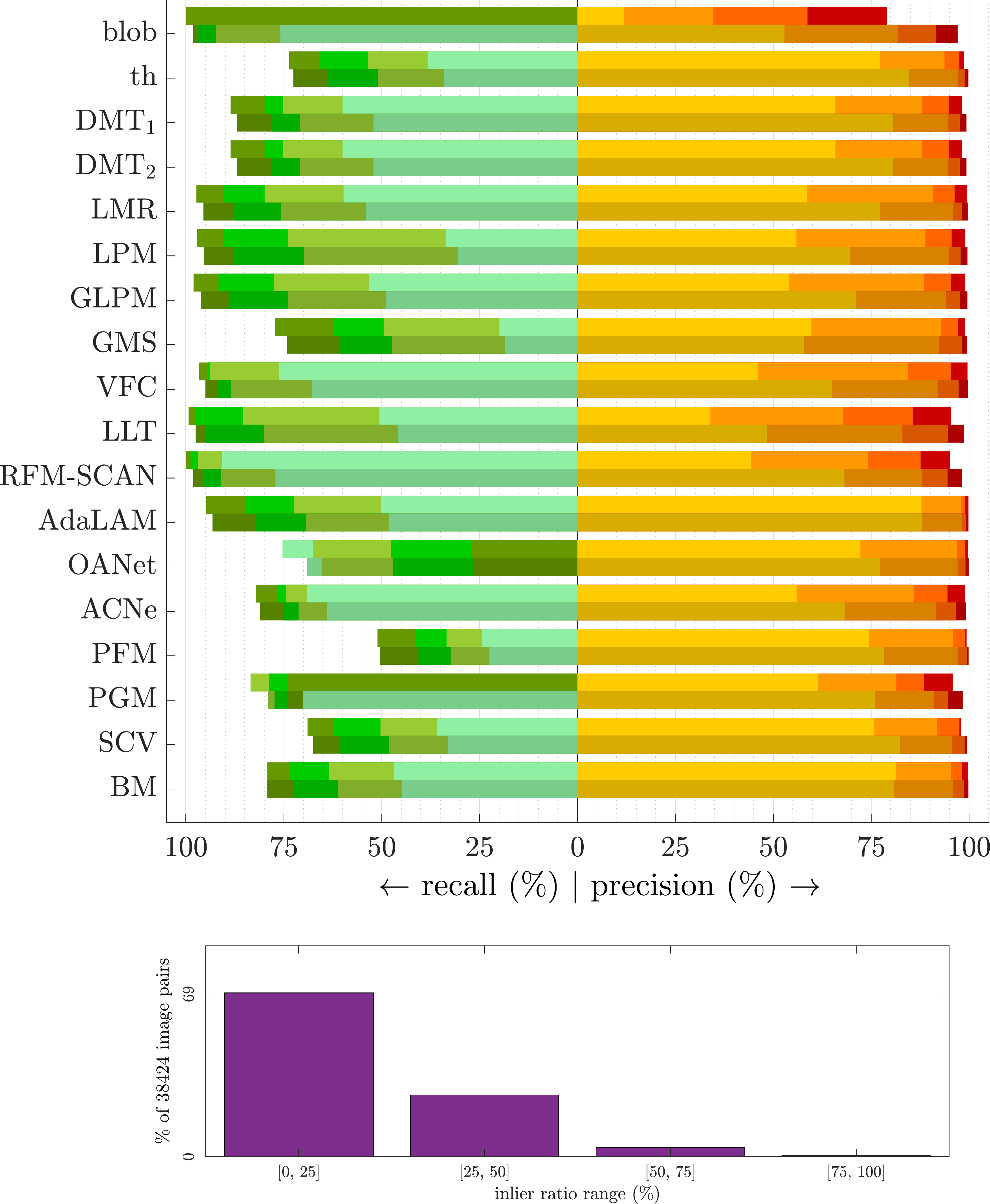}
		\end{minipage}
	}
	\subfloat[baseline configuration]{\label{plot_sift_rsift_inl}
		\begin{minipage}[c]{0.21\textwidth}
			\includegraphics[height=\textwidth]{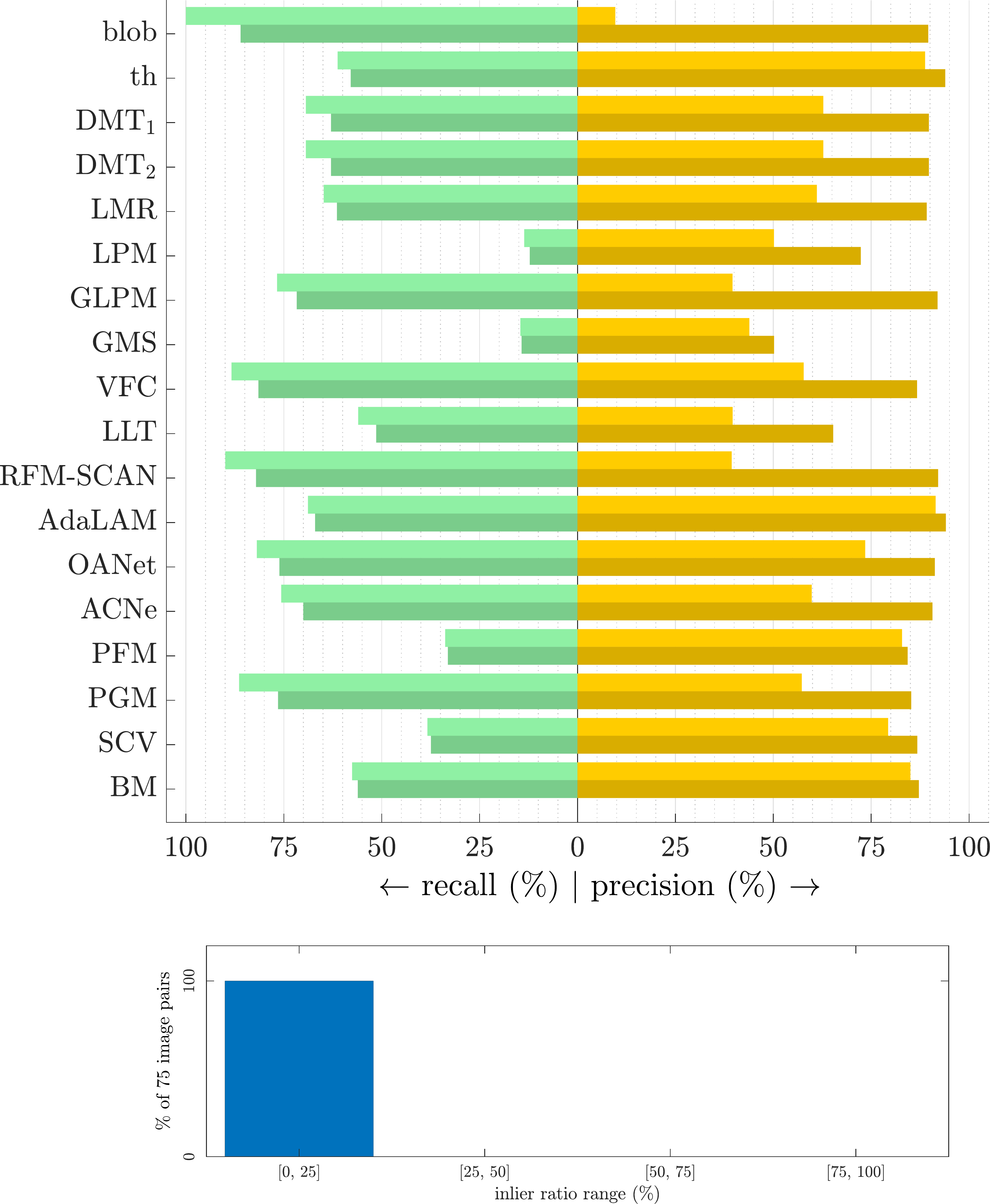}
			\\[1em]
			\includegraphics[height=\textwidth]{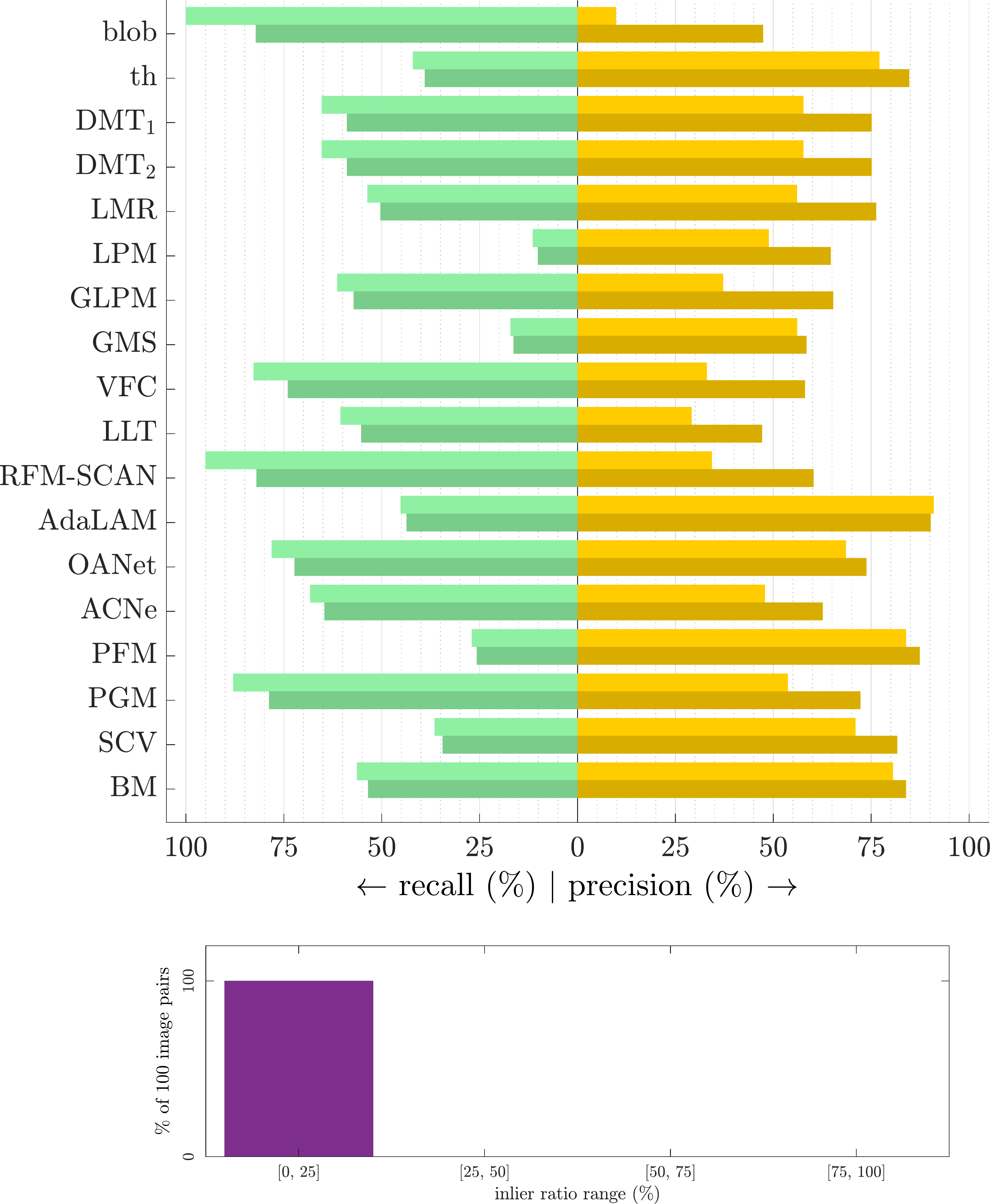}
		\end{minipage}
	}
	\subfloat[best configuration]{\label{plot_hz_sos_inl}
		\begin{minipage}[c]{0.21\textwidth}
			\includegraphics[height=\textwidth]{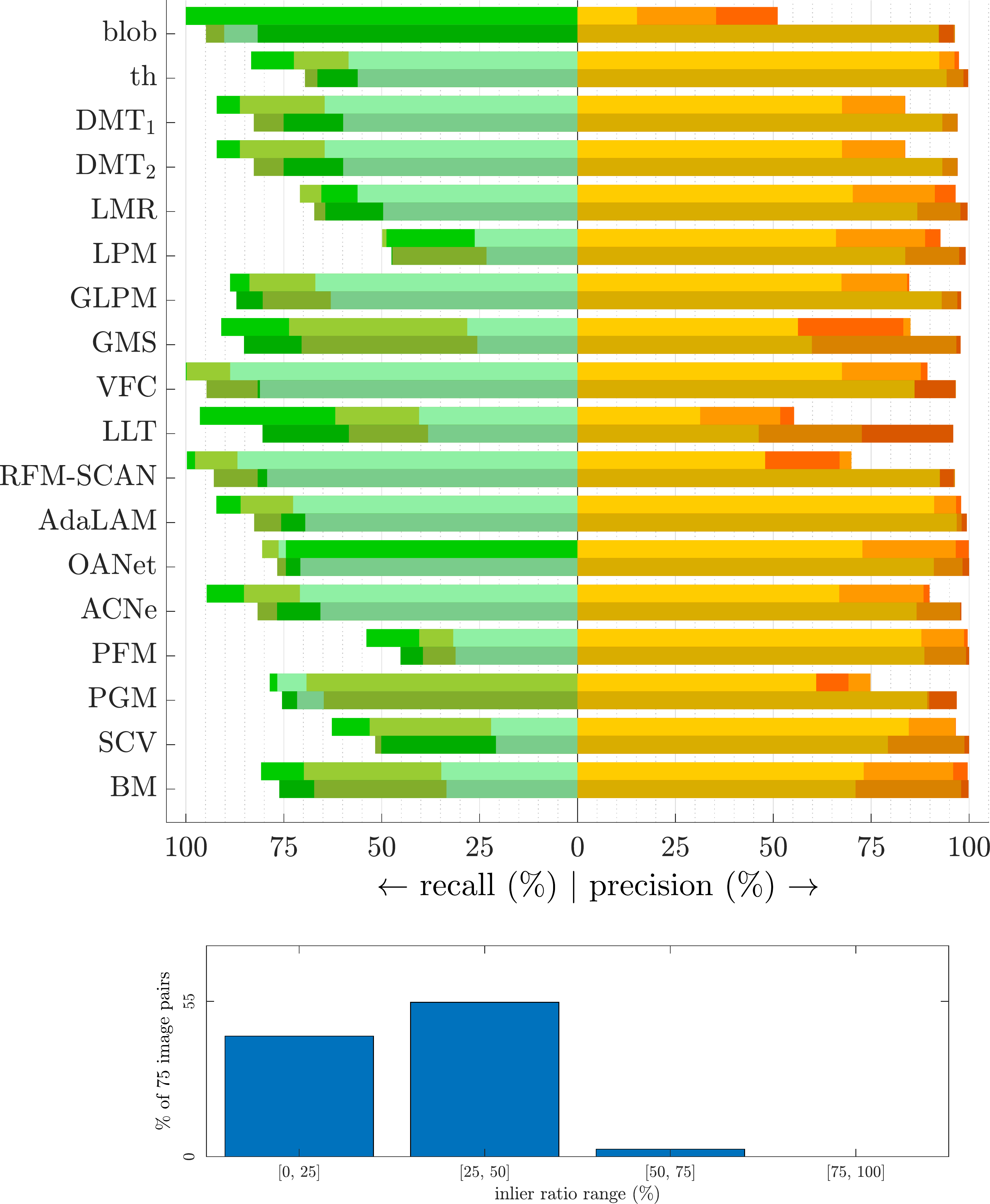}
			\\[1em]
			\includegraphics[height=\textwidth]{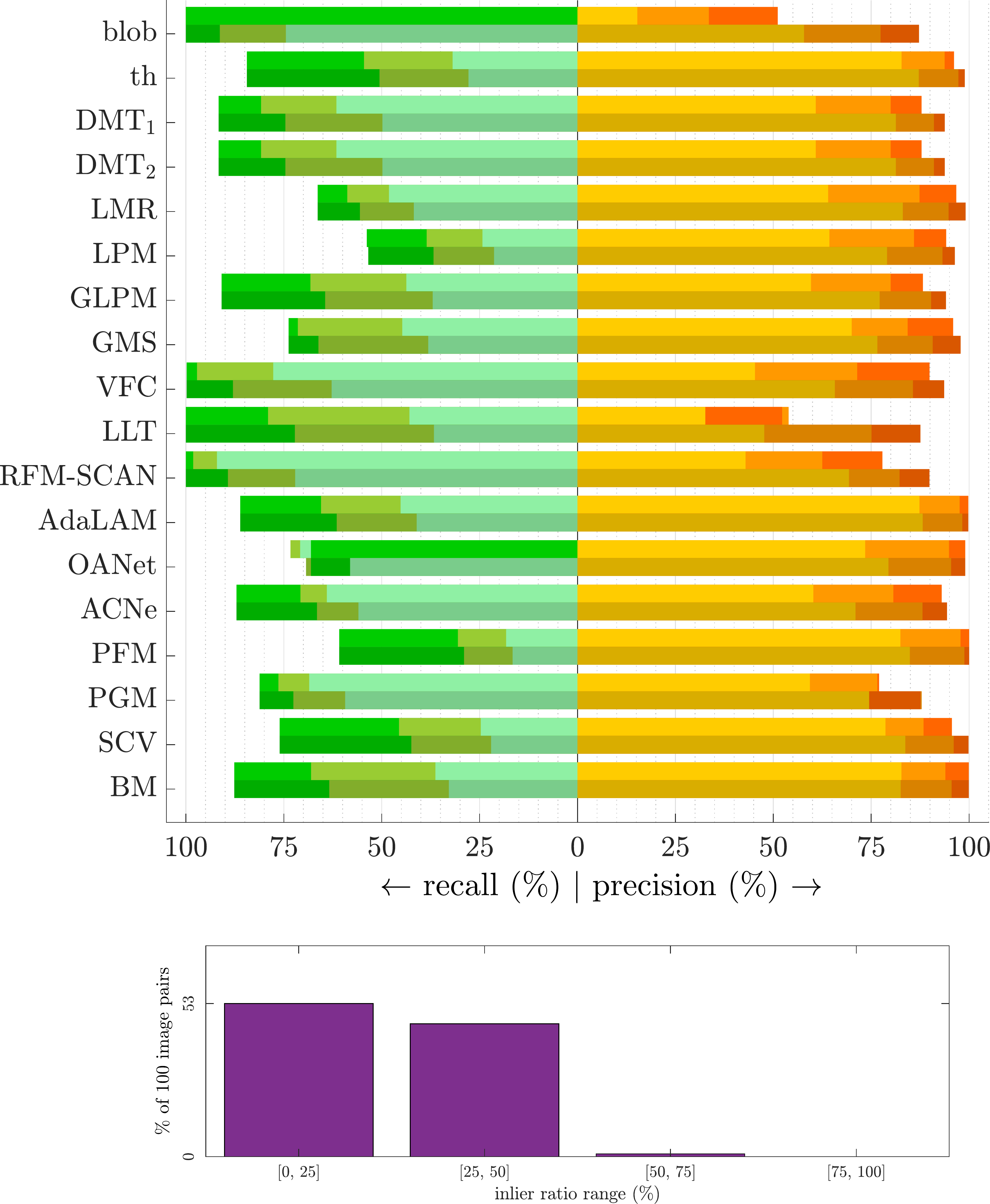}
		\end{minipage}
	}
	\hspace{0.5em}	
	\rotatebox[origin=c]{90}{\hspace{3.5em}SIFT+SOSNet\hspace{5em}HarrisZ+SOSNet}
	\subfloat[SUN3D]{\label{plot_sun3d}
		\begin{minipage}[c]{0.21\textwidth}
			\includegraphics[height=\textwidth]{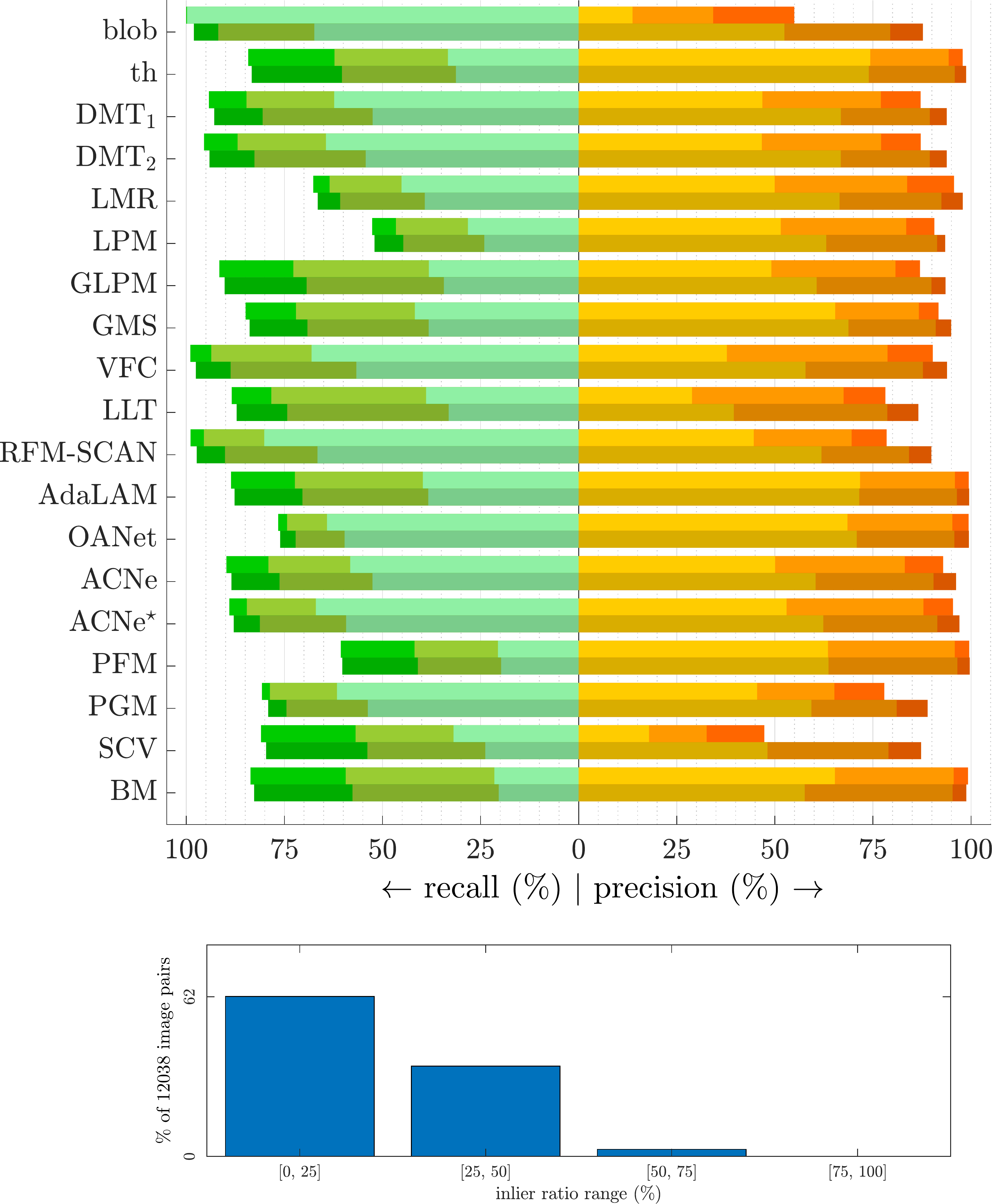}
			\\[1em]
			\includegraphics[height=\textwidth]{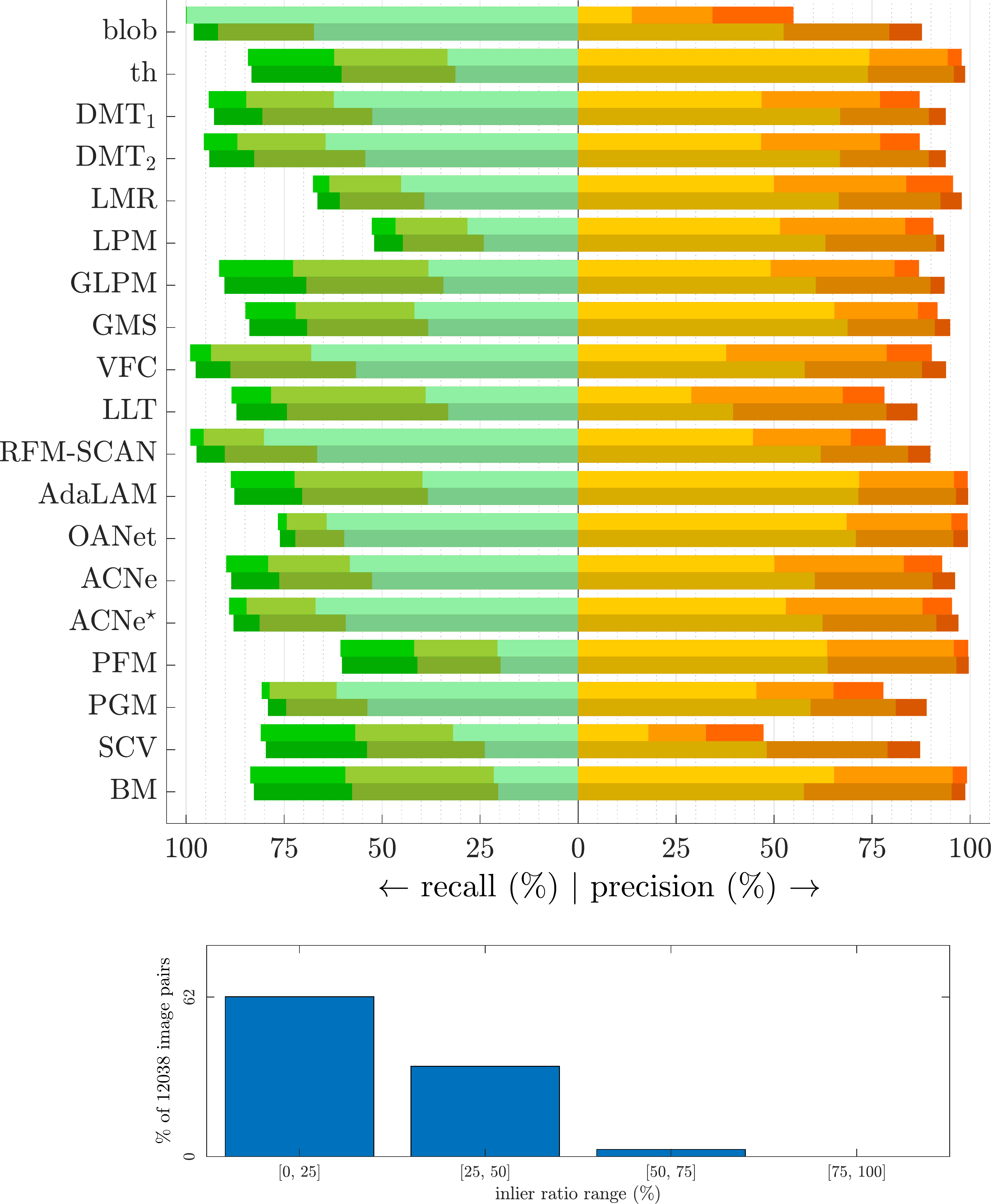}
		\end{minipage}
	}
	\rotatebox[origin=c]{90}{\hfil \includegraphics[height=0.04\textwidth]{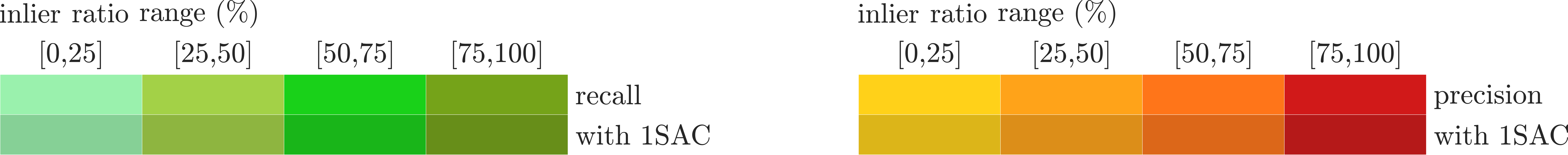}\hfil}	
	\caption{\label{inl_plot}
		Average precision/recall values of the local spatial filters according to the inlier ratio. The recall is computed with respect to the ground truth matches found by blob matching (see text for details, best viewed in color and zoomed in).}
\end{figure*}

Figure~\ref{inl_plot} shows the precision and recall histograms according to different inlier ranges for the corresponding scatter plots of Fig.~\ref{pr_plot}. In addition, further histograms indicating the proportion of image pairs having inliers within a specific range for each setup are reported. From the analysis of these histograms, the same observations discussed in Sec.~\ref{eval_dt} can be drawn out. In particular, it appears that OANet and ACNE suffer less than DTM of outlier contamination. Nevertheless, these differences after 1SAC are reduced. OANet and ACNe implicitly contains epipolar constraints which are forced into DTM by 1SAC. Under this consideration DTM appears still quite robust. 

\begin{table}[t!]
	\rowcolors{3}{lightgray}{}	
	\caption{Global statistics for the local spatial filters. The average planar/non-planar number of ground-truth matches per image pair is 457/382 (blob, for recall), 792/602 (all matches, for recall$^\star$)}\label{table_all}		
	\centering
	\resizebox{0.47\textwidth}{!}{
		\begin{tabular}{rrrrrrrrrr}
			& & & \multicolumn{1}{l}{\rot[37]{precision (\%)}} & \multicolumn{1}{l}{\rot[37]{recall (\%)}} & \multicolumn{1}{l}{\rot[37]{recall$^\star$ (\%)}} & \multicolumn{1}{l}{\rot[37]{\shortstack[l]{correct\vspace{-0.25em}\\matches}}} & \multicolumn{1}{l}{\rot[37]{\shortstack[l]{output\vspace{-0.35em}\\matches}}} & \multicolumn{1}{l}{\rot[37]{failures}} & \multicolumn{1}{l}{\rot[37]{time (s)}} \\
			\toprule
			\cellcolor{white}{} & \cellcolor{white}{} & blob & 23.48 & 99.79 & 56.23 & 457 & 3806 & 0 & -- \\
			\cellcolor{white}{} & \cellcolor{white}{} & th & 90.29 & 61.21 & 36.49 & 310 & 370 & 0 & 0.000 \\
			\cellcolor{white}{} & \cellcolor{white}{} & DTM$_1$ & 78.78 & 67.84 & 40.05 & 356 & 593 & 1 & 0.283 \\
			\cellcolor{white}{} & \cellcolor{white}{} & DTM$_2$ & 79.82 & 78.85 & 45.84 & 399 & 651 & 0 & 0.519 \\
			\cellcolor{white}{} & \cellcolor{white}{} & LMR & 74.14 & 74.32 & 43.11 & 384 & 588 & 2 & 0.065 \\
			\cellcolor{white}{} & \cellcolor{white}{} & LPM & 69.05 & 49.26 & 27.99 & 259 & 362 & 4 & 0.017 \\
			\cellcolor{white}{} & \cellcolor{white}{} & GLPM & 69.48 & 76.34 & 45.46 & 398 & 952 & 3 & 0.029 \\
			\cellcolor{white}{} & \cellcolor{white}{} & GMS & 56.51 & 27.62 & 18.37 & 200 & 365 & 28 & 0.253 \\
			\cellcolor{white}{} & \cellcolor{white}{} & VFC & 78.91 & 89.48 & 51.52 & 440 & 781 & 4 & 0.031 \\
			\cellcolor{white}{} & \cellcolor{white}{} & LLT & 46.39 & 52.47 & 31.63 & 298 & 591 & 28 & 1.665 \\
			\cellcolor{white}{} & \cellcolor{white}{} & RFM-SCAN & 57.71 & 92.84 & 52.65 & 438 & 1237 & 0 & 0.774 \\
			\cellcolor{white}{} & \cellcolor{white}{} & AdaLAM & 93.11 & 78.23 & 46.40 & 395 & 542 & 3 & 0.226 \\
			\cellcolor{white}{} & \cellcolor{white}{} & OANet & 83.47 & 75.06 & 42.22 & 362 & 552 & 1 & 0.082 \\
			\cellcolor{white}{} & \cellcolor{white}{} & ACNe & 74.69 & 78.61 & 45.33 & 385 & 638 & 0 & 0.047 \\
			\cellcolor{white}{} & \cellcolor{white}{} & PFM & 85.40 & 36.39 & 21.79 & 180 & 195 & 8 & 0.114 \\
			\cellcolor{white}{} & \cellcolor{white}{} & PGM & 71.10 & 78.06 & 43.78 & 341 & 486 & 6 & 5.051 \\
			\cellcolor{white}{} & \cellcolor{white}{} & SCV & 55.52 & 40.30 & 24.35 & 240 & 916 & 4 & 0.658 \\
			\cellcolor{white}{} & \multirow{-18}{*}{\rotatebox[origin=c]{90}{\cellcolor{white}{no model}}} & BM & 85.27 & 55.35 & 33.06 & 328 & 445 & 9 & 0.892 \\
			\cmidrule(l{2pt}r{2pt}){2-10}
			\cellcolor{white}{} & \cellcolor{white}{} & blob & 89.94 & 84.15 & 49.54 & 418 & 620 & 4 & -- \\
			\cellcolor{white}{} & \cellcolor{white}{} & th & 91.35 & 55.81 & 33.76 & 294 & 340 & 5 & -- \\
			\cellcolor{white}{} & \cellcolor{white}{} & DTM$_1$ & 89.68 & 60.72 & 36.57 & 333 & 482 & 5 & -- \\
			\cellcolor{white}{} & \cellcolor{white}{} & DTM$_2$ & 90.61 & 70.26 & 41.72 & 372 & 529 & 5 & -- \\
			\cellcolor{white}{} & \cellcolor{white}{} & LMR & 89.64 & 69.35 & 40.69 & 366 & 475 & 6 & -- \\
			\cellcolor{white}{} & \cellcolor{white}{} & LPM & 80.62 & 46.18 & 26.51 & 248 & 311 & 12 & -- \\
			\cellcolor{white}{} & \cellcolor{white}{} & GLPM & 91.54 & 71.87 & 42.98 & 378 & 534 & 4 & -- \\
			\cellcolor{white}{} & \cellcolor{white}{} & GMS & 59.58 & 26.73 & 17.79 & 193 & 316 & 29 & -- \\
			\cellcolor{white}{} & \cellcolor{white}{} & VFC & 89.56 & 83.11 & 48.39 & 417 & 614 & 5 & -- \\
			\cellcolor{white}{} & \cellcolor{white}{} & LLT & 59.60 & 48.94 & 29.77 & 281 & 408 & 28 & -- \\
			\cellcolor{white}{} & \cellcolor{white}{} & RFM-SCAN & 93.43 & 84.57 & 49.13 & 414 & 612 & 2 & -- \\
			\cellcolor{white}{} & \cellcolor{white}{} & AdaLAM & 94.41 & 75.59 & 44.86 & 380 & 513 & 3 & -- \\
			\cellcolor{white}{} & \cellcolor{white}{} & OANet & 91.95 & 70.67 & 40.21 & 346 & 472 & 4 & -- \\
			\cellcolor{white}{} & \cellcolor{white}{} & ACNe & 91.10 & 72.79 & 42.67 & 366 & 493 & 5 & -- \\
			\cellcolor{white}{} & \cellcolor{white}{} & PFM & 85.38 & 34.15 & 20.63 & 174 & 186 & 10 & -- \\
			\cellcolor{white}{} & \cellcolor{white}{} & PGM & 86.48 & 69.92 & 40.02 & 318 & 365 & 7 & -- \\
			\cellcolor{white}{} & \cellcolor{white}{} & SCV & 78.80 & 37.10 & 22.55 & 228 & 277 & 14 & -- \\
			\multirow{-36}{*}{\rotatebox[origin=c]{90}{\cellcolor{white}{planar}}} & \multirow{-18}{*}{\rotatebox[origin=c]{90}{\cellcolor{white}{1SAC}}} & BM & 85.00 & 53.69 & 32.11 & 316 & 419 & 10 & -- \\
			\midrule
			\cellcolor{white}{} & \cellcolor{white}{} & blob & 19.86 & 100.00 & 60.99 & 382 & 3834 & 0 & -- \\
			\cellcolor{white}{} & \cellcolor{white}{} & th & 82.26 & 44.89 & 28.71 & 185 & 235 & 1 & 0.000 \\
			\cellcolor{white}{} & \cellcolor{white}{} & DTM$_1$ & 73.01 & 64.77 & 41.34 & 269 & 501 & 0 & 0.271 \\
			\cellcolor{white}{} & \cellcolor{white}{} & DTM$_2$ & 72.76 & 72.71 & 45.90 & 298 & 546 & 0 & 0.512 \\
			\cellcolor{white}{} & \cellcolor{white}{} & LMR & 68.60 & 66.21 & 40.85 & 269 & 471 & 1 & 0.067 \\
			\cellcolor{white}{} & \cellcolor{white}{} & LPM & 65.92 & 46.48 & 26.73 & 177 & 264 & 3 & 0.017 \\
			\cellcolor{white}{} & \cellcolor{white}{} & GLPM & 65.26 & 62.86 & 40.62 & 271 & 644 & 4 & 0.018 \\
			\cellcolor{white}{} & \cellcolor{white}{} & GMS & 69.98 & 29.54 & 20.40 & 161 & 304 & 21 & 0.251 \\
			\cellcolor{white}{} & \cellcolor{white}{} & VFC & 58.20 & 81.63 & 51.75 & 345 & 897 & 6 & 0.037 \\
			\cellcolor{white}{} & \cellcolor{white}{} & LLT & 45.01 & 61.68 & 40.20 & 279 & 737 & 23 & 1.602 \\
			\cellcolor{white}{} & \cellcolor{white}{} & RFM-SCAN & 54.03 & 92.71 & 57.27 & 365 & 1145 & 0 & 0.735 \\
			\cellcolor{white}{} & \cellcolor{white}{} & AdaLAM & 91.21 & 58.95 & 37.50 & 254 & 339 & 3 & 0.165 \\
			\cellcolor{white}{} & \cellcolor{white}{} & OANet & 79.83 & 72.00 & 44.49 & 282 & 468 & 1 & 0.083 \\
			\cellcolor{white}{} & \cellcolor{white}{} & ACNe & 65.47 & 70.86 & 43.98 & 281 & 564 & 0 & 0.047 \\
			\cellcolor{white}{} & \cellcolor{white}{} & PFM & 82.27 & 28.81 & 18.46 & 112 & 125 & 8 & 0.100 \\
			\cellcolor{white}{} & \cellcolor{white}{} & PGM & 67.62 & 80.37 & 50.02 & 312 & 539 & 1 & 7.119 \\
			\cellcolor{white}{} & \cellcolor{white}{} & SCV & 54.99 & 44.20 & 28.35 & 186 & 812 & 2 & 0.729 \\
			\cellcolor{white}{} & \multirow{-18}{*}{\rotatebox[origin=c]{90}{\cellcolor{white}{no model}}} & BM & 86.24 & 53.59 & 35.00 & 244 & 351 & 5 & 0.729 \\
			\cmidrule(l{2pt}r{2pt}){2-10}
			\cellcolor{white}{} & \cellcolor{white}{} & blob & 61.91 & 80.71 & 51.25 & 334 & 740 & 1 & -- \\
			\cellcolor{white}{} & \cellcolor{white}{} & th & 88.80 & 40.77 & 26.46 & 172 & 204 & 2 & -- \\
			\cellcolor{white}{} & \cellcolor{white}{} & DTM$_1$ & 85.08 & 58.01 & 37.57 & 248 & 394 & 2 & -- \\
			\cellcolor{white}{} & \cellcolor{white}{} & DTM$_2$ & 84.92 & 64.78 & 41.50 & 273 & 427 & 1 & -- \\
			\cellcolor{white}{} & \cellcolor{white}{} & LMR & 82.89 & 60.92 & 38.00 & 253 & 351 & 3 & -- \\
			\cellcolor{white}{} & \cellcolor{white}{} & LPM & 77.19 & 43.01 & 24.93 & 167 & 218 & 6 & -- \\
			\cellcolor{white}{} & \cellcolor{white}{} & GLPM & 78.79 & 58.55 & 37.98 & 255 & 410 & 5 & -- \\
			\cellcolor{white}{} & \cellcolor{white}{} & GMS & 68.63 & 27.88 & 19.28 & 153 & 267 & 25 & -- \\
			\cellcolor{white}{} & \cellcolor{white}{} & VFC & 73.39 & 74.14 & 47.31 & 318 & 573 & 7 & -- \\
			\cellcolor{white}{} & \cellcolor{white}{} & LLT & 59.41 & 56.94 & 37.28 & 260 & 463 & 24 & -- \\
			\cellcolor{white}{} & \cellcolor{white}{} & RFM-SCAN & 74.42 & 81.51 & 51.33 & 333 & 631 & 1 & -- \\
			\cellcolor{white}{} & \cellcolor{white}{} & AdaLAM & 91.50 & 56.32 & 35.85 & 243 & 322 & 3 & -- \\
			\cellcolor{white}{} & \cellcolor{white}{} & OANet & 83.33 & 66.98 & 41.65 & 267 & 409 & 1 & -- \\
			\cellcolor{white}{} & \cellcolor{white}{} & ACNe & 75.58 & 66.27 & 41.41 & 267 & 440 & 1 & -- \\
			\cellcolor{white}{} & \cellcolor{white}{} & PFM & 85.46 & 26.93 & 17.41 & 106 & 116 & 8 & -- \\
			\cellcolor{white}{} & \cellcolor{white}{} & PGM & 80.59 & 72.88 & 45.76 & 290 & 402 & 2 & -- \\
			\cellcolor{white}{} & \cellcolor{white}{} & SCV & 76.81 & 40.53 & 26.25 & 173 & 265 & 4 & -- \\
			\multirow{-36}{*}{\rotatebox[origin=c]{90}{\cellcolor{white}{non-planar}}} & \multirow{-18}{*}{\rotatebox[origin=c]{90}{\cellcolor{white}{1SAC}}} & BM & 86.28 & 51.35 & 33.56 & 234 & 325 & 7 & -- \\
			\bottomrule
		\end{tabular}
	}
\end{table}

\begin{table}		
	\rowcolors{3}{lightgray}{}
	\vspace{-1.2em}
	\caption{Local spatial filter statistics for the baseline configuration. The average planar/non-planar number of ground-truth matches per image pair is 293/214 (blob, for recall), 740/519 (all matches, for recall$^\star$)}\label{table_rootsift}
	\centering
	\resizebox{0.47\textwidth}{!}{		
		\begin{tabular}{rrrrrrrrrr}
			& & & \multicolumn{1}{l}{\rot[37]{precision (\%)}} & \multicolumn{1}{l}{\rot[37]{recall (\%)}} & \multicolumn{1}{l}{\rot[37]{recall$^\star$ (\%)}} & \multicolumn{1}{l}{\rot[37]{\shortstack[l]{correct\vspace{-0.25em}\\matches}}} & \multicolumn{1}{l}{\rot[37]{\shortstack[l]{output\vspace{-0.35em}\\matches}}} & \multicolumn{1}{l}{\rot[37]{failures}} & \multicolumn{1}{l}{\rot[37]{time (s)}} \\
			\toprule
			\cellcolor{white}{} & \cellcolor{white}{} & blob & 20.63 & 100.00 & 37.60 & 293 & 1329 & 0 & -- \\
			\cellcolor{white}{} & \cellcolor{white}{} & th & 85.02 & 61.62 & 25.99 & 212 & 227 & 1 & 0.000 \\
			\cellcolor{white}{} & \cellcolor{white}{} & DTM$_1$ & 84.23 & 63.99 & 25.69 & 228 & 244 & 2 & 0.168 \\
			\cellcolor{white}{} & \cellcolor{white}{} & DTM$_2$ & 85.19 & 77.91 & 31.02 & 265 & 284 & 2 & 0.240 \\
			\cellcolor{white}{} & \cellcolor{white}{} & LMR & 70.29 & 79.58 & 31.90 & 274 & 320 & 6 & 0.036 \\
			\cellcolor{white}{} & \cellcolor{white}{} & LPM & 71.61 & 74.84 & 29.54 & 266 & 311 & 3 & 0.008 \\
			\cellcolor{white}{} & \cellcolor{white}{} & GLPM & 84.71 & 68.62 & 29.37 & 258 & 275 & 7 & 0.014 \\
			\cellcolor{white}{} & \cellcolor{white}{} & GMS & 45.48 & 15.69 & 7.31 & 92 & 94 & 40 & 0.260 \\
			\cellcolor{white}{} & \cellcolor{white}{} & VFC & 87.12 & 89.57 & 34.56 & 282 & 298 & 4 & 0.019 \\
			\cellcolor{white}{} & \cellcolor{white}{} & LLT & 40.34 & 39.94 & 16.89 & 190 & 227 & 37 & 0.485 \\
			\cellcolor{white}{} & \cellcolor{white}{} & RFM-SCAN & 59.56 & 94.78 & 35.96 & 283 & 482 & 0 & 0.075 \\
			\cellcolor{white}{} & \cellcolor{white}{} & AdaLAM & 91.45 & 78.08 & 32.41 & 266 & 273 & 5 & 0.090 \\
			\cellcolor{white}{} & \cellcolor{white}{} & OANet & 85.96 & 81.80 & 30.78 & 248 & 267 & 2 & 0.043 \\
			\cellcolor{white}{} & \cellcolor{white}{} & ACNe & 77.48 & 84.92 & 33.13 & 266 & 313 & 0 & 0.023 \\
			\cellcolor{white}{} & \cellcolor{white}{} & PFM & 75.38 & 35.38 & 14.92 & 122 & 127 & 15 & 0.076 \\
			\cellcolor{white}{} & \cellcolor{white}{} & PGM & 78.24 & 84.82 & 32.65 & 259 & 283 & 7 & 2.759 \\
			\cellcolor{white}{} & \cellcolor{white}{} & SCV & 84.15 & 47.13 & 19.92 & 198 & 209 & 6 & 0.445 \\
			\cellcolor{white}{} & \multirow{-18}{*}{\rotatebox[origin=c]{90}{\cellcolor{white}{no model}}} & BM & 83.91 & 57.48 & 24.36 & 238 & 245 & 11 & 0.532 \\
			\cmidrule(l{2pt}r{2pt}){2-10}
			\cellcolor{white}{} & \cellcolor{white}{} & blob & 81.07 & 75.11 & 31.20 & 263 & 270 & 12 & -- \\
			\cellcolor{white}{} & \cellcolor{white}{} & th & 82.39 & 52.84 & 23.18 & 200 & 205 & 12 & -- \\
			\cellcolor{white}{} & \cellcolor{white}{} & DTM$_1$ & 84.62 & 55.10 & 22.89 & 212 & 218 & 9 & -- \\
			\cellcolor{white}{} & \cellcolor{white}{} & DTM$_2$ & 85.61 & 67.27 & 27.75 & 246 & 252 & 8 & -- \\
			\cellcolor{white}{} & \cellcolor{white}{} & LMR & 85.78 & 72.99 & 29.78 & 259 & 265 & 9 & -- \\
			\cellcolor{white}{} & \cellcolor{white}{} & LPM & 82.44 & 68.23 & 27.84 & 252 & 259 & 12 & -- \\
			\cellcolor{white}{} & \cellcolor{white}{} & GLPM & 86.70 & 66.24 & 28.44 & 248 & 254 & 9 & -- \\
			\cellcolor{white}{} & \cellcolor{white}{} & GMS & 45.93 & 15.61 & 7.28 & 91 & 93 & 40 & -- \\
			\cellcolor{white}{} & \cellcolor{white}{} & VFC & 91.47 & 84.42 & 32.93 & 268 & 274 & 5 & -- \\
			\cellcolor{white}{} & \cellcolor{white}{} & LLT & 48.31 & 38.39 & 16.29 & 183 & 188 & 37 & -- \\
			\cellcolor{white}{} & \cellcolor{white}{} & RFM-SCAN & 88.36 & 79.95 & 32.29 & 266 & 272 & 7 & -- \\
			\cellcolor{white}{} & \cellcolor{white}{} & AdaLAM & 91.83 & 75.75 & 31.47 & 256 & 262 & 5 & -- \\
			\cellcolor{white}{} & \cellcolor{white}{} & OANet & 90.57 & 75.76 & 29.42 & 238 & 243 & 5 & -- \\
			\cellcolor{white}{} & \cellcolor{white}{} & ACNe & 90.40 & 76.44 & 31.08 & 252 & 257 & 6 & -- \\
			\cellcolor{white}{} & \cellcolor{white}{} & PFM & 74.92 & 32.62 & 14.02 & 118 & 121 & 18 & -- \\
			\cellcolor{white}{} & \cellcolor{white}{} & PGM & 81.96 & 73.75 & 29.46 & 238 & 245 & 12 & -- \\
			\cellcolor{white}{} & \cellcolor{white}{} & SCV & 80.16 & 43.87 & 18.70 & 189 & 194 & 14 & -- \\
			\multirow{-36}{*}{\rotatebox[origin=c]{90}{\cellcolor{white}{planar}}} & \multirow{-18}{*}{\rotatebox[origin=c]{90}{\cellcolor{white}{1SAC}}} & BM & 78.92 & 55.83 & 23.72 & 230 & 236 & 15 & -- \\
			\midrule
			\cellcolor{white}{} & \cellcolor{white}{} & blob & 16.41 & 100.00 & 38.91 & 214 & 1370 & 0 & -- \\
			\cellcolor{white}{} & \cellcolor{white}{} & th & 75.17 & 53.09 & 22.95 & 130 & 155 & 2 & 0.000 \\
			\cellcolor{white}{} & \cellcolor{white}{} & DTM$_1$ & 78.93 & 64.57 & 27.04 & 155 & 179 & 2 & 0.166 \\
			\cellcolor{white}{} & \cellcolor{white}{} & DTM$_2$ & 78.08 & 75.28 & 31.30 & 178 & 209 & 2 & 0.241 \\
			\cellcolor{white}{} & \cellcolor{white}{} & LMR & 64.29 & 76.50 & 31.95 & 182 & 257 & 4 & 0.040 \\
			\cellcolor{white}{} & \cellcolor{white}{} & LPM & 67.36 & 74.63 & 31.33 & 180 & 240 & 2 & 0.008 \\
			\cellcolor{white}{} & \cellcolor{white}{} & GLPM & 82.37 & 60.41 & 27.19 & 161 & 180 & 4 & 0.010 \\
			\cellcolor{white}{} & \cellcolor{white}{} & GMS & 68.75 & 15.22 & 7.54 & 55 & 56 & 30 & 0.266 \\
			\cellcolor{white}{} & \cellcolor{white}{} & VFC & 66.94 & 78.26 & 32.25 & 189 & 243 & 7 & 0.018 \\
			\cellcolor{white}{} & \cellcolor{white}{} & LLT & 45.50 & 53.14 & 23.45 & 154 & 227 & 28 & 0.531 \\
			\cellcolor{white}{} & \cellcolor{white}{} & RFM-SCAN & 57.61 & 90.48 & 35.60 & 200 & 370 & 0 & 0.070 \\
			\cellcolor{white}{} & \cellcolor{white}{} & AdaLAM & 90.24 & 66.81 & 28.78 & 168 & 176 & 4 & 0.064 \\
			\cellcolor{white}{} & \cellcolor{white}{} & OANet & 81.66 & 77.16 & 30.30 & 167 & 192 & 1 & 0.043 \\
			\cellcolor{white}{} & \cellcolor{white}{} & ACNe & 67.16 & 75.56 & 30.78 & 175 & 242 & 1 & 0.023 \\
			\cellcolor{white}{} & \cellcolor{white}{} & PFM & 75.95 & 35.13 & 15.53 & 83 & 91 & 12 & 0.075 \\
			\cellcolor{white}{} & \cellcolor{white}{} & PGM & 75.23 & 88.81 & 36.34 & 204 & 256 & 2 & 3.732 \\
			\cellcolor{white}{} & \cellcolor{white}{} & SCV & 81.93 & 55.10 & 23.72 & 136 & 156 & 2 & 0.455 \\
			\cellcolor{white}{} & \multirow{-18}{*}{\rotatebox[origin=c]{90}{\cellcolor{white}{no model}}} & BM & 85.25 & 61.61 & 26.68 & 163 & 177 & 6 & 0.424 \\
			\cmidrule(l{2pt}r{2pt}){2-10}
			\cellcolor{white}{} & \cellcolor{white}{} & blob & 62.87 & 80.34 & 33.83 & 194 & 269 & 2 & -- \\
			\cellcolor{white}{} & \cellcolor{white}{} & th & 85.52 & 47.78 & 21.33 & 124 & 132 & 4 & -- \\
			\cellcolor{white}{} & \cellcolor{white}{} & DTM$_1$ & 87.61 & 58.72 & 25.18 & 147 & 157 & 2 & -- \\
			\cellcolor{white}{} & \cellcolor{white}{} & DTM$_2$ & 87.23 & 69.04 & 29.26 & 169 & 182 & 2 & -- \\
			\cellcolor{white}{} & \cellcolor{white}{} & LMR & 79.46 & 70.05 & 29.97 & 174 & 191 & 5 & -- \\
			\cellcolor{white}{} & \cellcolor{white}{} & LPM & 82.10 & 70.23 & 29.89 & 174 & 192 & 3 & -- \\
			\cellcolor{white}{} & \cellcolor{white}{} & GLPM & 81.59 & 57.97 & 26.31 & 157 & 167 & 11 & -- \\
			\cellcolor{white}{} & \cellcolor{white}{} & GMS & 52.14 & 14.63 & 7.31 & 54 & 55 & 47 & -- \\
			\cellcolor{white}{} & \cellcolor{white}{} & VFC & 78.29 & 74.22 & 30.87 & 181 & 201 & 8 & -- \\
			\cellcolor{white}{} & \cellcolor{white}{} & LLT & 56.95 & 50.47 & 22.39 & 148 & 169 & 29 & -- \\
			\cellcolor{white}{} & \cellcolor{white}{} & RFM-SCAN & 76.32 & 79.06 & 32.86 & 189 & 225 & 1 & -- \\
			\cellcolor{white}{} & \cellcolor{white}{} & AdaLAM & 89.59 & 64.51 & 27.85 & 164 & 171 & 5 & -- \\
			\cellcolor{white}{} & \cellcolor{white}{} & OANet & 85.30 & 73.76 & 29.32 & 163 & 178 & 1 & -- \\
			\cellcolor{white}{} & \cellcolor{white}{} & ACNe & 77.22 & 72.62 & 29.86 & 170 & 202 & 1 & -- \\
			\cellcolor{white}{} & \cellcolor{white}{} & PFM & 79.97 & 32.42 & 14.64 & 79 & 84 & 13 & -- \\
			\cellcolor{white}{} & \cellcolor{white}{} & PGM & 83.54 & 81.13 & 33.78 & 191 & 214 & 3 & -- \\
			\cellcolor{white}{} & \cellcolor{white}{} & SCV & 86.84 & 51.50 & 22.49 & 130 & 139 & 3 & -- \\
			\multirow{-36}{*}{\rotatebox[origin=c]{90}{\cellcolor{white}{non-planar}}} & \multirow{-18}{*}{\rotatebox[origin=c]{90}{\cellcolor{white}{1SAC}}} & BM & 86.15 & 60.41 & 26.25 & 160 & 170 & 7 & -- \\
			\bottomrule
		\end{tabular}
	}
\end{table}

\begin{table}
	\rowcolors{3}{lightgray}{}
	\vspace{-1.2em}
	\caption{Local spatial filter statistics for the baseline configuration. The average planar/non-planar number of ground-truth matches per image pair are 632/572 (blob, for recall), 844/685 (all matches, for recall$^\star$)}\label{table_sosnet}
	\centering
	\resizebox{0.47\textwidth}{!}{
		\begin{tabular}{rrrrrrrrrr}
			& & & \multicolumn{1}{l}{\rot[37]{precision (\%)}} & \multicolumn{1}{l}{\rot[37]{recall (\%)}} & \multicolumn{1}{l}{\rot[37]{recall$^\star$ (\%)}} & \multicolumn{1}{l}{\rot[37]{\shortstack[l]{correct\vspace{-0.25em}\\matches}}} & \multicolumn{1}{l}{\rot[37]{\shortstack[l]{output\vspace{-0.35em}\\matches}}} & \multicolumn{1}{l}{\rot[37]{failures}} & \multicolumn{1}{l}{\rot[37]{time (s)}} \\
			\toprule
			\cellcolor{white}{} & \cellcolor{white}{} & blob & 27.18 & 100.00 & 74.59 & 632 & 6081 & 0 & -- \\
			\cellcolor{white}{} & \cellcolor{white}{} & th & 95.17 & 66.38 & 50.93 & 449 & 758 & 0 & 0.000 \\
			\cellcolor{white}{} & \cellcolor{white}{} & DTM$_1$ & 76.75 & 77.48 & 59.05 & 526 & 1510 & 0 & 0.477 \\
			\cellcolor{white}{} & \cellcolor{white}{} & DTM$_2$ & 77.45 & 83.94 & 63.57 & 561 & 1592 & 0 & 0.902 \\
			\cellcolor{white}{} & \cellcolor{white}{} & LMR & 82.77 & 65.04 & 49.39 & 448 & 929 & 0 & 0.096 \\
			\cellcolor{white}{} & \cellcolor{white}{} & LPM & 79.02 & 39.27 & 30.69 & 288 & 684 & 0 & 0.026 \\
			\cellcolor{white}{} & \cellcolor{white}{} & GLPM & 77.05 & 76.81 & 58.77 & 523 & 1518 & 0 & 0.062 \\
			\cellcolor{white}{} & \cellcolor{white}{} & GMS & 72.71 & 54.69 & 43.07 & 433 & 1329 & 9 & 0.246 \\
			\cellcolor{white}{} & \cellcolor{white}{} & VFC & 79.13 & 95.09 & 71.88 & 626 & 1877 & 1 & 0.035 \\
			\cellcolor{white}{} & \cellcolor{white}{} & LLT & 44.49 & 56.98 & 43.67 & 328 & 1354 & 26 & 3.583 \\
			\cellcolor{white}{} & \cellcolor{white}{} & RFM-SCAN & 60.18 & 92.96 & 69.89 & 610 & 2525 & 0 & 1.534 \\
			\cellcolor{white}{} & \cellcolor{white}{} & AdaLAM & 92.98 & 80.22 & 61.24 & 536 & 1175 & 1 & 0.429 \\
			\cellcolor{white}{} & \cellcolor{white}{} & OANet & 86.50 & 78.58 & 58.77 & 525 & 1173 & 0 & 0.119 \\
			\cellcolor{white}{} & \cellcolor{white}{} & ACNe & 79.21 & 79.33 & 60.47 & 524 & 1308 & 0 & 0.067 \\
			\cellcolor{white}{} & \cellcolor{white}{} & PFM & 93.90 & 36.65 & 28.39 & 236 & 293 & 3 & 0.165 \\
			\cellcolor{white}{} & \cellcolor{white}{} & PGM & 68.78 & 72.63 & 54.10 & 420 & 728 & 5 & 4.974 \\
			\cellcolor{white}{} & \cellcolor{white}{} & SCV & 21.14 & 45.54 & 35.79 & 353 & 2136 & 0 & 1.203 \\
			\cellcolor{white}{} & \multirow{-18}{*}{\rotatebox[origin=c]{90}{\cellcolor{white}{no model}}} & BM & 86.26 & 54.88 & 42.69 & 439 & 954 & 8 & 1.654 \\
			\cmidrule(l{2pt}r{2pt}){2-10}
			\cellcolor{white}{} & \cellcolor{white}{} & blob & 94.59 & 92.54 & 69.53 & 593 & 1527 & 1 & -- \\
			\cellcolor{white}{} & \cellcolor{white}{} & th & 96.75 & 63.42 & 48.66 & 431 & 715 & 1 & -- \\
			\cellcolor{white}{} & \cellcolor{white}{} & DTM$_1$ & 95.35 & 72.95 & 55.80 & 500 & 1211 & 1 & -- \\
			\cellcolor{white}{} & \cellcolor{white}{} & DTM$_2$ & 95.40 & 78.96 & 60.03 & 532 & 1278 & 1 & -- \\
			\cellcolor{white}{} & \cellcolor{white}{} & LMR & 94.13 & 60.04 & 45.79 & 417 & 793 & 2 & -- \\
			\cellcolor{white}{} & \cellcolor{white}{} & LPM & 91.31 & 36.40 & 28.60 & 270 & 577 & 4 & -- \\
			\cellcolor{white}{} & \cellcolor{white}{} & GLPM & 95.24 & 73.16 & 56.10 & 498 & 1218 & 1 & -- \\
			\cellcolor{white}{} & \cellcolor{white}{} & GMS & 82.35 & 51.73 & 40.84 & 410 & 1077 & 10 & -- \\
			\cellcolor{white}{} & \cellcolor{white}{} & VFC & 92.69 & 88.75 & 67.37 & 588 & 1510 & 2 & -- \\
			\cellcolor{white}{} & \cellcolor{white}{} & LLT & 62.13 & 53.63 & 41.14 & 307 & 825 & 26 & -- \\
			\cellcolor{white}{} & \cellcolor{white}{} & RFM-SCAN & 94.69 & 86.62 & 65.54 & 574 & 1485 & 1 & -- \\
			\cellcolor{white}{} & \cellcolor{white}{} & AdaLAM & 96.25 & 76.58 & 58.44 & 512 & 1085 & 1 & -- \\
			\cellcolor{white}{} & \cellcolor{white}{} & OANet & 94.94 & 74.08 & 55.64 & 500 & 1037 & 2 & -- \\
			\cellcolor{white}{} & \cellcolor{white}{} & ACNe & 92.35 & 74.60 & 57.12 & 499 & 1073 & 3 & -- \\
			\cellcolor{white}{} & \cellcolor{white}{} & PFM & 94.54 & 35.65 & 27.56 & 230 & 284 & 3 & -- \\
			\cellcolor{white}{} & \cellcolor{white}{} & PGM & 89.74 & 67.95 & 50.76 & 393 & 545 & 5 & -- \\
			\cellcolor{white}{} & \cellcolor{white}{} & SCV & 81.88 & 41.76 & 32.89 & 335 & 548 & 12 & -- \\
			\multirow{-36}{*}{\rotatebox[origin=c]{90}{\cellcolor{white}{planar}}} & \multirow{-18}{*}{\rotatebox[origin=c]{90}{\cellcolor{white}{1SAC}}} & BM & 86.46 & 52.75 & 41.04 & 420 & 881 & 9 & -- \\
			\midrule
			\cellcolor{white}{} & \cellcolor{white}{} & blob & 24.01 & 100.00 & 81.61 & 572 & 5933 & 0 & -- \\
			\cellcolor{white}{} & \cellcolor{white}{} & th & 90.37 & 42.26 & 36.15 & 262 & 424 & 0 & 0.000 \\
			\cellcolor{white}{} & \cellcolor{white}{} & DTM$_1$ & 70.25 & 71.38 & 59.59 & 432 & 1291 & 0 & 0.430 \\
			\cellcolor{white}{} & \cellcolor{white}{} & DTM$_2$ & 70.25 & 75.44 & 62.88 & 455 & 1348 & 0 & 0.856 \\
			\cellcolor{white}{} & \cellcolor{white}{} & LMR & 75.29 & 53.49 & 44.65 & 328 & 698 & 0 & 0.093 \\
			\cellcolor{white}{} & \cellcolor{white}{} & LPM & 74.09 & 29.90 & 25.29 & 180 & 451 & 1 & 0.024 \\
			\cellcolor{white}{} & \cellcolor{white}{} & GLPM & 69.64 & 55.79 & 47.42 & 348 & 1002 & 3 & 0.032 \\
			\cellcolor{white}{} & \cellcolor{white}{} & GMS & 76.84 & 57.30 & 48.49 & 379 & 1131 & 5 & 0.237 \\
			\cellcolor{white}{} & \cellcolor{white}{} & VFC & 57.76 & 86.87 & 72.55 & 535 & 2055 & 4 & 0.037 \\
			\cellcolor{white}{} & \cellcolor{white}{} & LLT & 40.50 & 57.53 & 49.15 & 376 & 1646 & 27 & 3.243 \\
			\cellcolor{white}{} & \cellcolor{white}{} & RFM-SCAN & 52.30 & 95.06 & 77.99 & 557 & 2508 & 0 & 1.305 \\
			\cellcolor{white}{} & \cellcolor{white}{} & AdaLAM & 92.15 & 54.95 & 46.67 & 353 & 691 & 2 & 0.295 \\
			\cellcolor{white}{} & \cellcolor{white}{} & OANet & 83.54 & 71.94 & 59.18 & 423 & 949 & 0 & 0.114 \\
			\cellcolor{white}{} & \cellcolor{white}{} & ACNe & 69.94 & 67.33 & 55.65 & 396 & 1047 & 0 & 0.063 \\
			\cellcolor{white}{} & \cellcolor{white}{} & PFM & 91.80 & 24.73 & 21.15 & 145 & 177 & 2 & 0.131 \\
			\cellcolor{white}{} & \cellcolor{white}{} & PGM & 67.76 & 72.30 & 59.92 & 417 & 829 & 1 & 7.961 \\
			\cellcolor{white}{} & \cellcolor{white}{} & SCV & 22.66 & 40.03 & 34.16 & 248 & 1530 & 1 & 1.024 \\
			\cellcolor{white}{} & \multirow{-18}{*}{\rotatebox[origin=c]{90}{\cellcolor{white}{no model}}} & BM & 88.58 & 51.39 & 44.14 & 349 & 737 & 4 & 1.261 \\
			\cmidrule(l{2pt}r{2pt}){2-10}
			\cellcolor{white}{} & \cellcolor{white}{} & blob & 66.97 & 82.58 & 68.72 & 497 & 1624 & 0 & -- \\
			\cellcolor{white}{} & \cellcolor{white}{} & th & 93.05 & 38.48 & 32.99 & 240 & 379 & 1 & -- \\
			\cellcolor{white}{} & \cellcolor{white}{} & DTM$_1$ & 85.71 & 62.32 & 52.51 & 387 & 976 & 1 & -- \\
			\cellcolor{white}{} & \cellcolor{white}{} & DTM$_2$ & 85.40 & 65.59 & 55.22 & 407 & 1018 & 0 & -- \\
			\cellcolor{white}{} & \cellcolor{white}{} & LMR & 89.08 & 48.10 & 40.48 & 301 & 560 & 0 & -- \\
			\cellcolor{white}{} & \cellcolor{white}{} & LPM & 85.54 & 27.57 & 23.47 & 169 & 375 & 3 & -- \\
			\cellcolor{white}{} & \cellcolor{white}{} & GLPM & 84.01 & 50.91 & 43.42 & 318 & 782 & 3 & -- \\
			\cellcolor{white}{} & \cellcolor{white}{} & GMS & 83.66 & 51.43 & 43.67 & 340 & 917 & 5 & -- \\
			\cellcolor{white}{} & \cellcolor{white}{} & VFC & 74.84 & 75.19 & 63.26 & 472 & 1351 & 4 & -- \\
			\cellcolor{white}{} & \cellcolor{white}{} & LLT & 57.15 & 52.21 & 44.77 & 344 & 1002 & 27 & -- \\
			\cellcolor{white}{} & \cellcolor{white}{} & RFM-SCAN & 75.16 & 80.64 & 67.08 & 489 & 1466 & 0 & -- \\
			\cellcolor{white}{} & \cellcolor{white}{} & AdaLAM & 92.85 & 50.94 & 43.32 & 324 & 628 & 2 & -- \\
			\cellcolor{white}{} & \cellcolor{white}{} & OANet & 86.86 & 63.55 & 52.70 & 381 & 811 & 1 & -- \\
			\cellcolor{white}{} & \cellcolor{white}{} & ACNe & 79.01 & 60.91 & 50.67 & 365 & 850 & 0 & -- \\
			\cellcolor{white}{} & \cellcolor{white}{} & PFM & 92.89 & 23.17 & 19.84 & 137 & 165 & 2 & -- \\
			\cellcolor{white}{} & \cellcolor{white}{} & PGM & 80.57 & 65.48 & 54.63 & 388 & 646 & 1 & -- \\
			\cellcolor{white}{} & \cellcolor{white}{} & SCV & 67.69 & 35.99 & 30.89 & 225 & 473 & 2 & -- \\
			\multirow{-36}{*}{\rotatebox[origin=c]{90}{\cellcolor{white}{non-planar}}} & \multirow{-18}{*}{\rotatebox[origin=c]{90}{\cellcolor{white}{1SAC}}} & BM & 87.80 & 47.29 & 40.65 & 319 & 659 & 6 & -- \\
			\bottomrule
		\end{tabular}
	}
\end{table}
\cleardoublepage

\cleardoublepage

\end{document}